\documentclass[twoside,11pt]{article}

\usepackage{blindtext}

\usepackage[preprint,abbrvbib]{jmlr2e}

\usepackage[utf8]{inputenc} 
\usepackage[T1]{fontenc}    
\usepackage{url}            
\usepackage{booktabs}       
\usepackage{amsfonts}       
\usepackage{nicefrac}       
\usepackage{microtype}      
\usepackage{xcolor}         
\hypersetup{breaklinks=true}
\usepackage{placeins}

\usepackage{amsmath}
\usepackage{xspace}
\usepackage{relsize}
\usepackage{graphicx}
\usepackage{mathtools}
\usepackage{cancel}
\usepackage{bm}
\usepackage{dsfont}
\usepackage{graphicx} 
\usepackage{subcaption}
\usepackage{tikz}
\usetikzlibrary{bayesnet,calc,decorations.pathreplacing}
\usepackage{lastpage}
\definecolor{darkpink}{rgb}{0.91, 0.33, 0.5}
\definecolor{darkgreen}{rgb}{0.0, 0.5, 0.0}

\newcommand{\fjadd}[1]{}

\usepackage[draft]{changes}
\definechangesauthor[name={Javier Sáez}, color=red]{fjsm}

\definechangesauthor[name={Juan Maroñas}, color=blue]{jmm}

\definechangesauthor[name={New}, color=orange]{new}

\definechangesauthor[name={Daniel Hernández}, color=darkgreen]{dhl}

\newcommand{\acro}[1]{\textsc{#1}\xspace}

\newcommand{\DGP}{\acro{\smaller DGP}}

\newcommand{\DSVI}{\acro{\smaller DSVI}}
\newcommand{\ELBO}{\acro{\smaller ELBO}}
\newcommand{\ELL}{\acro{\smaller ELL}}
\newcommand{\GP}{\acro{\smaller GP}}
\newcommand{\GPFLOW}{\acro{\smaller GPFLOW}}
\newcommand{\GPYTORCH}{\acro{\smaller GPYTORCH}}
\newcommand{\GPFLUX}{\acro{\smaller GPFLUX}}
\newcommand{\GPJAX}{\acro{\smaller GPJAX}}
\newcommand{\KLD}{\acro{\smaller KLD}}
\newcommand{\LL}{\acro{\smaller LL}}
\newcommand{\MCMC}{\acro{\smaller MCMC}}
\newcommand{\RBF}{\acro{\smaller RBF}}
\newcommand{\RMSE}{\acro{\smaller RMSE}}
\newcommand{\SVGP}{\acro{\smaller SVGP}}
\newcommand{\UCI}{\acro{\smaller UCI}}
\newcommand{\VI}{\acro{\smaller VI}}
\newcommand{\ZERO}{\acro{\smaller ZERO}}

\newcommand{\MO}{\acro{\smaller M0}}
\newcommand{\MY}{\acro{\smaller MY}}
\newcommand{\PCA}{\acro{\smaller PCA}}
\newcommand{\PCANWR}{\acro{\smaller PCA-NWR}}
\newcommand{\PCANW}{\acro{\smaller PCA-NW}}
\newcommand{\PCAW}{\acro{\smaller PCA-W}}
\newcommand{\W}{\acro{\smaller W}}
\newcommand{\NWR}{\acro{\smaller NWR}}
\newcommand{\NW}{\acro{\smaller NW}}
\newcommand{\ZERONWR}{\acro{\smaller ZERO-NWR}}
\newcommand{\ZERONW}{\acro{\smaller ZERO-NW}}
\newcommand{\ZEROW}{\acro{\smaller ZERO-W}}
\newcommand{\ZEROMO}{\acro{\smaller ZERO-Points-M0}}
\newcommand{\ZEROMY}{\acro{\smaller ZERO-Points-MY}}
\newcommand{\ZERONWRMO}{\acro{\smaller ZERO-Points-M0-NWR}}
\newcommand{\ZEROWMO}{\acro{\smaller ZERO-Points-M0-W}}
\newcommand{\ZERONWRMY}{\acro{\smaller ZERO-Points-MY-NWR}}
\newcommand{\ZEROWMY}{\acro{\smaller ZERO-Points-MY-W }}

\newcommand{\eg}{\emph{e.g.}\xspace}
\newcommand{\fig}{Fig.\xspace}
\newcommand{\ie}{\emph{i.e.}\xspace}
\newcommand{\usec}{Sec.\xspace}
\newcommand{\ueqn}{Eq.\xspace}
\newcommand{\ueqns}{Eqs.\xspace}
\newcommand{\utab}{Table\xspace}
\newcommand{\ualg}{Alg.\xspace}

\newcommand{\wrt}{w.r.t.\xspace}

\newcommand{\mb}[1]{\mathbf{#1}}

\newcommand{\bra}[1]{\left[#1\right]}
\newcommand{\braInv}[1]{\left[#1\right]^{-1}}
\newcommand{\braT}[1]{\left[#1\right]^T}
\newcommand{\pare}[1]{\left(#1\right)}
\newcommand{\pareT}[1]{\left(#1\right)^T}
\newcommand{\tr}[1]{\text{Tr}\left[#1\right]}

\newcommand{\R}{\mathbb R}

\newcommand{\Xspace}{\mathcal{X}}
\newcommand{\Yspace}{\mathcal{Y}}

\newcommand{\Expectation}[2]{\mathbb{E}_{#1}\left[#2\right]}
\newcommand{\myprod}[2]{\prod_{#1}^{#2}} 
\newcommand{\mysum}[2]{\sum_{#1}^{#2}} 
\newcommand{\argmin}{\text{argmin}\xspace} 
\newcommand{\norm}[1]{\lVert {#1}\rVert}
\newcommand{\diag}{\operatorname{diag}}

\newcommand{\matD}{\mathbf{D}}
\newcommand{\matW}{\mathbf{W}}
\newcommand{\matI}{\mathbf{I}}

\newcommand{\veczero}{\mb{0}}
\newcommand{\vecw}{\mb{w}}
\newcommand{\vecwT}{\mb{w}^T}

\newcommand{\vecsymbol}{\text{vec}}

\newcommand{\vvec}[1]{\vecsymbol\,#1}
\newcommand{\vvecB}[1]{\vecsymbol \left[#1\right]}

\newcommand{\vvecT}[1]{\left[ \vvec{#1}\right]^T}

\newcommand{\Ngauss}[1]{\mathcal{N}\left(#1\right)}

\newcommand{\dd}[1]{\text{d}#1}

\newcommand{\K}[2]{K_{{#1}{#2}}}
\newcommand{\Kinv}[2]{K_{{#1}{#2}}^{-1}}
\newcommand{\Lchol}[2]{L_{{#1}{#2}}}
\newcommand{\Linv}[2]{L_{{#1}{#2}}^{-1}}

\newcommand{\nupos}[2]{\boldsymbol{\nu}_{#1}^{#2}}
\newcommand{\nuallpos}[1]{\overline{\boldsymbol{\nu}}_{#1}}

\newcommand{\Kzz}[1]{K_{\Zsamples{#1}\Zsamples{#1}}}
\newcommand{\Kzzinv}[1]{K_{\Zsamples{#1}\Zsamples{#1}}^{-1}}
\newcommand{\Kxz}[1]{K_{\Xsamples\Zsamples{#1}}}
\newcommand{\Kxsz}[1]{K_{\Xsel\Zsamples{#1}}}
\newcommand{\Kzx}[1]{K_{\Zsamples{#1}\Xsamples}}
\newcommand{\Kxx}{K_{\Xsamples\Xsamples}}

\newcommand{\Lzz}[1]{L_{\Zsamples{#1}\Zsamples{#1}}}
\newcommand{\Lzzinv}[1]{L_{\Zsamples{#1}\Zsamples{#1}}^{-1}}
\newcommand{\LzzT}[1]{L_{\Zsamples{#1}\Zsamples{#1}}^T}

\newcommand{\muX}{\mu_\Xsamples}
\newcommand{\muZ}[1]{\mu_{\Zsamples{#1}}}

\newcommand{\f}[1]{\mb{f}_{#1}\xspace} 
\newcommand{\fpos}[2]{\mb{f}^{#2}_{#1}\xspace} 
\newcommand{\fallpos}[2]{\mb{\overline{f}}^{#2}_{#1}\xspace}

\newcommand{\X}[2]{\mb{x}^{#2}_{#1}\xspace} 
\newcommand{\Xsamples}{\mb{X}\xspace} 
\newcommand{\Xsel}{\mb{X_{s}}\xspace} 
\newcommand{\Xx}{\mb{X_{x}}\xspace} 
\newcommand{\Xz}{\mb{X_{z}}\xspace} 
\newcommand{\Xxz}{\mb{X_{xz}}\xspace} 

\newcommand{\Y}[2]{\mb{y}^{#2}_{#1}\xspace} 
\newcommand{\Ysamples}{\mb{Y}\xspace}

\newcommand{\varm}[2]{\mathbf{m}_{#1}^{#2}}
\newcommand{\varS}[2]{\mathbf{S}_{#1}^{#2}}
\newcommand{\varSinv}[2]{{\mathbf{S}_{#1}^{#2}}^{-1}}

\newcommand{\varmr}[1]{\mathbf{m}^{#1}_{\mb{r}}}

\newcommand{\varmw}[1]{\mathbf{m}^{#1}_{\uwhit}}
\newcommand{\varSw}[1]{\mathbf{S}^{#1}_{\uwhit}}

\newcommand{\varSwinv}[1]{\mathbf{S}_{\uwhit}^{{#1}^{-1}}}

\newcommand{\uu}[1]{\mb{u}^{#1}\xspace} 
\newcommand{\upos}[2]{\mb{u}^{#2}_{#1}\xspace}
\newcommand{\uallpos}[2]{\mb{\overline{u}}^{#2}_{#1}\xspace}

\newcommand{\uwhit}{\mb{v}\xspace} 

\newcommand{\Z}[1]{\mb{z}_{#1}\xspace} 
\newcommand{\Zsamples}[1]{\mb{Z}^{#1}\xspace} 
\newcommand{\Zsamplespos}[2]{\mb{Z}_{#1}^{#2}\xspace}

\newcommand{\qfm}[2]{\mathbf{\mu}^{#2}_{qf^{#1}}}
\newcommand{\qfmall}[2]{\mathbf{\overline{\mu}}^{#2}_{qf^{#1}}}
\newcommand{\qfS}[2]{K^{#2}_{qf^{#1}}}
\newcommand{\qfSall}[2]{\overline{K}^{#2}_{qf^{#1}}}

\newcommand{\qfmr}[2]{\mathbf{\mu}^{#2}_{qf_r^{#1}}}

\newcommand{\qfmw}[2]{\mathbf{\mu}^{#2}_{qf_w^{#1}}}
\newcommand{\qfSw}[2]{K^{#2}_{qf_w^{#1}}}

\newcommand{\cent}[1]{\mathbf{c}_{#1}}

\usepackage{comment}
\usepackage{float}
\usepackage{multirow} 
\usepackage{algorithm}
\usepackage{algpseudocode}
\usepackage{hyperref}
\usepackage{lastpage}

\jmlrheading{23}{2026}{1-\pageref{LastPage}}{1/21; Revised 5/22}{9/22}{21-0000}{Francisco Javier Sáez-Maldonado, Juan Maroñas and Daniel Hernández-Lobato}

\ShortHeadings{An Analysis of Posterior Collapse, Parameterization and Initialization in Variational \DGP{}s
}{An Analysis of Posterior Collapse, Parameterization and Initialization in Variational \DGP{}s
}
\firstpageno{1}

\begin{document}

\title{An Analysis of Posterior Collapse, Parameterization and Initialization in Variational Deep Gaussian Processes}

\author{\name Francisco Javier Sáez-Maldonado\thanks{Equal Contribution. Juan Maroñas is also with the Machine Learning Group at Universidad Autónoma de Madrid.} \email fjaviersaezm@ugr.es \\
       \addr Department of Computer Science and Artificial Intelligence\\
       Universidad de Granada\\
       Granada, Spain
       \AND
       \name Juan Maroñas$^*$ \email juan.maronas@cunef.edu \\
       \addr Department of Quantitative Methods\\
       CUNEF Universidad\\
       Madrid, Spain
       \AND
       \name Daniel Hernández-Lobato \email daniel.hernandez@uam.es \\
       \addr Department of Computer Science and\\ Centro de Investigación Avanzada en Física Fundamental\\
       Universidad Autónoma de Madrid\\
       Madrid, Spain.
}

\editor{My editor}

\maketitle

\begin{abstract}
Deep Gaussian Processes (\DGP{}s) are probabilistic models with remarkable prediction 
performance that concatenate Gaussian Processes (\GP{}s) across several layers. 
Exact inference in \DGP{}s is, however, intractable, and 
variational inference (\VI) is often used to approximate the posterior with a parametric
distribution tuned by minimizing the Kullback-Leibler divergence. Moreover, finding a good \VI 
approximation is challenging due to the model's strong nonlinearities and multiple posterior modes.
In particular, a problem of \VI is posterior collapse, where \VI converges to a variational posterior that matches the prior. In variational \DGP{}s, this implies explaining the data as noise. In 
this work, we study posterior collapse in \DGP{}s and identify its 
connection to the Double Stochastic \VI algorithm and the widely 
used linear prior mean function employed in all but the last layer of a \DGP{}. 
We show that the benefit of the linear prior mean does not arise from avoiding the non-injective pathology 
in very deep \DGP{}s, as previously believed, but from improving the conditioning of the optimization 
problem at initialization. Thus, we propose an alternative initialization of a zero prior mean
\DGP that mimics a \DGP with a linear prior mean at initialization. This enables successful training of 
\DGP{}s without imposing optimization-driven constraints on the prior, allowing to choose the prior based on modeling assumptions rather than optimization convenience. Our analysis considers 
three common parameterizations of \DGP{}s and shows that not all of them benefit from a linear prior mean. 
We also explain why a whitened parameterization of the \DGP provides more stable convergence, something often 
assumed from experience, but lacking a rigorous analysis. Furthermore, we show that
this stability is also beneficial to avoid the posterior collapse problem.
Extensive experiments on synthetic and \UCI datasets validate our findings. Namely, the proposed initialization prevents posterior collapse, improves stability, and achieves 
performance comparable to (and sometimes better than) \DGP{}s with a linear prior mean. 
\end{abstract}


\tableofcontents

\section{Introduction}\label{sec:intro}
\looseness=-1 Gaussian Processes (\GP{}s) \citep{gps_for_ml_rasmussen} are flexible non-parametric models that have shown promising results across many applications, such as molecule optimization \citep{gomez2018automatic,tripp2024tanimoto} or uncertainty estimation in Neural Networks \citep{ortegavariational, JMLR:v24:22-0479}. They are used as prior distributions over some target function $f$, \ie \(f \sim \mathcal{GP}_{\nu}(\mu(\cdot), K(\cdot, \cdot))\), where \(\mu(\cdot)\) and  \(K(\cdot, \cdot)\) are, respectively, the mean and covariance functions of the \GP, parameterized by the set of parameters $\nu$.  Deep Gaussian Processes (\DGP{}s) \citep{pmlr-v31-damianou13a} extend \GP{}s by concatenating multiple \GP{}s in a layered network, thereby defining a hierarchical structure that leads to a more flexible probabilistic model. This concatenation enlarges the class of functions that can be modeled, allowing to capture complex patterns within the data. \fig{} \ref{fig:dgp_prior} (a) shows a graphical illustration of a \DGP. Typically, the inner layers have as many units as there are input dimensions \citep{Salimbeni2017}.

In spite of the aforementioned advantages of \DGP{}s, these models have the drawback that exact Bayesian inference is intractable. Thus, the community has been exploring different ways of computing posterior approximations, most of them relying on sparse GP representations based on inducing points and variational methods \citep{pmlr-v31-damianou13a,Salimbeni2017,IPVI_DGP,convolutionalDGP}, on Expectation Propagation \citep{EPDGP}, on Markov Chain Monte Carlo (\MCMC) \citep{havasi2018inference}, and, recently, on Denoising Diffusion Variational Inference \citep{DDVI2024xu}. One of the most popular variational inference approximations is the Doubly Stochastic Variational Inference (\DSVI) algorithm \citep{Salimbeni2017}, due to its implementation simplicity, performance, scalability, and its widespread adoption into various \DGP{} inference frameworks and alternative formulations of \DGP{} priors \citep{havasi2018inference,jankowiak2019neural,MultiResDGP,dtgp,DGPIW,rudner2020inter}.

Two design choices have gained popularity over the last years when training \DGP{} models using \DSVI.
The first one involves using a whitened representation of the process values at the inducing points \citep{hensman2015mcmc}. The second one consists of
using a non-learnable linear-mean function in the inner layers of a \DGP, firstly proposed by \cite{Salimbeni2017}.

The motivation for the first design choice is that it is well known that parameterizations that remove correlations enhance mixing when fitting probabilistic models using \MCMC \citep{papaspiliopoulos2007general, HMCHierachicalModels}. The use of this parameterization is nowadays the \textit{de facto} way of fitting these models when using Variational Inference  (\VI), as implemented by many popular \GP{}/\DGP{} libraries \citep{GPflow2017, gardner2018gpytorch, dutordoir2021gpflux}. Beyond libraries, the community has also adopted the whitened representation in different research directions involving \GP{}s or \DGP{}s \citep{DGPIW,tgp,etgp}. However, to the best of our knowledge, the reason why this parameterization is beneficial for \VI algorithms has not been adequately studied. In fact, these modern libraries do not explicitly argue why using a whitened representation is a better choice, beyond being recommended in its source code as seen, \eg, in the library documentation by \cite{gardner2018gpytorch}\footnote{\url{https://github.com/cornellius-gp/gpytorch/blob/44993efcc180bdbdeaaf2107c7cc1ba532b2da9b/gpytorch/variational/unwhitened_variational_strategy.py\#L30}}. Some works also outline that whitening can facilitate optimization \citep{tutorialSparseGPs}, but provide no theoretical nor empirical evidence for it.

The second design choice was motivated to avoid the non-injective pathologies of very deep \DGP{}s with a \ZERO prior mean function \citep{pathologies_vdn}. To alleviate these pathologies, \cite{Salimbeni2017} introduced a prior mean function that is simply equal to the identity function, or a \PCA projection of the inputs, when the number of units in each hidden layer is smaller than the number of input dimensions. 
For these reasons, it is also known as the \PCA prior mean function in the literature. Notwithstanding, such a mean function is used even in $2$-layer \DGP{}s, which are far from being close to the models studied by \cite{pathologies_vdn}, with up to $11$ layers. Moreover, \cite{pmlr-v31-damianou13a} successfully trained $5$ layers \ZERO prior mean \DGP{}s\footnote{Neil Lawrence's blog also showed successful optimization of $5$-layers zero mean \DGP{}s \url{https://inverseprobability.com/talks/notes/deep-gaussian-processes.html}.} which suggests that the original motivation to change the prior mean function of the inner layers is not completely justified. Similar to the whitening reparameterization, the \PCA prior mean function has become the default choice in the inner layers of a \DGP{} in modern software \citep{dutordoir2021gpflux,gardner2018gpytorch} and research \citep{DGPIW,rudner2020inter,tgp,etgp,dtgp}, with all attributing its success to the non-injective pathological problem of \DGP{}s \citep{rudner2020inter,DGPIW}.

In this work, we found that the success of the \PCA prior mean function is linked to optimization issues that arise from the selection of the initial variational parameters within the \DSVI inference framework. In particular, issues that arise when initializing the variational mean to $\varm{}{}=\veczero$, a common choice in the literature \citep{Salimbeni2017, DGPIW,GPflow2017, dutordoir2021gpflux, gardner2018gpytorch}. Specifically, we observed that the \DSVI algorithm applied to a \ZERO prior mean \DGP{} with this initialization can result in a bad local optimum. The variational learning objective collapses the approximate posterior to the prior, and the model learns to explain the data as noise. This is a well-known effect studied in Variational Autoencoders equipped with an expressive decoder, known as \emph{posterior collapse} \citep{pos_collapse_1,chen2017variationallossyautoencoder,pos_collapse_2}. However, the sources of the collapse problem differ from those we found in the context of \DGP{}s. One may find this behavior unsurprising since previous works have shown that sparse \GP{}s fitted via \VI tend to learn observation uncertainty through the noise parameter \citep{sparseGPVarAnalysis}. Moreover, some works suggest training sparse \GP{}s by first freezing model hyper-parameters \citep{tutorialSparseGPs,GPs-Big-Data}, or just the noise parameter \citep{MultiResDGP}, before fitting everything end-to-end, including variational parameters. Here, we show that the posterior collapse problem in \DGP{}s is alleviated by using the \PCA prior mean function in the inner layers of the \DGP model.

Since the architectural choice of the prior should be dictated by modeling considerations, and not by limitations due to an optimization issue \citep{GVI_jer}, we propose a straightforward initialization of the variational parameters that allows successful training of \DGP{}s using the \DSVI inference algorithm with independence of the specified prior mean function. For this, we propose to initialize the parameters of the model so that the output of each \GP within the inner layers of the \DGP mimics the output of a linear-mean prior \GP at initialization.  This can be efficiently achieved by initializing the variational means so that the inner layer predictions \emph{just at the inducing locations} are similar to the output of the \PCA prior mean function, \ie, the identity function, when the number of hidden units is equal to the number of input dimensions, or a \PCA projection of the inputs, when this number is smaller than the input dimensions. Furthermore, we extend this idea to initialize the last layer of the \DGP{} so that the initial predictive mean of the \DGP explains the observed targets well, which is expected to facilitate fitting the \DGP model.

Our analysis considers both a whitened and a non-whitened parameterization of a \DGP. We show that posterior collapse can occur under both parameterizations. Thus, the parameterization itself is not the root cause of collapse. However, the whitened parameterization consistently provides much more stable optimization, because whitening removes prior‑induced correlations and yields better‑conditioned gradients. By contrast, the non‑whitened parameterization is significantly more unstable, exhibiting noisy updates, large fluctuations in the KL divergence, and a higher chance of converging to poor local optima.

Regarding the non-whitened reparameterization, we also investigate the impact of a reparameterization of the variational mean (\ie, the variational mean is simply the variational mean minus the prior mean). This reparameterization is used, in libraries such as \GPFLOW, and is originally motivated in the context of inter-domain \GP{}s \citep{tutorialSparseGPs}. Our initial analysis shows that such a non-whitened reparameterization benefits further from the \PCA prior mean function than the standard non-whitened parameterization implemented, \eg, in \GPYTORCH or \GPJAX. However, both non-whitened parameterizations suffer from optimization instabilities, even if the \PCA prior mean function is used, which may result in poorly fitted solutions that suffer from the posterior collapse problem. An initial study of optimization instabilities, in terms of Jacobians and Hessians of the Kullback-Leibler divergence regularizer of \VI, is carried out in a separate technical report \citep{maronas2025jacobianhessiankullbackleiblerdivergence}.

The experimental evaluation carried out includes a series of studies on both a controlled synthetic dataset and eight real‑world \UCI benchmarks, designed to validate the theoretical analysis of posterior collapse and the proposed initialization strategy. These experiments clearly show that \ZERO prior mean \DGP{}s often collapse under standard initialization, while \PCA prior mean \DGP{}s avoid collapse only because of their favorable initialization behavior. Furthermore,  the results obtained show that the proposed initialization consistently stabilizes optimization, avoids collapse in cases where standard \DGP{}s fail, and achieves predictive performance comparable to (and sometimes surpassing) \PCA prior mean \DGP{}s models, confirming its effectiveness in practical scenarios.

Summing up, this paper extends our unpublished preliminary results \citep{workshop-nips-mode-collapse} through the following contributions:
\begin{itemize}
    \item We provide a comprehensive analysis of \emph{posterior collapse} in \DGP{}s trained with \DSVI, showing that collapse arises from the interaction between 
	standard variational parameter initialization and the inner-layer mean function, rather than from the non-injective pathologies traditionally cited in the literature.

    \item We reinterpret the role of the widespread \PCA prior mean function, demonstrating that its effectiveness stems from how it conditions the optimization 
	    landscape at initialization, rather than from improved modeling properties.

    \item We propose a simple and computationally efficient initialization scheme for \ZERO prior mean \DGP{}s that mimics the beneficial initialization behavior 
	    of \PCA prior mean \DGP{}s. This enables successful training of \DGP{}s under principled \ZERO mean priors, avoiding optimization-driven architectural constraints.

    \item We analyze three commonly used parameterizations of \DGP{}s. Namely, the standard non-whitened, the whitened, and the \GPFLOW reparameterized non-whitened forms, 
	    and show that they exhibit fundamentally different optimization behaviors, with only some benefiting from \PCA prior mean functions.

    \item We provide the first detailed explanation of why the whitened parameterization yields more 
	    stable convergence, clarifying an assumption that has been widely accepted in practice but lacked rigorous justification.

    \item We identify conditions under which even whitened \DGP{}s with \PCA prior mean functions can suffer 
	    posterior collapse, particularly when kernel output scale or inducing-point configurations introduce excessive variance in the inner layers.

    \item Through extensive experiments on synthetic and \UCI benchmark datasets, we show that the proposed initialization prevents posterior collapse, 
	    stabilizes optimization, and achieves predictive performance comparable to or better than that of \DGP{}s with \PCA prior mean functions.

\end{itemize}

The rest of the paper is as follows: \usec \ref{sec:related_work} describes important related work from the literature that is relevant for the current investigation.
\usec \ref{sec:dgp_intro} describes the \DGP models, the role of prior mean functions, and the typical initializations and parameterizations that are used for these models.
\usec \ref{sec:optimization:difficulties}  analyzes the problem of posterior collapse in the context of \DGP and investigates when it is more likely to happen, for each prior mean function
considered and initialization. \usec \ref{sec:proposed_initialization} proposes novel initializations of a \ZERO prior mean \DGP to alleviate the posterior collapse problem.
\usec \ref{sec:experiments} includes synthetic and real-world experiments to validate our findings and the proposed initializations.
Finally, \usec \ref{sec:conclusions} gives the conclusions of the paper.
\section{Related Work}
\label{sec:related_work}

This section reviews important work from the literature that is relevant to the current investigation.

First, inference sub-optimality in variational methods has been widely studied and is associated with a poorly expressive approximate posterior distribution, which is only able to fit some modes from the true posterior \citep{bishop2007}. This effect is known as mode collapse. The problem arises from the way in which the Kullback-Leibler Divergence (\KLD) is minimized, which forces the variational distribution to avoid regions of low posterior density. Consequently, another issue derived from this minimization problem is that the variational posterior tends to underestimate the posterior variance.

In modern machine learning, the mode collapse problem has mainly been studied in the context of deep generative models based on variational inference (\VI), such as the variational autoencoder \citep{VAESeminal}. For a long time, the community has focused on creating expressive posterior approximations that can circumvent the limitations of factorized Gaussian distributions \citep{pmlr-v37-rezende15}, showing improved quality and diversity of generated samples. While not explicitly stated as a solution to \emph{mode collapse}, improving upon a unimodal Gaussian distribution directly targets this issue. That is, better approximations to the posterior necessarily lead either to unimodal non-Gaussian densities or to multimodal distributions. However, other research directions have also attributed this collapse problem to the decoder \citep{suboptVAE}. Beyond deep generative models, the mode collapse problem is still studied in probabilistic graphical models such as the Gaussian mixture model. For instance, \cite{theoreticalmodecollapseVI2025soletskyi} provides a theoretical explanation of this problem using the gradient flow associated with the optimization procedure.

On the other hand, in the deep generative models literature, one often finds the term mode collapse used to define the issue in which a generative model is only able to generate samples from a subset of the data distribution, a problem initially observed in generative adversarial networks \citep{GANs}. However, this term has also been adopted and studied in many different generative models, such as variational autoencoders \citep{modecollapsevaes2022}, or diffusion models \citep{modecollapsedifussionRL2024barcelo}, where the collapse problem is attributed to the presence of conflicting gradients during training.

Another issue associated with a collapse problem is what is known as posterior collapse. Again adopted by the deep generative community, this term refers to the approximate posterior collapsing to the prior in generative models trained via \VI. This effect is favored because that is precisely what minimizes the \KLD term in the objective. Some works attribute this issue to variational autoencoders equipped with expressive decoders  \citep{chen2017variationallossyautoencoder,pos_collapse_1,pos_collapse_2} showing that optimization results in a model working as a standard autoregressive model. This is because the decoder model is so expressive that it is able to map the noise distribution parameterized by the prior and transform it into a sample from the data distribution, which is the fitting mechanism used in many of such models \citep{PixelCNN,PixelRNN}. This results in a model that does not need to encode meaningful information in the inference network (the encoder) to reconstruct the data samples, which makes the objective function set the approximate posterior equal to the prior. Other works show that posterior collapse arises due to spurious local maxima in the training objective \citep{spuriousCollapse} or attribute it to a non-identifiability problem  \citep{nonIdenPosCollapse}. In our particular case, the issues we have found in the context of \DGP{}s can be associated with the problem studied by \cite{spuriousCollapse}, since under certain initialization, there are solutions in the training objective of the \DGP{} where the model learns everything as noise and collapses the variational distribution to the prior. 

Both the mode collapse and posterior collapse problems naturally appear in Bayesian neural networks fitted via \VI. In the modern era of machine learning, \cite{BNN_graves} and \cite{BNN_bundell} studied unimodal posterior approximations to the true posterior, which were then improved in follow-up works to yield more expressive approximate posteriors, directly targeting the mode collapse problem \citep{BNN_louizos1,BNN_louizos2}. On the other hand, \cite{wu2018deterministic} studied a hierarchical prior with parameters fitted via empirical Bayes. The authors state that by allowing too many degrees of freedom in the prior distribution, the empirical Bayes update would always lead to a \KLD of zero because the prior parameters would be set to the variational parameters. This results in a posterior collapse problem derived from an architectural design and optimization procedure. Our work studies posterior collapse in the context of \DGP{}s where the problem arises from a different optimization issue.

In the context of sparse Gaussian processes, the standard regression likelihood (homoscedastic Gaussian) results in a posterior distribution that is Gaussian. Thus, variational sparse \GP{}s were originally designed to scale computations \citep{pmlr-v5-titsias09a}. Here, the choice of a Gaussian variational distribution was justified, and no collapse problem is expected. This idea opened the possibility of using variational distributions to fit sparse \GP{}s with non-Gaussian likelihoods such as classification \citep{hensman2015mcmc} or heteroscedastic observation models \citep{variatonalHetereo}. Naturally, this approximation suffers from both the collapse and variance underestimation problems. In consequence, inference sub-optimality has also been addressed using 
\MCMC \citep{hensman2015mcmc} to accommodate non-Gaussian posteriors arising from the choice of the likelihood.

In \DGP{}s, the posterior distribution is non-Gaussian. The initial variational approach by \cite{pmlr-v31-damianou13a} considers a joint variational Gaussian distribution between all the latent functions in the hierarchy. Later, \cite{Salimbeni2017} proposed a variational joint posterior where the distributions are conditionally Gaussian, given a sample from the previous layer, since they observed that the original inference distribution in \cite{pmlr-v31-damianou13a} tends to underestimate the posterior variance. This is because the approximate posterior is not able to model correlations between the latent functions within layers. While the joint distribution over the latent functions suggested by \cite{Salimbeni2017} is non-Gaussian, and thus aims at directly tackling more complex posteriors, \cite{havasi2018inference} shows further that the posterior over just the inducing points is both non-Gaussian and multimodal. Therefore, those authors propose using \MCMC to draw samples from it. Following this observation \cite{IPVI_DGP,convolutionalDGP} draw inspiration from the deep generative community and propose improving the variational distribution to make it non-Gaussian and multimodal using normalizing flows. Furthermore, there are recent alternatives based on approximating the posterior distribution through denoising diffusion models\citep{DDVI2024xu,DDBridge}. Although none of these works explicitly mention the problem of mode collapse, they, similarly to deep generative models, target both the mode collapse and the variance underestimation problem that characterizes variational inference by considering more flexible posteriors. Notwithstanding, none of these works explicitly studied or reported inference suboptimality due to a posterior collapse, as we do.

In the context of \DGP{}s, there is also a line of research work studying variance collapse in the inner layers of the model. In particular, \cite{pmlr-v124-ustyuzhaninov20a} found that, in noise-free settings, \DGP{}s fitted via variational inference with factorized distributions result in the variational distributions of the inner layers collapsing to a deterministic function. In principle, with respect to any parameterization of the model. Later, \cite{popescu2022hierarchicalgaussianprocesseswasserstein2} performed a similar study and showed that when the number of inducing points increases, the variance collapses under a \DGP{} with a \ZERO prior mean function in the inner layer. They also show that the \PCA prior mean function solves this problem. While this is probably the closest line of work related to ours, we do not attribute the posterior collapse nor optimization issues found to the variance tending to zero, but to the actual signal seen by the objective function to be optimized at initialization. 

Regarding the parameterization of the process values at the inducing points, the use of a whitened representation of random variables goes beyond the field of optimization in the context of \DGP{}s. For instance, in latent Gaussian models, \cite{slicesampling2010murray} discussed the use of a whitened representation of the random variables to make those variables independent under the prior, simplifying posterior sampling using \MCMC. This is also used in single-layer sparse \GP{}s \citep{hensman2015mcmc}. In general, a reparameterization of probabilistic graphical models is often used to remove correlations in the posterior that enhance mixing \citep{papaspiliopoulos2007general, HMCHierachicalModels}. Moreover, in the context of \GP{}s, using a whitened representation of the process is commonly assumed. For instance, \cite{decoupledconditionalsvgps2023zhu} decouples the kernel hyper-parameters in the mean and covariance conditionals of the \SVGP{}. Then, these authors highlight that, empirically, they found it useful to use whitening in the mean kernel matrix. Also, \cite{sparseortogonalVI2020shi}  proposed to decompose the \SVGP{} as a sum of two independent processes, pointing out the importance of using a whitened representation to make optimization easier by reducing the correlation in the posterior distributions. The justifications provided in these previous works for using a whitening representation are, however, scarce and rely on experimental arguments, as also indicated in \GP libraries \citep{gardner2018gpytorch}, \GP tutorials \citep{tutorialSparseGPs} or other research works \citep{DGPIW}. Here, we provide explanations on why using the whitening reparameterization helps the optimization process, finding the prior covariance as the main cause for this. Notwithstanding, note that the non-whitened parameterization has also been used as the default parameterization in other \DGP{} research works \citep{Salimbeni2017,rudner2020inter}. 

Our work also allows us to conclude that, when using a non-whitening parameterization, a reparameterization performed of the variational mean (the prior mean is subtracted from the variational mean), as done in \GPFLOW, does not provide significant additional convergence stability. However, it contributes to the effectiveness of the \PCA prior mean function in avoiding the posterior collapse problem. In the literature, this extra reparameterization is simply introduced as a way to unlock the versatility of interdomain \GP{} frameworks  \citep{tutorialSparseGPs}. Furthermore, its adoption into \DGP{}s or \GP libraries \citep{Salimbeni2017,dutordoir2021gpflux,DGPIW,GPflow2017} has not been properly justified, and is probably just due to a library implementation simplicity. We would like to remark that, to our knowledge, none of these previous works study the relation between optimization results and the different parameterizations considered in the context of \DGP{}s.

Regarding the architectural designs of \DGP{}s, \cite{decoupledinducingDGPs2018havasi} proposes a decoupled parameterization of the \GP{} based on the duality between \GP{}s and Gaussian measures, which was first considered by \cite{decoupledinducing2017chen}. To optimize the proposed model, the authors use the \PCA prior mean function in the inner layers, arguing that it avoids degenerate solutions for the covariance matrix and that it also helps with the initialization of the inducing points. Furthermore, the proposed parameterization is compared to the standard \SVGP{} and to the whitened reparameterization of the variational mean parameter, mentioning that both of them are more stable than the one from \cite{decoupledinducing2017chen}. Also, in deep inter-domain \GP{}s, \cite{rudner2020inter} state that the use of a \PCA prior mean function in the inner layers is required to avoid the pathologies presented by \cite{pathologies_vdn}, similar to \cite{Salimbeni2017} and \cite{DGPIW}. In the three papers, even though the decision of using a \PCA prior mean function is supported by practical evidence, no strong theoretical or empirical arguments are provided to justify that decision. Additionally, \cite{de2020you} provides a purely experimental study on the role of the prior mean function in \GP{}s, limiting the scope of the study to Bayesian optimization. By contrast, our work analyzes the role of the prior mean function in \DGP{}s trained with the \DSVI algorithm from an optimization viewpoint, something that, to our knowledge, has not been done yet.

Finally, the initialization of the \SVGP{} parameters is studied in \cite{GPinitKLS2022zhu}, where a novel method for initializing the inducing points, based on kernel-based least squares, is proposed. While this initialization helps to improve the performance of the model, it also requires solving an optimization problem iteratively before training the model. Furthermore, it is only applied to standard \GP{}s and the study does not consider \DGP{}s. In our work, by contrast, the proposed initialization requires solving a simpler optimization problem and naturally considers \DGP{}s.
\section{Deep Gaussian Processes}
\label{sec:dgp_intro}
In this section, we will detail the theoretical background on \DGP{}s needed to follow the rest of this work. Throughout the exposition, we will assume the reader is familiarized enough with sparse variational \GP{}s \citep{pmlr-v5-titsias09a}, their training using stochastic variational inference \citep{GPs-Big-Data}, and their use in the context of the Doubly Stochastic Variational Inference algorithm for \DGP{}s \citep{Salimbeni2017}. 

\subsection{Problem Formulation}

Consider the observed data \(\mathcal{D} = \{(\X{}{n}, \Y{}{n}) \ \mid \ \X{}{n} \in \Xspace \subseteq \R^D, \Y{}{n} \in \Yspace \subseteq \R^C, \ n = 1,\dots, N\}\), where $D$ and $C$ are the input and output dimensionality, respectively. We write it in matrix form as \(\Xsamples{} = (\X{}{1},\cdots, \X{}{N})^T\) and \(\Ysamples = (\Y{}{1}, \cdots, \Y{}{N})^T\). Our goal is to learn a predictive distribution for $\Y{}{}$ given $\X{}{}$ and the observed data, \ie, $p(\Y{}{}\mid \X{}{},\mathcal{D})$. We can model this problem by placing a prior distribution over a function that maps $\X{}{}$ to a distribution over $\Y{}{}$, and then use the posterior distribution to integrate over all possible functions to make predictions on unseen data instances. We consider a \DGP as the prior distribution of this function.

\begin{figure}[ht]
   
    \centering
    \begin{subfigure}[b]{\textwidth}
        \centering
        \includegraphics[width=0.9\textwidth]{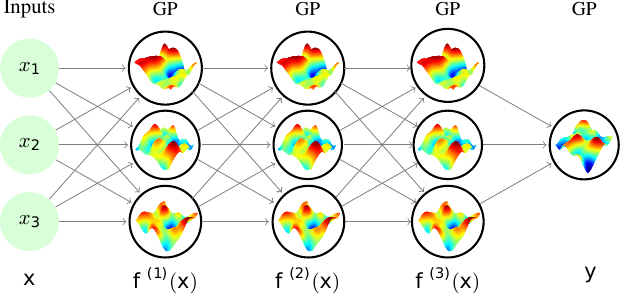} 
        \caption{Neural Network inspired graphical representation of a 4-layer \DGP.}
        \label{fig:subfig2}
    \end{subfigure}

    \vspace{0.5cm} 

    \begin{subfigure}[b]{0.9\textwidth}
        \centering
        \resizebox{\textwidth}{!}{
        \begin{tikzpicture}[baseline]


    \tikzstyle{plate}=[draw, rectangle, rounded corners, minimum width=2.5cm, minimum height=4.6cm, anchor=center]
    \tikzstyle{RV}=[shape=circle,line width=1.0pt,minimum size = 1.25cm, draw=black,fill=gray!20]
    \tikzstyle{RVo}=[shape=circle,line width=1.0pt,minimum size = 1.25cm, draw=black,fill=white]
    \tikzstyle{param}=[fill,circle,inner sep=1.0pt,minimum size=4.0pt]

    
    \node[plate] (r1) at (2.5,0) {};
    \node[anchor=north east] at (r1.north east) {$D^1$};
    \node[RV, yshift = 0.5cm] (c1) at (r1.center) {$\fpos{}{1}$};
    \node [param, label = below:{ $\nu^1$}, yshift = 0.25cm] (h1) at (2.5,-2) {};
    
    \node[plate] (r2) at (5.5,0) {};
    \node[anchor=north east] at (r2.north east) {$D^2$};
    \node[RV, yshift = 0.5cm] (c2) at (r2.center) {$\fpos{}{2}$};
    \node [param, label = below:{ $\nu^2$}, yshift = 0.25cm] (h2) at (5.5,-2) {};

    \node at (7.5,0.5) {$\hdots$};
    
    \node[plate] (r3) at (9.5,0) {};
    \node[anchor=north east] at (r3.north east) {$D^{L-1}$};
    \node[RV, yshift = 0.5cm] (c3) at (r3.center) {$\fpos{}{L-1}$};
    \node [param, label = below:{ $\nu^{L-1}$}, yshift = 0.25cm] (h3) at (9.5,-2) {};
    
    \node[plate] (r4) at (12.5,0) {};
    \node[anchor=north east] at (r4.north east) {$D^L$};
    \node[RV, yshift = 0.5cm] (c4) at (r4.center) {$\fpos{}{L}$};
    \node [param, label = below:{ $\nu^L$}, yshift = 0.25cm] (h4) at (12.5,-2) {};
    
    \node[RVo, yshift = 0.5cm] (x) at (0,0) {$\X{}{}$};
    \node[RVo, yshift = 0.5cm] (y) at (15,0) {$\Y{}{}$};

    \node [plate,minimum height = 3.75cm, inner xsep = 0.5cm, fit = (c1)(c2)(c3)(c4)(x)(y)] (pointsplate) {};
    \node[anchor=south east] at (pointsplate.south east) {$N$};

    \draw[->,  thick]  (x.east) -- (c1.west);
    \draw[->,  thick]  (c1.east) -- (c2.west);
    \draw[->,  thick]  (c3.east) -- (c4.west);
    \draw[->,  thick]  (c4.east) -- (y.west);
    
    \draw[->,  thick]  (h1.north) -- (c1.south);
    \draw[->,  thick]  (h2.north) -- (c2.south);
    \draw[->,  thick]  (h3.north) -- (c3.south);
    \draw[->,  thick]  (h4.north) -- (c4.south);

\end{tikzpicture} 
        }
	    \caption{Probabilistic Graphical Model of the Deep Gaussian Process as originally formulated by \cite{pmlr-v31-damianou13a}.}
        \label{fig:subfig1}
    \end{subfigure}
    \caption{Graphical representations of a Deep Gaussian Process.}
    \label{fig:dgp_prior}
\end{figure}
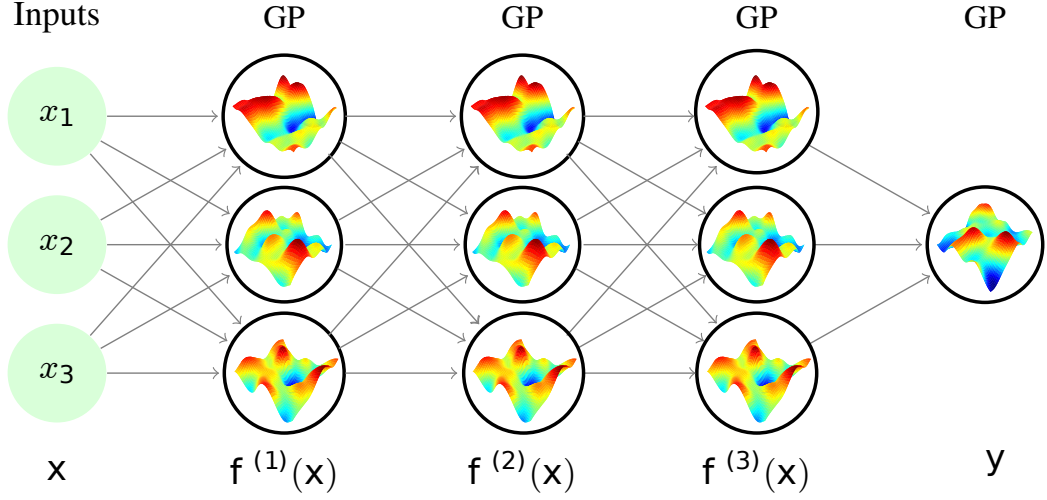
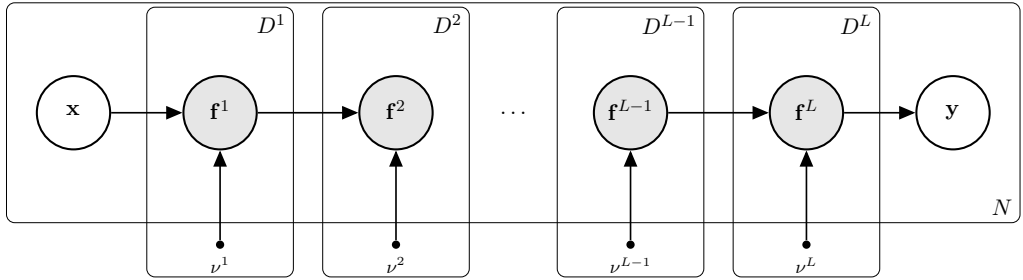

\DGP{}s define a prior distribution over vector-valued functions that surpass some of the limitations of standard \GP models, such as misspecification from assumed joint Gaussianity between points in the function, or smoothness. The joint prior probability over these functions is obtained by a hierarchical concatenation of \GP{}s. The idea is to organize \GP{}s into layers, as in a fully connected neural network, and feed the \GP{} output of one layer as input to the next one. In its original formulation \citep{pmlr-v31-damianou13a}, within the same layers,  
the \GP{}s are conditionally independent given a sample realization of the previous layer. \fig{} \ref{fig:dgp_prior} (a) shows a graphical depiction of a 4-layered \DGP. 
\fig{} \ref{fig:dgp_prior} (b) shows the probabilistic graphical model of a $L$-layered \DGP{}, where each layer has $D^l$ \GP{}s (units).

In our problem formulation, $\fpos{d}{l}$ is a vector random variable with the values of the $d$-th latent function in the layer $l$ evaluated at a particular set of points $\Xsamples{}$. 
We have a \GP{} as the prior of that latent function. Such a GP is characterized by a kernel $K(\cdot,\cdot)$ with parameters $\nupos{d}{l}$. 
A particular sample $s$ of such a latent variable, as in a Monte Carlo method, is denoted as $\fpos{d,s}{l}$, and for a single point $\X{}{n}$, we might also denote 
with $f_{d,n,s}^{l}$ the sample of the corresponding random variable.  We assume each layer has $D^l$ processes, and denote with $\fallpos{l}{} = (\fpos{1}{l},\hdots,\fpos{D^l}{l})$ the set of 
vector random variables belonging to layer $l$. Then, we can write the joint distribution  between the observed data and the \DGP latent variables as:
\begin{align}
    p(\Ysamples{},\Xsamples{},\fallpos{1}{},\hdots,\fallpos{L}{}) &= p(\Ysamples{} \mid \fallpos{L}{})\myprod{l=1}{L} p(\fallpos{l}{}\mid \fallpos{l-1}{})
    = p(\Ysamples{} \mid \fallpos{L}{})\myprod{l=1}{L}\myprod{d=1}{D^l} p(\fpos{d}{l}\mid \fallpos{l-1}{})\,,
\end{align}
with $\fallpos{0}{}\coloneq \Xsamples{}$ and for simplicity we have removed the dependence at each layer with respect to 
$\nuallpos{l}=(\nupos{1}{l},\ldots,\nupos{D}{l})$. 
The form of $ p(\Ysamples{} \mid \fallpos{L}{})$ will depend on the task considered. For example, for multiclass classification, we can use a categorical likelihood or, 
for multivariate regression, a multivariate Gaussian distribution with some assumed noise model that may include dependencies between the regressed dimensions.
Each of the factors $p(\fpos{d}{l}\mid \fallpos{l-1}{})$ describes the prior distribution given by a \GP at inputs $\fallpos{l-1}{}$.

Furthermore, one can easily model dependencies within the processes in the same layer by aggregating the processes using a mixing matrix \citep{multioutputGPs,jankowiak2019neural}, a warping function \citep{tgp,etgp,dtgp}, or a combination of both \citep{multidimTGP}, and then, at inference, make some assumptions to achieve scalability, such as removing dependencies. This is a practice implemented in many common \GP{} libraries \citep{GPflow2017}, as already noted by \cite{etgp}.

\subsection{Inference}

Inference in a \DGP, \ie, computing $p(\fallpos{1}{},\hdots,\fallpos{L}{} \mid \Xsamples{},\Ysamples{})$, is intractable. One of the most popular ways of performing approximate inference in this model is the Doubly Stochastic Variational Inference (\DSVI) algorithm \citep{Salimbeni2017}. Since this is our object of study, we present the formulation involved precisely. 

The \DSVI formulation first introduces a set of $M$ inducing points $\Zsamplespos{d}{l}$ and associated process values $\upos{d}{l}$ for each latent function in the \DGP.
Let $\uallpos{}{l}=(\upos{1}{l},\ldots,\upos{D_l}{l})$. The joint distribution of the observed data and all latent variables then becomes:
\begin{align}
	p(\Ysamples{},\Xsamples{},\{\fallpos{l}{} \}_{l=1}^L,\{\uallpos{l}{}\}_{l=1}^L) &= p(\Ysamples{} \mid \fallpos{L}{})
	\myprod{l=1}{L}\myprod{d=1}{D^l} p(\fpos{d}{l}\mid \fallpos{l-1}{},\upos{d}{l})p(\upos{d}{l})\,,
\end{align}
with $ p(\fpos{d}{l}\mid \fallpos{l-1}{},\upos{d}{l})$ a \GP predictive distribution at $\fallpos{l-1}{}$ given observed targets $\upos{d}{l}$ at $\Zsamplespos{d}{l}$, 
and with $p(\upos{d}{l})$ a \GP prior distribution for $\upos{d}{l}$.
Again, for simplicity, we have removed the dependence at each layer with respect to 
$\nuallpos{l}=(\nupos{1}{l},\ldots,\nupos{D}{l})$ and with respect to $\Zsamplespos{d}{l}$. 

Then, the intractable posterior distribution $ p(\{\fallpos{l}{} \}_{l=1}^L,\{\uallpos{l}{}\}_{l=1}^L\mid\Ysamples{},\Xsamples{})$ is approximated via variational inference, by minimizing the \KLD between an approximate family and the true posterior. The novel contribution from \cite{Salimbeni2017} is to define a variational family that maintains the conditional structure between the layers, as in the prior distribution, while assuming independence within the layers. To yield an efficient algorithm, they also used the uncollapsed version \citep{GPs-Big-Data} of the variational sparse \GP{} formulation \citep{pmlr-v5-titsias09a}. More precisely, the approximate posterior distribution is
\begin{align}
	q(\{\fallpos{l}{} \}_{l=1}^L,\{\uallpos{l}{}\}_{l=1}^L\mid\Ysamples{},\Xsamples{}) &= 
        \myprod{l=1}{L}\myprod{d=1}{D^l} p(\fpos{d}{l}\mid \fallpos{l-1}{},\upos{d}{l})q(\upos{d}{l}) \,,
	\label{eq:vi_posterior}
\end{align}
with $q(\upos{d}{l})$ a Gaussian variational distribution with mean and covariances given by $\varm{d}{l}$ and $\varS{d}{l}$, respectively.
Thus, the approximate predictive distribution for the $d$-th \GP{} process values in layer $l$ is:
\begin{align}
	q(\fpos{d}{l} \mid  \fallpos{l-1}{}) &= \int p(\fpos{d}{l} \mid \upos{d}{l},\fallpos{l-1}{}) q(\upos{d}{l}) \dd \upos{d}{l}\,, 
	\label{eq:vi_predictive}
\end{align}
which is Gaussian with mean and covariances given by:
\begin{align}
    \qfm{}{} &= \Kxz{}\Kzzinv{}\pare{\varm{}{}-\muZ{}} + \muX\,, \label{equ:pos_pred_params_non_whit_mean}\\
   \qfS{}{} &= \Kxx - \Kxz{}\Kzzinv{}\Kzx{} + \Kxz{}\Kzzinv{}\varS{}{}\Kzzinv{}\Kzx{}\,,
   \label{equ:pos_pred_params_non_whit}
\end{align}
where we have removed the dependence with respect to the layer, $l$, and the hidden dimension, $d$, to simplify the notation. We denote with $\muX = \mu(\fallpos{l-1}{})$ and $\Kxx = K(\fallpos{l-1}{}, \fallpos{l-1}{})$, the prior mean and corresponding kernel matrix (prior covariances) evaluated at a set of input points $\fallpos{l-1}{}$,
respectively.  Similarly, we denote with the kernel matrix $\Kzz{}=K(\Zsamplespos{d}{l},\Zsamplespos{d}{l})$ 
the prior covariances associated with the inducing points.
The kernel matrix $\Kxz{}=K(\fallpos{l-1}{},\Zsamplespos{d}{l})$
denotes the prior cross-covariances between the process values at the inducing points and the input locations. 
Note that at the first layer $\fallpos{l-1}{}=\Xsamples$.  This notation will simplify exposing 
the ideas described later on in our paper.

When using the variational posterior in \ueqn (\ref{eq:vi_posterior}), the learning objective results in the Evidence Lower Bound (\ELBO) given by:
\begin{align}
     \ELBO &= \Expectation{ q(\fallpos{1}{},\hdots,\fallpos{L}{})}{ \log p(\Ysamples \mid \fallpos{1}{},\hdots,\fallpos{L}{})} -
	\mysum{l=1}{L} \KLD\left[q(\uallpos{l}{}) \mid\mid p(\uallpos{d}{})\right] \nonumber \\
    &\approx\frac{1}{S}\mysum{s=1}{S}\Expectation{ q(\fallpos{L}{}\mid \fallpos{L-1,s}{})} {\log 
	p(\Ysamples \mid \fallpos{L}{})} -\mysum{l=1}{L}\mysum{d=1}{D^l} \KLD\left[q(\upos{d}{l}) \mid\mid p(\upos{d}{l})\right]\,,
\label{equ:dsvi_dgp_elbo}
\end{align}
where the first expectation is partially approximated using samples obtained at each level in the hierarchy. 
These samples are generated recursively by sampling from the Gaussian distribution in Eq. (\ref{eq:vi_predictive}). 
Namely, $\fallpos{l-1,s}{} \sim q(\fallpos{l-1}{} \mid \fallpos{l-2,s}{})$. 
The expectation over the last layer can be computed either via quadrature methods, Monte Carlo, or exact integration, 
depending on the likelihood. Each of the \KLD{}s between the prior and the approximate posterior in \ueqn (\ref{equ:dsvi_dgp_elbo}) are given by:
\begin{align}
    \KLD=\frac{1}{2}\bra{\log \mid\Kzz{}\mid - \log \mid\varS{}{}\mid -M + \tr{\Kzzinv{}\varS{}{}} + \pareT{\varm{}{}-\muZ{}}\Kzzinv{}\pare{\varm{}{}-\muZ{}}}\,,
    \label{equ:KLD_non_whit}
\end{align}
where, again, we omit subindices for simplicity. 

By using likelihoods that factorize across training instances and output dimensions, this training objective in Eq. (\ref{equ:dsvi_dgp_elbo}) can be computed using minibatches of data and expectations over univariate variational posterior distributions \citep{Salimbeni2017}. Moreover, in some regression problems, the last expectation in Eq. (\ref{equ:dsvi_dgp_elbo}) admits a closed-form solution.
Gradients are typically computed efficiently using the reparameterization trick \citep{VAESeminal} and automatic differentiation.
When the likelihood does not factorize across target dimensions,  expectations can still be computed by sampling from a multivariate Gaussian with at most $C$ dimensions per datapoint, as in multiclass classification, with $C$ the number of output dimensions. If a mixing matrix is used at any level of the hierarchy, one often ignores dependencies so that the joint multivariate Gaussian distribution results in independent sampling per \GP{}, alleviating the quadratic memory cost of keeping the full covariance per datapoint \citep{etgp}. Overall, this results in an efficient algorithm that justifies its wide adoption in many settings. 

When training the model, a single sample per observed instance, \ie, $S=1$, is usually enough.
A predictive distribution for test instances can be computed in a similar way to that of training instances. In this case, more samples $S$ can be used to increase 
the accuracy of the Monte Carlo approximation of the predictive distribution.

\subsection{Parameterizing \DGP{}s}
\label{subsec:parameterizing_dgps}

This section discusses how \DGP{} are parameterized in practice. Model parameterization is relevant to multiple parts of the modeling procedure. Here, we focus on how to parameterize the hierarchical prior defined by a \DGP{}. In particular, we will discuss the non-whitened and whitened parameterizations of the prior and approximate posterior over the process values associated with the inducing points. We also describe a reparameterized version of the non-whitened model used by \cite{Salimbeni2017,GPflow2017} and \cite{dutordoir2021gpflux}. Throughout this subsection, we will omit the layer $l$ and hidden dimension $d$ indices for brevity.

\subsubsection{Non-Whitened \DGP{}s}

In the previous section, we implicitly described the standard non-whitened parameterization. The prior and approximate posterior are given by the following distributions:
\begin{align}
    p(\uu{}) &= \Ngauss{\uu{} \mid \muZ{}, \Kzz{}}\,, \\
    q(\uu{}) &= \Ngauss{\uu{}\mid \varm{}{}, \varS{}{}}\,,
\end{align}
which results in a marginal variational posterior distribution with parameters 
given by \ueqns(\ref{equ:pos_pred_params_non_whit_mean}) and (\ref{equ:pos_pred_params_non_whit}), and a \KLD in the loss function given by \ueqn(\ref{equ:KLD_non_whit}). We can find this parameterization in libraries such as
\GPYTORCH \citep{gardner2018gpytorch} or \GPJAX \citep{GPJax}. We have also found the inter-domain \DGP{} of \cite{rudner2020inter} to use this parameterization.

\subsubsection{Whitened \DGP{}s}
\label{subsec:whitened_gps}

The term whitening is often referred to as a reparameterization of the distributions in a graphical model that removes correlations from the random variables being sampled using \MCMC algorithms. This reparameterization allows an efficient exploration of the target distribution \citep{papaspiliopoulos2007general}, something also known in the community as  \emph{mixing}. Intuitively, this reparameterization unstucks the sampling algorithm from regions of the distribution with high geometrical changes. More precisely, it often results in samples that are more representative of the target distribution and less correlated.
The funnel distribution is a good example of this phenomenon \citep{neal2003slice}.

This reparameterization accounts for transforming the prior process values associated with the inducing points to have $0$ mean and unit variance, 
and, in the context of sparse \GP{}s, was initially introduce to facilitate sampling via \MCMC \citep{hensman2015mcmc}). More precisely, instead of performing inference on $\uu{}$, one performs inference on $\uwhit$, whose prior is defined as a whitened Gaussian distribution. Namely,
\begin{align}
     p(\uwhit) &= \Ngauss{\uwhit \mid \veczero, \matI}\,,
\end{align}
with the inducing values $\uu{}$ being recovered by the following linear transformation:
\begin{align}
    \uu{} = \Lzz{} \cdot \uwhit + \muZ{}\,,
    \label{equ:whit_to_non_white}
\end{align}
where $\Lzz{}$ is the Cholesky factor of $\Kzz{}$, \ie, $\Kzz{} = \Lzz{}\LzzT{}$. When used in the context of variational inference, the variational distribution is also defined in the whitened space:
\begin{align}
q(\uwhit) &= \Ngauss{\uwhit \mid \varmw{}{},\varSw{}{}}\,.
\end{align}

With the reparameterization from \ueqn (\ref{equ:whit_to_non_white}), the prior and variational posterior on the unwhitened space takes the form:
\begin{align}
    p(\uu{}) &= \Ngauss{\uu{} \mid \muZ{}, \Kzz{}}\,,\\
    q(\uu{}) &= \Ngauss{\uu{} \mid \Lzz{}\varmw{}{} + \muZ{}, \Lzz{}\varSw{}{}\LzzT{}}\,,
\end{align}
which results in a predictive Gaussian distribution in Eq. (\ref{eq:vi_predictive}) with mean and covariance given by:
\begin{align}
    \qfmw{}{} &= \Kxz{}\braT{\Lzzinv{}}\varmw{}{} + \muX\,, \\
   \qfSw{}{} &= \Kxx{} - \Kxz{}\Kzzinv{}\Kzx{} + \Kxz{}\braT{\Lzzinv{}}\varSw{}{}\Lzzinv{}\Kzx{}\,.
\label{equ:pos_pred_params_whit}
\end{align}

Note that this reparameterization leaves the \GP prior invariant, and so the prior remains the same as in the non-whitened case, meaning that both parameterizations lead to the same statistical model. The \KLD in the \ELBO can be computed from the prior and posterior distributions in the whitened space. That is,
\begin{align}
	\KLD[q(\uwhit)\mid\mid p(\uwhit)]&=\frac{1}{2}\left[ - \log \mid \varSw{}{}\mid-M + \tr{\varSw{}{}} + \varmw{T}{}\varmw{}{}\right],
    \label{equ:KLD_whit}
\end{align}
since the \KLD is invariant under variable transformations. Note that now, the \KLD does not depend on the location of the inducing points $\Zsamples{}$, nor kernel parameters. Only the data-dependent term of \VI depends 
on them. This is expected to facilitate optimization.

While originally proposed for \MCMC algorithms, this reparameterization has become a \emph{de facto} standard in 
many \VI algorithms applied to \GP{}s and \DGP{}s \citep{DGPIW,tgp,etgp,dtgp} or directly 
recommended by \GP libraries \citep{gardner2018gpytorch} and sparse \GP tutorials 
\citep{tutorialSparseGPs}. The main reason is the association of this parameterization with better convergence 
results, but without a rigorous experimental or theoretical study. 
Our previous unpublished work showed that whitening results in more stable optimization \citep{workshop-nips-mode-collapse}. We discuss this later on in this work.

\subsubsection{Non-Whitened \DGP{}s with a Reparameterized Variational Mean}
\label{usec:non_whit_in_practice}

Another option to implement sparse non-whitened \GP{}s considers a reparameterization of the variational parameters. This is observed in the \GPFLUX \DGP library  \citep{dutordoir2021gpflux}, in the work by \cite{Salimbeni2017}, and also in the implementation of the sparse \GP{}s in the \GPFLOW library \citep{GPflow2017}, which is the framework considered in this work. 

In particular, the  \GPFLOW non-whitened parameterization reparameterizes the variational mean as:
\begin{align}
 \varmr{} = \varm{}{} - \muZ{}   \,.
\end{align}
Plugging this reparameterization into the variational mean and covariance yields a predictive Gaussian distribution in Eq. (\ref{eq:vi_predictive}) with parameters:
\begin{align}
	\qfmr{}{} &= \Kxz{}\Kzzinv{}\varmr{} + \muX\,, \label{equ:pos_pred_params_non_whit_rep_mean}\\
   \qfS{}{} &= \Kxx - \Kxz{}\Kzzinv{}\Kzx{} + \Kxz{}\Kzzinv{}\varS{}{}\Kzzinv{}\Kzx{}\,,
   \label{equ:pos_pred_params_non_whit_rep}
\end{align}
and the resulting \KLD in the \ELBO results in:
\begin{align}
    \KLD=\frac{1}{2}\bra{\log \mid\Kzz{}\mid - \log \mid\varS{}{}\mid -M + \tr{\Kzzinv{}\varS{}{}} + \varmr{T}\Kzzinv{}\varmr{}}\,.
    \label{KLD:non_whit_rep}
\end{align}
By inspection, one can see that this corresponds to the following prior and variational posterior:
\begin{align}
    p(\uu{}) &= \Ngauss{\uu{} \mid \veczero, \Kzz{}}\,, \\
    q(\uu{}) &= \Ngauss{\uu{}\mid \varmr{}, \varS{}{}}\,.
\end{align}
There are some concerns about the use of this reparameterization in \DGP{} models.  First,  the implications behind how this reparameterization modifies the statistical model defined by the \DGP{} prior remain underexplored. Note that, now, the prior mean over the values associated with the inducing points is zero, and different from the prior mean associated with any other input.  Moreover, we found its use in the context of \DGP{}s not justified by \cite{Salimbeni2017}. Similarly, we did not find a justification for it in the \GPFLOW and \GPFLUX library's documentation for both \GP{}s and \DGP{}s \citep{GPflow2017,dutordoir2021gpflux}. The only reference to the use of this parameterization is found in \cite{tutorialSparseGPs} and is related to inter-domain sparse \GP models. It turns out that this reparameterization unlocks the use of any prior mean function in these models, since it removes the convolution integral needed to compute the variational mean posterior, which is often analytically intractable. 

In spite of the aforementioned connection with inter-domain sparse \GP models, we have not found references to the advantages of this reparameterization in the seminal work of inter-domain \GP{}s \citep{seminalInterdomain}, nor in modern frameworks for these models \citep{vanderwilk2020frameworkinterdomainmultioutputgaussian}, nor in their practical implementation \citep{interdomainArmonical}. We think this is because single-layer \GP{} models often assume a \ZERO prior mean function, resulting in a marginal variational posterior mean common for both non-whitened parameterizations, which does not require the convolution integral. In inter-domain \DGP{}s \citep{rudner2020inter}, this reparameterization is not used since the convolutions involving the mean function in the framework do not need to be solved. Overall, this might explain why its adoption in other frameworks such as \GPYTORCH or general \DGP{}s models has not been popularized. Thus, in principle, for general \GP or \DGP models, this reparameterization just implies implementation simplicity, since the routine that evaluates the \KLD can be used for both the whitened and non-whitened parameterizations (\ie, by setting $\Kzz{} = \matI$). 

In any case, our analysis shows that this reparameterization is important to precondition the \VI algorithm at initialization, such that posterior collapse is avoided. More precisely, we found that the standard non-whitened parameterization with a big number of inducing points and the \PCA prior mean function can present a similar behavior to that of \DGP{}s with the \ZERO-mean prior function at initialization, locking the potential preconditioning that this mean function introduces. On the other hand, from an optimization viewpoint, this parameterization does not improve stability in the way the whitened parameterization does. 

Table \ref{tab:parameterizations} summarizes the different parameterizations described in this section.

\begin{table}[H]
\centering
\caption{A summary of the different \DGP prior parameterizations considered in this work.}
\resizebox{\textwidth}{!}{
\begin{tabular}{|c|c|c|c|}
\hline
\textbf{Parameterization} & $p(\uu{})$ and $q(\uu{})$& \textbf{Marginal Variational Posterior} & \textbf{KLD} \\
\hline
Non-whitened & \parbox{4.5cm}{\begin{align*}
    p(\uu{}) &= \Ngauss{\uu{} \mid \muZ{}, \Kzz{}} \\
    q(\uu{}) &= \Ngauss{\uu{}\mid \varm{}{}, \varS{}{}}
\end{align*}} & \parbox{4.5cm}{
\begin{align*}
    \qfm{}{} &= \Kxz{}\Kzzinv{}\pare{\varm{}{}-\muZ{}} + \muX \\
   \qfS{}{} &= \Kxx - \Kxz{}\Kzzinv{}(\Kzzinv{} - \varS{}{} )\Kzzinv{}\Kzx{} 
\end{align*}
}& 
\parbox{4.5cm}{
\begin{align*}
&\frac{1}{2}[\log \mid\Kzz{}\mid - \log \mid\varS{}{}\mid -M + \\
&\tr{\Kzzinv{}\varS{}{}} + \pareT{\varm{}{}-\muZ{}}\Kzzinv{}\pare{\varm{}{}-\muZ{}}]
\end{align*}
} \\\hline
Whitened & \parbox{4.5cm}{\begin{align*}
   p(\uu{}) &= \Ngauss{\uu{} \mid \muZ{}, \Kzz{}}\\
    q(\uu{}) &= \Ngauss{\uu{} \mid \Lzz{}\varmw{}{} + \muZ{}, \Lzz{}\varSw{}{}\LzzT{}}
\end{align*}} & \parbox{4.5cm}{
\begin{align*}
    \qfmw{}{} &= \Kxz{}\braT{\Lzzinv{}}\varmw{}{} + \muX \\
   \qfSw{}{} &= \Kxx{} - \Kxz{}\braT{\Lzzinv{}}\bra{\matI - \varSw{}{}}\Lzzinv{}\Kzx{}
\end{align*}
}& 
\parbox{4.5cm}{
\begin{align*}
&\frac{1}{2}\left[ - \log \mid \varS{}{}\mid-M + \tr{\varS{}{}} + \varm{}{T}\varm{}{}\right]
\end{align*}
} \\\hline
\GPFLOW Non-whitened & \parbox{4.5cm}{\begin{align*}
    p(\uu{}) &= \Ngauss{\uu{} \mid \veczero, \Kzz{}} \\
    q(\uu{}) &= \Ngauss{\uu{}\mid \varmr{}, \varS{}{}}
\end{align*}} &  \parbox{4.5cm}{
\begin{align*}
    \qfmr{}{} &= \Kxz{}\Kzzinv{}\varmr{} + \muX \\
   \qfS{}{} &=  \Kxx - \Kxz{}\Kzzinv{}(\Kzzinv{} - \varS{}{} )\Kzzinv{}\Kzx{} 
\end{align*}
} & 
\parbox{4.5cm}{
\begin{align*}
&\frac{1}{2}[\log \mid\Kzz{}\mid - \log \mid\varS{}{}\mid -M + \\
&\tr{\Kzzinv{}\varS{}{}} + \varmr{T}\Kzzinv{}\varmr{}]
\end{align*}
} \\
\hline
\end{tabular}}
\label{tab:parameterizations}
\end{table}

\subsection{Prior Mean Functions for \DGP{}s}
\label{subsec:computing_mean}

We give details in this section about the evaluation of the prior mean function in the context of \DGP{}s. 
We specifically describe the \PCA prior mean function suggested by \cite{Salimbeni2017}, which is a non-learnable 
linear mean function. The goal is to provide a solid basis to understand the proposed initialization of 
a \ZERO prior mean \DGP described in the following sections, designed to reduce the impact of the posterior collapse problem. 

We consider the standard \DGP architecture, suggested by \cite{Salimbeni2017}, which introduces hidden 
layers of dimensionality $D$ in the \DGP, with $D$ the input data dimension. Therefore, at each hidden layer 
there are $D$ \GP{}s that compute a non-linear transformation of the input.
Consider now a $D$ dimensional input $\X{}{n}$ which is fed into a hidden \DGP layer. 
If the prior mean is zero, then each \GP{} prior mean outputs a scalar zero, \ie, $\mu(\X{}{n})=0$. 
However, if the mean function is a linear mean function parameterized by $\vecw$, then each \GP{} prior mean outputs the 
dot product between the parameter and the input, \ie, $\mu(\X{}{n})=\vecwT\X{}{n}$. This means that, for an arbitrary 
number of \GP{}s per layer, the whole prior mean value for the layer is evaluated as $\matW\X{}{n}$, with $\matW$ a $D \times D$ matrix with rows being the 
parameters of the linear prior mean for each \GP{} unit within the layer.

Typically, $\matW$ is not learnable and has a predefined structure. For example, if the input dimension to a layer 
and the number of units in the layer are the same, as described above, then, we have $\matW=\matI$, as suggested by \cite{Salimbeni2017}. 
For this choice, the output of the prior mean of each \GP{} is the corresponding dimension of $\X{}{n}$. In other words, 
if $D=3$ then $\mu_1(\X{}{n})=x_{1}^{n}$, $\mu_2(\X{}{n})=x_{2}^{n}$ and $\mu_3(\X{}{n})=x_{3}^{n}$, where $\mu_d(\cdot)$ denotes 
the prior mean function of the $d$-th \GP{}. In this case, we can think of each mean function being parameterized 
by a one-hot-encoding vector $\vecw$ with $1$ in the $d$-th dimension and all the other dimensions equal to $0$. 

When the input dimension $D$ is large, it is no longer feasible to have $D$ units in each \DGP hidden layer.
In this case, a smaller number of units $D_l<D$ is considered at each layer $l$. Moreover, $\matW$ is set to a \PCA 
projection matrix from $D$ dimensions to $D_l$ dimensions in the first layer, as suggested by \cite{Salimbeni2017}. 
The parameters of $\matW$, in this case, are found using \PCA on the training data. In the other layers, we simply set $\matW=\matI$, 
as described before.

If the number of \GP{}s in a hidden layer is larger than the input dimensionality of that layer $D$, then $\matW$ is set equal
to the identity matrix with zero-padded column vectors. Thus, the remaining outputs of the layer are set to have a prior 
mean equal to $0$ \citep{Salimbeni2017}.

Summing up, when the \DGP has as many units in each layer as the input dimensionality $D$, which is the typical scenario, the \PCA prior mean 
function of \cite{Salimbeni2017} simply sets each hidden layer to model deviations from the identity function that maps each input 
vector $\X{}{n}$ to the same input vector $\X{}{n}$. 

Importantly, using this mean function implies that the statistical model defined by the \DGP{} in the whitened parameterization is different from the non-whitened with a reparameterized variational mean. This contrasts with the zero mean function, where both parameterizations lead to the same statistical model.

\subsection{Model Initialization and Specification}
\label{subsec:dgp_init}

In variational \DGP{}s, one usually finds the prior design choices and initialization of 
the model and variational parameters described in this section. Thus, our analysis will consider them,
but also other alternatives.

\paragraph{Prior design choices: }

\begin{enumerate}
    \item The covariance functions of the \DGP{} are chosen to be stationary since, by construction, the \DGP{} already models non-stationary processes. Stationary covariance functions depend on the distance between the points. It is common to find square exponential kernels combined with an output scale parameter \citep{Salimbeni2017,rudner2020inter,DGPIW}.  
    \item It is assumed independence between the latent functions at a layer. This is the original prior formulation by \cite{pmlr-v31-damianou13a}. 
	    However, we also discuss typical mixing procedures (such as linear combinations using a mixing matrix), showing that our analysis and proposed solutions are equivalent, under a correct initialization.
	\item We focus our experiments on single-output regression problems, but provide a small discussion in both coupled and independent multi-output Gaussian observation models, and other likelihoods such as those for classification. For regression problems, one usually considers a homoscedastic Gaussian noise model at the output of the \DGP. Namely, $p(y^n|f^L_n) = \Ngauss{y^n\mid f^L_n,\sigma^2}$, with $f^L_n$ the \DGP output associated to $\X{}{n}$. The target value is $y^n$, and the noise variance is $\sigma^2$. Thus, for $N$ training instances, we have:
    \begin{align}
	    p(\Ysamples{}\mid \fpos{}{L}) &=  \myprod{n=1}{N}\Ngauss{y^n\mid f^L_n,\sigma^2}\,,
    \end{align}
    which results in an expected log-likelihood (\ELL) term in the \ELBO given by:
    \begin{align}
	    \ELL &= \mysum{n=1}{N}\frac{1}{S}\mysum{s=1}{S}\Expectation{ q(f^L_n\mid \fpos{n,s}{L-1})} {\log p(y^n \mid f^L_{n})}\,.
	    \label{eq:ell}
    \end{align}
Note that for each training point, we are required to evaluate an expectation approximated by $S$ Monte Carlo samples. 
Importantly, however, the generated samples are independent for each different point in the training set or mini-batch $\Xsamples$ \citep{Salimbeni2017}.
		Thus, we only need to generate separate samples from the predictive distribution $q(\overline{f}_{1,n},\ldots,\overline{f}_{L-1,n})$ for each individual point $\X{}{n}$ in $\Xsamples$. 
These independent samples are represented in \fig \ref{fig:DSVI_zero_pca_pos_toy_at_init_NWR} using green dotted lines. They are used 
to compute the \ELL term in the \ELBO. The red dashed lines represent dependent samples from the predictive distribution at $N$ specific 
locations. Note that these last samples result in smooth functions sampled from the posterior, unlike the independent samples mentioned before.

\end{enumerate}

\paragraph{Model Initialization: }

\begin{enumerate}
	\item The inducing point locations are initialized following \cite{Salimbeni2017}. At the input layer, each $\mathbf{Z}_d^l$ is initialized by 
	    running the k-means algorithm over the training points with as many clusters as inducing points. 
		For the layers after the first one, the inducing points are initialized in the following way:
    \begin{itemize}
	    \item When the layer input dimension is equal to the data dimension, the inducing points are initialized with the same values as those in the input layer.
        \item  When the layer input dimension is smaller than the data dimension, a \PCA projection of the inducing points is performed.
        \item When the layer input dimension is bigger than the data dimension, the inducing points are copied, and the new dimensions are zero-padded.
    \end{itemize}

    \item The noise variance parameter $\sigma^2$  is initialized to values typically between $0.01$ and $1.0$. For example, \cite{Salimbeni2017} mentions an initial value of $0.01$, while their implementation uses an initial value of $1.0$ in practice. \cite{MultiResDGP} and \cite{dtgp} use $0.01$ for this parameter, while \cite{tgp} consider $0.05$, and \cite{DGPIW} use $0.1$. \GPFLOW implements a lower bound of $10^{-6}$ in the noise variance, for numerical stability. We use an intermediate noise variance value of $0.5$ to illustrate the optimization difficulties of \DGP{}s in the next section. However, in our experiments, we set the noise variance to $0.05$. The reason is that one of our conclusions is that smaller values of this parameter at initialization tend to reduce the posterior collapse problem. This is because larger initial noise variances force the model to represent the observed targets just as noise.

    \item The variational parameters of each \GP within the \DGP are often initialized to $\varm{}{} = \veczero, \varS{}{}=10^{-5}\matI$ for the hidden layers. The output layer is initialized with $\varS{}{}=\matI$ in the works by \cite{Salimbeni2017} and \cite{rudner2020inter}\footnote{The implementation released by \cite{Salimbeni2017} initializes $\varS{}{}=10^{-5}\Kzz{}$ (hidden layer) and $\varS{}{}=\Kzz{}$ (output layer) but the work mentions $\varS{}{}=10^{-5}\matI$ and $\varS{}{}=\matI$. In our analysis, we will show that these two initializations are valid.}. Other works that report results using the \DSVI algorithm consider $\varS{}{}=10^{-5}\matI$ in the output layer \citep{tgp,etgp,dtgp}. \cite{MultiResDGP} mentions a random initialization of the variational mean and variances between $0$ and $1$.  In any case, note that the variational variances should always be initialized to a value smaller than or equal to the prior variances.  Moreover, regarding the hidden layers, it seems reasonable to initialize the variational covariance to be small for two reasons. First, we start by being certain about the process values associated with the inducing points, enforcing their adaptation in input space, and we let the model learn the necessary uncertainty. Second, this provides samples from the \DGP{} hierarchy with less variance, improving convergence of the stochastic optimization algorithm. Such an initialization is also recommended in works such as that of \cite{tutorialSparseGPs}. We have, however, observed that \cite{DGPIW} uses $\matI$ to initialize the hidden layer variational covariance. In our work, we will analyze several settings of the initialization of the variational distribution at each layer, following the aforementioned observations and the initial values described.

    \item The kernel parameters depend upon the specific kernel. We use a Radial Basis Function kernel, given by
    \begin{equation}
    \label{eq:kernel:rbf}
     K(\X{}{},\X{}{'} )=\sigma_o \exp \left(-{\frac {\|\X{}{} -\X{}{'} \|^{2}}{2\ell ^{2}}}\right).
    \end{equation}
		\cite{Salimbeni2017} initialize $\ell = 2$, which results in smooth functions (assuming the data have been standardized). The output scale parameter $\sigma_o$ takes a value of $1.0$ at initialization and controls the variance of the covariance function, as in the demos given in their source code. In their paper, they use an output scale of $2.0$. For this reason, we use an output scale of $2.0$ in the  \UCI benchmark experiments. In our toy problem, we set the output scale to $1.0$. We also analyze the influence of this parameter at initialization.
    
    As we will justify and analyze later, in some of our models, we propose to adapt the length-scale to be more representative of the data. In particular, the length-scale is selected so that it minimizes the RMSE of the initialized model on the train set. Thus, we follow the common practice of choosing the length-scale adapted to the type of data we are trying to model \citep{MultiResDGP, tgp, etgp, rudner2020inter, tutorialSparseGPs}. As with the output scale, we also analyze the influence of the initial length-scale in the next section.

\end{enumerate}

The variational covariance $\varS{}{}$ is parameterized via its Cholesky factor, which results in an unconstrained parameterization. Parameters requiring some restriction, such as the variance being positive, are usually parameterized on an unconstrained domain and then mapped through an invertible mapping to the restricted domain \citep{GPflow2017}.

\section{Posterior Collapse in the Context of \DGP{}s}\label{sec:optimization:difficulties}

In this section, we present and analyze the optimization problems that often arise when training \DGP{}s with the \DSVI algorithm. Furthermore, we show some modeling choices that help overcome these issues. In particular, we focus our analysis on the choice of the inner layer mean function and on the three parameterizations previously introduced. We also analyze and discuss the use of different numbers of inducing points, or the initial variational and model parameters outlined in \usec \ref{subsec:dgp_init}. 

On one side, we show that a constant prior mean function, with the typical initialization discussed in the previous section (in particular $\varm{}{}=\varmw{}=\varmr{}{}=\veczero$), is likely to lead to the posterior collapse problem. That is, the model minimizes the \KLD in the \ELBO by setting the variational parameters to those of the prior. The \ELL term is then maximized by setting the variance of the noise equal to the variance of the targets, assuming they have zero mean. That is, the data are explained as pure noise. Our analysis starts with illustrative and intuitive descriptions of the problem to get a general idea. The problem is then formalized by noticing that a \DGP at initialization can be approximated via a \SVGP. With this observation, we then use the coordinate updates of the variational parameters of a \SVGP \citep{GPs-Big-Data,pmlr-v5-titsias09a} and derive the updates for the noise parameter, showing that a constant prior mean function enforces the model to learn a large noise variance while collapsing the variational parameters to the prior. This gives an idea of the type of local minima that the gradient ascent method often finds during the first iterations of optimization.

Furthermore, we show that the posterior collapse problem is present in both parameterizations (whitening and non-whitening), justifying the use of the \PCA prior mean function proposed by \cite{Salimbeni2017} in the non-whitened parameterization and subsequent usage in the whitened parameterization. We also show that, in the non-whitened case, the variational mean reparameterization of \usec~\ref{usec:non_whit_in_practice} benefits further from the \PCA prior mean function than the standard non-whitened parameterization. Notwithstanding, we observe that posterior collapse in non-whitened models can additionally arise from training instabilities resulting from this parameterization \citep{workshop-nips-mode-collapse}. Additionally, we provide insights into the source of these instabilities. Following this observation, we realize that any form of noise in the optimization procedure might result in posterior collapse, regardless of the parameterization. This additional noise might come from a small number of inducing points, a big variational variance $\varS{}{}$, or a big kernel output scale parameter $\sigma_o$. Based on this observation, the experiment section shows that a whitened \DGP{} with $10$ layers and the \PCA prior mean function, can also present posterior collapse when the variance coming from the inner layers is big at initialization, and a small number of inducing points is used. This result questions the single original motivation behind using the \PCA prior mean function, \ie, simply a large depth of the model \citep{Salimbeni2017}.
\subsection{Impact of Inner Layer Prior Mean Functions} \label{sec:optimization:subsec:inner:mean}

While \GP{}s usually have \ZERO prior mean functions, \cite{Salimbeni2017} suggested using a non-learnable linear mean function in the inner layers of a \DGP{}s, as described in \usec~\ref{subsec:computing_mean}. Whenever the input and output dimensionality of the layer is the same (the most common scenario), the mean function copies the input into the output $\Xsamples=\mu(\Xsamples)$. When the input dimensionality is higher, a \PCA projection of the inputs is performed. By contrast, when the output dimensionality is higher, the input is copied to the output and padded with zeros.

Attending to \cite{Salimbeni2017}, this design choice is motivated to avoid the pathologies discovered by \cite{pathologies_vdn} in very deep \DGP{}s with \ZERO prior mean functions. They mentioned that with this inner mean function, it is effective to initialize the variational mean $\varm{d}{l}=\veczero$ for any \GP{}. However, the following two observations raise concerns regarding the validity of this statement:  1) \cite{Salimbeni2017} use this mean function even in $2-$layer \DGP{}s, which are far from the deepness of the models (around $11$ hidden layers) studied by \cite{pathologies_vdn}; and 2) \cite{pmlr-v31-damianou13a} reported good predictive performance using \ZERO prior mean \DGP{}s with $5$ hidden layers, being the inference algorithm the main difference \wrt \cite{Salimbeni2017}. 

Building upon these two observations, we conduct a deeper analysis and illustrate that the poor performance of a \DGP{} with a constant prior mean function and the number of layers considered by \cite{Salimbeni2017} is a consequence of posterior collapse, which originates from the inference algorithm with the initialization outlined in \usec \ref{subsec:dgp_init}. In particular, the choice $\varm{}{}=\varmr{}{}=\varmw{}{}=\veczero$. This behavior is therefore not attributable to the non-injective mapping problem discussed by \cite{pathologies_vdn}. We show that setting the variational mean equal to zero conditions the output layer of a \ZERO prior mean \DGP model to \emph{see}, just after initialization, a virtual dataset consisting of pairs of $\{\veczero,\Y{}{n}\}^N_{n=1}$, for any training point $\X{}{n}$. That is, the output of the last inner layer, which is the input to the last \DGP layer, tends to predict only zeros (just after initialization) for any input point $\X{}{n}$. By contrast, a \PCA prior mean \DGP does not have this problem. At initialization, the last inner layer will tend to predict $\X{}{n}$ given $\X{}{n}$ as an input.

The consequence is that the \PCA prior mean function, when $\varm{}{}=\varmr{}{}=\varmw{}{}=\veczero$, conditions the optimization algorithm to run away from a minimum that occurs when the \ELL term in the \ELBO is minimized by learning a huge observation noise. The reason is that, on expectation, it makes the model output a predictive variance that does not agree with the observed data. This forces the model to adjust the variational parameters to compensate for this behavior.  By contrast, a constant \ZERO prior mean function with the aforementioned initialization outputs the same predictive variance at any point (recall that each input point $\X{}{n}$ is mapped to $\mathbf{0}$). This implies that a homoscedastic noise model will favor learning a huge observation noise, because the last \GP model cannot disaggregate information from training points when the input it receives is always $\veczero$ for each training point. This could be seen as a non-injective problem attributable to the inference algorithm, but not to the \DGP prior, which is the one studied by \cite{pathologies_vdn}.

Summing up, the origin of the posterior collapse problem is not the initial predictive mean at the training points \citep{workshop-nips-mode-collapse}, which is equal to zero at initialization for each prior mean function considered in the inner layers (\ie, \ZERO or \PCA), but to the initial predictive mean at the inner layers, and the final output predictive variance, which are different at initialization, for each inner-layer prior mean function.

\subsection{An Analysis of the Posterior Collapse Problem on a Toy Problem}
\label{usec:graphical_mean_collapse_description}

To validate our claims, we first analyze the training signals that are propagated through the \DGP{} hierarchy and used to evaluate the \ELL term in the \ELBO. To do so, we design a toy experiment which will allow us to analyze how the different mean functions precondition the \DSVI{} algorithm at initialization, under different settings, including the number of inducing points, different initialization parameters, or the different parameterizations described in \usec \ref{subsec:parameterizing_dgps}.  We depict how the \PCA prior mean function consistently shapes the predictive variance in a way such that the \ELL can be very low at points where data are present, and the only way to increase it is by modifying the variational parameters. Furthermore, we show that $\varm{}{}=\varmr{}{}=\varmw{}{}=\veczero$ is the primary factor contributing to the bad performance of any constant prior mean \DGP{}. This analysis reveals that the shape of the predictive variance is determined by the inputs received at the output layer from the inner layers at initialization. This observation allows us to formally extend our explanation by approximating a \DGP at initialization using a \SVGP. This enables formalizing a coordinate optimization algorithm that can be used to understand the parameter updates carried out in a \DGP during the initial training iterations. This section considers \DGP{}s with one single \GP per layer, for illustration purposes. However, the exact same analysis and behavior is expected to hold when an arbitrary number of \GP{}s is used per layer. 

The rest of the section is organized as follows: First, in \usec \ref{sec:ell_and_initial_predictive}, we describe the example setup, introduce our toy problem, and then proceed by examining different initializations and the whitened and non-whitened parameterizations. We also explain why the two different prior mean functions generate different predictive distributions at initialization in \usec \ref{sec:reasons}. After this, in \usec \ref{sec:virtual_dataset} and \usec \ref{sec:impact_on_ell}, we analyze how the initial predictive distribution affects model fitting and why it may lead to posterior collapse. This is done under a non-whitened with a reparameterized variational mean under the initialization $\varS{}{}=10^{-5}\matI$ in all layers. Then, we analyze different initialization settings and discuss the whitened parameterization. After this, in \usec \ref{sec:Iat_output}, we study the configuration of \cite{Salimbeni2017}, where the inner layer's variational covariance are initialized to $\varSw{}{}=\varS{}{}=10^{-5}\matI$, but the output layer variational covariances are initialized to $\varS{}{}=\varSw{}{}=\matI$. Later, we discuss the role of the kernel parameters in \usec \ref{sec:impact_hypers}, and the number of inducing points in \usec \ref{sec:opt_analysis_more_inducing}. Finally, we briefly discuss the standard non-whitened parameterization implemented in \GPYTORCH and \GPJAX, the effect of a mixing matrix and warping functions.

\subsubsection{\ELL Term and Initial Predictive Distribution}
\label{sec:ell_and_initial_predictive}

In a \DGP, the \ELL term is defined in Eq. (\ref{eq:ell}). When training single-output \DGP{}s with a homoscedastic Gaussian observation model, 
as outlined in \usec \ref{subsec:dgp_init}, the \ELL can be estimated via Monte Carlo sampling as:
\begin{align}
	\ELL & = \mysum{n=1}{N}\frac{1}{S}\mysum{s=1}{S} \log \Ngauss{y_n \mid f_{n,s}^{L},\sigma^2}\,,
    \label{equ:ell_homeoscedastic}
\end{align}
with $f_{n,s}^{L} \sim q(\fpos{n,s}{L}\mid \fpos{n,s}{L-1}), \hdots, \fpos{n,s}{1} \sim  q(\fpos{n,s}{1}\mid \X{}{n})$. 
While the expectation over the last layer, for this particular noise model, can be integrated out analytically, our 
explanation remains simpler considering this Monte Carlo approximation. Recall that the required samples are obtained from the 
predictive distribution at each training point $\X{}{n}$. Thus, we only need to analyze the propagated samples,
which are independently generated for each $\X{}{n}$.

\begin{figure}[!t]
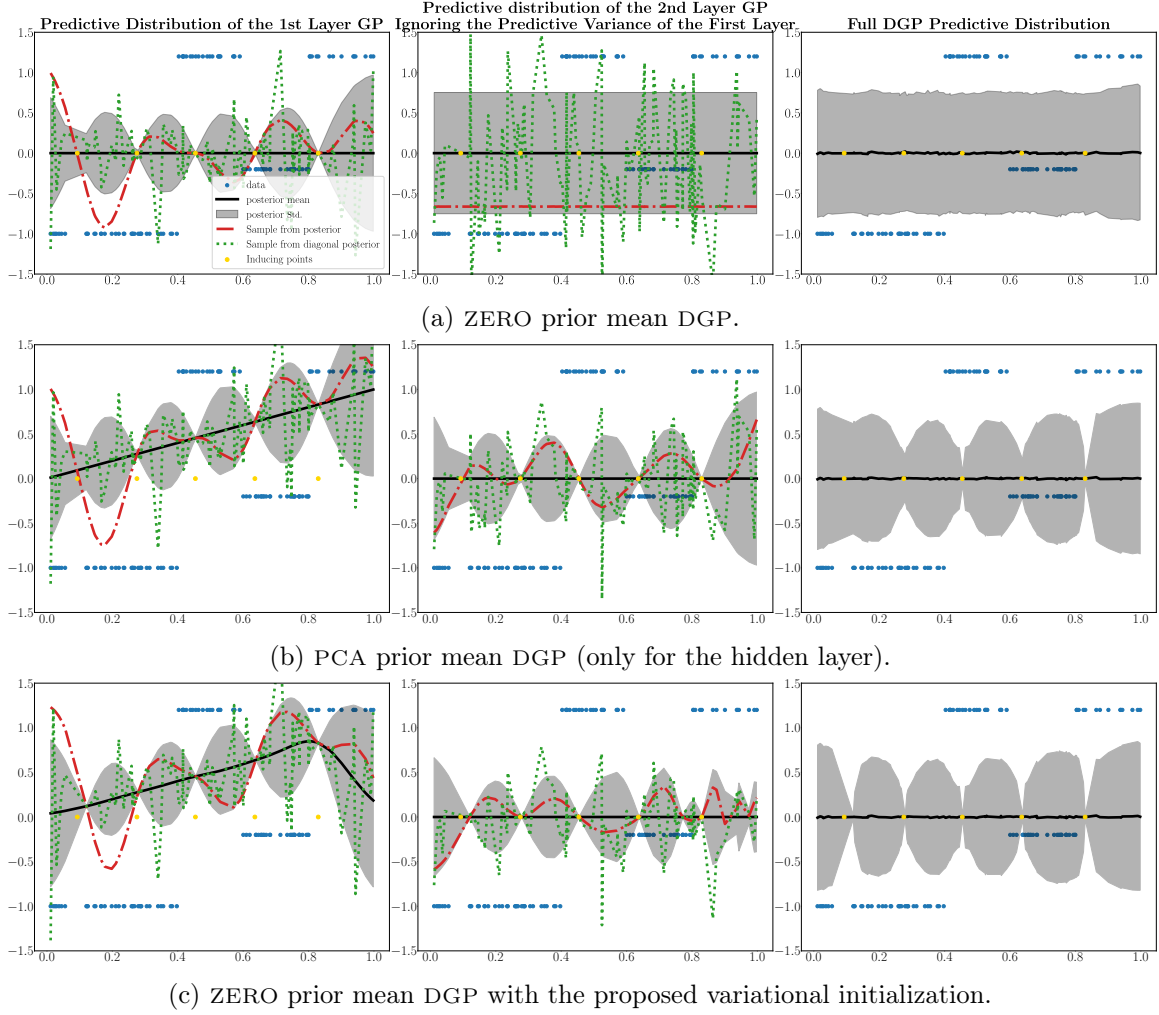

    \centering
    \begin{subfigure}{\textwidth}
        \centering
        \includegraphics[width=\textwidth]{imgs_jmlr/4-initialization_issues/S_out_e5_S_inner_e5/marginal_var_pos_at_init_model_zero_S_inner_1e-05_S_out_1e-05_num_Z_5_ls_0.1_whiten_False.pdf}
        \caption{\ZERO prior mean \DGP{}.}
    \end{subfigure}
    \begin{subfigure}{\textwidth}
        \centering
    \includegraphics[width=\textwidth]{imgs_jmlr/4-initialization_issues/S_out_e5_S_inner_e5/marginal_var_pos_at_init_model_PCA_S_inner_1e-05_S_out_1e-05_num_Z_5_ls_0.1_whiten_False.pdf} 
	    \caption{\PCA prior mean \DGP{} (only for the hidden layer).}
    \end{subfigure}
    \begin{subfigure}{\textwidth}
        \centering
        \includegraphics[width=\textwidth]{imgs_jmlr/4-initialization_issues/S_out_e5_S_inner_e5/marginal_var_pos_at_init_model_zero_m0_S_inner_1e-05_S_out_1e-05_num_Z_5_ls_0.1_whiten_False.pdf} 
        \caption{\ZERO prior mean \DGP{} with the proposed variational initialization.}
    \end{subfigure}
	\caption{From top to bottom, we display the predictive distribution of two layers \DGP{}s with \ZERO prior mean function (first row), with \PCA prior mean functions in the hidden layer (second row), and with \ZERO prior mean functions with the proposed variational initialization (third row). The hidden layer and output layer's $\varS{}{}$ is set to $10^{-5}\matI$ and the non-whitened parameterization implemented by \GPFLOW is considered. The left column shows samples and predictive distribution of the hidden layer \GP. The center column shows samples and predictive distribution of the output layer \GP, if we condition on the hidden layer's posterior mean, \ie by ignoring the uncertainty in the predictive distribution of the hidden layer. The right column shows the predictive distribution of the full \DGP{} obtained using the Monte Carlo approximation in \ueqn (\ref{equ:moments_full_posterior}).}
    \label{fig:DSVI_zero_pca_pos_toy_at_init_NWR}
\end{figure}

To understand the effect of the generated samples on the training objective,
we consider a one-dimensional single-output regression problem, 
where the goal is to model a latent function $f: \R \to \R$. 
We consider a \DGP with 2 layers. That is, a hidden layer and an output layer. The number of \GP{}s (units) in the hidden layer
is set equal to one. The problem considered is displayed in \fig \ref{fig:DSVI_zero_pca_pos_toy_at_init_NWR}.
This is the toy problem introduced by \cite{rudner2020inter}.
Blue points denote the training data. We observe that the target latent function is close to a step
function, with sharp transitions that are challenging to model using the squared exponential kernel and standard \GP{}s.

Each sub-figure in \fig \ref{fig:DSVI_zero_pca_pos_toy_at_init_NWR} corresponds to different \DGP{}s using the non-whitened parameterization implemented by \GPFLOW, 
where the prior mean or the initialization of the variational parameters is different. In this figure, the \DGP modes have not been trained, and we simply 
report how they behave at initialization. The first column of the figure shows the predictive distribution at initialization (mean and variance) of 
the \GP in the hidden layer $q(\fpos{}{1}\mid \Xsamples)$.  See \ueqn (\ref{equ:pos_pred_params_non_whit_rep_mean}) and Eq. (\ref{equ:pos_pred_params_non_whit_rep}).
Moreover, we also display joint samples from such a predictive distribution (using full covariances)
$\qfS{}{}$ (red dashed line) and from the marginal distribution, \ie with a diagonal covariance given by $\diag(\qfS{}{})$ (green dotted line). 
Note that the samples from the marginal distribution (green dotted line) are not smooth, unlike the samples from the joint distribution (red dashed line). 
Recall that the marginal samples are the ones used to evaluate the \ELL term, as mentioned previously.
We also report the initial location of the inducing points as yellow points. The second column of \fig \ref{fig:DSVI_zero_pca_pos_toy_at_init_NWR} displays 
similar information, at initialization, but for the \GP in the last layer of the \DGP model (second layer). This is the output layer. Importantly, however, 
we consider here the predictive distribution for each input point after computing the predictive mean of the first layer and letting that predictive mean
go through the second layer. This ignores prediction uncertainty resulting from the \GP in the first layer (only considers the nonlinearities given by the predictive mean), 
but gives a general idea about how the predictive distribution at the output layer changes with respect to the input of the \DGP in a tractable form. In other words, it is the predictive 
distribution of the output layer if we condition on the inner layer posterior mean (black line).
Last, the third column shows the full predictive distribution of the \DGP at initialization. This predictive distribution is obtained by
letting the input data go through the two layers of the \DGP. Since this distribution is intractable, we approximate it using $S=1000$ Monte Carlo samples as follows:
\begin{align}\label{equ:moments_full_posterior}
	\Expectation{}{\fpos{n}{2}} &= \frac{1}{S}\mysum{s=1}{S} \fpos{n,s}{2}\,, \quad 
   \mathbb{VAR}[\fpos{n}{2}] = \frac{1}{S}\mysum{s=1}{S} \pare{\fpos{n,s}{2}}^2 -  \Expectation{}{\fpos{n}{2}}^2\,,
\end{align}
with $\fpos{n,s}{2} \sim q(\fpos{n,s}{2}\mid \fpos{n,s}{1}); \fpos{n,s}{1} \sim  q(\fpos{n,s}{1}\mid \X{}{n})$.  

When we look at the figure row-wise instead of column-wise, we observe similar information, but when the prior mean or the variational 
initialization is changed.  Specifically, the first row of \fig \ref{fig:DSVI_zero_pca_pos_toy_at_init_NWR} shows
the predictive distributions, at initialization of a \DGP{} with the \ZERO prior mean function in each \GP{}.
The second row corresponds to the \PCA prior mean function (only in the hidden layer, in the output layer, the \ZERO prior mean function is employed). 
In each row, the variational initialization considered sets $\varS{}{}=10^{-5}\matI$ in the inner layer and the output layer. 
We also set a length-scale of $0.1$, and an output scale of $1.0$. The noise variance $\sigma^2$ is set to a small value.
Finally, the third row corresponds to a \ZERO prior mean \DGP with our proposed initialization, which 
we describe in detail later in \usec \ref{sec:proposed_initialization}. 
In all cases, the number of inducing points is set to $5$, initialized via K-means (for both the inner and output layer).

\fig \ref{fig:DSVI_zero_pca_pos_toy_at_init_NWR} shows that a different prior mean function, or a different variational initialization,
has a strong effect on the resulting initial \DGP predictive distribution (third column), but only in terms of the predictive standard deviation. In all cases,
the \DGP predictive mean is equal to zero at initialization. The predictive variance is small at the inducing points
when the \PCA prior mean function is used in the hidden layer \GP, and when the proposed variational initialization is used.
By contrast, when the \ZERO prior mean function is used, the predictive variance is constant for each input point.

\subsubsection{Explaining the Differences in the Initial Predictive Variance}
\label{sec:reasons}

The most striking difference in \fig \ref{fig:DSVI_zero_pca_pos_toy_at_init_NWR} is found in the \DGP initial predictive variance when \ZERO and \PCA prior mean functions are used (first row and second row, right column). The initial \DGP predictive means are, however, similar and equal to zero. 

\ueqn (\ref{equ:pos_pred_params_non_whit_rep_mean}) indicates that when $\varmr{}{}=\veczero$, the predictive mean $\qfmr{}{}$ at the inner layer is $\veczero$ for the \ZERO mean prior function, and $\Xsamples{}$, \ie, the \DGP inputs, for the \PCA prior mean function. Thus, samples from the predictive distribution of the inner layer will be around zero (the top-left sub-figure)  and around values of $\Xsamples{}$ (middle-left sub-figure), respectively. However, since the output layer prior mean is zero for both models, the output layer predictive mean is zero for each model at initialization (top-middle and center sub-figures). Thus, when we consider the full \DGP predictive distribution, we obtain zero as the initial predictive mean (top-right and middle-right sub-figures). 

Regarding the initial predictive variance, there is a clear difference in the output layer. In the \ZERO mean prior \DGP{}, it is constant.
By contrast, in the \PCA prior mean \DGP, we observe a wavy shape, with zero variance at the inducing points. For the input layer, the initial predictive variance 
next to an inducing point $\Z{i}$ will take the value corresponding to the $i$-th entry of $\diag(\varS{}{})$. In particular, in  
\ueqn (\ref{equ:pos_pred_params_non_whit_rep}), when $\Xsamples{}=\mathbf{Z}$, the predictive covariances are equal to $\varS{}{}$.
Since the variational covariance is initialized to $\varS{}{}=10^{-5}\matI$, the predictive variance is almost zero at the inducing point locations. 
This is something that happens both in \PCA and \ZERO mean prior \DGP{} models. The reason is that the predictive covariances do not depend on the prior 
mean in \ueqn (\ref{equ:pos_pred_params_non_whit_rep}). 

For input points far from the inducing points, one should expect that $\K{\X{}{i}}{\Zsamples{}}\approx\veczero$. Thus, the predictive covariance in \ueqn (\ref{equ:pos_pred_params_non_whit_rep}) will approximately be given by $\K{\X{}{i}}{\X{}{i}}$. This implies that any point far from the inducing points will get a predictive variance given by the kernel output scale parameter, $\sigma_o$, which is initialized to $1.0$. This is visible in \fig \ref{fig:DSVI_zero_pca_pos_toy_at_init_NWR} for the inner layer (left column) in each \DGP model. We do not observe a value exactly at $\sigma_o=1.0$ since $\K{\X{}{i}}{\Zsamples{}}$ is not exactly zero, due to the value of the length-scale.

Regarding the predictive variance of the \GP at the output layer (middle-column in \fig \ref{fig:DSVI_zero_pca_pos_toy_at_init_NWR}), 
we observe that in the \ZERO prior mean \DGP it is constant. Recall that, for any layer after the first one, the predictive covariance depends on the inducing locations $\Zsamples{}$ (which, at initialization, are the same across layers, see \usec \ref{subsec:dgp_init}), and the samples of the previous layer.
Importantly, however, the samples from the previous layer will be similar to 
 $\Xsamples{}$ for the \PCA prior mean  \DGP{}, and to a vector of $\veczero$ for the \ZERO prior mean \DGP{} (ignoring the prediction uncertainty from the first layer). 
Consequently, for the \ZERO mean prior function, the predictive covariances are:
\begin{align}
	\qfS{}{} & = \K{\veczero}{\veczero} - \K{\veczero}{\Zsamples{}}\Kzzinv{}\K{\Zsamples{}}{\veczero}+\K{\veczero}{\Zsamples{}}\Kzzinv{}\varS{}{}\Kzzinv{}\K{\Zsamples{}}{\veczero},
\end{align}
for any point from the input space, \emph{even at the inducing points locations}. 
This results in similar predictive variances for each point in the input space.
Furthermore, the generated samples at this layer are expected to be constant in terms of 
the \DGP inputs, as illustrated by the red dashed curve in the top-middle sub-figure in \fig \ref{fig:DSVI_zero_pca_pos_toy_at_init_NWR}.
The reason is that the predictive mean of the first layer will map each input to a value equal to zero. This implies a posterior covariance $\qfS{}{}$ with the same values in all its entries.
Since the marginal variances are constant, we also observe a constant variance across independent samples, as illustrated by the green dotted curve.

Whether the output layer predictive variance is close to the output scale kernel parameter or the initial values specified in $\varS{}{}$ depends on how far $\veczero$ is from the inducing points. A greater number of inducing points will result in a posterior more concentrated at the initial $\varS{}{}$ values. The reason is that there is a bigger chance for an inducing point being close to $\veczero$.

We note, however, that for the \PCA prior mean \DGP{}, things are different. In this case, at initialization, the hidden layer predictive mean is equal to the identity function. Namely, it outputs $\Xsamples$ given $\Xsamples$ as an input. Thus, the predictive variance for the output layer has exactly the same behavior as the input layer for the \ZERO prior mean \DGP. Specifically, rather than always evaluating on $\veczero$, we evaluate on $\Xsamples{}$ when $\Xsamples{}$ is the \DGP input. The consequence is that the output layer predictive variance has the same wavy shape as shown in the input layer, but around a mean value of zero, since the output layer prior mean is the \ZERO mean function. The last row in \fig \ref{fig:DSVI_zero_pca_pos_toy_at_init_NWR} shows that our proposed initialization results in a \ZERO mean \DGP with a similar behavior as that of the \PCA prior mean \DGP at initialization.

While our reasoning regarding the shape of the output layer posterior variance relies on ignoring the predictive variance from the hidden layer, we expect a similar behavior in the full \DGP predictive distribution. This is indeed the case, as illustrated by  \fig \ref{fig:DSVI_zero_pca_pos_toy_at_init_NWR} (right column).
The reason is that the initial predictive variances are fairly small at initialization. Thus, the \ZERO prior mean \DGP will output values $\overset{\sim}{\veczero}$ close to $\veczero$ in the hidden layer, for each input in $\Xsamples$. Similarly, the \PCA prior mean \DGP will output $\tilde{\mathbf{x}}^n$ close to $\mathbf{x}^n$, for each input $\mathbf{x}^n$ in $\Xsamples$, in the hidden layer.
Overall, this implies that, at initialization, the \ELL for each training sample in the \ZERO mean \DGP{} is evaluated using a sampling procedure similar to
$\overset{\sim}{\veczero{}{}}  \sim  q(\fpos{n,s}{1}\mid \X{}{n})$ and $\fpos{n,s}{2} \sim q(\fpos{n,s}{2}\mid \overset{\sim}{\veczero{}{}})$.
By contrast, the \PCA mean \DGP{} is evaluated using a sampling procedure similar to
$\tilde{\mathbf{x}}^n  \sim  q(\fpos{n,s}{1}\mid \X{}{n})$ and $\fpos{n,s}{2} \sim q(\fpos{n,s}{2}\mid \tilde{\mathbf{x}}^n)$.

\subsubsection{Virtual Dataset Seen by a \ZERO Prior Mean \DGP{} at Initialization}
\label{sec:virtual_dataset}

An intuitive argument to understand why the \ZERO prior mean \DGP is likely to have a posterior collapse problem is related to the \emph{virtual} dataset that it sees during training, just after initialization. Specifically, we have seen that the output layer of the \ZERO prior mean \DGP{} receives input values $\overset{\sim}{\veczero{}{}}$ close to $\veczero$, for any point in the training set. This means that the virtual dataset that is being used to optimize the \ELL is made up from pairs $\{(\overset{\sim}{\veczero{}{}},y^n)\}^N_{n=1}$. Thus, from the model’s perspective, every input vector is the same (all of them being close to the zero vector). Therefore, in the output layer, there is no variability in $\X{}{n}$ that could explain the variability in $y^n$. As a consequence, for the model, it is easier to represent $y^n$ purely as noise and set the \KLD equal to zero.

\subsubsection{Impact of Initial Predictive Variances in the \ELL}
\label{sec:impact_on_ell}

How do the differences in the initial predictive variance influence the \ELL?  An intuitive explanation starts by noting that the \ELL term involves the expected value of the score function $\log \Ngauss{y^n \mid f_n^L,\sigma^2}$ for each training point, under the marginal distribution $q(f_n^L)$ over the last layer. Assuming $q(f_n^L)$ has finite mean and variance, then the \ELL can be obtained analytically. 
That is,
\begin{align}
	\ELL &= \sum_{n=1}^N \mathds{E}_{q}[\log \Ngauss{y^n \mid f_n^L,\sigma^2}]\\
    &= 
	\sum_{n=1}^N \log \Ngauss{y^n \mid \mathds{E}_{q}[f_n^L],\sigma^2} - \frac{1}{2\sigma^2}\mathbb{VAR}(f_n^L)\\
    &= \mysum{n=1}{N} -\log\sigma\sqrt{2\pi} -\frac{1}{2\sigma^2}(y^n -\qfm{n}{L})^2 -\frac{1}{2\sigma^2}\tr{\qfS{}{L}} \,.
	\label{eq:ell_expansion}
\end{align}
Here, while we use a similar notation to that of the \GP predictive mean and variance, $\qfm{n}{}=\mathds{E}_q[f_n^L]$ and $\qfS{}{L}$ represent the mean and covariance of a non-Gaussian distribution. Therefore, the learning algorithm finds parameter values so that this score function is high, which, broadly speaking, implies that the resulting predictive mean is close to the targets, and the predictive variance $\mathbb{VAR}(f_n^L)$ is as small as possible, ideally $0$. The optimal parameter $\sigma^2$ is the expected value of the average squared distance between the target $y^n$ and the \DGP output $f_n^L$, see appendix \ref{sec:app:A}. In other words, it is the sample variance given optimal posterior mean, \ie how much uncertainty remains under-explained, plus a contribution that comes from the trace term.

The right column of \fig \ref{fig:DSVI_zero_pca_pos_toy_at_init_NWR} shows that the \PCA prior mean \DGP{} provides a predictive distribution at initialization that can highly disagree with the observed data. For example, at nearly $\X{}{n}=0.5$ (which coincides with an inducing point location $i$), the variance is almost zero while the corresponding $y^n$ is around $1.2$. This will incur a very low \ELL term, and so the model will be forced to update the parameters to improve the prediction of $y^n$. At this point, $\X{}{n}$, the kernel parameters ($\nupos{}{}$) and the inducing point locations ($\Zsamples{}$) do not influence the \ELL term. This is so because the predictive variance is given by $\varS{}{}$ (since we are computing predictions at an inducing point) and the predictive mean is zero due to the initialization. More precisely, as shown by the two equations below, any hyper-parameter gradient resulting from the predictive mean will be multiplied by $\veczero$ (due to the predictive mean being linearly related to $\varm{r}{}=\veczero$) and any hyper-parameter gradient from the predictive variance will be zero, as initial predictive variances are given just by $\varS{}{}$. In particular, note that if we approximate the predictive mean of the \DGP through Monte Carlo, then using \ueqn (\ref{equ:pos_pred_params_non_whit_rep_mean}) and \ueqn (\ref{equ:pos_pred_params_non_whit_rep}), the gradients of the \ELL \wrt the hyper-parameters and inducing points at this location $\X{}{n}$ at the last layer are given by:
\begin{align}
    \frac{\partial \ELL_n}{\partial \nupos{}{}} &= \frac{1}{S}\mysum{s=1}{S}\frac{1}{\sigma^2}(y^n-\qfm{n,s}{L})\braT{\frac{\partial \braT{\K{\fpos{n,s}{L-1}}{\Zsamples{}}\Kzzinv{}}}{\partial \nupos{}{T}}}\varmr{}{} -\frac{1}{2\sigma^2}\frac{\partial\diag(\varS{}{})_i}{\partial  \nupos{}{}} = \veczero\\
    \frac{\partial \ELL_n}{\partial \vvec{\Zsamples{}{}}} &= \frac{1}{S}\mysum{s=1}{S}\frac{1}{\sigma^2} (y^n-\qfm{n,s}{L})\braT{\frac{\partial \braT{\K{\fpos{n,s}{L-1}}{\Zsamples{}}\Kzzinv{}}}{\partial \vvecT{\Zsamples{}{}}}}\varmr{}{}
    -\frac{1}{2\sigma^2}\frac{\partial\diag(\varS{}{})_i}{\partial  \vvec{\Zsamples{}{}}} = \veczero
\end{align}
where we have omitted the $L$ superscript from some of the variables for clarity. Note that by the chain rule, the gradients at each layer will all be multiplied by $\varmr{L}{}$. Thus, such a low \ELL term can only be increased by moving $\varm{}{},\varS{}{}$ and $\sigma^2$ to compensate and explain the data, since they are the only gradients taking a value different from zero.  More precisely, since the predictive variance is very small (due to the initial value of $\varS{}{}$)
, the noise variance  $\sigma^2$ does not influence the \ELL term for the input $\X{}{n}$ via  $-\frac{1}{2\sigma^2}\mathbb{VAR}(f_n^L)$. Thus, the \ELL is maximized by setting this parameter to $\nicefrac{1}{N}\left(\mid\mid\Ysamples{} - \qfm{}{L} \mid\mid^2_2\right)$. From this point, making the variance greater will necessarily make the \ELL small, since $\lim_{\sigma \to \infty}-\log \sigma \sqrt{2\pi}= -\infty$. On the other hand, for an initial small $\sigma^2$ at initialization, gradient ascent pushes $\varS{}{}$ to stay at low values. This is so because the gradient \wrt $\varS{}{}$ at the $i-th$ inducing location is $-\nicefrac{1}{2\sigma^2}$ and zero at other point. Thus, it is expected to be large, and because $\varS{}{}$ cannot be negative, this keeps $\varS{}{}$ to small values. This is to be expected since maximizing the \ELL \wrt $\varS{}{}$ is achieved by a small trace term, which is achieved by keeping the diagonal terms in $\varS{}{}$ small.
Thus, the only way we can make the \ELL higher is by making the predictive mean to be closer to the observed targets, which necessarily implies moving $\varm{}{}$ away from its initial value. 
Therefore, it is expected to be more efficient to update $\varm{}{}$ to improve the \ELL term. Once $\varm{}{}$ is different from $\veczero$, the kernel hyper-parameters and the inducing points start being optimized. This avoids having big variational variances $\varS{}{}$ (note that the \ELL forces $\varS{}{}$ to be small) or big noise values $\sigma^2$ during the first optimization iterations, reducing the possibility of explaining everything as noise and, consequently, the posterior collapse problem.

In contrast, for the \ZERO mean prior \DGP, at initialization, we have a constant predictive variance. This implies that there are no predictions that are in greater disagreement with the observed data than others. Thus, since the variational mean $\varm{}{}$ and the variational covariance $\varS{}{}$ in the last layer are initialized close to the prior mean and covariances, it is easier to keep them to these values.  Moreover, in this case, the predictive variance will just depend on values coming from the previous layer, which are close to $\veczero$ for each input (see the previous section). Thus, making the term $\frac{1}{2\sigma^2}\mathbb{VAR}(\fpos{n}{L})$ small is more easily done by increasing the noise variance than by moving the inducing points or the kernel parameters so that $\Kxz{}\approx \veczero$,  $ \Kxx{} \approx \veczero$ is satisfied.  This results in a low \KLD term, and the algorithm is expected to maximize the \ELL simply by increasing the noise parameter, indicating that everything is noise. That will maximize the \ELL term for all points without incurring a strong disagreement in terms of the \KLD. 
The consequence is that it is more likely to have a posterior collapse. From this analysis, we can understand why more inducing points or a small output scale parameter $\sigma_o$ are likely to induce less collapse. These design choices contribute to making the trace in the \ELL small, forcing the model to move $\varm{}{}$ from its initial value. 
\subsubsection{Whitened Parameterization}
\label{sec:whitened_param}

\begin{figure}[!t]
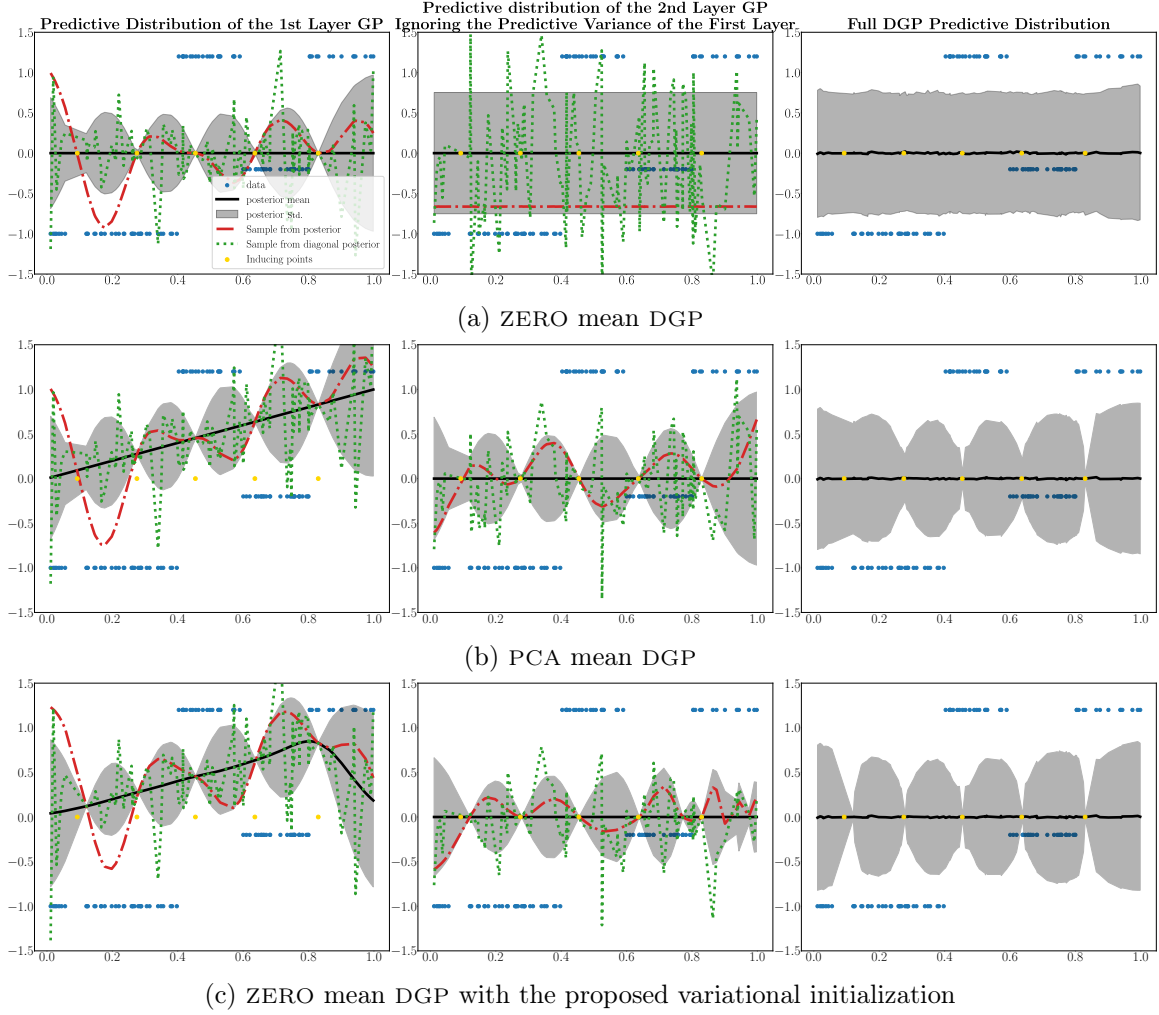

    \centering
    \begin{subfigure}{\textwidth}
        \centering
        \includegraphics[width=\textwidth]{imgs_jmlr/4-initialization_issues/S_out_e5_S_inner_e5/marginal_var_pos_at_init_model_zero_S_inner_1e-05_S_out_1e-05_num_Z_5_ls_0.1_whiten_True.pdf}
        \caption{\ZERO mean \DGP{}}
    \end{subfigure}
    \begin{subfigure}{\textwidth}
        \centering
        \includegraphics[width=\textwidth]{imgs_jmlr/4-initialization_issues/S_out_e5_S_inner_e5/marginal_var_pos_at_init_model_PCA_S_inner_1e-05_S_out_1e-05_num_Z_5_ls_0.1_whiten_True.pdf} 
        \caption{\PCA mean \DGP{}}
    \end{subfigure}
    \begin{subfigure}{\textwidth}
        \centering
        \includegraphics[width=\textwidth]{imgs_jmlr/4-initialization_issues/S_out_e5_S_inner_e5/marginal_var_pos_at_init_model_zero_m0_S_inner_1e-05_S_out_1e-05_num_Z_5_ls_0.1_whiten_True.pdf} 
        \caption{\ZERO mean \DGP{} with the proposed variational initialization}
    \end{subfigure}
    \caption{Same predictive distributions as the ones displayed in \fig \ref{fig:DSVI_zero_pca_pos_toy_at_init_NWR}, but when the the whitened parameterization is used.}
    \label{fig:DSVI_zero_pca_pos_toy_at_init_W}
\end{figure}

So far, we have analyzed the initial predictive variance of a non-whitened reparameterized \DGP implemented in \GPFLOW. In this section, we show that the posterior collapse problem may also appear under the whitened parameterization, for similar reasons. Specifically, \fig \ref{fig:DSVI_zero_pca_pos_toy_at_init_W} shows the equivalent predictive distributions to those displayed \fig \ref{fig:DSVI_zero_pca_pos_toy_at_init_NWR}, for different \DGP{}s, using the whitened parameterization. All variational parameters and hyper-parameters are initialized as before. We observe that the initial predictive distributions resulting from both parameterizations look very similar. Therefore, a similar behavior should be expected under the whitened parameterization. Namely, the \ZERO prior mean \DGP is more likely to have a posterior collapse problem. We also observe how the proposed initialization works with this parameterization as well. In general, however, the predictive distribution under non-whitened and whitened parameterizations will not always be that similar. Initialization of the output layer variational covariances to $\matI$ reveals some differences, as illustrated later on.

The results obtained, however, show that the origin of the problem is independent of the parameterization, and that the \ZERO prior mean \DGP is consistently more likely to have a posterior collapse problem. This analysis also motivates a set of ablation studies carried out in the experimental section and provides guidance for more principled initializations of both variational parameters and model hyper-parameters.

\subsubsection{Initializing $\varS{}{}=\varSw{}{}=\matI$ at the Output Layer}
\label{sec:Iat_output}

First, we consider the non-whitened case. \fig~\ref{fig:DSVI_zero_pca_pos_toy_at_init_Sout_I_Sinner1e-5_unwhit} shows the predictive distributions that are obtained in this case.
This figure illustrates that in the output \GP the predictive variance at the inducing points, when ignoring the prediction uncertainty from the inner layer, 
is now larger and equal to $1$, as expected from $\varS{}{}=\matI$ (middle column).  We observe, however, that in the \PCA prior mean \DGP, the variance is slightly smaller 
than the prior variance when the input lies between two inducing points. This seems counterintuitive, as we could expect that far from the inducing points the uncertainty is close to that of the prior, \ie, $\Kxx{}{}$, with possibly added uncertainty coming from $\varS{}{}$. Nevertheless, it is a consequence of the variational posterior $\varS{}{}$ providing more uncertainty than what the prior kernel and the location of the inducing points reduce. Specifically, we initialize $\varS{}{}=\matI$, while the prior covariances are given by $\Kzz{}$. Therefore, 
\begin{align}
    \qfS{}{} &= \underbrace{\Kxx}_{(1)} - \underbrace{\Kxz{}\Kzzinv{}\Kzx{}}_{(2)} + \underbrace{\Kxz{}\Kzzinv{}\varS{}{}\Kzzinv{}\Kzx{}}_{(3)}\,,\nonumber\\
    &= \Kxx - \Kxz{}\Kzzinv{}\bra{\Kzz{}-\varS{}{}}\Kzzinv{}\Kzx{}\,, \label{equ:NW_posterior_variance}
\end{align}
may lead to a smaller predictive variance than the prior, \ie, $\Kxx$. To see so, the predictive variance has three terms. The first term (1) is the prior variance. The second term (2) is how much variance is reduced by the prior distribution and the position of the inducing points (which has no uncertainty associated). The third term (3) reintroduces uncertainty coming
from our actual lack of knowledge about the inducing point function evaluations. Thus,
despite uncertainty arising from the inducing points (through $\varS{}{}$) and from the prior (through
$\Kxx{}{}$), the information provided by the inducing point locations together with the prior
kernel (through $\Kxz{}\Kzzinv{}\Kzx{}$) might be stronger, consequently contributing to reducing the predictive variance at points far from the inducing points. This can be formally stated through the Loewner order of positive definite matrices. Formally, whenever ($\Kzz{}-\varS{}{}$) is positive definite, then by  Loewner order $\Kzz{}\succ\varS{}{}$. Since for any point $\mathbf{v}^T=\K{\X{}{n}}{\Zsamples{}}\Kzzinv{}{}$ is a vector, then $\mathbf{v}^T\Kzz{}\mathbf{v} > \mathbf{v}^T\varS{}{}\mathbf{v}$ which implies that the uncertainty will never exceed the prior variance far from the inducing points. This will be true whenever $\diag(\varS{}{}) \leq \sigma_o$.

On the other hand, the \ZERO prior mean \DGP does not show this behavior, as in the first layer each input is mapped to a vector of zeros.
Finally, independently of the prior mean used, the full \DGP predictive variance also increases, as expected, since now $\varS{}{}=\matI$ at the output layer. 
The third column in \fig~\ref{fig:DSVI_zero_pca_pos_toy_at_init_Sout_I_Sinner1e-5_unwhit} shows the corresponding predictive distribution.

\begin{figure}[!t]
    \centering
    \begin{subfigure}{\textwidth}
        \centering
        \includegraphics[width=\textwidth]{imgs_jmlr/4-initialization_issues/S_out_1_S_inner_e5/marginal_var_pos_at_init_model_zero_S_inner_1e-05_S_out_1_num_Z_5_ls_0.1_whiten_False.pdf}
        \caption{\ZERO mean \DGP{}}
    \end{subfigure}
    \begin{subfigure}{\textwidth}
        \centering
        \includegraphics[width=\textwidth]{imgs_jmlr/4-initialization_issues/S_out_1_S_inner_e5/marginal_var_pos_at_init_model_PCA_S_inner_1e-05_S_out_1_num_Z_5_ls_0.1_whiten_False.pdf} 
        \caption{\PCA mean \DGP{}}
    \end{subfigure}
    \caption{\ZERO (top) and \PCA (bottom) \DGP models with the output layer variational covariance initialized to $\varS{}{}=\matI$, in the \GPFLOW's unwhitened parameterization.}
    \label{fig:DSVI_zero_pca_pos_toy_at_init_Sout_I_Sinner1e-5_unwhit}
\end{figure}

Because the full \DGP predictive variances are bigger in the output layer when we set $\varS{}{}=\matI$, one should expect that it is easier for the \DGP model to suffer from 
a posterior collapse problem. This is confirmed by our experiments later on. In addition, in this setting, the \ZERO prior 
mean \DGP is also expected to be more likely to suffer from a posterior collapse problem than the \PCA prior mean \DGP. The reason is that the first layer remains unchanged. Therefore, 
the \ZERO prior mean \DGP will still  observe the virtual dataset $\{(\overset{\sim}{\veczero{}{}},y^n)\}^N_{n=1}$ at initialization, which will make difficult
explaining $y^n$ from $\X{}{n}$.

When we consider a whitened parameterization, and we set $\varSw{}{}=\matI$, the initial predictive distributions that are obtained are very similar to those obtained in the 
unwhitened case, as illustrated by \fig \ref{fig:DSVI_zero_pca_pos_toy_at_init_Sout_1_Sinner1e-5_Whiten}. In this case, however, the predictive variance of the output \GP is 
constant and equal to $1$ in the \PCA prior mean \DGP, when the prediction uncertainty from the inner layer is ignored (bottom-middle figure). The reason is 
that now $\varSw{}{}=\matI$ equals the prior whitened covariances associated with the inducing points and, consequently, at any point, at initialization, the predictive covariances are given by those of the prior, \ie, $\Kxx{}{}$. Note that:
\begin{align*}
     \qfSw{}{} &= \Kxx{} -\Kxz{}\Kzzinv{}\Kzx{} + \Kxz{}\braT{\Lzzinv{}}\matI\Lzzinv{}\Kzx{} \\
     &= \Kxx{} -\Kxz{}\Kzzinv{}\Kzx{} + \Kxz{}\Kzzinv{}\Kzx{} \\
     &= \Kxx{}
\end{align*}
Note that once we start optimizing the variational parameters, and $\varSw{}{}$ gets a different value from $\matI$, the same wavy shape appears, as we illustrate in \fig \ref{fig:DSVI_zero_pca_pos_toy_at_init_Sout_1.30.8_Sinner1e-5} in the appendix. Thus, the full \DGP predictive distributions look very similar 
to the non-whitened case in this setting, and one should expect a similar behavior of each \DGP with respect to the posterior collapse problem.

\begin{figure}[!t]
    \centering
    \begin{subfigure}{\textwidth}
        \centering
        \includegraphics[width=\textwidth]{imgs_jmlr/4-initialization_issues/S_out_1_S_inner_e5/marginal_var_pos_at_init_model_zero_S_inner_1e-05_S_out_1_num_Z_5_ls_0.1_whiten_True.pdf}
        \caption{\ZERO mean \DGP{}}
    \end{subfigure}
    \begin{subfigure}{\textwidth}
        \centering
        \includegraphics[width=\textwidth]{imgs_jmlr/4-initialization_issues/S_out_1_S_inner_e5/marginal_var_pos_at_init_model_PCA_S_inner_1e-05_S_out_1_num_Z_5_ls_0.1_whiten_True.pdf} 
        \caption{\PCA mean \DGP{}}
    \end{subfigure}
    \caption{\ZERO (top) and \PCA (bottom) \DGP models with the output layer variational covariance initialized to $\varSw{}{}=\matI$, in the whitened parameterization.}
\label{fig:DSVI_zero_pca_pos_toy_at_init_Sout_1_Sinner1e-5_Whiten}
\end{figure}

One may ask what may happen when we set $\varS{}{}=\varSw{}{}=\matI$ also at the inner layer. In this case, we should expect an increment in the initial predictive 
variance at that layer, for each parameterization, see \fig \ref{fig:varying_inducing_points_Sinner1_Sout1e-5} in the appendix. This bigger variance will result in an even bigger initial predictive variance at the output layer. Therefore, such an initialization is expected to lead to a higher probability of having a posterior collapse problem, and is hence not recommended. This initialization provides samples in the inner layers, for the \ZERO prior mean \DGP model, which might be more different from $\veczero$ than when $\varS{}{}=10^{-5}\matI$. Thus, at first, it might seem to help to alleviate the posterior collapse problem since now it is less likely that the output layer receives as input a dataset of virtual pairs $\{(\overset{\sim}{\veczero{}{}},y^n)\}^N_{n=1}$. However, in general, we have observed that this additional noise in the inner layer during training usually leads to sub-optimal models. In our experiments, we show that these instabilities, with independence of the parameterization, might lead to posterior collapse. More precisely, in the experimental section, we show that a very deep \PCA prior mean \DGP with the whitened parameterization can lead to posterior collapse when extra noise is introduced 
in the inner layers, and that reducing this noise alleviates the problem.

We note that some works also consider random initialization of $\varS{}{}$ \citep{MultiResDGP}. 
In this setting, we shall expect a similar behavior to the one already described, for both the 
non-whitened and the whitened parameterizations. Namely, that the \ZERO prior mean \DGP is more
likely to suffer from a posterior collapse problem. The reason for this will be again the difficulty 
to model the dependence of $y^n$ with respect to $\X{}{n}$, at initialization, as a consequence of the 
first layer predicting values close to $\veczero{}{}$ for any input. However, one should be careful when using a random initialization of $\varS{}{}$, 
since that may result in inconsistent variational covariances that make the predictive variance go beyond the prior, as illustrated by \fig \ref{fig:DSVI_zero_pca_pos_toy_at_init_Sout_-I_Sinner1e-5_unwhit} in the appendix. This happens when random $\varS{}{}$ results in $\Kzz{}-\varS{}{}$ not being positive definite.

From this section, we conclude that spherical initializations $\varS{}{}=\varSw{}{}=\alpha\matI$ with small $\alpha\leq \sigma_o$, \eg, $10^{-5}$, yields an initialization that remains conceptually well-motivated, keeps the model within a plausible region of the parameter space, and avoids convergence issues associated with pathological starting points. This is because $\Kzz{}-\varS{}{}$ results in positive definite matrices in both parameterizations, resulting in a reduction of predictive variance beyond the prior. The suitability of this initialization is confirmed later on in the experimental section.
\subsubsection{Impact of the Kernel Hyper-parameters}
\label{sec:impact_hypers}

The impact of the kernel hyper-parameters in the initial predictive distribution can be easily understood. 
In the \RBF kernel, the output scale parameter, $\sigma_o$, is directly related to the diagonal of $\Kxx{}$, which is
the initial prior variance. Reducing $\sigma_o$ is expected to reduce the initial predictive variance when we move 
away from the inducing points.  At the inducing points, however, the predictive variance will still be given by $\varS{}{}$.
The length-scale, on the other hand, affects the relative distance between the points, 
making the prior covariances $\Kxx{}$ more or less isotropic. It also has an impact on how quickly the predictive mean and 
variance go back to those of the prior as we move away from the inducing points. This will be the case for both the whitened 
and non-whitened parameterizations. Therefore, changing the output scale or the length-scale will have a small impact on 
the posterior collapse problem of the \ZERO prior mean \DGP. More precisely, the influence of $\X{}{n}$ on the \DGP predictive distribution
will still be rather small, making the posterior collapse likely to occur. By contrast, in the \PCA prior mean \DGP
there will still be a significant dependence of the predictive distribution of the output \GP \wrt $\X{}{n}$, reducing
the possibility of a posterior collapse. However, if posterior collapse appears, it might be beneficial to reduce $\sigma_o$. This will reduce 
the variance in the inner layers at locations far from the inducing points, 
reducing overall noise in the optimization process. At the output layer, it will also reduce the predictive variance, and thus the 
trace term of the \ELL, which results in a larger \ELL.

Summing up, changing the kernel hyper-parameters will affect mostly the predictive distribution in regions 
of the input space far from the inducing points or, similarly, far from the observed data, which is expected to 
have a smaller impact on the evaluation of the \ELL term during the first iterations of the optimization problem, as compared with the initial choice of the variational parameters.

\subsubsection{The Effect of More Inducing Points}
\label{sec:opt_analysis_more_inducing}
\begin{figure}[!t]
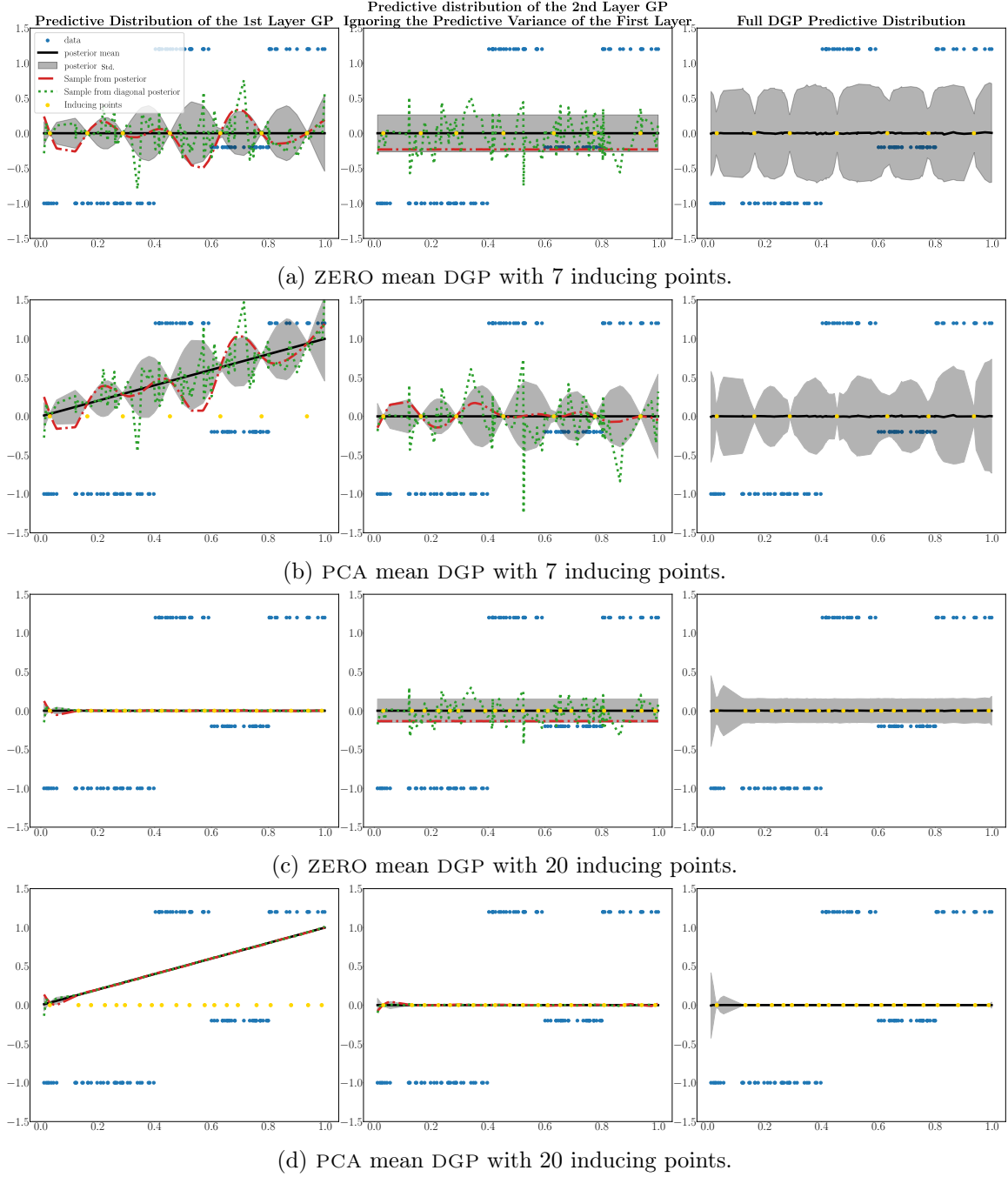

    \centering
    \begin{subfigure}{\textwidth}
        \centering
        \includegraphics[width=\textwidth]{imgs_jmlr/4-initialization_issues/S_out_e5_S_inner_e5_more_inducing/marginal_var_pos_at_init_model_zero_S_inner_1e-05_S_out_1e-05_num_Z_7_ls_0.1_whiten_False.pdf}
        \caption{\ZERO mean \DGP{} with $7$ inducing points.}
    \end{subfigure}
    \begin{subfigure}{\textwidth}
        \centering
        \includegraphics[width=\textwidth]{imgs_jmlr/4-initialization_issues/S_out_e5_S_inner_e5_more_inducing/marginal_var_pos_at_init_model_PCA_S_inner_1e-05_S_out_1e-05_num_Z_7_ls_0.1_whiten_False.pdf}
        \caption{\PCA mean \DGP{} with $7$ inducing points.}
    \end{subfigure}
        \begin{subfigure}{\textwidth}
        \centering
        \includegraphics[width=\textwidth]{imgs_jmlr/4-initialization_issues/S_out_e5_S_inner_e5_more_inducing/marginal_var_pos_at_init_model_zero_S_inner_1e-05_S_out_1e-05_num_Z_20_ls_0.1_whiten_False.pdf}
        \caption{\ZERO mean \DGP{} with $20$ inducing points.}
    \end{subfigure}
    \begin{subfigure}{\textwidth}
        \centering
        \includegraphics[width=\textwidth]{imgs_jmlr/4-initialization_issues/S_out_e5_S_inner_e5_more_inducing/marginal_var_pos_at_init_model_PCA_S_inner_1e-05_S_out_1e-05_num_Z_20_ls_0.1_whiten_False.pdf}
        \caption{\PCA mean \DGP{} with $20$ inducing points.}
    \end{subfigure}
    \caption{Initial predictive distributions of a \ZERO prior mean and a \PCA prior mean \DGP{} with $7$ and $20$ inducing points using the \GPFLOW non-whitened parameterization.}
    \label{fig:varying_inducing_points_Sinner1e5_Sout1e5}
\end{figure}

We now study the effect of increasing the number of inducing points when $\varS{}{}=10^{-5}\matI$ in all layers. 
\fig~\ref{fig:varying_inducing_points_Sinner1e5_Sout1e5} displays the initial predictive distributions of the \ZERO prior 
mean \DGP and the \PCA prior mean \DGP when $7$ and $20$ inducing points are considered, instead of $5$. We only consider here 
the non-whitened parameterization.
We observe that when a large number of inducing points are used, the predictive variance becomes closer to zero, which 
is the posterior variance specified by $\varS{}{}$ associated to the inducing points. 
In this setting, we can expect the \ELL term to be very low, because of the small variance. As we saw in \usec~\ref{sec:impact_on_ell}, this favors not making the noise variance $\sigma^2$ large in order to explain the observed data, but updating the variational mean parameter. Note that the more inducing points, the smaller will be the trace term in the \ELL if $\varS{}{}$ is initialized with a small $\alpha$. This is confirmed by our coordinate update approach in the next section.

For completeness, appendix \ref{sec:app:c:1:opt:difficulties} provides figures with more inducing points and other initializations for $\varS{}{}$ in the inner and output layers. In those figures, we observe that with more inducing points, the same behaviors we have been analyzing are observed. For example, $\varS{}{}$ with smaller $\alpha$ is also preferable. We also observe a wiggling shape appearing in the \ZERO \DGP{}. This is because with more inducing points, it is likely that function evaluations around $\veczero$ are closer to the inducing points, making  $\Kxz{} \neq \veczero$. In consequence, it is expected that more inducing points help in avoiding the posterior collapse problem in the \ZERO \DGP{}, which we confirm in the experimental section.

Therefore, increasing the number of inducing points should favor avoiding the collapse problem. Moreover, it is also expected to make the 
dependence of the predictive variance on the output scale parameter $\sigma_o$ less relevant since now the predictive variance is likely 
to depend mostly on $\varS{}{}$. We confirm these arguments in the experimental section.
\subsubsection{The Standard Non-whitened Parameterization}

\begin{figure}[!t]
    \centering
    \includegraphics[width=\linewidth]{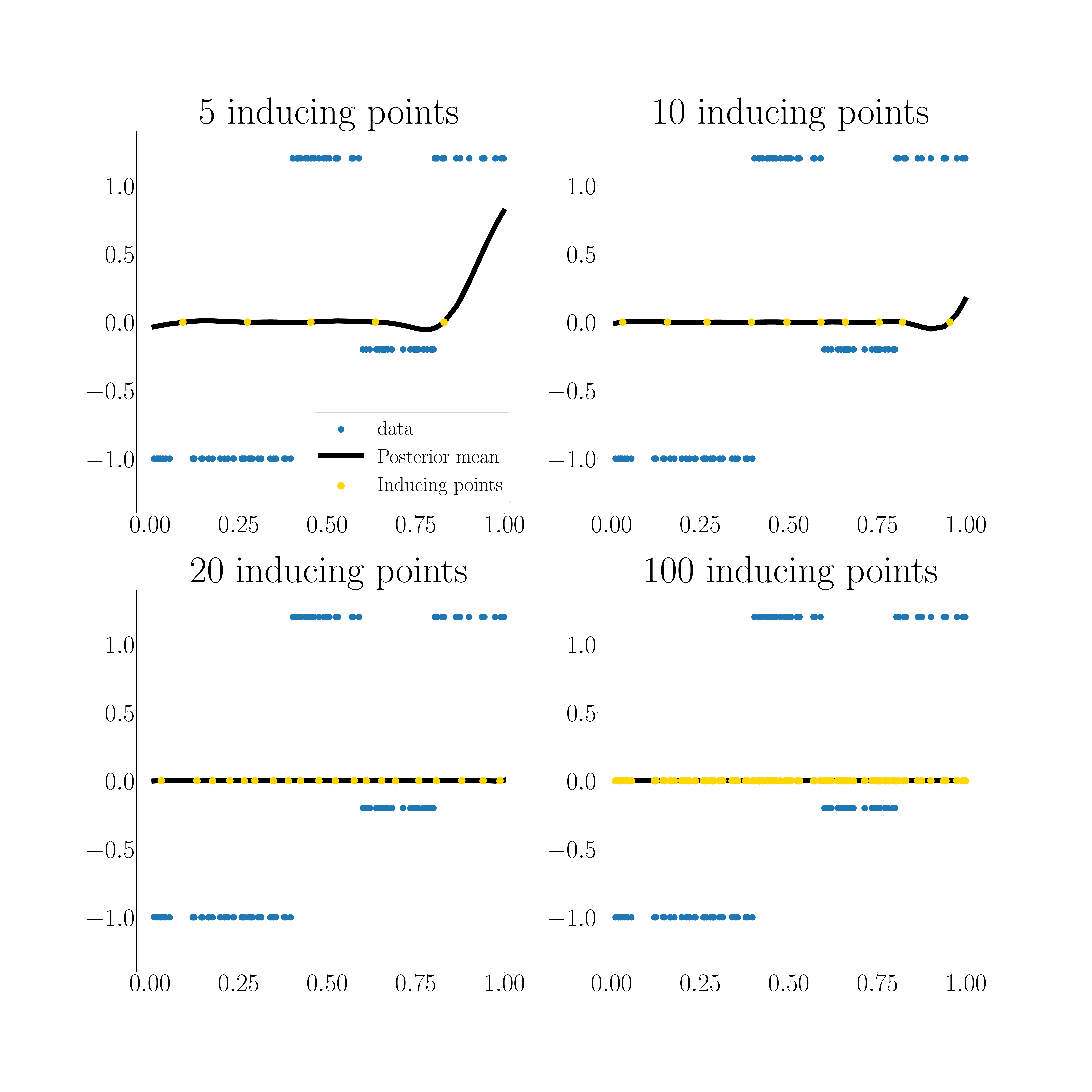}
    \caption{Predictive mean of the inner layer \GP of a \DGP with the standard non-whitened parameterization for different numbers of inducing points and the \PCA prior mean function.}
    \label{fig:posterior_mean_non_whitened}
\end{figure}

So far, we have analyzed what happens under the non-whitened parameterization by \GPFLOW, in which the tuned parameter of the posterior approximation
is the posterior mean minus the prior mean (see \usec \ref{usec:non_whit_in_practice}).  However, as we mentioned earlier, both \GPYTORCH and \GPJAX implement the standard non-whitened parameterization. 
For a \PCA prior mean \DGP, this parameterization results in the following predictive mean at initialization:

\begin{align}
	\qfm{}{} & = \Xsamples + \Kxz{}\Kzzinv{}(\veczero - \Zsamples{})\,.
	\label{eq:standard_non_whitened}
\end{align}
Thus, in the first layer \GP of the \DGP we will not observe a straight increasing line, as we have seen previously (see \fig \ref{fig:DSVI_zero_pca_pos_toy_at_init_NWR} middle row). In particular, right at the inducing points, the predictive mean will be $\qfm{}{}=\veczero$, as indicated by \ueqn (\ref{eq:standard_non_whitened}). \fig \ref{fig:posterior_mean_non_whitened} shows the predictive mean of the inner layer \GP of a \DGP with the standard non-whitened parameterization and the \PCA prior mean, for different numbers of inducing points. We observe that, as the number of inducing points increases, the predictive mean becomes nearly constant and equal to zero. 
It is only equal to $\Xsamples$ when we are far away from the inducing points.
This indicates that under this parameterization, a \PCA prior mean \DGP will behave similarly to a \ZERO prior mean \DGP in the inner layers, if the number of inducing points is big. 
Therefore, it is expected to be more likely to suffer from the posterior collapse problem than the non-whitened parameterization of \GPFLOW.
Even though \cite{rudner2020inter} reports successfully training \DGP{}s with the standard non-whitened parameterization, our analysis shows later on that such a parameterization is of less practical interest since it results in unstable optimization.

\subsubsection{Mixing Matrices and Warping Functions}

While the original formulation of \DGP{}s considered independence between processes within a layer, we can easily introduce dependencies either using mixing matrices \citep{jankowiak2019neural} or warping functions \citep{dtgp}. Note that random initialization of these transformations might impact both the \PCA and \ZERO prior mean \DGP predictive distribution at initialization. Thus, a practitioner might need to carefully choose the initial values. For mixing matrices, it seems reasonable to assume that the processes are (at the beginning) independent by initializing the mixing matrix to the identity, and let the marginal likelihood estimate any possible dependency. Also, warping functions are usually initialized to an identity mapping \citep{tgp,etgp}. In those cases, at initialization, the \DGP models will behave in the same way as we have analyzed.

\subsection{An Approximate Explanation for the Posterior Collapse Problem}

So far, we have used an intuitive explanation to understand the source of the posterior collapse problem, under different initializations, on a toy dataset. 
We have found that the main reason for the posterior collapse problem is that, at initialization, the output \GP in the \DGP sees a virtual dataset of 
input points very similar to zero.  Even if most computations involving \DGP{}s with the \DSVI algorithm are not analytically tractable, we show in this section that we can make some approximations 
that yield tractable analytical conclusions.  More precisely,  at initialization, 
we approximate the output layer of a \DGP with an \SVGP that receives as an input either $\veczero$ or $\X{}{n}$, and then analyze the corresponding coordinate gradient 
ascent updates of the variational and noise parameters. This gives extra insights into the reasons for posterior collapse.
In this section, we focus on the standard non-whitened parameterization.

\subsubsection{Approximating at Initialization a \DGP Using a \SVGP }

While in \usec \ref{sec:impact_on_ell} we showed via the \ELL term how a \DGP is more or less likely to have a posterior collapse problem, we ignored the role of the \KLD term. To take it into account, we note that many of the initializations considered in the previous sections lead to a small predictive variance in the inner layer of a \DGP{}, making it nearly zero. This included a small variational variance $\varS{}{}=10^{-5}\matI$, a big number of inducing points, or a small output scale kernel parameter. This means that, at initialization, we will mainly propagate the predictive means $\qfm{}{}$ until the last layer. Given this, we can see that, at initialization, the output layer resembles a sparse Variational \GP \citep{pmlr-v5-titsias09a,GPs-Big-Data}. As mentioned earlier, for the \PCA prior mean, the \SVGP will be trained on $\{(\X{}{n}, y^n)\}^N_{n=1}$, while the \ZERO prior mean \DGP will be trained on $\{(\veczero,y^n)\}^N_{n=1}$. Thus, by studying the \SVGP{}'s objective at initialization and the gradient updates, we can draw conclusions on what early local minima a \DGP{} might find during optimization, revealing interesting properties.

A \SVGP{} can be trained by coordinate ascent on the variational and model parameters, where we mix exact and gradient-based updates. 
In particular, \cite{pmlr-v5-titsias09a} showed that the optimal variational updates on $\varm{}{}$ and $\varS{}{}$ are:
\begin{align}
	\varS{\text{opt}}{}  & \leftarrow  \Kzz{}{}\braInv{\Kzz{} + \frac{1}{\sigma^2}\Kzx{}\Kxz{}}\Kzz{}{}\,,\label{eq:update_S}\\
	\varm{\text{opt}}{}  & \leftarrow  \frac{1}{\sigma^2}\varS{\text{opt}}{}\Kzzinv{}\Kzx{}\mathbf{y}\,,\label{eq:update_m}
\end{align}
for the non-whitened parameterization.
Furthermore,  the optimal noise parameter $\sigma^2$ is:
\begin{align}
	\sigma_\text{opt}^2 & \leftarrow \frac{1}{N}\pare{\underbrace{\mid\mid\Ysamples{} - \qfm{}{} \mid\mid^2_2}_{\text{aleatoric}} + \underbrace{\tr{\qfS{}{}}}_{\substack{\text{aleatoric + } \\ \text{approximation error}}}}\,,
	\label{eq:update_noise}
\end{align}
where $\qfm{}{}$ and $\qfS{}{}$ are the predictive mean and the predictive covariance, respectively.
Appendix \ref{sec:app:A} provides the derivation. The coordinate update on $\sigma^2$ is the same for the whitened and non-whitened parameterizations. We just need to plug 
in the corresponding predictive mean and covariance.

The coordinate update on the noise in \ueqn (\ref{eq:update_noise}) reveals that optimality is achieved as the expected value of the squared distances between the targets and the expected mean (the sample variance), plus another term which has an aleatoric component, plus some uncertainty that comes from representing the problem through the inducing points. Note that attending to \ueqn (\ref{eq:ell_expansion}), the optimal \ELL is achieved when this trace term is $\veczero$. This is only achieved when $\varS{}{}=\veczero$ and $\Xsamples{}{}=\Zsamples{}$. Attending to the coordinate update for $\varS{}{}$, this term is zero when either $\sigma^2 \rightarrow \veczero$ or the kernel scale parameter $\sigma_o$ is zero. $\sigma^2$ contributes as a source of aleatoric uncertainty, while $\sigma_o$ is only driven to zero by the marginal likelihood if the data comes from a constant function. Thus, this is a source of aleatoric uncertainty. When $\varS{}{}$ is zero, we require $\Xsamples{}=\Zsamples{}$ for the trace to be zero. This is variance coming from an approximation error. It is not aleatoric nor epistemic, because it is not made zero by collecting more data, nor is noise implicit in the data to be modeled.

The immediate consequence of this observation in the \ZERO{} prior mean \DGP is that the contribution from $\tr{\qfS{}{}}$ is larger since the predictive variance is constant across training data, 
as illustrated in \fig \ref{fig:DSVI_zero_pca_pos_toy_at_init_NWR} and \fig \ref{fig:DSVI_zero_pca_pos_toy_at_init_W}. By contrast, the \PCA prior mean \DGP shrinks the predictive variance at some points. Clearly, the variational or hyper-parameters initialization will also contribute to this term, but given similar initializations, the \ZERO prior mean \DGP{} will increase the predictive variance more. Furthermore, in the limit of $\sigma^2$ being arbitrarily big, the coordinate updates on the variational parameters will result exactly in those that make the \KLD be zero, \ie $\varm{}{}=\veczero, \varS{}{}=\Kzz{}$, resulting in a posterior collapse problem. See \ueqn (\ref{eq:update_S}) and \ueqn(\ref{eq:update_m}). Thus, a higher $\sigma^2$ is expected to drive the variational parameters towards the prior.

\paragraph{Coordinate Updates for the \ZERO prior mean \DGP:}

For a \ZERO mean \DGP, we can see the output layer as a \SVGP{} that receives $\veczero$ as every input. 
Using the initial variational parameters $\varm{}{}=\veczero,\varS{}{}=\alpha\matI$, which make $\qfm{}{}=\veczero$, the coordinate update is given by:
\begin{align}
	\sigma_\text{opt}^2 & \leftarrow \frac{1}{N}\mysum{n=1}{N}(\Y{}{n})^2 +\frac{1}{N}\mysum{n=1}{N}\pare{  \K{\veczero}{\veczero} - \K{\veczero}{\Zsamples{}}\Kzzinv{}\K{\Zsamples{}}{\veczero}+\alpha\K{\veczero}{\Zsamples{}}\Kzzinv{}\Kzzinv{}\K{\Zsamples{}}{\veczero}}\,.
    \label{equ:coord_sigma_zero}
\end{align}
Note that if we condition on $\mathbf{0}$ for each training location $\X{}{n}$, the diagonals from the marginal variational posterior covariance evaluated at all pairs $\Xsamples{}$ are exactly the same, which results in $N$ equal terms being added up. As we already saw, this posterior covariance can be above or below the prior variance, depending on how much uncertainty from the prior we are able to subtract due to the inducing point locations. For $\varS{}{}=\alpha\matI$, with $\alpha\leq \sigma_o$, is below due to Lowener's order. Assuming the initial output scale kernel parameter is $\sigma_o=1$, we have:
\begin{align}
	\sigma_\text{opt}^2 & \leftarrow \left[ \frac{1}{N}\mysum{n=1}{N}(\Y{}{n})^2 \right] + 1 - \K{\veczero}{\Zsamples{}}\Kzzinv{}\K{\Zsamples{}}{\veczero}+\alpha\K{\veczero}{\Zsamples{}}\Kzzinv{}\Kzzinv{}\K{\Zsamples{}}{\veczero}\,,
\end{align}
which implies that initializing with smaller $\alpha$ values should favor smaller noise variances, reducing the posterior collapse problem,
a similar conclusion as the one from \usec~\ref{sec:Iat_output}. 

\paragraph{Coordinate update for the \PCA prior mean \DGP:}

For the \PCA prior mean \DGP model, the coordinate update is different, since we now observe $\X{}{n}$ as an input. Thus, we have
that when $\varm{}{}=\veczero,\varS{}{}=\alpha\matI$
\begin{align}
	\sigma_\text{opt}^2 & \leftarrow \frac{1}{N}\mysum{n=1}{N}(\Y{}{n})^2 +\frac{1}{N}\mysum{n=1}{N}\pare{  \K{\X{}{n}}{\X{}{n}} - \K{\X{}{n}}{\Zsamples{}}\Kzzinv{}\K{\Zsamples{}}{\X{}{n}}+\alpha\K{\X{}{n}}{\Zsamples{}}\Kzzinv{}\Kzzinv{}\K{\Zsamples{}}{\X{}{n}}}\,,
    \label{equ:coord_sigma_pca}
\end{align}
which for $\K{\X{}{n}}{\X{}{n}}=1$ we have:
\begin{align}
	\sigma_\text{opt}^2 & \leftarrow \left[\frac{1}{N}\mysum{n=1}{N}(\Y{}{n})^2\right] + 1 - \left[\frac{1}{N}\mysum{n=1}{N}\pare{  \K{\X{}{n}}{\Zsamples{}}\Kzzinv{}\K{\Zsamples{}}{\X{}{n}}- \alpha \K{\X{}{n}}{\Zsamples{}}\Kzzinv{}\Kzzinv{}\K{\Zsamples{}}{\X{}{n}}}   \right]\,. 
\end{align}
The difference with respect to the \ZERO prior mean \DGP is clear. Here, the resulting noise variance parameter does not receive a constant contribution but an average value at each training point. Recall that some of them will have a smaller predictive variance when $\alpha=10^{-5}$, due to the input points being closer to the inducing points.  The value of $\alpha$ will have a similar effect as the one described previously. Namely, a larger $\alpha$ will increase the noise variance and is expected to make the posterior collapse problem more likely. However, as we observed previously, the \PCA prior mean \DGP with $\alpha=1$ slightly shrinks the predictive variance compared to the \ZERO model, and so it is expected to suffer less from posterior collapse under this initialization as well.

In conclusion, the  \PCA prior mean \DGP is expected to result in a smaller noise variance estimate than the \ZERO prior mean \DGP and to reduce the posterior collapse problem, when the same initialization is used for both models.
Our results also highlight that a larger number of inducing points is more likely to make the estimate of $\sigma^2$ smaller, even more so when $\alpha=10^{-5}$, and to reduce the likelihood of a posterior collapse problem.
More precisely, more inducing points will result in smaller predictive variances for each $\X{}{n}$. This analysis reveals that an initial small variance kernel output scale parameter $\sigma_o$ also favors small values for $\sigma^2$. Only in the particular case of the inducing points being close to $\veczero$, the variance
of the \ZERO \DGP will be smaller. However, the initialization of inducing points very often tends to cover the whole training domain and, hence, this pathological case is highly unlikely in
practice.

\paragraph{Coordinate updates for the variational parameters: } We have shown that the \ZERO prior mean \DGP is expected to consistently learn higher variances than the \PCA prior mean \DGP because, as we mentioned, the noise variance parameter receives a constant contribution rather than an average value at each training point. However, the resulting noise variances will not be the only cause of the posterior collapse. By looking at the coordinate updates for the variational parameters in \ueqn (\ref{eq:update_S}) and \ueqn (\ref{eq:update_m}), we observe that $\varS{\text{opt}}{}$ will tend to $\Kzz{}$ also when $\Kzx{}\Kxz{}$ is a matrix of zeros, with independence of the value of the noise variance $\sigma^2$. A careful look at how this matrix is computed reveals that it will be closer to a matrix of zeros for the \ZERO prior mean \DGP{}. Specifically, each position in the matrix is given by:
\begin{align}
	\pare{\Kzx{}\Kxz{}}_{ij} & = \mysum{n=1}{N}K(\Z{i},\X{}{n})K(\X{}{n},\Z{j})\,.
\end{align}
Unless an inducing point is close to $\veczero$ (unlikely for a small number of inducing points), most of the entries of the matrix will be close to zero for the \ZERO prior mean \DGP, for a small enough length-scale. This will make the  \ZERO prior mean \DGP{} more likely to collapse the variational covariance to the prior $\Kzz{}$. 
Consider now that the optimal value for $\varS{}{}$ is approximately$\Kzz{}{}$, the optimal update for the variational mean is:
\begin{align}
	\varm{\text{opt}}{}  & \leftarrow  \frac{1}{\sigma^2}\Kzx{}\Ysamples\,.
\end{align}
Again, given that $\Kzx{}$ takes smaller values for the \ZERO prior mean \DGP and, in conjunction with $\sigma^2$ taking higher values, 
the variational mean will also tend to the prior, \ie, $\varm{}{}=\veczero$, more likely for the \ZERO prior mean \DGP than for the \PCA prior mean \DGP. 
When this collapse happens, the coordinate updates on the noise variance parameter are given by:
\begin{align}
     \sigma_\text{opt}^2 &\leftarrow \sigma_o + \frac{1}{N}\mysum{n=1}{N}(\Y{}{n})^2 
     \label{equ:max_noise_var}
\end{align}
for each model (when $\K{\X{}{n}}{\X{}{n}}=\K{\veczero}{\veczero}=\sigma_o$). See \ueqns (\ref{equ:coord_sigma_zero}),(\ref{equ:coord_sigma_pca}). Thus, both models learn the same noise variance in this scenario.

These results also show why a small number of inducing points is more likely to result in a posterior collapse. With fewer inducing points, $\Kxz{}$ is more likely to be close to $\veczero$ for many training points. Common sense also supports this reasoning. With very few inducing points, most of the targets will have a high predictive variance given by the prior (\ie a bigger trace term), and the model will tend to learn everything as noise. Overall, coordinate updates on the variational parameters also support why the \ZERO \DGP{} is more likely to suffer from posterior collapse.
\subsubsection{Simulating the Coordinate Updates}
\label{subsec:simul:coord:updates}

We have shown that coordinate updates make the \ZERO prior mean \DGP{} learn, in general, higher noise variances than the \PCA prior mean \DGP{}. This, in conjunction with the coordinate update on the variational parameters, where $\Kxz{}{}\rightarrow \veczero$ in some settings of the \ZERO prior mean \DGP{}, indicates that the model is more likely to collapse. However, showing that this actually leads to a posterior collapse after some optimization steps can only be done experimentally because: 1) the limit $\sigma\rightarrow\infty$ is never achieved since it results in $\ELL \rightarrow -\infty$, 2) achieving $\Kxz{}=\veczero$ depends on the kernel parameters, number of inducing points and its location, 3) analyzing the convergence of the iterative maps that result from the coordinate updates cannot be done analytically. Therefore, we perform several simulations under different conditions to validate our claims, fixing the kernel parameters and including points to the same initial values we have been using in this section.

Since the order in which coordinates are updated matters, we perform two different ordering schemes. The first ordering scheme implies updating  $\varS{}{}-\varm{}{}-\sigma^2$, and the second one uses the order $\sigma^2-\varS{}{}-\varm{}{}$. We will see that regardless of the order, the \ZERO prior mean \DGP{} consistently learns higher noise variances and collapses the variational parameters to the prior. Before proceeding with the simulation, we analyze the two ordering schemes to validate what we should expect from the simulations. 

\paragraph{Order update  $\varS{}{}-\varm{}{}-\sigma^2$:}

This order can consistently lead to smaller noise variances.
Specifically, for the initial value $\varm{}{}=\veczero$, changes in $\varm{}{}$ necessarily maximize the \ELL. This is because for $\varm{}{}=\veczero$ the \KLD in the \ELBO takes its minimum possible value \wrt this parameter. Thus, maximizing the \ELBO \wrt $\varm{}{}$ implies that the \ELL is increased more than the \KLD, after the update.

Note that maximizing the \ELL \wrt $\varm{}{}$ can only be achieved by making the term $\nicefrac{1}{2}(y^{n}-\qfm{n,s}{L})^2$ smaller. This is exactly the sample variance term appearing in the update for $\sigma^2$ in \ueqn (\ref{eq:update_noise}). Since this term is necessarily smaller than
that of $\varm{}{} = \veczero$, because we have shown that the \ELL is increased, updating first $\varm{}{}$ and then $\sigma^2$ consistently provides smaller variance estimates than updating first $\sigma^2$ and then $\varm{}{}$.
By contrast, it is not possible to guarantee that updating $\varS{}{}$ first will result in a trace term in Eq. (\ref{eq:update_noise}) that is smaller than in the previous step, and hence in smaller noise variances, unless $\varS{}{}$ is initialized to the prior. However, we can show that updating $\varS{}{}$ first results in a variance that never exceeds the prior variance. Specifically, if we replace $\varS{\text{opt}}{}$ into the predictive covariances we have:
\begin{align}
    \qfS{}{} = \Kxx{} -\Kxz{}\Kzzinv{}\Kzx{} + \Kxz{}\braInv{\Kzz{}+\frac{1}{\sigma^2}\Kzx{}\Kxz{}}\Kzx{}\,.
\end{align}
To show that any update on $\varS{}{}$ does not increase the variance beyond the prior, we need to show that  $\Kxz{}\braInv{\Kzz{}+\frac{1}{\sigma^2}\Kzx{}\Kxz{}}\Kzx{}$ is always smaller than $\Kxz{}\Kzzinv{}\Kzx{}$.  For a single point $\X{}{}$, $\mathbf{v}=\Kxz{}{}$ is a column vector. Thus, showing that $\mathbf{v}^T\Kzzinv{} \mathbf{v} \geq  \mathbf{v}^T\braInv{\Kzz{}+\frac{1}{\sigma^2}\Kzx{}\Kxz{}}\mathbf{v}$
 implies showing that $\Kzz{}+\frac{1}{\sigma^2}\Kzx{}\Kxz{} \succeq \Kzz{}{}$. Since  $\Kzz{}+\frac{1}{\sigma^2}\Kzx{}\Kxz{} - \Kzz{} = \frac{1}{\sigma^2}\Kzx{}\Kxz{}$ is positive definite\footnote{Note that $\Kzx{}\Kxz{}$ is the product of a matrix by its transpose.}, we can use Lowener's order to show that the inequality holds. A similar procedure can be followed to show that $\varS{\text{opt}}{}$ never exceeds the prior covariance $\Kzz{}$. One just needs to apply the Woodbury matrix identity.

This shows that the predictive variances will never exceed the prior variances given by $\Kxx{}$.
Thus, updating $\sigma^2$ first may consider bigger $\qfS{}{}$ than the prior variance, if $\varS{}{}$ is initialized in a pathological way
(\ie, with higher uncertainty than the prior, $\Kzz{}-\varS{}{}$ is not positive definite). Furthermore, this proof shows that $\varS{\text{opt}}{}$ in Eq. (\ref{eq:update_S}) 
will never result in a value that makes the \KLD exactly zero, \ie, $\varS{}{}=\Kzz{}$, unless $\Kxz{}=\veczero$ is zero. This is because the variance can never grow to $\infty$ and the update never exceeds the prior since $\Kzz{}\succeq\varS{\text{opt}}{}$ with equality only when $\Kxz{}=\veczero$.
In summary, this update order leads to more controlled predictive variances (which never go beyond the prior), independently of the initial variational parameters, and should be preferred. 

\paragraph{Order update  $\sigma^2-\varS{}{}-\varm{}{}$:} This order scheme is sensitive to the initial variational parameters. Attending to our previous results, for badly selected initial $\varS{}{}$, we could add variance beyond that of the prior, or be exactly the prior if $\varS{}{}=\Kzz{}$. Also, initializations  with $\varS{}{}=\alpha\mathbf{I}$ and $\alpha=1$ will provide higher variances when compared to $\alpha=10^{-5}$. Posterior collapse can happen here earlier since higher variances will contribute to making the matrix $\nicefrac{1}{\sigma^2}\Kzx{}\Kxz{}$ be closer to all zeros. As in the other order, total collapse can only happen if $\Kxz{}{}=\veczero$.

\begin{figure}[!p]
\centering
\resizebox{0.78\textwidth}{!}{%
\begin{tabular}{cc}
\raisebox{0.10\height}{\begin{subfigure}{0.48\linewidth}
    \centering
    \includegraphics[width=\linewidth]{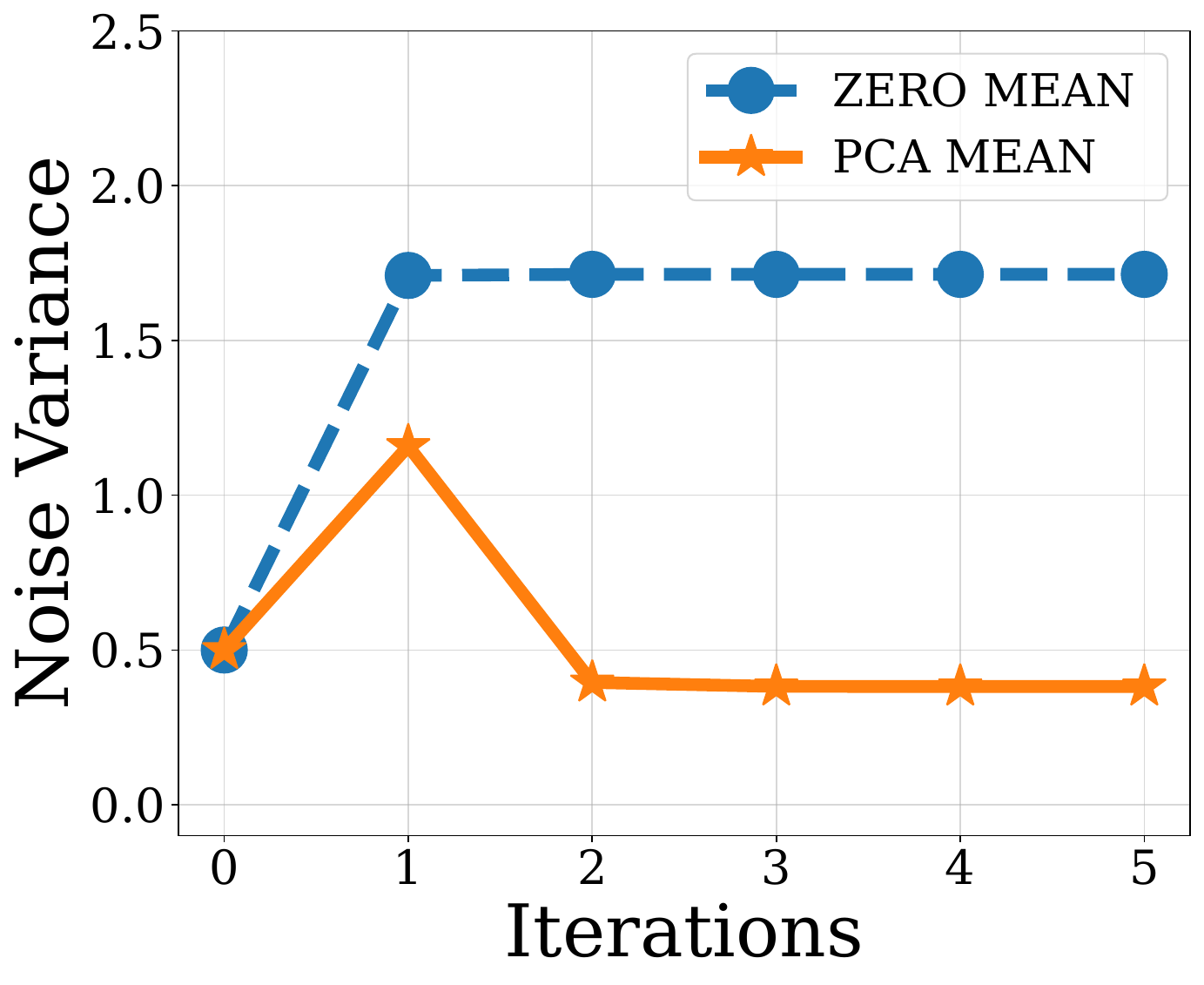}
\end{subfigure}}
&
\begin{subfigure}{0.48\linewidth}
    \centering
    \begin{subfigure}{\linewidth}
        \centering
        \includegraphics[width=\linewidth]{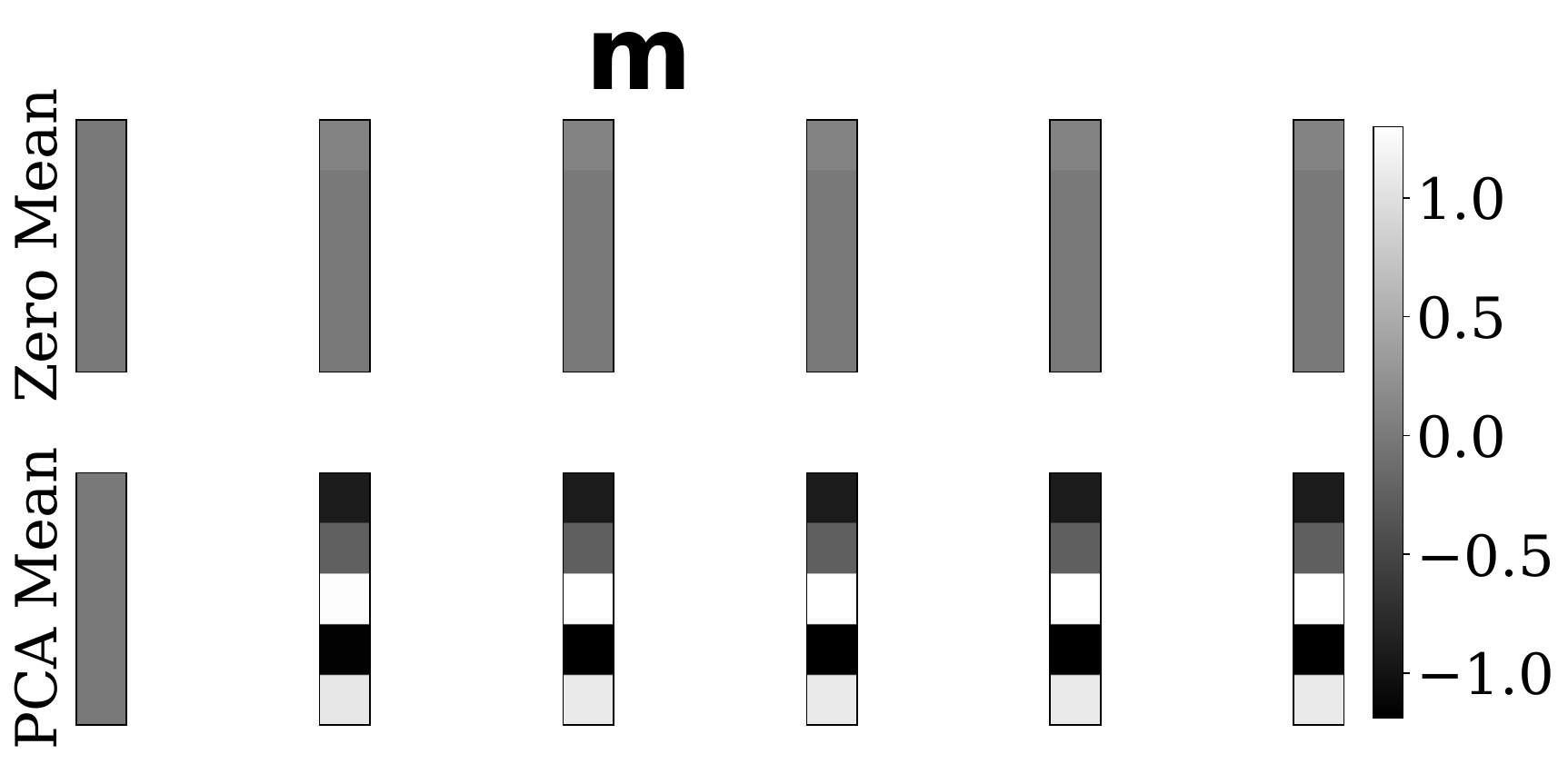}
\end{subfigure}
\begin{subfigure}{\linewidth}
        \centering
        \includegraphics[width=\linewidth]{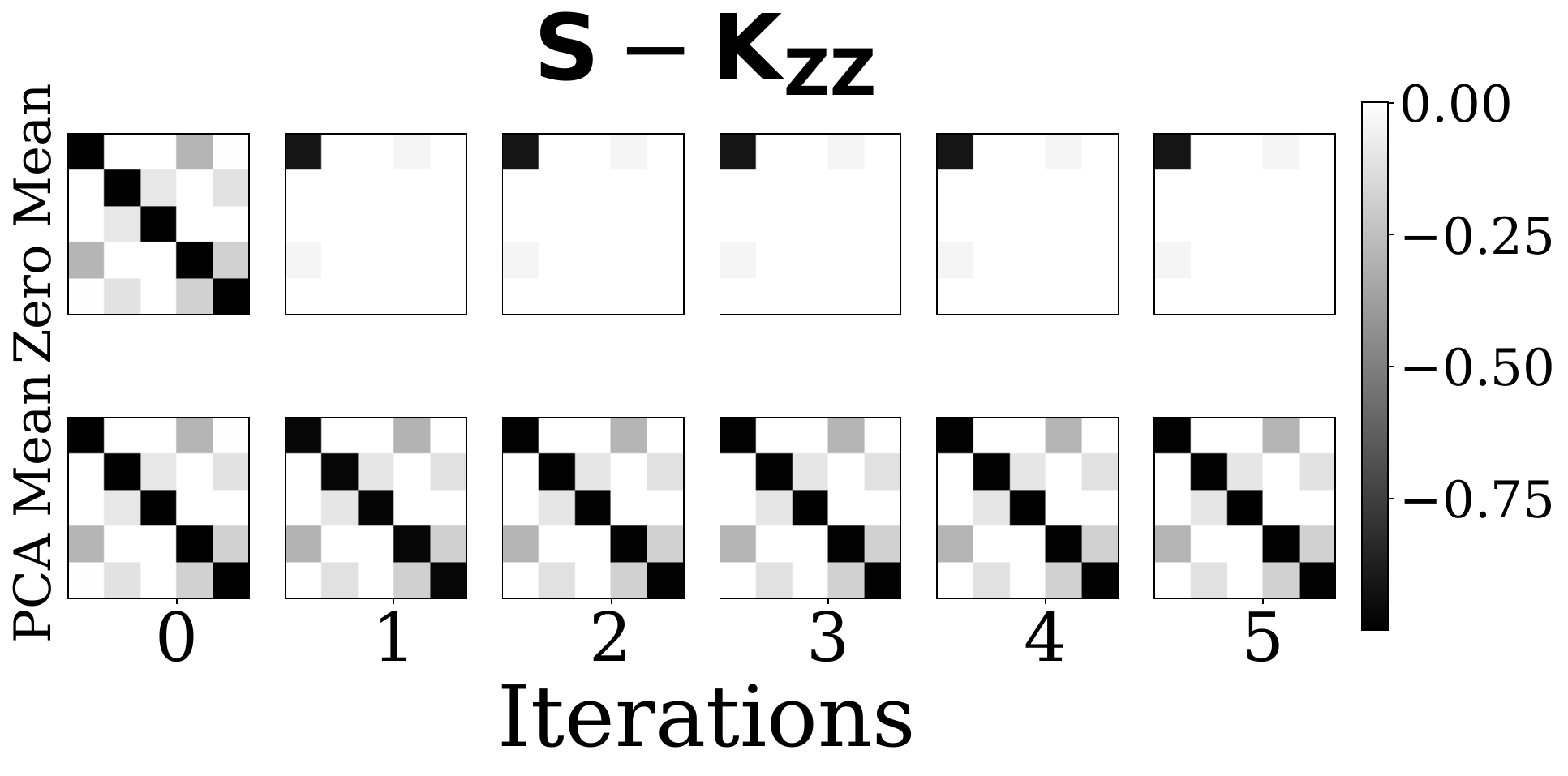}
    \end{subfigure}
\end{subfigure}\\
\multicolumn{2}{c}{
    \begin{subfigure}{\linewidth}
        \centering
        \caption{Order of updates: $\sigma^2$-$\varS{}{}$-$\varm{}{}$, with initialization $\varS{}{}=10^{-5}\matI$, $\varm{}{}=\veczero$}
    \end{subfigure}
} \\
\raisebox{0.10\height}{\begin{subfigure}{0.48\linewidth}
    \centering
    \includegraphics[width=\linewidth]{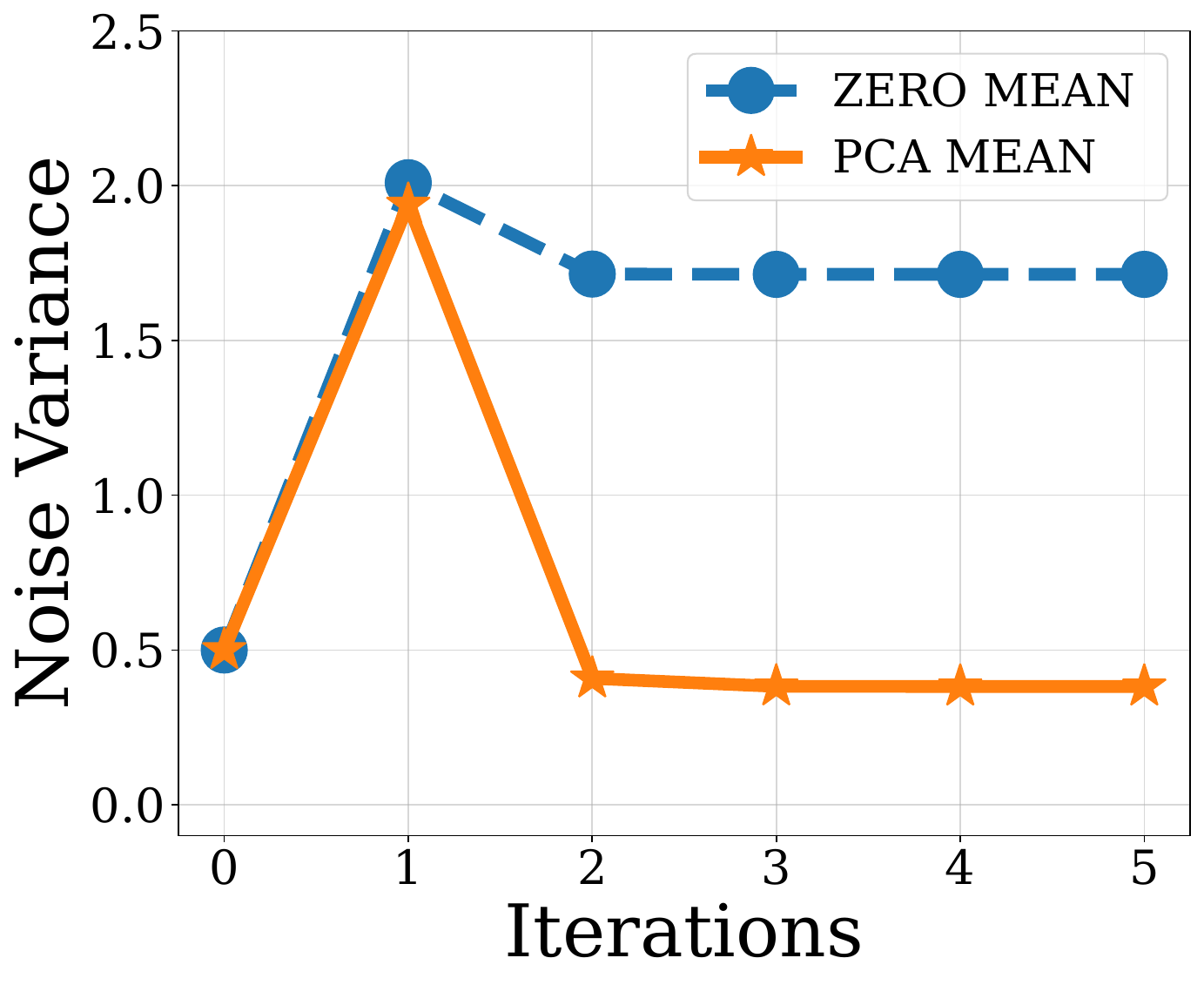}
\end{subfigure}}
&
\begin{subfigure}{0.48\linewidth}
    \centering
    \begin{subfigure}{\linewidth}
        \centering
        \includegraphics[width=\linewidth]{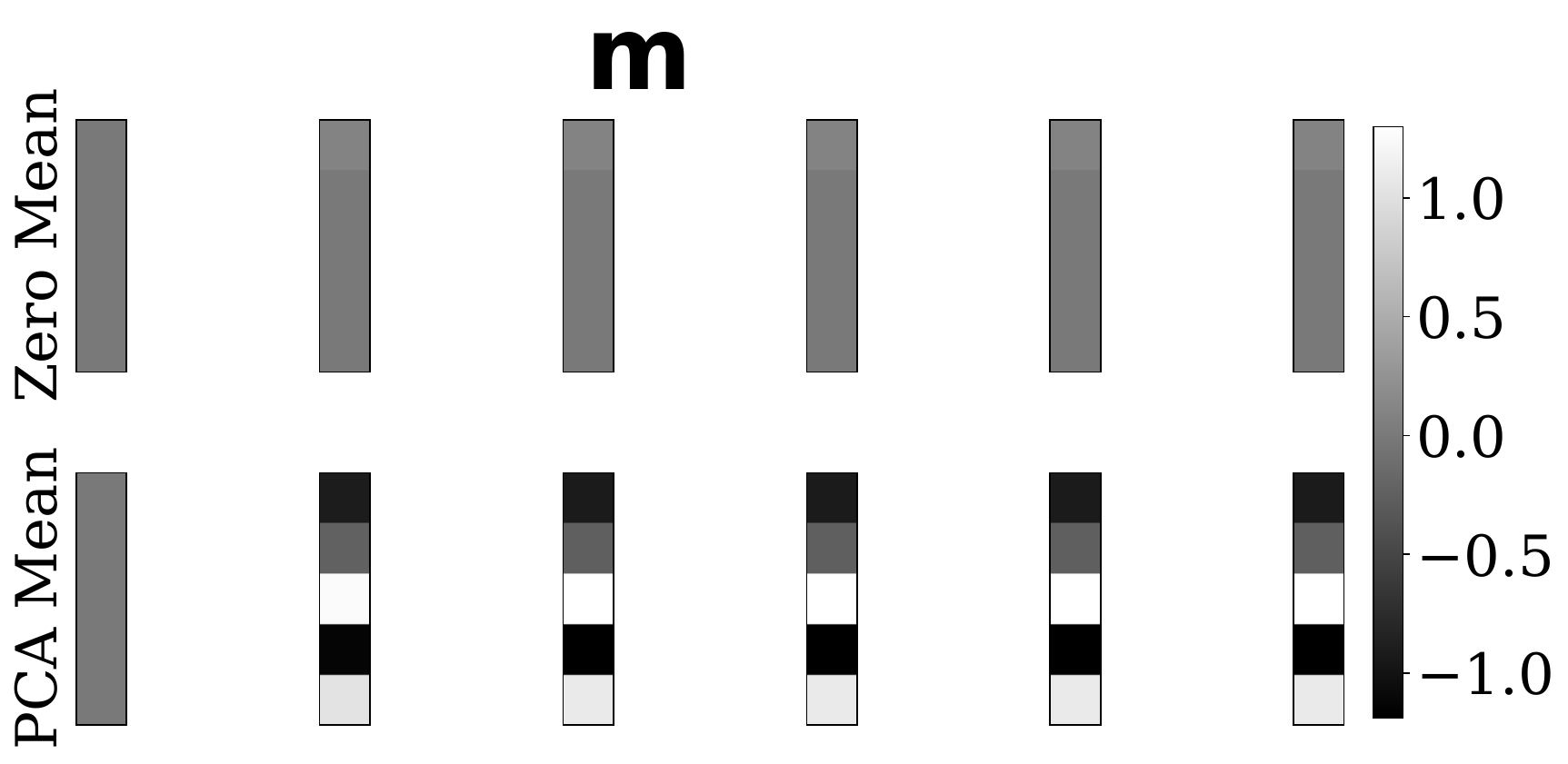}
    \end{subfigure}
\begin{subfigure}{\linewidth}
        \centering
        \includegraphics[width=\linewidth]{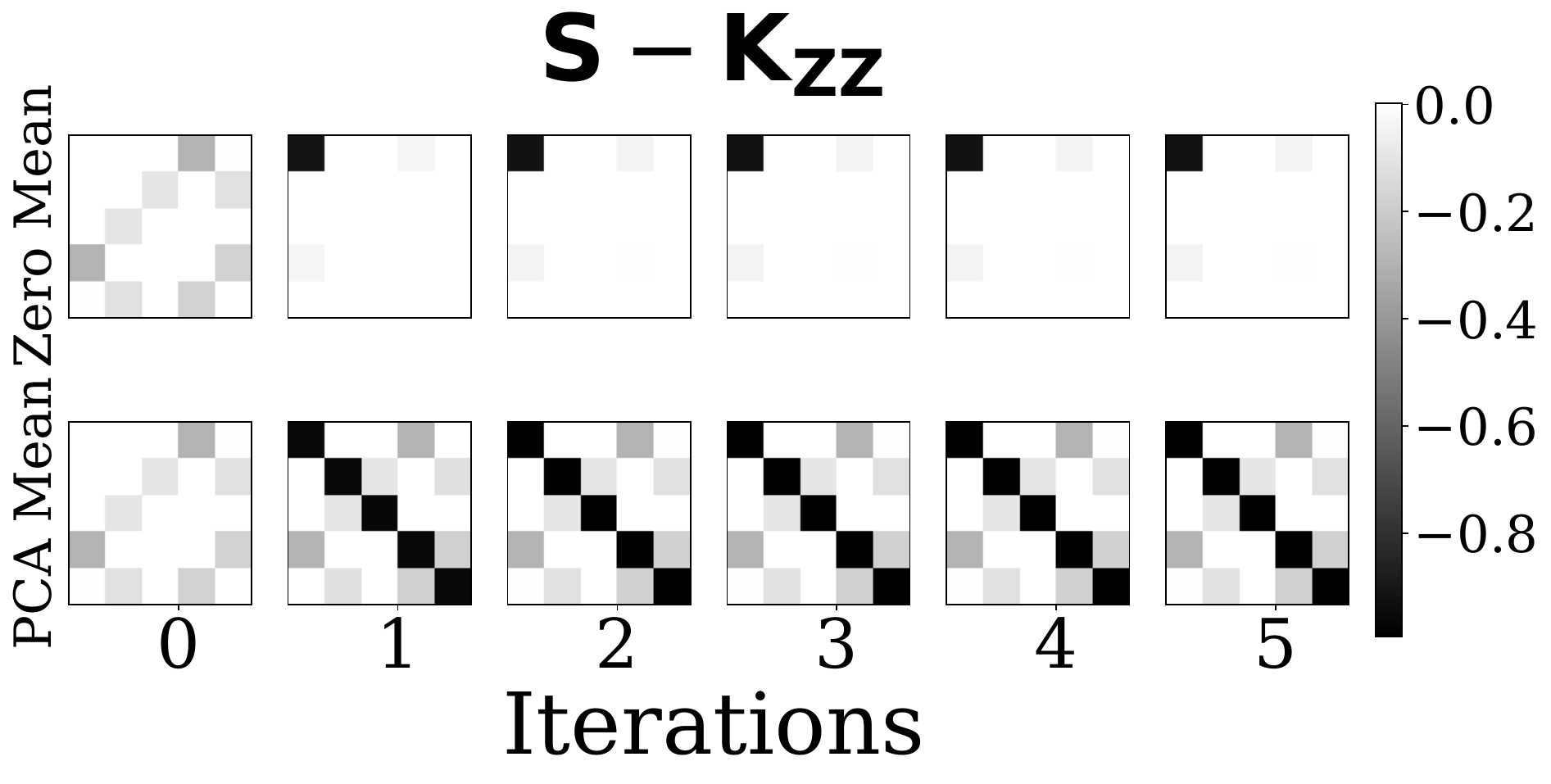}
    \end{subfigure}
\end{subfigure}
\\
\multicolumn{2}{c}{
    \begin{subfigure}{\linewidth}
        \centering
        \caption{Order of updates: $\sigma^2$-$\varS{}{}$-$\varm{}{}$, with initialization $\varS{}{}=\matI$, $\varm{}{}=\veczero$}
    \end{subfigure}
} \\
\raisebox{0.10\height}{\begin{subfigure}{0.48\linewidth}
    \centering
    \includegraphics[width=\linewidth]{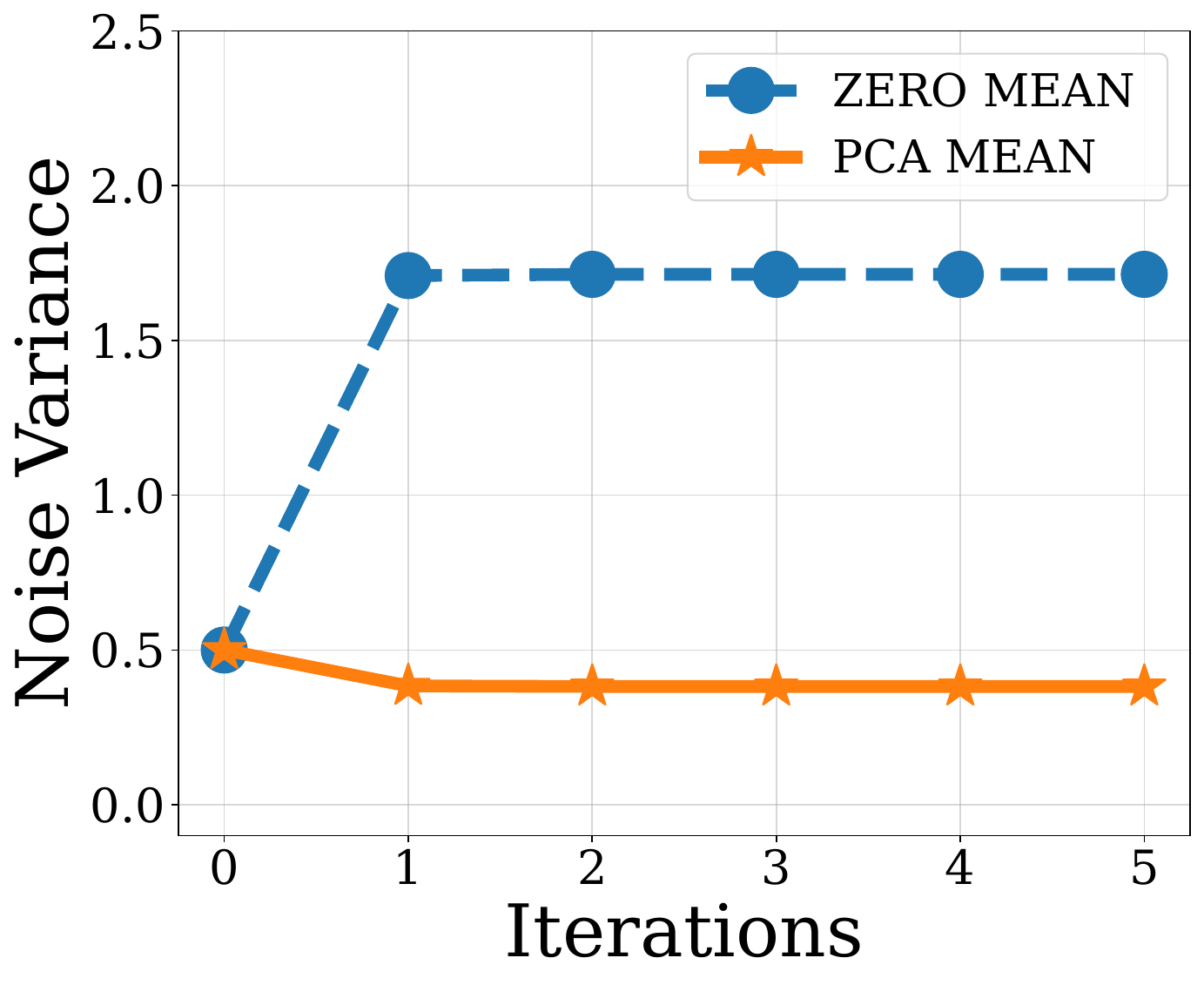}
\end{subfigure}}
&
\begin{subfigure}{0.48\linewidth}
    \centering
    \begin{subfigure}{\linewidth}
        \centering
        \includegraphics[width=\linewidth]{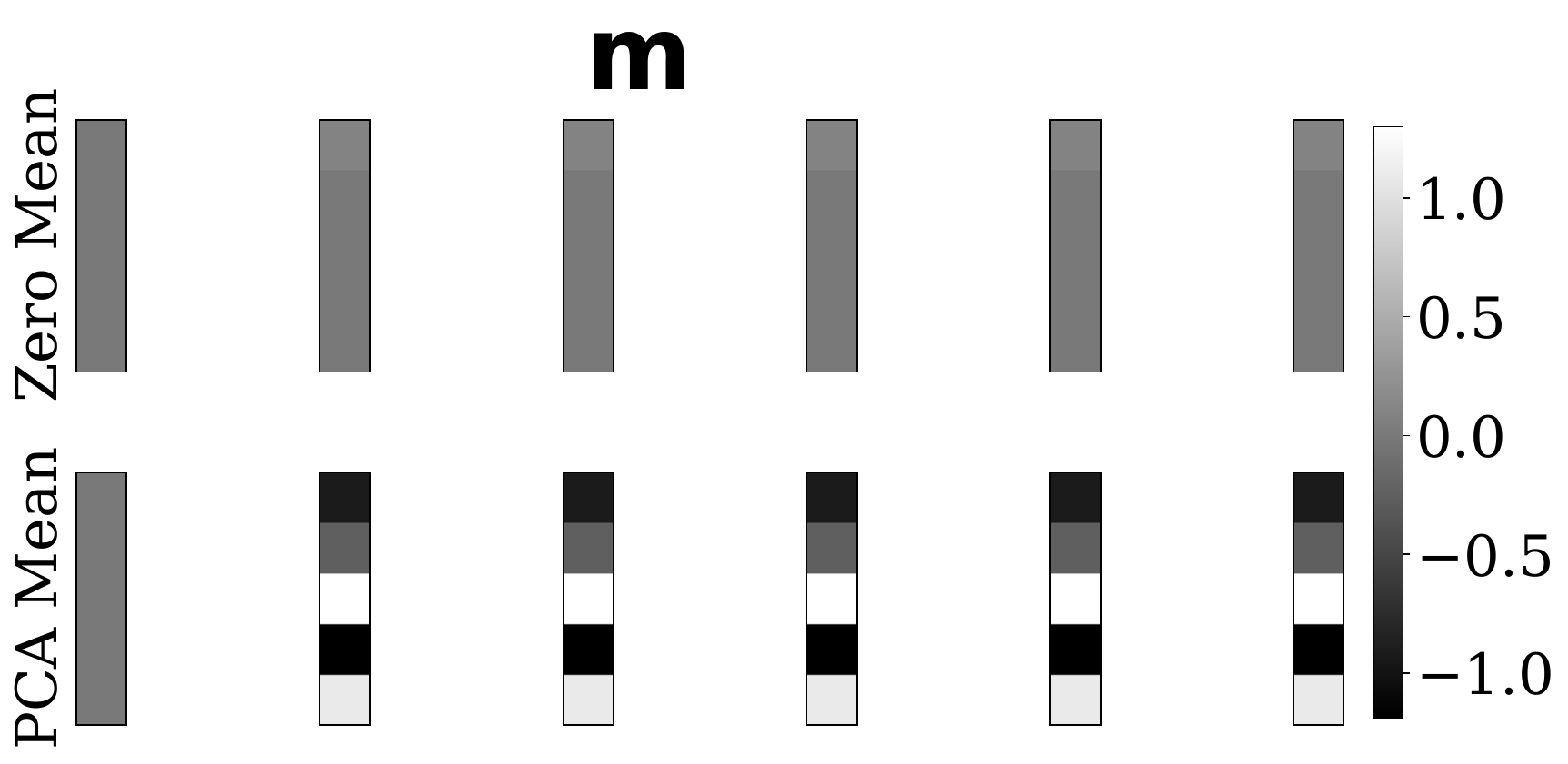}
    \end{subfigure}
    \begin{subfigure}{\linewidth}
        \centering
        \includegraphics[width=\linewidth]{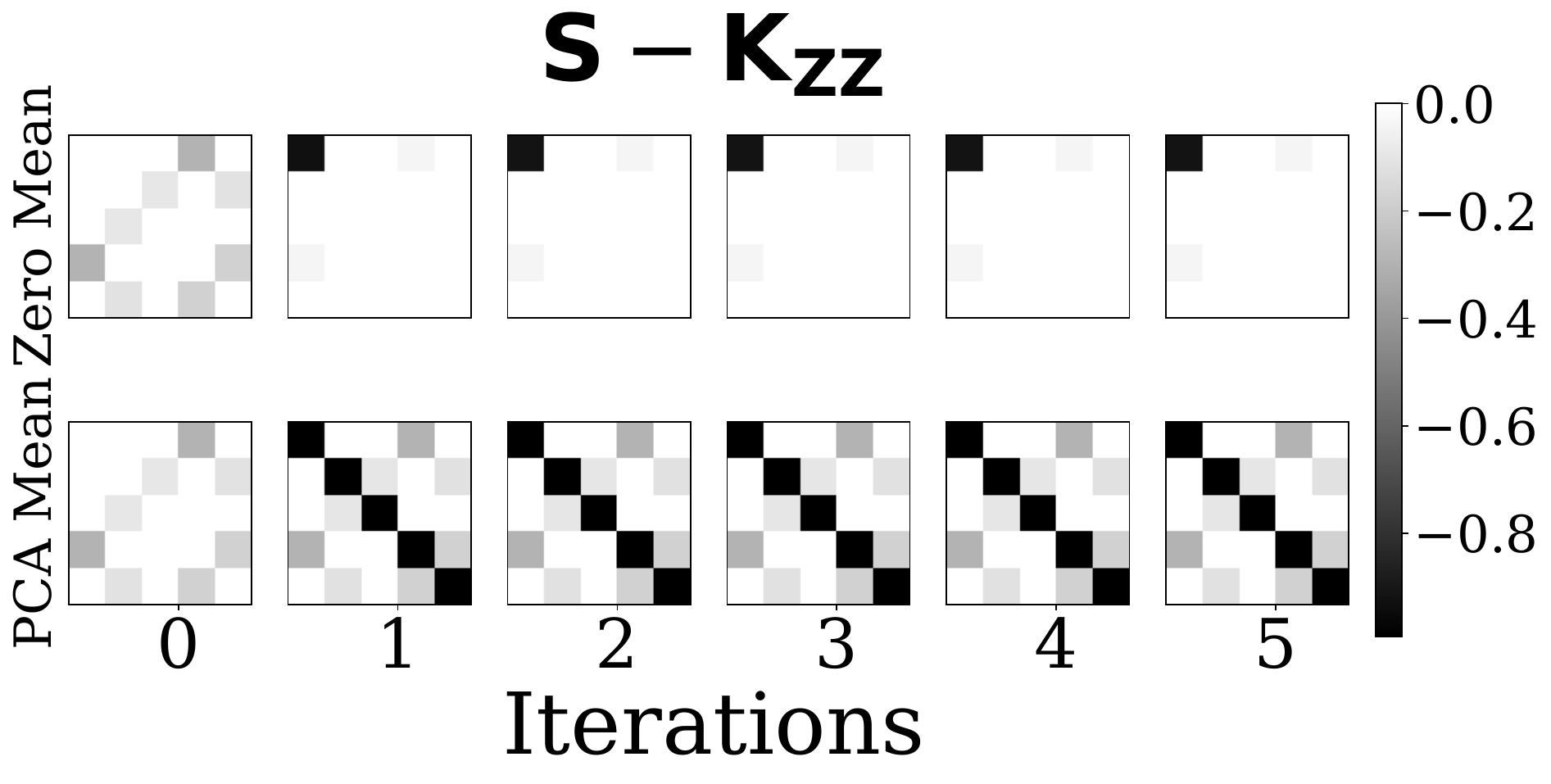}
    \end{subfigure}
\end{subfigure}\\
\multicolumn{2}{c}{
    \begin{subfigure}{\linewidth}
        \centering
        \caption{Order of updates: $\varS{}{}$-$\varm{}{}$-$\sigma^2$. Initialization does not matter since coordinate updates on $\varS{}{}$ do not depend on previous variational parameter values.}
    \end{subfigure}
}
\end{tabular}}

\caption{Each row shows coordinate updates, different initializations, and update order (indicated within each sub-caption). The left figures show the noise variance parameter. The right figures top show the evolution of the variational parameter $\varm{}{}$ and right-bottom figures show the value of $\varS{}{}-\Kzz{}$. This allows us to illustrate whether the parameters are close to a posterior collapse whenever $\varm{}{}$ and  $\varS{}{}-\Kzz{}$ are close to zero. Simulations are shown for both \PCA and \ZERO prior mean sparse \GP{}s using $5$ inducing points.}
\label{fig:coordinate_simulation_Z5}
\end{figure}

\paragraph{Simulations:} We simulate different scenarios on the toy problem described in \usec \ref{usec:graphical_mean_collapse_description}. 
The inputs for the \PCA prior mean sparse \GP are $\X{}{n}$. In the case of the \ZERO mean sparse \GP, they are a vector of zeros.
\fig \ref{fig:coordinate_simulation_Z5} report the optimization of the \ELBO using coordinate updates for $5$ inducing points (see \fig \ref{fig:coordinate_simulation_Z64} in the appendix for simulations using $64$ inducing points). In the figure, we find $3$ rows. The first and second rows consider the update order $\sigma^2-\varS{}{}-\varm{}{}$ for $\varS{}{}=10^{-5}\matI$ and $\varS{}{}=\matI$, respectively. The third row shows coordinate updates for the order $\varS{}{}-\varm{}{}-\sigma^2$. In this case, the initial value of $\varS{}{}$ is irrelevant. The left plots show the estimated noise variance for both $\ZERO$ and $\PCA$ prior mean models, for five iterations. Iteration $0$ corresponds to the $\sigma^2$ value at initialization, which is $0.5$. The right plots show the optimization, for both models, of the variational mean and $\varS{}{}-\Kzz{}$ using images. When $\varm{}{}$ and $\varS{}{}-\Kzz{}$ take values around $0$, we expect a posterior collapse to the prior. That is, $\varm{}{}=\veczero$ and $\varS{}{}=\Kzz{}$, which are the \KLD minimizers.

\fig \ref{fig:coordinate_simulation_Z5} shows that the model with the \ZERO prior mean learns higher noise variances, and mostly collapses the variational parameters to the prior (the variational mean is slightly different from zero), while the \PCA prior mean model does not. Total collapse  is not produced in practice unless $\Kxz{}{}=\veczero$, because $\sigma^2$ cannot grow to $\infty$. This supports our previous analysis, where we have shown that the predictive variances are bigger for the ZERO prior mean, and also the
probability that $\Kxz{} = \veczero$, effectively yielding higher noise variance parameter updates.

Regarding the order of the updates, $\varS{}{}-\varm{}{}-\sigma^2$ consistently provides better convergence, as we expected. Importantly, the \PCA prior mean model updates significantly reduce the noise variance parameter. In each model, both ordering updates converge to the same maximum of the \ELBO. We also observe in \fig \ref{fig:coordinate_simulation_Z5} that smaller values of $\alpha$ yield smaller values of $\sigma^2$ at the first iteration (order $\sigma^2-\varS{}{}-\varm{}{}$) since they result in a smaller initial predictive variance.  These coordinate updates also show that an initialization of $\varS{}{}=\matI$ in the output layer is more prone to the posterior collapse problem, as already discussed. Finally, an experiment with more inducing points supports these conclusions and is analyzed in appendix \ref{apx:coordupdates:64}.

The coordinate updates provide an intuitive idea behind the local minima that we may find early at the training stage of our \DGP{}. Since we have closed-form updates, we are removing any noise coming from stochastic gradients or from steps coming from different learning rates, which might help to jump out of the pathological local optima. Nevertheless, overall our analysis shows that early in the training,  it is likely that a \ZERO prior mean \DGP{} will end up learning the targets as just noise and collapsing the variational parameters, unless previous layers move out quickly from providing a zero input vector signal at the last layer. 

Finally, to complete our analysis, we need to consider what happens when all the parameters get optimized. In this case, there are no exact coordinate updates for the inducing points nor the kernel parameters. These parameters get optimized by gradient-based coordinate updates. In some cases, even if no exact coordinate updates are given, it is easy to understand what happens to some of these parameters. For instance, we have previously shown in \ueqn (\ref{equ:max_noise_var}) that, when collapse happens, the likelihood noise variance is given by the sample variance and the kernel output scale variance $\sigma_o$. However, when $\sigma_o$ gets optimized, this is not true. In this case, it is tempting to think that $\sigma_o$ will grow without bound. This leads to a noise parameter $\sigma_{\text{opt}}^2$ that attains arbitrarily large values and, consequently, favors overestimating the variance and explaining the data as noise. However, in the whitening case, when collapse happens, $\varS{}{}=\matI$. Thus, attending to the objective in \ueqn \ref{eq:ell_expansion}, the trace only depends on $\Kxx$ for any point, and takes its smallest value by making $\sigma_o=0$. This is because in the whitening case, $\sigma_o$ does not influence the \KLD. This is observed in one of our experiments.

\subsection{Impact of the Parameterization on the Optimization Process}
\label{collapse_parameterization_instabilities}

\begin{figure}[!t]
    \centering

    \begin{subfigure}[b]{\linewidth}
        \centering
        \includegraphics[width=0.32\linewidth]{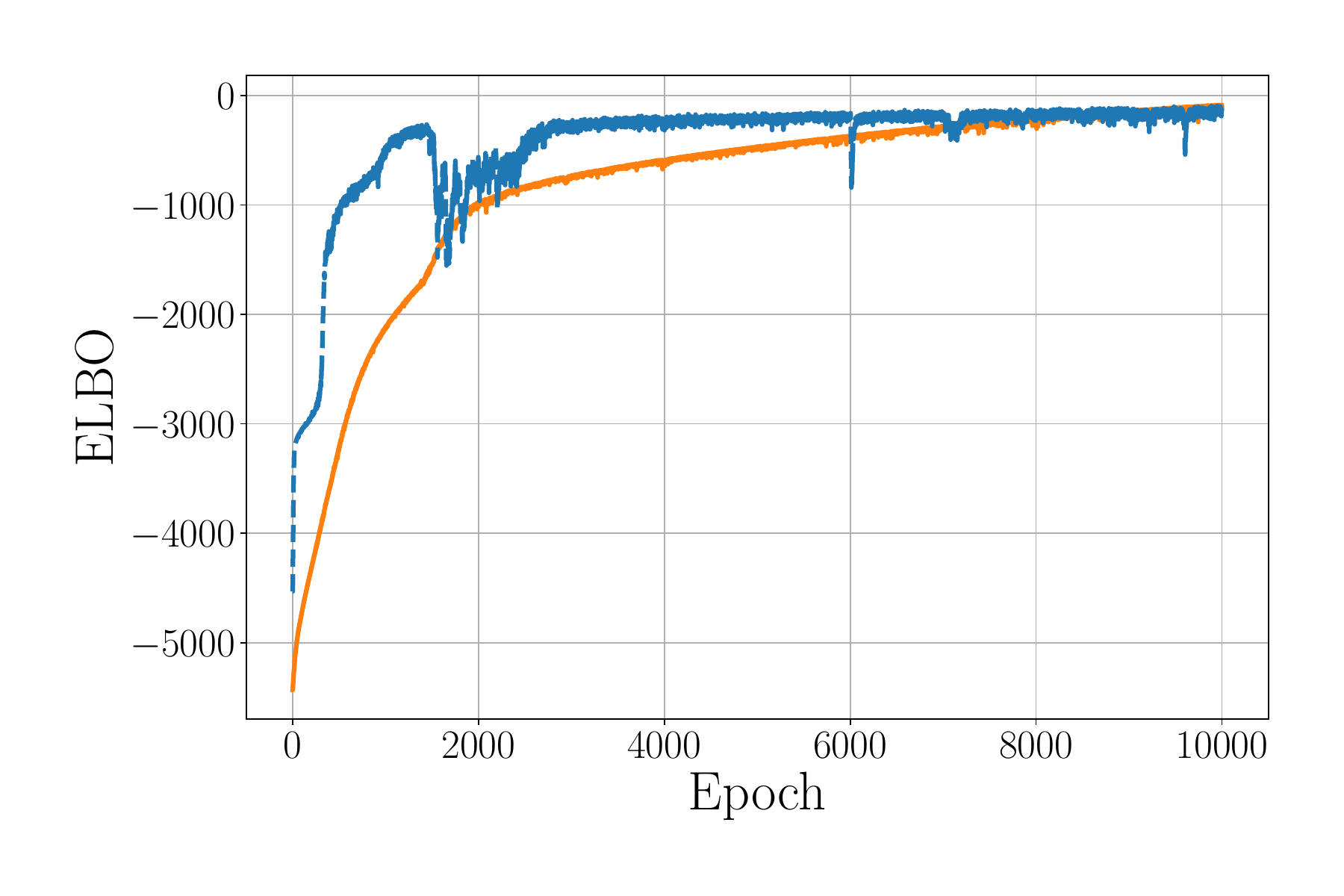}
        \includegraphics[width=0.32\linewidth]{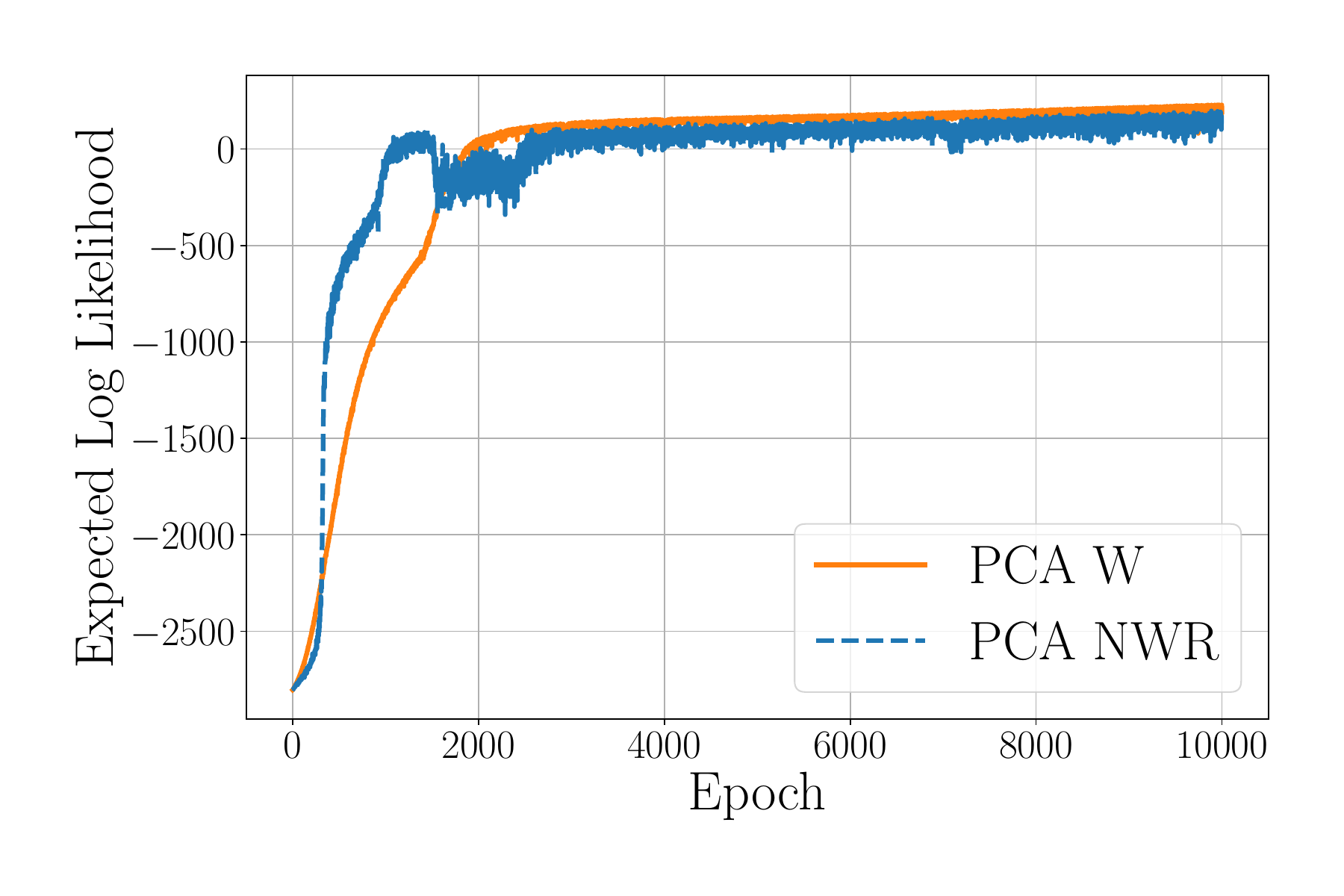}
        \includegraphics[width=0.32\linewidth]{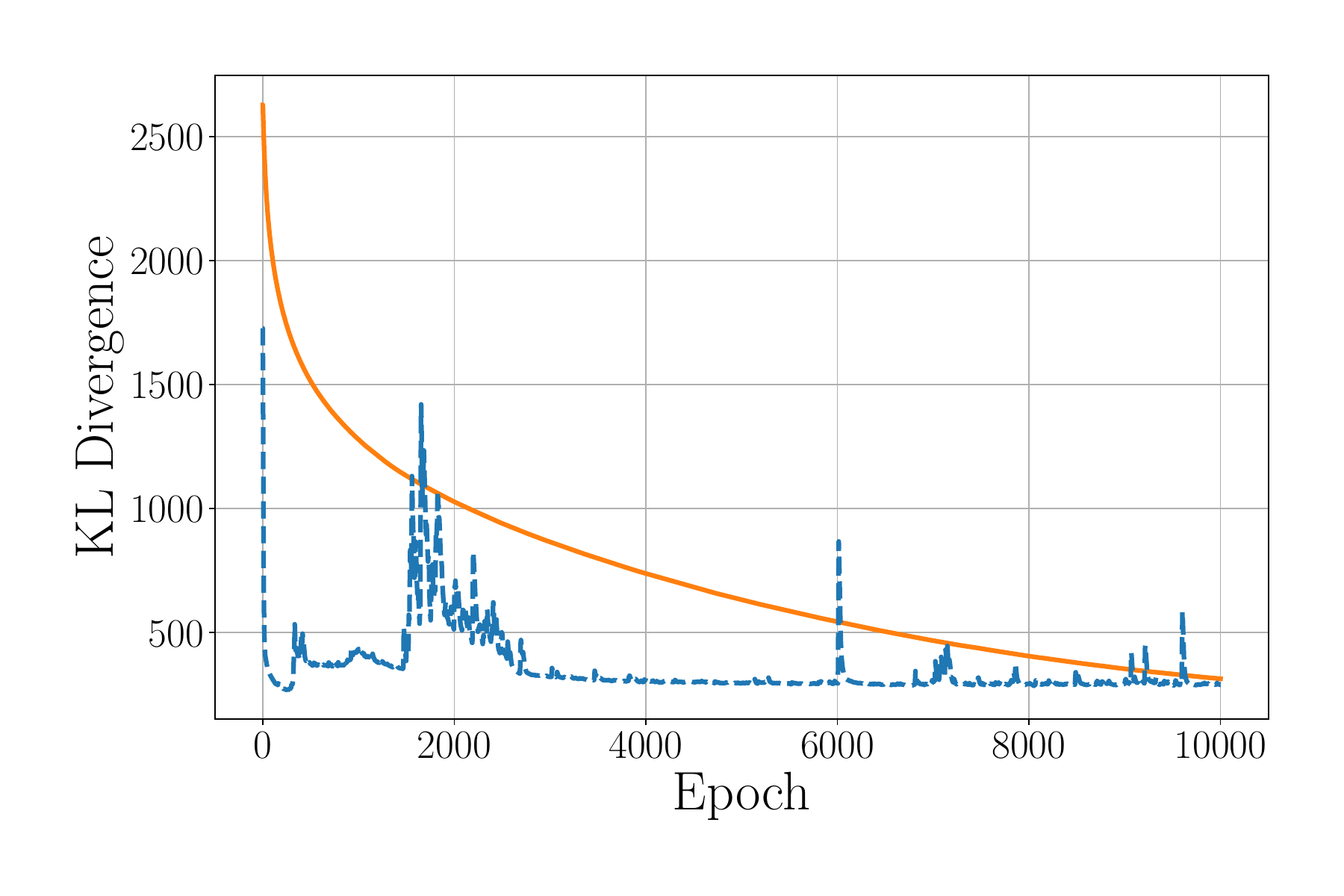}
        \caption{Optimization curves for unnormalized data}
    \end{subfigure}

    \begin{subfigure}[b]{\linewidth}
        \centering
        \includegraphics[width=0.32\linewidth]{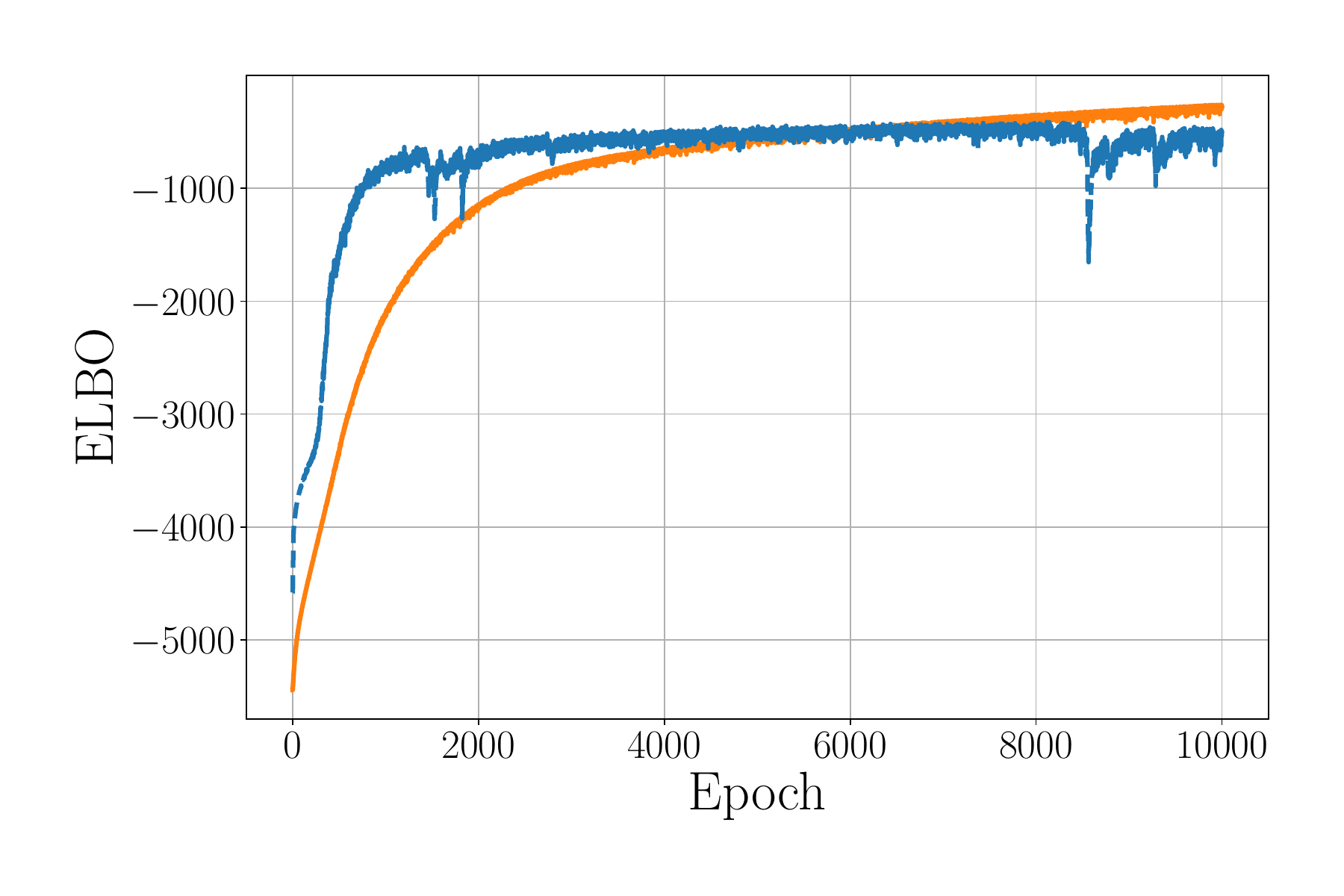}
        \includegraphics[width=0.32\linewidth]{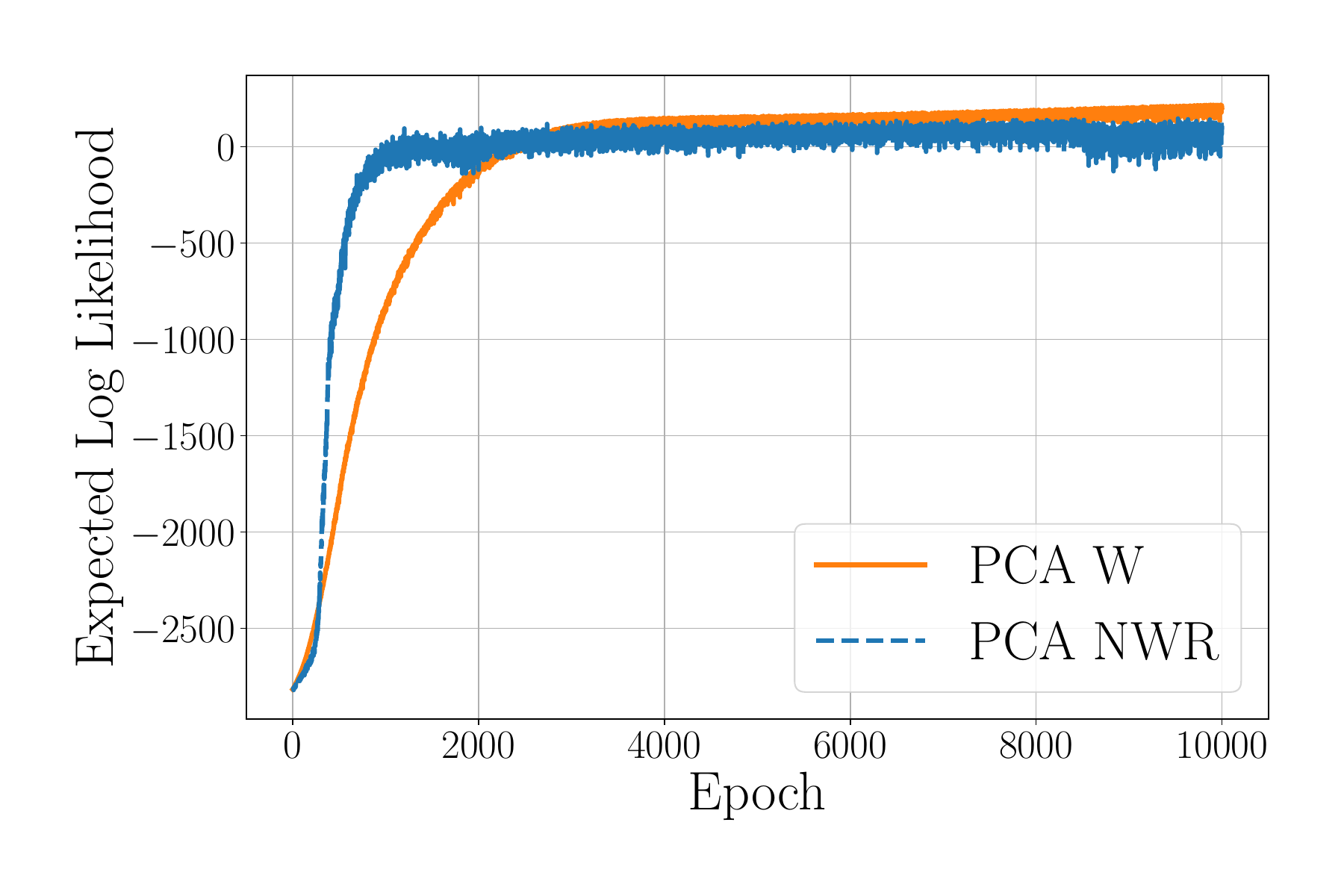}
        \includegraphics[width=0.32\linewidth]{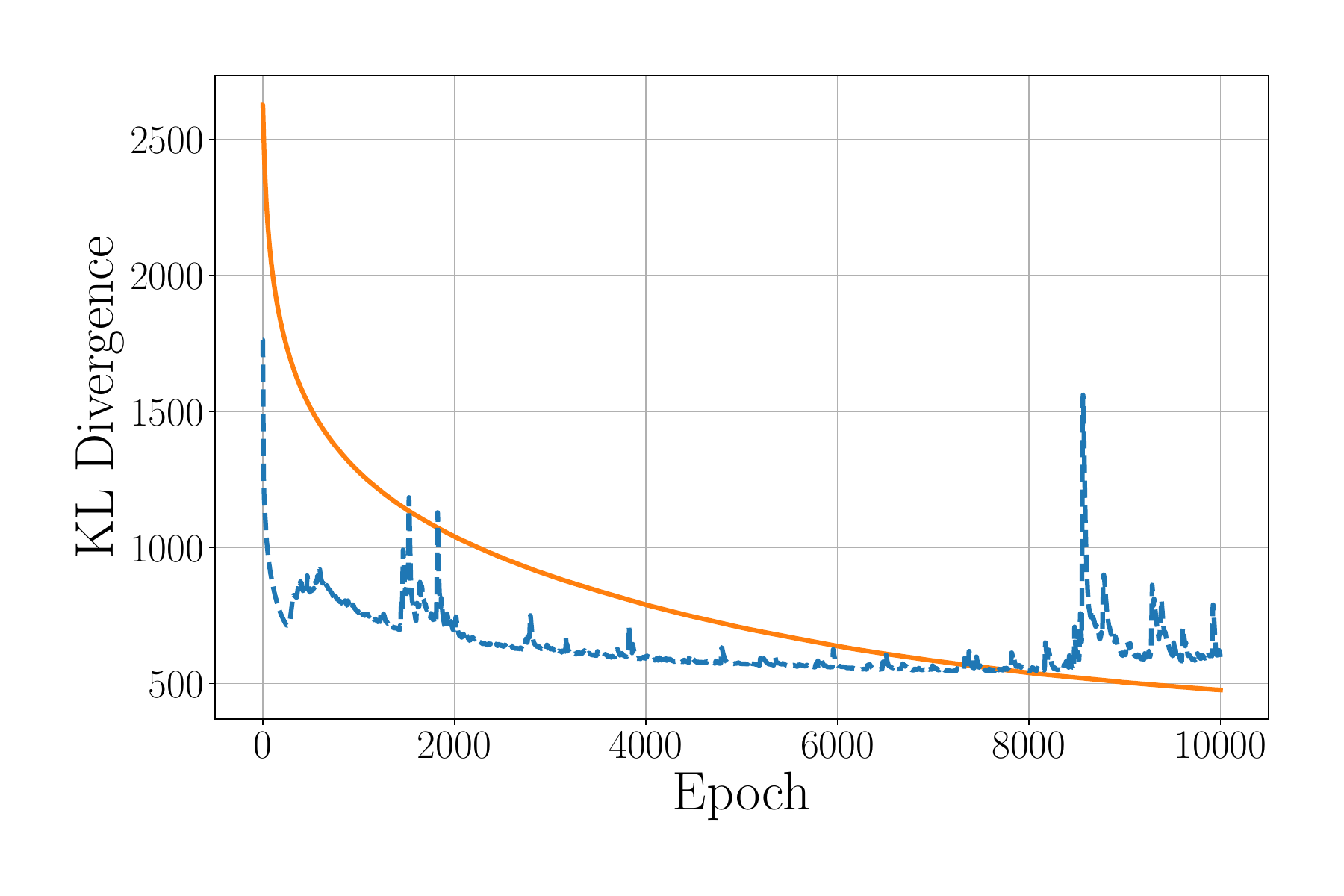}
        \caption{Optimization curves for normalized data}
    \end{subfigure}

    \caption{Training curves reporting the \ELBO, the \ELL term, and the \KLD term of a \PCA prior mean \DGP on the toy problem, for 
	both the whitened and the non-whitened parameterization of \GPFLOW. Figures show that the non-whitened parameterization is more unstable.}
    \label{fig:optimization_curves_NW_vs_W}
\end{figure}

In \usec \ref{sec:whitened_param}, we observed that the posterior collapse problem may happen independently of the parameterization used. 
Intuitively, however, the whitened parameterization should yield a more stable optimization procedure than the non-whitened parameterization. 
This verified empirically in \fig \ref{fig:optimization_curves_NW_vs_W}. This figure shows, in the toy problem, the optimization of the \ELBO and its
decomposition in the \ELL term and the \KLD term, when the \PCA prior mean function is used, for the whitened parameterization (solid orange line) and the
non-whitened parameterization of \GPFLOW (dashed blue line).  Here, we initialize $\varm{}{}=\veczero$ and $\varSw{}{}=\varS{}{}=10^{-5}\matI$.
In this section, we show that the reason behind these instabilities for the non-whitened parameterization
is a poorly conditioned Hessian of the \KLD term. Our analysis also shows that, in the non-whitened model, the reparameterization of the variational mean of \GPFLOW  does not 
provide better convergence guarantees. In general, the training instabilities of the non-whitened parameterization can make the model arrive at 
the local minimum in which the variational parameters are set equal to the prior, making it more likely for the posterior collapse problem.

\subsubsection{The \KLD at Initialization}

First, in the non-whitened representation, the \KLD at initialization, for $\varm{}{}=\veczero$, is expected to be smaller
for the \GPFLOW reparameterization of the variational mean (see Table \ref{tab:parameterizations}). 
The reson is the impact of the quadratic term $(\varm{}{} - \muZ{})^T\Kzzinv{}(\varm{}{} - \muZ{})$, which makes the \KLD greater for 
mean functions such as the \PCA. Regarding the non-whitened vs. whitened parameterization, it will depend on the initial choice of the 
variational parameters. Initially, we may think the whitened representation will have smaller \KLD on initialization, since the variational and 
prior mean are zero and the covariance matrix $\varS{}{}=\alpha\matI$ is diagonal at initialization. For $\alpha=1$, this is clearly true, 
and $\KLD=0$. However, a careful inspection of the \KLD at initialization reveals opposite observations for other values of $\alpha$. 
In particular, 
\begin{align}
	\KLD_{\text{non-whitened}}&=\frac{1}{2}\bra{\log \mid\Kzz{}\mid - M \log \alpha -M + \alpha\tr{\Kzzinv{}}}\,,\\
	\KLD_{\text{whitened}}&=\frac{1}{2}\left[ - M \log \alpha-M + M\alpha \right]\,.
\end{align}
Ignoring common terms $ -M \log \alpha$ and $-M$, and expressing computations in terms of the eigenvalues of $\Kzz{}{}$, $\lambda_k$, we have: 
\begin{align}
	\KLD_{\text{non-whitened}} &= \frac{1}{2}\left(\mysum{k=1}{M}\log \lambda_k + \alpha \mysum{k=1}{M}\frac{1}{\lambda_k}\right) + \text{const.}\,,\\
	\KLD_{\text{whitened}}&=\frac{1}{2}M\alpha + \text{const.}
\end{align}
Thus, there is no reason to establish that any of the parameterizations will have a higher \KLD at initialization. For small $\alpha$, the \KLD in the non-whitened case is dominated by the term $\mysum{k=1}{M}\log \lambda_k$. This implies that only when the length-scale is small relative to the data, we will have $\Kzz{}\approx \matI$ (for an output scale parameter of $1$), and this term will vanish. If the matrix has a large number of large eigenvalues, we can expect this term to be above $0$, and \KLD to be greater. For moderately big $\alpha$, the term is dominated by $\mysum{k=1}{M}\nicefrac{1}{ \lambda_k}$. In any case, any degenerated direction will incur a high \KLD. However, for the usual length-scales used at initialization (from $0.1$ to $2.0$) and using the fact that smaller values for $\alpha$ are generally used, we observe that the \KLD of the non-whitened parameterization will be smaller at initialization, as reported in \fig \ref{fig:optimization_curves_NW_vs_W}.
A smaller initial \KLD is expected to increase the probability of a posterior collapse problem. 

\subsubsection{Optimization Instabilities in the \KLD Term}
\label{subsec:source_of_instabilities}

\fig \ref{fig:optimization_curves_NW_vs_W} shows that the \KLD has a great impact on the instabilities of the non-whitened parameterization of \GPFLOW. Even when the data is normalized, the non-whitened parameterization presents unstable convergence.
Therefore, we look at the Jacobian of the \KLD, for the non-whitened parameterization of \GPFLOW, reported in 
Table \ref{tab:parameterizations}. At initialization, such a Jacobian is: 
\begin{align}
	\frac{\partial \KLD}{\partial \varm{}{}} &= \veczero\,, &
	\frac{\partial \KLD}{\partial \muZ{}{}}  &= \veczero\,, \\
	\frac{\partial \KLD}{\partial \varS{}{}} &= \frac{1}{2}\left(\Kzzinv{}-\varSinv{}{}\right)\,, &
	\frac{\partial \KLD}{\partial \Kzz{}} &= \frac{1}{2} \left(\Kzzinv{}-\Kzzinv{}\varS{}{}\Kzzinv{}\right)\,.
\end{align}
By contrast, in the case of the whitened parameterization, the Jacobian is:
\begin{align}
	\frac{\partial \KLD}{\partial \varmw{}} &= \veczero\,, &
	\frac{\partial \KLD}{\partial \muZ{}}  &= \veczero\,, \\
	\frac{\partial \KLD}{\partial \varSw{}{}} &= \frac{1}{2}\left(\matI -\varSwinv{}{}\right)\,, &
	\frac{\partial \KLD}{\partial \Kzz{}} &= \veczero \veczero^T\,.
\end{align}
These equations show that the bigger the \KLD, (\emph{\ie}, the bigger the difference between prior and variational covariances), the bigger the Jacobians are. This would imply using an adaptive learning rate algorithm to take more time to adjust and produce stable convergence. Moreover, since, in our toy problem, the whitened model has higher \KLD, we should expect worse-behaved convergence. However, this is not the case.
We believe this may be because in the whitened parameterization, the \KLD does not depend on $\Kzz{}$ and hence the \KLD Jacobian is zero \wrt the inducing points, 
which should lead to more stable gradient updates.

One way to analyze the optimization stability of a loss function is via its Hessian, which measures the curvature of the loss in all parameter-space directions. If the Hessian is poorly conditioned, \ie, its eigenvalues differ widely across directions, first–order gradient optimizers struggle, and convergence slows down due to exceedingly small steps in some directions and excessively large steps in others. This produces updates that oscillate around the minima, which is the noisy behavior observed in the non-whitened reparameterized model. We have compiled the derivations needed to analyze the Hessian of both parameterizations in a separate work \citep{maronas2025jacobianhessiankullbackleiblerdivergence}. This work shows that the Hessian is highly dependent on $\Kzz{}$, in the case of the non-whitened parameterization, and that the reparameterization implemented by \GPFLOW (see \usec \ref{usec:non_whit_in_practice}) does not provide any convergence improvement. 

Overall, this justifies why the non-whitened parameterization provides unstable convergence, and we show that these instabilities may result in poor solutions where collapse appears. However, we 
note that these observations open the door to a broader and more detailed theoretical analysis of optimization stability that falls outside the scope of this paper. A comprehensive treatment of these aspects will be presented in a separate work.

\subsection{Multi-output Models, Other Likelihoods and Alternative Assumptions}

Regarding other likelihoods, the first natural question is what we can expect from a $C$ multi-output \DGP{} regression model with a Multivariate Gaussian 
likelihood. This likelihood has a noise covariance parameter $\bm{\Sigma}$, which might not be diagonal. One can show that coordinate updates in 
this model can be written down in a similar way to the coordinate update of the 1-dimensional regression setting (see appendix \ref{sec:app:A}). Namely,
\begin{align}
	\bm{\Sigma}_{\text{opt}} & = \frac{1}{N}\mysum{n=1}{N}\pare{\Y{}{n}-\qfm{}{n}}\pareT{\Y{}{n}-\qfm{}{n}} + \qfS{}{n}\,,
\end{align}
where now $\Y{}{n}$ is a $C$-dimensional target, $\qfm{}{n}$ is the $C$-dimensional predictive mean vector and $\qfS{}{n}$ is the $C\times C$ predictive covariances at $\X{}{n}$. From this coordinate update, we can identify similar aleatoric and epistemic terms. The predictive covariances $\qfS{}{n}$ are expected to maintain the same pathological properties we have observed so far, both if the processes are independent or dependent (assuming we have initialized the mixing matrix to $\matI$). Specifically, $\qfS{}{n}$ is obtained via similar expressions as those given in \ueqns~(\ref{equ:pos_pred_params_non_whit}) and \ueqns~(\ref{equ:pos_pred_params_whit}) using block (sometimes diagonal) matrices. Thus, the problem of the output layer receiving as input a vector of zeros is exactly the same. In general, any likelihood that has a noise parameter to model aleatoric uncertainty, such as the Student-t for robust regression or the Laplace, may be more likely to suffer from posterior collapse. This includes, obviously,  heteroscedastic regression, where the model can learn high variances per datapoint.

Other likelihoods for other tasks, such as classification (categorical or Bernoulli likelihoods) or regression over count data (Poisson Likelihood), might need more time to achieve convergence, but do not have a parameter to learn everything as noise. For example, in classification problems, aleatoric and epistemic uncertainty are obtained from the average entropy and mutual information after sampling from the posterior distribution, so uncertainty estimation depends purely on the latent functions \citep{pmlr-v80-depeweg18a}. This means that posterior collapse is expected to be less relevant and is likely to depend only on the number of datapoints and processes. For a categorical predictive distribution, the training objective in a $C$ classification problem can be written down (see, for instance, \cite{etgp} for a formal derivation) as:
\begin{align}
	\ELBO &= 
    \frac{1}{S}\mysum{s=1}{S}\mysum{n=1}{N}\int q(\fpos{1,s}{L}\hdots\fpos{C,s}{L}) \log p(\Y{}{n}\mid \text{Softmax}(\fpos{1,s}{L}\hdots\fpos{C,s}{L})) \nonumber \\
	& \quad -\mysum{c=1}{C}\KLD\bra{q(\upos{}{c})\mid\mid p(\upos{}{c})}\,.
\end{align}
Now, if we consider the model at initialization, we know that the inputs to the last layer will be around $\veczero$, for the \ZERO prior mean \DGP{}. 
This implies that the \ELL evaluates to $\log(\nicefrac{1}{C})$ for each training sample at initialization. For large $C$ this term becomes very small, the \KLD is not going to dominate, and the model is forced to push the parameters to stop predicting latent functions around $\veczero$. For small $C$, it might be more beneficial to output constant predictive distributions and minimize the \KLD term. Therefore, for classification problems, we might expect either posterior collapse or a substantial time to converge. The \PCA prior mean \DGP will not have this problem.

We have seen that most of the issues related to the posterior collapse problem come from the distance between $\veczero$ and the inducing points. 
This suggests that the \ZERO prior mean \DGP might benefit, at initialization, from using non-stationary kernels. However, previous work has demonstrated 
that non-stationary kernels can incur severe misspecification, and that well non-stationary inductive biases need to be encoded for successful performance. 
See the work of \cite{etgp} for further discussion and comparison.

Regarding the use of more Monte Carlo samples to estimate the \ELL in the \ELBO, we shall expect an exacerbated effect if more samples are used. 
We believe that fewer samples relate to higher variance of the \ELBO gradients, which may be useful to escape bad local optima such as the one associated with the posterior collapse problem.  On the other hand, as the number of Monte Carlo samples increases, the estimator of the \ELL converges to its expected value. Since this expectation is taken under a distribution with zero mean, this convergence amplifies the degeneracy we have been illustrating, bringing the optimization process closer to the pathological regime associated with posterior collapse. Since in practice one sample is enough for the model to achieve good convergence, and due to computational cost, we have not carried out extra experiments exploring this direction.

\subsection{Conclusions about how to Initialize \DGP{}s}

	In this section, we have consciously analyzed the posterior collapse problem under different parameterizations and initializations of \DGP{}s. The takeaways from the analysis are mainly concerned with removing most of the noise from the optimization procedure. This is, in turn, achieved by initializing the likelihood noise parameter $\sigma^2$ to a small value, and the variational variance $\varSw{}{}=\varS{}{}=\alpha\matI$  with $\alpha\leq \sigma_o$. Other valid options are $\varS{}{}=\alpha\Kzz{}$ with $\alpha\leq \sigma_o$. In general, select $\varS{}{}$ so that $\Kzz{} - \varS{}{}$ is positive definite since it implies $\Kzz{}\succeq\varS{}{}$.  This will result in a predictive variance that does not go beyond the prior at initialization, resulting in a smaller trace term in the \ELL. Setting $\varSw{}{}=\varS{}{}=\alpha\matI$, for $\alpha\leq \sigma_o$, always satisfies this requirement in both parameterizations. A careful look at some \DGP{} implementations has revealed that these two options are the most common in many works under any of the three parameterizations. Moreover, more inducing points help to alleviate posterior collapse. Finally, if posterior collapse is still present, one can reduce the kernel scale parameter $\sigma_o$, which might result in smaller predictive variances at input locations far from the inducing points. Also, using a whitened parameterization results in better-behaved gradients, reducing the noise in the optimization and reducing the posterior collapse problem.

\section{An Initialization to Alleviate Posterior Collapse}
\label{sec:proposed_initialization}

In the previous section, we have described the optimization problems suffered by \ZERO{} prior mean \DGP{}s, and explained why the \PCA prior mean function is able to reduce their impact. Since Bayesian theory dictates that the choice of the prior must be driven by modeling assumptions, and not by limitations of the optimization procedure, this section introduces an alternative way of initializing \ZERO prior mean \DGP{}s such that the possibility of a posterior collapse problem is reduced.

The idea relies on initializing the hyper-parameters and the variational parameters such that the predictive distribution of the \ZERO{} prior mean \DGP{} mimics the \PCA prior mean \DGP{} at initialization. By doing this, we can avoid the local optimum in which the posterior collapse occurs. We achieve this behavior by setting adequate initial values for the model and variational parameters so that the predictive mean $\qfm{}{},\qfmr{}{},\qfmw{}{}$ at initialization in the inner layers resembles that of a \DGP{} with the \PCA prior mean function. We present a general framework and impose restrictions to yield an efficient implementation. As we will show in the experiments section, this approach allows effective training of \DGP{}s with \ZERO prior mean functions. However, the proposed solution should be valid for a \DGP{} with any prior mean function. This general case is presented in \usec \ref{subsec:proposed_technique_arbitrary_mean_funs}, but has not been validated experimentally.

In this section, we will consider three parameterizations: whitened (\W), standard non-whitened (\NW), and non-whitened with the reparameterization of the variational mean of \GPFLOW 
(\NWR). Besides this, we will consider two mean functions $\PCA$ and $\ZERO$. Therefore, we define the following acronyms to 
refer to each of these models throughout the rest of the work, see \utab \ref{tab:acronyms}.

\begin{table}[h!]
\centering
\caption{Summary of the acronyms used to refer to the different models considered in this work, attending to their mean function and the parameterization of the inducing points.}
\begin{tabular}{l|c|l}
\hline
\textbf{Parameterization} & \textbf{Mean Function} & \textbf{Acronym} \\ 
\hline
Whitened                  & \ZERO                  & \ZEROW           \\ 
Non-whitened              & \ZERO                  & \ZERONW          \\ 
Non-whitened reparameterized & \ZERO              & \ZERONWR         \\ 
Whitened                  & \PCA                   & \PCAW            \\ 
Non-whitened              & \PCA                   & \PCANW           \\ 
Non-whitened reparameterized & \PCA               & \PCANWR          \\ 
\hline
\end{tabular}
\label{tab:acronyms}
\end{table}

\subsection{The Predictive Mean at Initialization}
\label{sec:subsec:marginal:variational:posterior:init}

Our goal is to mimic the \PCA prior mean \DGP{} at initialization using a \ZERO prior mean function. 
Therefore, we start by inspecting the predictive mean of this model, for the initialization outlined in \usec\ref{subsec:dgp_init} 
(\ie, $\varm{}{}=\varmw{}{}=\varmr{}=\veczero$). To lighten notation, we will be presenting our approach as if we had a single \GP per layer. 
Following \ueqns (\ref{equ:pos_pred_params_non_whit}), (\ref{equ:pos_pred_params_whit}), (\ref{equ:pos_pred_params_non_whit_rep}), 
the predictive mean of a \PCA{} prior mean \DGP{}s at initialization take values:
\begin{align}
	\qfm{}{} &= \Xsel - \Kxsz{}\Kzzinv{}\Zsamples{}\,, &
	\qfmw{}{} &=  \Xsel\,, &
	\qfmr{}{} &= \Xsel\,,
	\label{eq:initial_predictive_mean_pca_dgp}
\end{align}
for each parameterization, \W, \NW\ and \NWR, respectively, where $\Xsel$ is the set of points in which we are evaluating our posterior. In general, $\Xsel$ will represent the input points where we want our \ZERO prior mean \DGP to mimic the \PCA prior mean \DGP at initialization. This may include, \eg, the train set, a mini-batch of points, or the whole input domain $\Xspace$. Ideally, we would like to be able to mimic the \PCA prior mean \DGP{} at initialization on $\Xspace$, yet we will show in \usec \ref{sec:initialization:subsec:system} that this is not possible.

\begin{remark}
Following \usec \ref{subsec:computing_mean}, for a general \DGP{} with an arbitrary number of \GP{}s per layer, Eq. (\ref{eq:initial_predictive_mean_pca_dgp}) is extended by noting that $\Xsel$ represents the $d-$th coordinate of $\Xsel$ in the evaluation of the predictive mean of the \PCA prior mean \DGP. Also, when the number of \GP{}s at the level of the hierarchy is smaller than at the previous level, the predictive mean function at that layer is given by a \PCA projection of the inputs $\Xsel$. Thus, we just need to change $\Xsel$ by the corresponding \PCA projection. 
\end{remark}

When the model is a \ZERO prior mean \DGP, the predictive mean is:
\begin{align}
	\qfm{}{} &= \Kxz{}\Kzzinv{}\varm{}{}\,, &
	\qfmw{}{} &=  \Kxz{}\braT{\Lzzinv{}}\varmw{}\,, &
    \qfmr{}{} &= \Kxz{}\Kzzinv{}\varmr{}\,,
	\label{eq:initial_predictive_mean_zero_dgp}
\end{align}
for each parameterization. Clearly, at initialization, these predictive means evaluate to $\veczero$. Thus, to select initial parameters so that, at initialization, a \ZERO prior mean \DGP{} has the same predictive mean at initialization as that of a \PCA prior mean \DGP{}, we need to match Eq. (\ref{eq:initial_predictive_mean_pca_dgp}) with Eq. (\ref{eq:initial_predictive_mean_zero_dgp}), at each inner layer. This involves solving a system of linear equations. The learnable parameters are $\varm{}{}$, $\varmw{}$, $\varmr{}$, $\Zsamples{}$, $\Xsamples{}$, $\Xsel$ and $\nupos{}{}$.

\subsection{Solving the System of Equations}
\label{sec:initialization:subsec:system}
The three systems of equations that result from the above idea are:
\begin{align}
   \Kxz{}\Kzzinv{}\varm{}{} &= \Xsel -  \Kxsz{}\Kzzinv{}\Zsamples{} \tag{\NW}\,,\\
   \Kxz{}\braT{\Lzzinv{}}\varmw{} &= \Xsel \tag{\W}\,, \\
   \Kxz{}\Kzzinv{}\varmr{} &= \Xsel \tag{\NWR}\,,
\end{align}
for each parameterization.  We note that we do not require the set of points $\Xsel$ to be the same as $\Xsamples$ in order to define the system, but it is required that $\dim(\Xsel)=\dim(\Xsamples)$, where $\dim(\cdot)$ represents the dimensions (rows and columns) of its argument, for the three systems to be well-defined. Since it is counterintuitive to require the \ZERO prior mean \DGP to mimic the \PCA prior mean \DGP at different points in the domain, we simply set $\Xsamples=\Xsel$. Ideally, we would like to mimic the \PCA prior mean \DGP at the full domain, \ie $\Xsel=\Xspace$. However, this implies having an infinite number of equations.

With this in mind, under some assumptions, these systems can be turned into linear systems that can be solved without requiring numerical gradient-based optimization. However, this is not a restriction but a decision we have made. Thus, there is freedom to go through other approaches that might be interesting for other reasons (for example, incrementing the support $\Xsel$), and still yield the desired initialization. These additional details are presented in Appendix \ref{sec:app:B}.

The set of assumptions we have made includes fixing the values of the kernel hyper-parameters $\nupos{}{}$, $\Xsel$, and $\Zsamples{}$ and solving for $\varm{}{}$. In this case, the optimal values for the variational means, which would resemble the predictions of the \PCA prior mean \DGP 
at initialization, at some selected $\Xsel$ and $\Zsamples{}$, are given by 
\begin{align}
    \varm{}{} &= \braInv{\Kxsz{}\Kzzinv{}}\bra{\Xsel - \Kxsz{}\Kzzinv{}\Zsamples{}}\,, \tag{\NW{}}\\
    \varmw{} &= \braInv{\Kxsz{}\braT{\Lzzinv{}}}\Xsel\,, \tag{\W{}}\\
    \varmr{}{} &= \braInv{\Kxsz{}\Kzzinv{}}\Xsel\,, \tag{\NWR{}}
\end{align}
for the three parameterizations considered, assuming $\Kxsz{}\braT{\Lzzinv{}}$ and $\Kxsz{}\Kzzinv{}$ are invertible. If we additionally impose that the inducing points and $\Xsel$ are exactly the same points, \ie  $\Xsel = \Zsamples{} \coloneqq \Xxz$, for some set of inputs points $\Xxz$, the solutions become:
\begin{align}
	\varm{}{} &= \veczero\,, & 
	\varmw{} &= \Linv{\Xxz}{\Xxz}\Xxz\,,  &
    \varmr{}{} &= \Xxz \,.
	\label{equ:proposed_init_at_inducing}
\end{align}
This last assumption is used to remove many of the operations with cubic cost in the solutions.
\begin{remark}
For the \NW model, the optimal value for a \ZERO prior mean \DGP{} to mimic a \PCA prior mean \DGP at initialization is $\varm{}{}=\veczero$. However, this is exactly the initialization value we want to avoid. In other words, with the assumptions taken up to this point, the \ZERO prior mean \DGP{} with the standard initialization already mimics the \PCA prior mean \DGP{} at the inducing points values. However, in this work, we consider the non-whitened parameterization implemented by \GPFLOW (\NWR{}) and also give the necessary steps to develop initializations for the $\NW{}$ model implemented by \GPYTORCH and \GPJAX. These can be found in Appendix \ref{sec:app:B}.
\end{remark}

\subsection{Arbitrary Prior Mean Functions}
\label{subsec:proposed_technique_arbitrary_mean_funs}

The presented derivations consider a \ZERO prior mean \DGP{} that mimics a \PCA prior mean \DGP. However, our results can easily be extended 
for \DGP{}s with arbitrary mean functions to mimic a  \PCA prior mean \DGP. In particular, for an arbitrary prior mean function $\mu$, the means of the predictive distribution posteriors are:
\begin{align}
    \qfm{}{} &= \Kxz{}\Kzzinv{}\pare{\varm{}{}-\muZ{}} + \muX{}\,,\\
    \qfmw{}{} &=  \Kxz{}\braT{\Lzzinv{}}\varmw{} + \muX{}\,,\\
    \qfmr{}{} &= \Kxz{}\Kzzinv{}\varmr{} + \muX{}\,,
\end{align}
for each parameterization, respectively. Therefore, the systems of equations to solve are:
\begin{align}
  \Kxz{}\Kzzinv{}\pare{\varm{}{}-\muZ{}} + \muX{} &= \Xsel -  \Kxsz{}\Kzzinv{}\Zsamples{}\,, \\
  \Kxz{}\braT{\Lzzinv{}}\varmw{} + \muX{} &= \Xsel\,, \\
  \Kxz{}\Kzzinv{}\varmr{} + \muX{} &= \Xsel\,.
\end{align}
Given the same assumptions we have taken so far (fix values for $\nupos{}{}, \Xsel,\Zsamples{}$ and setting $\Xsel = \Zsamples{} = \Xxz$), the solutions are:
\begin{align}
	\varm{}{} &= \veczero\,,  &
	\varmw{} &= \Linv{\Xxz}{\Xxz}\pare{\Xxz-\mu_{\Xxz}}\,,  &
    \varmr{}{} &= \Xxz{} - \mu_{\Xxz}\,.
\end{align}

\begin{remark}
As a sanity check, note that if the prior PCA mean function is used, we obtain $\varm{}{}=\varmw{}{} = \varm{}{} = \veczero$, as expected, since $\mu_{\Xxz} = \Xxz$. Since both models are the same, they should have exactly the same parameters at initialization if one wants to mimic the other. 
\end{remark}

\subsection{Fixing $\Xxz$ and Initializing the Output Layer To Predict the Targets}
\label{sec:subsec:datapoints:Z:selection}

Above, we assumed that $\Xsel = \Zsamples{} \coloneqq \Xxz$, for some set of input points $\Xxz$, to reduce the computational cost of the approach.
There are two ways in which we can initialize $\Xxz$:

\begin{enumerate}
    \item Initialize the inducing points and let $ \Xxz = \Zsamples{} \coloneqq \Xz$. In this way, our initialization will resemble the predictive mean of the \PCA prior mean \DGP{} at the inducing locations $\Zsamples{}$, which are usually initialized via \emph{K-means}. The inducing points obtained will be representative of the training data, and will guarantee that we match the initial predictions on a representative set.
    \item Select $\Xxz = \Zsamples{} \subset \Xsamples{} \coloneqq \Xx$ to be representative of the training data. This means that the inducing points are a subset of the training inputs.
\end{enumerate}
Note that the solutions to the initialization problem for both options are given by:
\begin{align}
	\varm{}{} &= \veczero\,, & 
	\varmw{} &= \Linv{\Zsamples{}}{\Zsamples{}}\Zsamples{}\,,  &
    \varmr{}{} &= \Zsamples{}\,,
    \label{equ:m0_init}
\end{align}
where $\Zsamples{}$ is obtained either through k-means or as a subset of the training data.

The second option, however, has the advantage that it allows us to initialize the output \GP of the \DGP so that it predicts values that are closer to the corresponding targets associated with each selected point. Usually, a \ZERO prior mean function is used at the output layer, with $\varm{}{}=\varmr{}{}=\varmw{}=\veczero$. However, by using a similar reasoning to that used in \usec~\ref{sec:subsec:marginal:variational:posterior:init}, we can initialize the output  \GP{} so that it has a predictive mean equal to the corresponding target values at some of the training points. This is expected to start the optimization process at a good initial solution and to help avoid the posterior collapse problem. We can think of this approach as mimicking a \DGP{} model whose predictive mean function takes as input $\Xsamples$ and outputs the corresponding $\mathbf{y}$. Initializing the output layer like this is similar to what we have described so far. One only has to replace $\Xsel$ by the corresponding $\mathbf{y_s}$. This results in the following systems of equations:
\begin{align}
	\Kxz{}\Kzzinv{}\varm{}{} &= \mathbf{y_s}-  \Kxsz{}\Kzzinv{}\mathbf{y_z}\,, \\
	\Kxz{}\braT{\Lzzinv{}}\varmw{} &= \mathbf{y_s}\,, \\
	\Kxz{}\Kzzinv{}\varmr{} &= \mathbf{y_s}\,,
\end{align}
for each parameterization. Note that we require the targets, $\mathbf{y_s}$, associated with $\Xsel$. This is the reason why selecting $\Zsamples{} \subset \Xsamples{}$ (option $2$ above) becomes useful. By selecting the inducing points from the training data, we have the associated targets, something not possible in the first option. Using the same assumptions as before, namely, the inducing points and training points are the same locations, the solutions for this particular case are: 
\begin{align}
	\varm{}{} &= \veczero\,, &
	\varmw{} &= \Linv{\Xx}{\Xx}\mathbf{y_x}\,, &
	\varmr{}{} &= \mathbf{y_x}\,.
\label{eq:proposed:varm:init}
\end{align}
The fact that the inducing and training points are the same, and that the inducing points are selected from the training set, implies that $\mathbf{y_s}=\mathbf{y_z}=\mathbf{y_x}$ and $\Xxz=\Xx$.
\subsection{Initialization of the Inducing Points}
\label{subsec:init_inducing}

\begin{figure}[ht]
    \centering
    \begin{subfigure}{0.32\textwidth}
        \includegraphics[width=\textwidth]{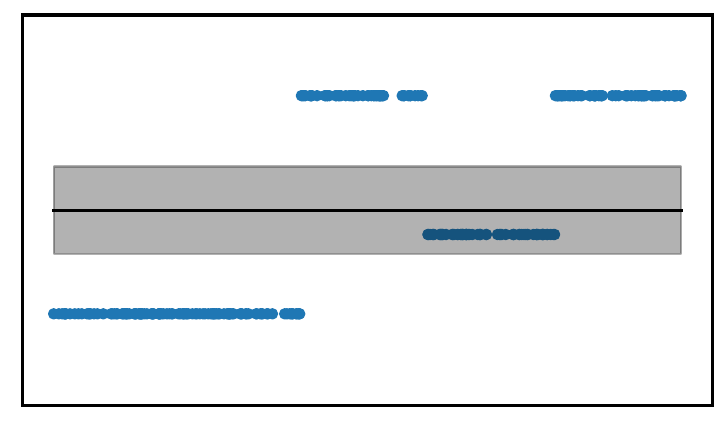}
        \caption{Standard initialization:\\ \(\varm{l}{} = \mathbf{0}, \varS{l}{} = 10^{-5}\mathbf{I}\).}
    \end{subfigure}
    \hfill
\begin{subfigure}{0.32\textwidth}
        \includegraphics[width=\textwidth]{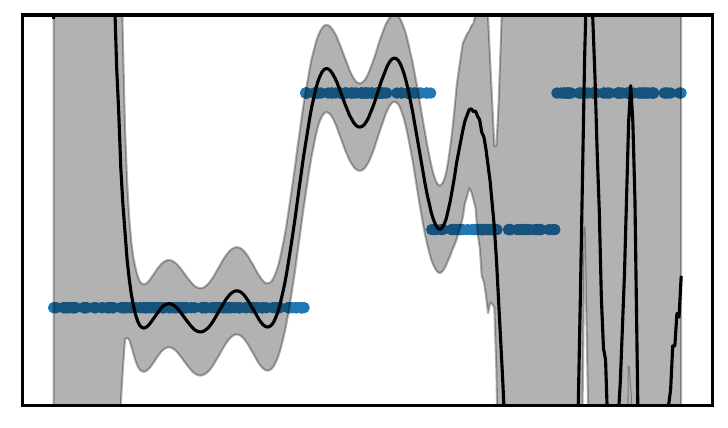}
        \caption{Proposed initialization\\ using random $\Zsamples{}\subset\Xsamples{}$.}
    \end{subfigure}
    \hfill
\begin{subfigure}{0.32\textwidth}
        \includegraphics[width=\textwidth]{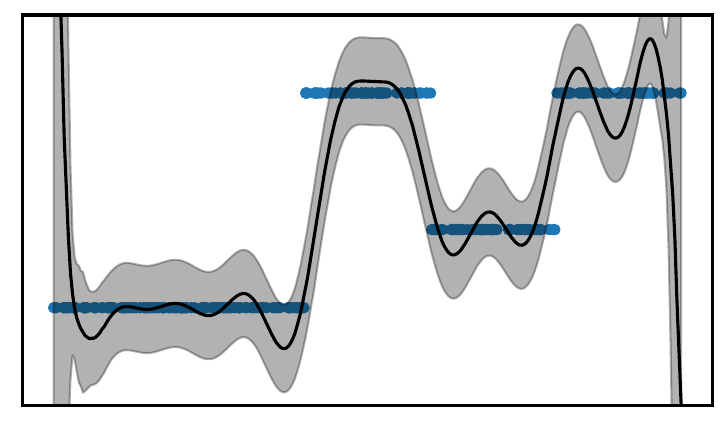}
        \caption{Proposed initialization selecting $\Zsamples{}\subset\Xsamples{}$      using Eq. \eqref{eq:cosine:inducing}.}
    \end{subfigure}
    \caption{Initial predictions of whitened 2-layer \DGP{}s using the \ZERO prior mean. Each sub-figure considers a different initialization of the variational mean of the output layer.}
    \label{fig:toy:initializations}
\end{figure}

As mentioned earlier, our initialization requires a subset from the training data that is as representative as possible of the full support over where we are making predictions. Selecting a random subset of $M$ instances, $\Xsamples{}$, from the training set may concentrate the inducing points in a specific region of the domain. Thus, to alleviate this problem, we propose the following initialization of the inducing points $\mathbf{Z}$:

\begin{enumerate}
\item We compute $M$ centroids $\mathcal{C} =\{\cent{1},\cdots, \cent{M}\}$ from the training data, \(\Xsamples_{\text{train}}\), using \emph{k-means}.
\item We select the inducing locations \(\Z{j}\) from \(\Xsamples_{\text{train}}\) using the cosine distance to each centroid:
\begin{align}
    \label{eq:cosine:inducing}
	    \Z{j} &= \underset{\X{}{j} \in \Xsamples_{\text{train}}}{\argmin} d(\cent{j}, \X{}{j}) \nonumber \\
        &=  \underset{\X{}{j} \in \Xsamples_{\text{train}}}{\argmin} \cent{j} \cdot \X{}{j} / (\norm{\cent{j}} \norm{\X{}{j}}), \quad \forall \cent{j} \in \mathcal C\,.
\end{align}
When a training point is selected as the closest one to a centroid, we remove it from the training set so that it is not selected again. 
In this way, all the inducing points differ from each other, but they are the training instances closest to the generated centroids.
\end{enumerate}
\fig \ref{fig:toy:initializations} shows the predictive distribution of a \DGP with 2-layers with the \ZERO{} prior mean. 
The whitened parameterization is used. In all cases, the variational covariances are initialized at $10^{-5}\matI$.
We consider three scenarios: (a) when the variational mean is initialized to zero; (b) when the proposed initialization is used
for the variational mean at each layer, but the inducing points are randomly chosen from the training set; and (c) when the proposed initialization is used, 
but the inducing points are selected using the algorithm described above.  We observe that the last scenario is the one producing a more accurate initial predictive distribution that best matches the observed data.

In the following sections, we will label as \ZERONWRMO and \ZEROWMO the models that use 
$\varm{}{}=\veczero$ in the output layer, but use the proposed initialization of the inner layers given in \ueqn (\ref{equ:m0_init}),
for the non-whitened and the whitened parameterizations, respectively. Similarly, we will label 
with \ZERONWRMY and \ZEROWMY to the model that initializes the predictive mean of the output layer so that the corresponding 
targets are predicted well, \ie using \ueqn (\ref{equ:m0_init}) in the inner layers and \ueqn (\ref{eq:proposed:varm:init}) in the output layer. In all cases, we consider the non-whitened parameterization of \GPFLOW.
\begin{figure}[!b]
    \centering
    \begin{subfigure}[t]{0.32\linewidth}
        \centering
        \footnotesize{$\ell = 0.01$} \\[0.3em]
        \includegraphics[width=\linewidth]{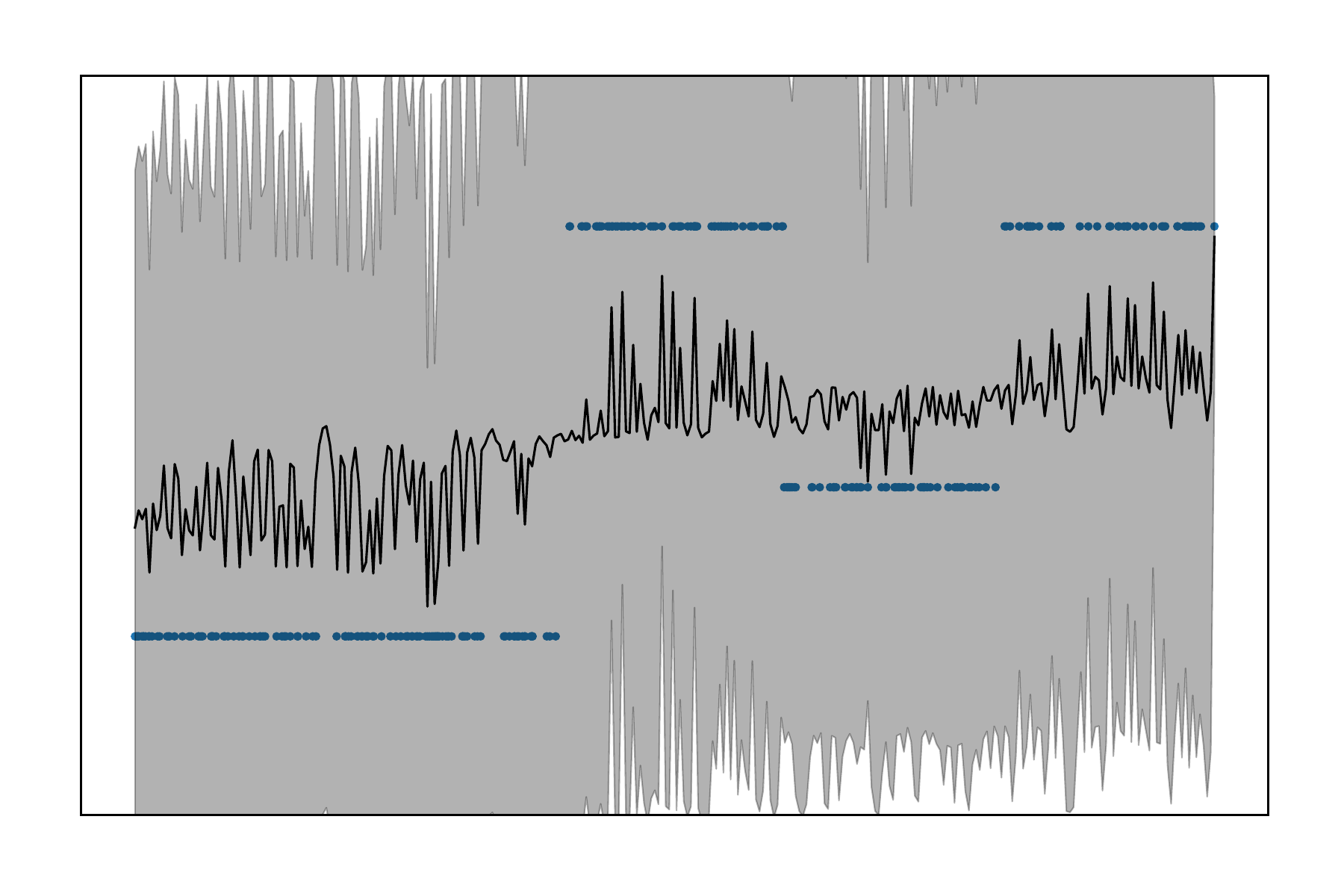}
        \label{fig:init:my:ls:0.01}
    \end{subfigure}
    \hfill
    \begin{subfigure}[t]{0.32\linewidth}
        \centering
        \footnotesize{$\ell=0.1005$} \\[0.3em]
        \includegraphics[width=\linewidth]{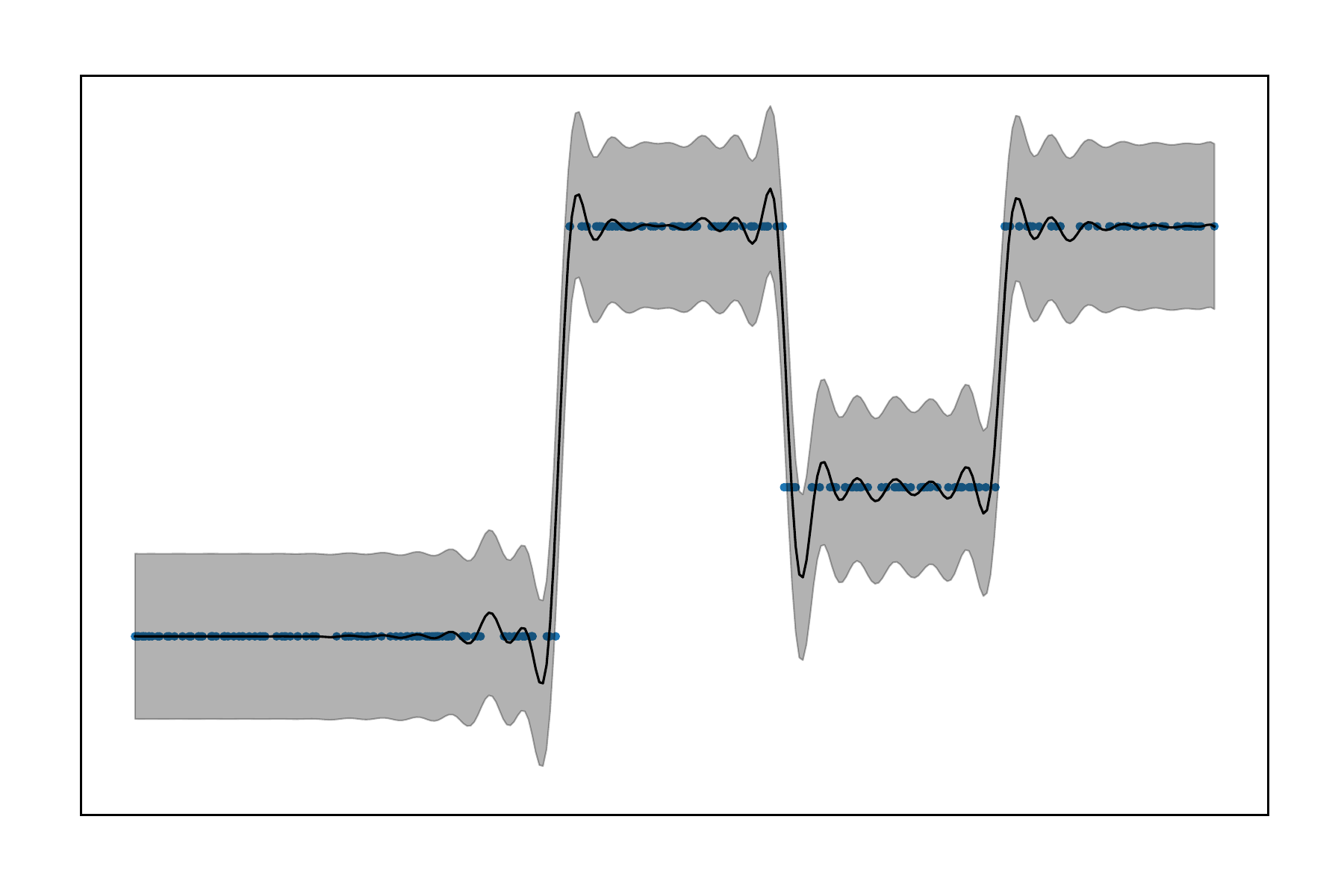}
        \label{fig:init:my:ls:0.1005}

    \end{subfigure}
    \hfill
    \begin{subfigure}[t]{0.32\linewidth}
        \centering
        \footnotesize{$\ell=1.0$} \\[0.3em]
        \includegraphics[width=\linewidth]{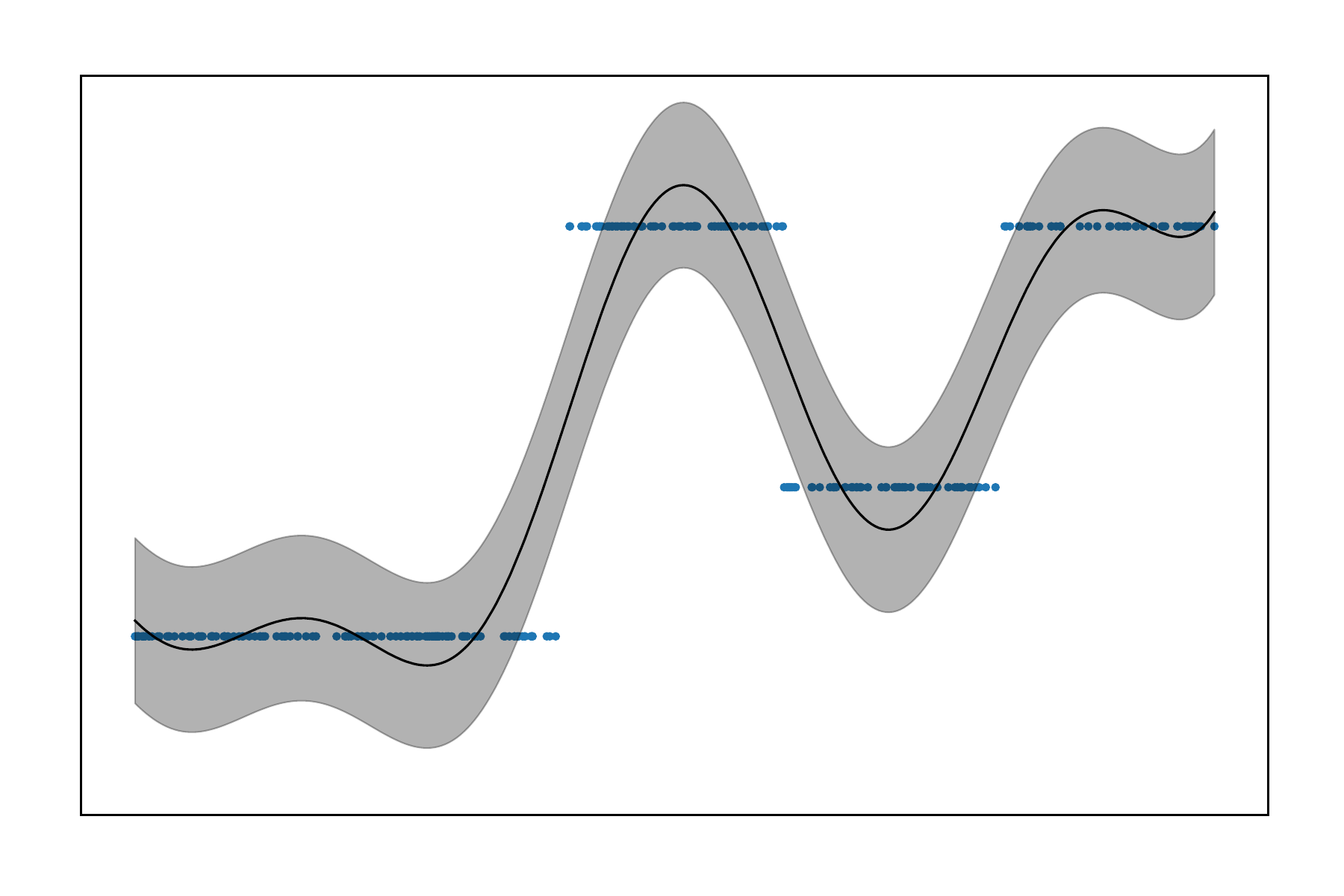}
        \label{fig:init:my:ls:1.0}
    \end{subfigure}

    \caption{Initial predictive distributions of the \ZEROWMY{} model using different initial length-scale $\ell$ values.}
    \label{fig:init:three:ls:my}
\end{figure}
\subsection{Selecting the Initial Kernel Length-scale $\ell$}
\label{sec:initialization:inital:lengthscale}

The proposed initialization for the variational mean aims at predicting the observed target values. Therefore, it becomes relevant to carefully set the initial kernel length-scale, $\ell$, for the particular given dataset in the case of using the initialization in the output layer (\MY models). In particular, in such a case, the quality of the initial predictive distribution may strongly depend on the length-scale, as illustrated by \fig \ref{fig:init:three:ls:my}. A heuristic used to initialize $\ell$ may guarantee that the initial predictive distribution on the observed data is good. With this goal, given a training set, we split it into training and validation, and compute the median Euclidean distance, $m_d$, between the datapoints in the training set. Then, we create a grid $\mathcal{G}$ of evenly-spaced length-scales with the middle point equal to $m_d$. For each length-scale in $\mathcal G$, we initialize the model using that length-scale and evaluate \RMSE{} of the corresponding initial distribution on the validation set. We finally select the length-scale $\ell^\star \in \mathcal G$ with the lowest \RMSE{} value. In practice, however, we have found that on some datasets, $m_d$ may vary a lot depending on the train-validation split. To avoid this issue, we repeat the selection in multiple train-validation splits and select the length-scale that provided the lowest \RMSE{} across different splits. The full algorithm is described in \ualg~\ref{alg:lengthscale:selection}. 

\begin{algorithm}[!t]
\caption{Lengthscale $\ell$ Selection via \RMSE{} Minimization}
\begin{algorithmic}[1]
\Require Train Dataset $D$, number of trials $N_{\text{trials}}$, number of steps $S$,  step size $\delta$
\State \texttt{pairs}$\gets [\,]$
\For{$s = 0$ to $N_{\text{trials}} - 1$}
    \State $D^t_s, D^v_s, \gets$ \texttt{random\_partition}$(D)$ 
    \Comment{Randomly select $10\%$ of $D$ as validation}
    
    \State $d \gets \texttt{median\_pairwise\_distance}(D^t_s)$
    
    \State $\mathcal{G} \gets \{ d \pm i \cdot \delta \mid i = 1,\cdots, S\} \cup \{ d \}$
    \Comment{Generate candidate lengthscales}

    \State $\mathcal{G} \gets \{ \ell \in \mathcal{G} \mid \ell > 0 \}$


    \For{each $l \in \mathcal{G}$}
        \State $M_\ell \gets \texttt{initialize\_model}(D^t_s, \ell)$
        \Comment{Init model using lengthscale $\ell$}
        
        \State $r_\ell \gets \texttt{evaluate\_rmse}(M_\ell, D^v_s)$
        \Comment{Compute $\RMSE{}$ on $D_s$}
        
        \State Append ($\ell$, $r_\ell$) to \texttt{pairs}
    \EndFor

    
\EndFor\\
\Return $\ell^\star$ with lowest $r_{\ell^\star}$ in \texttt{pairs}
\end{algorithmic}
\label{alg:lengthscale:selection}
\end{algorithm}

Note that the proposed heuristic focuses on selecting $\ell$ with the lowest \RMSE{}, without considering the resulting \KLD{} at initialization. We noted that adjusting the length-scale to improve \RMSE{} hardness the \KLD. This comes from replacing $\varmr{}{},\varmw{}$ by $\Linv{\Xx,\Xx}{}\mathbf{y_x}$ and $\mathbf{y_x}$ in the quadratic terms $\varmr{T}\Kzzinv{}\varmr{}$ and $\varmw{T}\varmw{}$. \fig \ref{fig:kld:rmse:init:my} in the appendix provides a comparison between the \KLD value and the length-scale. Future work may consider improving this selection strategy by designing a new heuristic that also takes into account the initial \KLD{}, which may increase, depending on the chosen initial length-scale.

\subsection{Summary of the Proposed Initialization Strategy}
\label{sec5_summary}
As a summary, the set of steps taken to yield the initial variational parameters are:
\begin{enumerate}
	\item Find $\Zsamples{}$ via the procedure described by \ueqn \eqref{eq:cosine:inducing} for the \MY models. The \MO models can also be initialized through \emph{k-means}. 
	\item Set $\Xsel = \Zsamples{} = \Xx$.
	\item In the case of initializing the output layer to predict the targets, heuristically determine the length-scale. Otherwise, use the default length-scale.
	\item Set the optimal variational mean by \ueqn \eqref{equ:proposed_init_at_inducing}.
\end{enumerate}

The proposed initialization has the limitation of being able to mimic the predictive mean of the \PCA prior mean \DGP, at the inner layers, only at a subset of the training points, given by the inducing points. Increasing the support of the initialization can be done with the procedure described in Appendix \ref{sec:app:B}, but results in more complex systems of equations. \emph{Importantly}, our initialization should only improve the results substantially, in those \ZERO mean models that suffer from posterior collapse. Our initialization does not change the statistical model, nor the approximate inference distribution. This implies that the space of possible solutions is the same across all the \ZERO models. Thus, we shall only expect either solving the posterior collapse problem (through either \MO or \MY initializations) or, given the same number of training iterations, faster convergence to a solution of the optimization problem (\MY initialization). As a consequence, the \MY initialized model may surpass the rest of the models.

\subsection{Dependent and Non-Gaussian Processes at Each Layer}

This work considers \DGP{} where the \GP{}s at each layer are independent a priori. However, the proposed initialization strategy can work as well with models in 
which the \GP{}s at each layer are dependent or where the latent processes are transformed in some way, yielding non-Gaussian processes as building blocks.
Such modifications of the standard \DGP{} can be achieved either by using a mixing matrix \citep{multioutputGPs} or by considering warping functions
\citep{tgp,etgp,dtgp,multidimTGP}. Since these models use \GP{}s as building blocks, and act directly on the samples or distributions over the function evaluations, 
our initialization and analysis remain valid. We just need to proceed in the same way to mimic the \GP{} at initialization, and then possibly initialize the 
next elements of the modeling processes in a way that at initialization the behavior is the same as the standard \DGP{}. For this, we just need 
the mixing matrix to be initialized to the identity matrix, and the warping functions to be an identity function.

\section{Experiments}\label{sec:experiments}

In this section, we will detail the experiments performed to validate our theoretical observations and to test the proposed initialization. We have considered two scenarios for this validation: the first one is the toy dataset used in \cite{rudner2020inter}, which has already been used to analyze the optimization issues of \DGP{}s in \usec \ref{sec:optimization:difficulties}. The experiments performed on this dataset are useful to exhibit the pathologies of \DGP{}s supported by the visualizations of the predictive distribution. The second scenario involves real datasets from the \UCI{} repository \citep{Dua:2019:UCI}. In these experiments, we will check that the conclusions obtained in the toy dataset are extrapolated to real-world datasets.

The baseline models considered are \DGP{}s with the \ZERO prior mean, and the whitened and non-whitened parameterization of \GPFLOW. We refer to them as 
\ZEROW{}, \ZERONWR{}. Similarly, we consider equivalent \DGP models with the \PCA prior mean. We refer to them as  \PCAW and \PCANWR. See \utab~\ref{tab:acronyms}. 
Besides this, we also consider our proposal initializations, labeled as \ZEROWMO{}, \ZERONWRMO{}, \ZEROWMY{}, \ZERONWRMY{}, 
where \MO indicates that the output layer's variational mean is $\varmr{L}{}=\varm{}{L}=\veczero$ at initialization, and \MY indicates 
that we use our proposed initialization for $\varm{}{L}$ in the output layer to predict the observed targets. We note that the standard non-whitened parameterization is not implemented in \GPFLOW{}. Therefore, we do not evaluate it. The covariance function is, for all the models, parameterized by the \RBF{} kernel defined in \ueqn~\eqref{eq:kernel:rbf}.

\subsection{Toy Dataset}
\label{sec:experiments:toy}

As seen in \usec \ref{sec:optimization:difficulties}, the toy dataset of \cite{rudner2020inter}
describes a 1-dimensional regression problem with a multi-step function. This problem can be hard 
for \DGP{}s since they need to learn to use a very small length-scale in order to model the non-smooth steps of the target function. 
We utilize this dataset to analyze the optimization issues that arise within the \DSVI algorithm and different 
initialization parameters. We divide this section into two fundamental parts. The first aims at supporting the arguments provided 
in \usec~\ref{sec:optimization:difficulties} for the posterior collapse problem under the \DSVI{} algorithm. The second one shows 
a comparison of all the models in this dataset.

\subsubsection{An Illustration of the Posterior Collapse Problems of \DSVI}

We begin by validating the statements made in \usec~\ref{sec:Iat_output} and \usec~\ref{sec:opt_analysis_more_inducing}. We focus on the case of \ZEROW{} and \PCAW{}, standard \DGP{} models, as the whitened parameterization is expected to facilitate the training of the models. As a common setup, in this section, we train all the models for $10.000$ epochs using Adam and a learning rate of $10^{-3}$. We use $2,3$ and $5$ layer \DGP{}s. We analyze the effect of the different initializations for $\varS{}{}$ and its relation with the number of inducing points. $\varS{}{l}$ will denote the variational covariance of the inner layers, and $\varS{}{L}$ of the output layer.

\begin{itemize}
\item $\varS{}{l}=10^{-5}\mathbf{I}$ and  $\varS{}{L} = 10^{-  5}\mathbf{I}$:
\fig~\ref{fig:more_inducings_all_1e-5} shows the predictive distribution of the \ZEROW{} and \PCAW{} models for each configuration. We observe that as the number of inducing points decreases and the number of layers increases, the posterior collapse problem of the \ZEROW{} model becomes more evident. Specifically, $5$ inducing points show total collapse with $3$ and $5$ layers. With $20$ inducing points, the model learns a poor predictive distribution when the \DGP has 3 layers, and results in posterior collapse when $5$ layers are considered. This effect is reduced when the number of inducing points is increased to $100$. In this case, there is no posterior collapse, but the predictive distribution is very poor in the deeper model. The \PCAW{} model is able to escape posterior collapse. These results corroborate the ideas presented in \usec~\ref{sec:opt_analysis_more_inducing}, indicating that more inducing points should alleviate the posterior collapse problem. Furthermore, they also illustrate that the \ZERO mean prior \DGP is more likely to have posterior collapse than the \PCA prior mean \DGP.
\begin{figure}[htbp]
\vspace{-0.5cm}
    \centering



    \begin{minipage}[c]{0.05\textwidth}
    \end{minipage}%
    \begin{minipage}[c]{0.93\textwidth}
        \centering
        \hfill
        \makebox[0.32\linewidth][c]{\textbf{2 Layers}}%
        \hfill
        \makebox[0.32\linewidth][c]{\textbf{3 Layers}}%
        \hfill
        \makebox[0.32\linewidth][c]{\textbf{5 Layers}}%
    \end{minipage}

\begin{minipage}[c]{0.05\textwidth}
    \centering
    \rotatebox{90}{\textbf{5 Inducing}}
\end{minipage}%
\begin{minipage}[c]{0.93\textwidth}
    \centering

    \begin{subfigure}[c]{0.32\linewidth}
        \includegraphics[width=\linewidth]{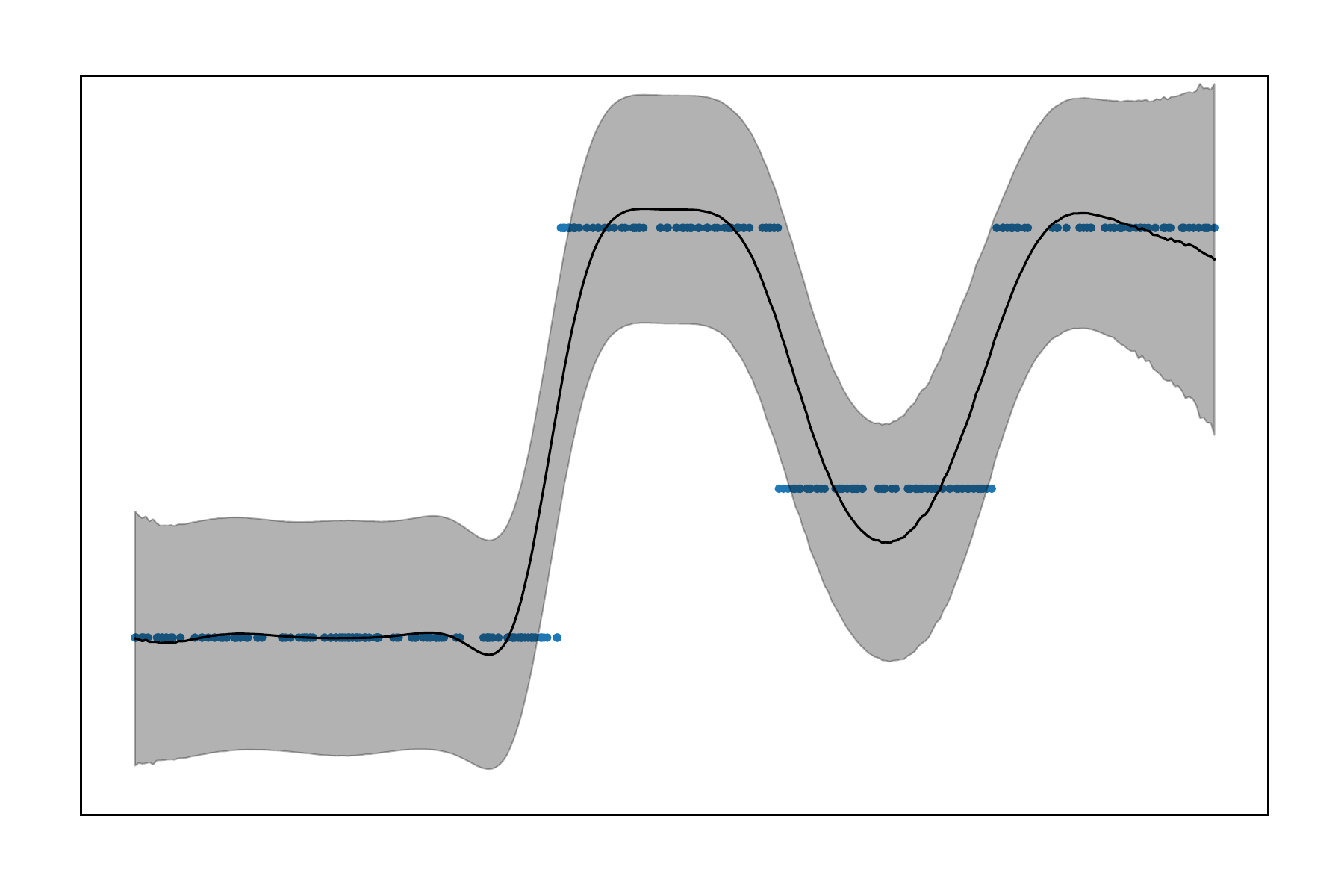}
    \end{subfigure}
    \hfill
    \begin{subfigure}[c]{0.32\linewidth}
        \includegraphics[width=\linewidth]{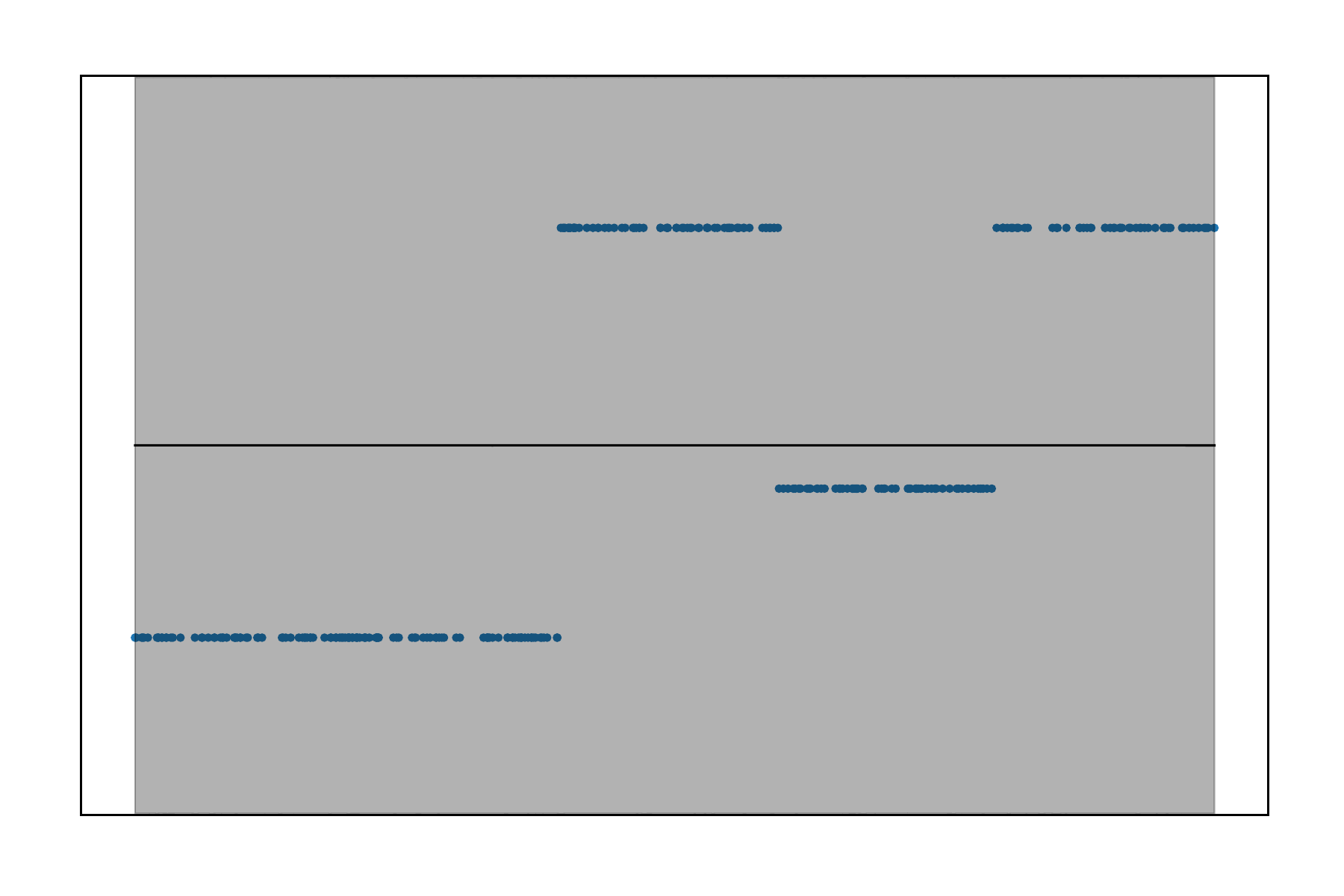}
    \end{subfigure}
    \hfill
    \begin{subfigure}[c]{0.32\linewidth}
        \includegraphics[width=\linewidth]{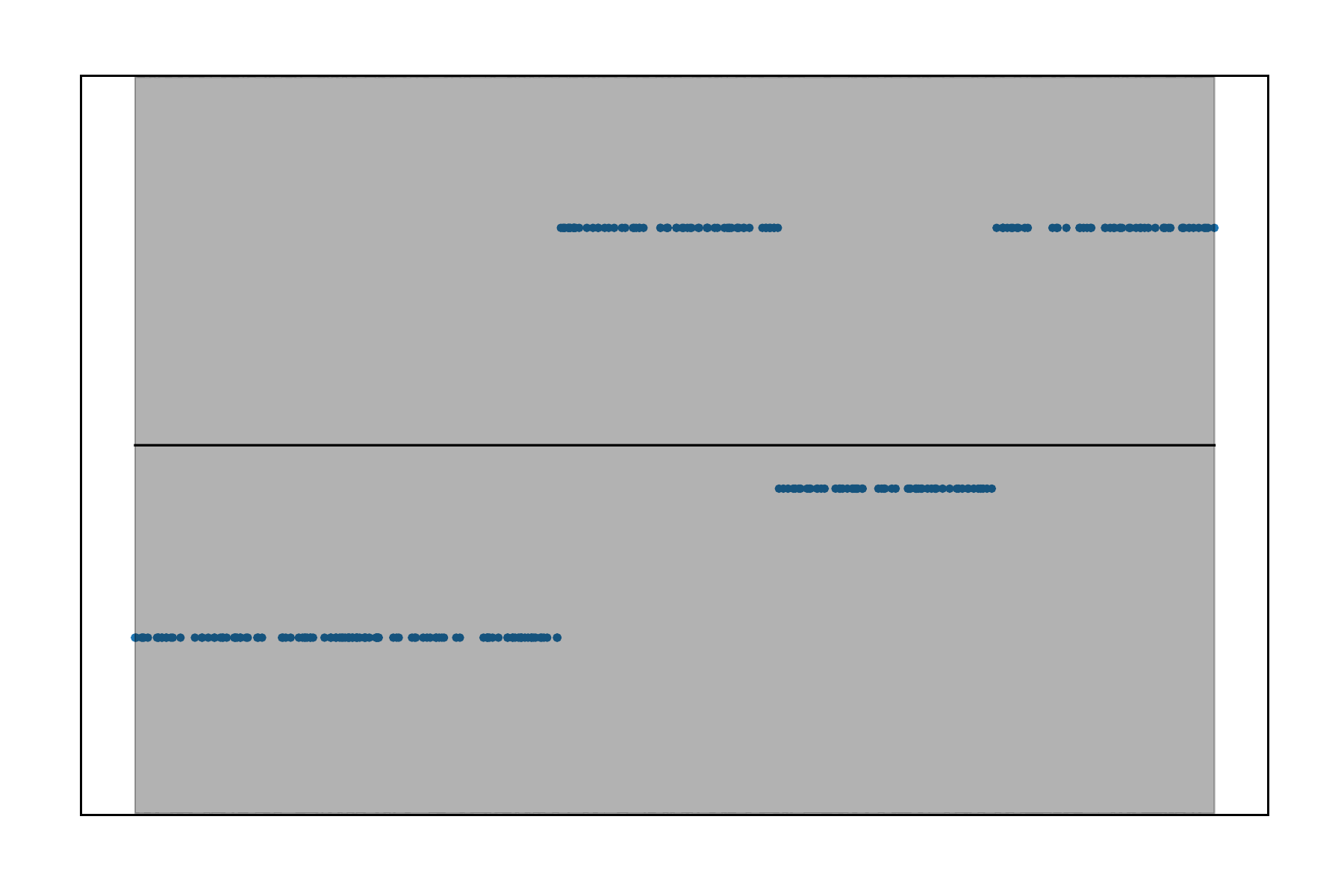}
    \end{subfigure}
\end{minipage}
    
    \begin{minipage}[c]{0.05\textwidth}
        \centering
        \rotatebox{90}{\textbf{20 Inducing}}
    \end{minipage}%
    \begin{minipage}[c]{0.93\textwidth}
        \centering
        \begin{subfigure}[c]{0.32\linewidth}
            \includegraphics[width=\linewidth]{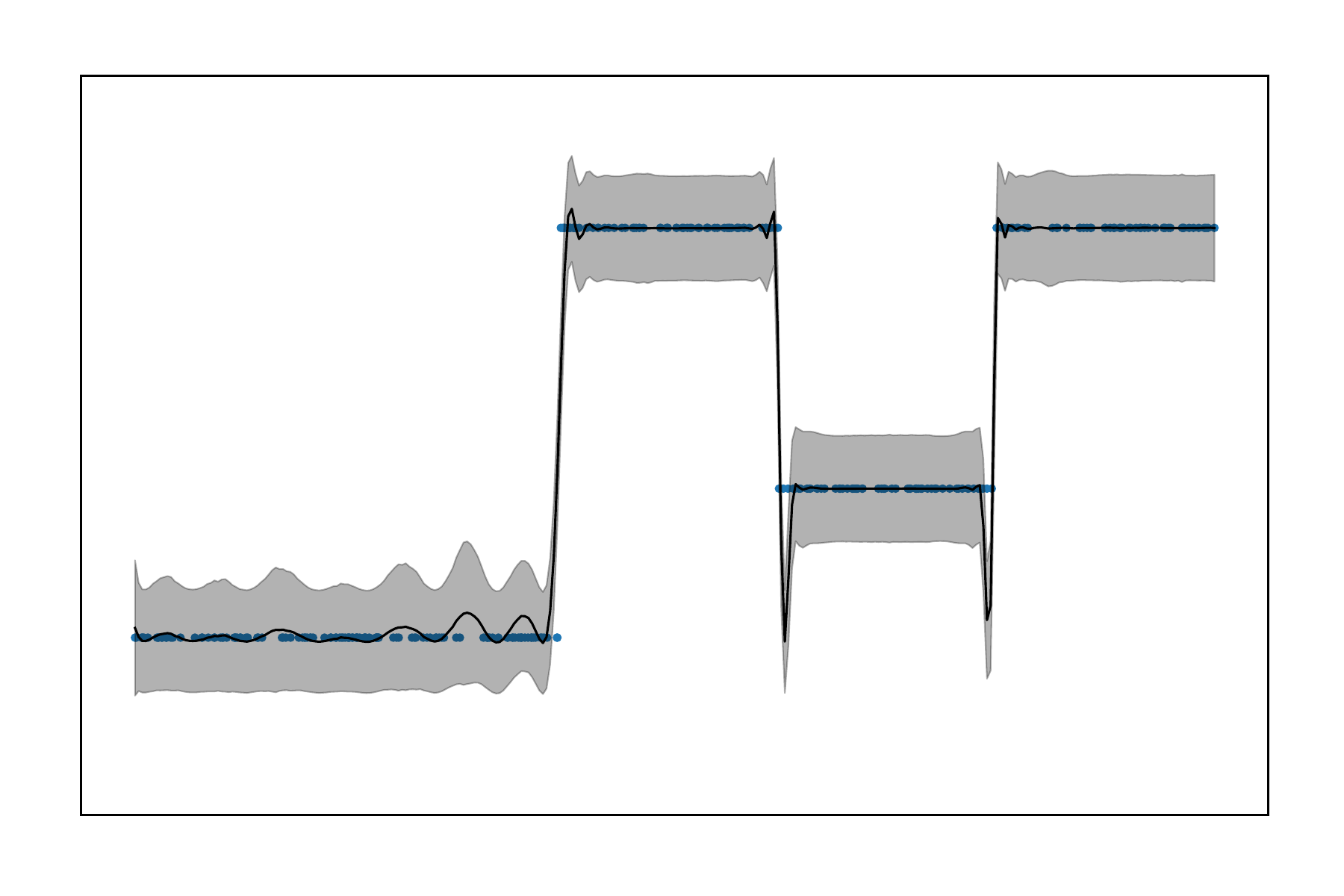}
        \end{subfigure}
        \hfill
        \begin{subfigure}[c]{0.32\linewidth}
            \includegraphics[width=\linewidth]{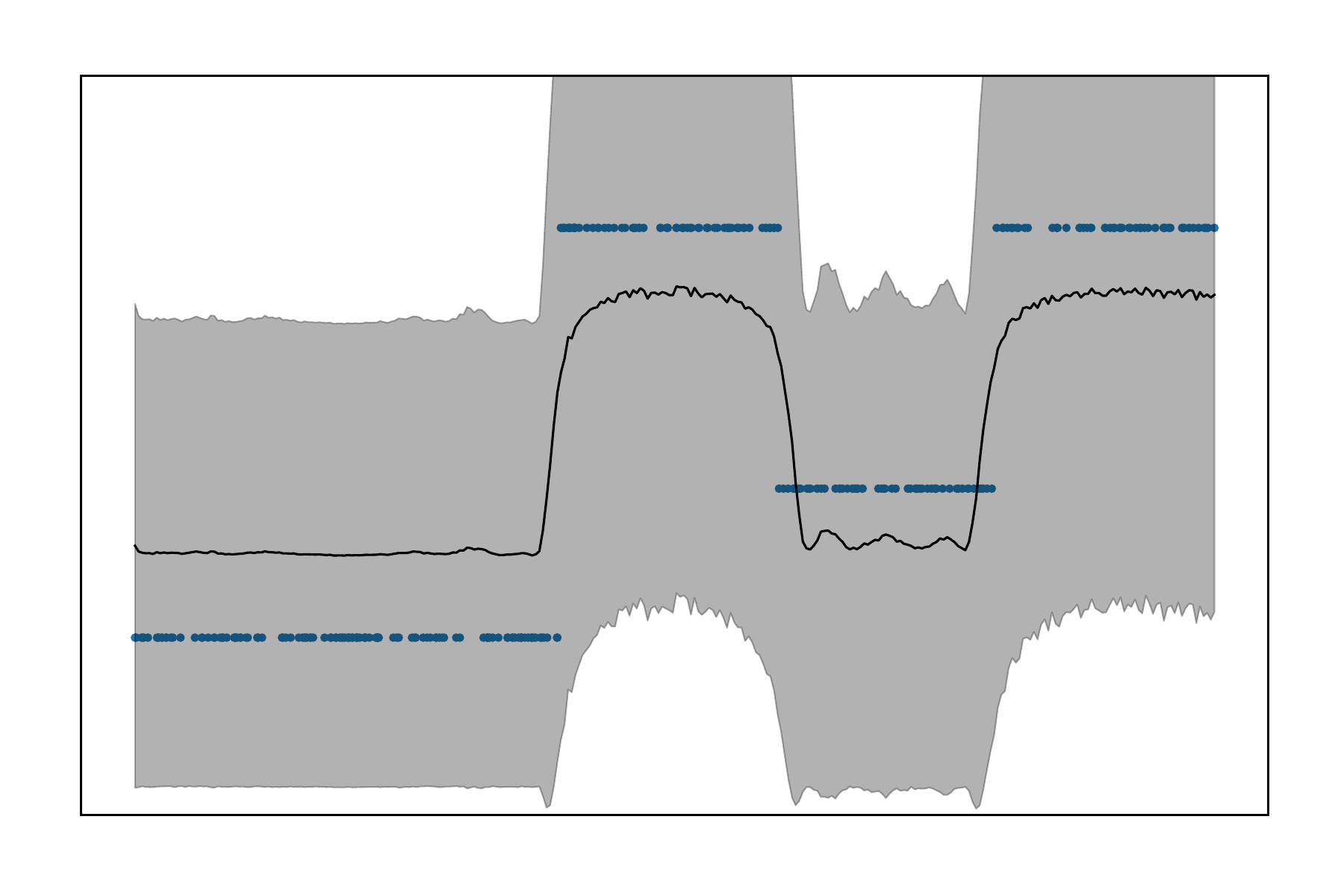}
        \end{subfigure}
        \hfill
        \begin{subfigure}[c]{0.32\linewidth}
            \includegraphics[width=\linewidth]{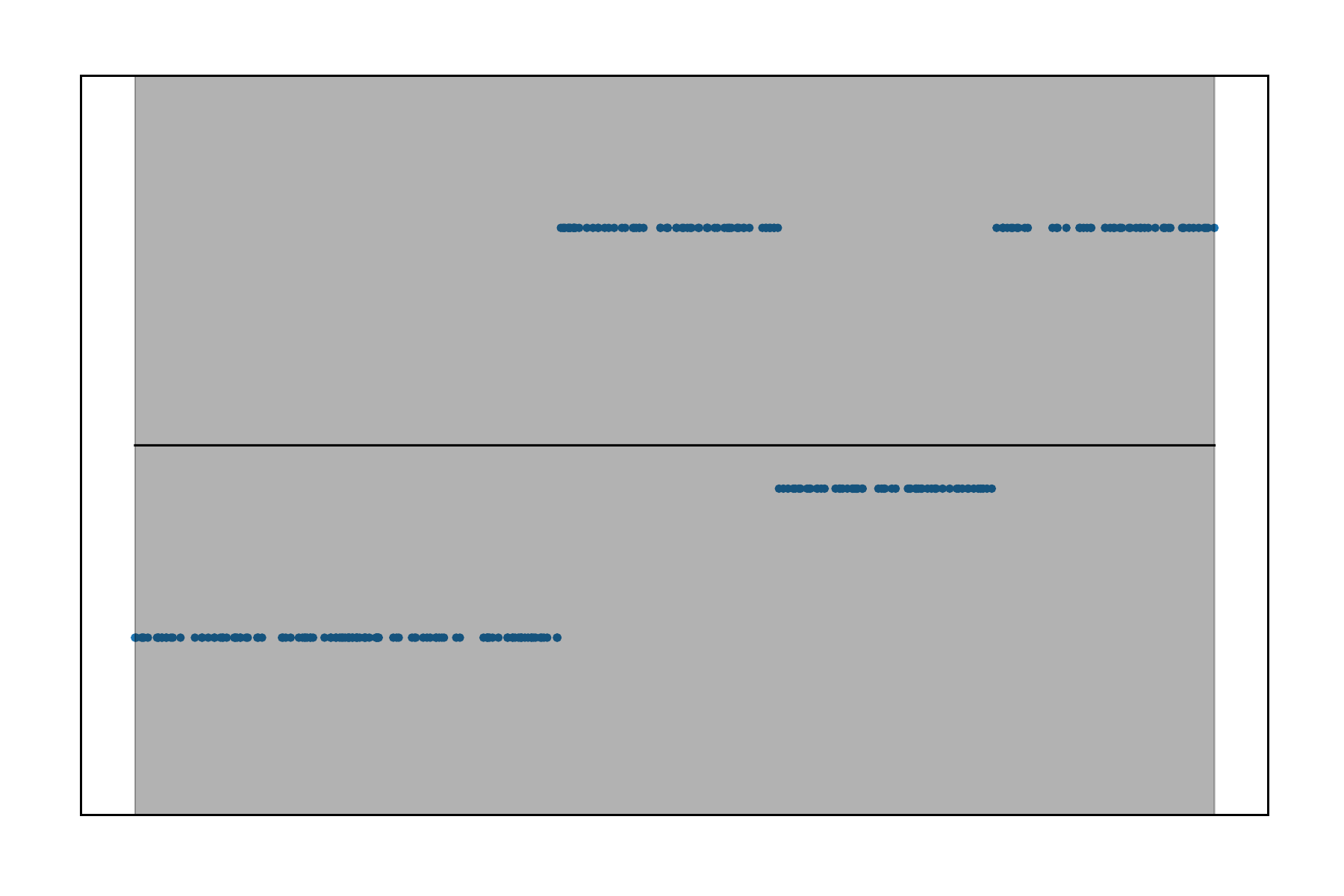}
        \end{subfigure}
    \end{minipage}

    \vspace{0.25cm} 

    \begin{minipage}[c]{0.05\textwidth}
        \centering
        \rotatebox{90}{\textbf{100 Inducing}}
    \end{minipage}%
    \begin{minipage}[c]{0.93\textwidth}
        \centering
        \begin{subfigure}[c]{0.32\linewidth}
            \includegraphics[width=\linewidth]{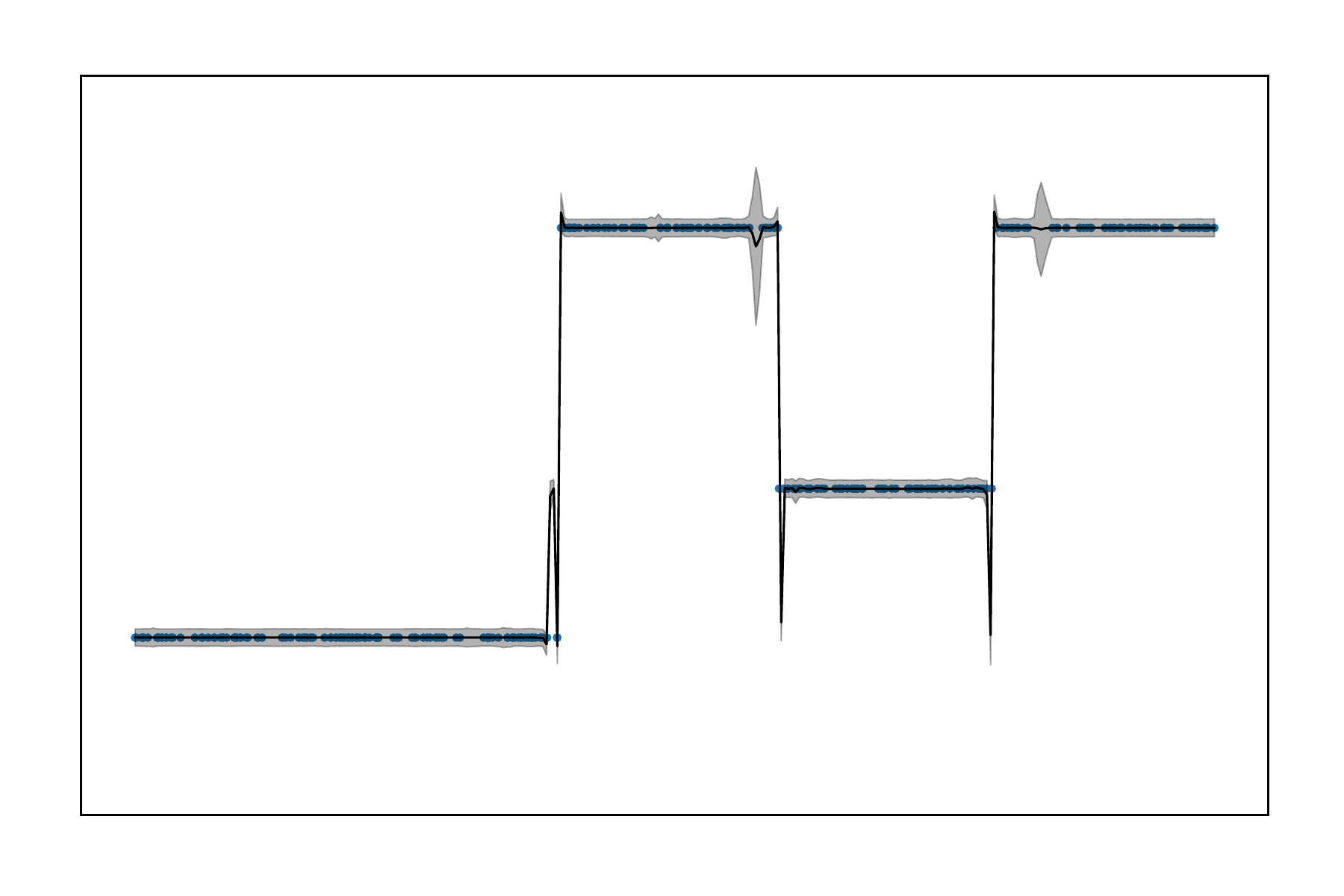}
        \end{subfigure}
        \hfill
        \begin{subfigure}[c]{0.32\linewidth}
            \includegraphics[width=\linewidth]{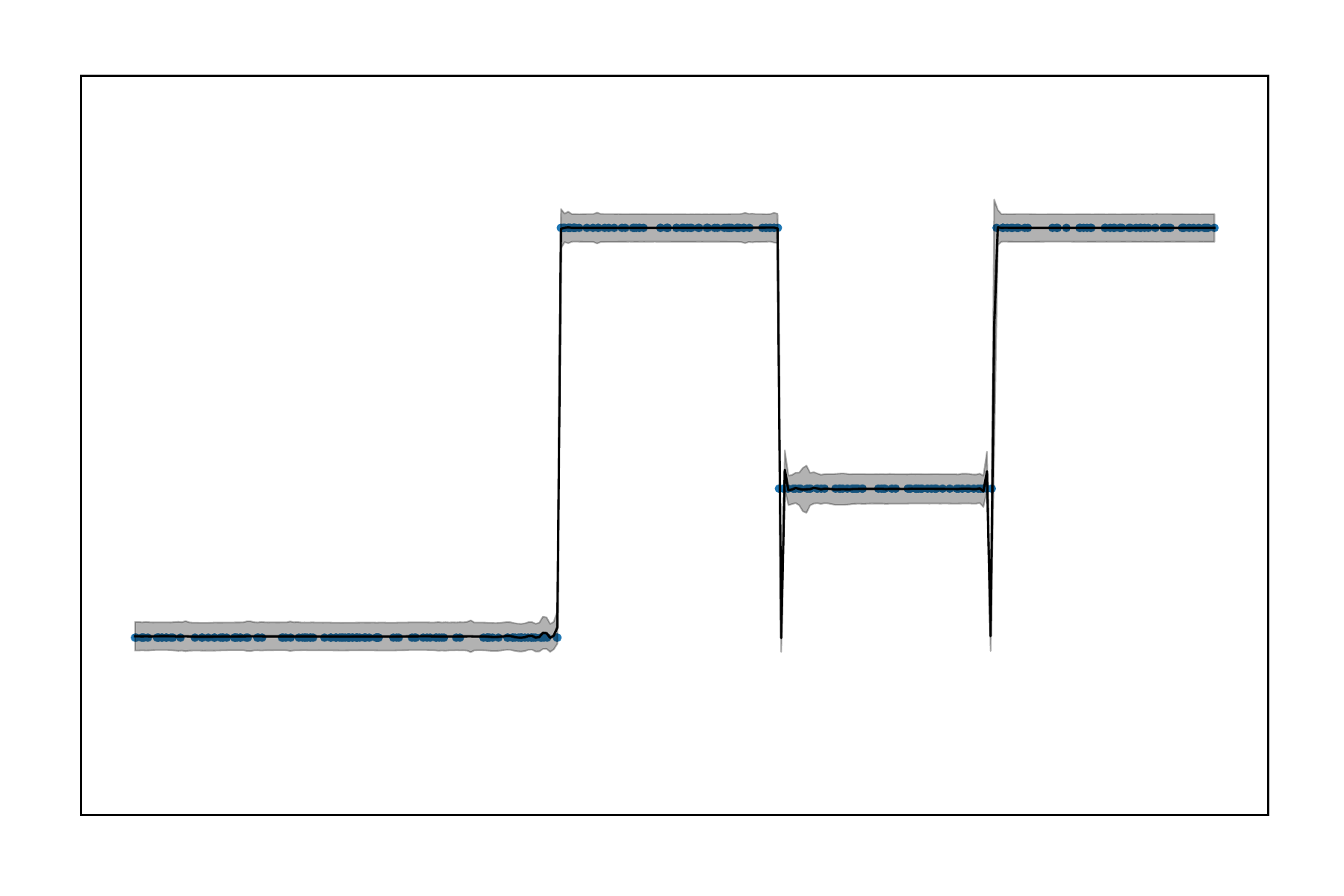}
        \end{subfigure}
        \hfill
        \begin{subfigure}[c]{0.32\linewidth}
            \includegraphics[width=\linewidth]{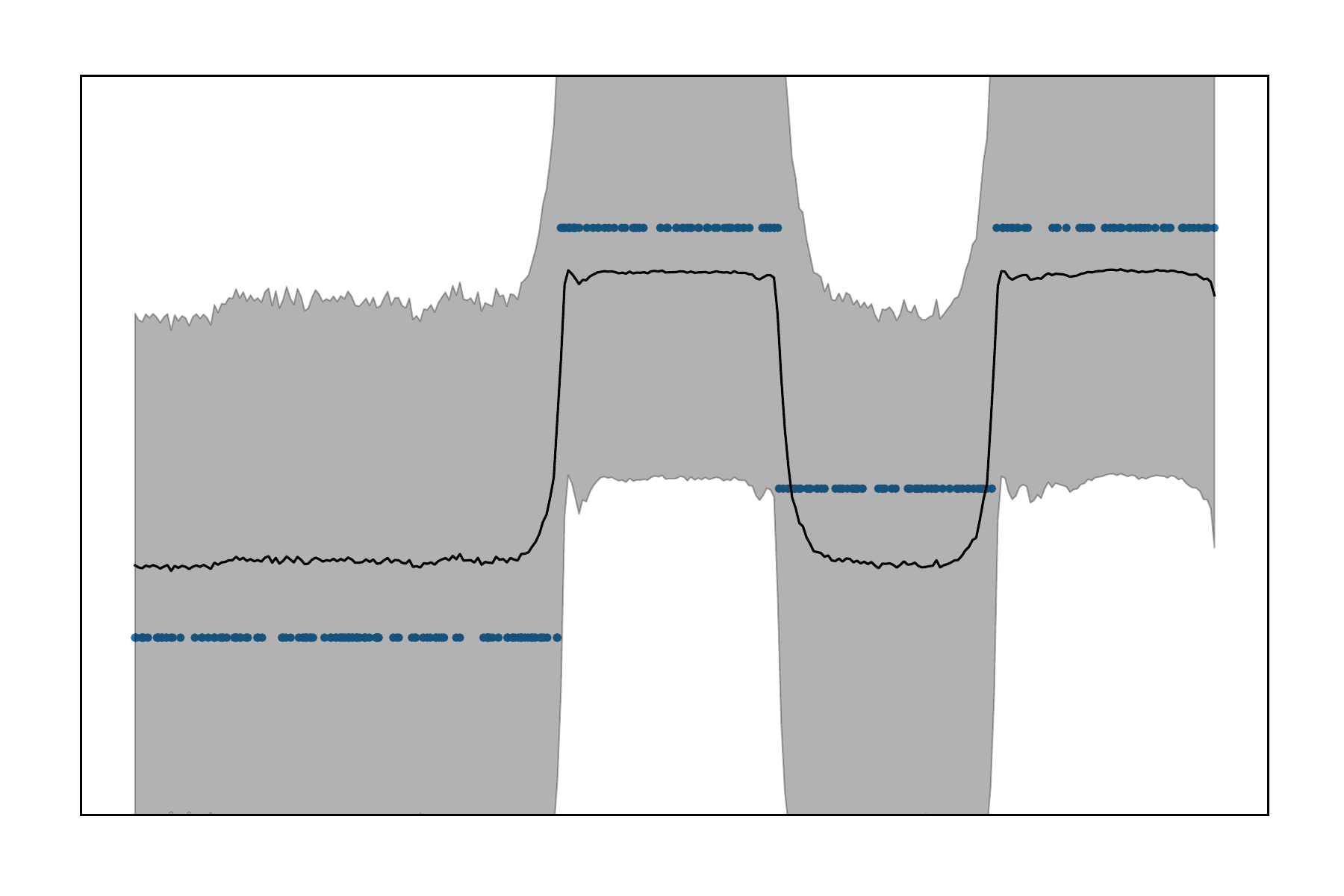}
        \end{subfigure}
    \end{minipage}
    \begin{minipage}[c]{\textwidth}
        \makebox[\linewidth][c]{\textbf{\ZEROW{}}}%
    \end{minipage}

    \begin{minipage}[c]{0.05\textwidth}
        \hfill 
    \end{minipage}%

\begin{minipage}[c]{0.05\textwidth}
    \centering
    \rotatebox{90}{\textbf{5 Inducing}}
\end{minipage}%
\begin{minipage}[c]{0.93\textwidth}
    \centering

    \begin{subfigure}[c]{0.32\linewidth}
        \includegraphics[width=\linewidth]{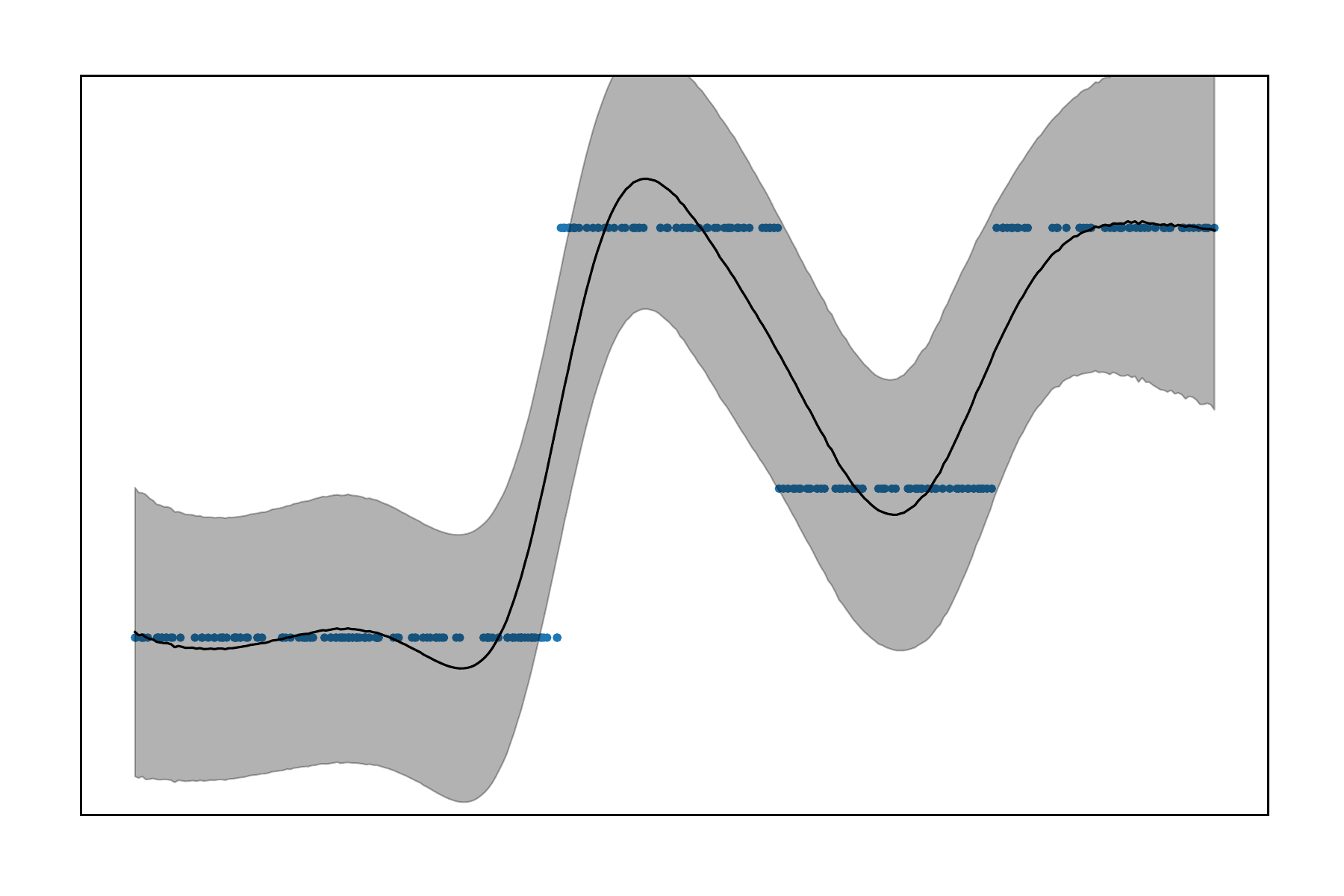}
    \end{subfigure}
    \hfill
    \begin{subfigure}[c]{0.32\linewidth}
        \includegraphics[width=\linewidth]{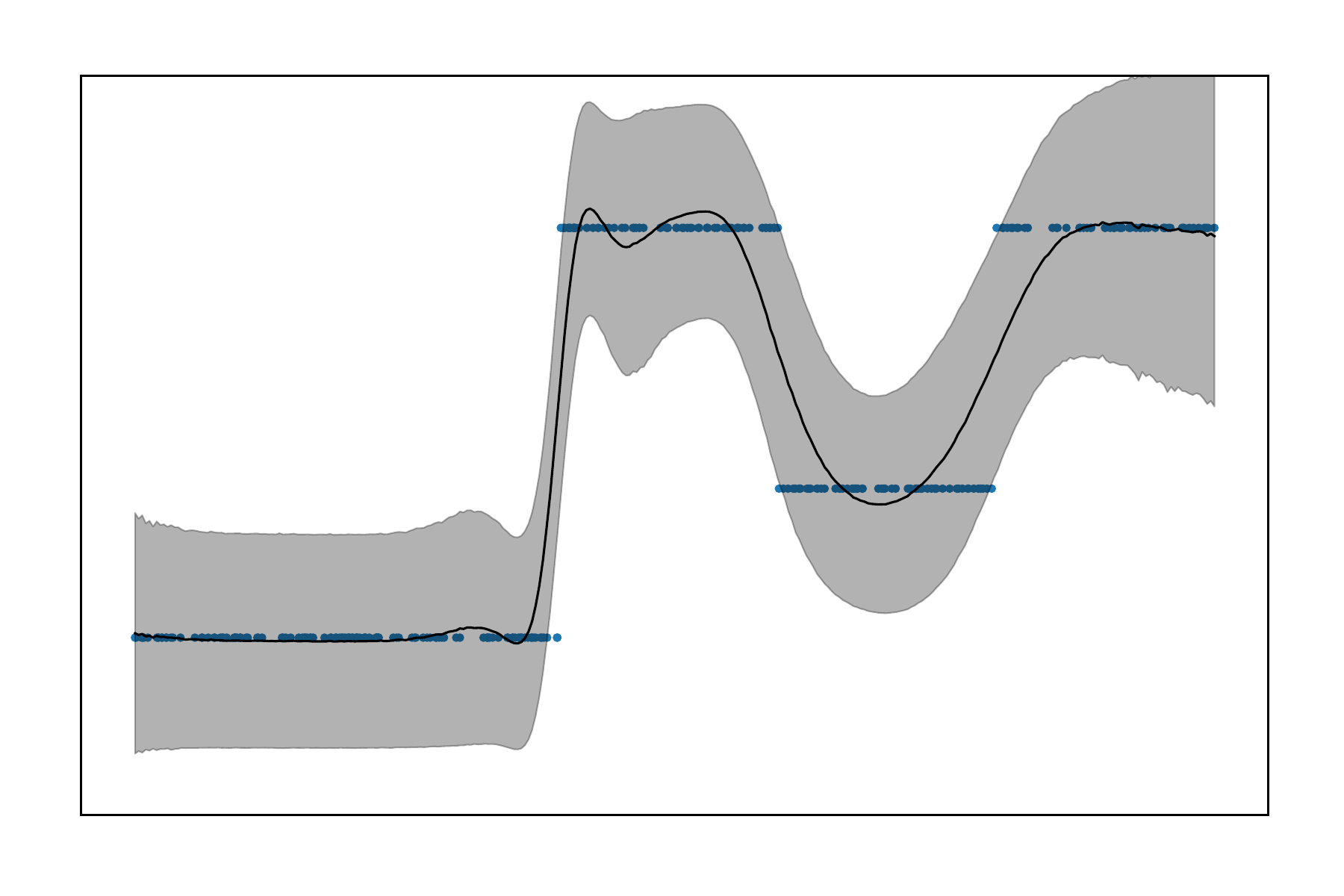}
    \end{subfigure}
    \hfill
    \begin{subfigure}[c]{0.32\linewidth}
        \includegraphics[width=\linewidth]{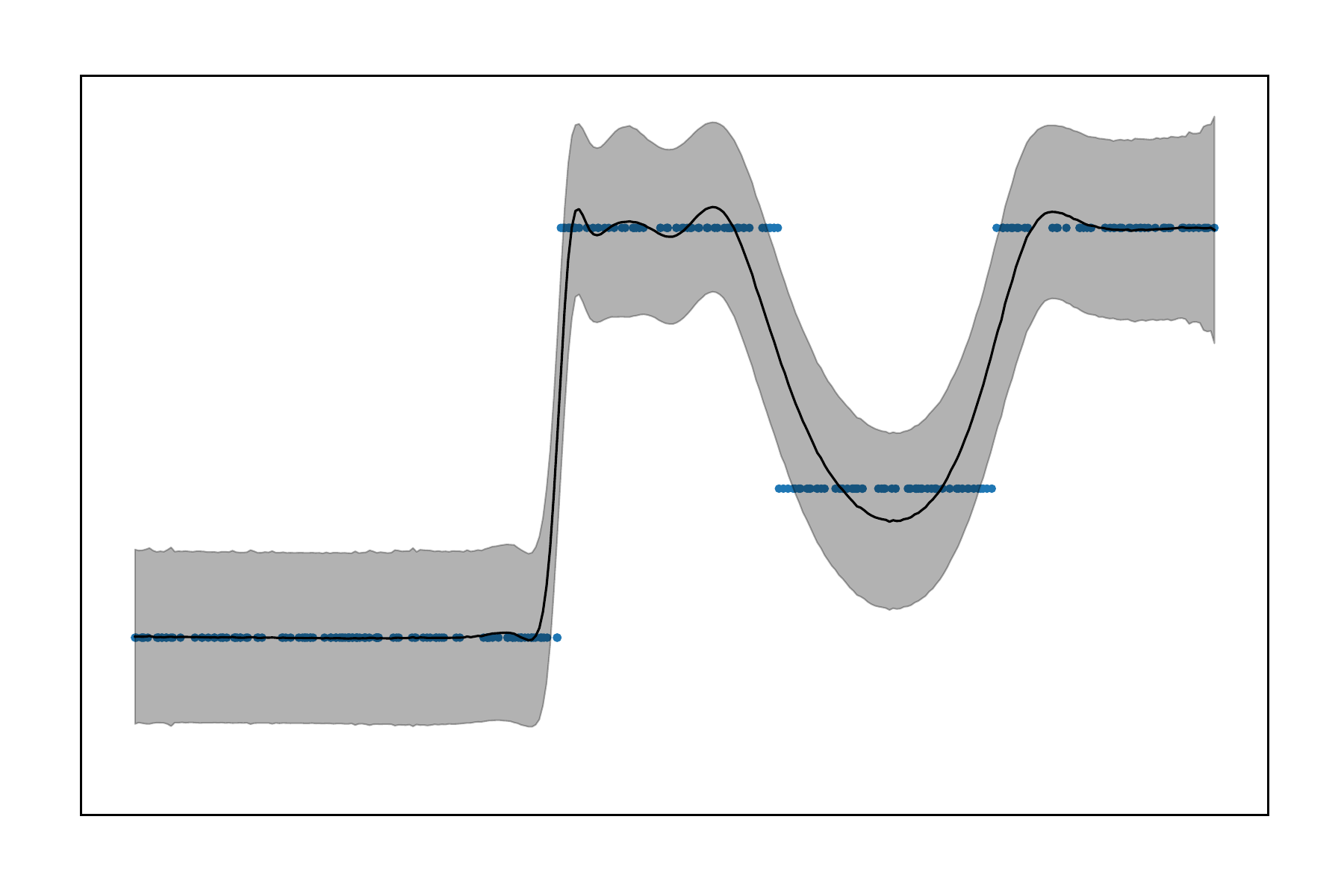}
    \end{subfigure}
\end{minipage}
    
    \begin{minipage}[c]{0.05\textwidth}
        \centering
        \rotatebox{90}{\textbf{20 Inducing}}
    \end{minipage}%
    \begin{minipage}[c]{0.93\textwidth}
        \centering
        \begin{subfigure}[c]{0.32\linewidth}
            \includegraphics[width=\linewidth]{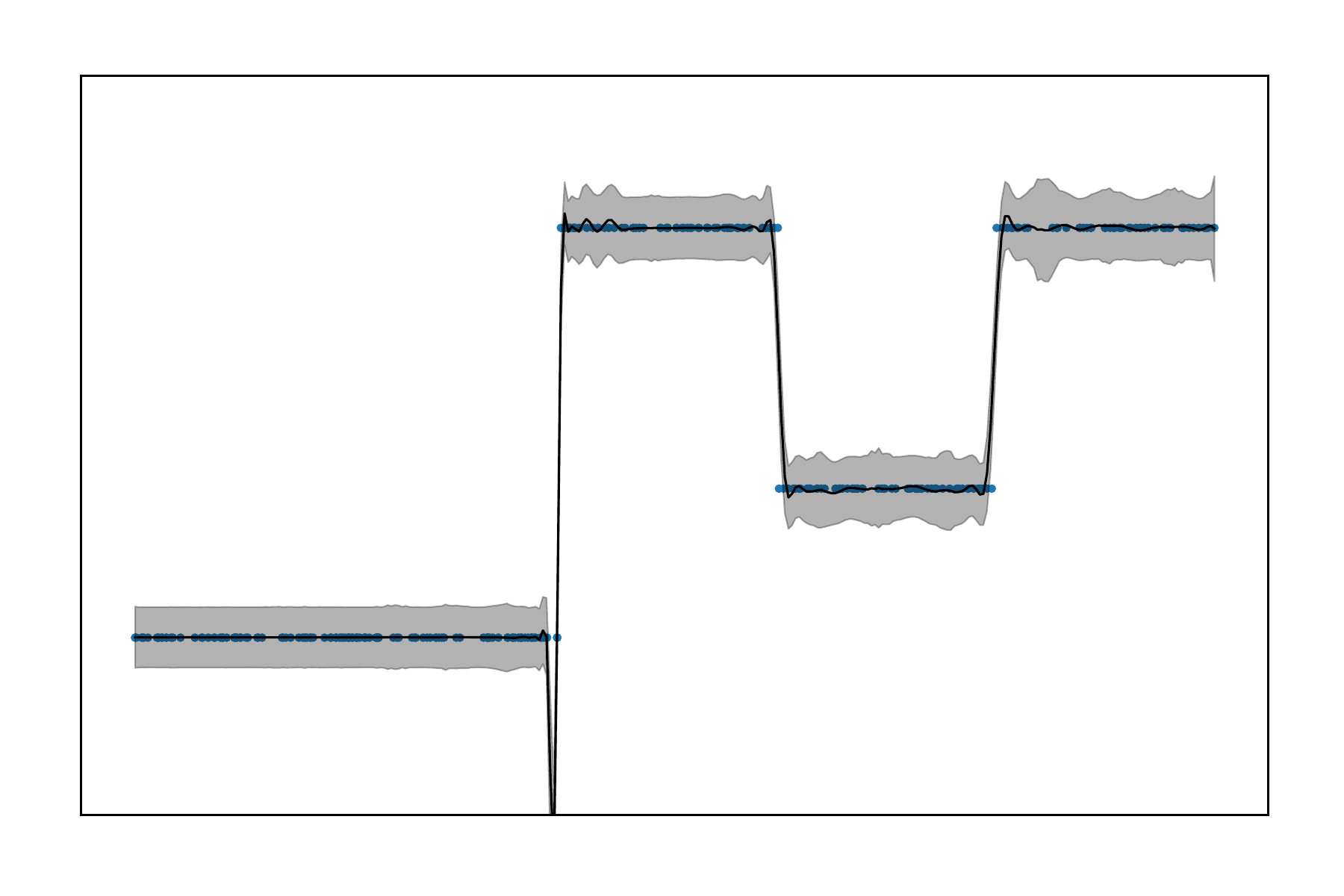}
        \end{subfigure}
        \hfill
        \begin{subfigure}[c]{0.32\linewidth}
            \includegraphics[width=\linewidth]{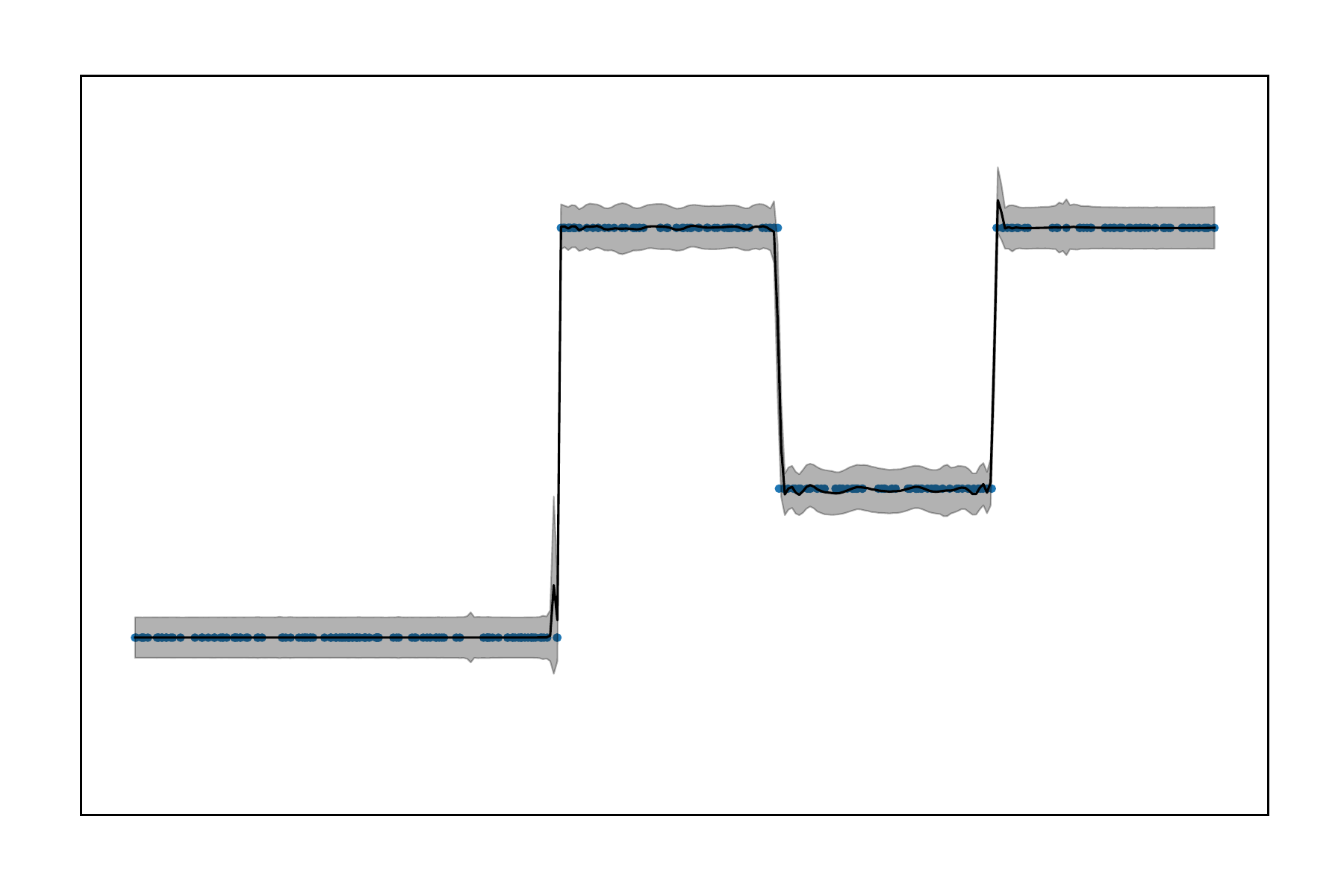}
        \end{subfigure}
        \hfill
        \begin{subfigure}[c]{0.32\linewidth}
            \includegraphics[width=\linewidth]{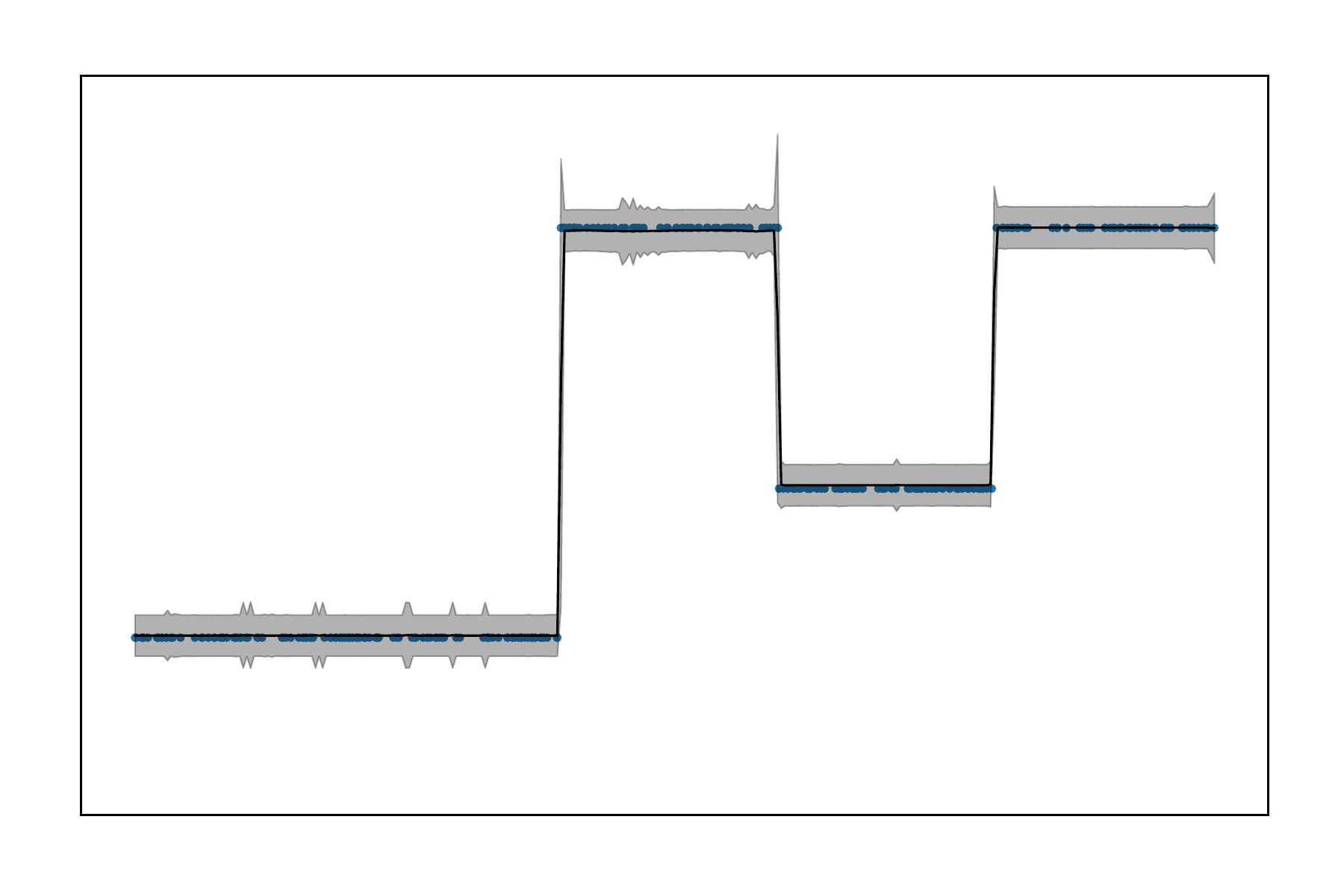}
        \end{subfigure}
    \end{minipage}

    \vspace{0.25cm} 

    \begin{minipage}[c]{0.05\textwidth}
        \centering
        \rotatebox{90}{\textbf{100 Inducing}}
    \end{minipage}%
    \begin{minipage}[c]{0.93\textwidth}
        \centering
        \begin{subfigure}[c]{0.32\linewidth}
            \includegraphics[width=\linewidth]{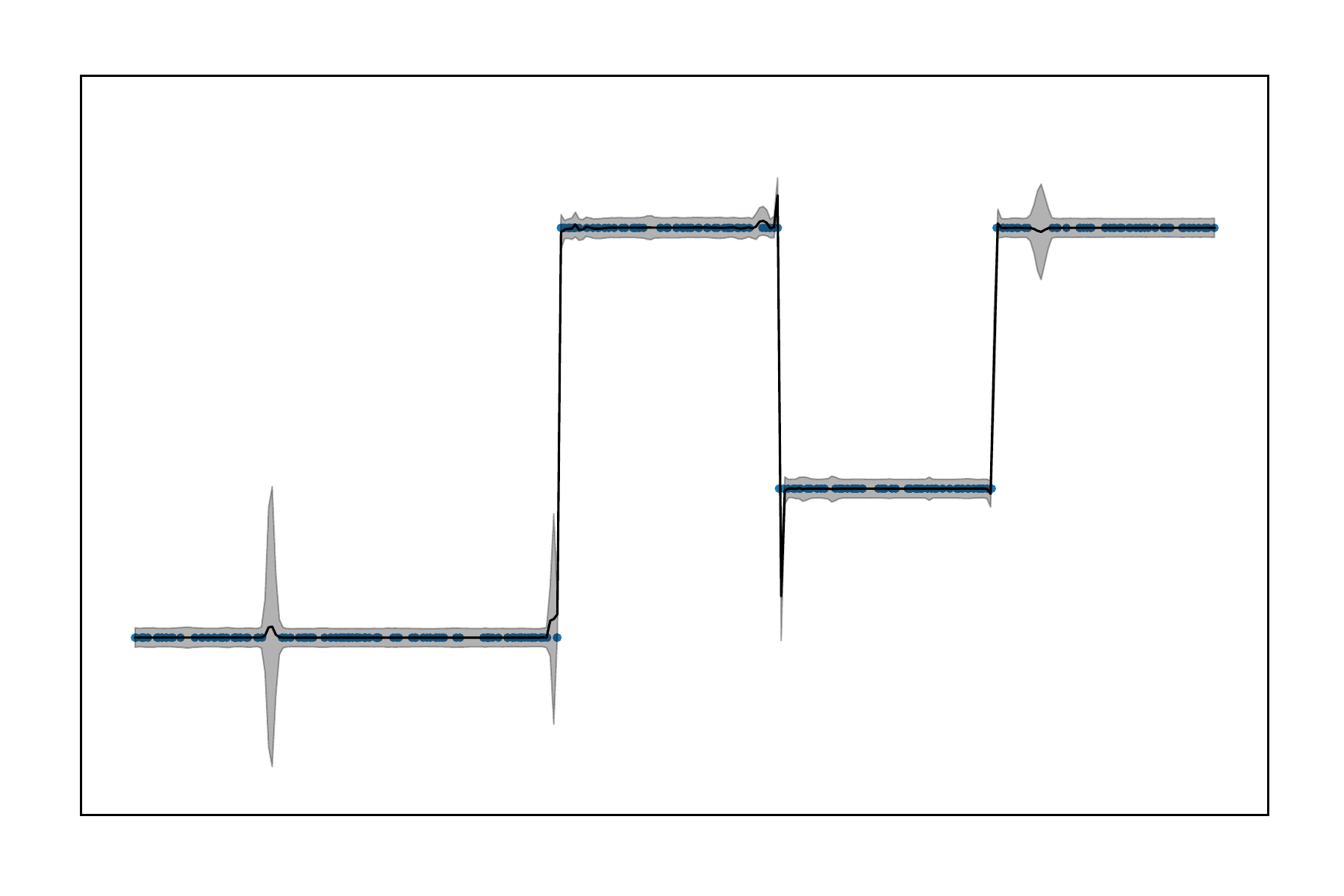}
        \end{subfigure}
        \hfill
        \begin{subfigure}[c]{0.32\linewidth}
            \includegraphics[width=\linewidth]{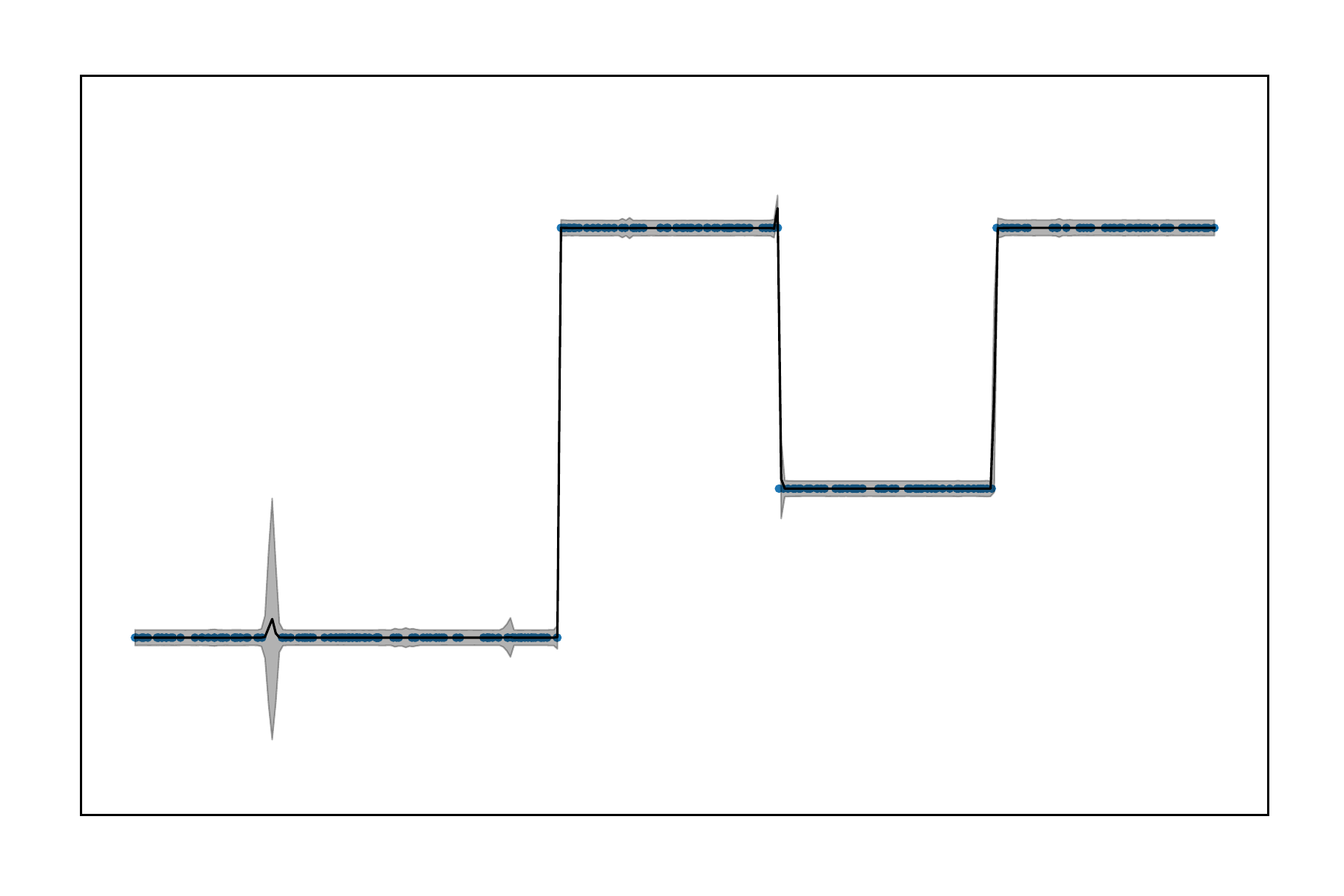}
        \end{subfigure}
        \hfill
        \begin{subfigure}[c]{0.32\linewidth}
            \includegraphics[width=\linewidth]{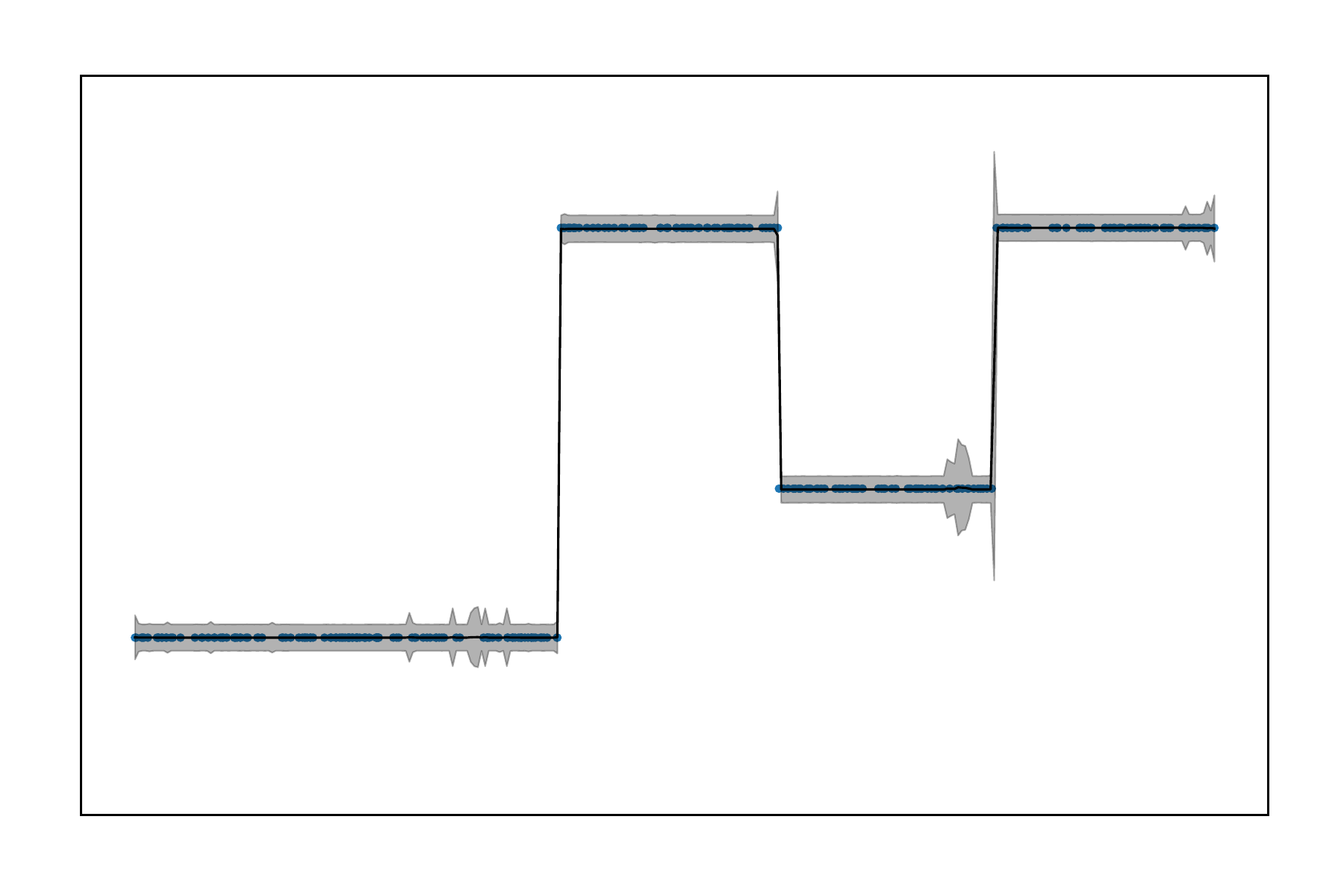}
        \end{subfigure}
    \end{minipage}
    
    \begin{minipage}[c]{\textwidth}
        \makebox[\linewidth][c]{\textbf{\PCAW{}}}%
    \end{minipage}
    \caption{Predictive distribution (mean and standard deviation) of \ZEROW{} (top 3 rows) and \PCAW{} (bottom 3 rows) \DGP models initialized with $\varSw{}{} = 10^{-5}$ in all layers. Blue points show training data. Columns correspond to different number of layers. Rows correspond to different number of inducing points: 5, 20 and 100 inducing points.}
    \label{fig:more_inducings_all_1e-5}
\end{figure}

\item $\varS{}{l}=10^{-5}\mathbf{I}$ and $\varS{}{L} = \mathbf{I}$:
\fig~\ref{fig:more_inducings_inner_low_output_high} shows the results obtained when the output layer covariances are initialized to those of the prior. We observe that, in this case, the \ZEROW{} model with $5$ layers and $100$ inducing points suffers from posterior collapse. When $\varS{}{L} = 10^{-5}\mathbf{I}$, the model does not collapse, but the predictive distribution is poor. This confirms that setting $\varS{}{L} = \mathbf{I}$ increases the probability of a posterior collapse since $4$ out of $9$ models have collapsed.  Again, we observe that using a higher number of inducing points leads to better predictive distributions and reduces the probability of a posterior collapse. Specifically, the predictive distribution of the \ZEROW{} \DGP with 3 layers improves substantially when $100$ inducing points are considered, similar to the previous setting considered in \fig~\ref{fig:more_inducings_all_1e-5}. Again, the \PCAW{} model does not result in posterior collapse, unlike the \ZEROW{} model. However, in this model we can observe a poorer fit with $5$ inducing points, especially with $3$ layers. This confirms that, in general, setting $\varS{}{L} = \mathbf{I}$ results in suboptimal fitting compared to $\varS{}{L} = 10^{-5}\mathbf{I}$.
\begin{figure}[htbp]
\vspace{-0.75cm}
    \centering



    \begin{minipage}[c]{0.05\textwidth}
    \end{minipage}%
    \begin{minipage}[c]{0.93\textwidth}
        \centering
        \hfill
        \makebox[0.32\linewidth][c]{\textbf{2 Layers}}%
        \hfill
        \makebox[0.32\linewidth][c]{\textbf{3 Layers}}%
        \hfill
        \makebox[0.32\linewidth][c]{\textbf{5 Layers}}%
    \end{minipage}

\begin{minipage}[c]{0.05\textwidth}
    \centering
    \rotatebox{90}{\textbf{5 Inducing}}
\end{minipage}%
\begin{minipage}[c]{0.93\textwidth}
    \centering

    \begin{subfigure}[c]{0.32\linewidth}
        \includegraphics[width=\linewidth]{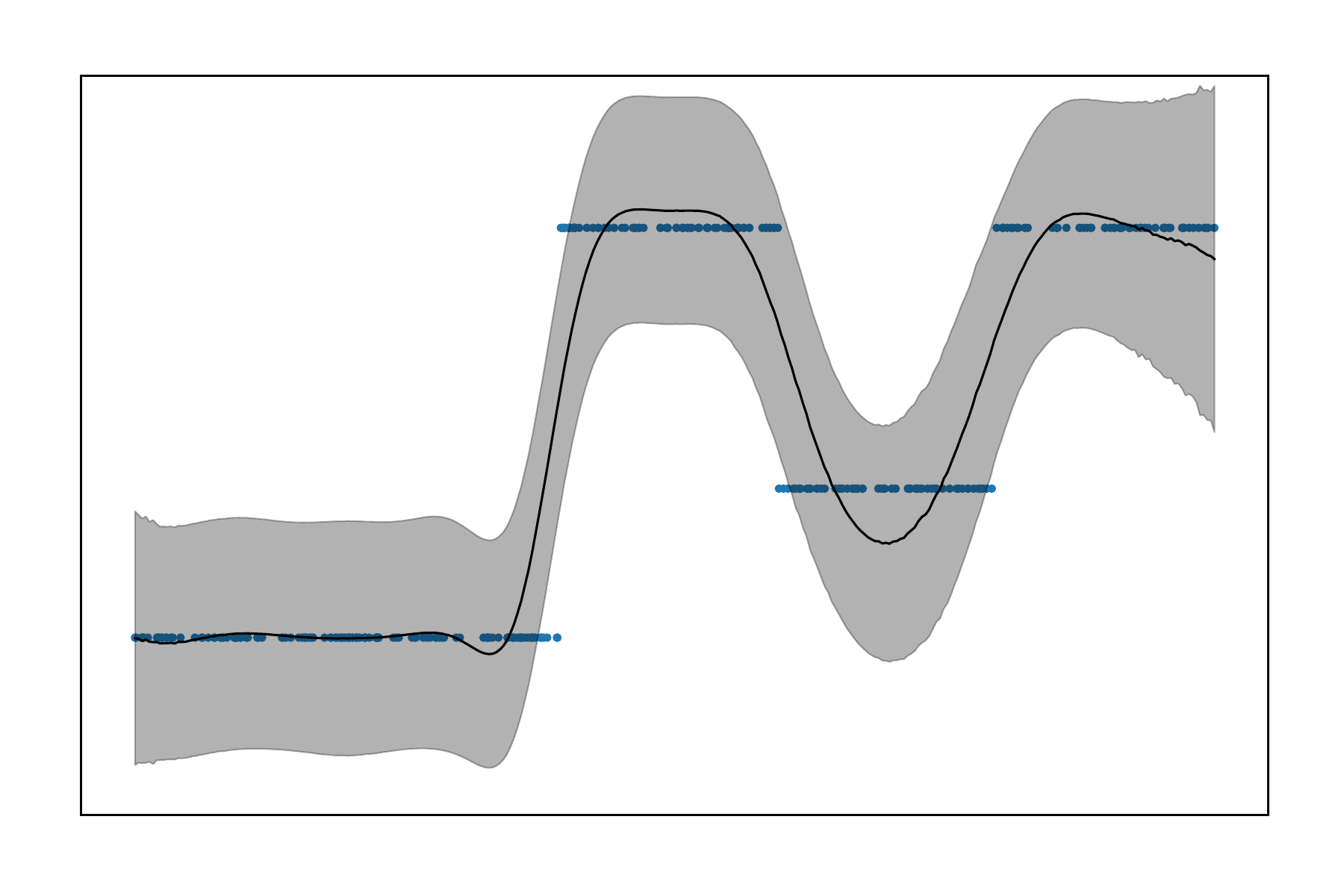}
    \end{subfigure}
    \hfill
    \begin{subfigure}[c]{0.32\linewidth}
        \includegraphics[width=\linewidth]{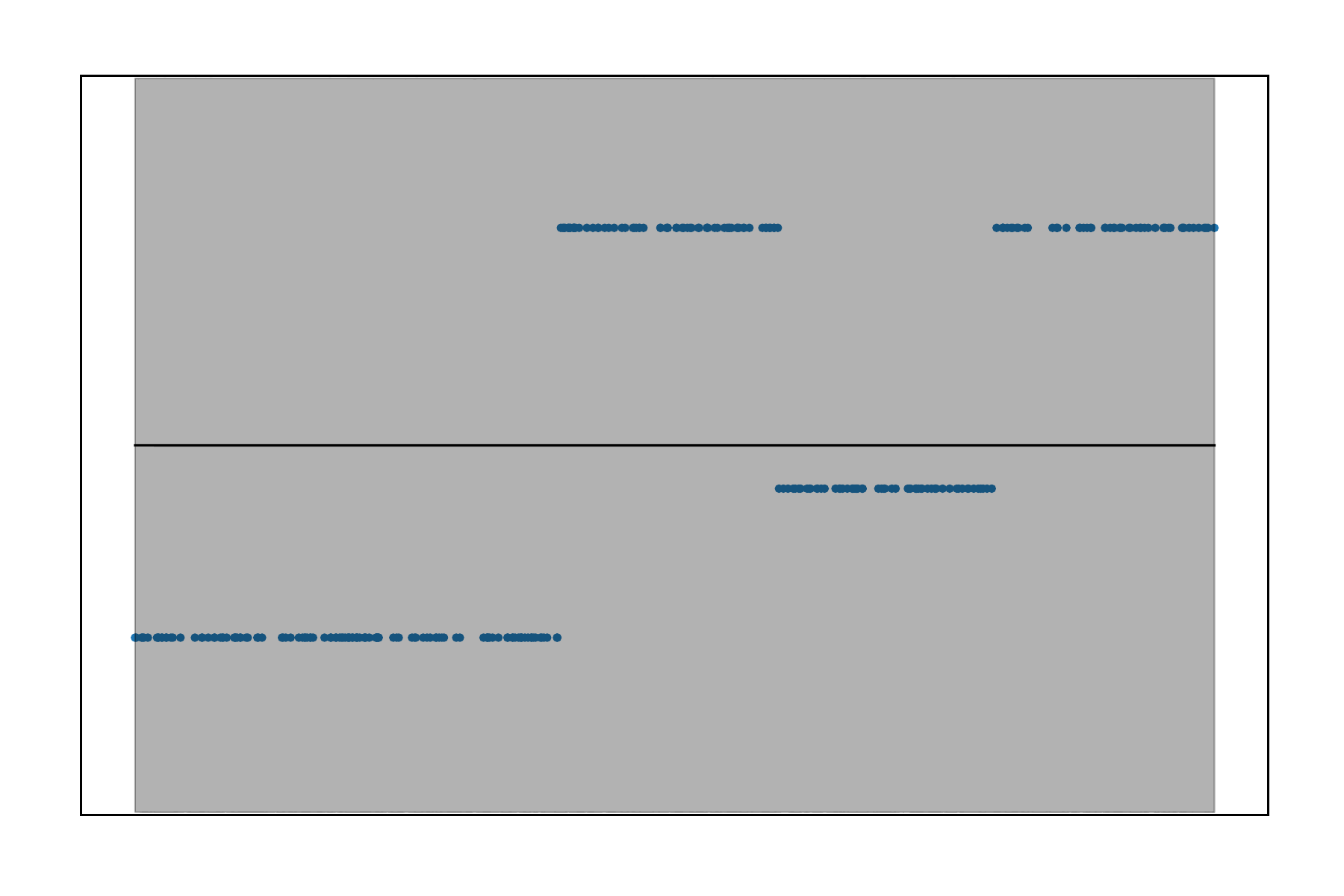}
    \end{subfigure}
    \hfill
    \begin{subfigure}[c]{0.32\linewidth}
        \includegraphics[width=\linewidth]{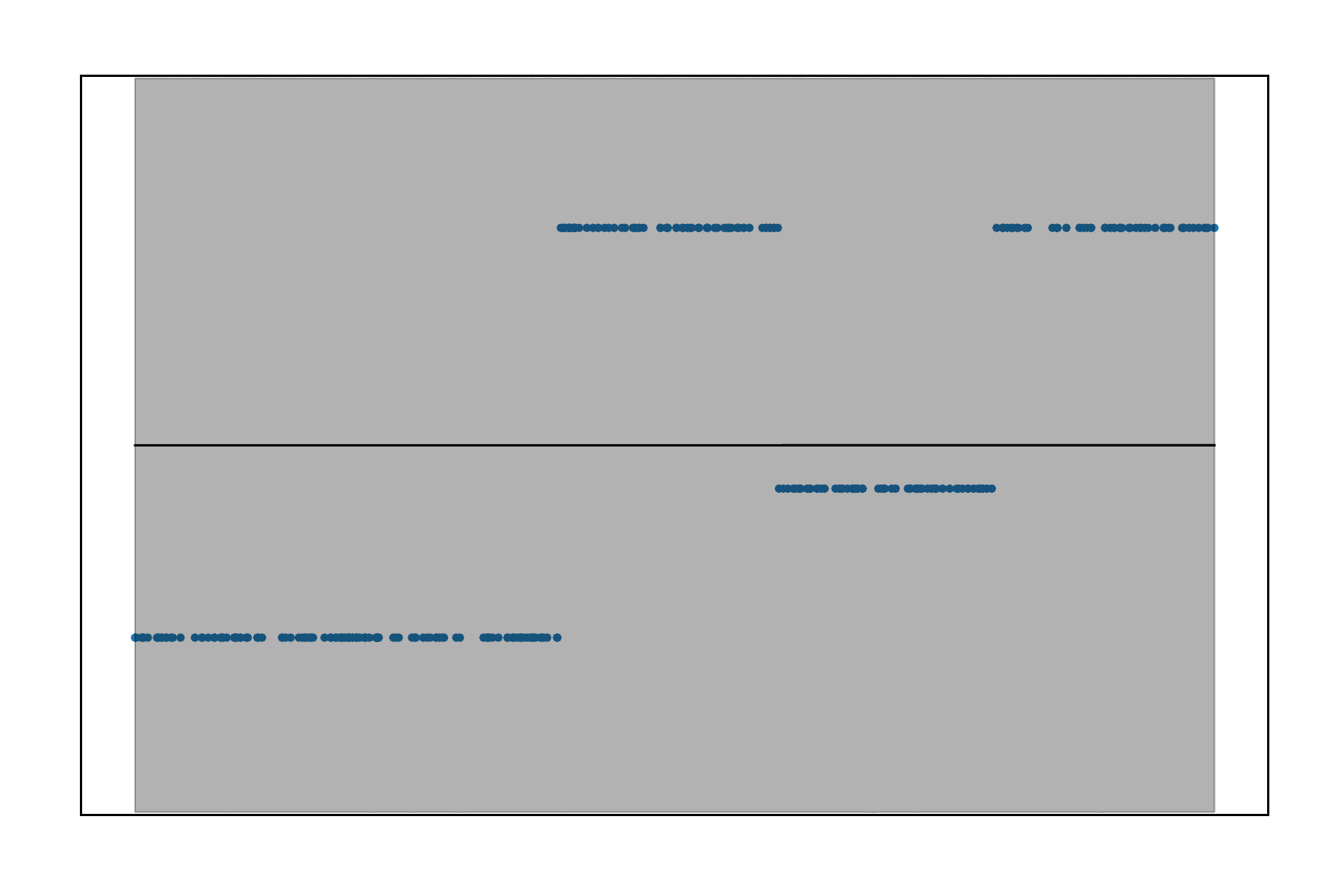}
    \end{subfigure}
\end{minipage}
    
    \begin{minipage}[c]{0.05\textwidth}
        \centering
        \rotatebox{90}{\textbf{20 Inducing}}
    \end{minipage}%
    \begin{minipage}[c]{0.93\textwidth}
        \centering
        \begin{subfigure}[c]{0.32\linewidth}
            \includegraphics[width=\linewidth]{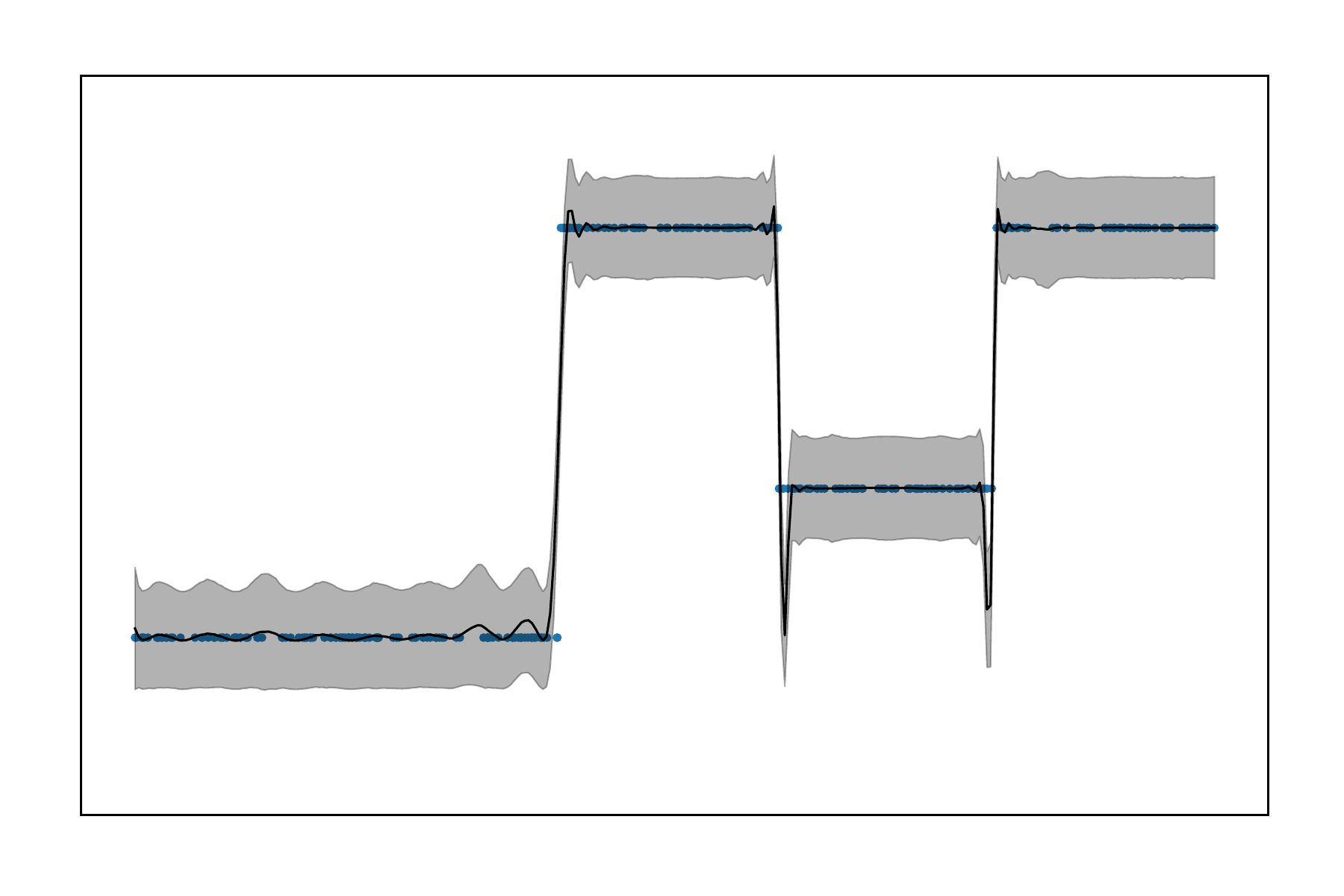}
        \end{subfigure}
        \hfill
        \begin{subfigure}[c]{0.32\linewidth}
            \includegraphics[width=\linewidth]{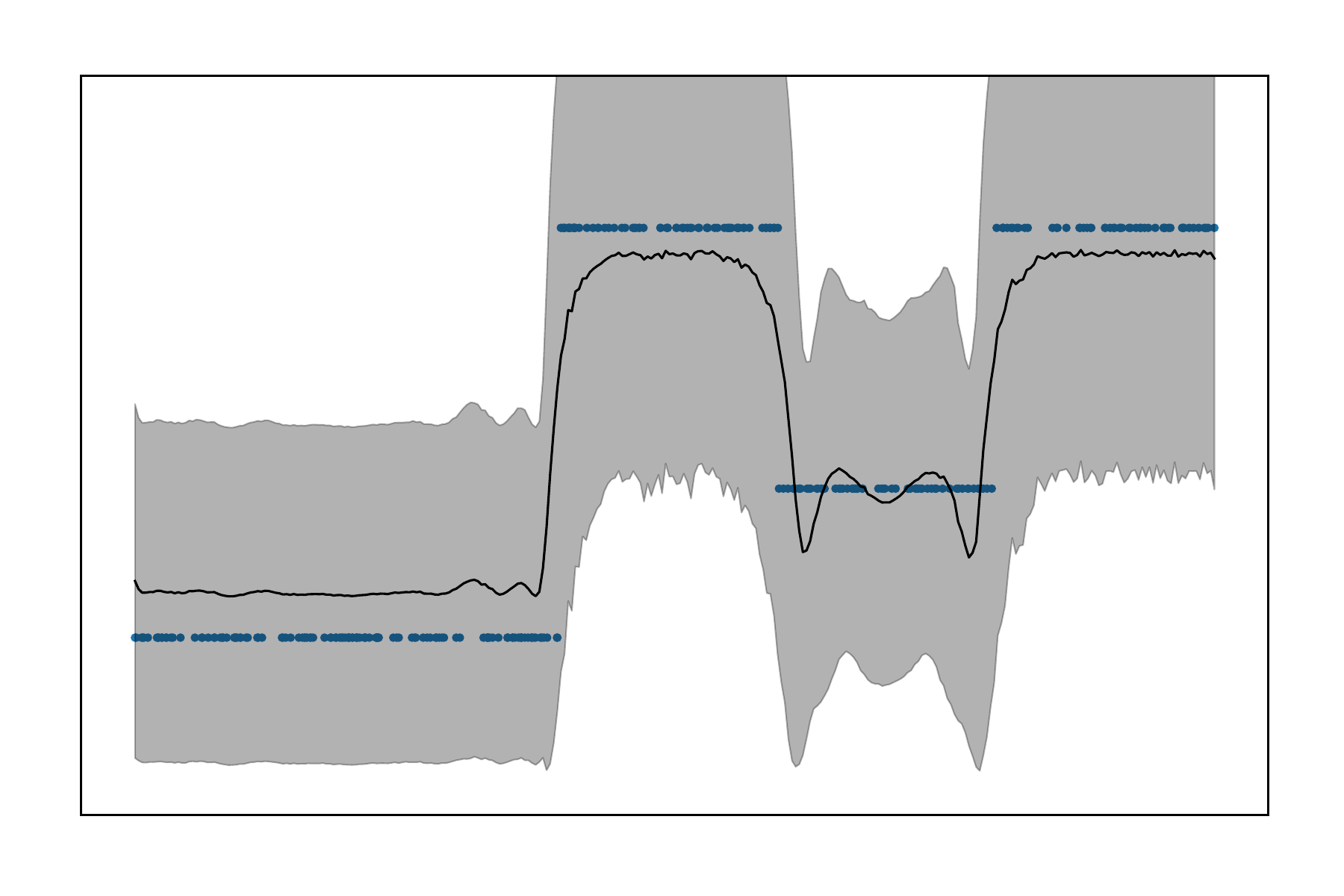}
        \end{subfigure}
        \hfill
        \begin{subfigure}[c]{0.32\linewidth}
            \includegraphics[width=\linewidth]{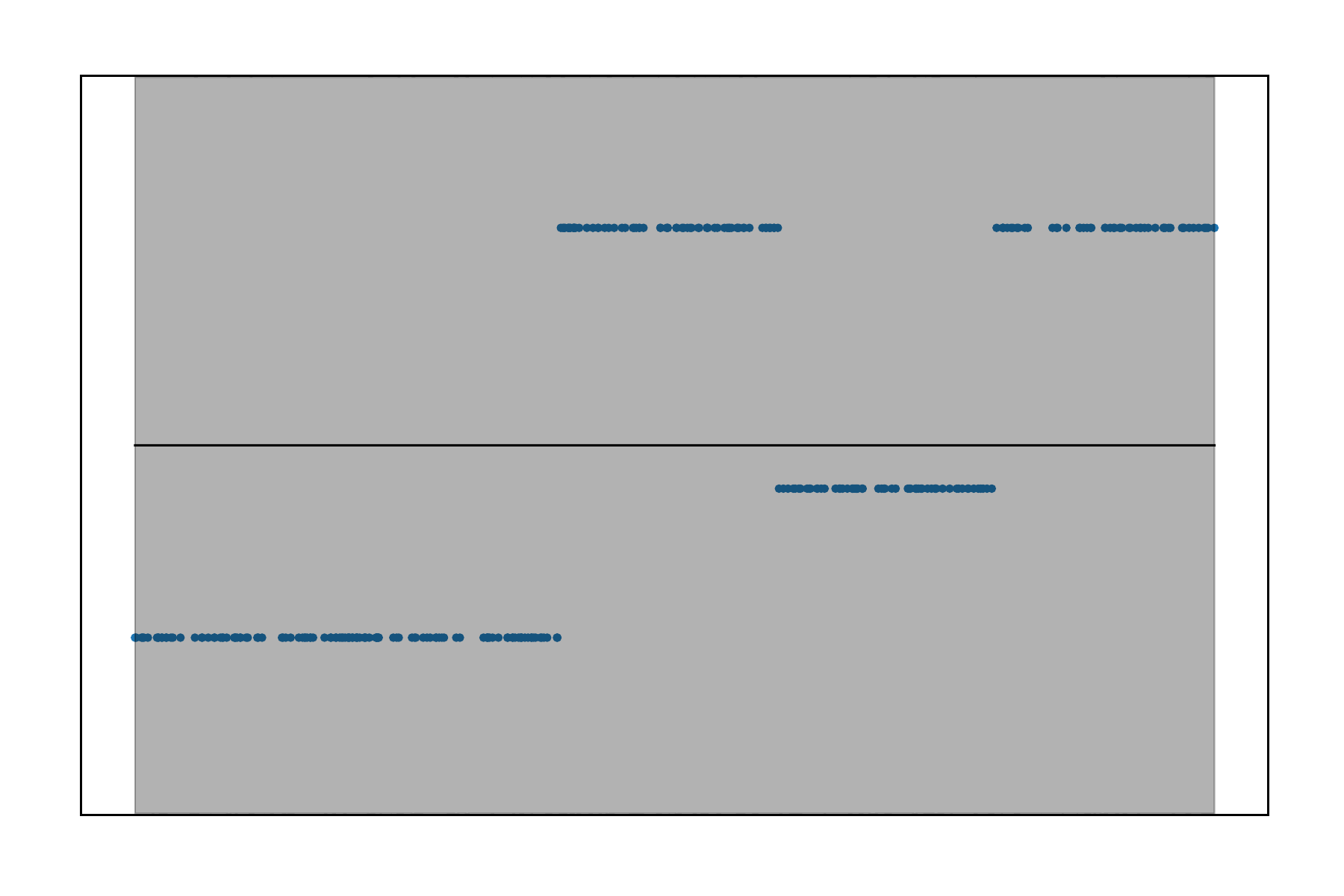}
        \end{subfigure}
    \end{minipage}

    \vspace{0.25cm} 

    \begin{minipage}[c]{0.05\textwidth}
        \centering
        \rotatebox{90}{\textbf{100 Inducing}}
    \end{minipage}%
    \begin{minipage}[c]{0.93\textwidth}
        \centering
        \begin{subfigure}[c]{0.32\linewidth}
            \includegraphics[width=\linewidth]{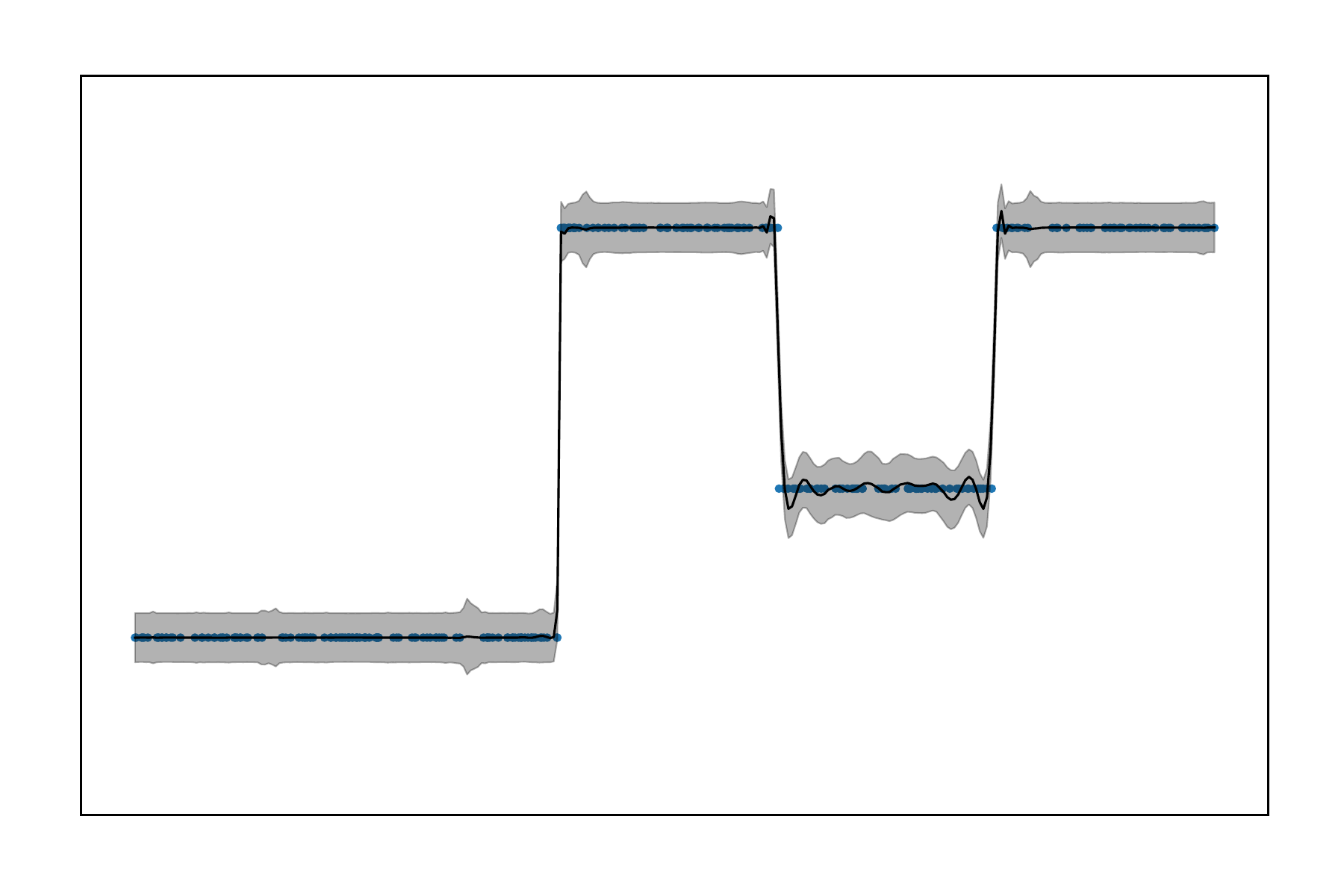}
        \end{subfigure}
        \hfill
        \begin{subfigure}[c]{0.32\linewidth}
            \includegraphics[width=\linewidth]{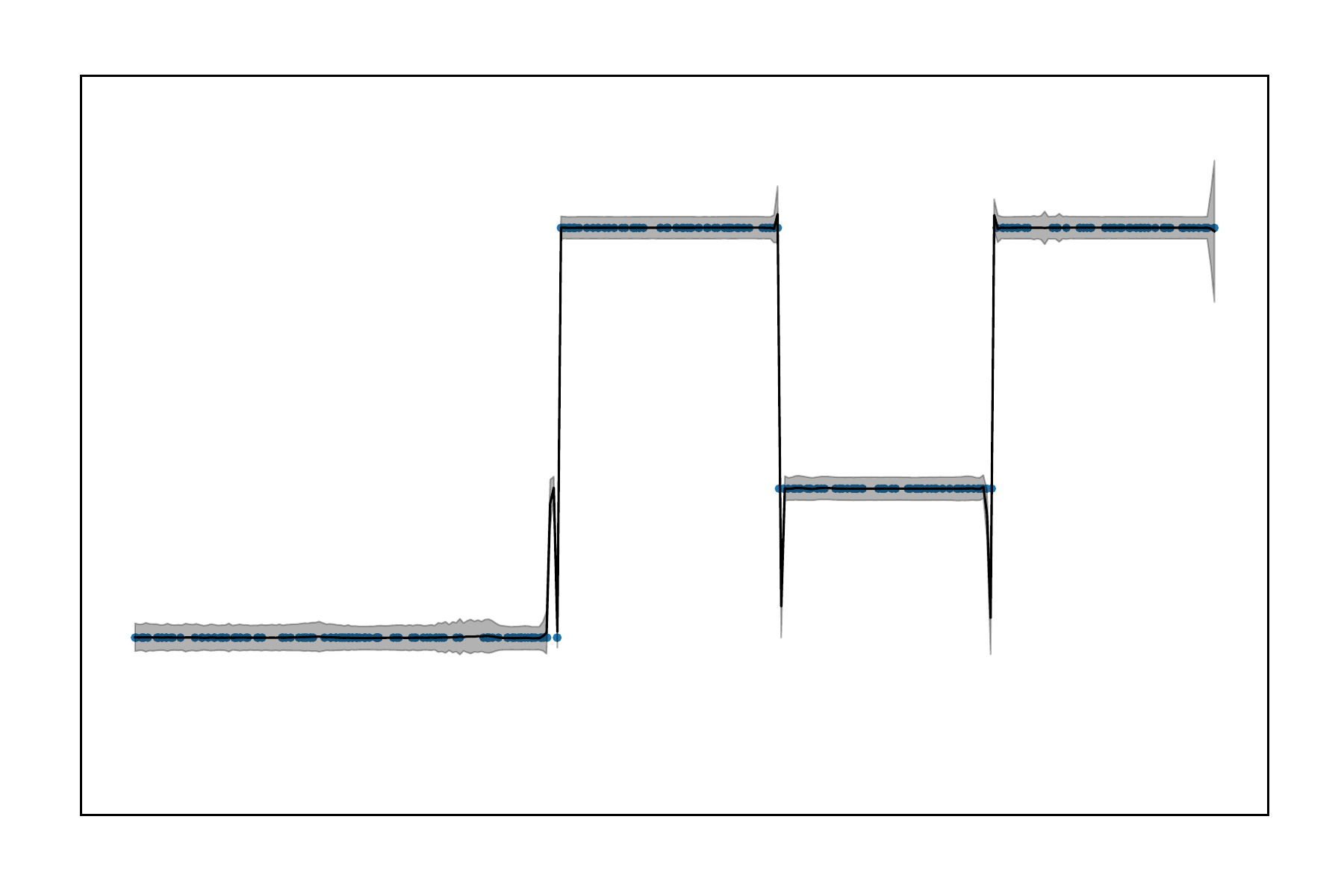}
        \end{subfigure}
        \hfill
        \begin{subfigure}[c]{0.32\linewidth}
            \includegraphics[width=\linewidth]{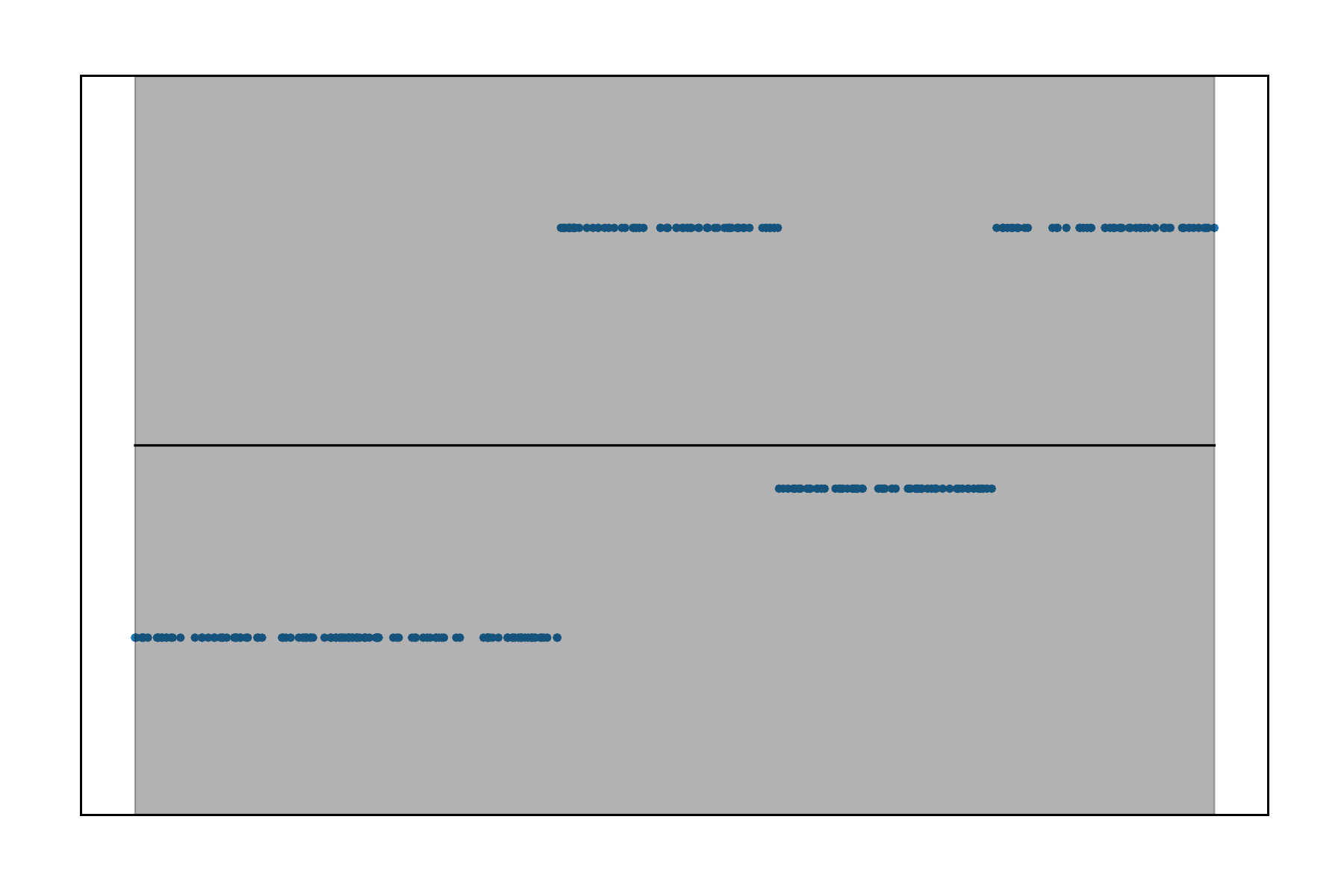}
        \end{subfigure}
    \end{minipage}
    \begin{minipage}[c]{\textwidth}
        \makebox[\linewidth][c]{\textbf{\ZEROW{}}}%
    \end{minipage}

    \begin{minipage}[c]{0.05\textwidth}
        \hfill 
    \end{minipage}%

\begin{minipage}[c]{0.05\textwidth}
    \centering
    \rotatebox{90}{\textbf{5 Inducing}}
\end{minipage}%
\begin{minipage}[c]{0.93\textwidth}
    \centering

    \begin{subfigure}[c]{0.32\linewidth}
        \includegraphics[width=\linewidth]{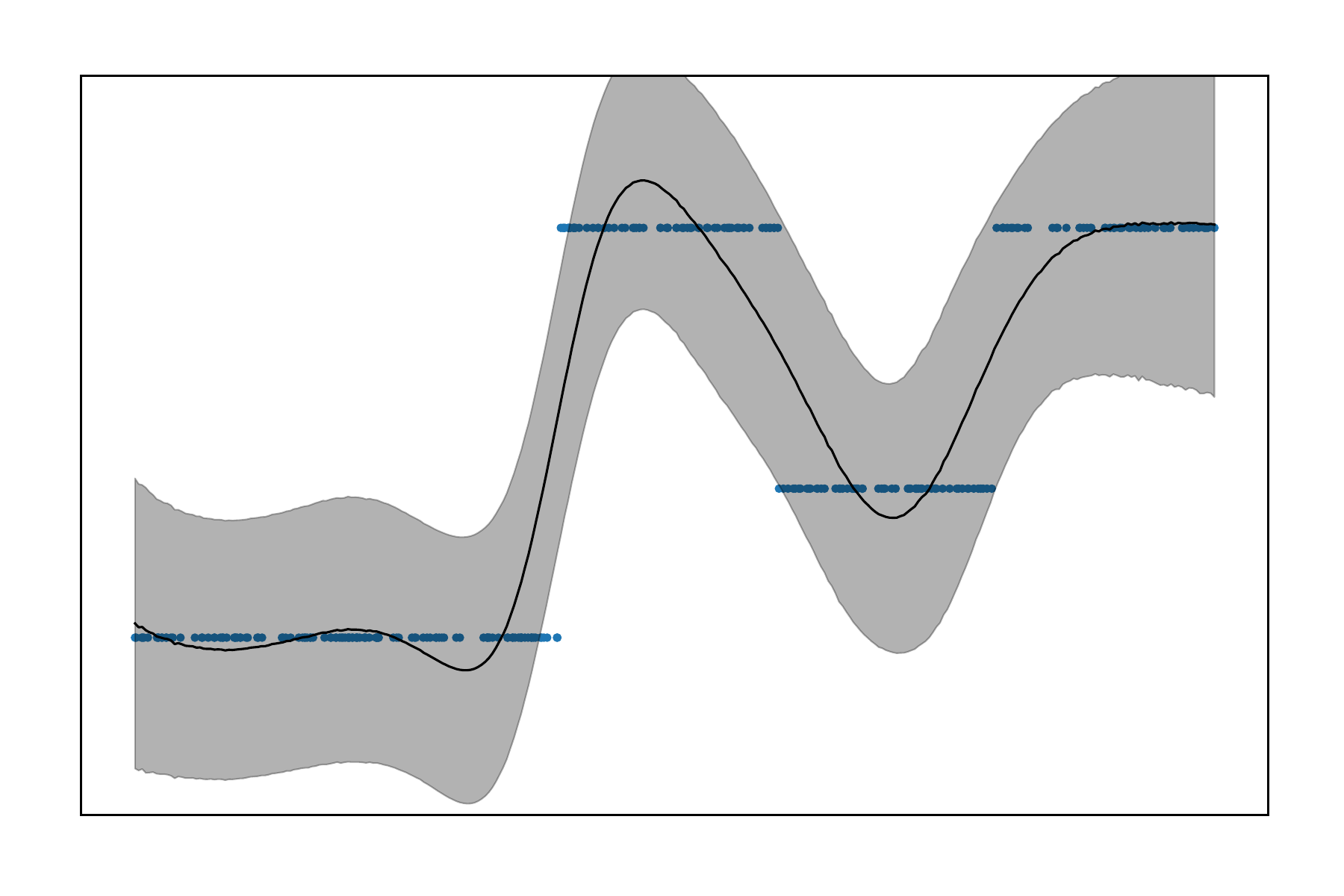}
    \end{subfigure}
    \hfill
    \begin{subfigure}[c]{0.32\linewidth}
        \includegraphics[width=\linewidth]{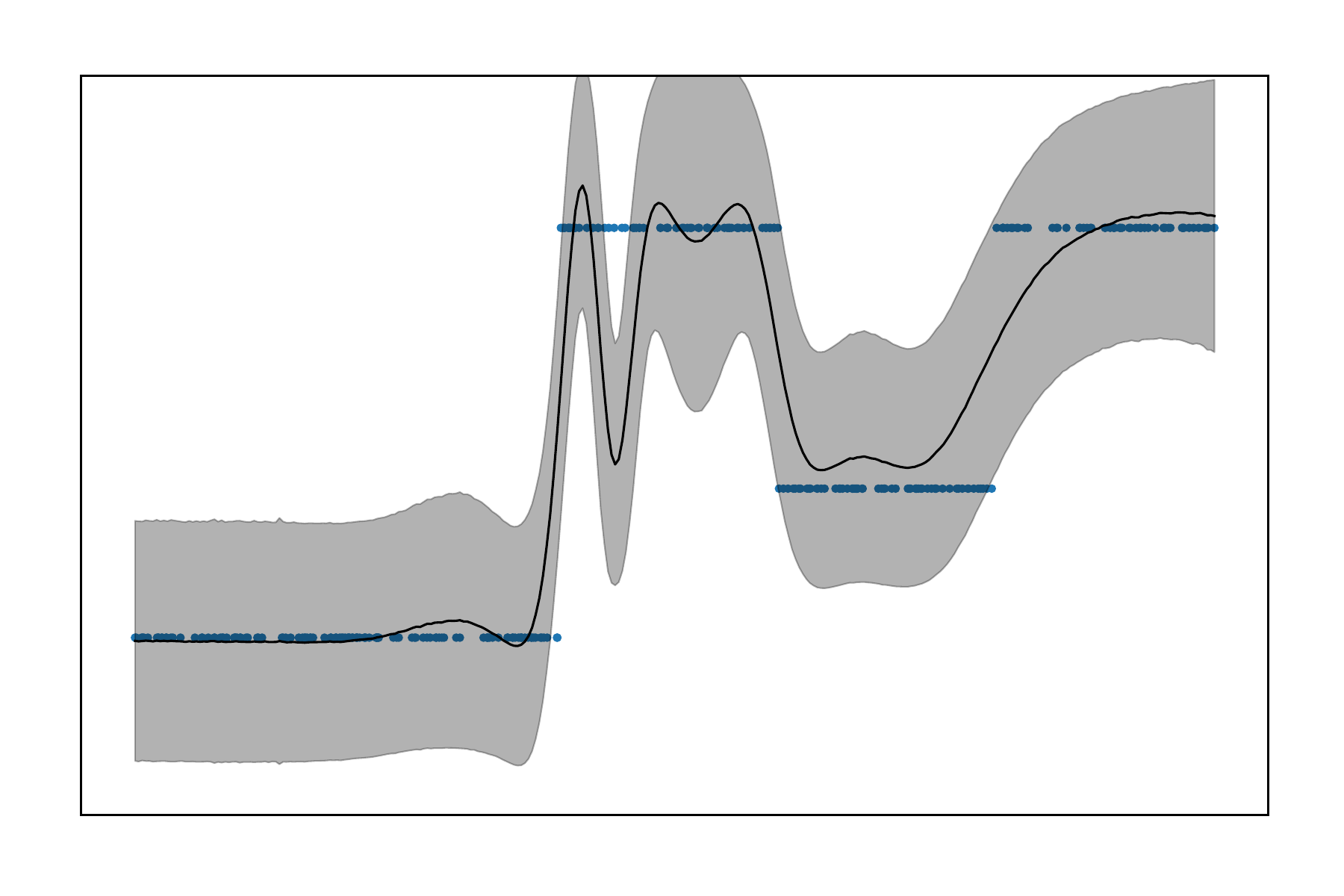}
    \end{subfigure}
    \hfill
    \begin{subfigure}[c]{0.32\linewidth}
        \includegraphics[width=\linewidth]{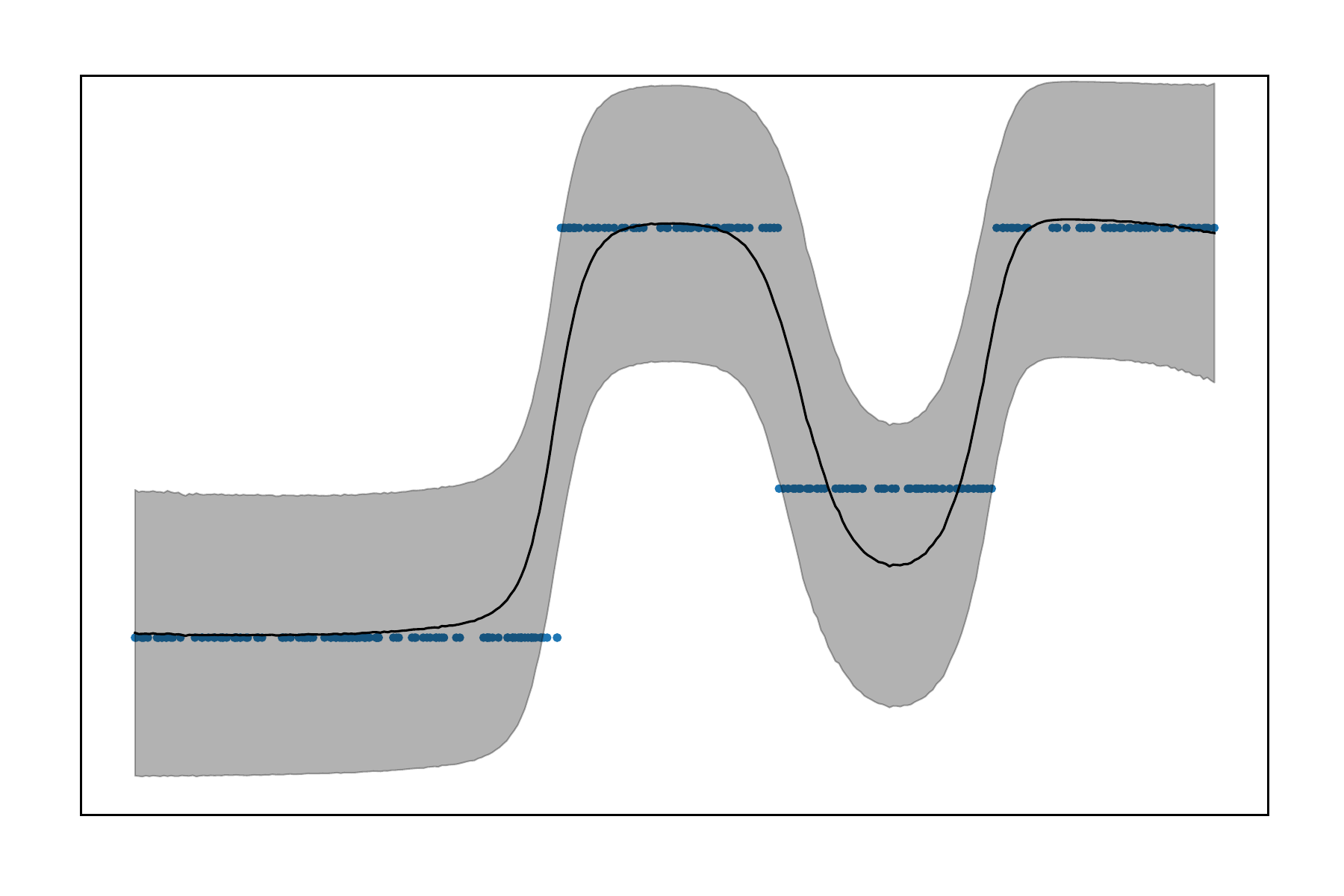}
    \end{subfigure}
\end{minipage}
    
    \begin{minipage}[c]{0.05\textwidth}
        \centering
        \rotatebox{90}{\textbf{20 Inducing}}
    \end{minipage}%
    \begin{minipage}[c]{0.93\textwidth}
        \centering
        \begin{subfigure}[c]{0.32\linewidth}
            \includegraphics[width=\linewidth]{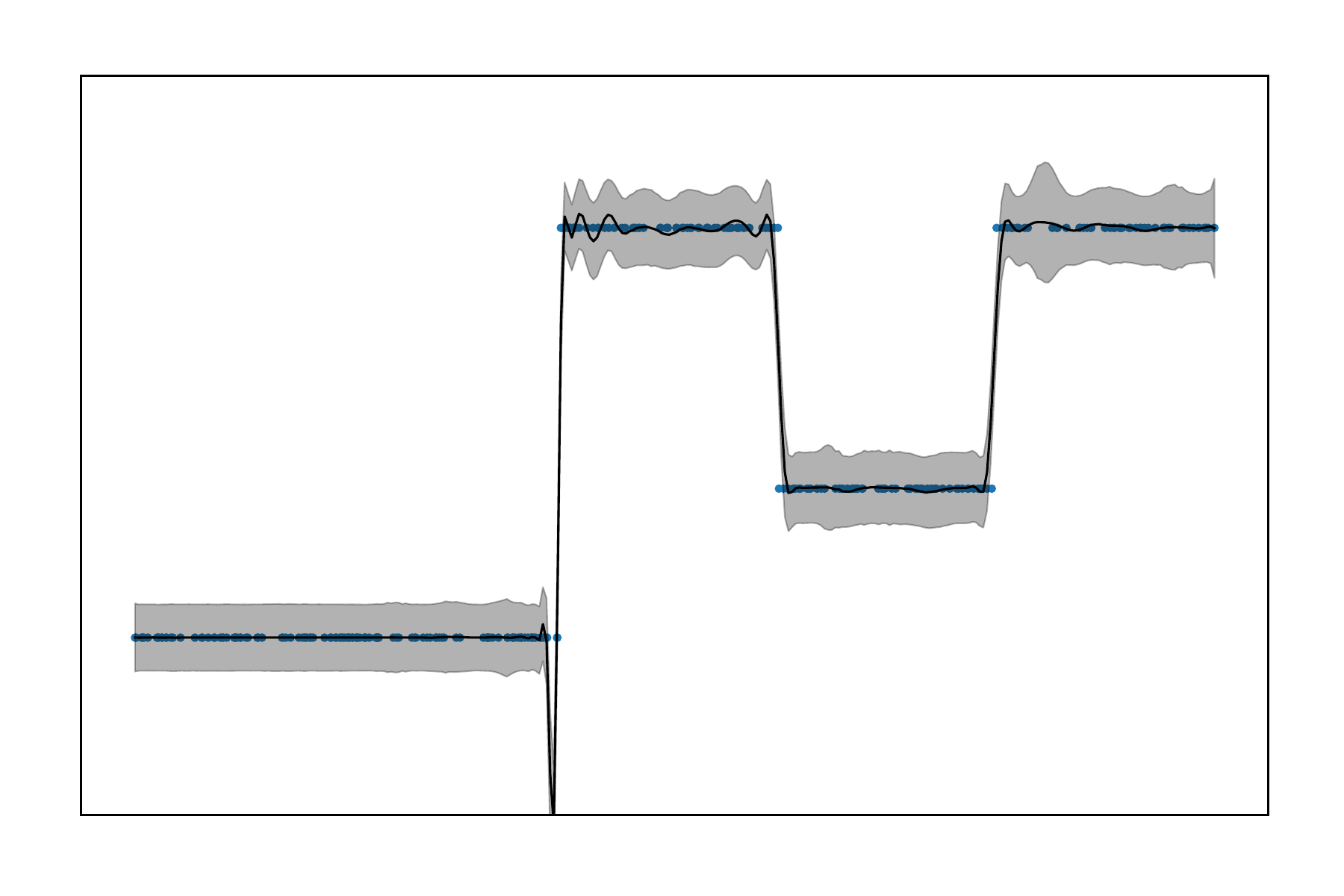}
        \end{subfigure}
        \hfill
        \begin{subfigure}[c]{0.32\linewidth}
            \includegraphics[width=\linewidth]{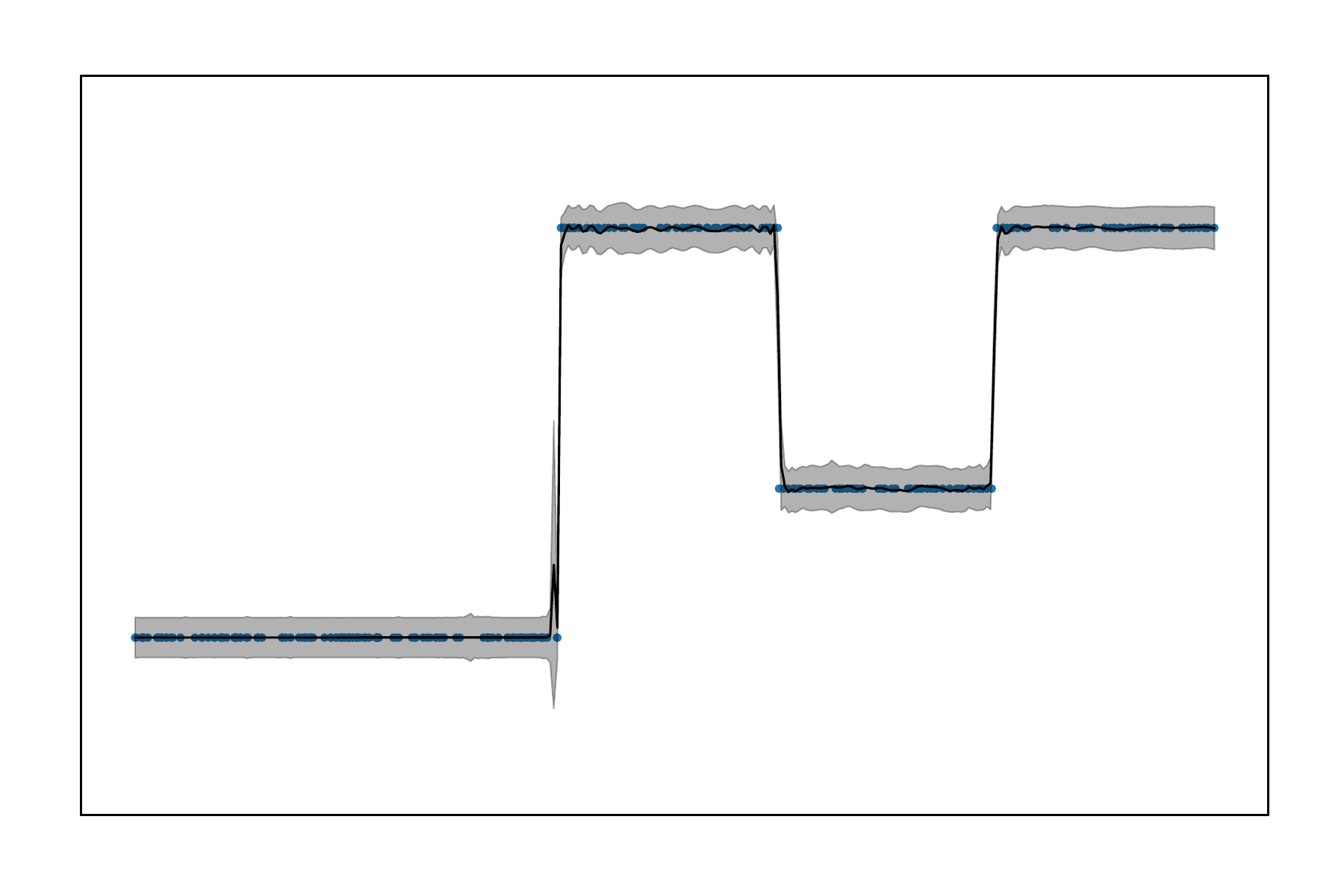}
        \end{subfigure}
        \hfill
        \begin{subfigure}[c]{0.32\linewidth}
            \includegraphics[width=\linewidth]{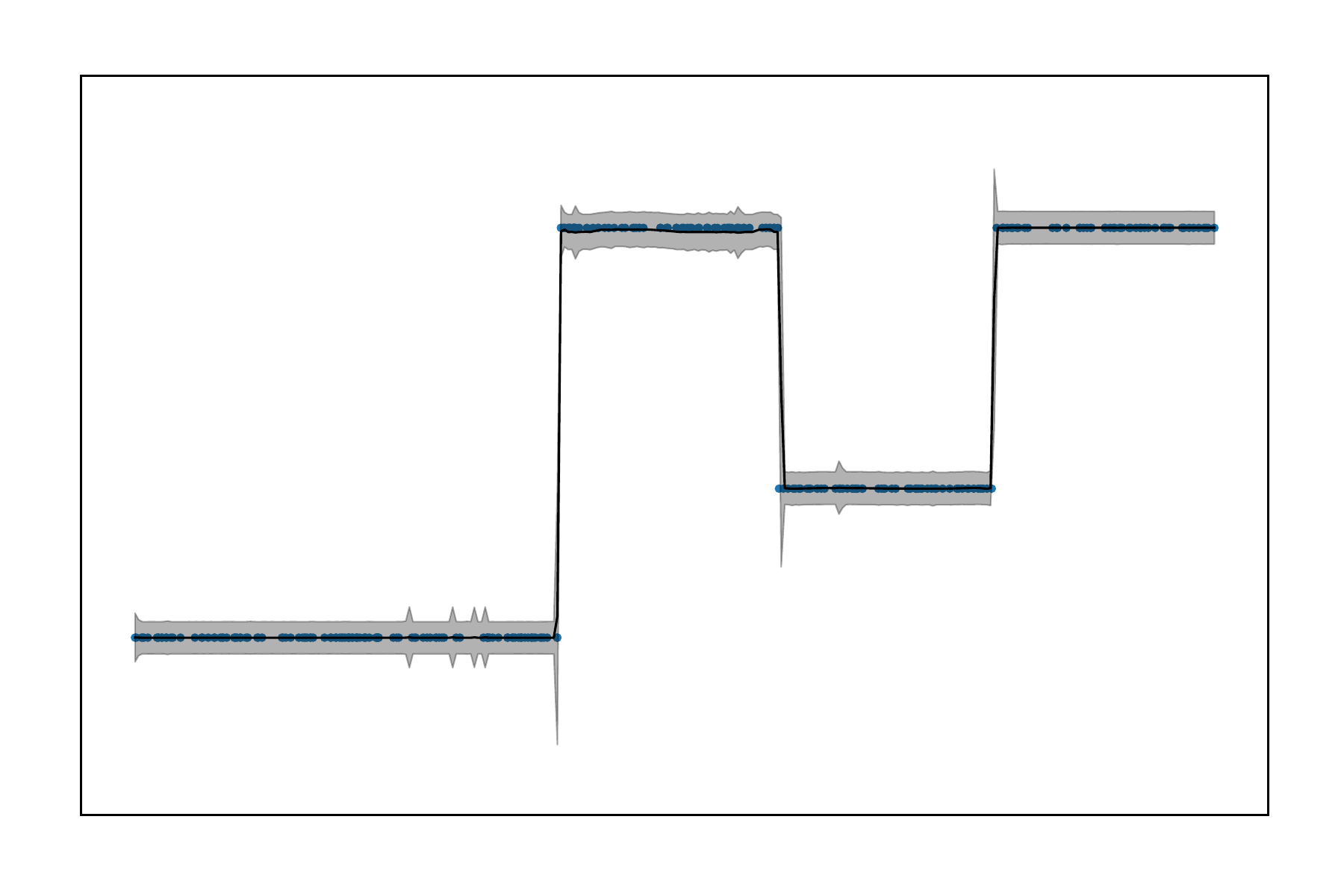}
        \end{subfigure}
    \end{minipage}

    \vspace{0.25cm} 

    \begin{minipage}[c]{0.05\textwidth}
        \centering
        \rotatebox{90}{\textbf{100 Inducing}}
    \end{minipage}%
    \begin{minipage}[c]{0.93\textwidth}
        \centering
        \begin{subfigure}[c]{0.32\linewidth}
            \includegraphics[width=\linewidth]{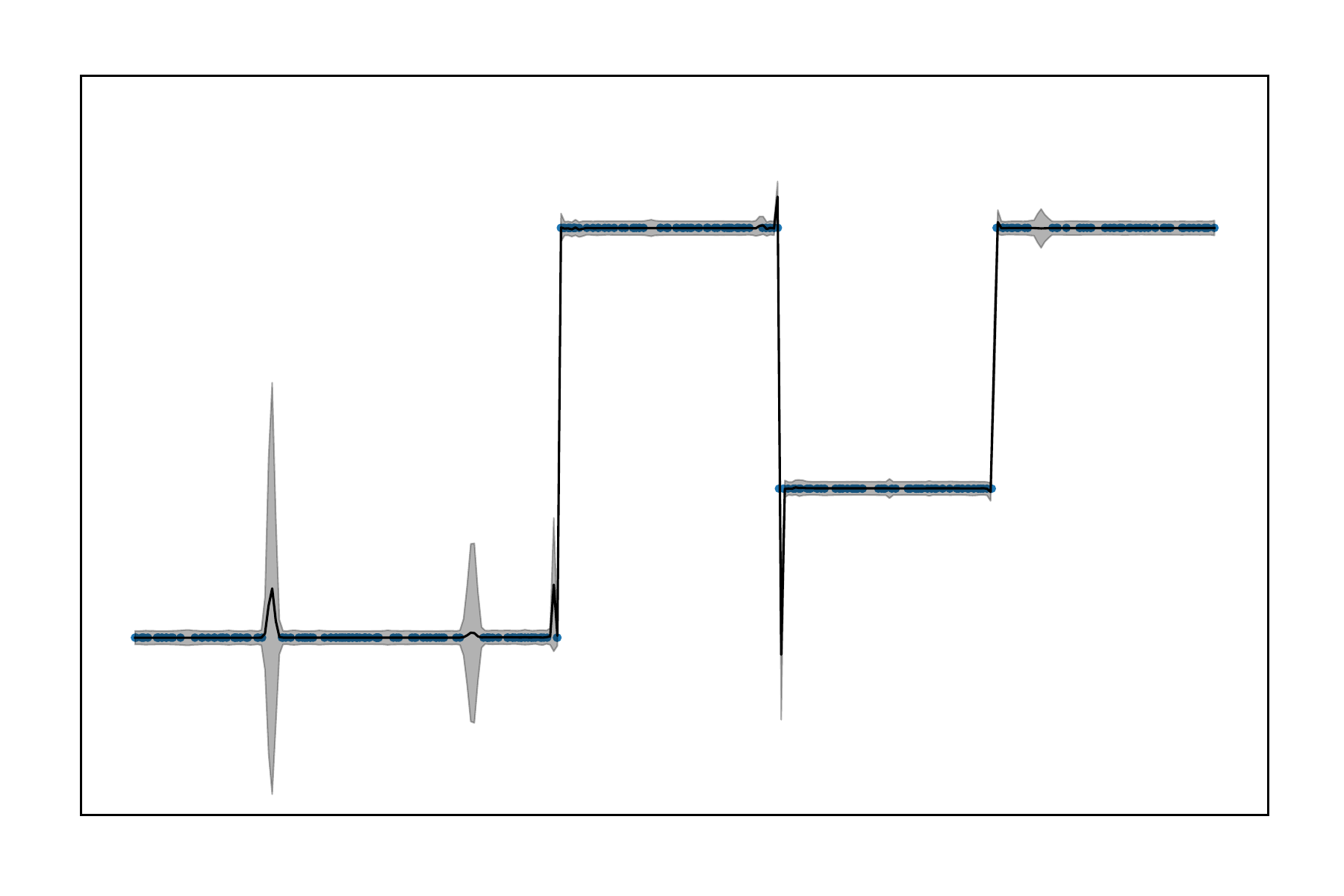}
        \end{subfigure}
        \hfill
        \begin{subfigure}[c]{0.32\linewidth}
            \includegraphics[width=\linewidth]{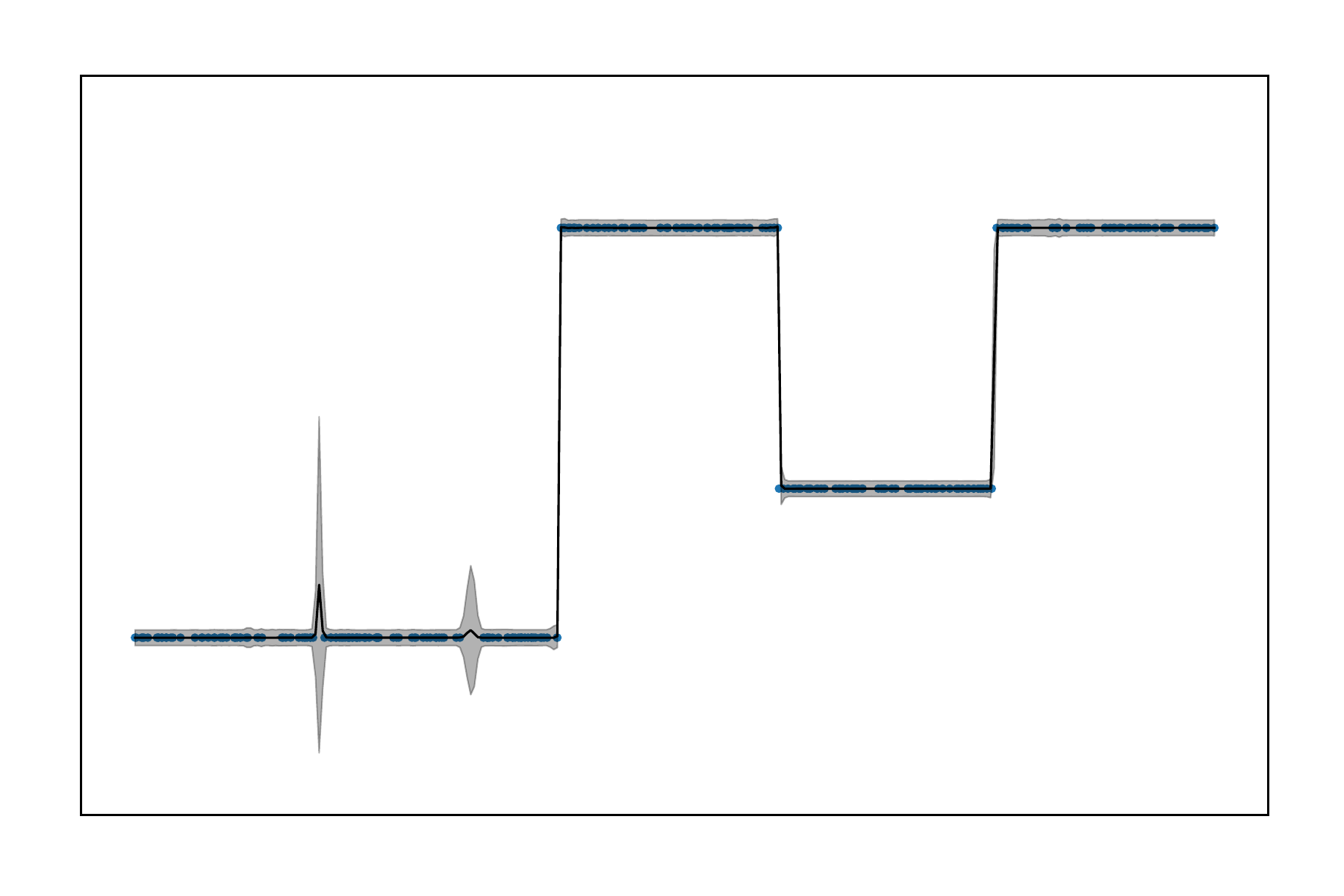}
        \end{subfigure}
        \hfill
        \begin{subfigure}[c]{0.32\linewidth}
            \includegraphics[width=\linewidth]{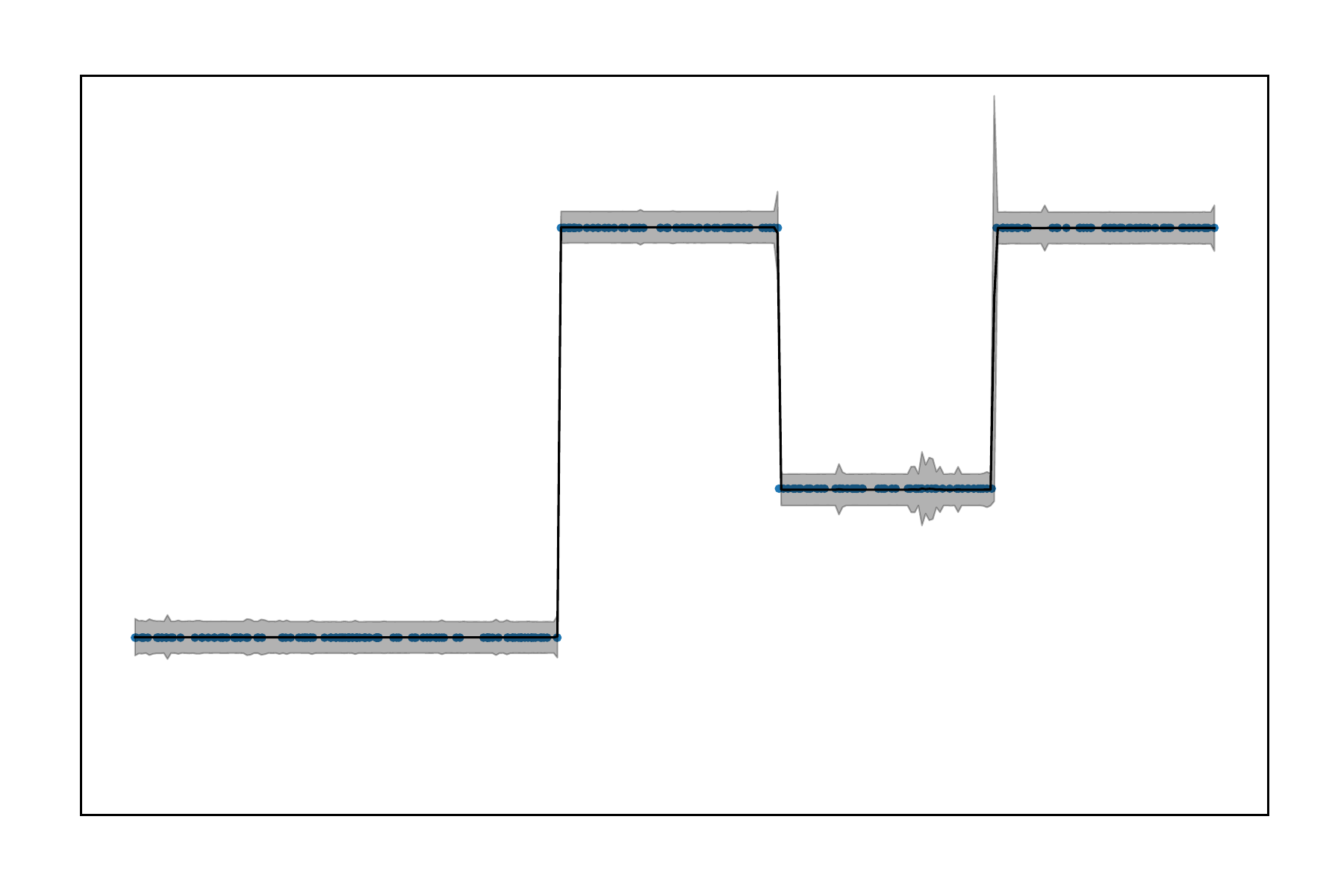}
        \end{subfigure}
    \end{minipage}
    
    \begin{minipage}[c]{\textwidth}
        \makebox[\linewidth][c]{\textbf{\PCAW{}}}%
    \end{minipage}
    \caption{Predictive distribution (mean and standard deviation) of \ZEROW{} (top 3 rows) and \PCAW{} (bottom 3 rows) \DGP models initialized with $\varSw{l}{} = 10^{-5}$ in the hidden layers and $\varSw{L}{} = \mathbf{I}$ in the output layer. Blue points show training data. Higher variance in the output layer results in poorer solutions. Columns correspond to different numbers of layers. Rows correspond to different numbers of inducing points: 5, 20, and 100 inducing points.}
 
\label{fig:more_inducings_inner_low_output_high}
\end{figure}

\item $\varS{}{l}=\mathbf{I}$ and $\varS{}{L} = 10^{-5}\mathbf{I}$: \fig~\ref{fig:more_inducings_inner_high_output_low} shows the predictive distributions in this case, where the inner layer variance is higher. Again, we observe that the \ZEROW{} model suffers from posterior collapse in four out of the nine cases. This confirms that setting $\varS{}{l}=\mathbf{I}$ in the inner layers increases the probability of posterior collapse. Compared to the previous setting, we observe that the \ZERO{} prior mean model with $3$ layers and $5$ inducing points slightly escapes from collapse. As we discussed at the end of \usec \ref{sec:Iat_output}, this might be because, since the variability in the samples from the inner layer is higher, the chance of samples being far from $\veczero$ increases, which avoids the pathological behavior of the \ZERO \DGP{} at initialization. Nevertheless, the solution is poor. In Appendix \ref{sec:app:C} we provide further insights on this configuration. Furthermore, in this setting, the predictive distribution of the \PCAW{} model is also slightly worse than when $\varS{}{l}=10^{-5}\mathbf{I}$ for $20$ inducing points. These results confirm that one should avoid setting high initial uncertainties in the inner layers of the \DGP.

\begin{figure}[htbp]
\centering

\begin{minipage}[c]{0.05\textwidth}
\end{minipage}%
\begin{minipage}[c]{0.93\textwidth}
\centering
\hfill
\makebox[0.32\linewidth][c]{\textbf{2 Layers}}%
\hfill
\makebox[0.32\linewidth][c]{\textbf{3 Layers}}%
\hfill
\makebox[0.32\linewidth][c]{\textbf{5 Layers}}
\end{minipage}

\begin{minipage}[c]{0.05\textwidth}
    \centering
    \rotatebox{90}{\textbf{5 Inducing}}
\end{minipage}%
\begin{minipage}[c]{0.93\textwidth}
    \centering

    \begin{subfigure}[c]{0.32\linewidth}
        \includegraphics[width=\linewidth]{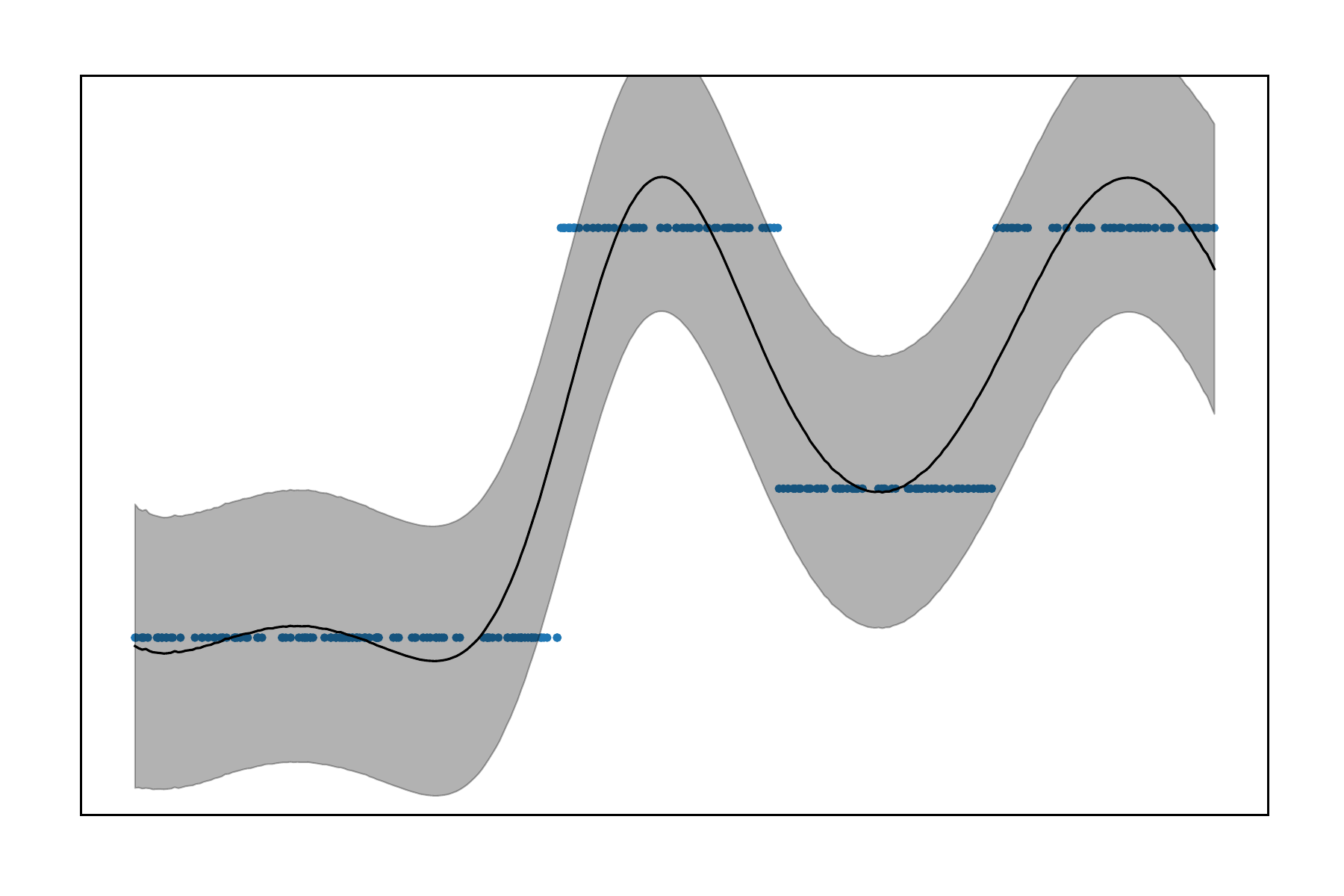}
    \end{subfigure}
    \begin{subfigure}[c]{0.32\linewidth}
        \includegraphics[width=\linewidth]{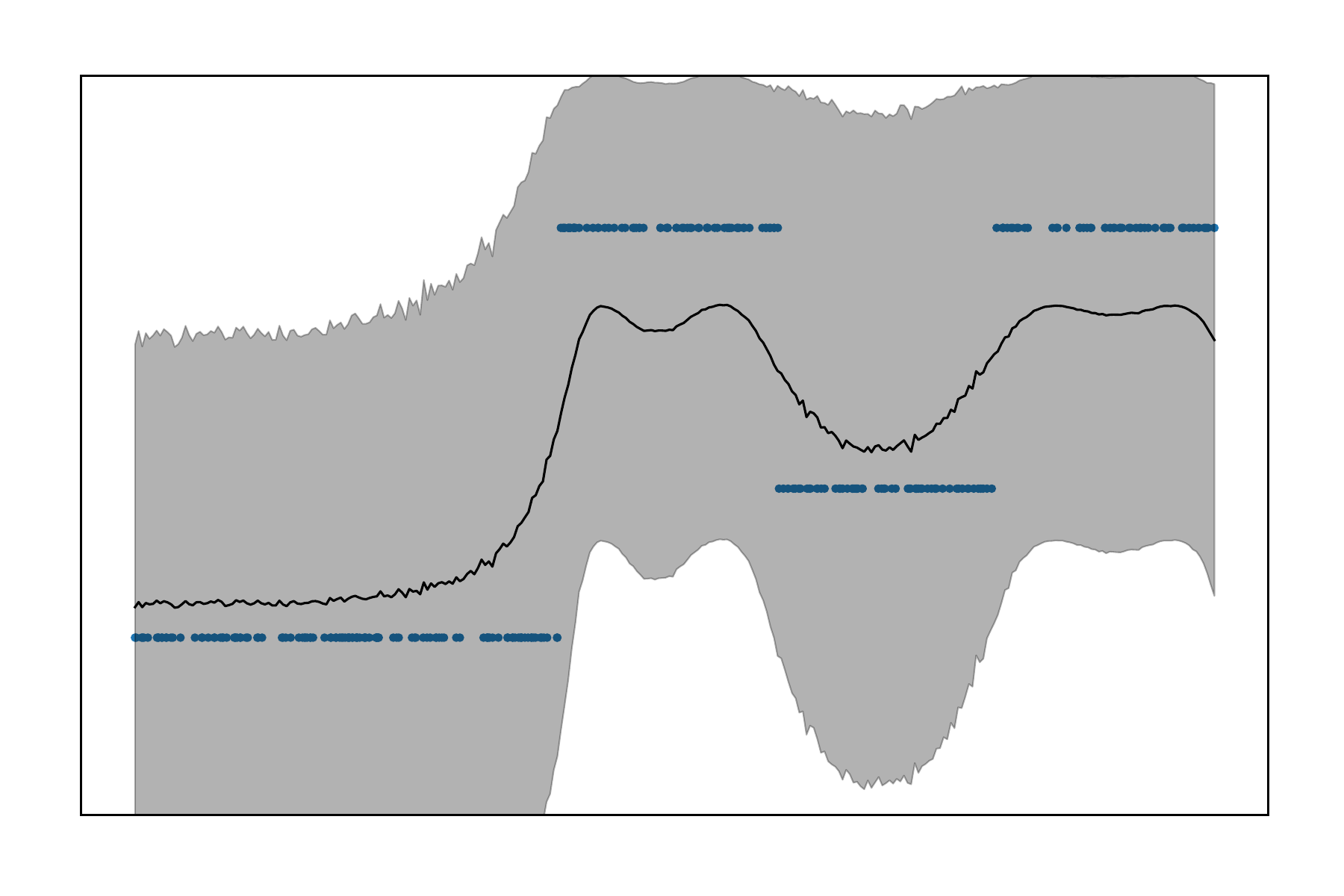}
    \end{subfigure}
    \begin{subfigure}[c]{0.32\linewidth}
        \includegraphics[width=\linewidth]{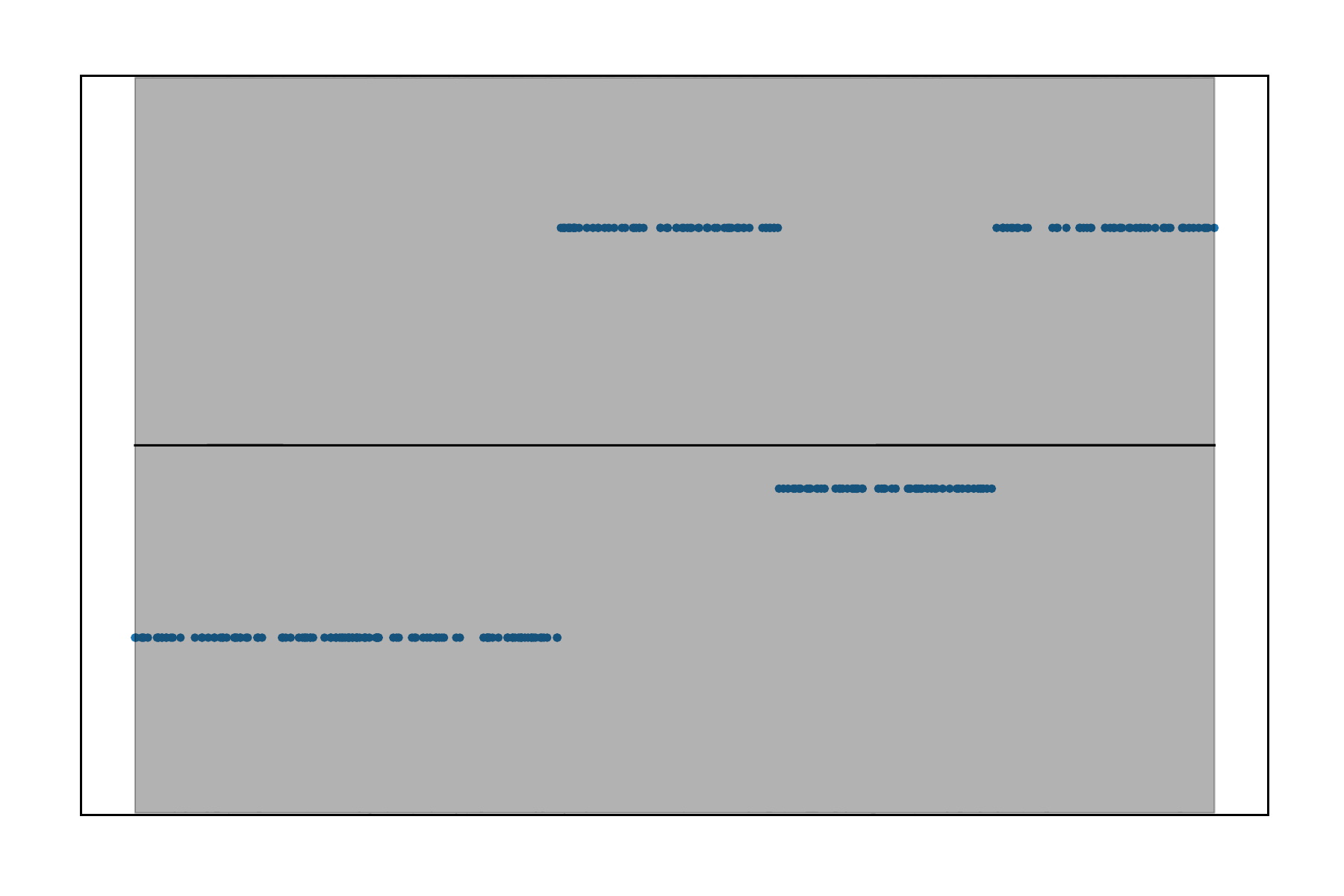}
    \end{subfigure}
\end{minipage}

\begin{minipage}[c]{0.05\textwidth}
    \centering
    \rotatebox{90}{\textbf{20 Inducing}}
\end{minipage}%
\begin{minipage}[c]{0.93\textwidth}
    \centering

    \begin{subfigure}[c]{0.32\linewidth}
        \includegraphics[width=\linewidth]{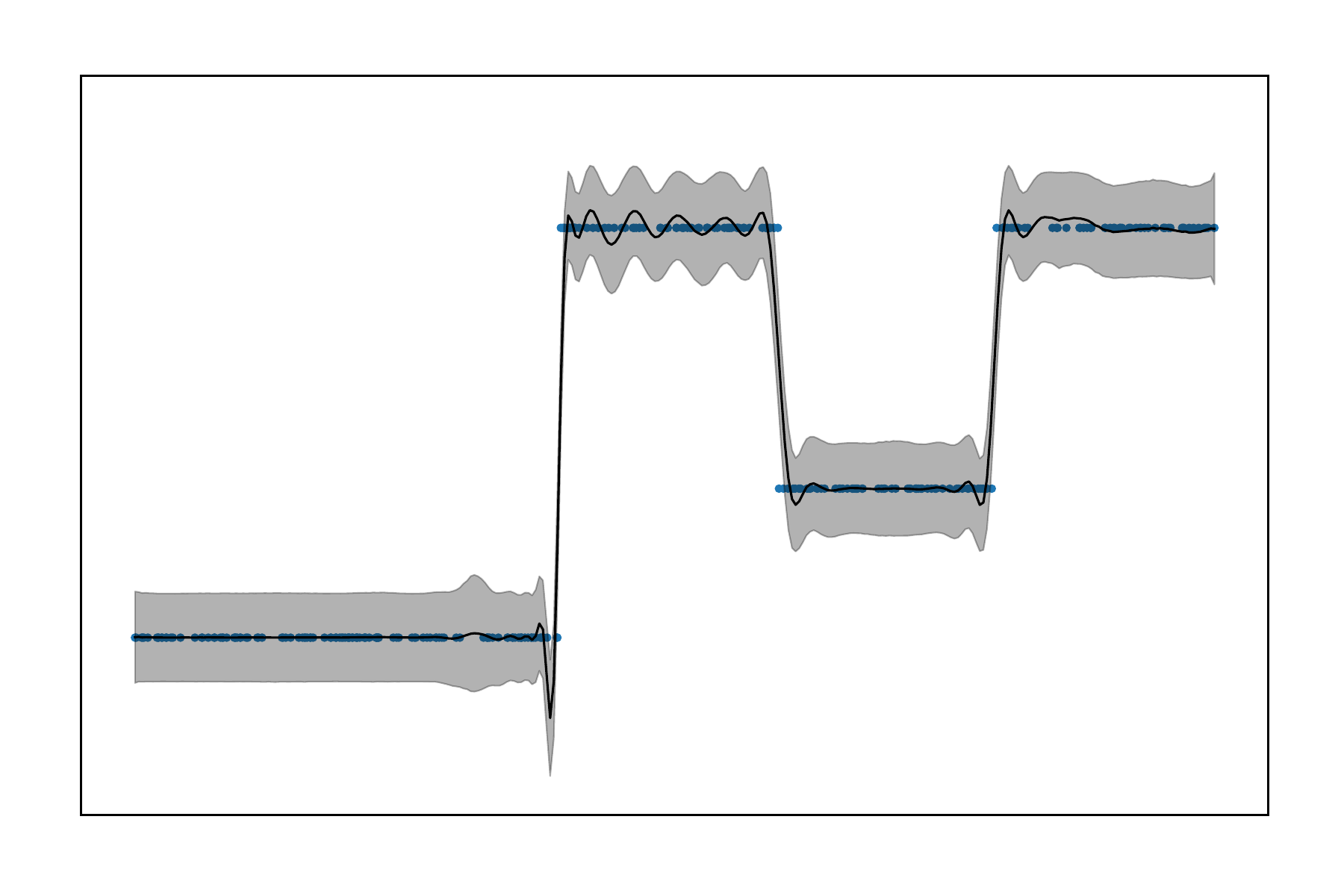}
    \end{subfigure}
    \begin{subfigure}[c]{0.32\linewidth}
        \includegraphics[width=\linewidth]{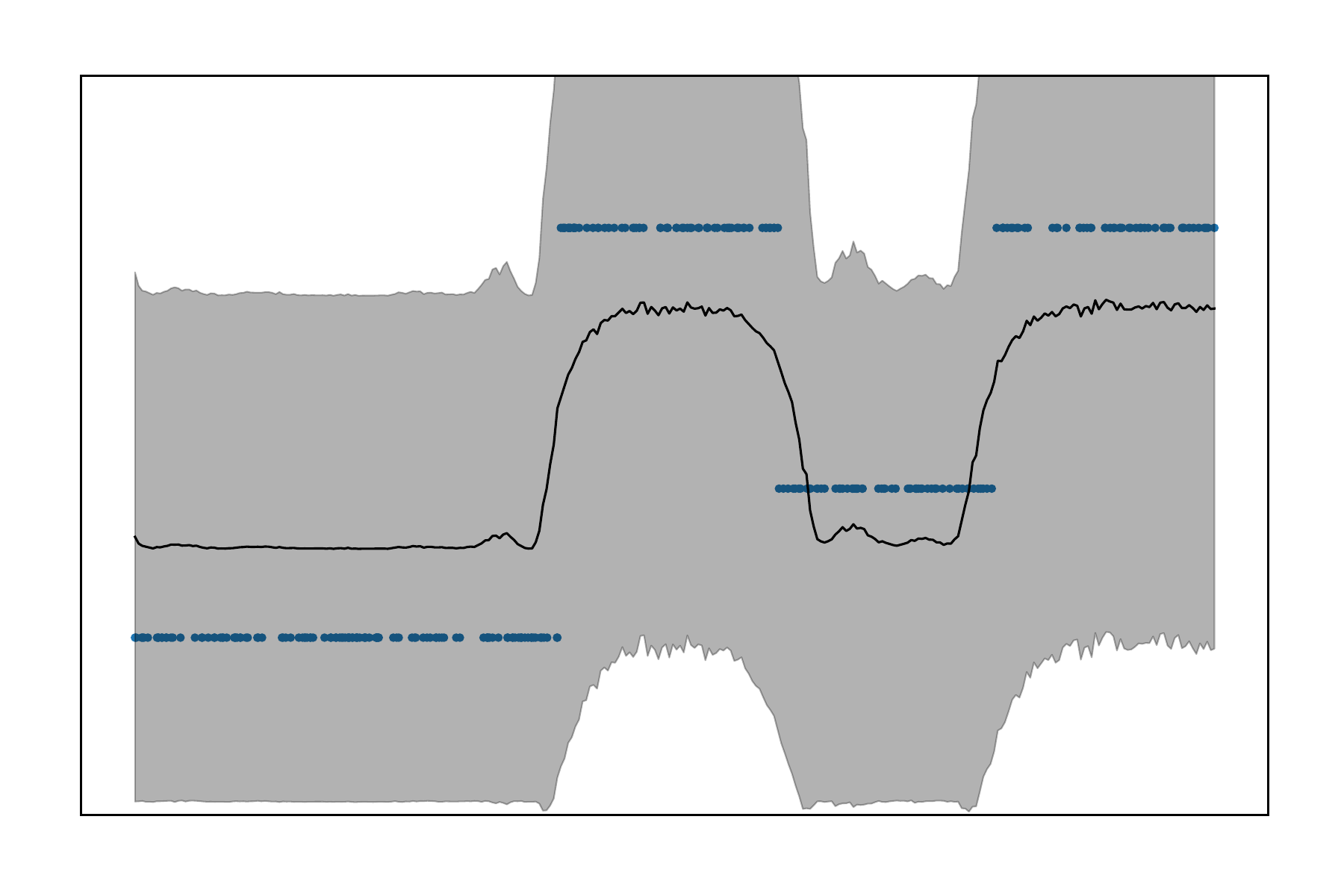}
    \end{subfigure}
    \begin{subfigure}[c]{0.32\linewidth}
        \includegraphics[width=\linewidth]{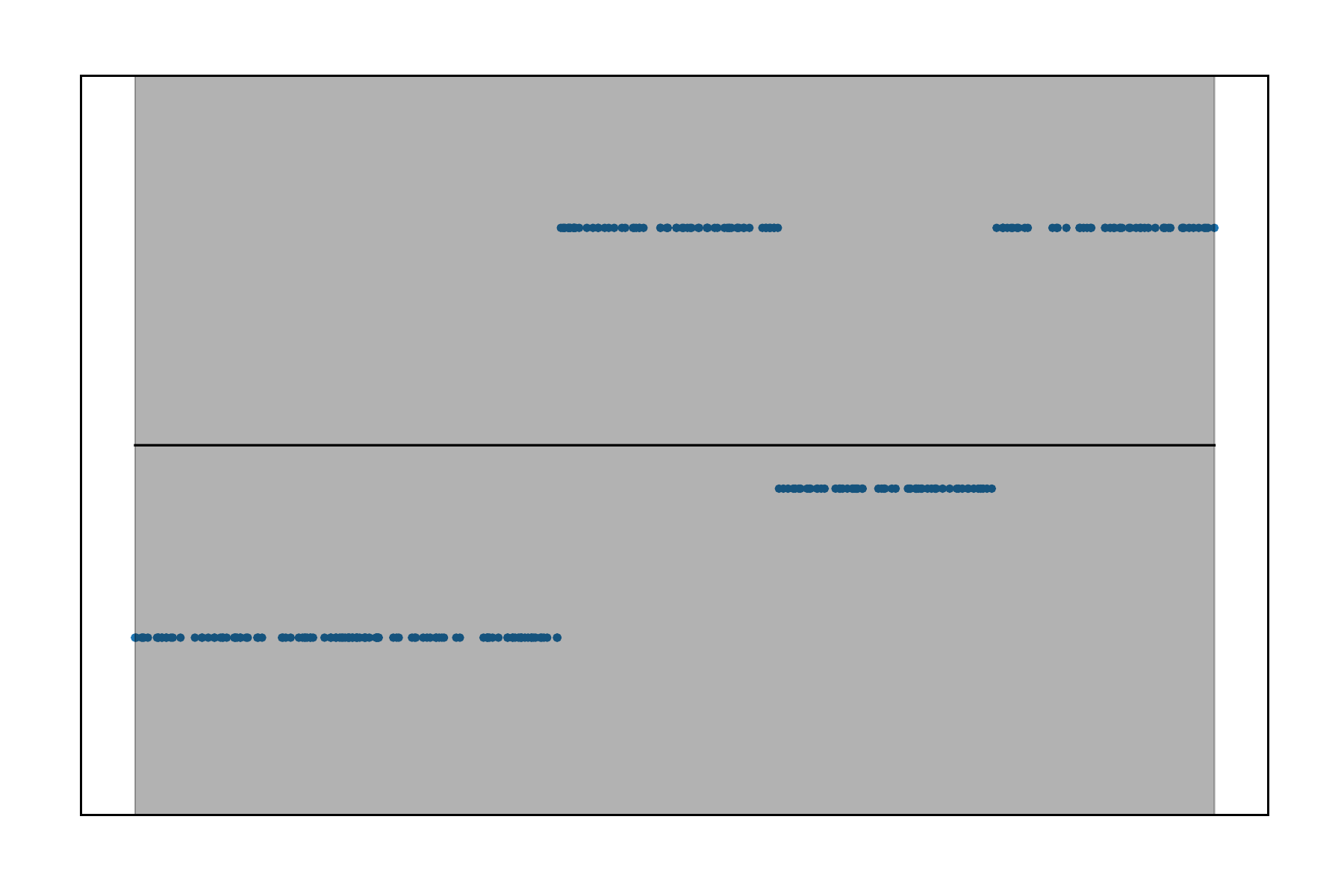}
    \end{subfigure}
\end{minipage}

\begin{minipage}[c]{0.05\textwidth}
    \centering
    \rotatebox{90}{\textbf{100 Inducing}}
\end{minipage}%
\begin{minipage}[c]{0.93\textwidth}
    \centering

    \begin{subfigure}[c]{0.32\linewidth}
        \includegraphics[width=\linewidth]{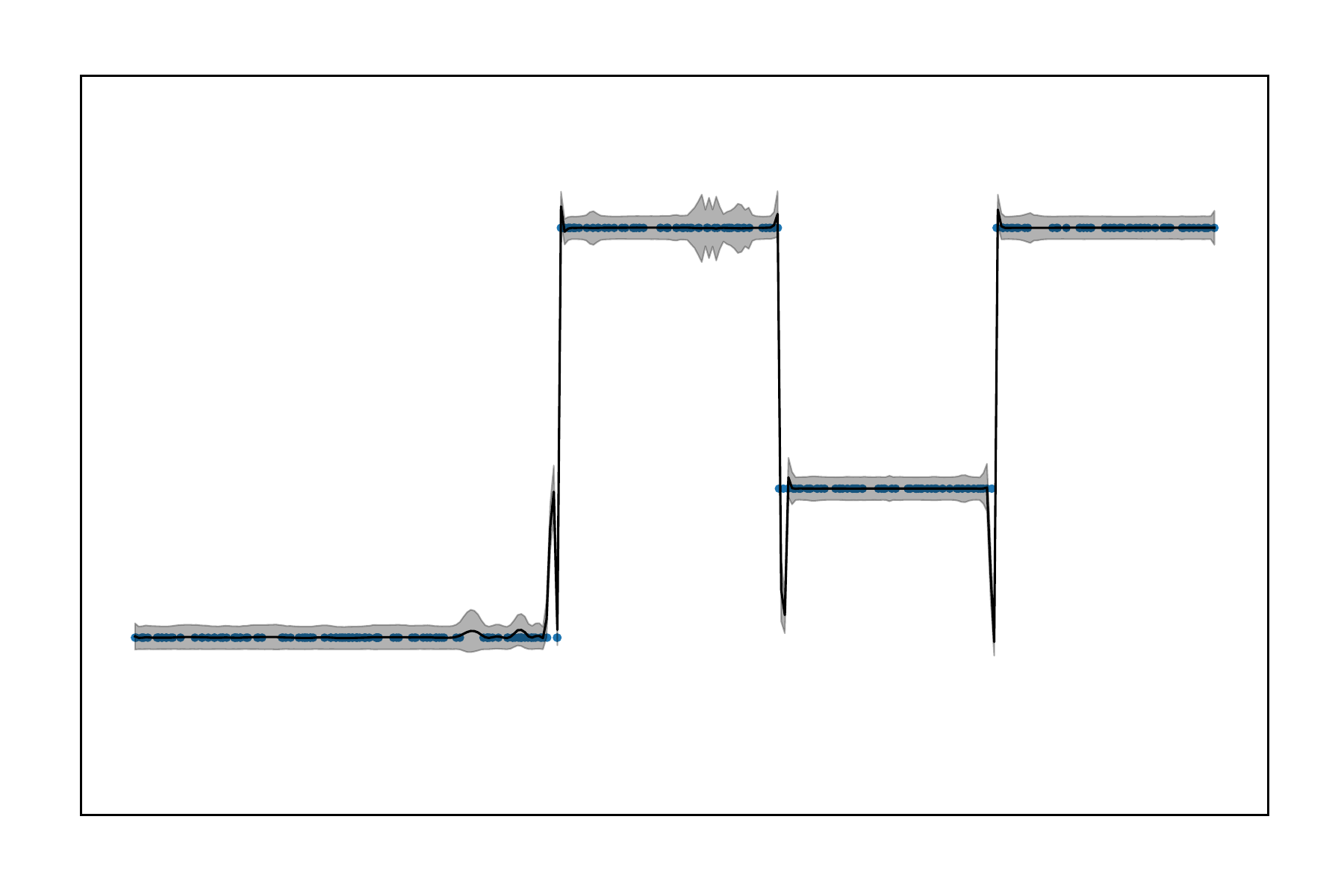}
    \end{subfigure}
    \begin{subfigure}[c]{0.32\linewidth}
        \includegraphics[width=\linewidth]{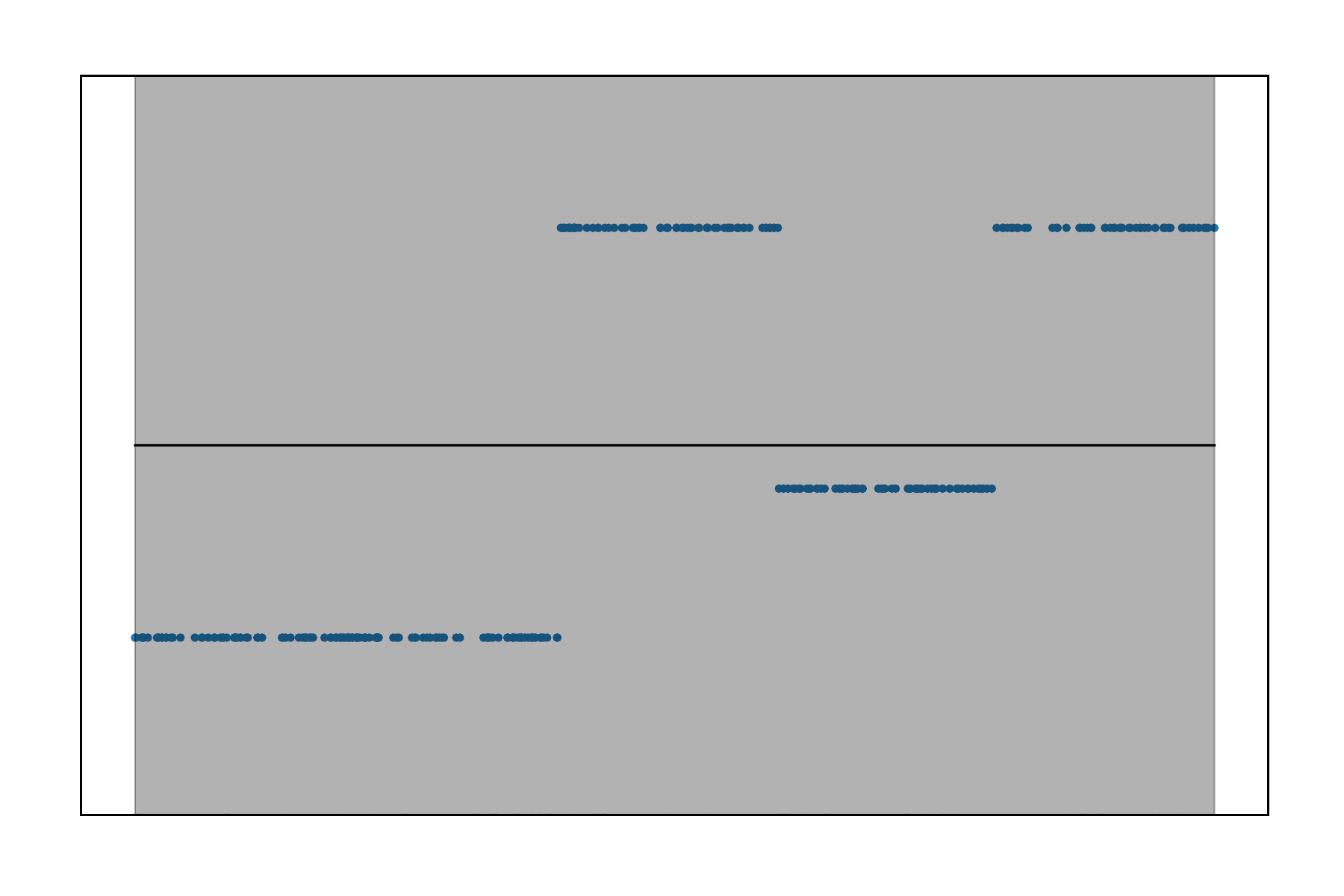}
    \end{subfigure}
    \begin{subfigure}[c]{0.32\linewidth}
        \includegraphics[width=\linewidth]{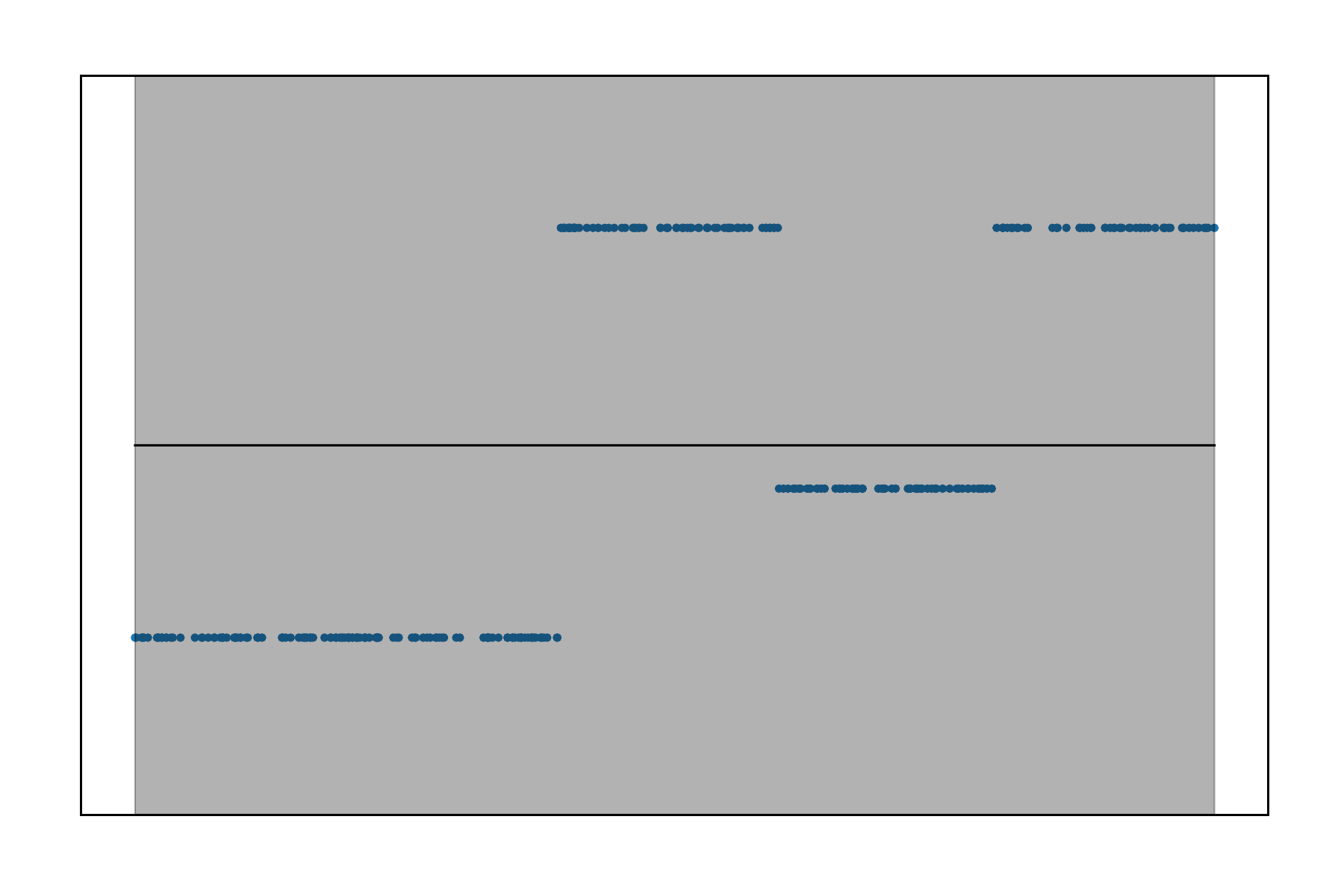}
    \end{subfigure}
\end{minipage}

\begin{minipage}[c]{\textwidth}
\makebox[\linewidth][c]{\textbf{\ZEROW{}}}
\end{minipage}

\begin{minipage}[c]{0.05\textwidth}
    \centering
    \rotatebox{90}{\textbf{5 Inducing}}
\end{minipage}%
\begin{minipage}[c]{0.93\textwidth}
    \centering

    \begin{subfigure}[c]{0.32\linewidth}
        \includegraphics[width=\linewidth]{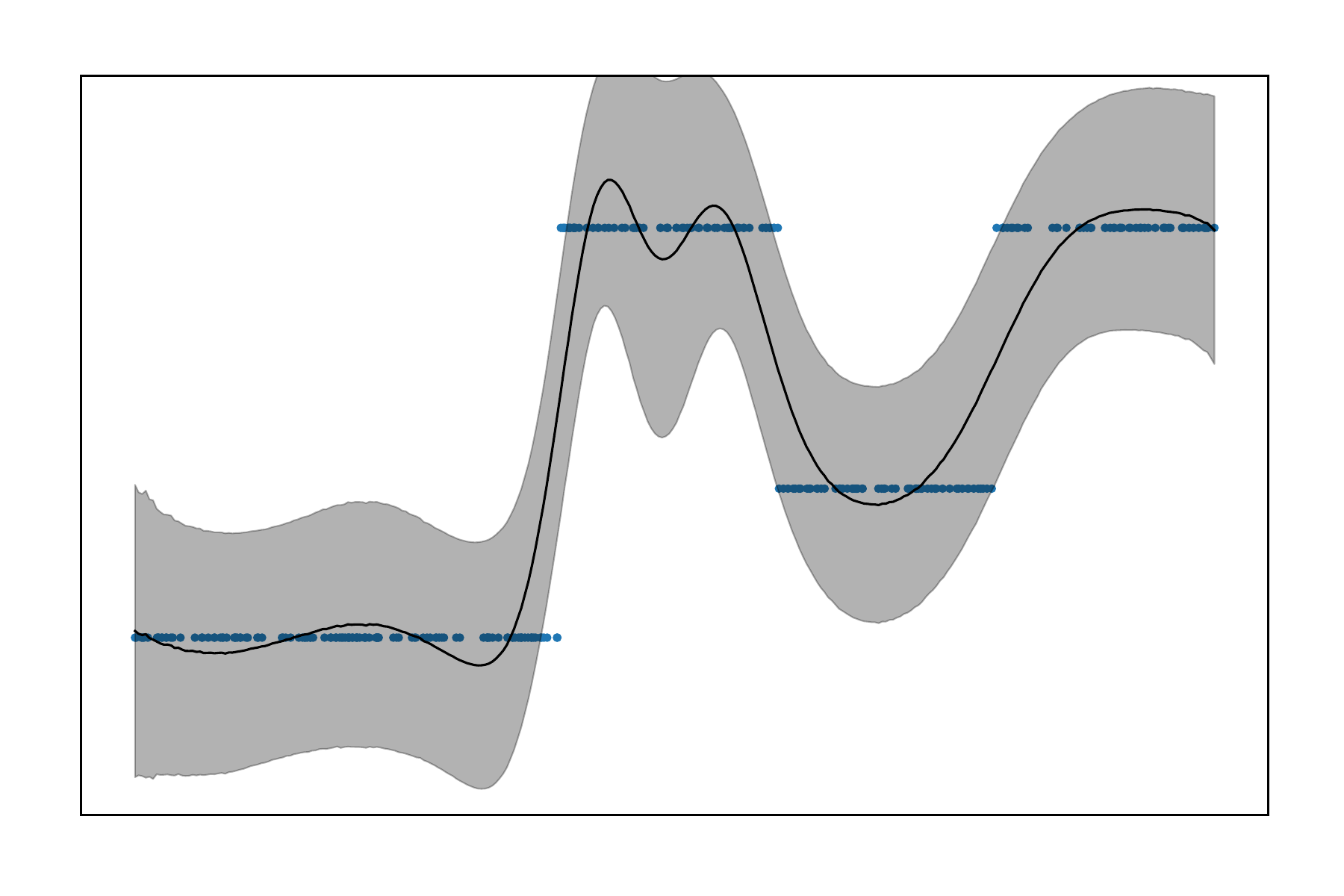}
    \end{subfigure}
    \begin{subfigure}[c]{0.32\linewidth}
        \includegraphics[width=\linewidth]{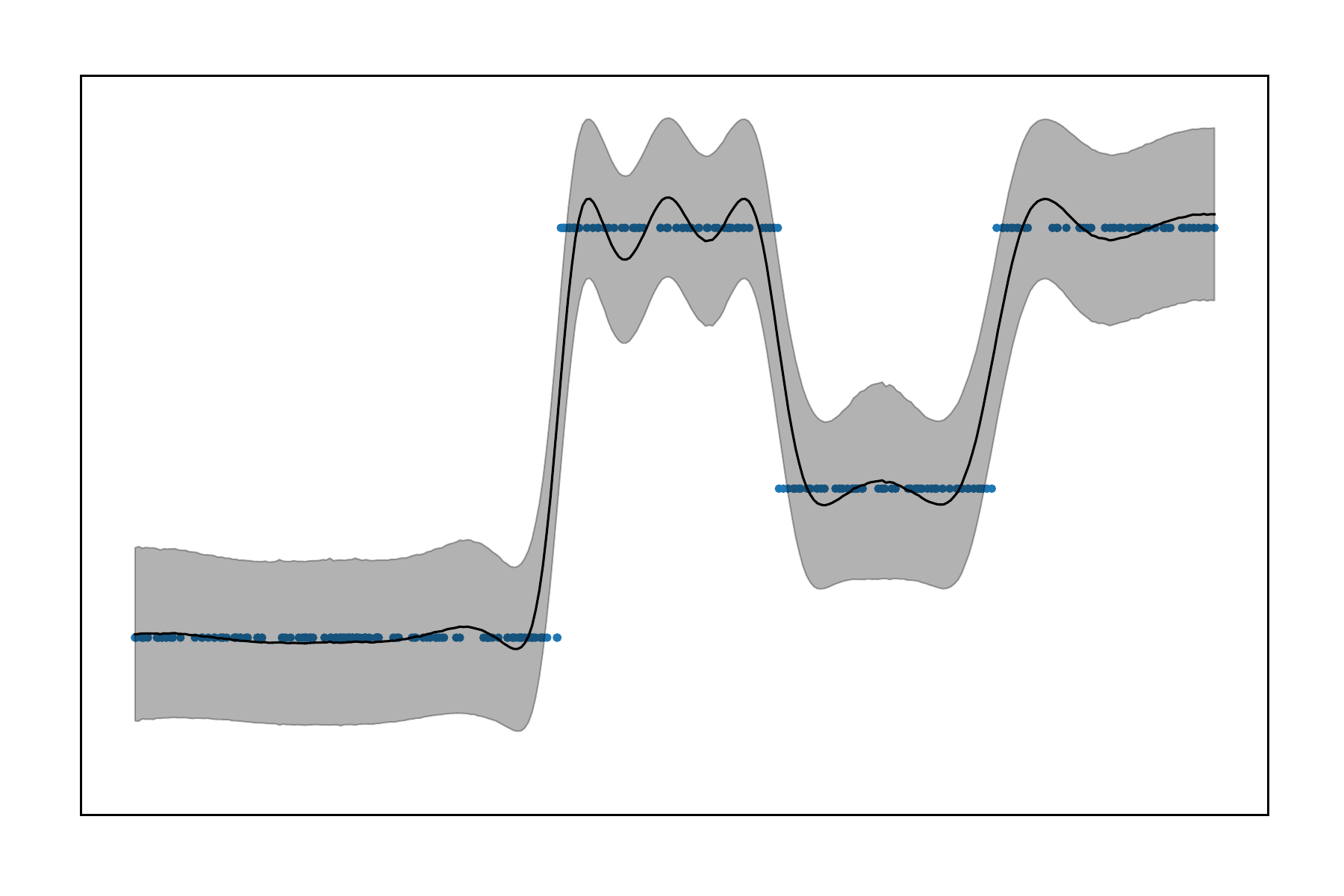}
    \end{subfigure}
    \begin{subfigure}[c]{0.32\linewidth}
        \includegraphics[width=\linewidth]{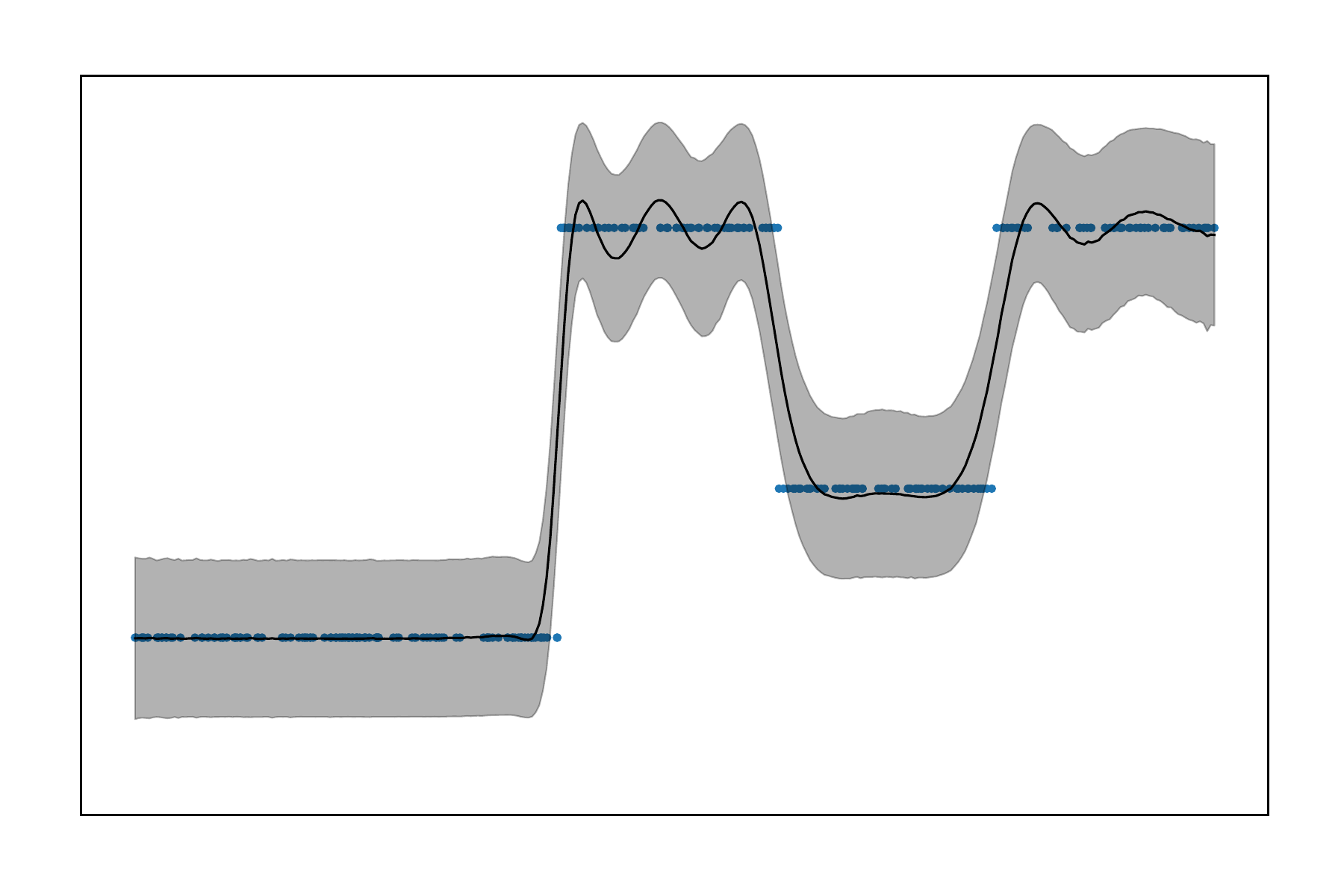}
    \end{subfigure}
\end{minipage}

\begin{minipage}[c]{0.05\textwidth}
    \centering
    \rotatebox{90}{\textbf{20 Inducing}}
\end{minipage}%
\begin{minipage}[c]{0.93\textwidth}
    \centering

    \begin{subfigure}[c]{0.32\linewidth}
        \includegraphics[width=\linewidth]{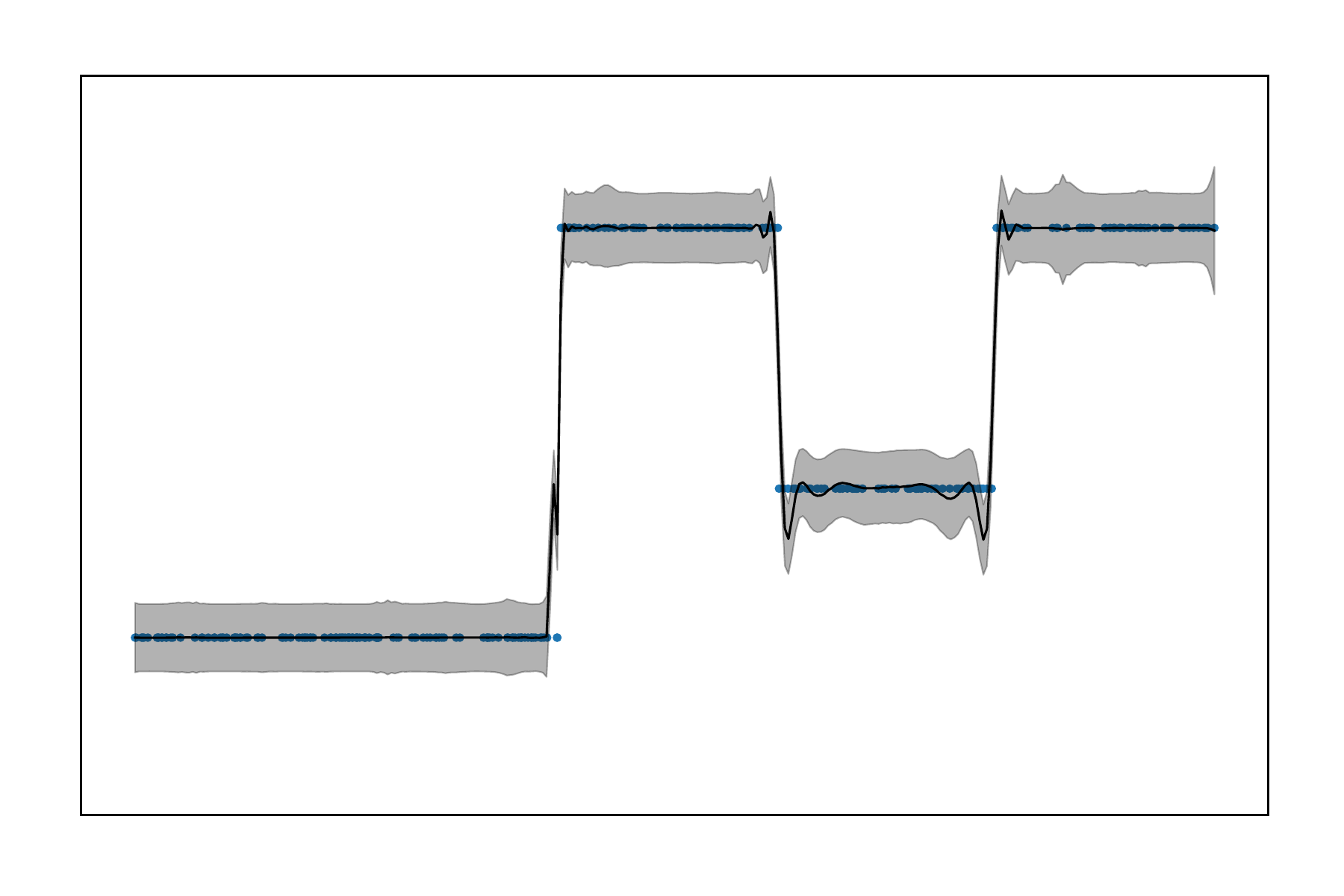}
    \end{subfigure}
    \begin{subfigure}[c]{0.32\linewidth}
        \includegraphics[width=\linewidth]{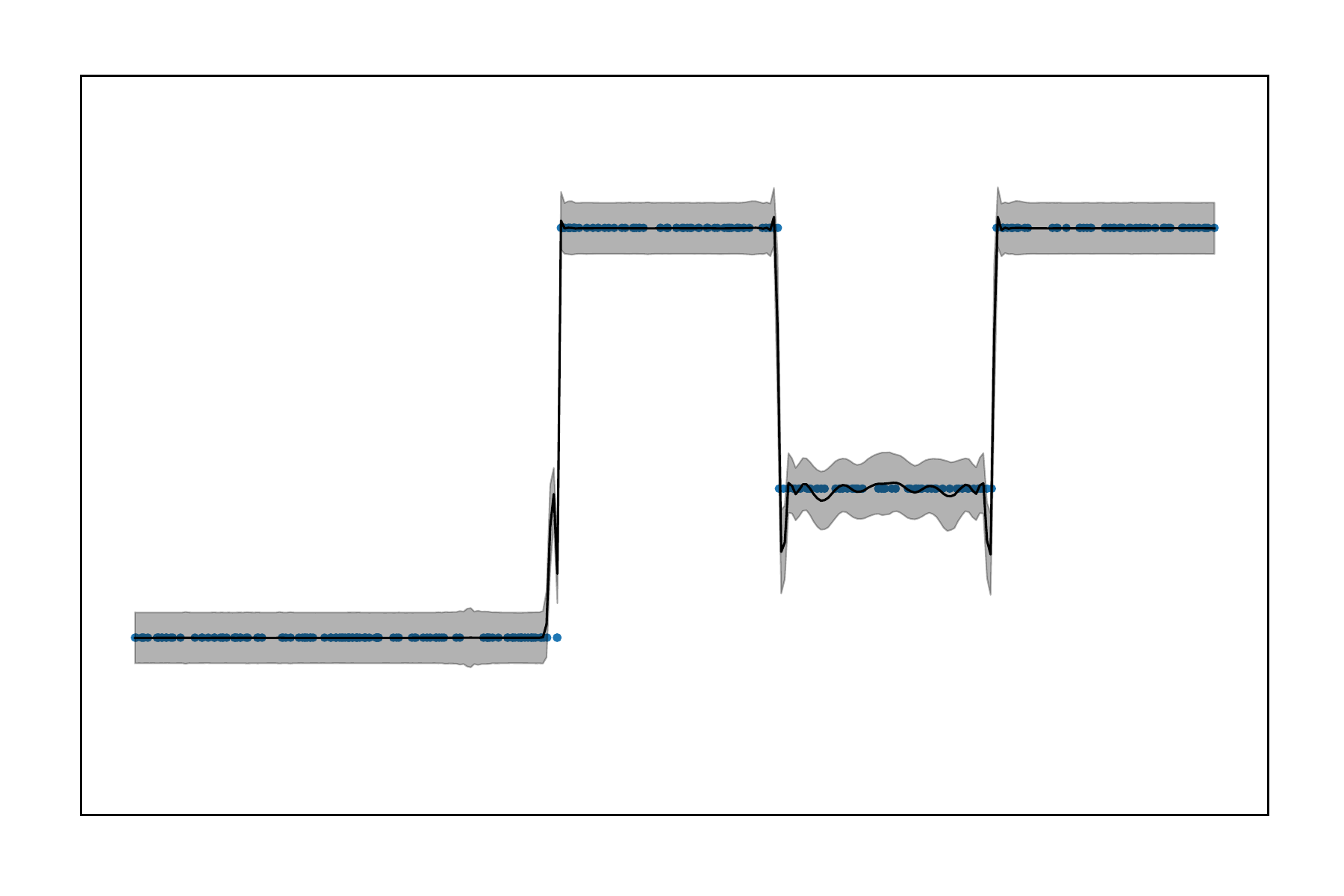}
    \end{subfigure}
    \begin{subfigure}[c]{0.32\linewidth}
        \includegraphics[width=\linewidth]{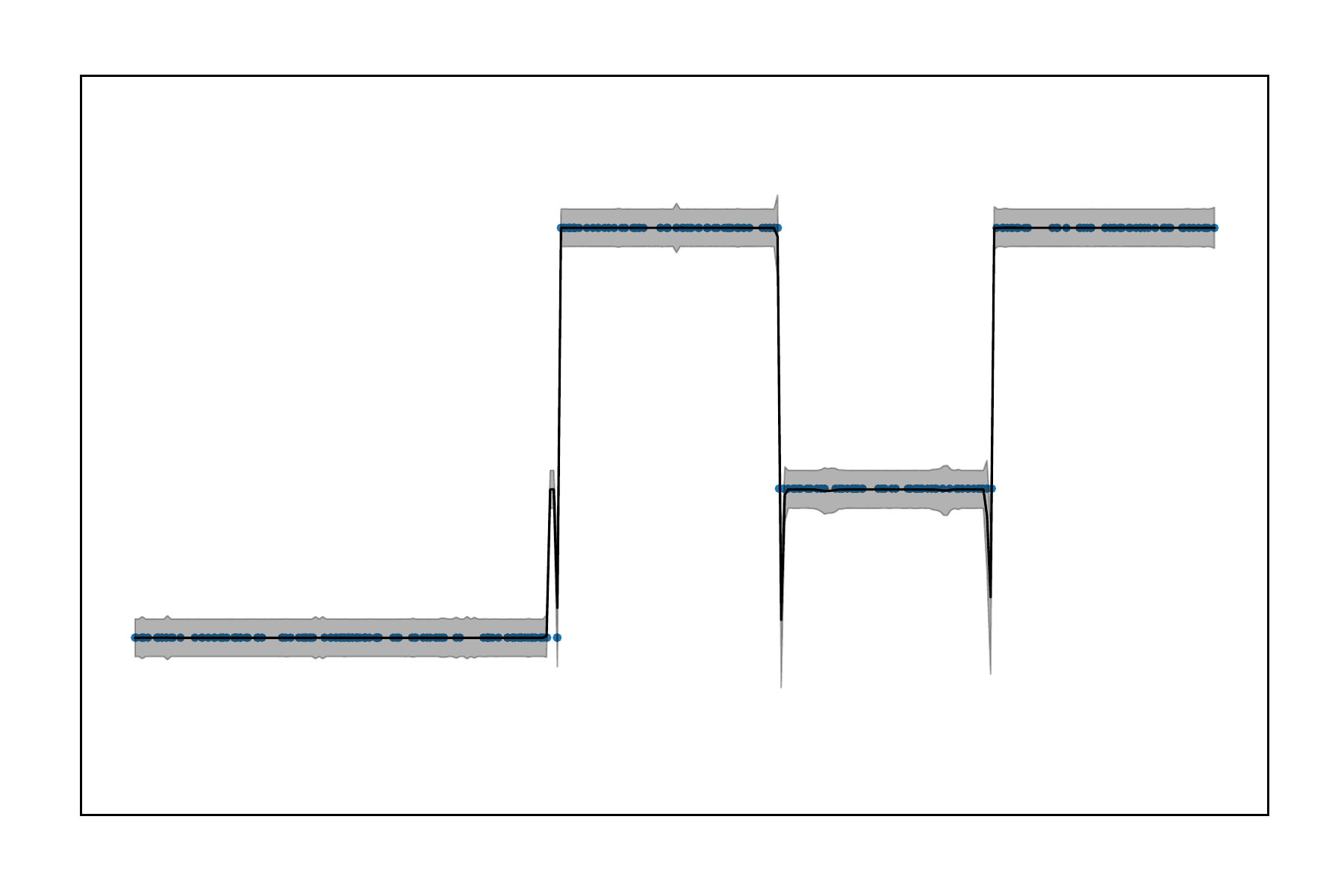}
    \end{subfigure}
\end{minipage}

\begin{minipage}[c]{0.05\textwidth}
    \centering
    \rotatebox{90}{\textbf{100 Inducing}}
\end{minipage}%
\begin{minipage}[c]{0.93\textwidth}
    \centering

    \begin{subfigure}[c]{0.32\linewidth}
        \includegraphics[width=\linewidth]{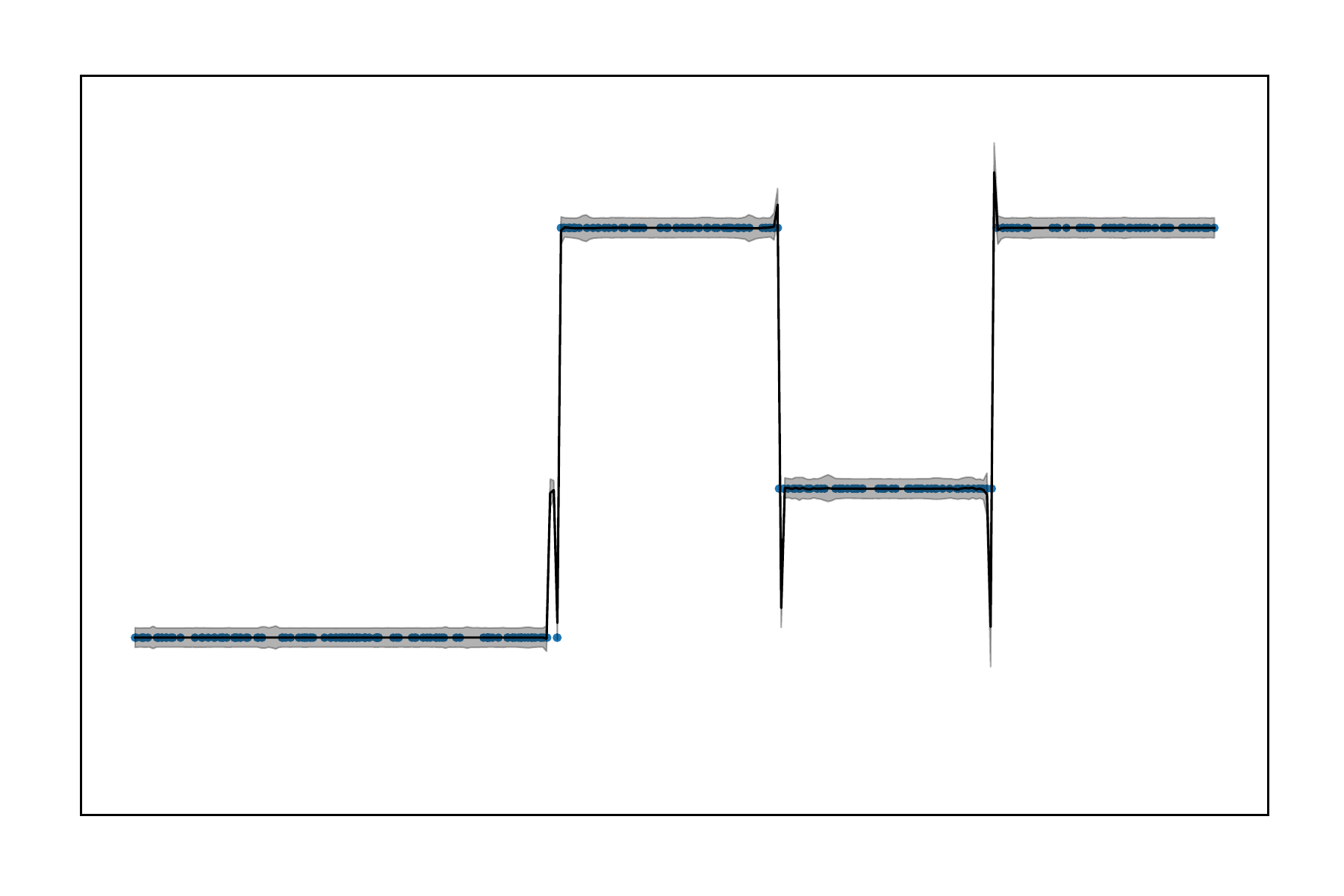}
    \end{subfigure}
    \begin{subfigure}[c]{0.32\linewidth}
        \includegraphics[width=\linewidth]{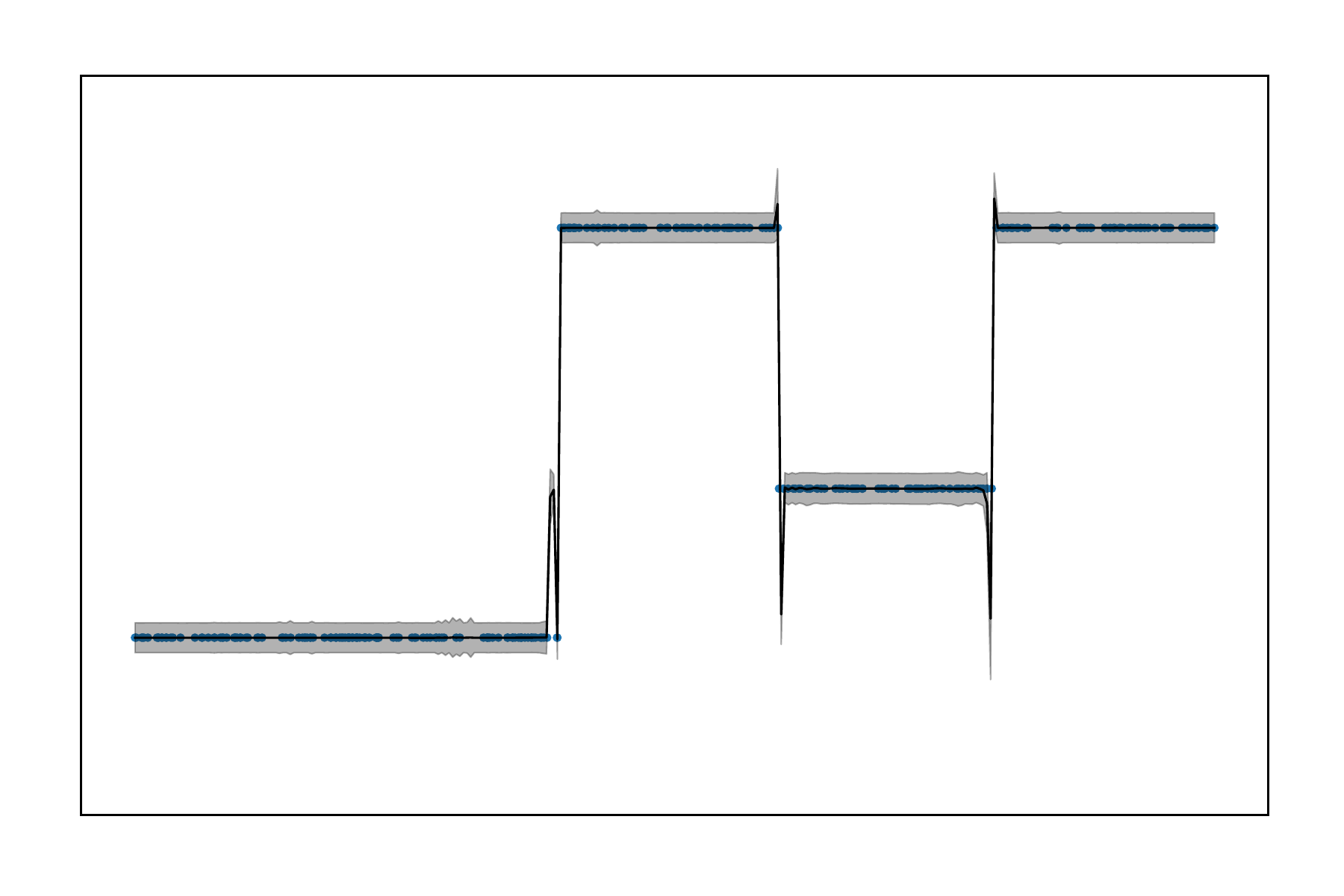}
    \end{subfigure}
    \begin{subfigure}[c]{0.32\linewidth}
        \includegraphics[width=\linewidth]{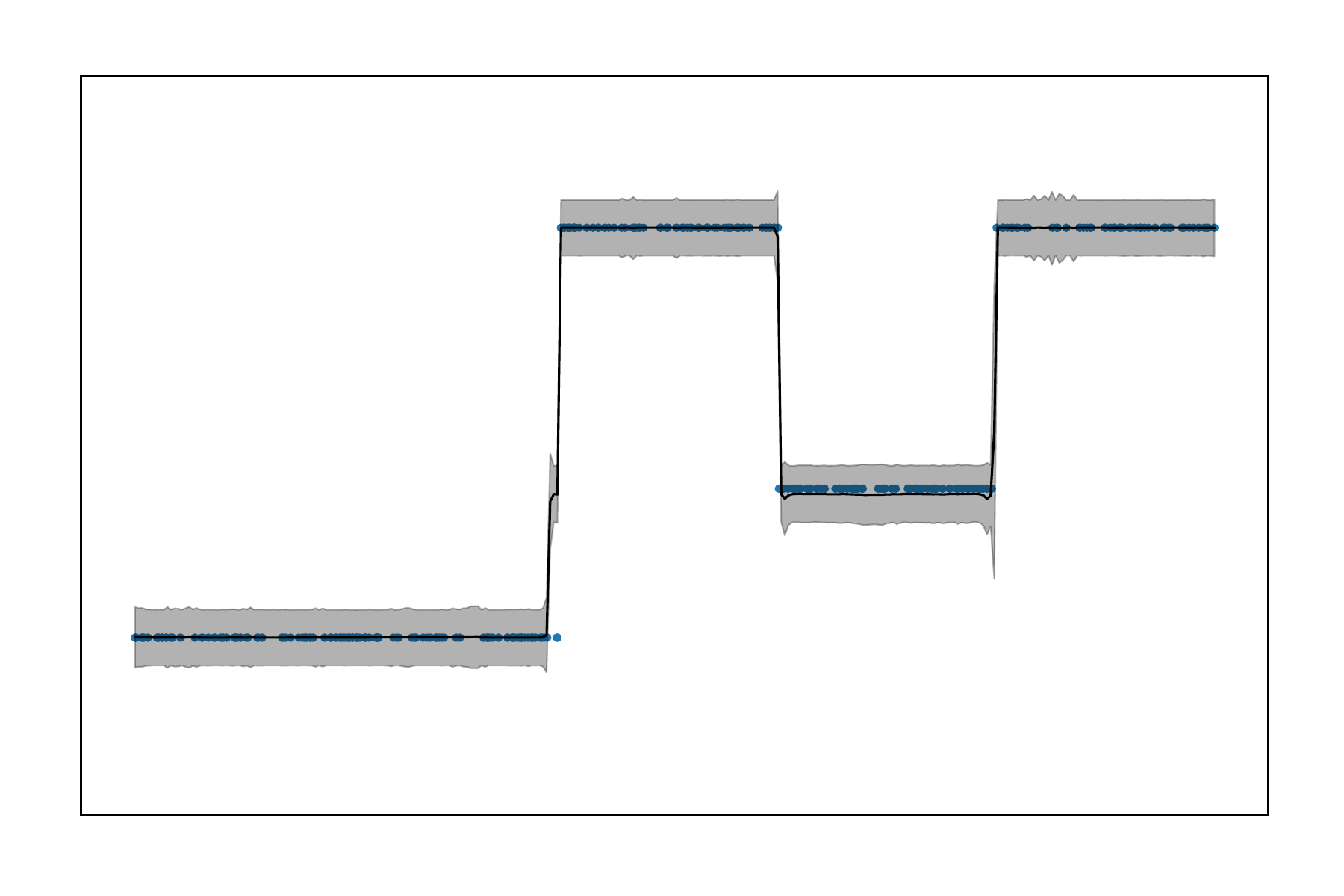}
    \end{subfigure}
\end{minipage}

\begin{minipage}[c]{\textwidth}
\makebox[\linewidth][c]{\textbf{\PCAW{}}}
\end{minipage}

\caption{Predictive distribution of both \ZEROW{} prior mean \DGP (top 3 rows) and the \PCAW{} prior mean \DGP (bottom 3 rows) initialized with $\varSw{l}{} = \mathbf{I}$ in the inner layers and $\varSw{L}{} = 10^{-5}\mathbf{I}$ in the output layer. Blue points show training data. Using a high variance on the inner layers leads to a noisy optimization which yields poor solutions.}
\label{fig:more_inducings_inner_high_output_low}
\end{figure}

\item $\varS{}{l}=\mathbf{I}$ and $\varS{}{L} = \mathbf{I}$: For completeness, the pictures from this configuration are provided in Appendix \ref{sec:app:c:2:toy:optimization}.
\end{itemize}
Overall, this subsection corroborates our analysis carried out in \usec \ref{sec:optimization:difficulties}, confirming that using smaller initial variational variances $\varS{}{}$ and a bigger number of inducing points favors avoiding posterior collapse. Moreover, injected noise in the optimization results in suboptimal fitted models. In this case, the injected noise came either from a bigger initial $\varS{l}{}$ or a smaller number of inducing points, which makes the inner layer predictive variance take the value of the kernel output scale parameter $\sigma_o$ at locations far from the inducing points.
\subsubsection{Evaluating a Very Deep \DGP Model} 

\begin{figure}[htbp]
    \centering



    \begin{minipage}[c]{0.05\textwidth}
    \end{minipage}%
    \begin{minipage}[c]{0.93\textwidth}
        \centering
        \hfill
        \makebox[0.32\linewidth][c]{\textbf{Predictive distribution}}%
        \hfill
        \makebox[0.32\linewidth][c]{\textbf{Layer-wise \KLD{}}}%
        \hfill
        \makebox[0.32\linewidth][c]{\quad\quad\quad \textbf{Likelihood Variance}}%
    \end{minipage}
    
    \begin{minipage}[c]{0.05\textwidth}
        \centering
        \rotatebox{90}{\textbf{$\sigma = 1$}}
    \end{minipage}%
    \begin{minipage}[c]{0.93\textwidth}
        \centering
        \begin{subfigure}[c]{0.32\linewidth}
            \includegraphics[width=\linewidth]{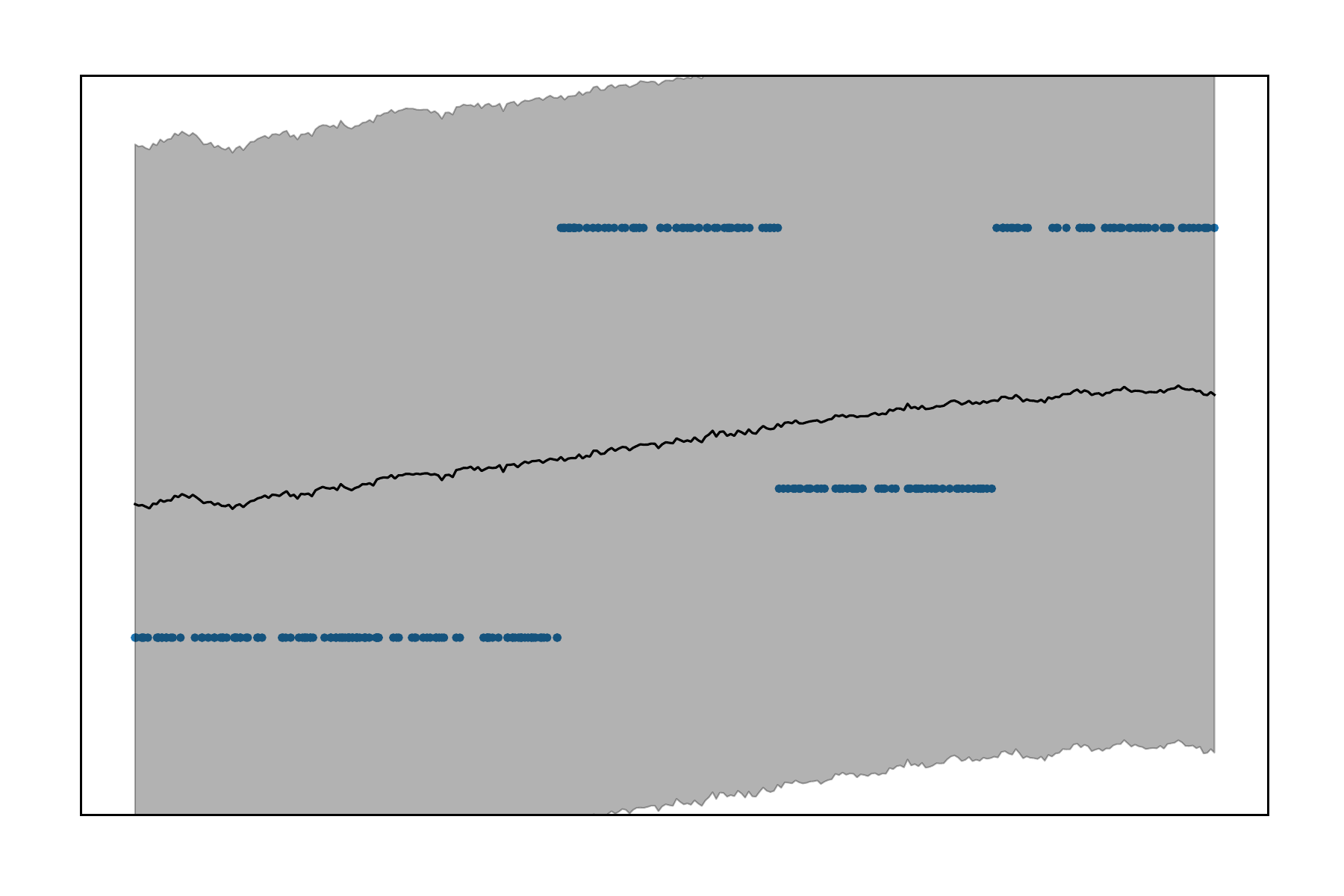}
        \end{subfigure}
        \hfill
        \begin{subfigure}[c]{0.32\linewidth}
            \includegraphics[width=\linewidth]{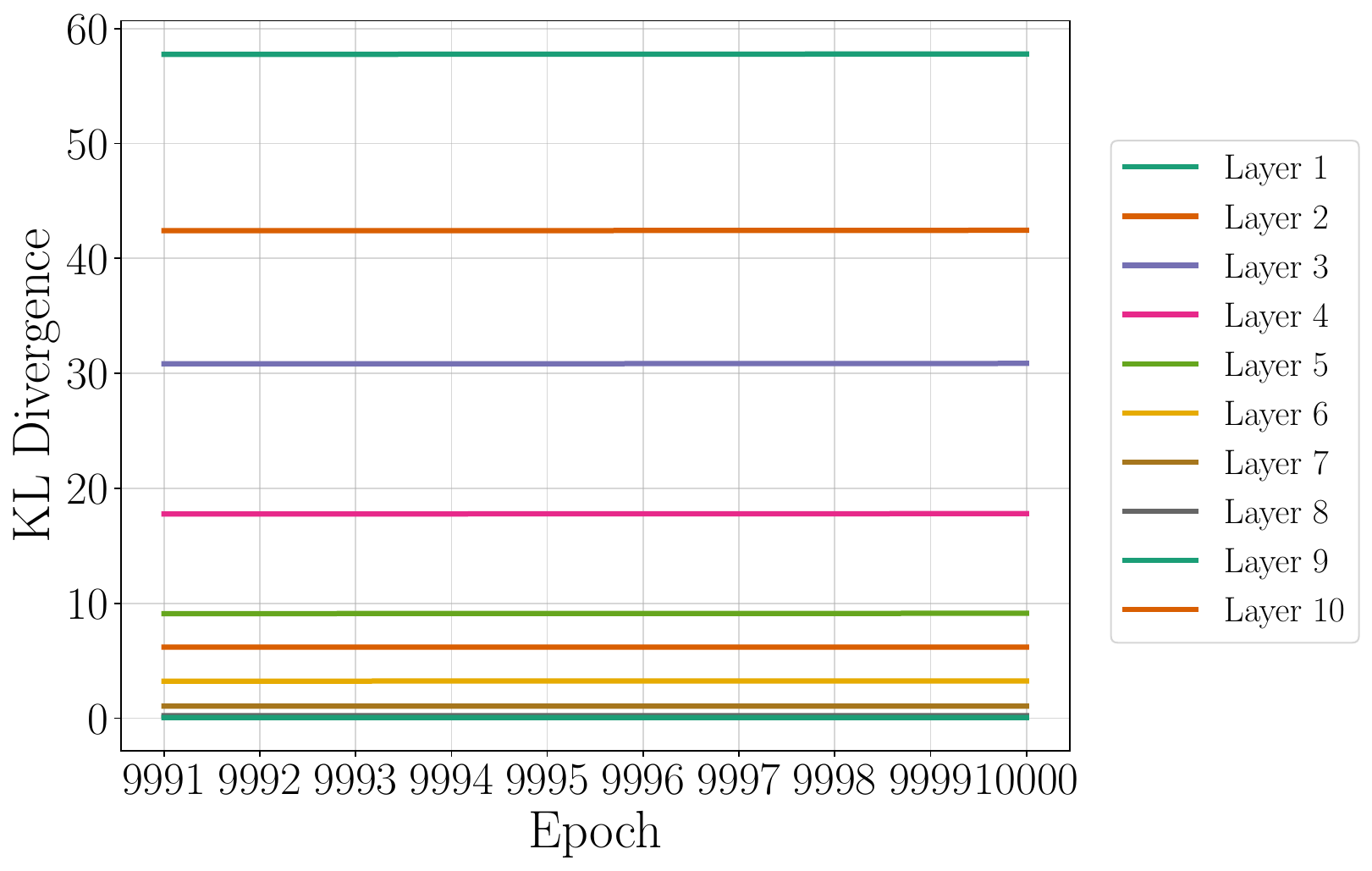}
        \end{subfigure}
        \hfill
        \begin{subfigure}[c]{0.32\linewidth}
            \includegraphics[width=\linewidth]{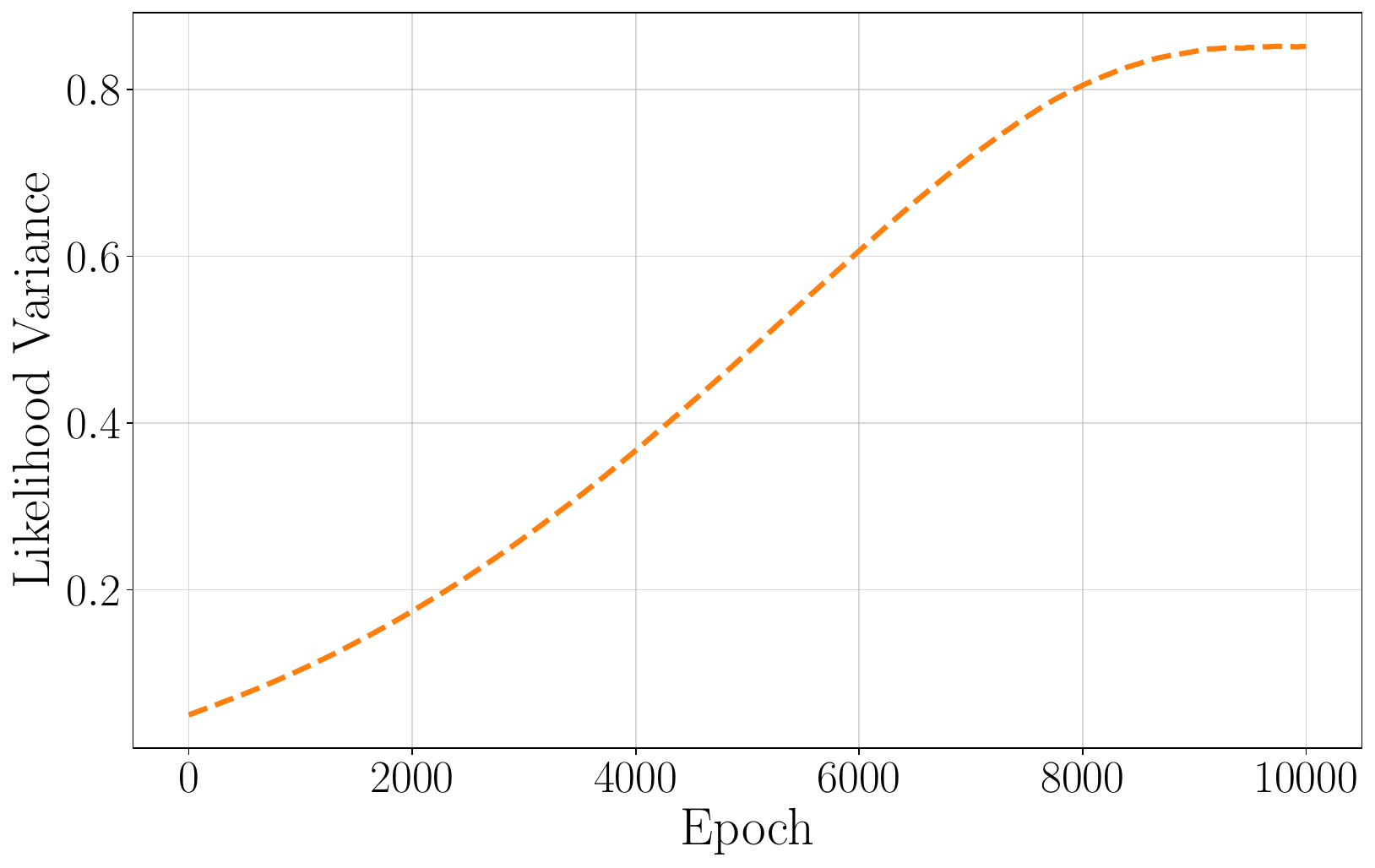}
        \end{subfigure}
    \end{minipage}
    \vspace{0.25cm} 

    \begin{minipage}[c]{0.05\textwidth}
        \centering
        \rotatebox{90}{\textbf{$\sigma=0.1$}}
    \end{minipage}%
    \begin{minipage}[c]{0.93\textwidth}
        \centering
        \begin{subfigure}[c]{0.32\linewidth}
            \includegraphics[width=\linewidth]{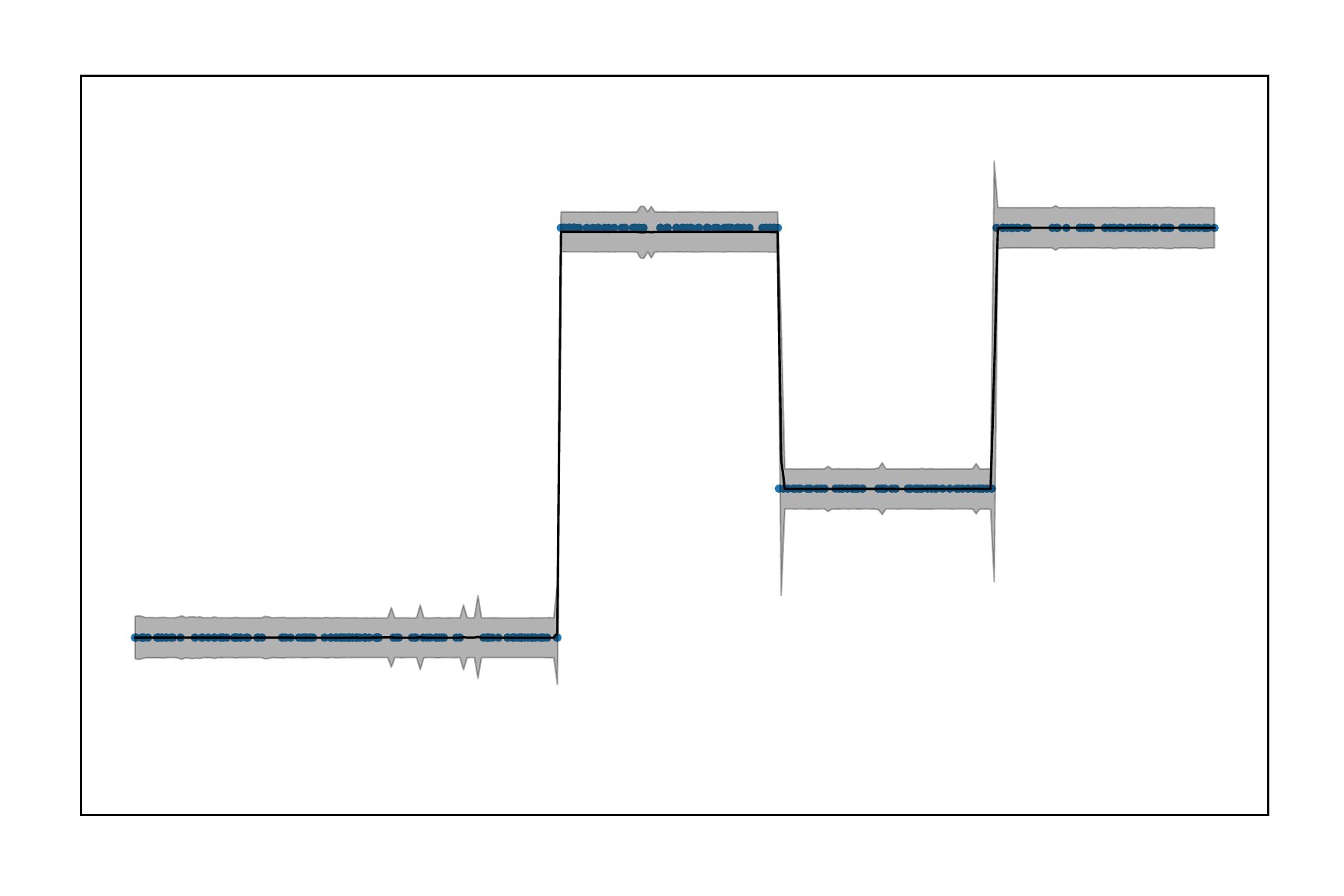}
        \end{subfigure}
        \hfill
        \begin{subfigure}[c]{0.32\linewidth}
            \includegraphics[width=\linewidth]{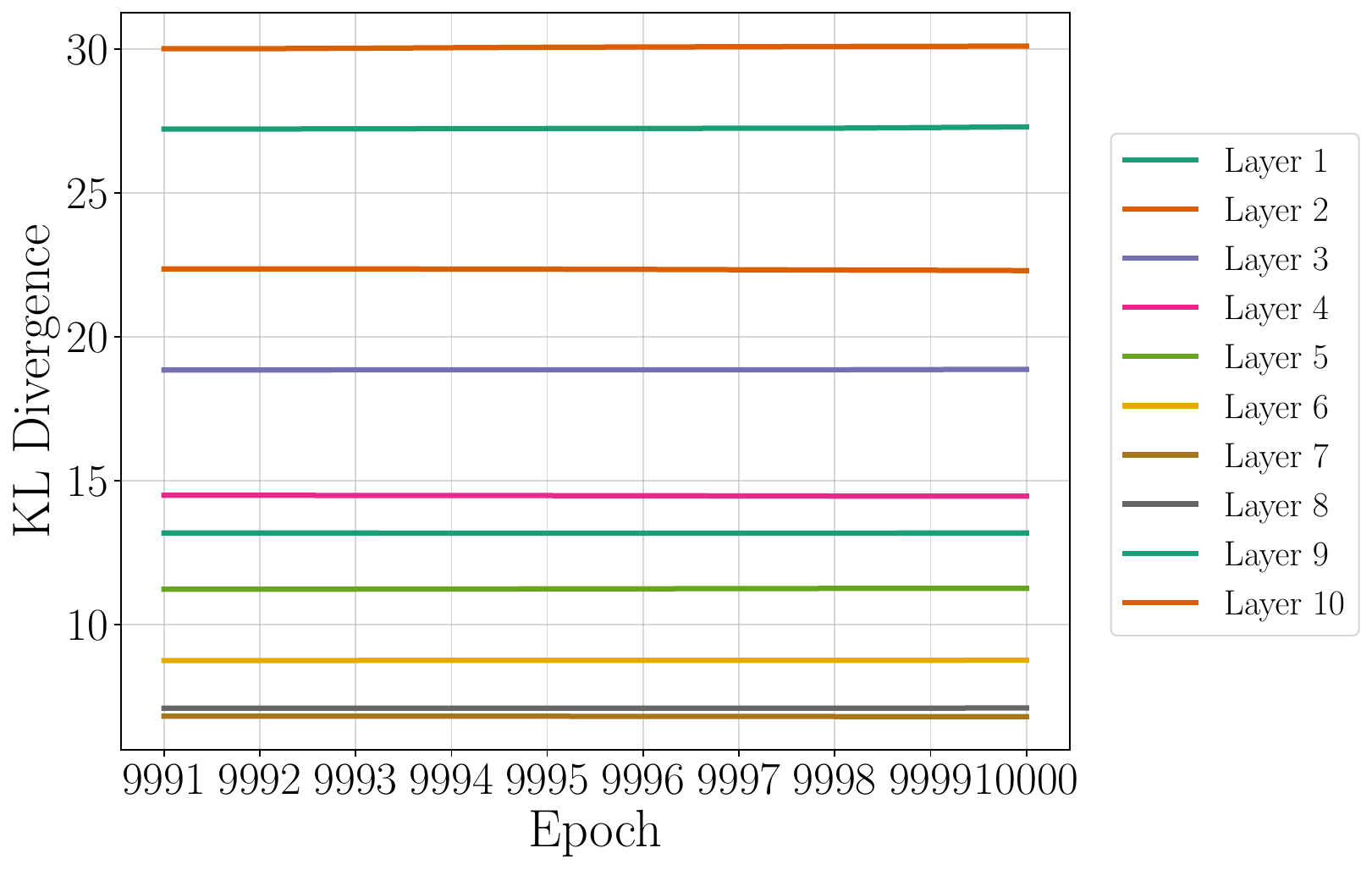}
        \end{subfigure}
        \hfill
        \begin{subfigure}[c]{0.32\linewidth}
            \includegraphics[width=\linewidth]{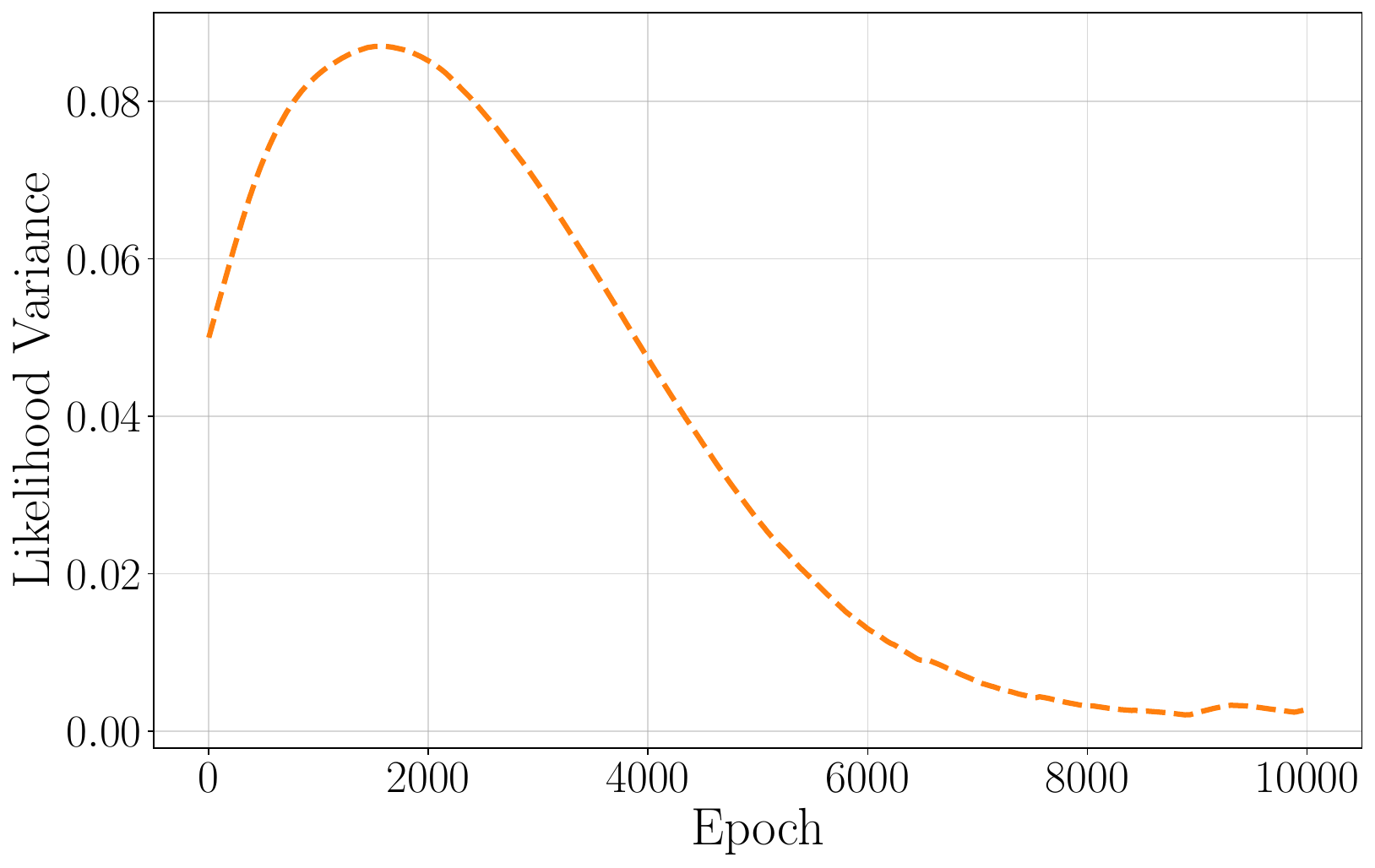}
        \end{subfigure}
    \end{minipage}
    \begin{minipage}[c]{\textwidth}
        \makebox[\linewidth][c]{\textbf{10 layer \PCAW{}}}%
    \end{minipage}

    \caption{(top-row) Predictive distribution, layer-wise \KLD{} and noise variance during training of the 10-layer \PCAW{} model with $20$ inducing points. The kernel output scale parameter $\sigma$ is initialized to $1.0$. (bottom-row) Same information, but when $\sigma$ is initialized to $0.1$. We observe that $\sigma=0.1$ results in a better model. By contrast, $\sigma=1$ results in a model with near posterior collapse.}
    \label{fig:duvenaud_collapse}
\end{figure}

The previous results show that the \PCAW{} model is robust to posterior collapse in the evaluated configurations. 
However, we also observed that the models evaluated are more likely to suffer from posterior collapse when the number of 
layers increases and the number of inducing points is reduced.  Therefore, in this section we train a $10$-layer \PCAW{} \DGP model with $20$ inducing points on the 
toy problem using $\varSw{l}{}=\varSw{L}{}=10^{-5}\matI$. The predictive distribution obtained is shown in the top-left corner of \fig~\ref{fig:duvenaud_collapse}. 
We observe that the model is close to posterior collapse. This is supported by the zero \KLD{} shown in some of its hidden layers 
during the last training epochs, as displayed in the top-middle sub-figure. Moreover, the noise variance, displayed in the 
top-right sub-figure, is also increasing with each training epoch, suggesting that the model is simply explaining the observed data as noise.

We hypothesize that the observed behavior may be due to a big initial prior variance, as specified by the kernel output scale parameter, \ie, $\sigma_o$ in \ueqn~\eqref{eq:kernel:rbf}. 
The bottom row of \fig~\ref{fig:duvenaud_collapse} shows that when this initial parameter is reduced, the model does not suffer posterior collapse and learns to perfectly 
fit the observed data. Since there are few inducing points in this setting, far from them, the model will output the prior variance, which is big when $\sigma_o = 1.0$. This 
introduces noise in the optimization procedure, resulting in suboptimal fitting. By reducing the kernel output scale, we reduce this noise, which results in more effective training of the model. 

Summing up, these experiments show that the \PCA prior mean function with a stable optimization procedure coming from the whitened parameterization might not necessarily avoid the posterior collapse problem, nor the non-injective pathologies \citep{pathologies_vdn}, and also that
setting smaller initial prior variances may be beneficial to avoid the posterior collapse problem. This further confirms the result from the previous section, where injected noise in the optimization procedure drives the model to collapse.

\subsubsection{Results for Each \DGP Model and Initialization Considered}

We now proceed with a detailed performance comparison of all models in this dataset. The results obtained here will determine the model 
configurations that are run in the real-world datasets. For this comparison, we use $5$ layers in all the experiments and train the models 
for \(10.000\) epochs, which is enough to achieve convergence of the \ELBO{}. We use $100$ inducing points. We initialize the inducing locations $\Zsamples{}$ using \emph{K-means} for the \ZERO{} and \PCA{} prior mean function models, and we use $\Zsamples{} = \Xx$ for the \MO and \MY models. This means that the inducing points between the \ZERO and \PCA models and the \MO and \MY differ. We use $\varS{}{} = 10^{-5} \matI$ in all layers since attending 
to our first experiments, this configuration is more likely to produce the best results.
Here, we consider the non-whitened parameterization of \GPFLOW, \ie, \NWR{}, and the whitened parameterization, \ie, \W{}. 

\begin{figure}[!t]
    \centering
    \begin{subfigure}[t]{0.32\linewidth}
        \centering
        \footnotesize{ \ZERO output mean, $\varm{}{L} = 0$} \\[0.3em]
         \includegraphics[width=\linewidth]{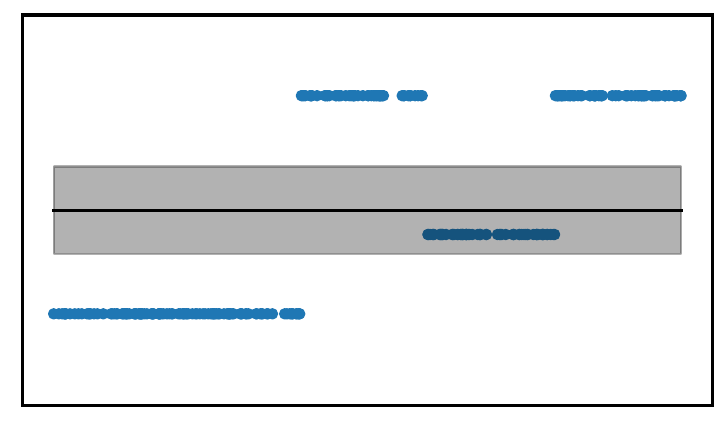}
        \label{fig:init:zero_w}
    \end{subfigure}
    \hfill
    \begin{subfigure}[t]{0.32\linewidth}
        \centering
        \footnotesize{\ZEROWMY} \\[0.3em]
        \includegraphics[width=\linewidth]{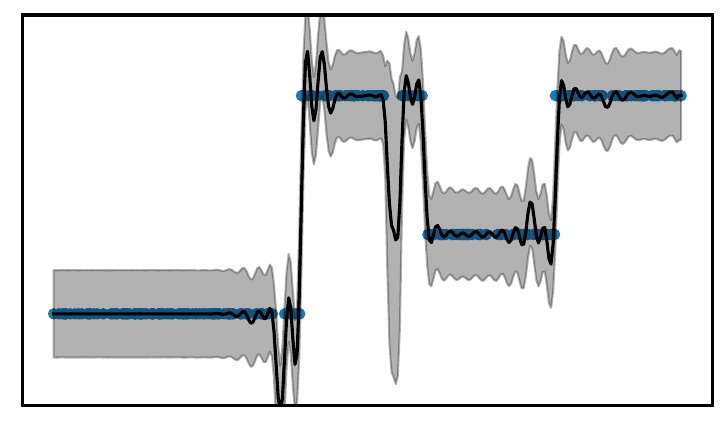}
        \label{fig:init:zero_dmy_w}
    \end{subfigure}
    \hfill
    \begin{subfigure}[t]{0.32\linewidth}
        \centering
        \footnotesize{\ZERONWRMY} \\[0.3em]
        \includegraphics[width=\linewidth]{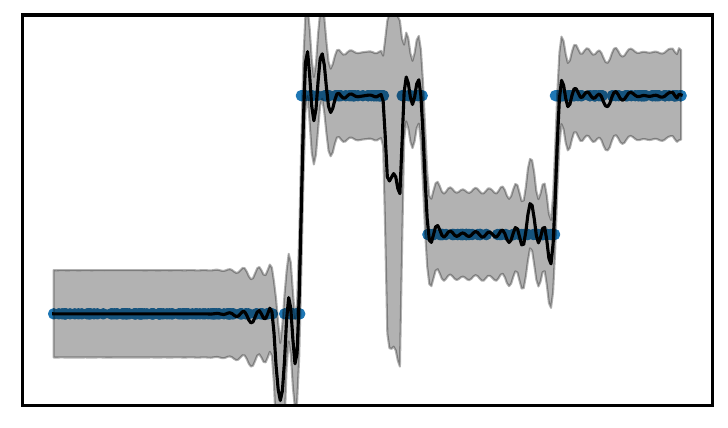}
        \label{fig:init:placeholder}
    \end{subfigure}
	\caption{(left) Initial predictive distributions of all the models that have $\varm{}{L} = \mathbf{0}$ (\ZEROW{}, \ZERONWR, \PCAW{}, \PCANWR, \ZEROWMO, \ZERONWRMO), and for the proposed initialization in both \W{} (middle) and \NWR{} cases (right). The proposed initialization shows a very accurate initial predictive distribution.}
    \label{fig:init:three_models}
\end{figure}

First, we compare \MO{} and \MY at initialization. We illustrate the initial predictive distribution of the \DGP models using the standard value for the variational mean in the last layer, $\varm{}{L} = \mathbf{0}$, and our proposed initialization \MY in \fig~\ref{fig:init:three_models}. The initial predictive distribution of all the models using $\varm{}{L} = \mathbf{0}$ is the same, and is displayed in the left sub-figure. In the case of the \MY{} models, we observe small differences in the \W{} and \NWR{} initializations. However, both are able to explain in a reasonable way the observed targets (middle and right sub-figures). \utab \ref{tab:toy:initial:combined} shows the initial \ELL{}, \KLD{} and \RMSE{} of each model and initialization. As expected, all the models with a zero variational parameter in the output layer have the same \ELL{} and \RMSE{} since they output the same predictive distribution at initialization (the kernel hyper-parameters are the same). The models with the proposed initialization, \ie, \ZEROWMY and \ZERONWRMY, obtain much better initial \ELL{} and \RMSE{} values. The reason is that the output layer aims to predict the targets at the inducing locations $\Zsamples{}$, which were selected as a subset of $\Xsamples{}$. The initial \KLD{} is, however, different. The big differences appear in the \MY models, in which the variational mean is strongly modified in the last layer to predict the targets at the inducing points $\Zsamples{}$. This results in a big difference between the variational distribution and the prior, which yields a high \KLD. In the case of the \MO models, there is an increment in the \KLD due to the modified variational distribution in the inner layers. The large initial \KLD{} value of the \MY{} models results in a significantly larger \ELBO{}, which can affect its optimization.

\begin{table}[!htb]
    \caption{Initial values for \RMSE, \ELL and \KLD for each model on the toy dataset. Values are reported for the optimal length-scale for \MY{} models (obtained using \ualg~\ref{alg:lengthscale:selection}), \ie, $\ell = 0.1$, and for the standard initial value $\ell = 2.0$ for the other models.}
    \centering
        \begin{tabular}{lccc|ccc}
		\hline
\multirow{2}{*}{\bf Model Name} & \multicolumn{3}{c|}{$\ell = 0.1$ } & \multicolumn{3}{c}{$\ell = 2.0$} \\
 & \ELL & \KLD & \RMSE & \ELL & \KLD & \RMSE \\
		\hline
\ZEROW      & -2816.35 & 2628.23   & 1.00 & -2816.35 & 2628.23   & 0.99 \\
\ZERONWR    & -2816.34 & 1763.54   & 1.00 & -2816.32 & 1691.91   & 0.99 \\
\PCAW       & -2816.40 & 2628.23   & 1.00 & -2816.35 & 2628.23   & 0.99 \\
\PCANWR     & -2817.12 & 1763.54   & 1.00 & -2816.32 & 1691.91   & 0.99 \\
\ZEROWMO    & -2816.35 & 2665.76   & 1.00 & -2816.35 & 2672.95   & 0.99 \\
\ZERONWRMO  & -2816.50 & 1802.41   & 1.00 & -2816.32 & 1736.62   & 0.99 \\
\ZEROWMY    &    79.12 & 772874.04 & 0.16 &   -339.79 & 8797481.98 &  0.41 \\
\ZERONWRMY  &    83.15 & 772010.58 & 0.16 &   -339.21 & 8796545.65 &  0.41 \\
	\hline
\end{tabular}
    \label{tab:toy:initial:combined}
\end{table}

We now analyze the final predictive distribution of each method after training. For this, in this toy problem, we use the learning rate $\lambda = 10^{-4}$. 
\utab\ref{tab:toy:test:results} shows the final results obtained by each method in terms of the \RMSE, the \KLD, and the estimated noise variance.
We observe that a non-whitened model, \ie, \ZERONWR{}, is the worst performing one in terms of \RMSE, exhibiting a much lower \KLD{} and a higher noise variance than the 
rest of the models. These are indicators of posterior collapse. The \KLD{} does not go to zero due to noise in the optimization process. 
\NWR models have, in general, worse \RMSE, higher noise variances, and smaller \KLD than their corresponding whitened versions. This indicates that the \NWR parameterization results in models that tend to present posterior collapse, due to the noise injected coming from the parameterization. We observe how the \ZEROW{} model also provides suboptimal results with considerably worse results in terms of \RMSE and a higher noise variance. The \ZERONWR{} model presents nearly a collapsed distribution.
Importantly, our proposed initialization obtains the best \RMSE. However, we observe that the \MY initialization typically results in a larger \KLD{} term. 

\begin{table}[!htb]
    \centering
	\caption{Test metrics by all the models on the toy dataset using $\lambda=10^{-4}$.}
\begin{tabular}{l|ccc}
\hline
    Model & \KLD{} & Noise Var. & \RMSE{} \\
    \hline
    \ZEROW{}       & $241.4252$     & $0.1098$ & $0.4591$ \\
    \ZERONWR{}     & $39.7501$      & $0.1205$ & $0.9896$ \\
    \PCAW{}        & $259.3910$     & $0.0329$ & $0.0321$ \\
    \PCANWR{}      & $229.2737$     & $0.0344$ & $0.0423$ \\
    \ZEROWMO{}     & $281.4474$     & $0.0340$ & $0.0294$ \\
    \ZERONWRMO{}   & $228.2416$     & $0.0331$ & $0.0406$ \\
    \ZEROWMY{}     & $316030.0797$ & $0.0190$ & $\mathbf{0.0055}$ \\
    \ZERONWRMY{}   & $483.9357$     & $0.0199$ & $0.0166$ \\
    \hline
\end{tabular}
    \label{tab:toy:test:results}
\end{table}

\begin{figure}[!htb]
    \centering
    \begin{subfigure}[t]{0.24\linewidth}
        \centering
        \scriptsize{\ZEROW} \\[0.3em]
        \includegraphics[width=\linewidth]{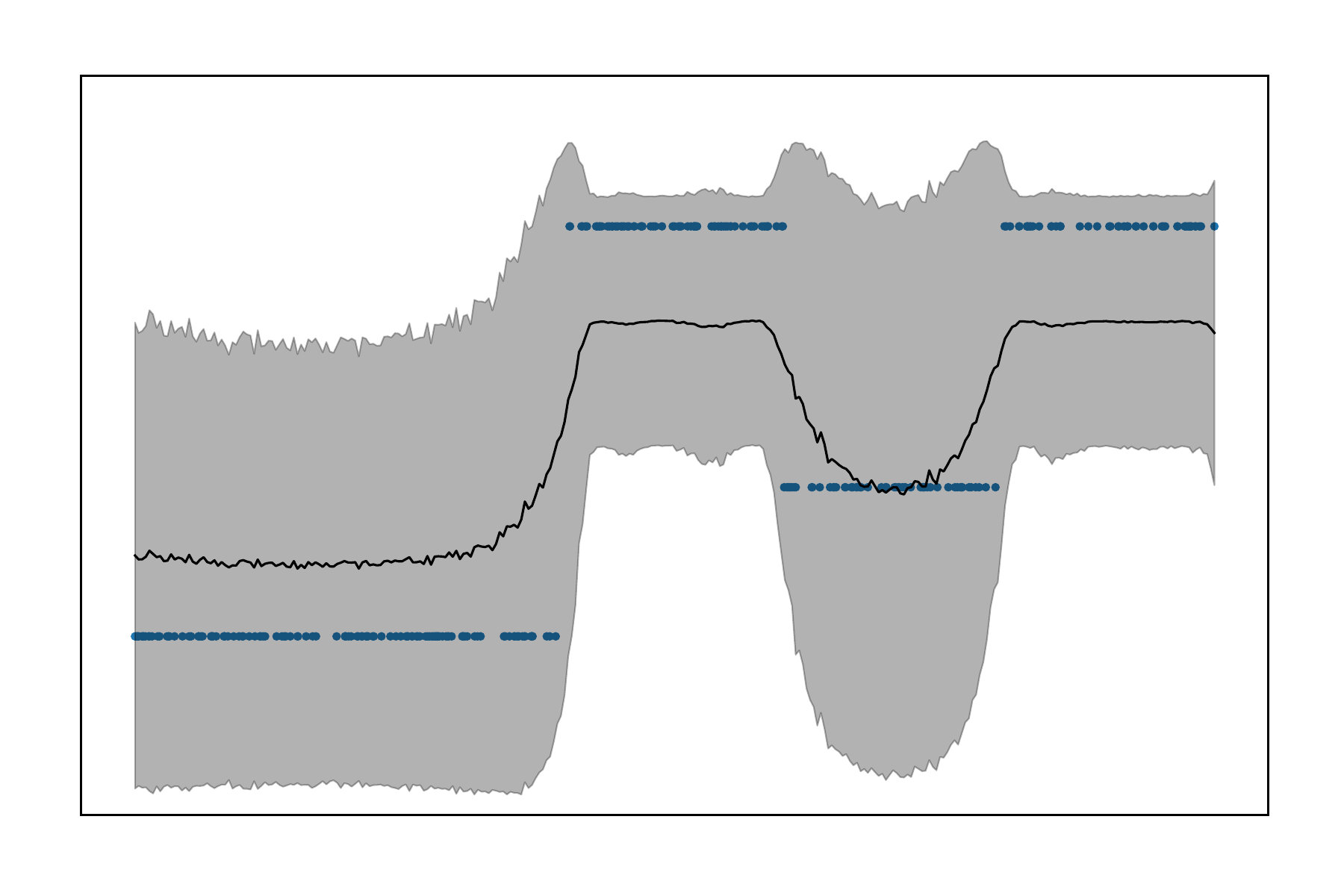}
    \end{subfigure}
    \begin{subfigure}[t]{0.24\linewidth}
        \centering
        \scriptsize{\PCAW} \\[0.3em]
        \includegraphics[width=\linewidth]{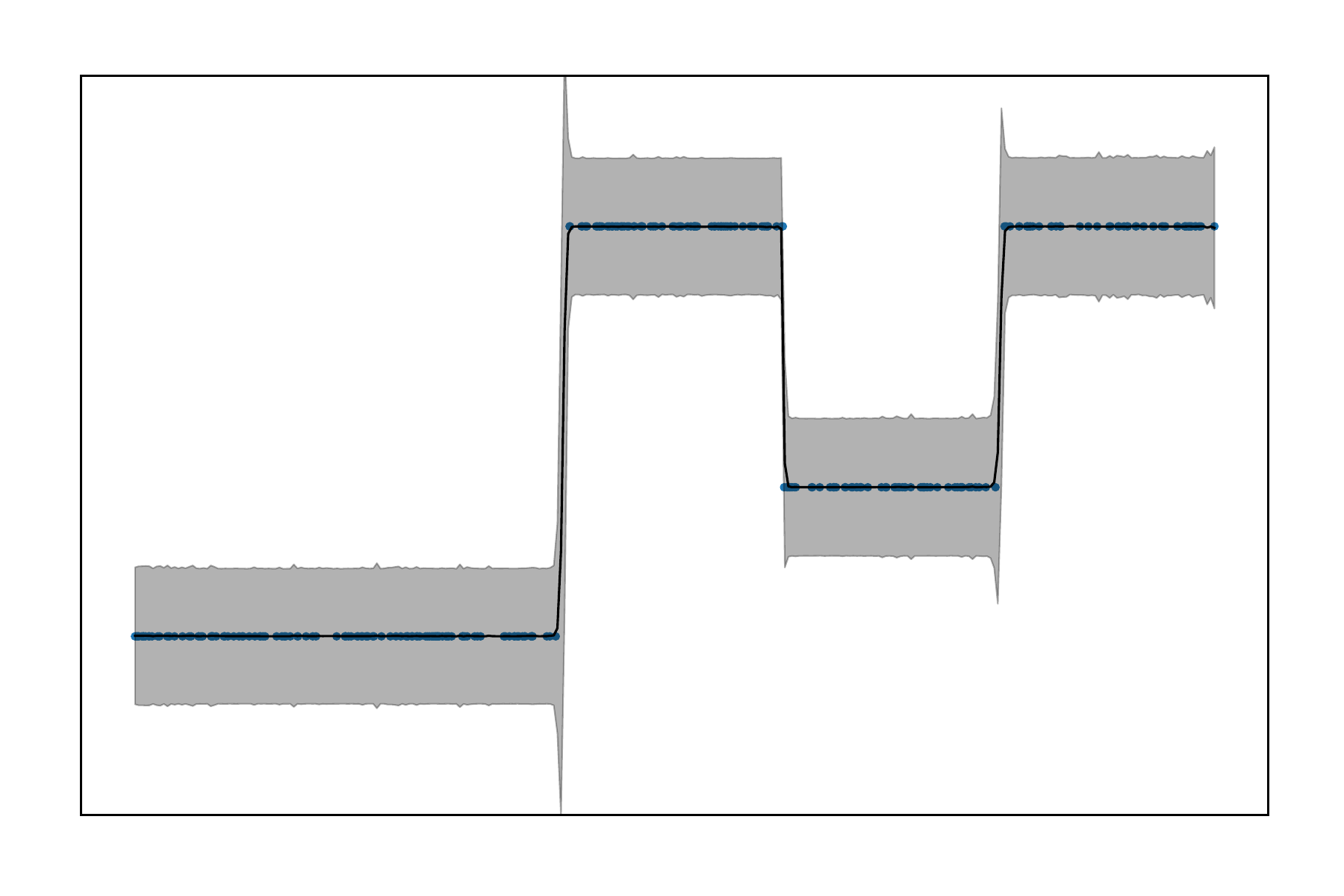}
    \end{subfigure}
    \begin{subfigure}[t]{0.24\linewidth}
        \centering
        \scriptsize{\ZEROWMO{}} \\[0.3em]
        \includegraphics[width=\linewidth]{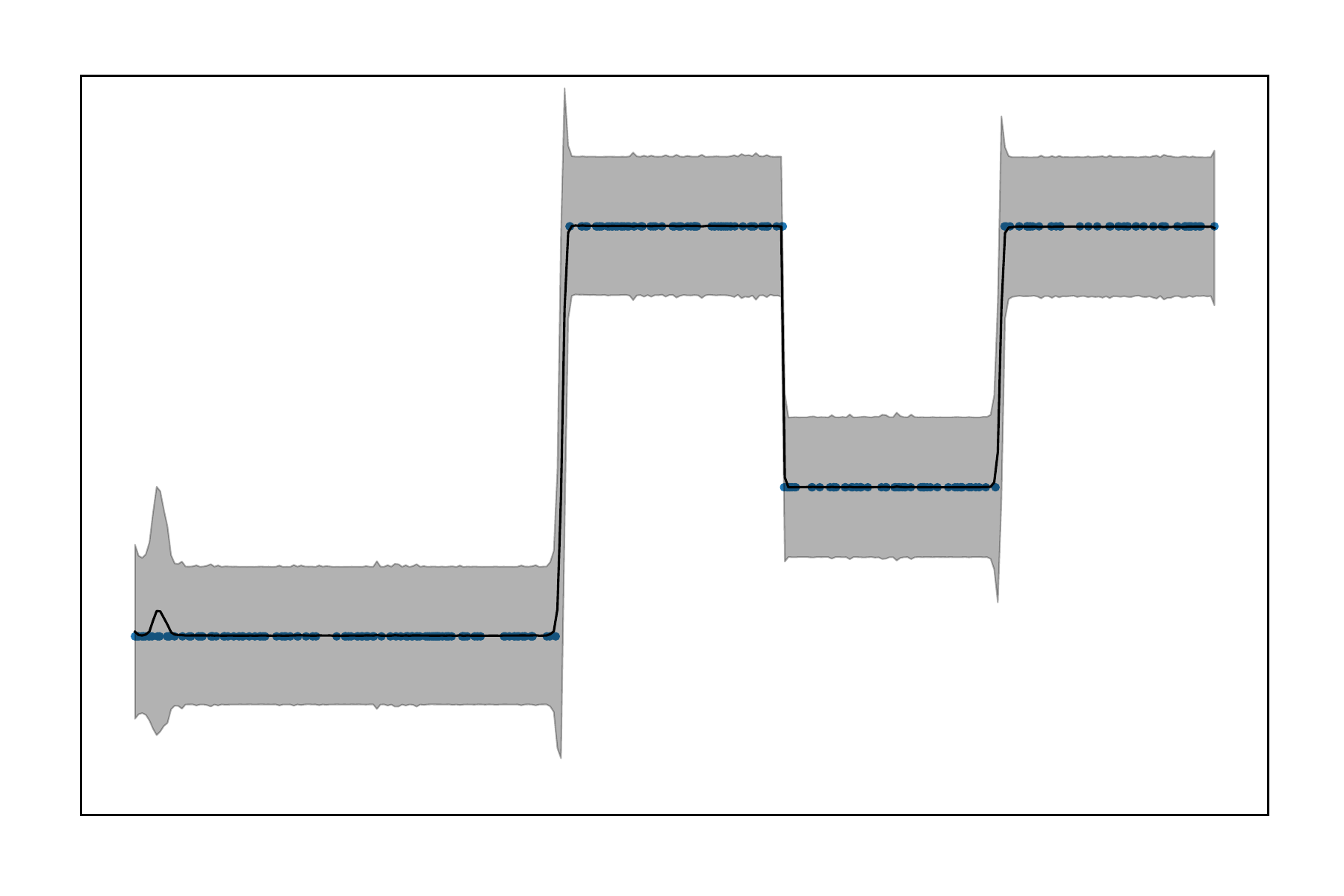}
    \end{subfigure}
    \begin{subfigure}[t]{0.24\linewidth}
        \centering
        \scriptsize{\ZEROWMY} \\[0.3em]
        \includegraphics[width=\linewidth]{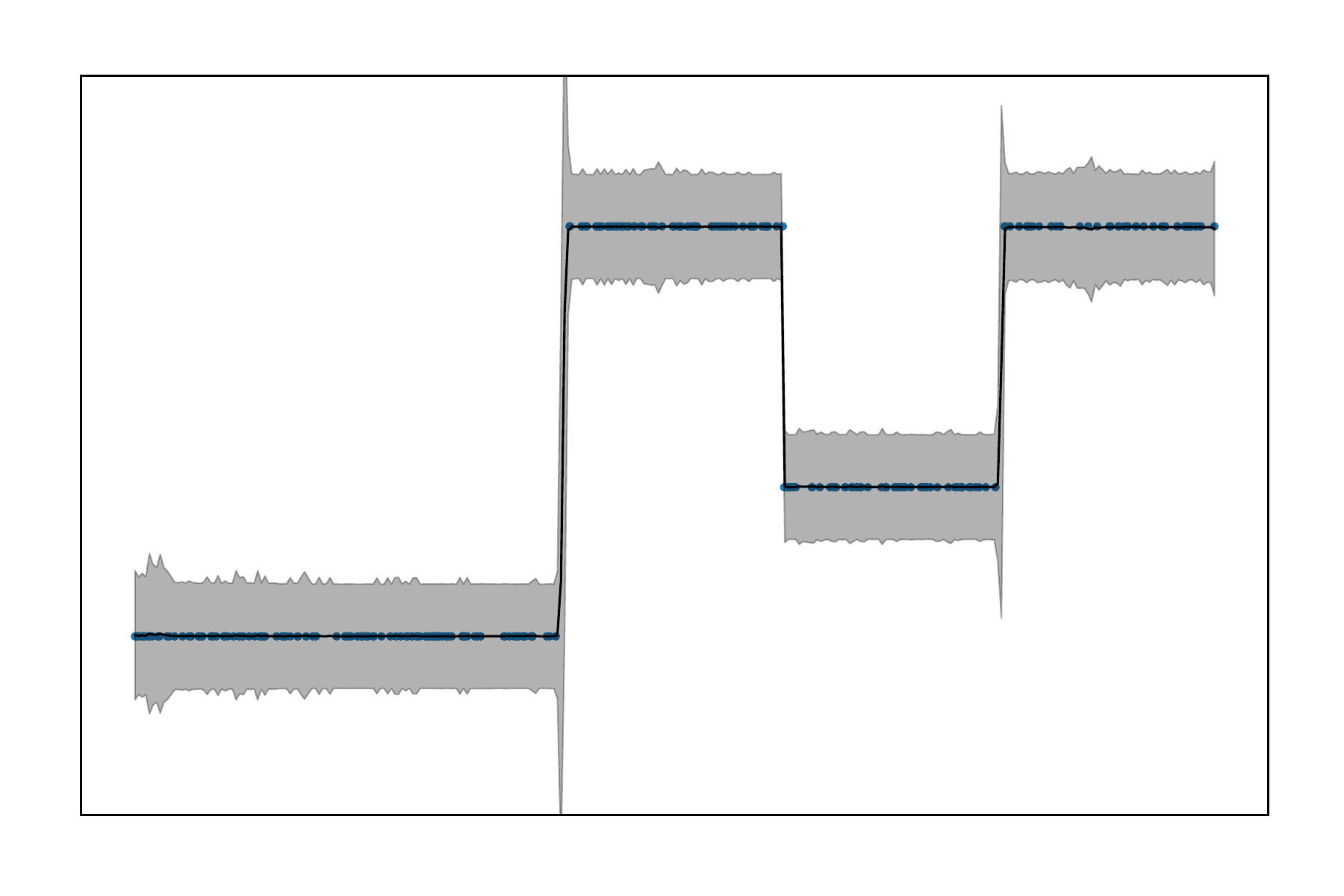}
    \end{subfigure}
    \vspace*{0.5cm}
    \begin{subfigure}[t]{0.24\linewidth}
        \centering
        \scriptsize{\ZERONWR} \\[0.3em]
        \includegraphics[width=\linewidth]{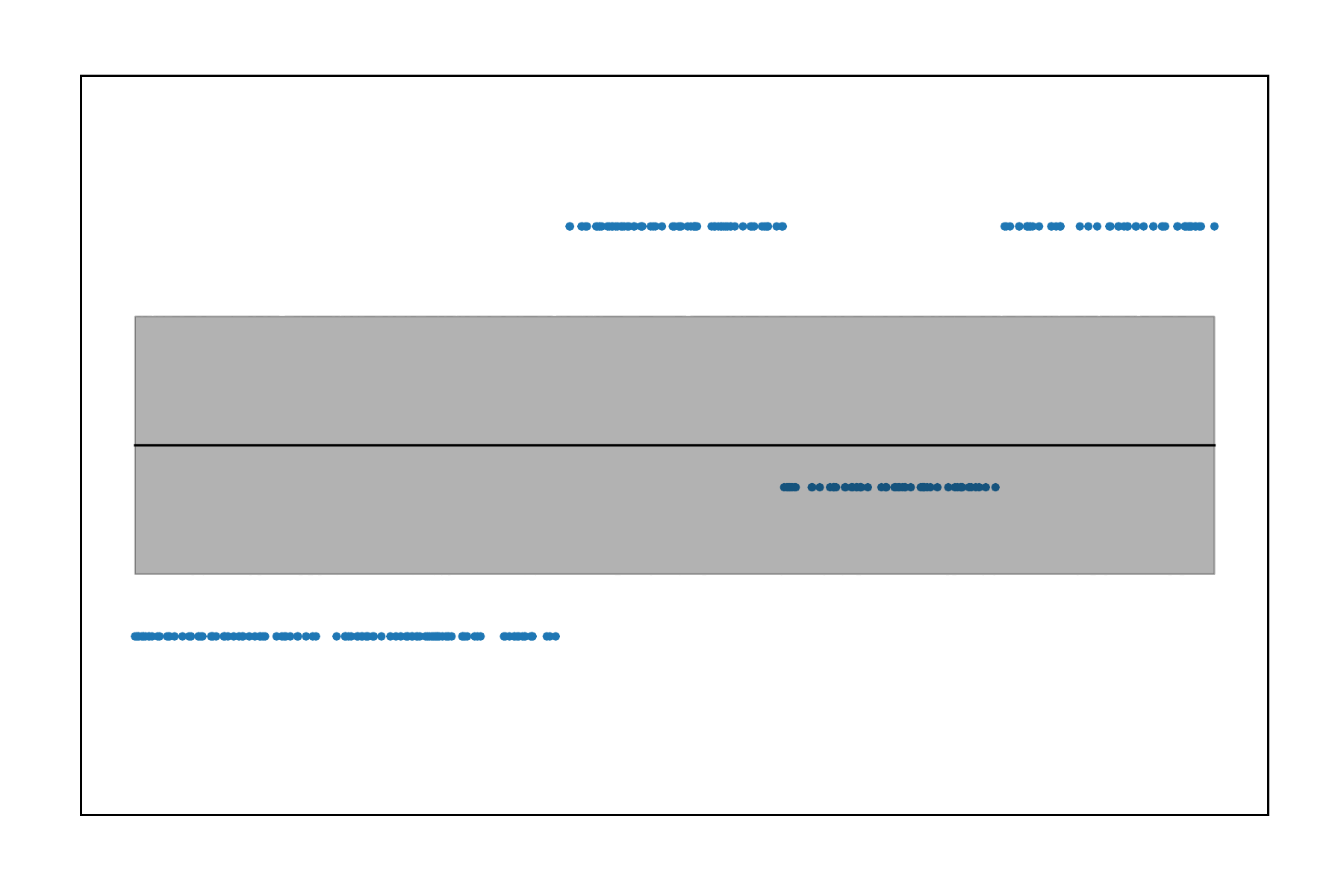}
    \end{subfigure}
    \begin{subfigure}[t]{0.24\linewidth}
        \centering
        \scriptsize{\PCANWR} \\[0.3em]
        \includegraphics[width=\linewidth]{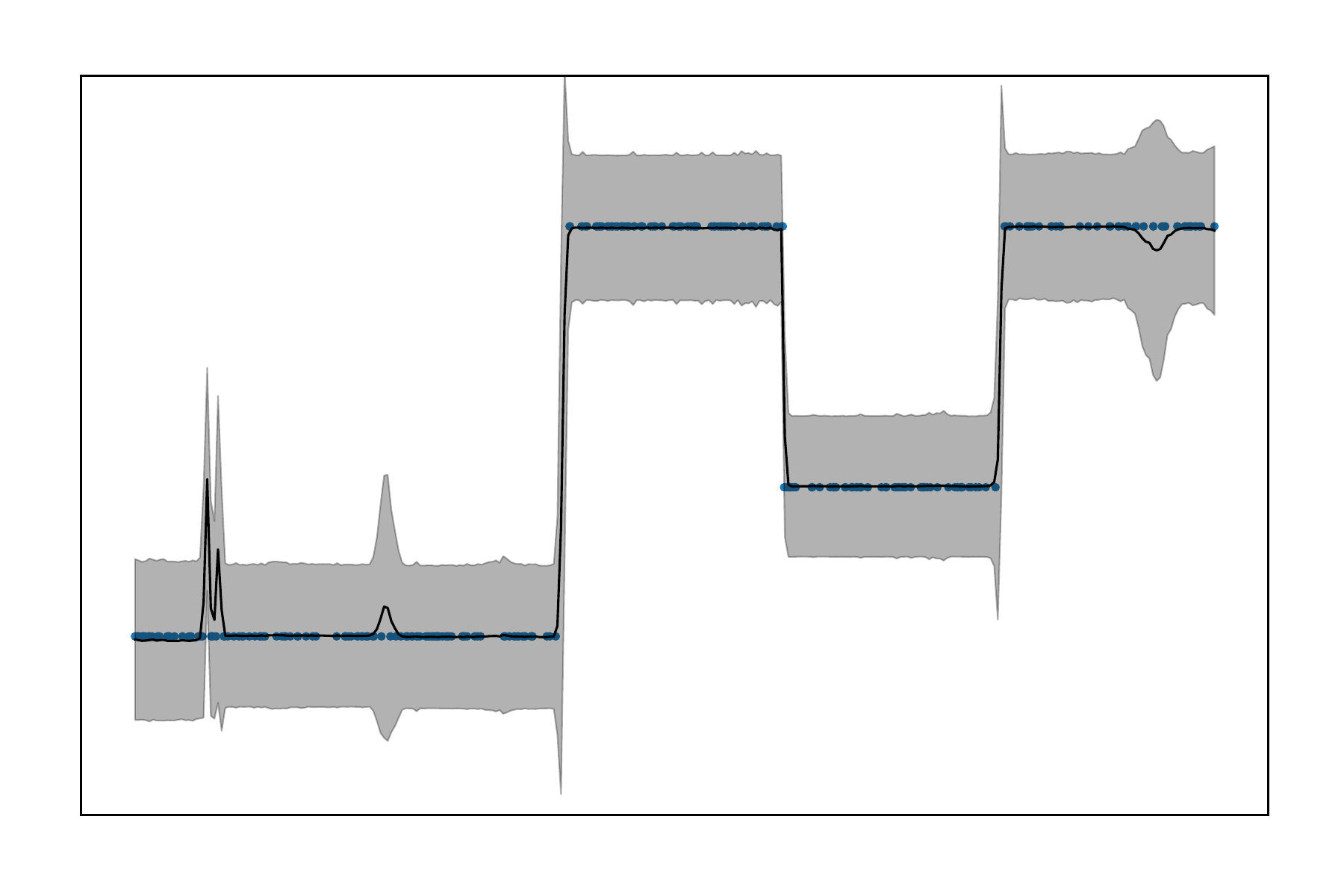}
   \end{subfigure}
    \begin{subfigure}[t]{0.24\linewidth}
        \centering
        \scriptsize{\ZERONWRMO{}} \\[0.3em]
        \includegraphics[width=\linewidth]{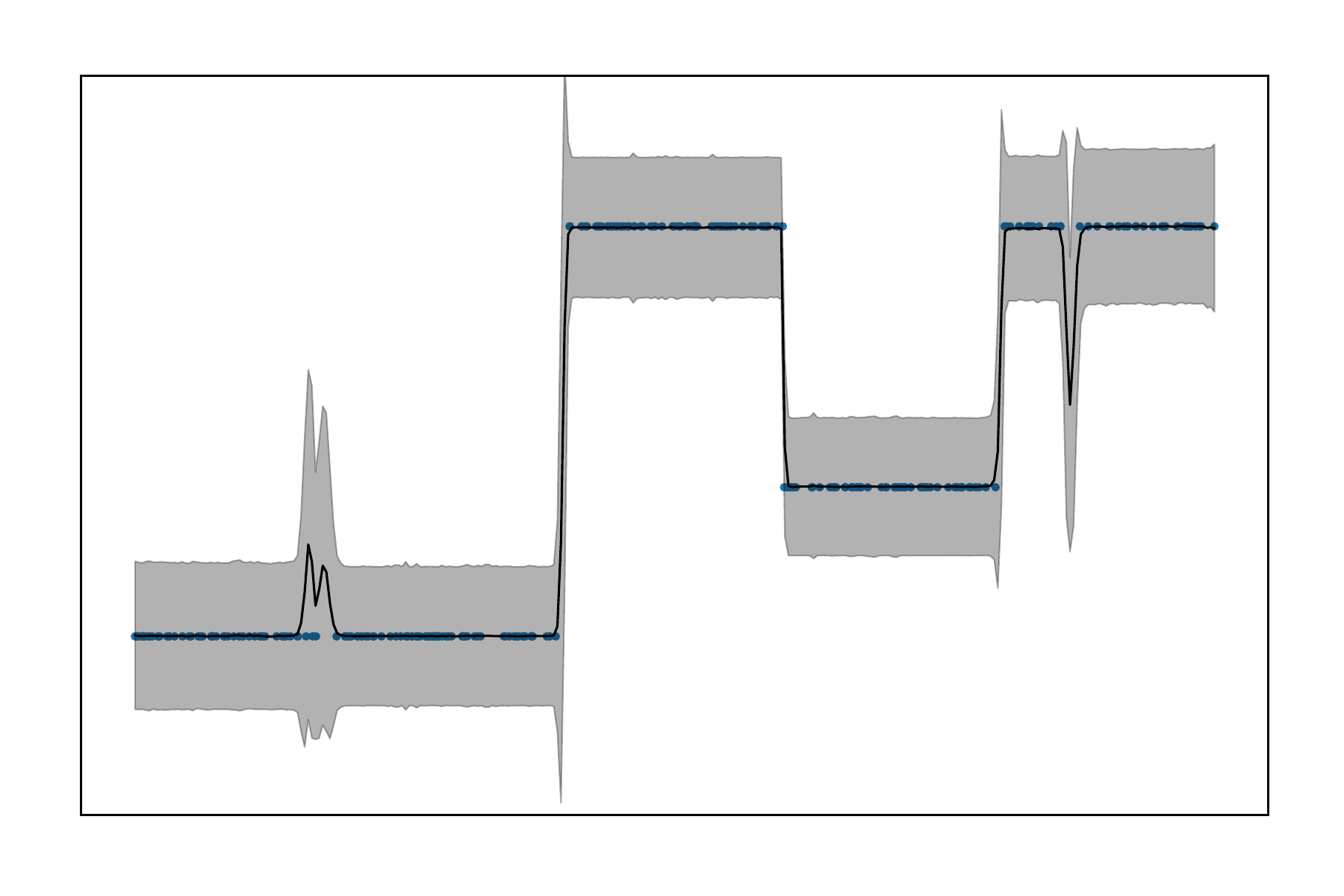}
    \end{subfigure}
    \begin{subfigure}[t]{0.24\linewidth}
        \centering
        \scriptsize{\ZERONWRMY} \\[0.3em]
        \includegraphics[width=\linewidth]{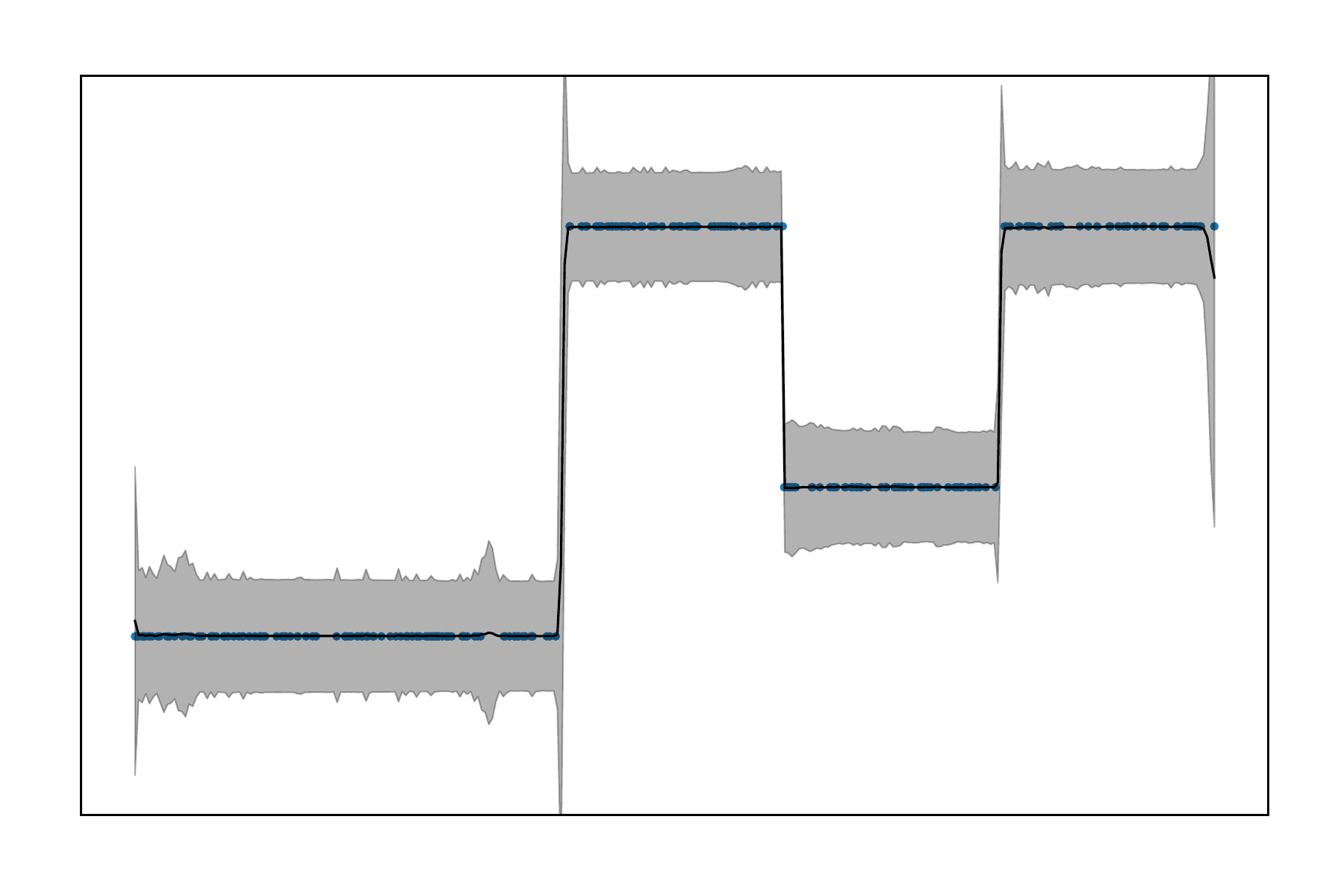}
    \end{subfigure}
    \caption{Predictive distributions of all trained models with $\lambda = 10^{-4}$.}
    \label{fig:toy:test:predictive:results}
\end{figure}

\fig{}~\ref{fig:toy:test:predictive:results} shows the predictive distribution of all the models, after training. 
We observe that, in general, standard \ZERO{} prior mean \DGP{}s with the non-whitened parameterizations suffer 
from a posterior collapse (bottom-left figure). Furthermore, a whitened parameterization alleviates such a problem, 
but results in poor predictive distributions (top-left figure). That is, the model tends to predict mostly noise. 
The \PCA prior mean function is very effective in alleviating these problems, resulting in better predictive distributions.
Similarly, our proposed initialization is also successful in avoiding the posterior collapse problem for all the evaluated 
configurations. It generates predictive distributions that look very similar to those of the \PCA prior mean \DGP{}s. In Appendix \ref{sec:app:c:3:toy:results}, we extend with experiments using a higher learning rate.

From the experiments carried out so far, we can conclude that:
\begin{itemize}
\item In this dataset, a \ZERO prior mean function with the standard initialization often leads to a posterior collapse problem or a bad predictive distribution, independently of the parameterization.
\item A whitened parameterization is to be preferred over a non-whitened one. Whitening often reduces the posterior collapse problem or improves the predictive distribution. 
\item The proposed initialization of $\varm{}{l}$ in the inner layers (that is, the \MO{} model), offers a similar behavior to the \PCA{} prior mean \DGP model and alleviates the posterior collapse problem, independently of the parameterization considered.
\item Using the proposed initialization in the last layer of the model (\ie, the \MY{} model) results in slightly more 
accurate predictive distributions. However, it may induce high \KLD{} values before and after training, which may complicate \ELBO optimization.
\end{itemize}
\subsection{Experiments on the \UCI{} Datasets}
\label{sec:experiments:uci}

In this section, following observations made in the toy problem experiments, we train several \DGP models on $8$ real-world datasets from the \UCI{} repository \citep{Dua:2019:UCI}. These are Boston, Concrete, Energy, Kin8nm, Power, Protein, Redwine, and Yacht. These datasets have been used in other works involving \DGP \citep{Salimbeni2017}.
The goal of these experiments is two-fold. First, comparing the performance of the proposed initializations \MO{} and \MY{} of the \ZERO prior mean \DGP{} model with the standard initializations used 
in the \ZERO{} and \PCA{} prior mean \DGP{}s. Second, studying the posterior collapse problem and optimization stability in real-world datasets. 

In these experiments, we consider the experimental setup considered in \cite{Salimbeni2017}. For optimization, we use Adam \citep{adamoptimizer}. We train each model for $20,000$ iterations using a learning rate of $\lambda=0.01$ and a batch size of at most $10,000$ instances. Each model is evaluated on $20$ different train/test splits, and we report average results across splits. We train $2$, $3$, $4$, and $5$-layer $\DGP{}$ models, using $100$ inducing points and $\min\{D,30\}$ different \GP{}s per layer, where $D$ is the dimensionality of the data. The kernels are shared between the \GP{}s in a layer, and the kernel parameters are initialized as $\ell = \sigma =  2.0$. The noise variance is initialized to $0.05$. For the \PCA{} and \ZERO{} mean prior models, and also for the \MO model, inducing points are initialized with \emph{K-means} in the real-world experiments, which means $\Zsamples{}=\Xz$. With this, the differences with respect to the \PCA and \ZERO prior mean \DGP{} models just rely on the initialization of $\varmr{}{},\varmw{}{}$. The \MY model can only use $\Zsamples{}=\Xx$, which means that inducing points also differ from those used by \PCA and \ZERO prior mean \DGP{} since now the inducing points must be selected from the training data using the proposal in \usec \ref{subsec:init_inducing}. For every model, we initialize the variational covariance to $\varS{}{}=10^{-5}\mathbf{I}$ in all the layers. We estimate the test metrics using $S=100$ Monte Carlo samples. In this section, we will show the results in terms of the test log-likelihood. The results in terms of \RMSE{} draw the same conclusions, and we place them in the Appendix \ref{sec:app:c:1:rmse} for brevity.

Although the \NWR{} parameterization showed poor performance in the toy example, we include results for both the \emph{whitened} and \emph{non-whitened} models for two main reasons. First, to the best of our knowledge, there has been no previous report in the literature comparing the performance of both parameterizations. Second, we have observed that, despite exhibiting greater optimization instability, the \NWR{} parameterization achieves better results on several datasets. A detailed comparison between the two parameterizations is provided in Appendix \ref{sec:app:c:4:w:vs:nwr}.

\subsubsection{Main Results For Both Parameterizations}
\begin{figure}[!p]
\vspace{-1.0cm}
    \centering
    \begin{subfigure}{0.9\linewidth}
        \centering
        \includegraphics[width=0.9\linewidth]{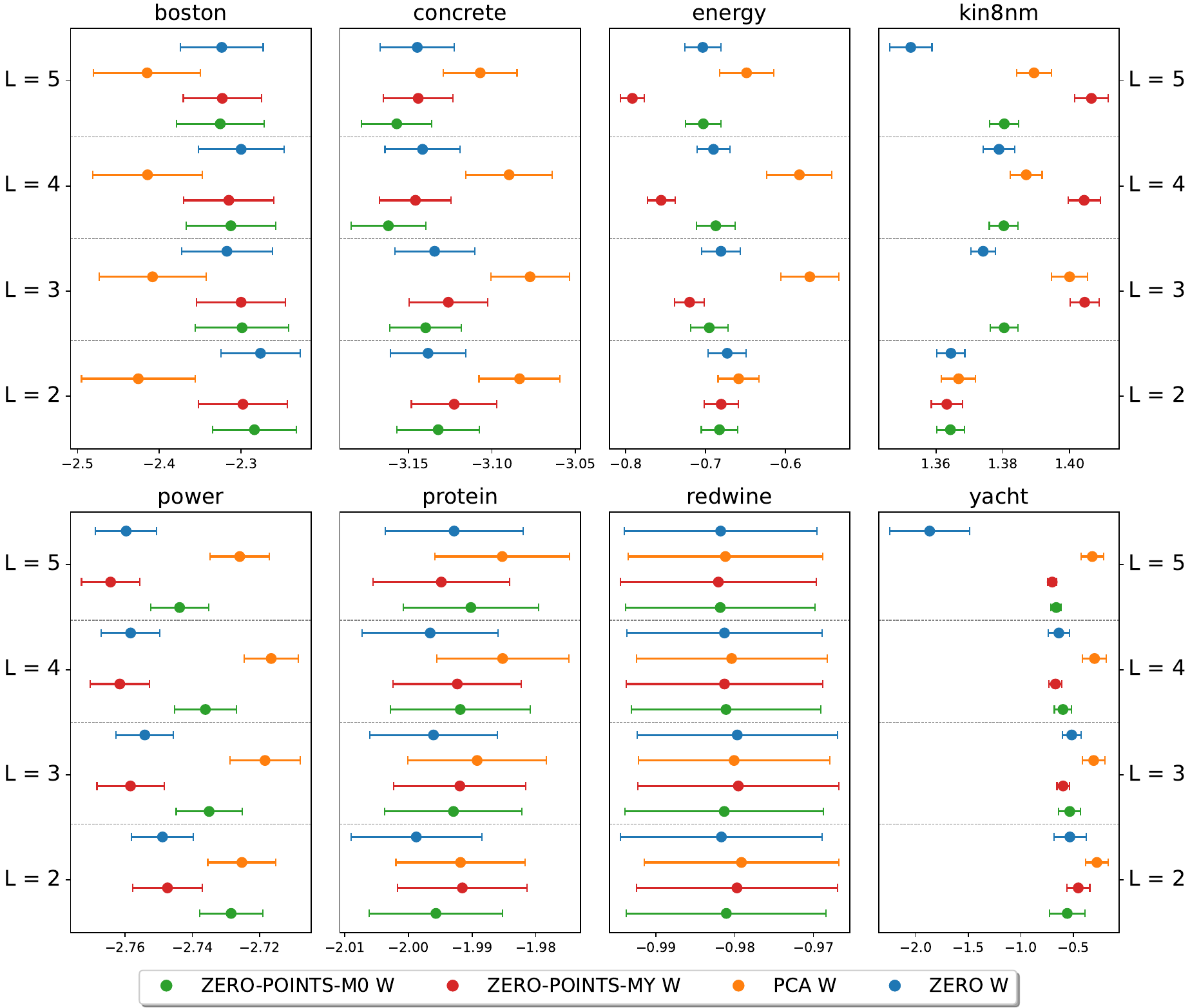}
         \caption{ Whitened models.}
        \label{fig:results:ll:whiten}
    \end{subfigure}
    \begin{subfigure}{0.9\linewidth}
        \centering
        \includegraphics[width=0.9\linewidth]{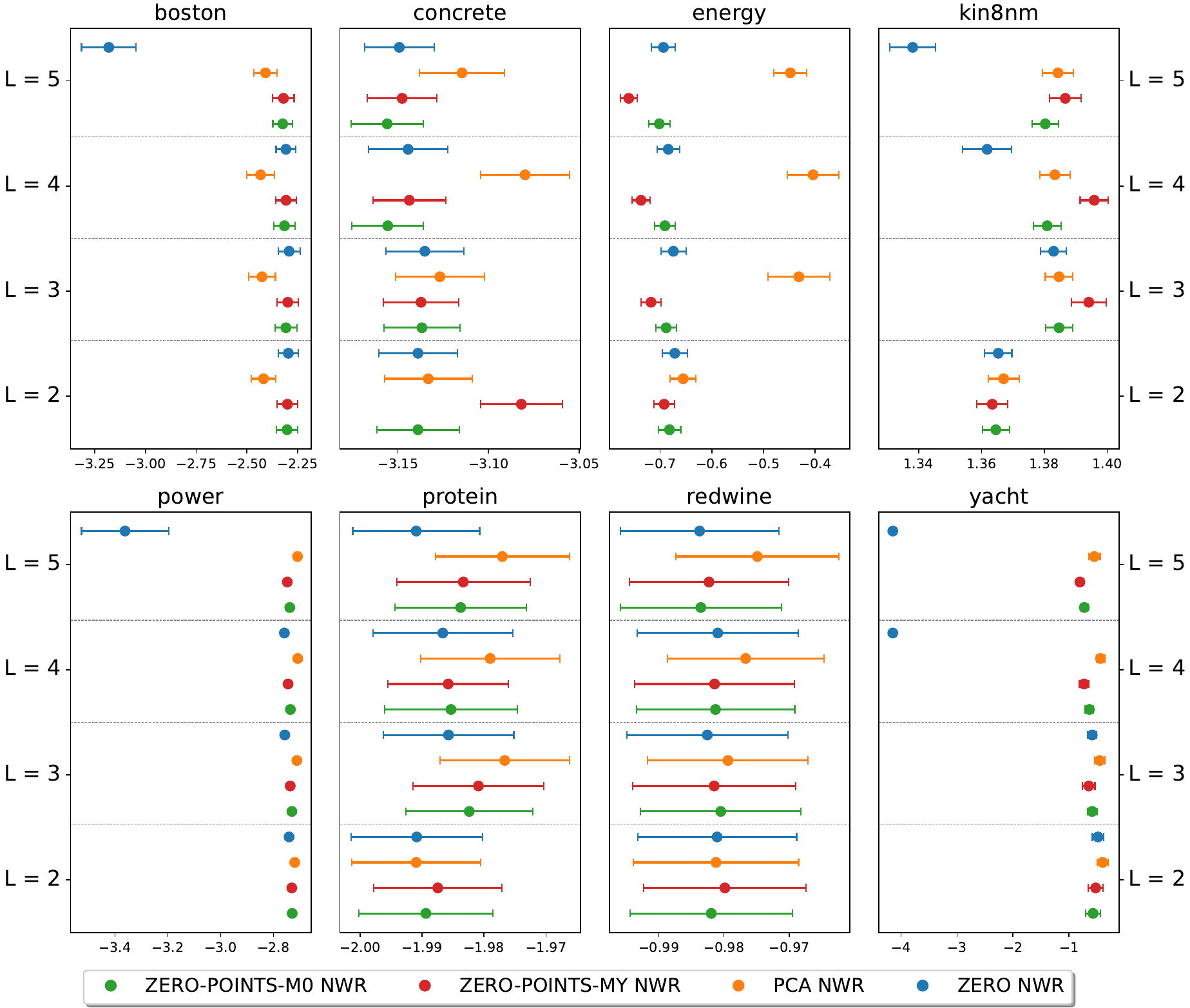}
        \caption{ Non-whitened models.}
        \label{fig:results:ll:nwr}
    \end{subfigure}
    \caption{Test log-likelihood (right is better) in all \UCI{} datasets.}
    \label{fig:results:uci}
\end{figure}

\fig~\ref{fig:results:ll:whiten} and \fig~\ref{fig:results:ll:nwr} show the test log-likelihood obtained by each whitened and non-whitened model, respectively. The figure shows that this metric is, in most of the datasets, very similar for all the models when we take into account the standard deviation of the results. In general, there are no significant differences when varying the number of layers. Importantly, the models using the proposed initializations, \ie, \MO{} and \MY{}, show competitive results when compared with the \PCA and \ZERO models.

Regarding the \emph{whitened} parameterization, shown in \fig~\ref{fig:results:ll:whiten}, looking first at the \MO model, we observe how this model is either better or equal to the \ZEROW{} model. As we mentioned in \usec \ref{sec5_summary}, we shall not expect the \MO model to surpass the \ZEROW model in predictive performance. Examples of clear improvements are given in Power, Kin8nm, and Yacht datasets. Among these, the most notable case is the Yacht dataset, where the \ZEROW{} model suffers from posterior collapse (as we will demonstrate later in \usec~\ref{sec:experiments:uci:collapse:analysis}), while the proposed \MO{} initialization mitigates this issue and achieves performance comparable to that of the \PCAW{} model. This case illustrates the beneficial properties of the proposed initialization.  Only in the Concrete dataset, with 4-layers, and the Energy dataset, with 3-layers, \MO models work slightly worse. In the rest of the datasets, we observe slightly better or equal results. 

Regarding the \MY model, we observe that this model can surpass the \PCAW model (Kin8nm). This model also performs similarly to the \ZEROW and \MO models (for instance, in Yacht or Boston). Nevertheless, we observe that the \MY model achieves worse results than the \MO or even the \ZEROW{} models in cases like the Power and Energy datasets.  We have further investigated and found out that this occurs due to the initial length-scale selection. In the Energy dataset, for instance, we have observed that the length-scale selected using the proposed algorithm (see \ualg~\ref{alg:lengthscale:selection})  is around $6$, which results in a big \KLD. Thus, for the same number of epochs, the model might not have achieved a maximum in the \ELBO; or the maximum achieved is poor. Looking at the \RMSE result in \fig \ref{fig:results:rmse:whiten}, we observed that the \MY model in this dataset is not degraded. According to the Generalized Variational Inference theory \citep{GVI_jer}, the \KLD is the uncertainty quantifier. Therefore, it is reasonable to expect that if optimization fails to achieve an appropriate \KLD value, the model will struggle to quantify uncertainty accurately. This primarily affects the test \LL{}, which is sensitive to uncertainty estimation, while leaving point-estimate metrics, like \RMSE, largely unchanged. 

Finally, we observed that the Boston dataset exhibits degraded performance when using the \PCA prior mean function. This finding suggests that architectural choices should be motivated by modeling considerations rather than optimization-related issues. Although the \PCAW{} variant consistently performs better, the results on the Boston, Kin8nm, Protein, and Power datasets, where either the \MO or \MY models match or outperform the \PCA model, indicate that the improvements associated with the \PCA prior mean model do not necessarily arise from a change in the underlying statistical model. Consequently, our results support the use of the \ZERO prior mean, provided that the model is initialized appropriately.

Regarding the \emph{non-whitened} parameterization,  shown in \fig~\ref{fig:results:ll:nwr}, we observe similar results. The most notable difference is that, here, the \ZERONWR{} yields consistently worse results in a greater number of datasets than the whitened case. While the whitened model showed degraded performance in Kin8nm and Yacht, the \NWR model also degrades performance in Boston and Power, and the 4th layer Yacht \DGP. The most important conclusion drawn from this figure is that the proposed \MO{} and \MY{} models effectively correct the poor solutions of the \ZERONWR{} model by simply changing the initialization of the underlying shared statistical model. This highlights the benefits of the proposed initialization also for non-whitened models. We further highlight that in Concrete $2$ layers, the \MY model clearly outperforms the rest of the models. This contrasts with the whitened parameterization where the \PCA clearly outperformed the rest of models in this dataset. As mentioned previously, these parameterizations are compared in the appendix for each of the models. Interestingly, we observe that models with the \ZERO mean and \NWR parameterization show degraded performance compared to the whitened parameterization (\eg, Kin8nm with 4 layers or Power with 5 layers). However, when using our \MO initialization, \NWR slightly surpasses the whitened version.

To provide additional insights on these figures, we summarize the obtained results using \emph{whitened}
models in two different ways: Average model rank, and standard deviation analysis. In the following tables, the best result is highlighted in bold, and the second best is underlined. \utab{}~\ref{table:rankings:ll} shows the average rank of each model per dataset, in terms of the \LL{}. The ranks were computed for each split. The table shows that the \PCAW{} model achieves the best results overall. Notwithstanding, the tables also show that the proposed initializations \ZEROWMO{} and \ZEROWMY{} perform better than the \ZEROW{} model. 
Furthermore, we observe that both \ZEROWMO{} and \ZEROWMY{} achieve results closer to those of the \PCAW{} model. This validates our design approach: we specifically initialized the inner layers of the \ZERO{} \DGP{} to approximate the \PCA{} mean function, which is expected to yield more similar performance. This confirms further that this initialization successfully narrows the gap between \PCAW{} \DGP{}s and \ZEROW{} \DGP{}s, highlighting the beneficial properties of the proposed initialization. 

To assess the beneficial properties of the proposed initialization with respect to the stability of the resulting model, in \utab{}~\ref{table:rank:var:ll} we display the ranking of the models based on the standard deviation of the results across the splits (lower is better). The table shows that the proposed initializations achieve more stable results than both \PCAW{} and \ZEROW{} models. This indicates that the initialization induces a smoother optimization process than those of the \ZEROW{} and \PCAW{} models, which is a highly beneficial property. 

For \emph{non-whitened} models, we draw the same conclusions as those obtained for whitened ones. In \utab~\ref{table:rankings:ll:nwr} show the split-wise average rank in terms of the \LL{} for each model. \utab~\ref{table:rank:var:ll:nwr} shows the ranks in terms of standard deviations of the test \LL{}.

\begin{table}[!h]
      \centering
      \caption{Average rank of each \emph{whitened} model in terms of the test \LL{} for each dataset.}
  \label{table:rankings:ll}
	\resizebox{\textwidth}{!}{
\begin{tabular}{lccccccccc}
\toprule
{} &  boston & concrete &  energy &  kin8nm &   power & protein & redwine &   yacht & Overall \\
\midrule
\PCA               &  $3.66$ &   $1.64$ &  $1.46$ &  $2.23$ &  $1.30$ &  $2.21$ &  $2.70$ &  $1.40$ &  $\mathbf{2.07}$ \\
\ZERO              &  $2.33$ &   $2.89$ &  $2.62$ &  $3.34$ &  $3.30$ &  $3.01$ &  $2.64$ &  $3.14$ &  $2.91$ \\
\ZEROMO &  $2.46$ &   $3.30$ &  $2.90$ &  $2.99$ &  $2.04$ &  $2.56$ &  $2.69$ &  $3.06$ &  $2.75$ \\
\ZEROMY  &  $1.55$ &   $2.17$ &  $3.01$ &  $1.45$ &  $3.36$ &  $2.21$ &  $1.98$ &  $2.40$ &  \underline{$2.27$} \\
\bottomrule
\end{tabular}

}
\end{table}
\begin{table}[!h]
\centering
      \caption{Average rank of each \emph{whitened} model in terms of the standard deviations (lower is better) of the \LL{} (horizontal bars) in \fig{}~\ref{fig:results:ll:whiten}.}
      \resizebox{\textwidth}{!}{
\begin{tabular}{lccccccccc}
\toprule
{} &  boston & concrete &  energy &  kin8nm &   power & protein & redwine &   yacht & Overall \\
\midrule
\PCA               &  $2.00$ &   $4.00$ &  $2.00$ &  $3.00$ &  $1.00$ &  $4.00$ &  $4.00$ &  $3.00$ &  $2.88$ \\
\ZERO              &  $4.00$ &   $2.00$ &  $3.00$ &  $4.00$ &  $4.00$ &  $3.00$ &  $2.00$ &  $2.00$ &  $3.00$ \\
\ZEROMO &  $3.00$ &   $1.00$ &  $1.00$ &  $1.00$ &  $3.00$ &  $2.00$ &  $3.00$ &  $1.00$ &  $\mathbf{1.88}$ \\
\ZEROMY &  $1.00$ &   $3.00$ &  $4.00$ &  $2.00$ &  $2.00$ &  $1.00$ &  $1.00$ &  $3.00$ &  \underline{$2.12$} \\
\bottomrule
\end{tabular}

}
      \label{table:rank:var:ll}
\end{table}
\begin{table}[!h]
      \centering
      \caption{Average rank of each \NWR{} model in terms of the test \LL{} for each dataset.}
  \label{table:rankings:ll:nwr}
	\resizebox{\textwidth}{!}{
\begin{tabular}{lccccccccc}
\toprule
{} &  boston & concrete &  energy &  kin8nm &   power & protein & redwine &   yacht & Overall \\
\midrule
PCA               &  $3.51$ &   $1.81$ &  $1.19$ &  $2.30$ &  $1.35$ &  $2.17$ &  $2.20$ &  $1.44$ &  $\mathbf{2.00}$ \\
ZERO              &  $2.41$ &   $2.77$ &  $2.46$ &  $3.19$ &  $3.66$ &  $2.83$ &  $2.81$ &  $3.40$ &  $2.94$ \\
ZERO DatapointsM0 &  $2.49$ &   $3.04$ &  $2.74$ &  $2.75$ &  $2.31$ &  $2.51$ &  $2.56$ &  $2.90$ &  $2.66$ \\
ZERO DatapointsMY &  $1.59$ &   $2.38$ &  $3.61$ &  $1.76$ &  $2.67$ &  $2.49$ &  $2.42$ &  $2.26$ &  \underline{$2.40$} \\
\bottomrule
\end{tabular}

}
\end{table}
\begin{table}[!h]
\centering
      \caption{Average rank of each \NWR{} model in terms of the standard deviations (lower is better) of the \LL{} (horizontal bars) in \fig{}~\ref{fig:results:ll:whiten}.}
      \resizebox{\textwidth}{!}{
\begin{tabular}{lccccccccc}
\toprule
{} &  boston & concrete &  energy &  kin8nm &   power & protein & redwine &   yacht & Overall \\
\midrule
PCA               &  $3.00$ &   $4.00$ &  $1.00$ &  $3.00$ &  $3.00$ &  $4.00$ &  $4.00$ &  $3.00$ &  $3.12$ \\
ZERO              &  $4.00$ &   $2.00$ &  $2.00$ &  $4.00$ &  $2.00$ &  $2.00$ &  $3.00$ &  $4.00$ &  $2.88$ \\
ZERO DatapointsM0 &  $2.00$ &   $1.00$ &  $3.00$ &  $2.00$ &  $1.00$ &  $3.00$ &  $1.00$ &  $1.00$ &  $\mathbf{1.75}$ \\
ZERO DatapointsMY &  $1.00$ &   $3.00$ &  $4.00$ &  $1.00$ &  $4.00$ &  $1.00$ &  $2.00$ &  $2.00$ &  \underline{$2.25$} \\
\bottomrule
\end{tabular}

}
      \label{table:rank:var:ll:nwr}
\end{table}
\subsubsection{Analyzing the Posterior Collapse Problem on the \UCI Datasets}
\label{sec:experiments:uci:collapse:analysis}

We now analyze the previous results to determine whether the models that exhibited poor performance were affected by posterior collapse. Posterior collapse occurs when the model effectively reverts to the prior when making predictions. This phenomenon is typically characterized by vanishing KL divergence values together with an increase in the estimated noise variance. To assess whether this behavior is present, we examine the evolution of these quantities during training.

\begin{figure}[!t]
    \centering

    \begin{subfigure}{0.49\textwidth}
        \centering
        \includegraphics[width=\linewidth]{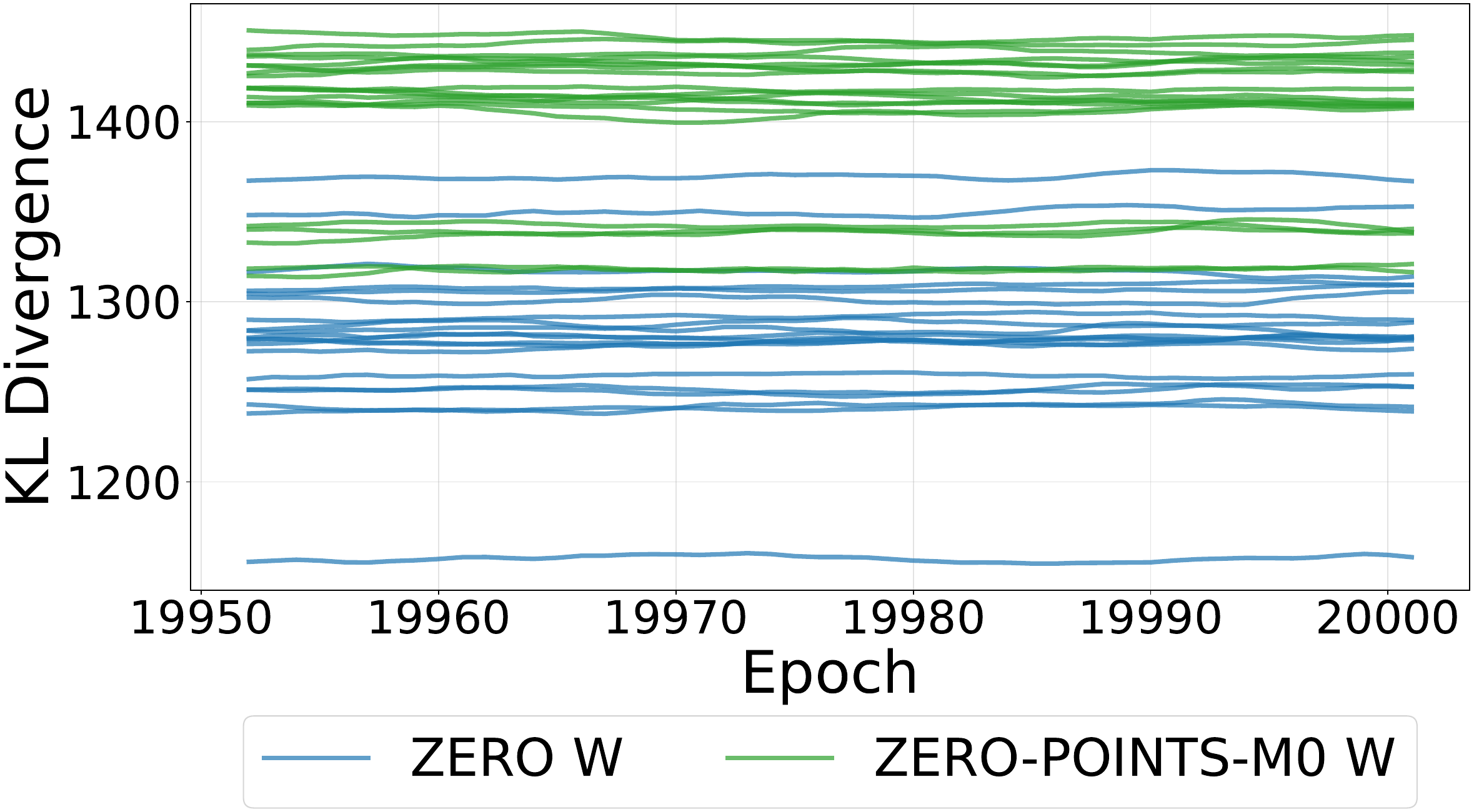}
        \caption{Kin8nm -- \KLD{}}
        \label{fig:results:Kin8nm:kld:last:50}
    \end{subfigure}
    \hfill
    \begin{subfigure}{0.49\textwidth}
        \centering
        \includegraphics[width=\linewidth]{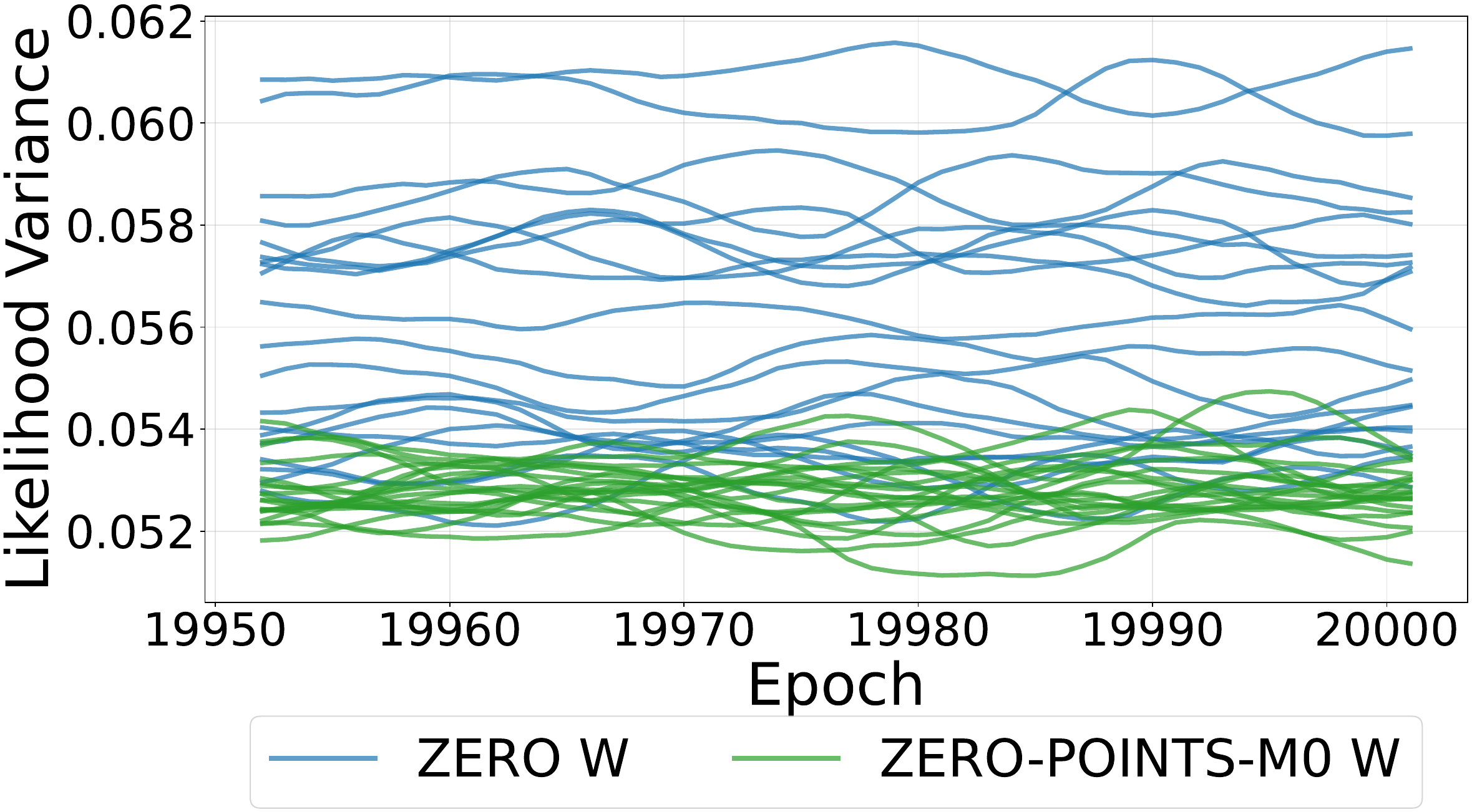}
        \caption{Kin8nm -- Noise variance}
        \label{fig:results:Kin8nm:llvar:last:50}
    \end{subfigure}

    \begin{subfigure}{0.49\textwidth}
        \centering
        \includegraphics[width=\linewidth]{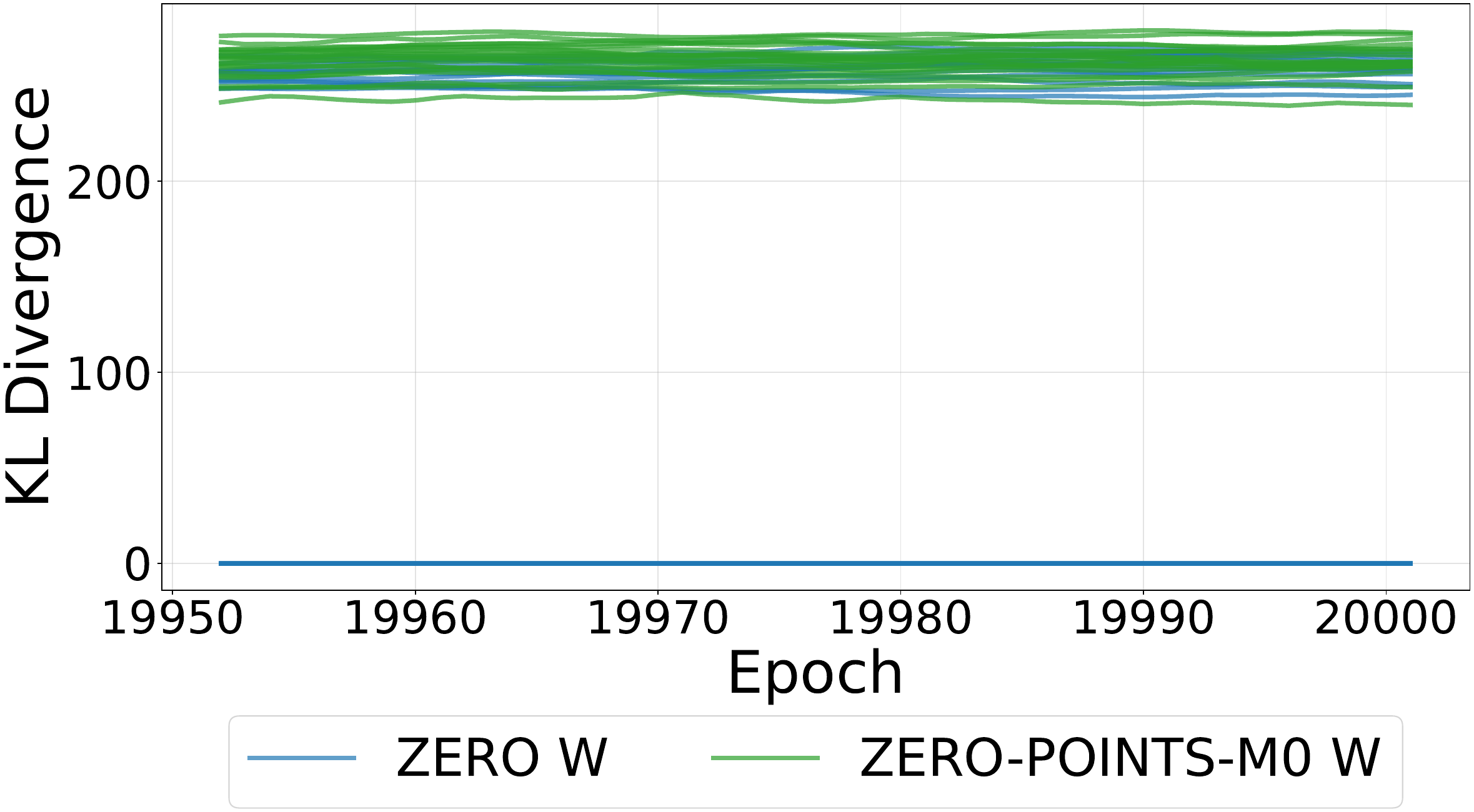}
        \caption{Yacht -- \KLD{}}
        \label{fig:results:Yacht:kld:last:50}
    \end{subfigure}
    \hfill
    \begin{subfigure}{0.49\textwidth}
        \centering
        \includegraphics[width=\linewidth]{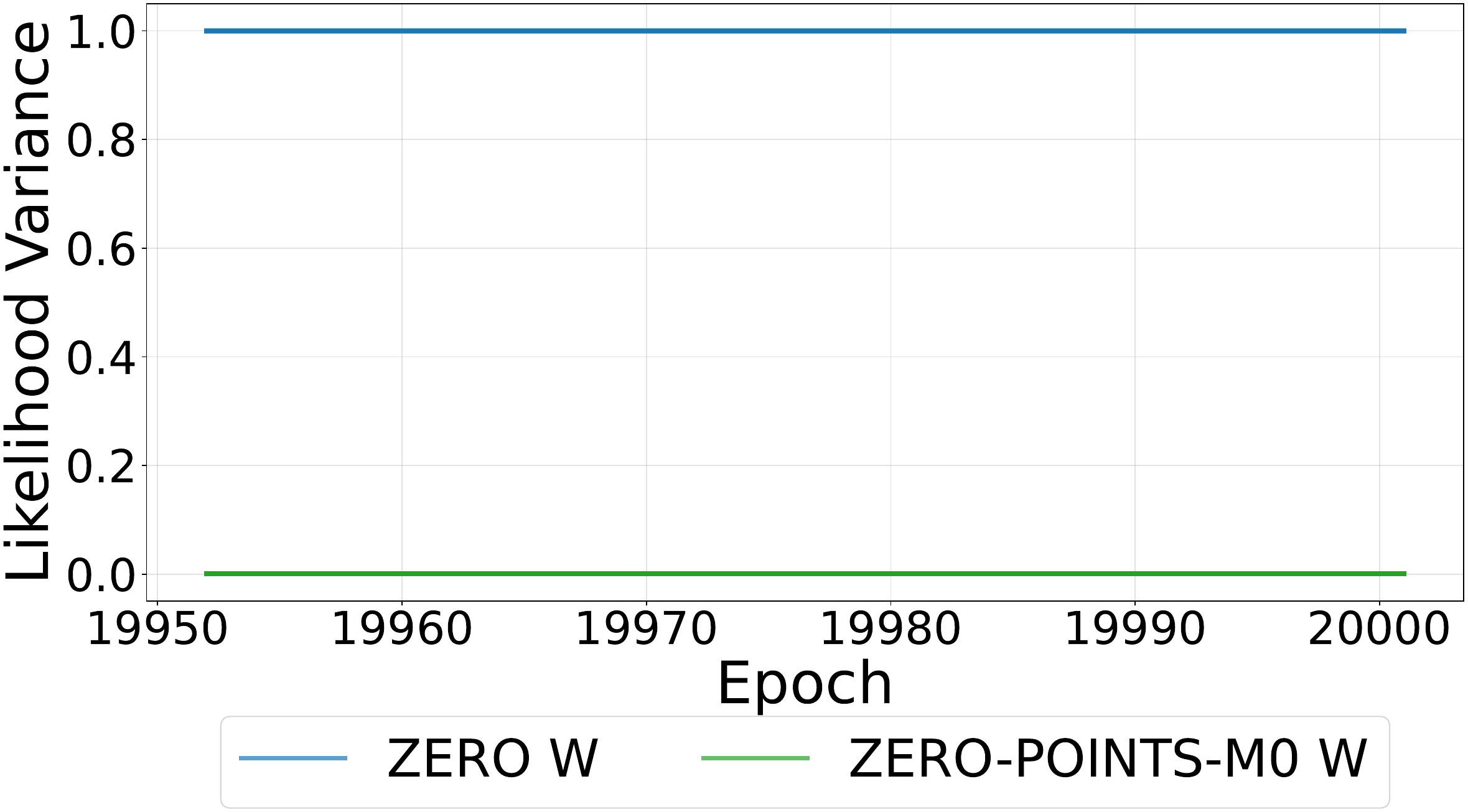}
        \caption{Yacht -- Noise variance}
        \label{fig:results:Yacht:llvar:last:50}
    \end{subfigure}

    \caption{
        \KLD{} (left) and likelihood variance (right) obtained by the
        \ZEROW{} and \ZEROWMO{} models in the Kin8nm and Yacht datasets,
        for each train/test split.
        No mode collapse is shown in Kin8nm.
        Seven out of $20$ splits suffer from posterior collapse in Yacht.
    }
    \label{fig:kld_llvar_comparison}
\end{figure}

\begin{figure}[!t]
    \centering
    \begin{subfigure}[t]{0.49\linewidth}
        \centering
        \includegraphics[width=\linewidth]{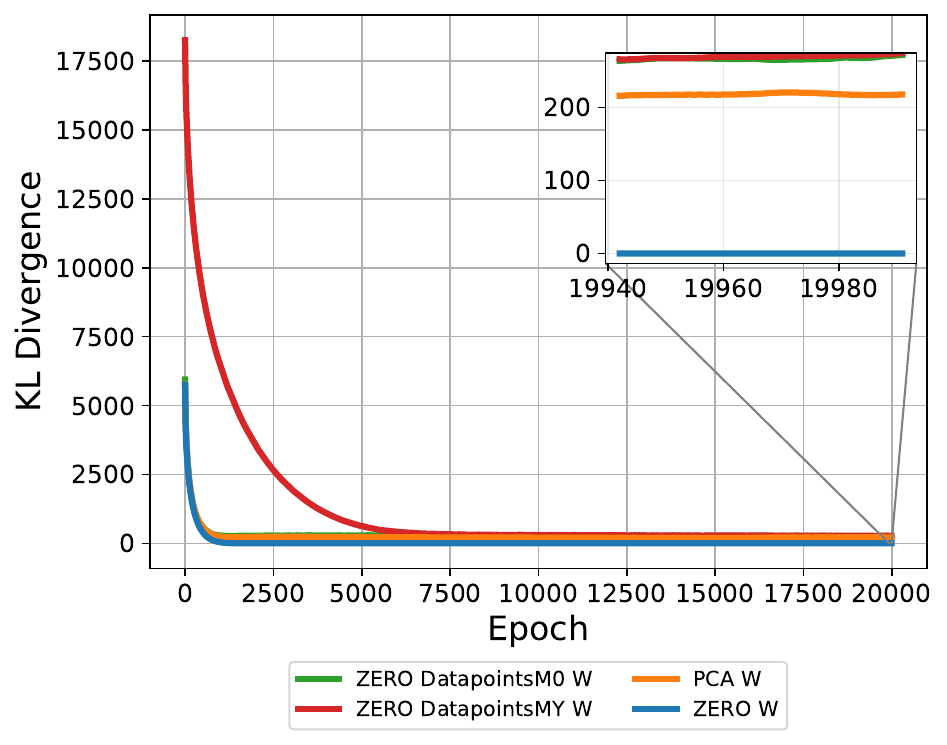}
        \caption{\KLD{}. The figure zooms in on the last 50 epochs.}
        \label{fig:results:Yacht:kld}
    \end{subfigure}
    \begin{subfigure}[t]{0.49\linewidth}
        \centering
        \includegraphics[width=\linewidth]{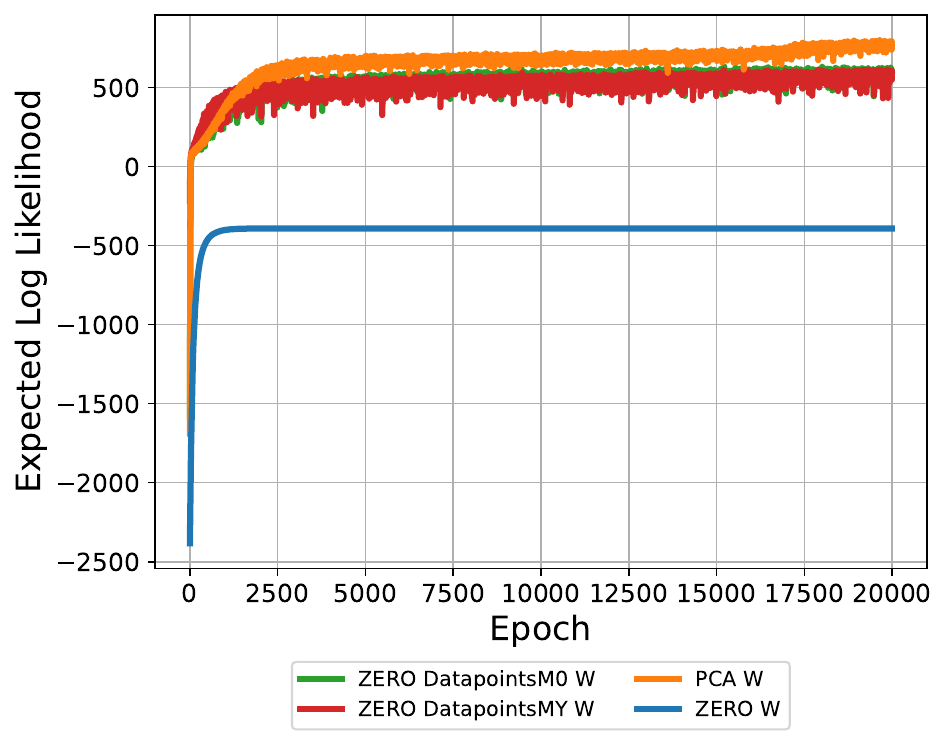}
        \caption{\ELL{}.}
        \label{fig:results:Yacht:ell}
    \end{subfigure}
    \caption{\KLD{} and \ELL{} obtained by the four models during training on the Yacht dataset, 
	when using $5$ layers, for a representative train/test split. The posterior collapse of the \ZEROW{} model is shown by the \KLD{}, which becomes zero, and 
	induces a poor \ELL{} term in the \ELBO.}
    \label{fig:kld:ell:Yacht}
\end{figure}

We first consider the whitened parameterization. In this case, \fig~\ref{fig:results:ll:whiten} shows poor performance of the 5-layer \ZEROW{} model on the Kin8nm and Yacht datasets, especially on the latter. To further analyze these results, we plot the \KLD{} and estimated noise variance of the \ZEROW{} and \ZEROWMO{} models over the last 50 epochs of the optimization process in \fig~\ref{fig:kld_llvar_comparison}, for each of the 20 training--test splits. The figure shows a horizontal line equal to $0$ for the Yacht dataset. More precisely, in this dataset, the results show that in $7$ out of the $20$ splits ($35\%$ of them) there is posterior collapse. This causes the model to estimate the prior distribution as its posterior, effectively providing a poor solution. This also explains the high standard deviation of the \ZEROW{} prior mean \DGP model in Yacht. The value of the likelihood variance coincides with $\nicefrac{1}{N}\sum_{n=1}^N (y_n)^2$ (horizontal line around $1.0$)\footnote{The data is standardized by subtracting the mean and dividing by the variance, hence the expected second moment coincides with the variance.}, which is the optimal coordinate update of the noise parameter. Since we are under a whitening parameterization, the model has learned $\sigma_0 = 0$, which is coherent with our discussion in \usec~\ref{subsec:simul:coord:updates}. Importantly, our initialization solves the posterior collapse problem, as \fig~\ref{fig:results:ll:whiten} shows. \fig~\ref{fig:kld_llvar_comparison} shows how the \MO model (green lines) presents, in general, higher \KLD and smaller likelihood variance than the \ZERO model (blue line). This behavior is consistent with a tendency toward posterior collapse, even when it does not occur completely. Furthermore, we compare all the models in \fig~\ref{fig:kld:ell:Yacht}, which shows for a representative train/test split, the \ELL and the \KLD of each method for each iteration epoch, on the Yacht dataset. We observe that the \ZEROW{} model results in a posterior collapse as the \KLD is zero. The \ZEROWMO{} and \ZEROWMY{} models, which also consider a \ZERO prior mean function, but use the proposed initialization, avoid this pathology, achieving a solution that is closer to that of the \DGP with the \PCA{} prior mean function, \ie, \PCAW{}, as indicated by \fig~\ref{fig:results:ll:whiten}. 

%
%
%
%
\begin{figure}[!p]
    \centering

    \begin{subfigure}{0.49\textwidth}
        \centering
        \includegraphics[width=\linewidth]{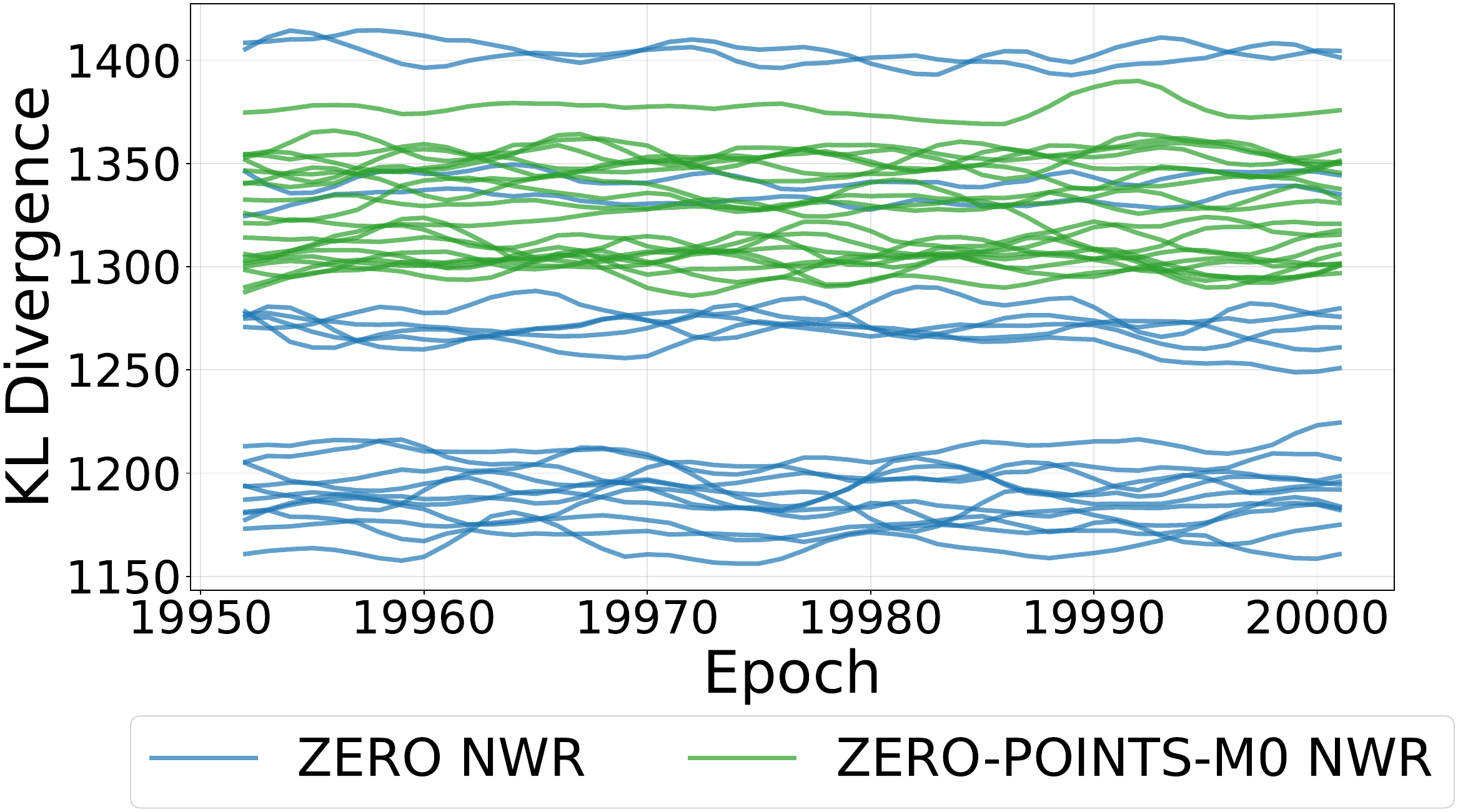}
        \caption{Kin8nm -- \KLD{}}
        \label{fig:results:kin8nm:kld:last:50:nwr}
    \end{subfigure}
    \hfill
    \begin{subfigure}{0.49\textwidth}
        \centering
        \includegraphics[width=\linewidth]{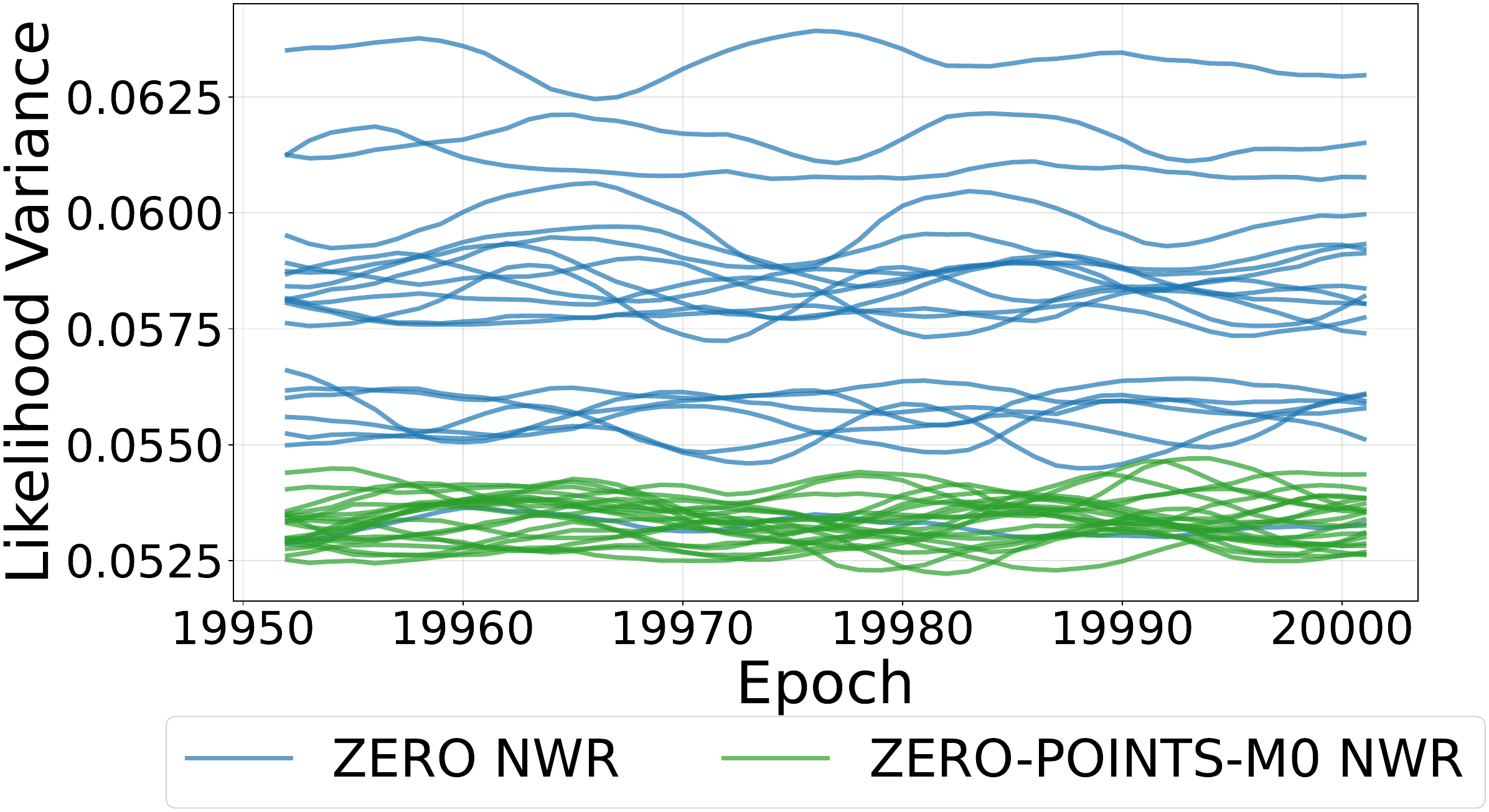}
        \caption{Kin8nm -- Noise variance}
        \label{fig:results:kin8nm:llvar:last:50:nwr}
    \end{subfigure}

    \begin{subfigure}{0.49\textwidth}
        \centering
        \includegraphics[width=\linewidth]{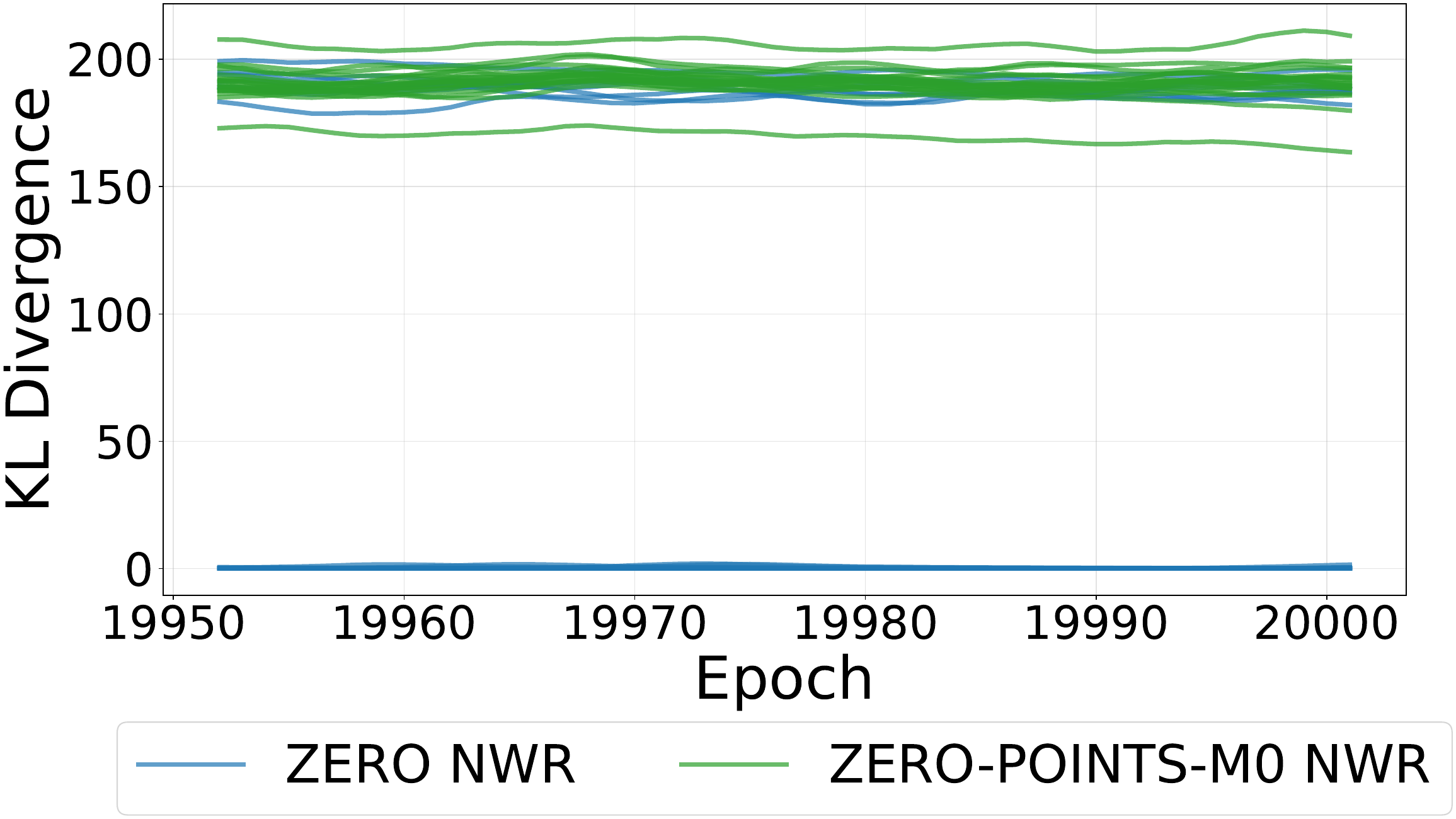}
        \caption{Boston -- \KLD{}}
        \label{fig:results:boston:kld:last:50:nwr}
    \end{subfigure}
    \hfill
    \begin{subfigure}{0.49\textwidth}
        \centering
        \includegraphics[width=\linewidth]{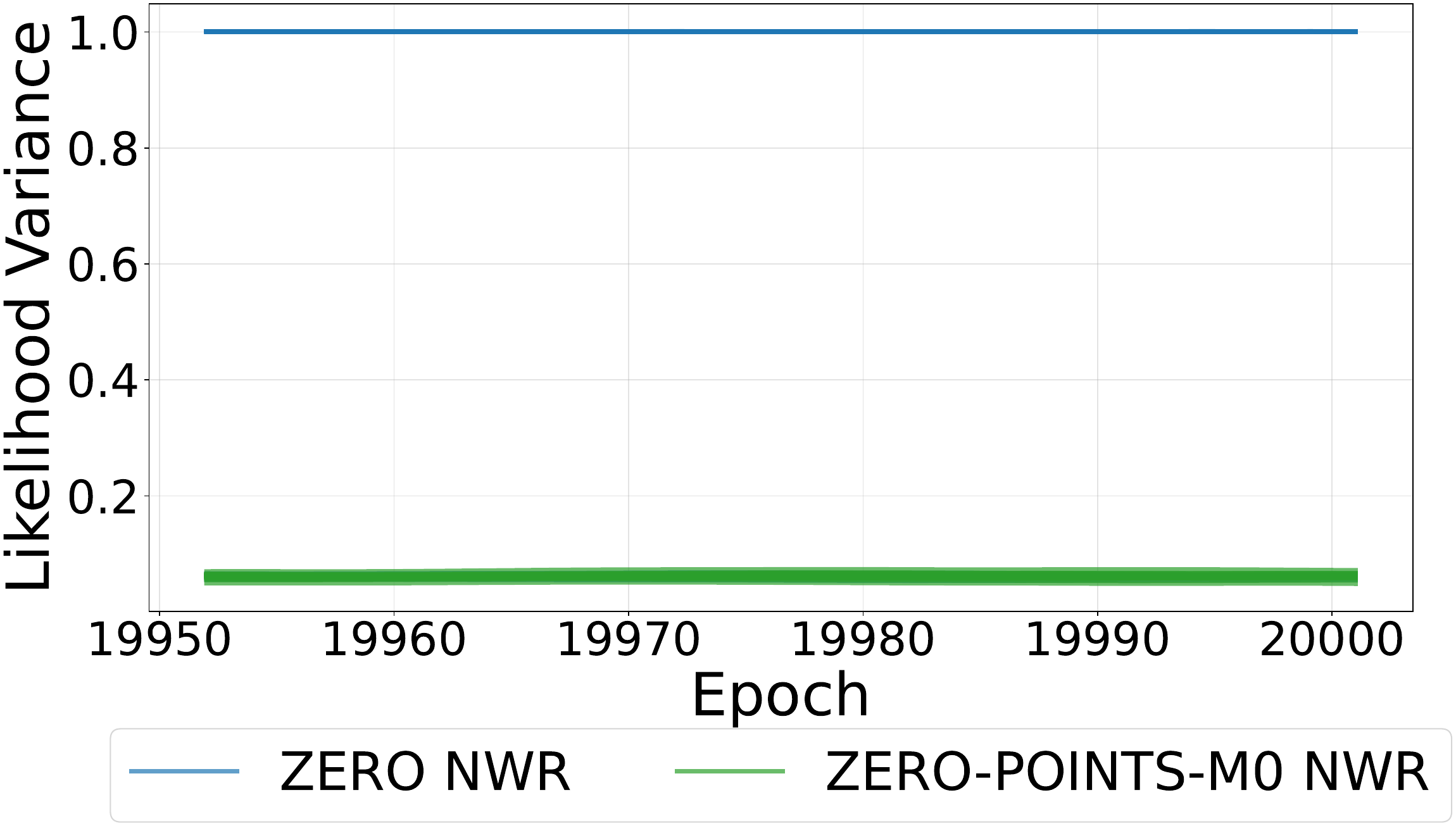}
        \caption{Boston -- Noise variance}
        \label{fig:results:boston:llvar:last:50:nwr}
    \end{subfigure}

    \begin{subfigure}{0.49\textwidth}
        \centering
        \includegraphics[width=\linewidth]{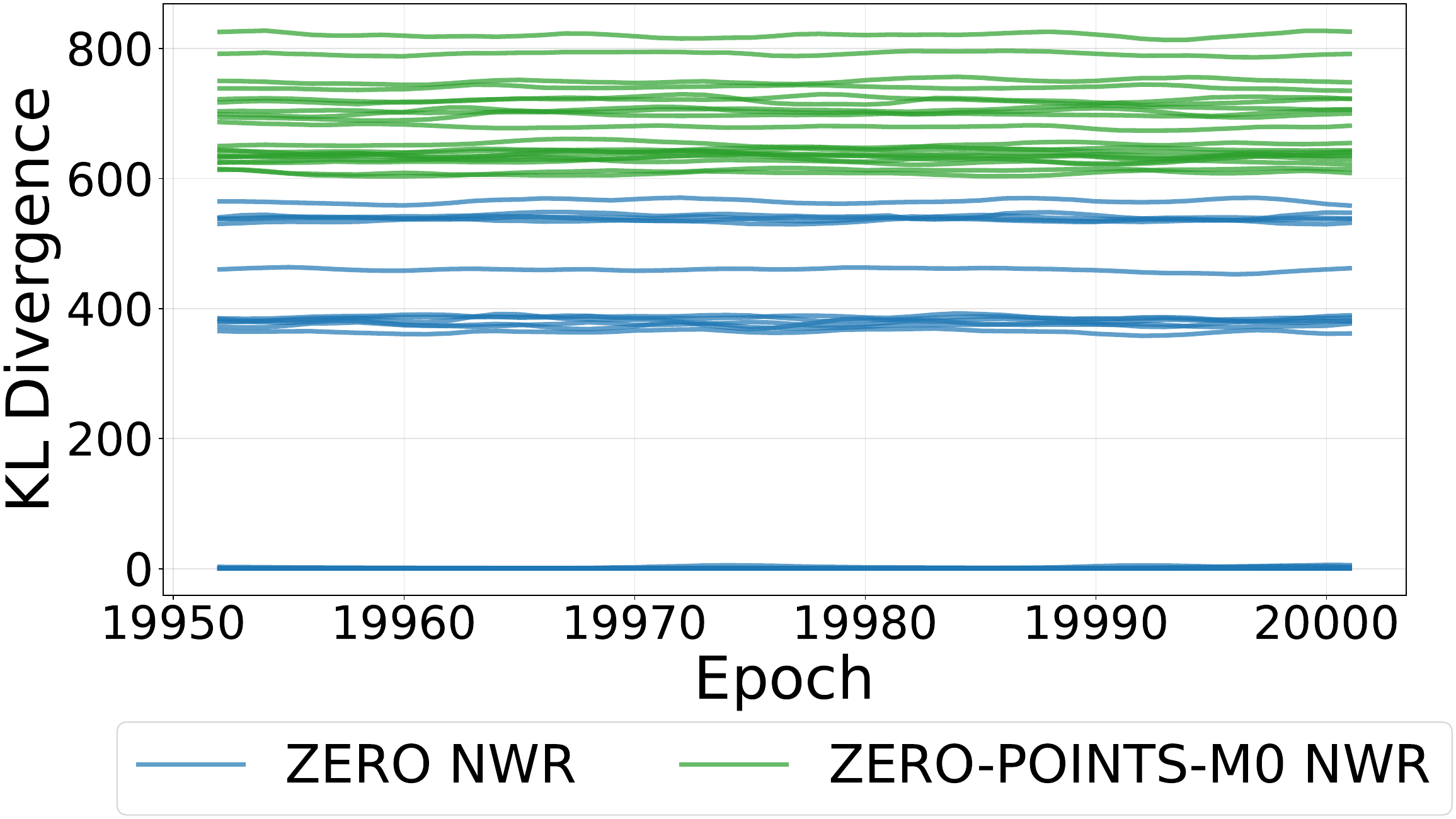}
        \caption{Power -- \KLD{}}
        \label{fig:results:power:kld:last:50:nwr}
    \end{subfigure}
    \hfill
    \begin{subfigure}{0.49\textwidth}
        \centering
        \includegraphics[width=\linewidth]{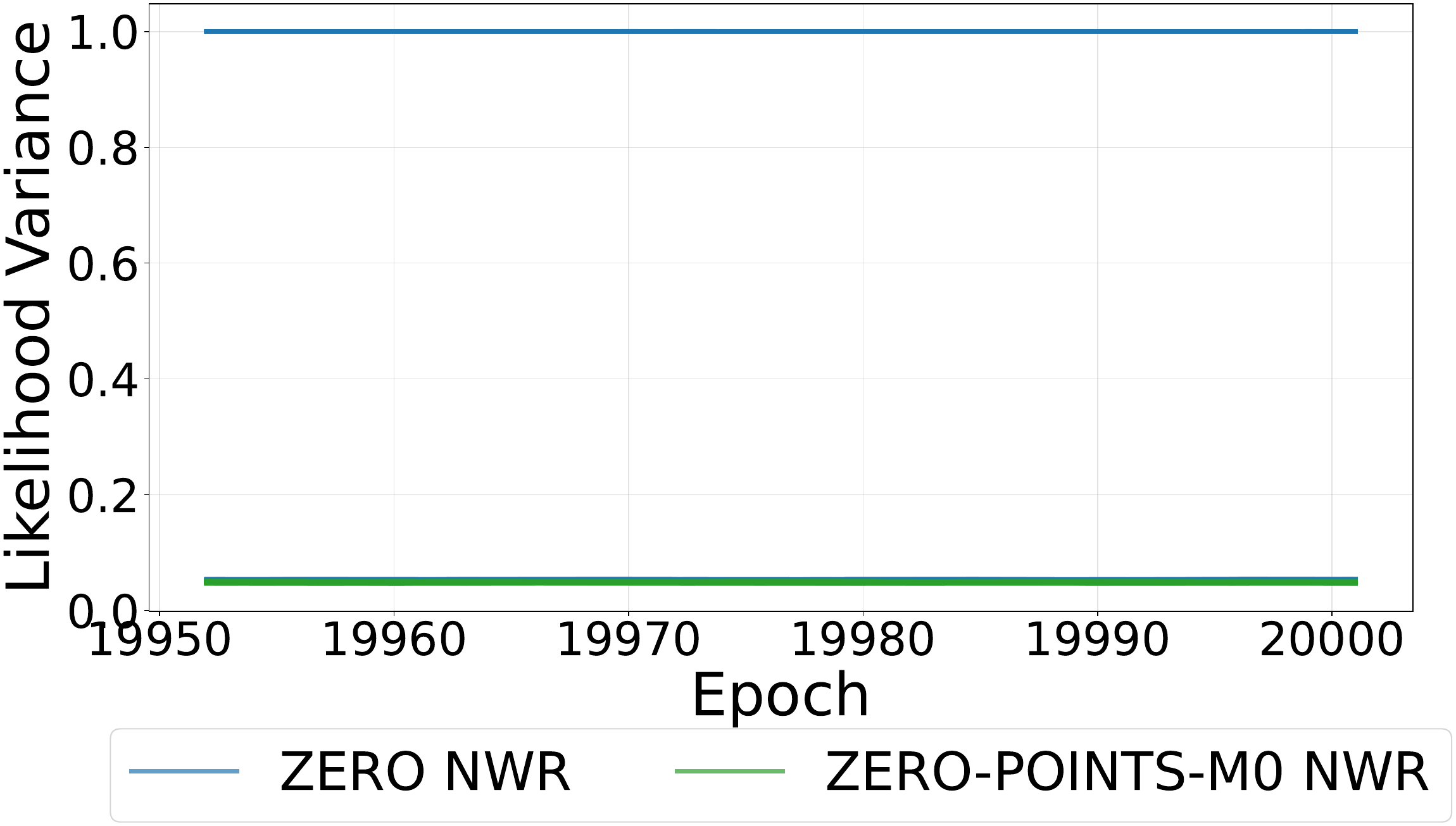}
        \caption{Power -- Noise variance}
        \label{fig:results:power:llvar:last:50:nwr}
    \end{subfigure}

    \begin{subfigure}{0.49\textwidth}
        \centering
        \includegraphics[width=\linewidth]{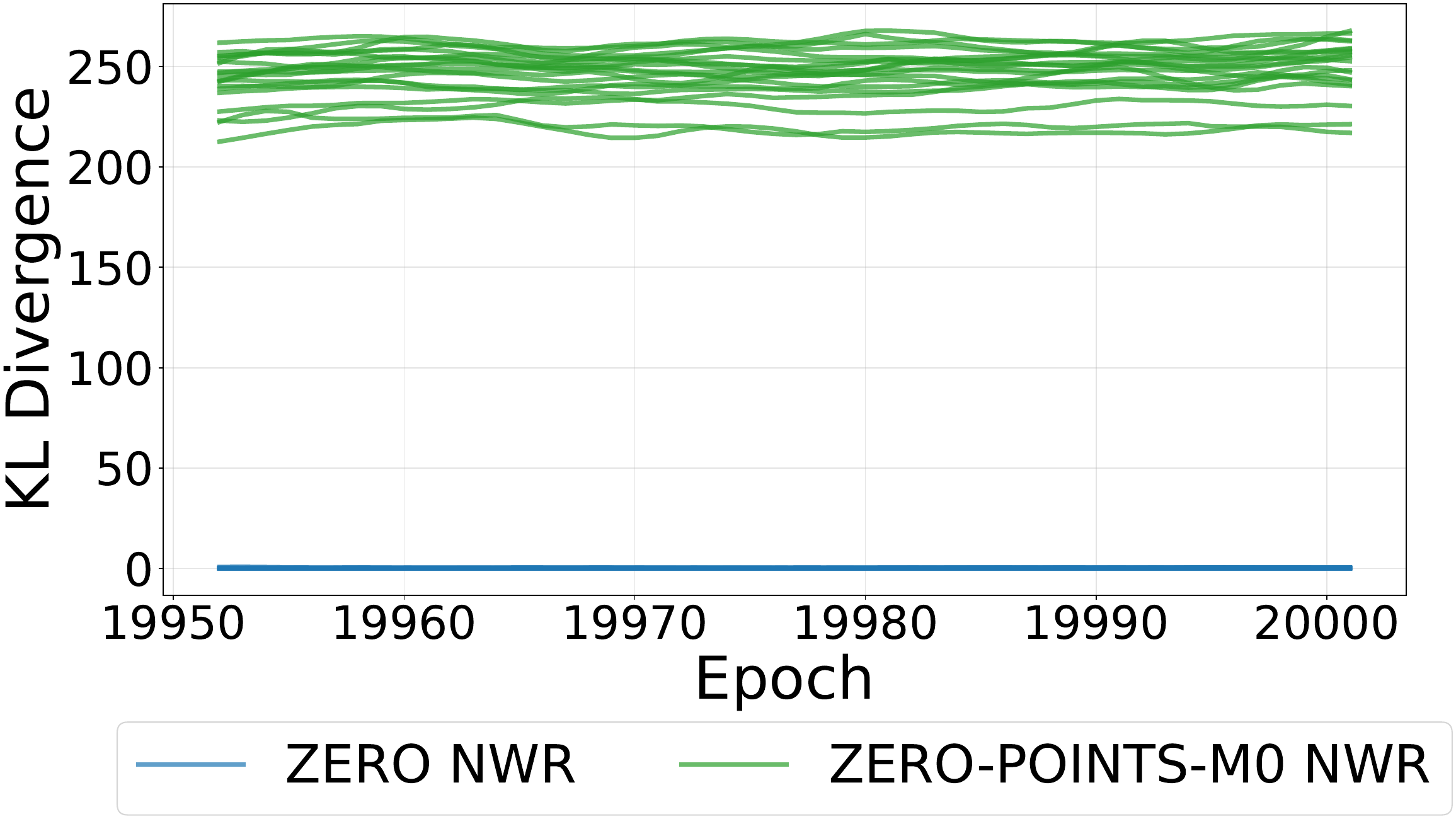}
        \caption{Yacht -- 4 Layers -- \KLD{}}
        \label{fig:results:yacht:4:kld:last:50:nwr}
    \end{subfigure}
    \hfill
    \begin{subfigure}{0.49\textwidth}
        \centering
        \includegraphics[width=\linewidth]{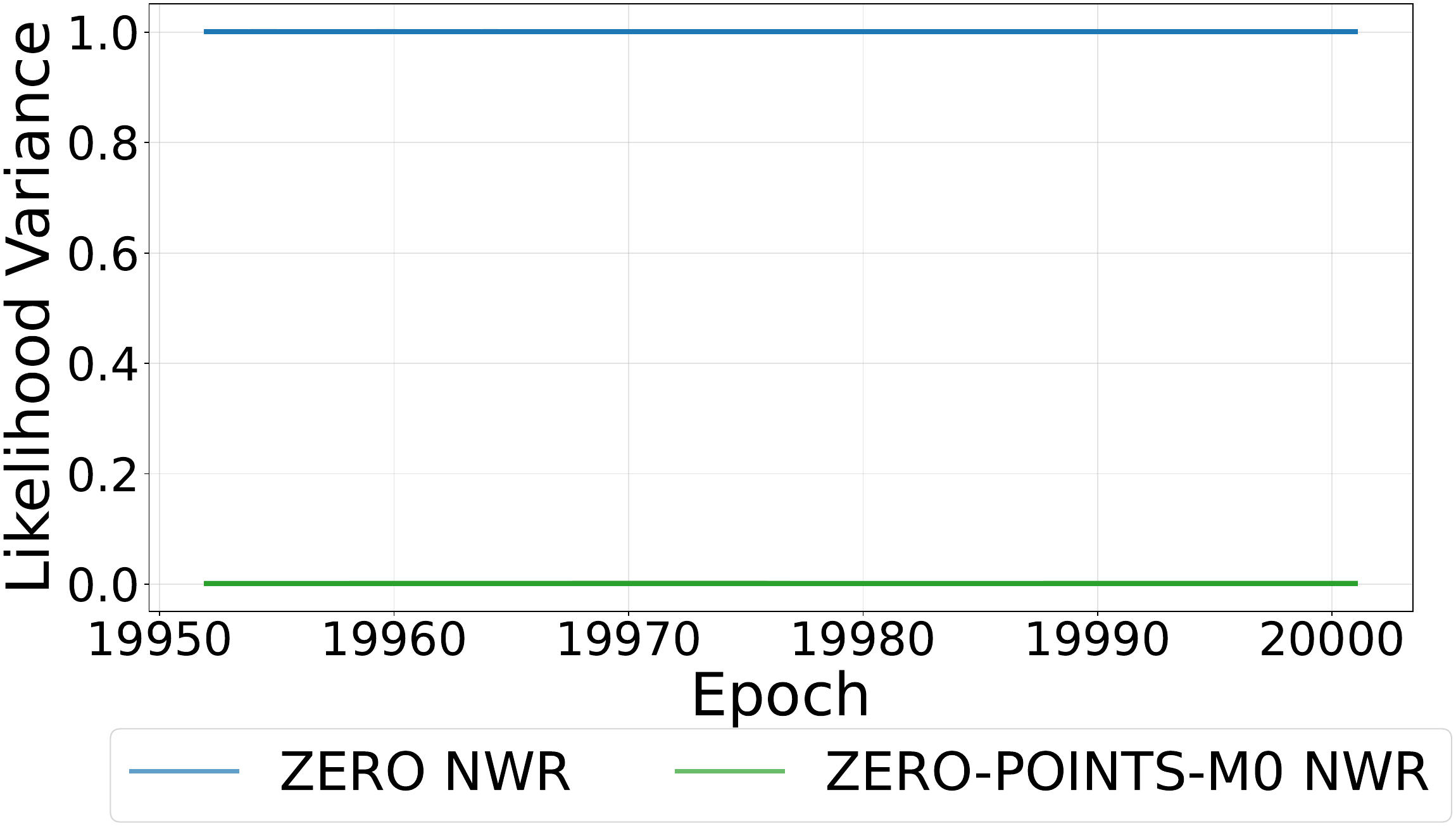}
        \caption{Yacht -- 4 layers-- Noise variance}
        \label{fig:results:yacht:4:llvar:last:50:nwr}
    \end{subfigure}

    \caption{
        \KLD{} (left) and likelihood variance (right) obtained by the
        \ZERONWR{} and \ZERONWRMO{} models in the Boston, Power, and Yacht (with $4$ layer model) datasets,
        for each train/test split.
    }
        \label{fig:kld_llvar_comparison_NWR}
\end{figure}

In the Kin8nm dataset, however, the poorer performance of the \ZEROW{} \DGP{} model is not accompanied by posterior collapse. As shown in \fig~\ref{fig:results:Kin8nm:kld:last:50}, the \KLD{} remains significantly above zero throughout training, providing no evidence of posterior collapse. Nevertheless, the \ZEROW{} model consistently exhibits a lower \KLD{} and a higher estimated noise variance than the \ZEROMO{} model (see \fig~\ref{fig:results:Kin8nm:kld:last:50} and \ref{fig:results:Kin8nm:llvar:last:50}). This suggests a tendency towards collapse, which results in explaining part of the data variability as noise, and accounts for the performance gap observed in \fig~\ref{fig:results:ll:whiten}. Notice that the \ZEROW{} model shows a test log-likelihood of $\approx 1.352$, compared to $\approx1.38$ for the \ZEROWMO{} model. To test whether this difference is significant, we perform a Wilcoxon signed-rank test which determines if the differences are centered around zero. The test outputs a $p-$value of $4.7\times 10^{-5}$, which is very close to zero, indicating that the differences are significant. Motivated by this observation, we further study the previous \emph{low \KLD{}--high observation noise}, which we observe in more scenarios, observation in Appendix \ref{sec:app:c:4:splits}. We confirm that many of the performance gaps between the \ZERO and \MO models come from this behavior, reaffirming the tendency of the \ZERO model towards collapsed solutions.

Lastly, we show that the poor solutions of the \ZERONWR{} models presented in \fig~\ref{fig:results:ll:nwr} often correspond to posterior collapse. First, \fig~\ref{fig:results:kin8nm:kld:last:50:nwr} and \ref{fig:results:kin8nm:llvar:last:50:nwr} indicate that the poor results on the Kin8nm dataset exhibit behavior similar to that observed in the whitened case: although \ZERONWR{} does not suffer from posterior collapse, it still shows a significant difference in test log likelihood ($p = 3.81\times10^{-6}$). In contrast, \ZERONWR{} suffers from posterior collapse in 13 out of 20 splits ($65\%$) on the Boston dataset and in 6 out of 20 splits ($30\%$) on the Power dataset. The corresponding \KLD{} and likelihood variance plots are shown in \fig~\ref{fig:results:boston:kld:last:50:nwr} and \ref{fig:results:boston:llvar:last:50:nwr} for Boston, and in \fig~\ref{fig:results:power:kld:last:50:nwr} and \ref{fig:results:power:llvar:last:50:nwr} for Power. Finally, the model exhibits complete posterior collapse in all 20 splits of the Yacht dataset when using four layers, making it the only configuration with fewer than five layers that displays this pathology. This behavior is illustrated in \fig~\ref{fig:results:yacht:4:kld:last:50:nwr} and \ref{fig:results:yacht:4:llvar:last:50:nwr}. Most importantly, both proposed initializations, \MO{} and \MY{}, consistently avoid the posterior collapse suffered by the \ZERO{} model, further highlighting the benefits of our initialization strategy.

\section{Conclusions and Future Work}
\label{sec:conclusions}

In this work, we presented a detailed analysis of the \DSVI{} algorithm used to train \DGP{}s and characterized the frequently encountered problem of posterior collapse. Specifically, the optimization process reaches a local optimum that minimizes the \KLD{} by simply setting the variational parameters equal to those of the prior. Consequently, the model's predictions collapse into noise, yielding a poor and uninformative solution.

We observed that posterior collapse is more likely to happen when using a \ZERO{} prior mean function in the inner layers of the \DGP{}. By contrast, using a \PCA{} prior mean function may sometimes solve this issue. However, this imposes an important constraint on the prior mean functions that may not be desirable in some problems or extensions of the \DGP{} model. An example is the use of transformed \GP{}s instead of standard \GP{}s \citep{tgp,dtgp}. Moreover, the choice of the model's prior should be done in such a way that reflects particular characteristics of the learning problem, not simply as a consequence of the optimization problems that arise when fitting the model \citep{GVI_jer}. In addition, our analysis indicates that avoiding non-injectivity pathologies is not the fundamental reason for the effectiveness of the \PCA prior mean function, in the context of \DGP{}s, as previously thought in the literature. In fact, we have shown that a very deep \DGP with a whitened parameterization and this prior mean function can lead to posterior collapse if enough noise is introduced in the inner layers. Besides this, we have also shown that the whitened parameterization yields more stable optimization for the \ZERO{} prior mean function model, which is often translated into better predictive distributions. Yet, our \UCI results show that the non-whitened parameterization can result in better predictive performance. Since the \PCA mean function under this parameterization imposes a different statistical model, this represents a line of research.

Based on our observations, we proposed a novel initialization of the \DGP{} model that, by modifying the initial value of the variational mean parameter, mimics the behavior of the \PCA{} prior mean in the inner layers of the model, at initialization. This enables maintaining the typical \ZERO{} prior mean function used in the context of \GP{}s. Furthermore, using a minimal modification of the proposed initialization in the output layer produces an initial predictive distribution that is accurate at the inducing points (selected from a subset of the training set). We then experimentally showed that, thanks to the proposed initialization, the \ZERO{} prior mean \DGP{} can successfully avoid the posterior collapse phenomenon. In addition, we demonstrate that the results obtained, when using our proposed initialization, outperform the results of standard initialized \ZERO{} prior mean \DGP{}s, and that can more closely match those of the \PCA prior mean \DGP{}. 

Even though our proposed initialization alleviates the posterior collapse problem and outperforms the standard initialization of \ZERO prior mean \DGP{}s, it presents two main limitations. Firstly, it does not match exactly the predictive performance of the \PCA{} mean function model, which is still better in most of the cases. Secondly, the proposed solution requires solving a system of linear equations. The numerical stability of the solution depends on the number of inducing points and the initial parameters of the kernel, which could compromise the quality of the proposed initialization. Additionally, in the case of initializing the output layer to accurately predict the targets at the inducing points, we also observed a big initial and final \KLD term, which can also complicate the fitting of the model. Specifically, a big initial \KLD results in a smaller initial \ELBO value. Improving the support where our initialization matches the \PCA mean function, and improving the numerical stability of the initialization are lines of future research. The first line can be suitably approached through Deep Transformed \GP{}s \citep{dtgp}.

In any case, by avoiding posterior collapse without constraining the prior mean, the method provides a practical strategy for challenging problems modeled by \DGP{}s and opens a new direction for research on initialization strategies in such models. Moreover, it also enables considering more flexible models in the context of \DGP{}, such as transformed \GP{}s \citep{tgp,dtgp}, in which using the \PCA prior mean function is challenging. 

\acks{The authors acknowledge financial support by Grant PID2022-140189OB-C22 funded by
MICIU AEI/10.13039/501100011033 and by NextGenerationEU/PRTR. 
The authors also acknowledge support from the project
PID2022-139856NB-I00, funded by MCIN\slash AEI\slash 10.13039\slash
501100011033\slash FEDER, UE; from project IDEA-CM (TEC-2024\slash COM-89),
funded by the Autonomous Community of Madrid; and from the ELLIS Unit Madrid.
The authors also acknowledge computational support from the Centro de 
Computaci\'on Cient\'ifica-Universidad Aut\'onoma de Madrid (CCC-UAM).
}

\FloatBarrier



\vskip 0.2in


\FloatBarrier
\clearpage
\appendix
\section{Coordinate Updates for the Noise Parameter}\label{sec:app:A}

By noting that a one-dimensional Gaussian likelihood function can be compactly expressed through:
\begin{equation}
    \myprod{n=1}{N} \Ngauss{y^n\mid f_n,\sigma^2} = \Ngauss{\Ysamples{}\mid \f{},\sigma^2\matI}
\end{equation}
The objective function of a one-dimensional \SVGP can be written as:
\begin{align*}
\ELBO &= \int q(\f{})\log \Ngauss{\Ysamples{}\mid \f,\sigma^2\matI}\dd\f{} - \KLD\bra{q(\uu{})\mid\mid p(\uu{})}\\
&= \log \Ngauss{\Y{}{} \mid \qfm{}{} , \sigma^2\matI+\qfS{}{}}-\frac{1}{2}\tr{\pare{\sigma^2\matI}^{-1}\qfS{}{}}  - \KLD\bra{q(\uu{})\mid\mid p(\uu{})}
\end{align*}
Getting the differential over $\sigma^2$:
\begin{align*}
    \dd_{\sigma^2} \ELBO =  \bra{-\frac{N}{2} + \frac{1}{2\sigma^2}\bra{\pareT{\Ysamples{} - \qfm{}{}}\pare{\Ysamples{}-\qfm{}{}} + \tr{\qfS{}{}}}} \dd\sigma^2
\end{align*}
Since this is a scalar-valued, scalar argument function, the differential directly identifies the gradient. Thus:
\begin{align*}
    &-\frac{N}{2} + \frac{1}{2\sigma^2}\bra{\pareT{\Ysamples{} - \qfm{}{}}\pare{\Ysamples{}-\qfm{}{}} + \tr{\qfS{}{}}} = \veczero\\
    &\sigma^2 = \frac{1}{N}\bra{\pareT{\Ysamples{} - \qfm{}{}}\pare{\Ysamples{}-\qfm{}{}} + \tr{\qfS{}{}}} \\
    &\sigma^2 = \frac{1}{N}\Big(\mid\mid\Ysamples{} - \qfm{}{} \mid\mid^2_2 + \tr{\qfS{}{}}\Big)
\end{align*}
This is equivalent to the expected value under $q(f_n)$ of the average squared distance between the target $y^n$ and the \DGP output $f_n$.

For the \ELL term in \ueqn \ref{eq:ell_expansion} for the \DGP, the derivation is exactly the same. Note that in the derivation we have done, the \KLD does not take part in the differential. Now for any distribution $q(\f{})$ with finite mean and covariance it holds:

\begin{equation}
\begin{split}
&\int q(\f{})\log \Ngauss{\Ysamples{}\mid \f{},\sigma^2\matI} \dd\f{}\\
&= \log \Ngauss{\Y{}{} \mid \qfm{}{} , \sigma^2\matI+\qfS{}{}}-\frac{1}{2}\tr{\pare{\sigma^2\matI}^{-1}\qfS{}{}}
\end{split}    
\end{equation}
Thus, if we substitute $q(\fpos{}{L})$ by the distribution over the last layer of a \DGP{}, then the \ELL takes the above expression and so the coordinate update for $\sigma^2$ is exactly the same as that of a \SVGP using the predictive mean and predictive covariance of the non-Gaussian distribution $q(\fpos{}{L})$.

For a $C$ Multi-output \SVGP one can follow a similar reasoning. Here, however, while we can express the likelihood function in a compact way using Kronecker products, the final expression does not match the same ordering as the latent processes of the variational posterior. Thus, here we explicitly write the likelihood function as a sum over samples. The noise parameter of the likelihood function is expressed through $\Sigma$, assuming the most general case, which is a dense matrix. This gives:
\begin{align*}
    \ELBO = \mysum{n=1}{N}\log p(\Y{}{n} \mid \qfmall{n}{}, \Sigma +\qfSall{n}{} ) -\frac{1}{2}\tr{\Sigma^{-1}\qfSall{n}{}} -\KLD\bra{q(\uallpos{}{})\mid\mid p(\uallpos{}{})}
\end{align*}
Here now $\Y{}{n} \in \mathbb{R}^C$ and is a vector $\qfmall{n}{}$ containing the variational mean of each process at point $n$. In contrast, $\qfSall{n}{}$ is a $C\times C$ matrix containing dependencies between all the processes at point $n$. The first order differential \wrt the noise matrix:
\begin{align*}
\dd_{\Sigma}\ELBO = \frac{1}{2}\mysum{n=1}{N}\vvecB{\Sigma^{-1}\pare{\Y{}{n} - \qfmall{n}{}}\pare{\Y{}{n} - \qfmall{n}{}}^T\Sigma^{-1}-\Sigma^{-1}+\Sigma^{-1}\qfSall{n}{}\Sigma^{-1}}^T\dd\vvec\Sigma
\end{align*}
From here, we can identify the gradient and undo the vec operator to retrieve the optimal update in terms of a matrix. Setting the expression to zero and operating yields:
\begin{align*}
    &\mysum{n=1}{N}\Sigma^{-1}\pare{\Y{}{n} - \qfmall{n}{}}\pare{\Y{}{n} - \qfmall{n}{}}^T\Sigma^{-1}-\Sigma^{-1}+\Sigma^{-1}\qfSall{n}{}\Sigma^{-1} = \veczero\\
    &\Sigma = \frac{1}{N}\mysum{n=1}{N}\pare{\Y{}{n} - \qfmall{n}{}}\pare{\Y{}{n} - \qfmall{n}{}}^T + \qfSall{}{}
\end{align*}
If one now assumes that all the outputs share the same noise parameter, or that we want independent outputs each with its own noise, then either the differential must be recomputed or just use the differential of a diagonal matrix, which is easier than using Lagrange multipliers \citep{minka}.
\clearpage

\section{Additional Details of the Proposed Initialization}
\label{sec:app:B}

This appendix provides additional insights into how to solve the systems of equations that result from the proposed initialization. While we have only evaluated the most efficient version, for completeness, we provide the full analysis and derivations. This might be used to initialize the non-whitened models implemented in \GPYTORCH or \GPJAX or to increase the support of points $\Xsel$ where the \ZERO model mimics the \PCA model at initialization. 

From the main paper, we know that the three systems of equations that result from our idea are:
\begin{align}
   \Kxz{}\Kzzinv{}\varm{}{} &= \Xsel -  \Kxsz{}\Kzzinv{}\Zsamples{} \tag{\NW}\\
   \Kxz{}\braT{\Lzzinv{}}\varmw{} &= \Xsel \tag{\W} \\
   \Kxz{}\Kzzinv{}\varmr{} &= \Xsel \tag{\NWR}
\end{align}
First, we noted that we do not require the set of points $\Xsel$ to be the same as $\Xsamples$ in order to define the system, but it is required that $\dim(\Xsel)=\dim(\Xsamples)$ for the three systems to be well-defined. However, since it is counterintuitive to require the \ZERO \DGP to mimic the \PCA \DGP at different points in the domain, our first assumption is $\Xsamples=\Xsel$.

With this in mind, we will now see that these systems can be solved in many ways. We will analyze all the options and show how we can yield a linear solvable system that can be solved without requiring numerical gradient-based optimization. As we mentioned, this is not a restriction of our approach but a decision we have made for our experiments. Thus,  there is freedom to go through other approaches that might be interesting for other reasons (for example, incrementing the support $\Xsel$ where the desired initialization is satisfied).

\subsection{Equations and Parameters from the System of Equations}
The resulting systems of equations have a total of $\dim(\Xsel)$ equations, which correspond to the points where we aim to mimic the \PCA \DGP{}. Ideally, we would like $\Xsel = \Xspace$. However, since this would imply having an infinite number of equations, we start by fixing a finite number of points $\Xsel$, yet to be determined, where the equation will be satisfied.

On the other hand we have a total of $\dim(\Xsel)+2\dim(\Zsamples{})+\dim(\nu)$ variables: the $\Xsel$ where the posterior is evaluated, the value of the inducing points $\Zsamples{}$, the variational mean $\varmw{}{},\varmr{},\varm{}{}$ and the kernel hyper-parameters $\nu{}$. Note that $\dim(\varm{}{})=\dim(\Zsamples{})$. Since there are more variables than equations, solving the system for all the variables results in an infinite number of solutions. Thus, we need to fix some values for the variables and solve for the others.
\subsection{Fixing the Kernel Parameters}

Having fixed the number of equations to $\dim(\Xsel)$, we need to fix a number of variables equal to the number of equations for the system to be compatible. If $\dim(\Xsel)$ is greater than the number of parameters, then the system is overestimated, which results in a system without a solution unless some of the equations are linear combinations of the others.

The immediate consequence is that fixing values for $\Xsel$, $\Zsamples{}$ and $\varm{}{}$ and solving for $\nu$ is not a good choice, because the number of kernel hyper-parameters is usually small, and that would imply that our initialization only resembles the \PCA model at very few points. For instance, for an \RBF kernel, we would only be able to mimic the \PCA \DGP{} at two points in the domain (assuming the kernel does not implement automatic relevance determination). Moreover, this case entails solving a nonlinear system of equations using gradient-based optimization, as the kernel parameters appear nonlinearly in the equation. 

Consequently, by fixing $\nu$, we now have $\dim(\Xsel)$ equations and $\dim(\Xsel)+2\dim(\Zsamples{})$ variables, and so we need to fix more variables. We describe two alternatives.
\subsection{Option 1: $2\dim(\Zsamples{})$ Equations}

The first option we can think about is a system in which we have $2\dim(\Zsamples{})$ equations. In other words, the number of points at which we want to mimic the \PCA \DGP{} is twice the number of inducing points. Since the number of variables $\dim(\Xsel)+2\dim(\Zsamples{})$ is greater than the number of equations, we have three possibilities for fixing values in order to solve the system:

\begin{itemize}
    \item Fix values for $\varm{}{}$ and $\Zsamples{}$ and solve the system for $\Xsel$.
    \item Fix either $\varm{}{}$ or $\Zsamples{}$ and a total of $\dim(\Zsamples{})$ points from $\Xsel$ and solve the system for the rest of $\Xsel{}$ locations and either $\varm{}{}$ or $\Zsamples{}$, depending on what we have fixed.
    \item Fix $\Xsel$ and solve the system for $\varm{}{}$ and $\Zsamples{}$.
\end{itemize}

However, this approach presents several difficulties. The first one is that when fixing a value for $\varm{}{}$ we need to choose one far from $\veczero{}$, since these are the values that can make the model collapse and because that would imply the solution $\Xsel=\veczero$ for the \NWR and \W model, with independence of the value selected for $\Zsamples{}$. Second, solving the system for $\Xsel$ may not be useful since optimization can solve for a $\Xsel$ that is far from our training data (since $\Xsel$ can be any point in the domain), and we would like our model to be initialized at points next to the training data or the points where we are likely to make predictions. Third, fixing $\Xsel$ or $\Zsamples{}$ can result in a system with no closed-form solution, due to the nonlinear relation in the equation from $\Xsel$ and $\Zsamples{}$. In this case, the system would be solved by numerical gradient-based optimization, minimizing least squares. For example fixing $\Xsel$ and solving for $\varmr{}{}$ and $\Zsamples{}$ can be done through:
\begin{equation}
    \underset{\varmr{}{},\Zsamples{}}{\argmin} ||\Kxsz{}\Kzzinv{}\varmr{} - \Xsel||_p
\end{equation}
which shows that this approach also has a cubic cost per gradient update. 
Thus, none of the options outlined in this section are good due to either an efficiency bottleneck or the plausible set of solutions achieved.
\subsection{Option 2: $\dim(\Zsamples{})$ Equations}
From the previous section, we have learned that it is interesting to solve the system at points in which $\Xsel$ is representative of the places where we make predictions, and we would like to avoid cubic cost and numerical gradient-based optimization.

As we now see, if we sacrifice the total number of points where we mimic the \PCA \DGP{} to the half, \ie from $2\dim(\Zsamples{})$ to $\dim(\Zsamples{})$, we can yield an easier system to solve. Again, we have three options.
\begin{itemize}
    \item Fix a value for $\varm{}{}$ and $\Zsamples{}$ and solve for $\Xsel{}$, which is not a good option since it requires numerical gradient-based optimization and does not ensure initializing the model in a desired region of the space $\Xspace$.
    \item Fix a representative value for $\Xsel{}$ (for example a subset of the training data or some iterations of k-means) and $\varm{}{}$ and solve the system for $\Zsamples{}$, which is not a good option because we require non-linear optimization of the least square function, and also require  $\varm{}{}$ to be different from $\veczero$.
    \item Fix values for $\Xsel$ and $\Zsamples{}$ and solve for $\varm{}{}$, which results in a linear system of equations that can be solved, at most, in two cubic operations, which makes it the most computationally efficient option for solving the systems.
\end{itemize}

With this last option, we have that the solution for the optimal value $\varm{}{}$, which would resemble the \PCA mean function at initialization, at some selected $\Xsel$ and $\Zsamples{}$ is given by (for the three parameterizations considered): 
\begin{align}
    \varm{}{} &= \braInv{\Kxsz{}\Kzzinv{}}\bra{\Xsel - \Kxsz{}\Kzzinv{}\Zsamples{}} \tag{\NW{}}\\
    \varmw{} &= \braInv{\Kxsz{}\braT{\Lzzinv{}}}\Xsel \tag{\W{}}\\
    \varmr{}{} &= \braInv{\Kxsz{}\Kzzinv{}}\Xsel \tag{\NWR{}}
\end{align}
Thanks to fixing $\Xsel$ and $\Zsamples{}$, we yield linear systems with exact solutions, but require two cubic operations in the number of inducing points, one per matrix inversion. Nevertheless, choosing $\Xsel$ and $\Zsamples{}$ to be exactly the same points, we can yield a much more efficient solution. There are two options here: either select $\Xsel$ to be the inducing points $\Zsamples{}$ (obtained through k-means at initialization), or select a subset from $\Xsel$ to be the inducing points. In any case since now $\Xsel = \Zsamples{} \coloneqq \Xxz$, the solutions to the systems are:
\begin{equation}\label{equ:proposed_init_at_inducing_2}
\begin{aligned}
    \varm{}{} &= \veczero;  && \text{using } \K{\Xxz}{\Xxz}\Kinv{\Xxz}{\Xxz} = \matI\\
    \varmw{} &= \Linv{\Xxz}{\Xxz}\Xxz;  && \text{using } \K{\Xxz}{\Xxz} = \Lchol{\Xxz}{\Xxz}\Lchol{\Xxz}{\Xxz}^T\\
    \varmr{}{} &= \Xxz;  && \text{using } \K{\Xxz}{\Xxz}\Kinv{\Xxz}{\Xxz} = \matI 
\end{aligned}
\end{equation}

\subsection{Selection of the Initial Length-scale}

\begin{figure}[!htb]
    \centering
    \begin{subfigure}[t]{0.49\linewidth}
        \centering
        \includegraphics[width=\linewidth]{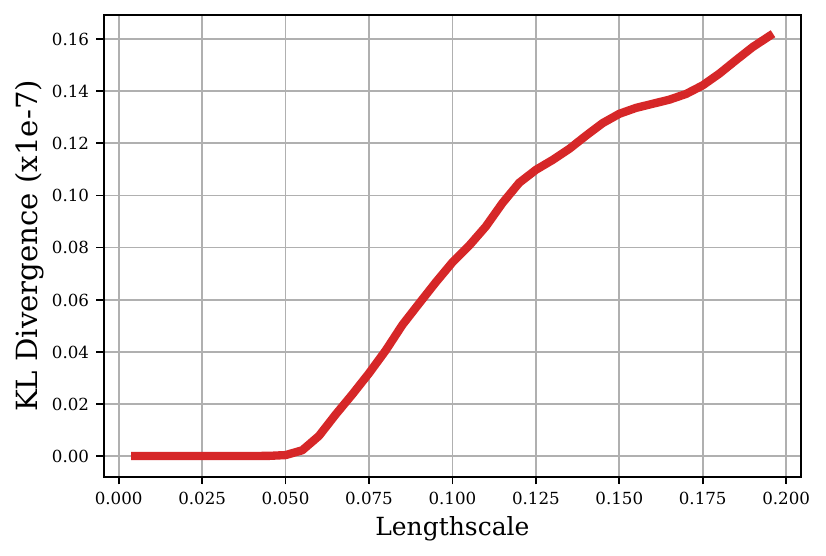}
        \caption{\KLD{}.}
        \label{fig:kld:init:my}
    \end{subfigure}
    \hfill
    \begin{subfigure}[t]{0.49\linewidth}
        \centering
        \includegraphics[width=\linewidth]{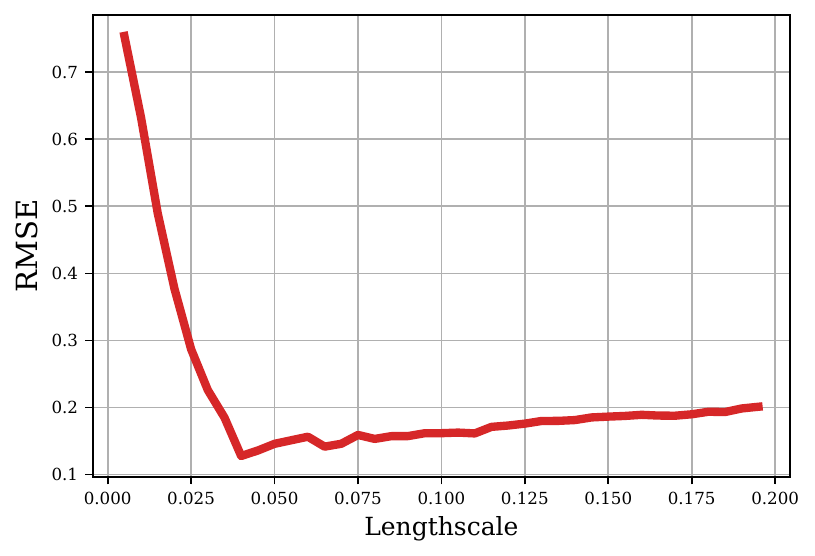}
        \caption{\RMSE{}.}
        \label{fig:rmse:init:my}
    \end{subfigure}
    \caption{\KLD{} and \RMSE{} obtained by the two layer \ZEROWMY{} model when varying the length-scale $\ell$ at initialization in the steps dataset.}
    \label{fig:kld:rmse:init:my}
\end{figure}

In \usec~\ref{sec:initialization:inital:lengthscale}, we highlighted the importance of selecting an appropriate initial length-scale for the \ZEROMY{} model to obtain good initial predictive performance. Nevertheless, this choice can entail a trade-off between the achieved \RMSE{} and the resulting \KLD{}. \fig~\ref{fig:kld:rmse:init:my} presents the variation of the initial \KLD{} and \RMSE{} of the \ZEROWMY{} model with respect to the initial length-scale.

\clearpage
\section{Further Experimental Results}\label{sec:app:C}

This section completes some of the experiments or simulations of the main paper. More precisely, we first provide additional figures for the theoretical analysis performed in \usec~\ref{sec:optimization:difficulties}. Then, we provide further observations from the results presented in \usec~\ref{sec:experiments}, both for the toy dataset and \UCI{} real-world datasets.

\subsection{Additional Figures for Section \ref{sec:optimization:difficulties}}
\label{sec:app:c:1:opt:difficulties}

\begin{figure}[!p]
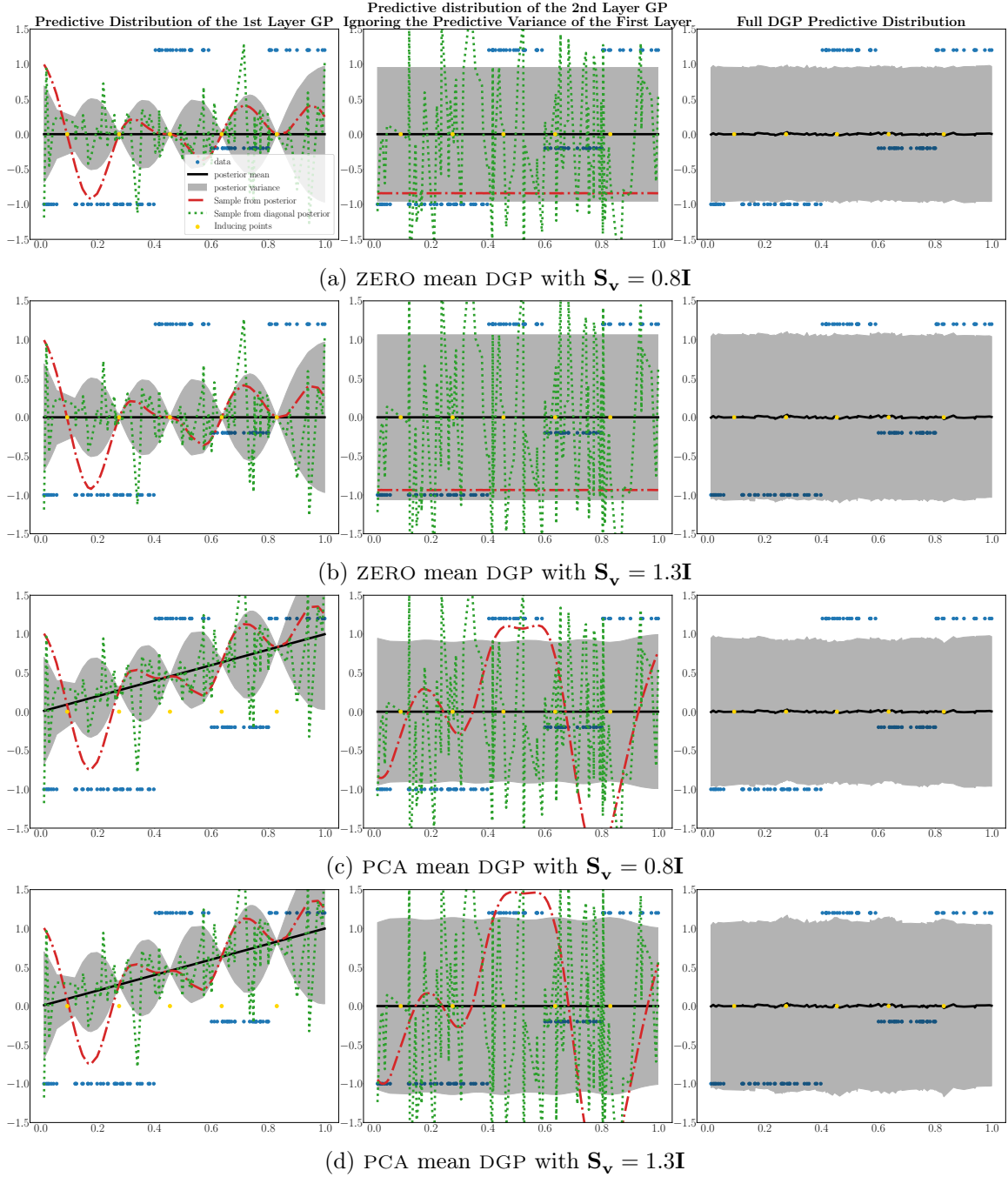

    \begin{subfigure}{\textwidth}
        \centering
        \includegraphics[width=\textwidth]{imgs_jmlr/4-initialization_issues/S_out_1_S_inner_e5/marginal_var_pos_at_init_model_zero_S_inner_1e-05_S_out_0.8_num_Z_5_ls_0.1_whiten_True.pdf}
        \caption{\ZERO mean \DGP{} with $\varSw{}{}=0.8\matI$}
    \end{subfigure}
    \begin{subfigure}{\textwidth}
        \centering
        \includegraphics[width=\textwidth]{imgs_jmlr/4-initialization_issues/S_out_1_S_inner_e5/marginal_var_pos_at_init_model_zero_S_inner_1e-05_S_out_1.3_num_Z_5_ls_0.1_whiten_True.pdf}
        \caption{\ZERO mean \DGP{} with $\varSw{}{}=1.3\matI$}
    \end{subfigure}
    \begin{subfigure}{\textwidth}
        \centering
        \includegraphics[width=\textwidth]{imgs_jmlr/4-initialization_issues/S_out_1_S_inner_e5/marginal_var_pos_at_init_model_PCA_S_inner_1e-05_S_out_0.8_num_Z_5_ls_0.1_whiten_True.pdf} 
        \caption{\PCA mean \DGP{} with $\varSw{}{}=0.8\matI$}
    \end{subfigure}
    \begin{subfigure}{\textwidth}
        \centering
        \includegraphics[width=\textwidth]{imgs_jmlr/4-initialization_issues/S_out_1_S_inner_e5/marginal_var_pos_at_init_model_PCA_S_inner_1e-05_S_out_1.3_num_Z_5_ls_0.1_whiten_True.pdf} 
        \caption{\PCA mean \DGP{} with $\varSw{}{}=1.3\matI$}
    \end{subfigure}
    \caption{\ZERO (top) and \PCA (bottom) \DGP models with the output layer variational covariance initialized to $\varSw{}{}=1.3\matI$ and $\varSw{}{}=0.8\matI$ , in the whitened parameterization.}
\label{fig:DSVI_zero_pca_pos_toy_at_init_Sout_1.30.8_Sinner1e-5}
\end{figure}

\begin{figure}[!htp]
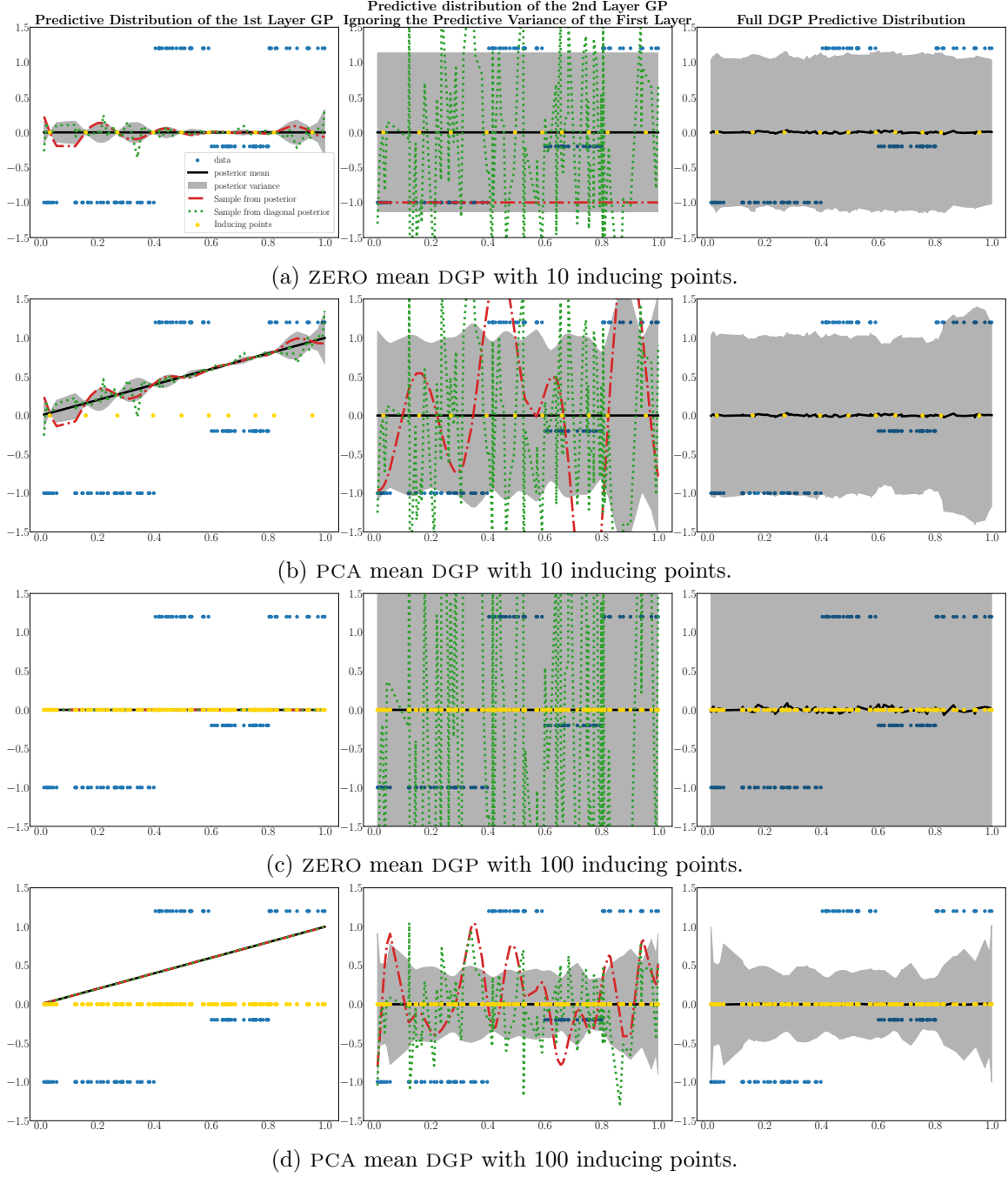

    \centering
    \begin{subfigure}{\textwidth}
        \centering
        \includegraphics[width=\textwidth]{imgs_jmlr/4-initialization_issues/S_out_1_S_inner_e5_more_inducing/marginal_var_pos_at_init_model_zero_S_inner_1e-05_S_out_1_num_Z_10_ls_0.1_whiten_False.pdf}
        \caption{\ZERO mean \DGP{} with $10$ inducing points.}
    \end{subfigure}
    \begin{subfigure}{\textwidth}
        \centering
        \includegraphics[width=\textwidth]{imgs_jmlr/4-initialization_issues/S_out_1_S_inner_e5_more_inducing/marginal_var_pos_at_init_model_PCA_S_inner_1e-05_S_out_1_num_Z_10_ls_0.1_whiten_False.pdf}
        \caption{\PCA mean \DGP{} with $10$ inducing points.}
    \end{subfigure}
    \begin{subfigure}{\textwidth}
        \centering
        \includegraphics[width=\textwidth]{imgs_jmlr/4-initialization_issues/S_out_1_S_inner_e5_more_inducing/marginal_var_pos_at_init_model_zero_S_inner_1e-05_S_out_1_num_Z_100_ls_0.1_whiten_False.pdf}
        \caption{\ZERO mean \DGP{} with $100$ inducing points.}
    \end{subfigure}
    \begin{subfigure}{\textwidth}
        \centering
        \includegraphics[width=\textwidth]{imgs_jmlr/4-initialization_issues/S_out_1_S_inner_e5_more_inducing/marginal_var_pos_at_init_model_PCA_S_inner_1e-05_S_out_1_num_Z_100_ls_0.1_whiten_False.pdf}
        \caption{\PCA mean \DGP{} with $100$ inducing points.}
    \end{subfigure}
    \caption{\ZERO and \PCA mean \DGP{}s with $10$ and $100$ inducing points using the \GPFLOW non-whitened parameterization, with $\varS{}{}=\matI$ in the output layer and $\varS{}{}=10^{-5}\matI$ in the inner layers.}
    \label{fig:varying_inducing_points_Sinner1e5_Sout1}
\end{figure}

In this section, we provide figures that complete the analysis of section \ref{sec:optimization:difficulties}. \fig~\ref{fig:DSVI_zero_pca_pos_toy_at_init_Sout_1.30.8_Sinner1e-5} complements the analysis of the initial predictive distribution according to a different initialization of $\varSw{}{}$. \fig~\ref{fig:varying_inducing_points_Sinner1_Sout1e-5} shows the wiggle effect when $\varS{}{l} = \matI$ and $\varS{}{L} = 10^{-5}\matI$ in \emph{non-whitened} models, for different number of inducing points. In \fig~\ref{fig:varying_inducing_points_Sinner1e5_Sout1}, we swap the values of $\varS{}{}$ with respect to the last figure, to show the wiggle effect when  $\varS{}{l} = 10^{-5}\matI$ and $\varS{}{L} = \matI$ in \emph{non-whitened} models.

\begin{figure}[!htp]
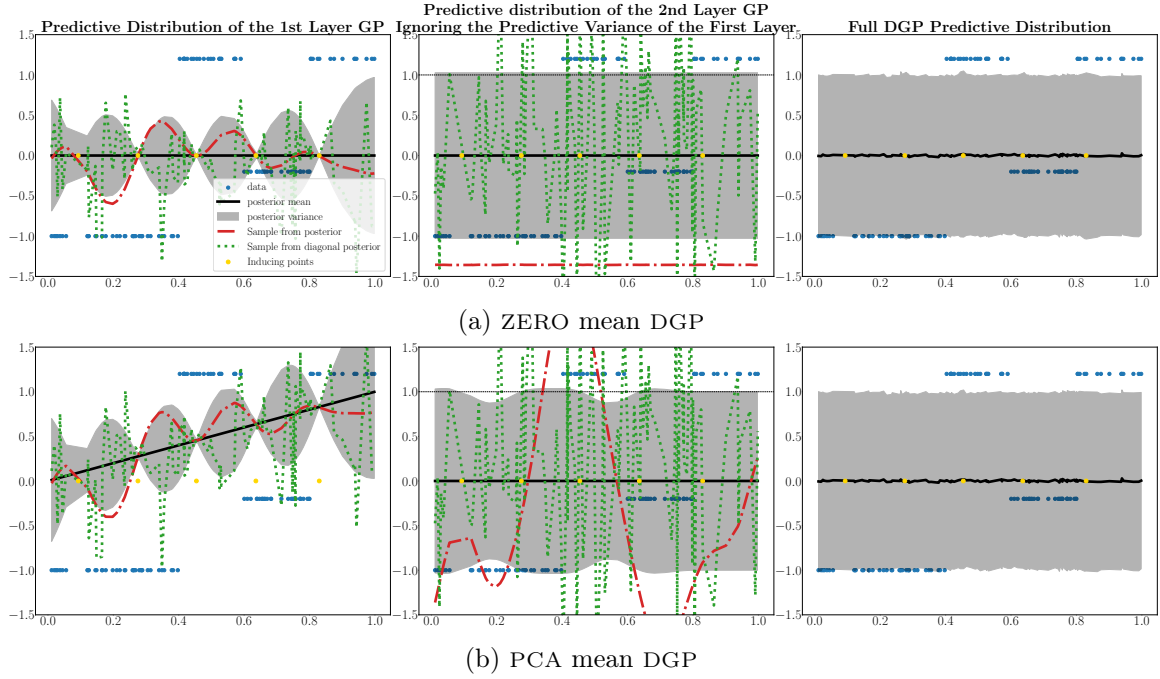

    \centering
    \begin{subfigure}{\textwidth}
        \centering
        \includegraphics[width=\textwidth]{imgs_jmlr/4-initialization_issues/S_out_1_S_inner_e5/marginal_var_pos_at_init_model_zero_S_inner_1e-05_S_out_-1_num_Z_5_ls_0.1_whiten_False.pdf}
        \caption{\ZERO mean \DGP{}}
        \label{}
    \end{subfigure}
    \begin{subfigure}{\textwidth}
        \centering
        \includegraphics[width=\textwidth]{imgs_jmlr/4-initialization_issues/S_out_1_S_inner_e5/marginal_var_pos_at_init_model_PCA_S_inner_1e-05_S_out_-1_num_Z_5_ls_0.1_whiten_False.pdf} 
        \caption{\PCA mean \DGP{}}
    \end{subfigure}
    \caption{\ZERO (top) and \PCA (bottom) \DGP models with the output layer variational covariance initialized to break prior information in the \GPFLOW's unwhitened parameterization. A horizontal line at $1$ shows how variance far from the inducing point goes beyond the prior. }
    \label{fig:DSVI_zero_pca_pos_toy_at_init_Sout_-I_Sinner1e-5_unwhit}
\end{figure}

To create \fig \ref{fig:DSVI_zero_pca_pos_toy_at_init_Sout_-I_Sinner1e-5_unwhit}, where predictive variance goes beyond the prior, we have initialized $\varS{}{}$ in a way such that the prior information coming from the kernel is broken. Remember, we saw that predictive variance depends on $\Kzz{}{}-\varS{}{}$. For $\varS{}{}=\matI$ being diagonal, this removes the information contained in the diagonal of $\Kzz{}{}$ but keeps covariances, effectively reducing predictive variance far from the inducing points. To yield $\varS{}{}$ which breaks prior information, we need to initialize $\varS{}{}$ such that i) is positive semi-definite and ii)  $\Kzz{}{}-\varS{}{}$ also breaks the information contained in the off-diagonal. This can be achieved by a diagonal matrix $\matD$ with entries taking the values of $1$ and $-1$ and setting:
\begin{align}
 \varS{}{} = \matD^T\Kzz{}\matD   
\end{align}
If $\matD$ is an orthogonal matrix, then $\matD^T\Kzz{}\matD$ results in a positive semi-definite matrix. By this initialization $\varS{}{}$ takes values in the off-diagonal such that some covariances from $\Kzz{}{}$ are removed, and some are duplicated. This breaks information contained in the prior and inducing point location and effectively exacerbates posterior variance. Thus, one must take care with random initialization of $\varS{}{}$. 

\begin{figure}[!htp]
    \centering
    \begin{subfigure}{\textwidth}
        \centering
        \includegraphics[width=\textwidth]{imgs_jmlr/4-initialization_issues/S_out_e5_S_inner_1_more_inducing/marginal_var_pos_at_init_model_zero_S_inner_1_S_out_1e-05_num_Z_5_ls_0.1_whiten_False.pdf}
        \caption{\ZERO \DGP{} with $5$ inducing points.}
    \end{subfigure}
    \begin{subfigure}{\textwidth}
        \centering
        \includegraphics[width=\textwidth]{imgs_jmlr/4-initialization_issues/S_out_e5_S_inner_1_more_inducing/marginal_var_pos_at_init_model_PCA_S_inner_1_S_out_1e-05_num_Z_5_ls_0.1_whiten_False.pdf}
        \caption{\PCA \DGP{} with $5$ inducing points.}
    \end{subfigure}
    \begin{subfigure}{\textwidth}
        \centering
        \includegraphics[width=\textwidth]{imgs_jmlr/4-initialization_issues/S_out_e5_S_inner_1_more_inducing/marginal_var_pos_at_init_model_zero_S_inner_1_S_out_1e-05_num_Z_20_ls_0.1_whiten_False.pdf}
        \caption{\ZERO \DGP{} with $20$ inducing points.}
    \end{subfigure}
    \begin{subfigure}{\textwidth}
        \centering
        \includegraphics[width=\textwidth]{imgs_jmlr/4-initialization_issues/S_out_e5_S_inner_1_more_inducing/marginal_var_pos_at_init_model_PCA_S_inner_1_S_out_1e-05_num_Z_20_ls_0.1_whiten_False.pdf}
        \caption{\PCA \DGP{} with $20$ inducing points.}
    \end{subfigure}
\end{figure}
\begin{figure}\ContinuedFloat
    \begin{subfigure}{\textwidth}
        \centering
        \includegraphics[width=\textwidth]{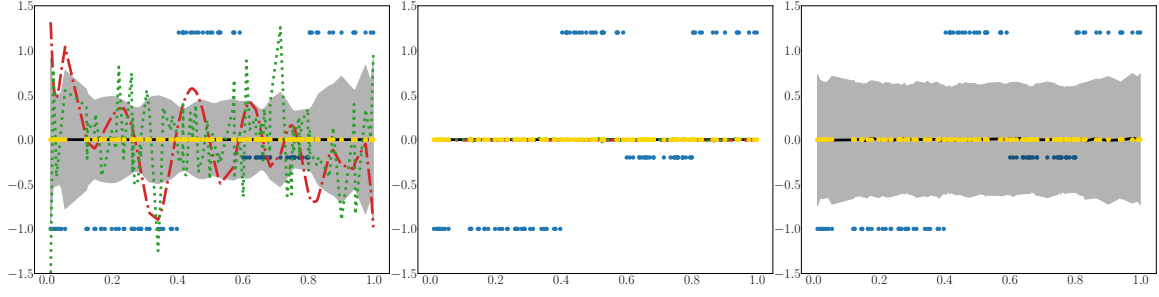}
        \caption{\ZERO \DGP{} with $100$ inducing points.}
    \end{subfigure}
    \begin{subfigure}{\textwidth}
        \centering
        \includegraphics[width=\textwidth]{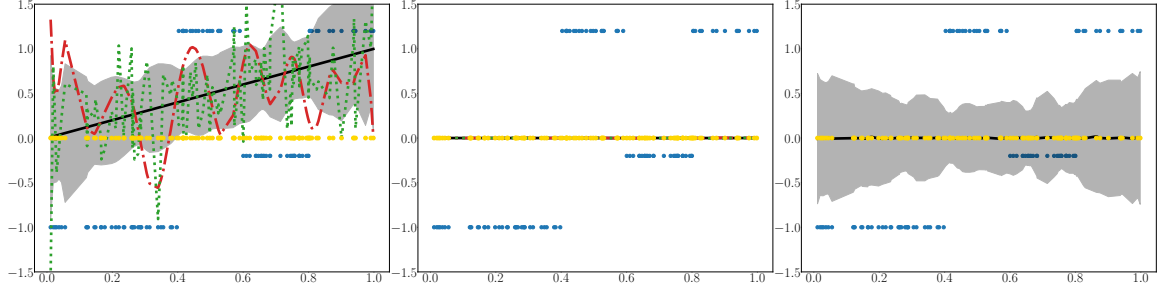}
        \caption{\PCA \DGP{} with $100$ inducing points.}
    \end{subfigure}
    \caption{\ZERO and \PCA mean \DGP{}s with the variational covariance $\varS{}{}=\matI$ for the inner layer and $\varS{}{}=10^{-5}\matI$ in the output layer, varying the number of inducing points in the non-whitened parameterization.}
    \label{fig:varying_inducing_points_Sinner1_Sout1e-5}
\end{figure}

\subsection{Coordinate Updates for 64 Inducing Points}
\label{apx:coordupdates:64}
\begin{figure}[!p]
\centering
\resizebox{0.78\textwidth}{!}{%
\begin{tabular}{cc}
%
\raisebox{0.05\height}{\begin{subfigure}{0.48\linewidth}
    \centering
    \includegraphics[width=\linewidth]{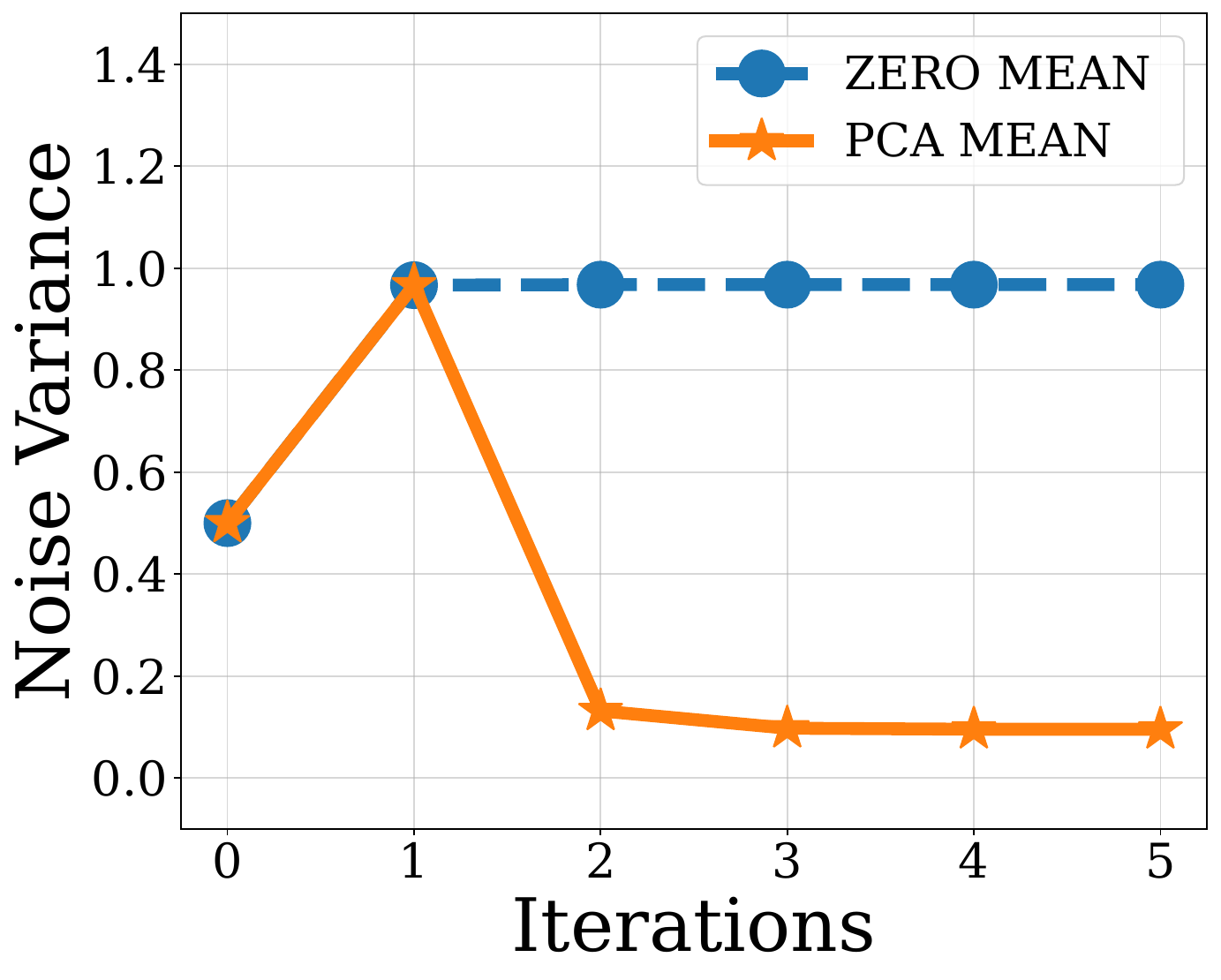}
\end{subfigure}}
&
\begin{subfigure}{0.48\linewidth}
    \centering
    \begin{subfigure}{\linewidth}
        \centering
        \includegraphics[width=\linewidth]{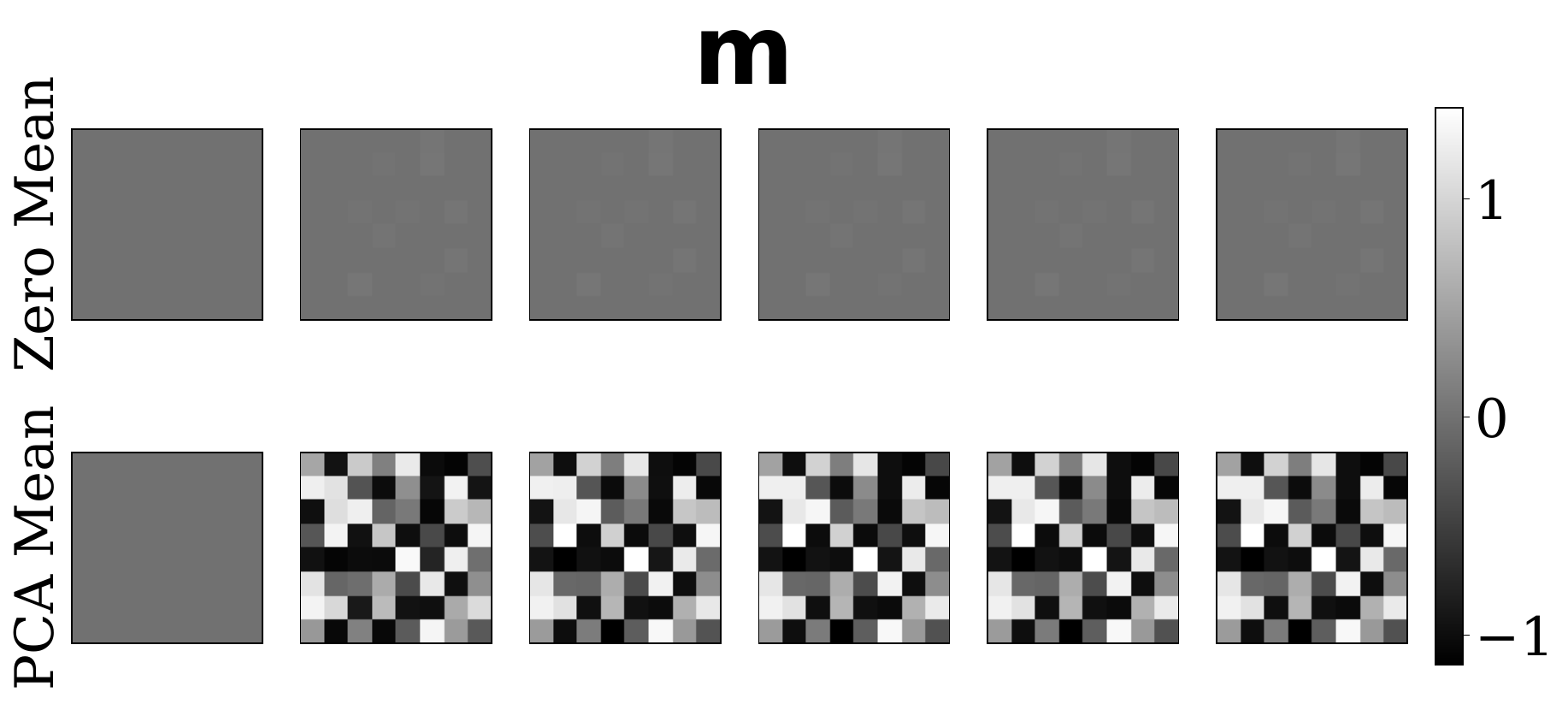}
\end{subfigure}
\begin{subfigure}{\linewidth}
        \centering
        \includegraphics[width=\linewidth]{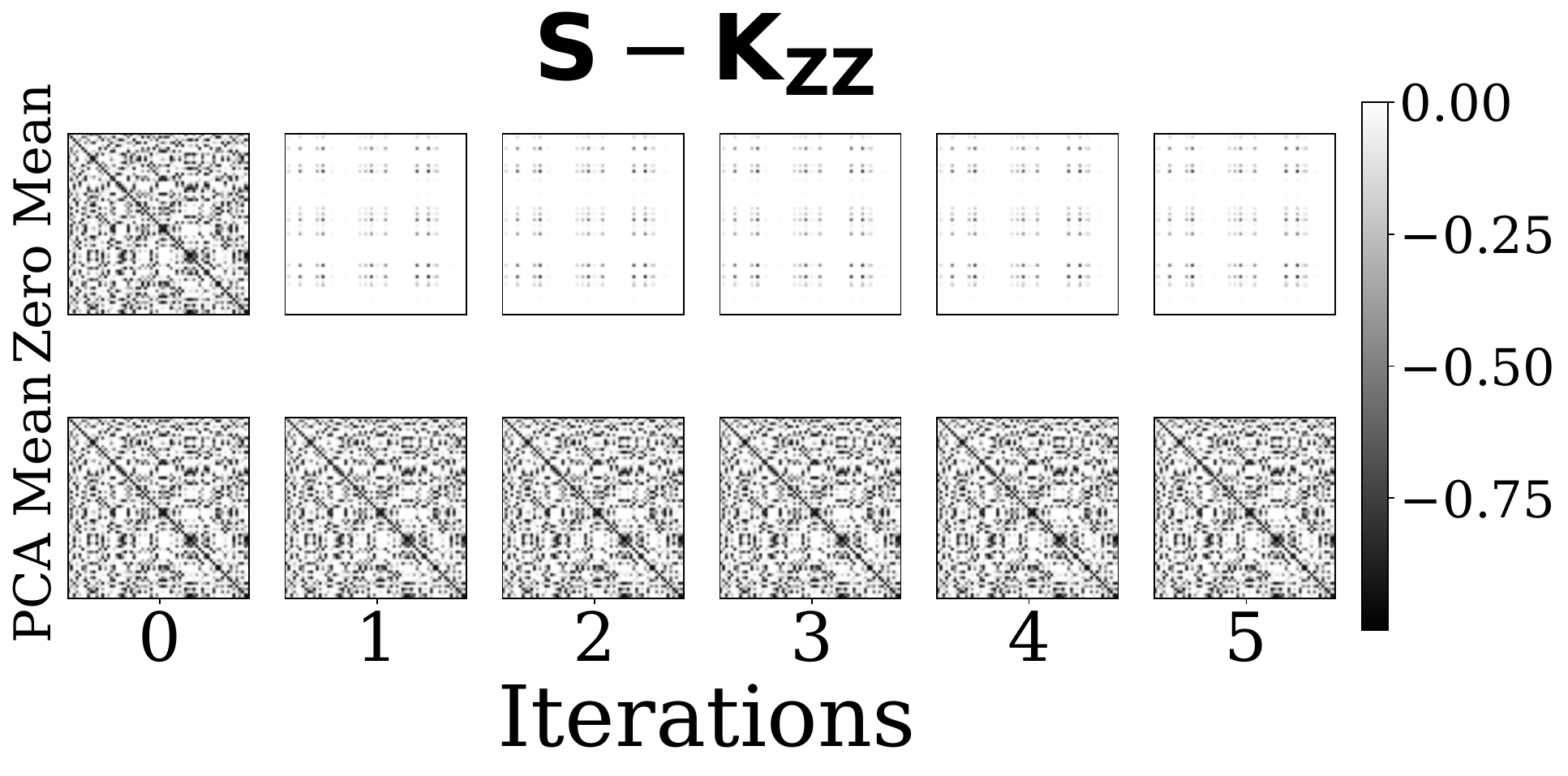}
    \end{subfigure}
\end{subfigure}\\
\multicolumn{2}{c}{
    \begin{subfigure}{\linewidth}
        \centering
        \caption{Order of updates: $\sigma^2$-$\varS{}{}$-$\varm{}{}$, with initialization $\varS{}{}=10^{-5}\matI$, $\varm{}{}=\veczero$}
    \end{subfigure}
} \\
\raisebox{0.05\height}{\begin{subfigure}{0.48\linewidth}
    \centering
    \includegraphics[width=\linewidth]{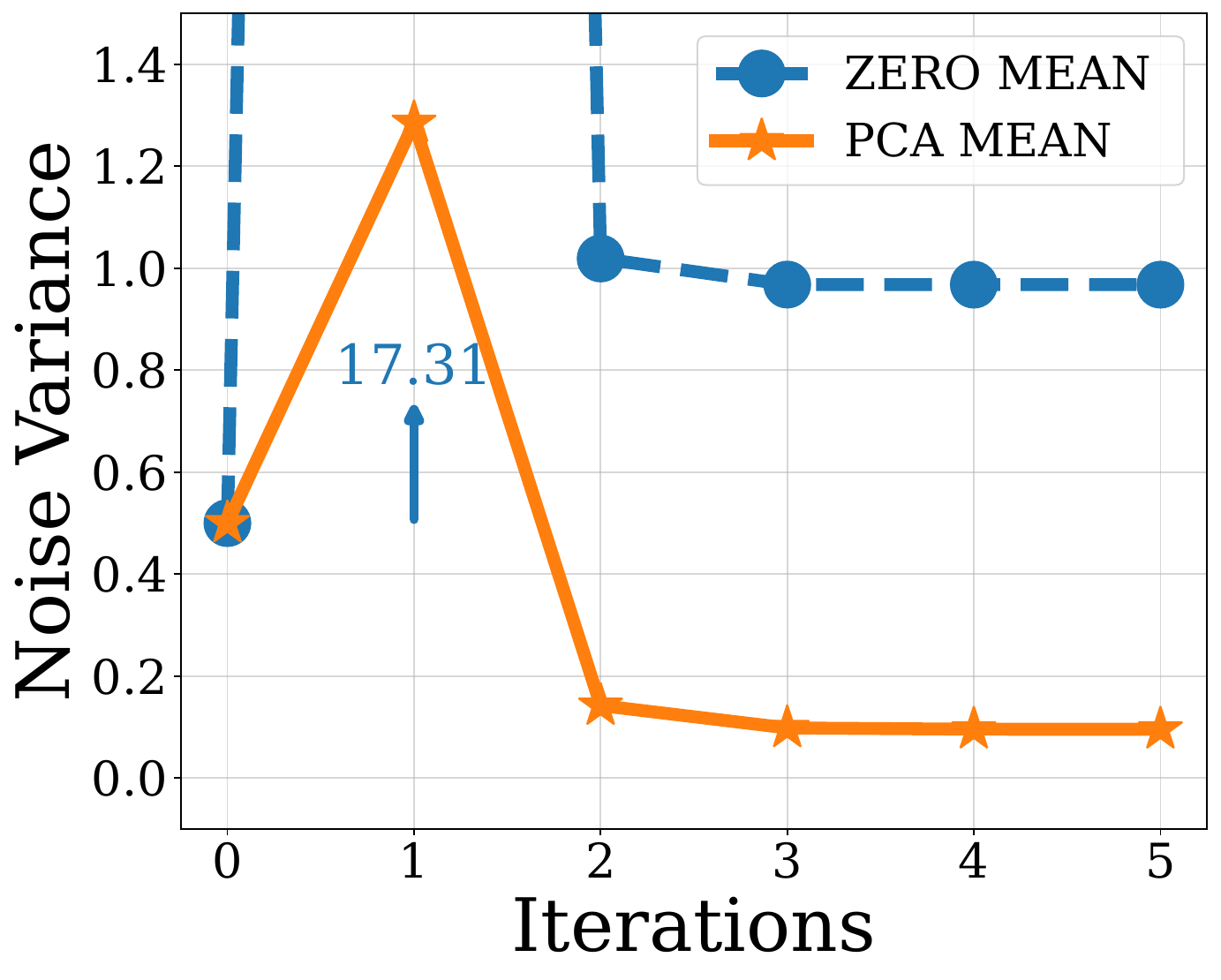}
\end{subfigure}}
&
\begin{subfigure}{0.48\linewidth}
    \centering
    \begin{subfigure}{\linewidth}
        \centering
        \includegraphics[width=\linewidth]{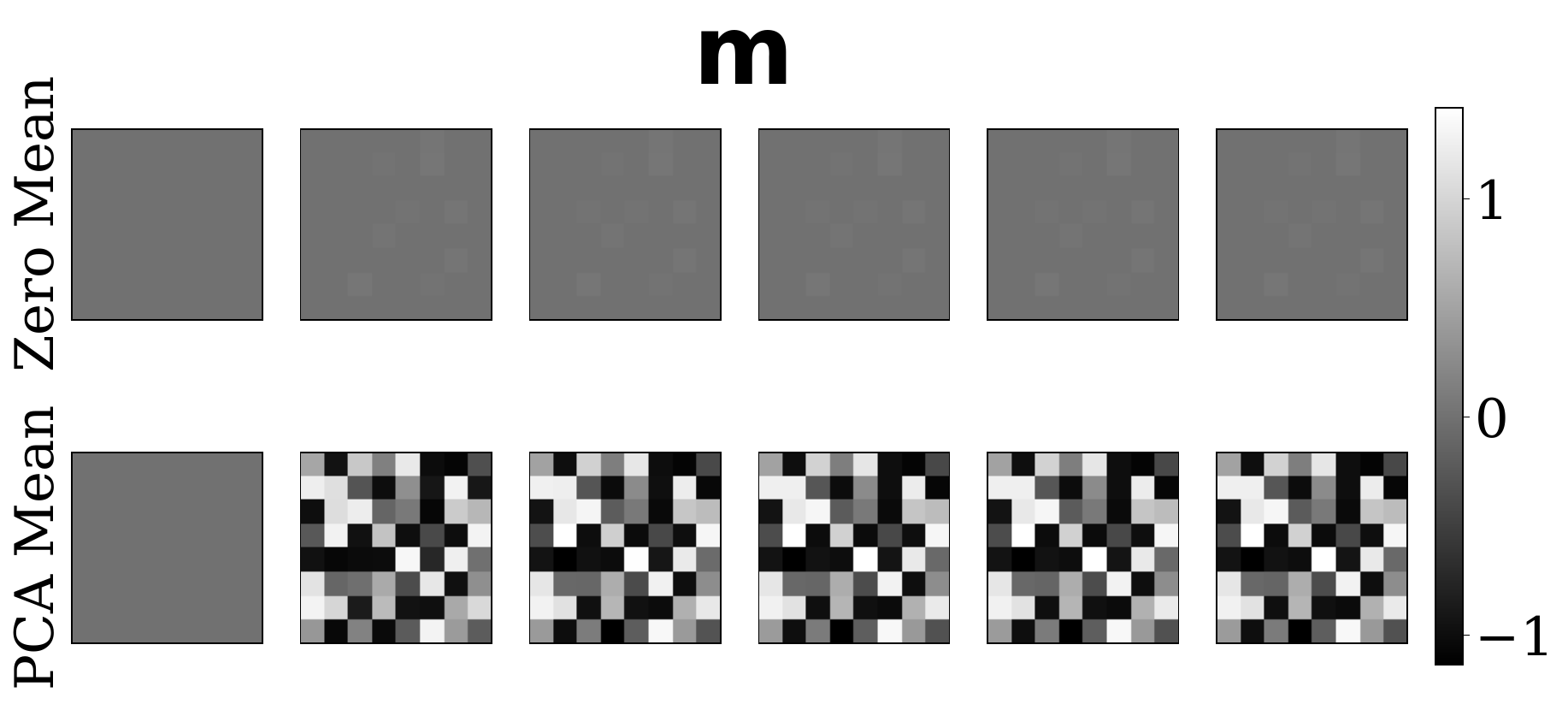}
    \end{subfigure}
    \begin{subfigure}{\linewidth}
        \centering
        \includegraphics[width=\linewidth]{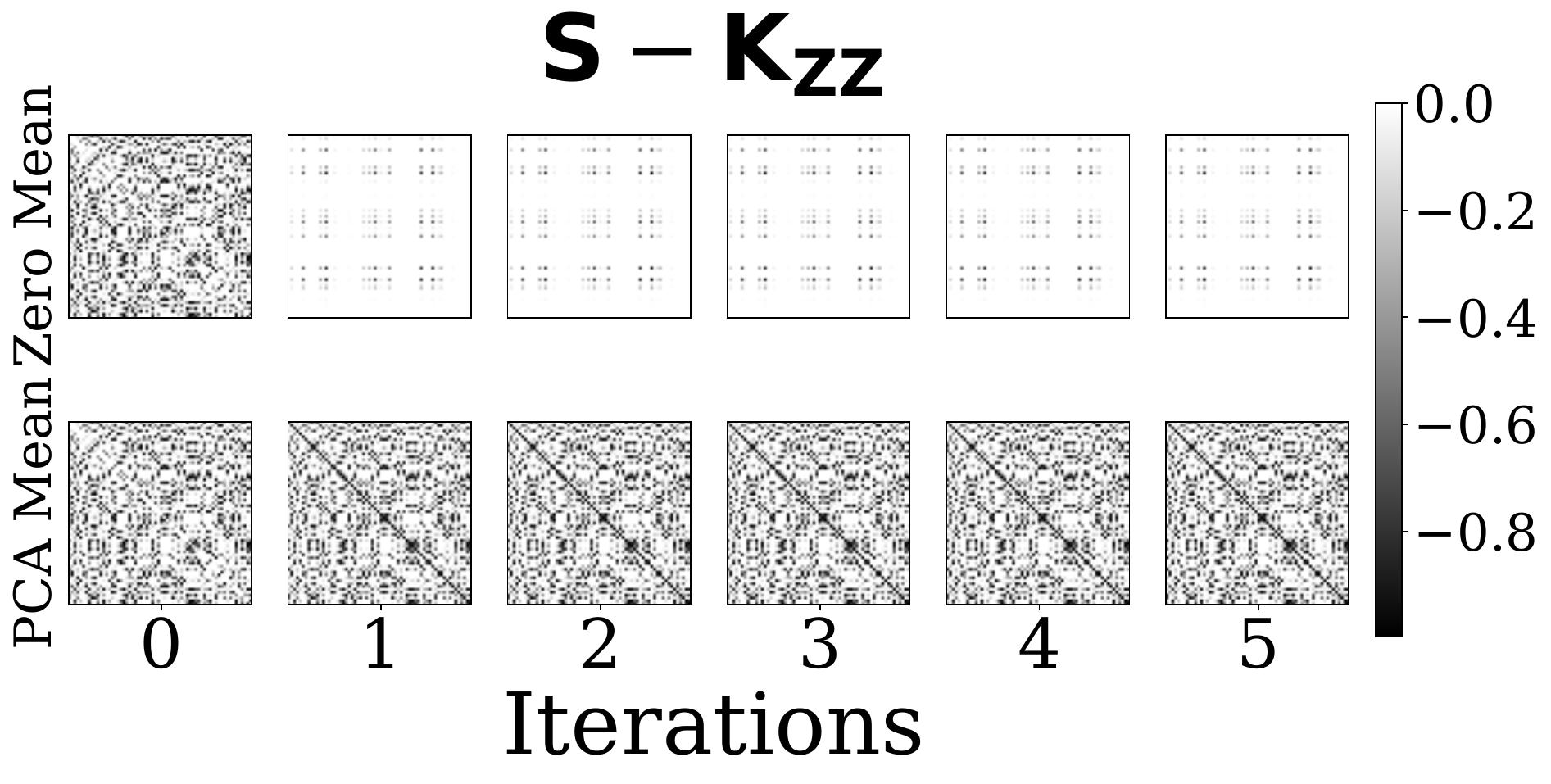}
    \end{subfigure}
\end{subfigure}
\\
\multicolumn{2}{c}{
    \begin{subfigure}{\linewidth}
        \centering
        \caption{Order of updates: $\sigma^2$-$\varS{}{}$-$\varm{}{}$, with initialization $\varS{}{}=\matI$, $\varm{}{}=\veczero$}
    \end{subfigure}
} \\
\raisebox{0.05\height}{\begin{subfigure}{0.48\linewidth}
    \centering
    \includegraphics[width = \linewidth]{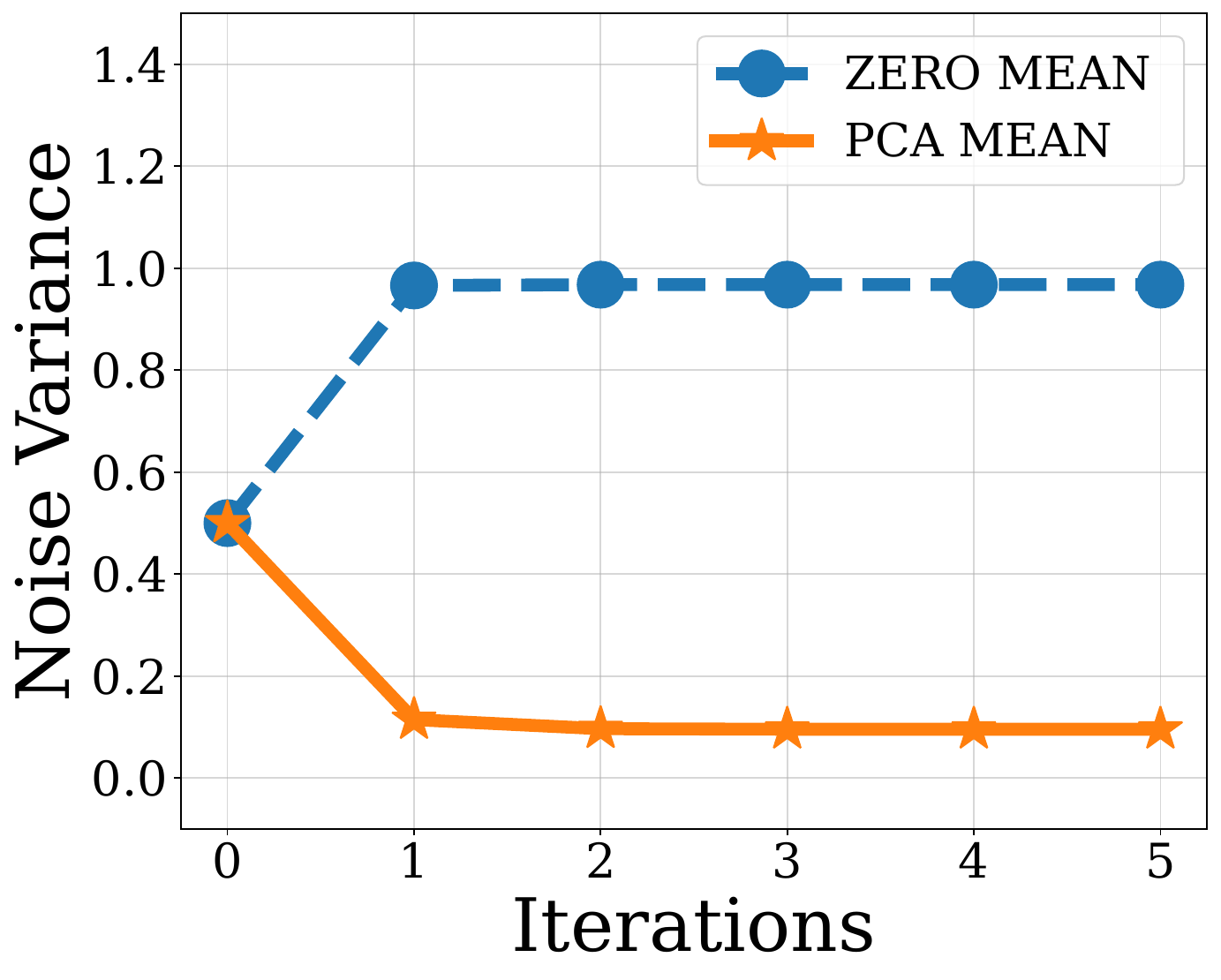}
\end{subfigure}}
&
\begin{subfigure}{0.48\linewidth}
    \centering
    \begin{subfigure}{\linewidth}
        \centering
        \includegraphics[width=\linewidth]{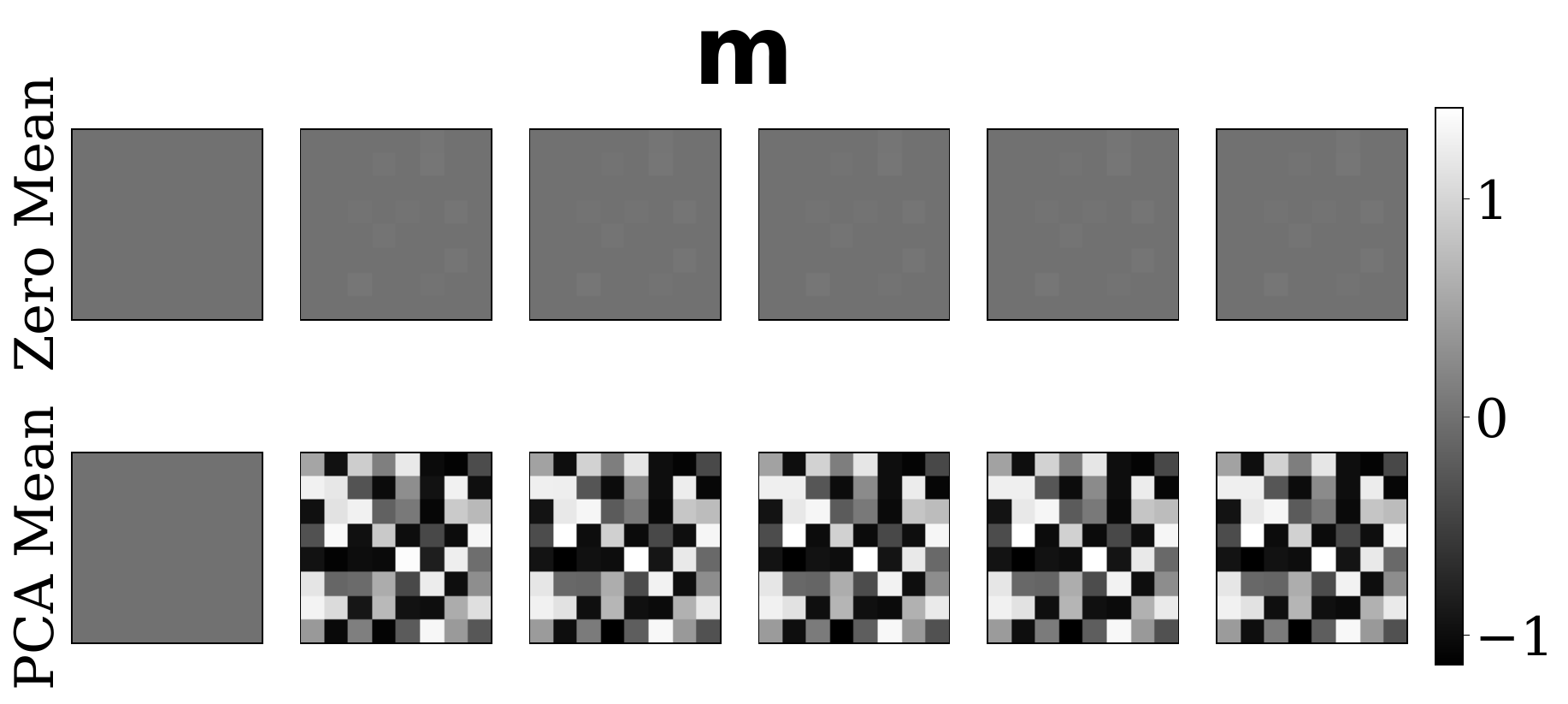}
    \end{subfigure}
    \begin{subfigure}{\linewidth}
        \centering
        \includegraphics[width=\linewidth]{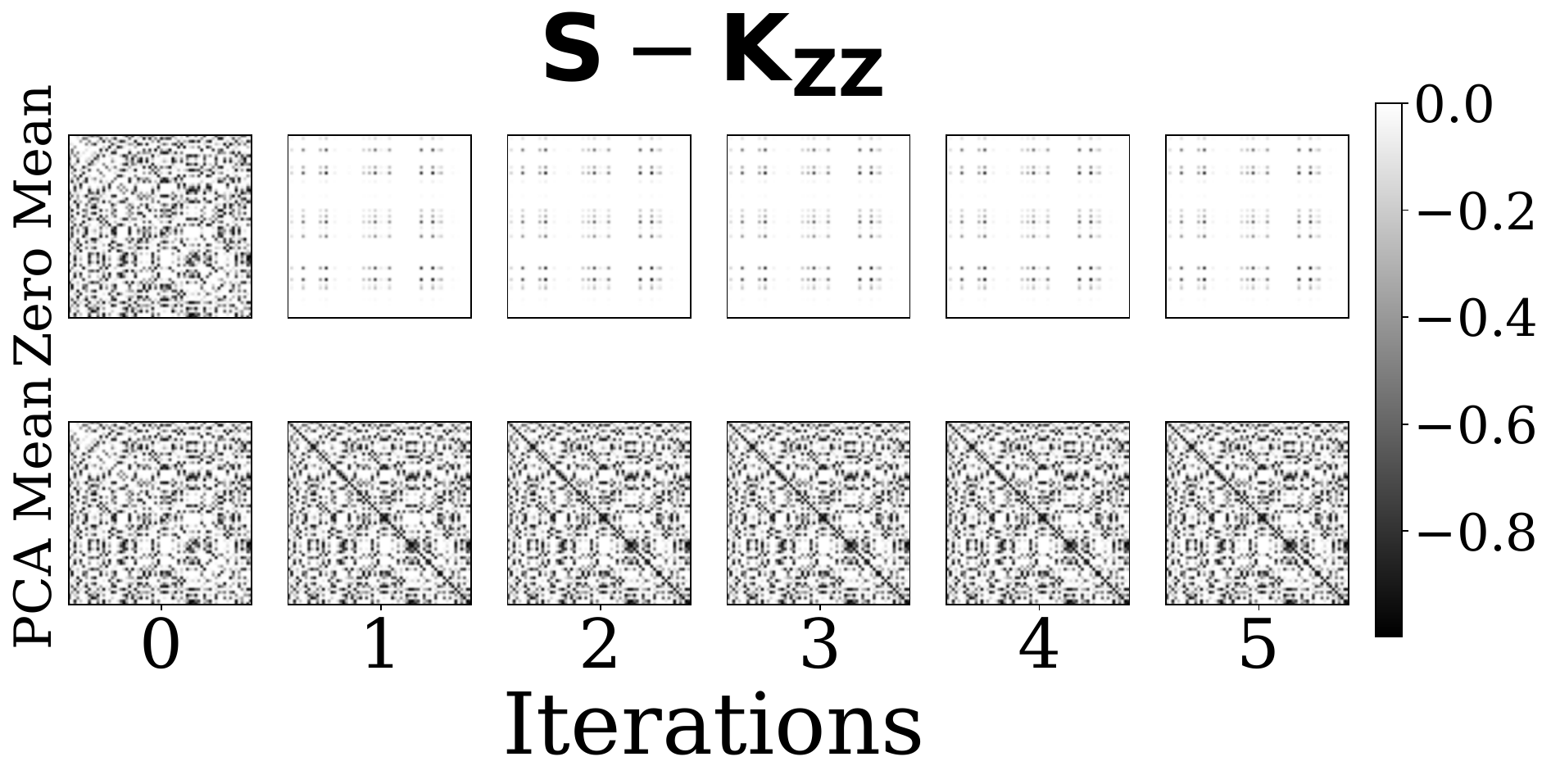}
    \end{subfigure}
\end{subfigure}\\
\multicolumn{2}{c}{
    \begin{subfigure}{\linewidth}
        \centering
        \caption{Order of updates: $\varS{}{}$-$\varm{}{}$-$\sigma^2$. Initialization does not matter since coordinate updates on $\varS{}{}$ does not depend on previous variational parameter values.}
    \end{subfigure}
}
\end{tabular}}

\caption{Coordinate updates experiment with $64$ inducing points.}
\label{fig:coordinate_simulation_Z64}
\end{figure}

\fig \ref{fig:coordinate_simulation_Z64}  shows coordinate updates for $64$ inducing points. In general, we observe the same behavior as with the $5$ inducing points. As a difference, we observe that more inducing
points yield lower noise variance as expected, since the uncertainty coming from approximation error is reduced. Particularly interesting is the fact that for $\varS{}{} = \matI$, $64$ inducing points and the order $\sigma^2-\varS{}{}-\varm{}{}$, the variance explodes in the first iteration to a value around $17$. This shows that early in the training, gradient-based optimization can be directed to a zone where posterior collapse is likely.

\clearpage
\subsection{Additional Toy Experiments Illustrating other Parameter Initializations}
\label{sec:app:c:2:toy:optimization}

\begin{figure}[!htbp]
    \centering
    %
    %
    %
    %
    \begin{minipage}[c]{0.05\textwidth}
    \end{minipage}%
    \begin{minipage}[c]{0.93\textwidth}
        \centering
        \hfill
        \makebox[0.32\linewidth][c]{\textbf{2 Layers}}%
        \hfill
        \makebox[0.32\linewidth][c]{\textbf{3 Layers}}%
        \hfill
        \makebox[0.32\linewidth][c]{\textbf{5 Layers}}%
    \end{minipage}

    \begin{minipage}[c]{0.05\textwidth}
        \centering
        \rotatebox{90}{\textbf{5 Inducing}}
    \end{minipage}%
    \begin{minipage}[c]{0.93\textwidth}
        \centering
        \begin{subfigure}[c]{0.32\linewidth}
            \includegraphics[width=\linewidth]{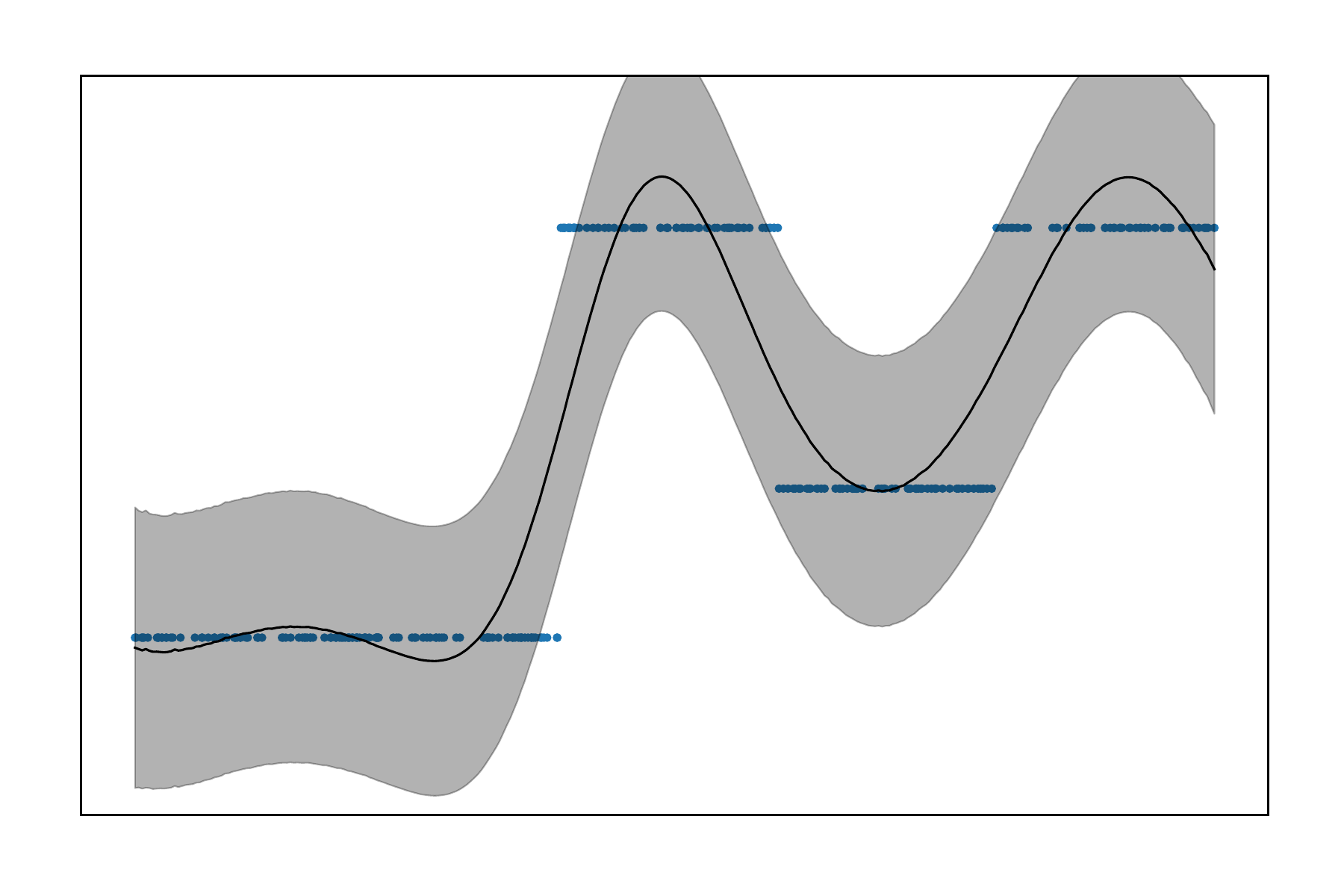}
        \end{subfigure}
        \hfill
        \begin{subfigure}[c]{0.32\linewidth}
            \includegraphics[width=\linewidth]{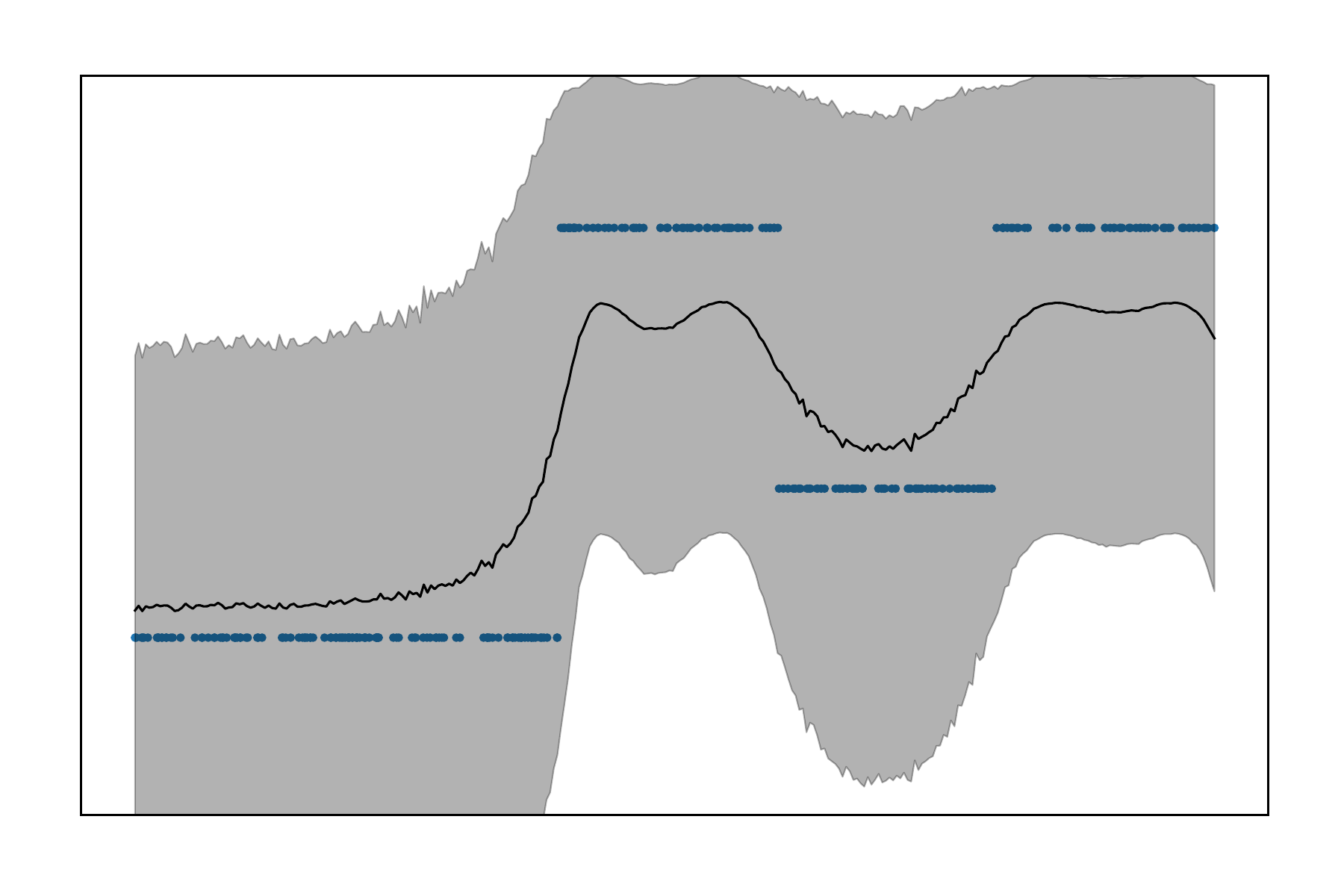}
        \end{subfigure}
        \hfill
        \begin{subfigure}[c]{0.32\linewidth}
            \includegraphics[width=\linewidth]{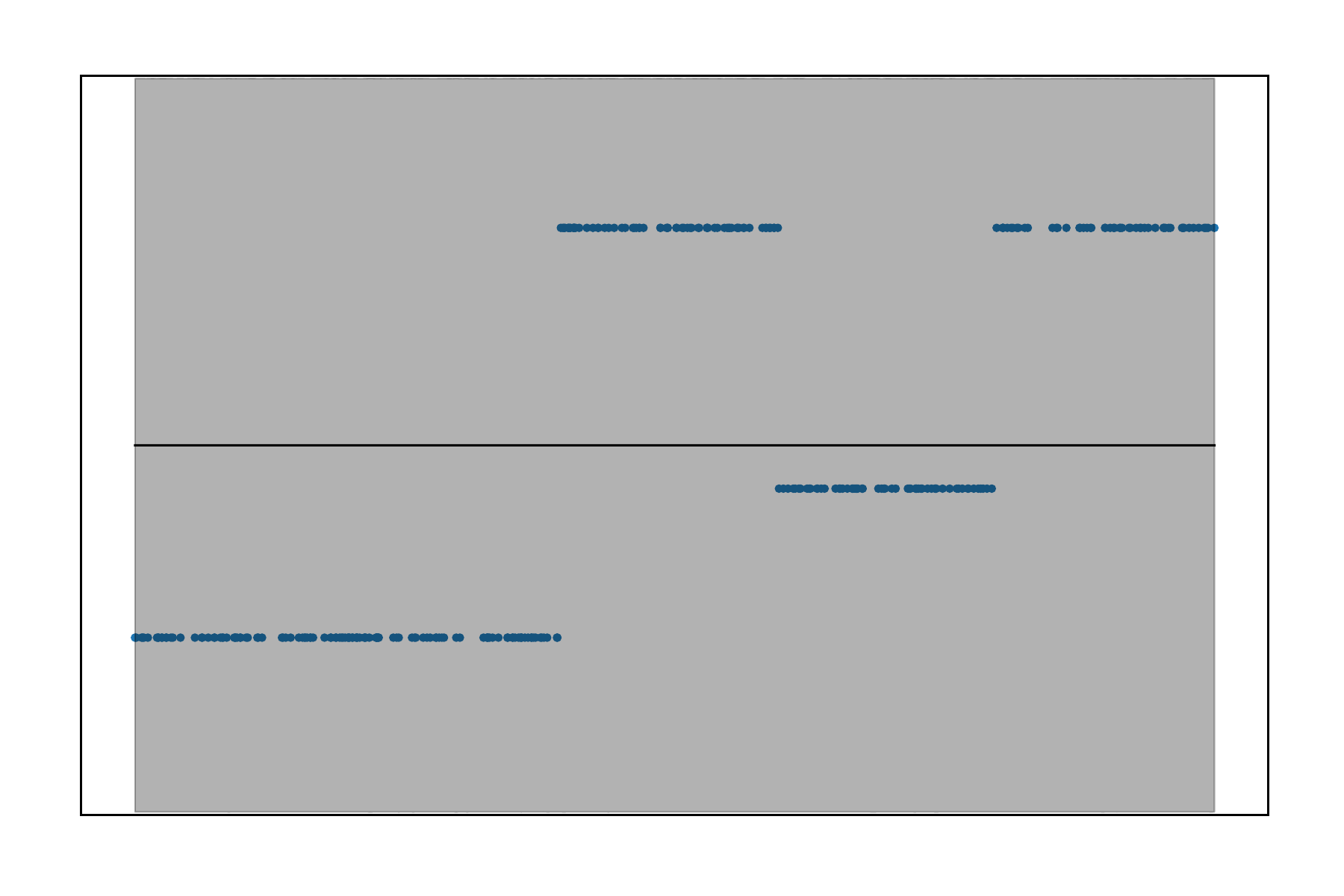}
        \end{subfigure}
    \end{minipage}

    \vspace{0.25cm} 
    
    \begin{minipage}[c]{0.05\textwidth}
        \centering
        \rotatebox{90}{\textbf{20 Inducing}}
    \end{minipage}%
    \begin{minipage}[c]{0.93\textwidth}
        \centering
        \begin{subfigure}[c]{0.32\linewidth}
            \includegraphics[width=\linewidth]{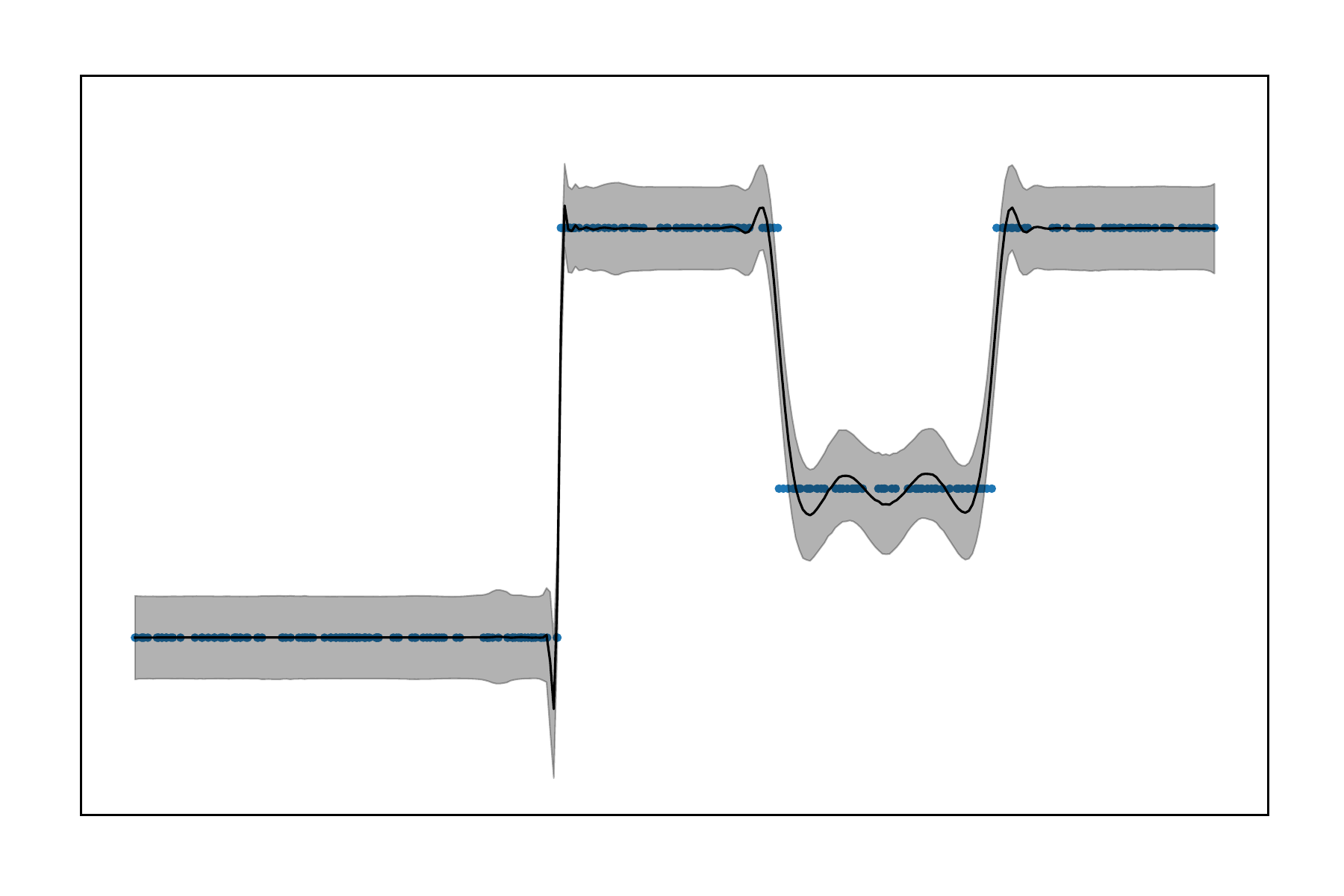}
        \end{subfigure}
        \hfill
        \begin{subfigure}[c]{0.32\linewidth}
            \includegraphics[width=\linewidth]{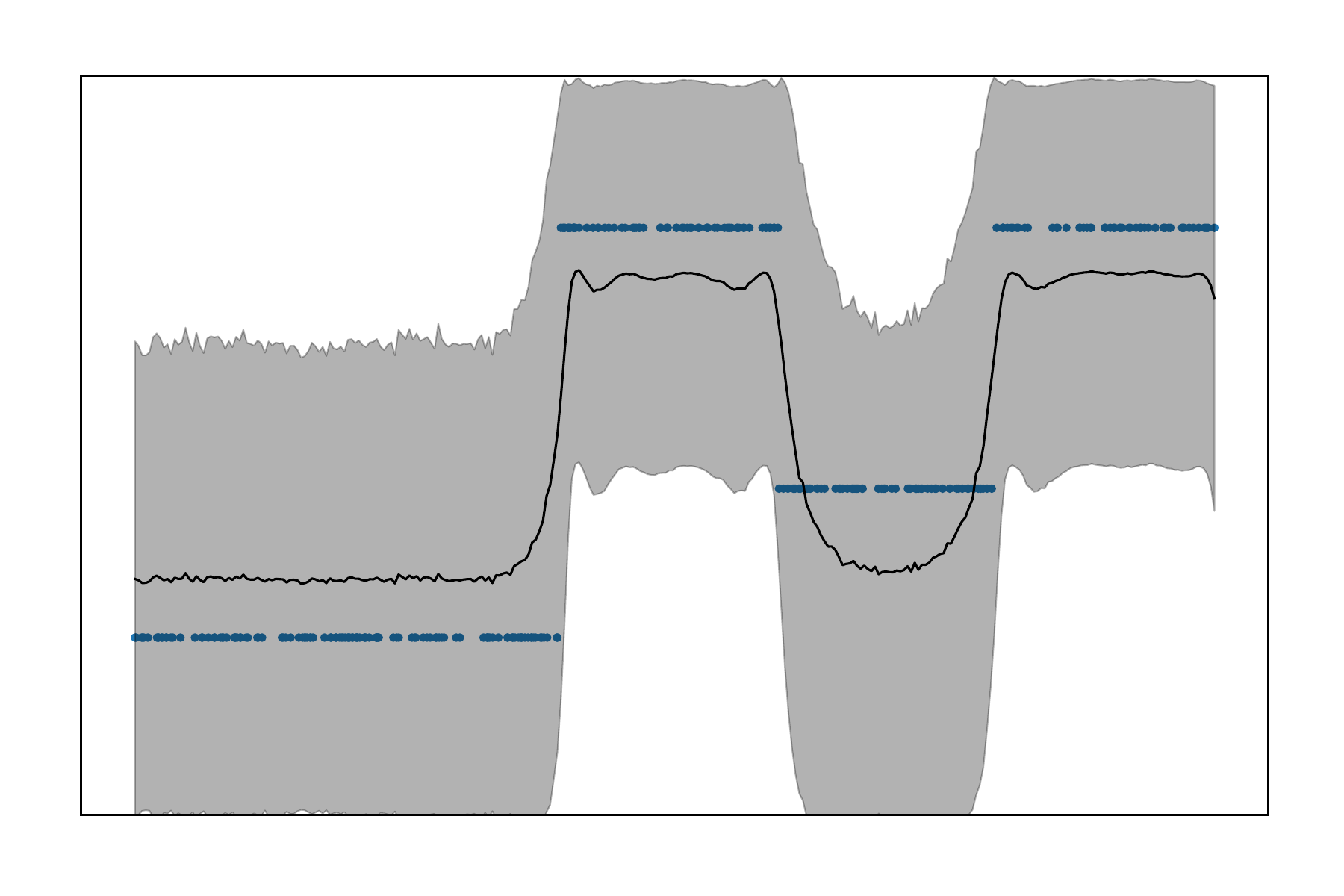}
        \end{subfigure}
        \hfill
        \begin{subfigure}[c]{0.32\linewidth}
            \includegraphics[width=\linewidth]{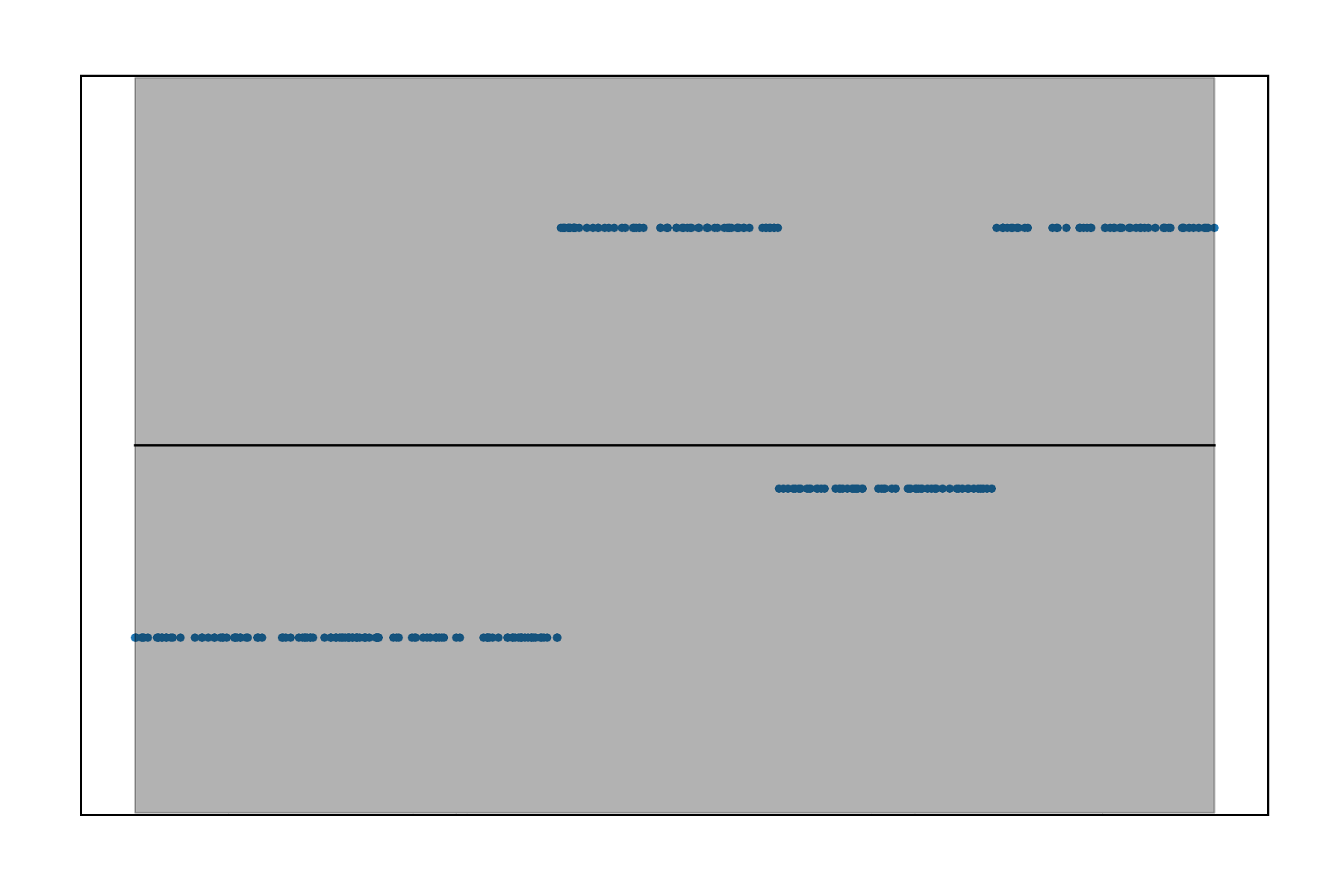}
        \end{subfigure}
    \end{minipage}

    \vspace{0.25cm} 

    \begin{minipage}[c]{0.05\textwidth}
        \centering
        \rotatebox{90}{\textbf{100 Inducing}}
    \end{minipage}%
    \begin{minipage}[c]{0.93\textwidth}
        \centering
        \begin{subfigure}[c]{0.32\linewidth}
            \includegraphics[width=\linewidth]{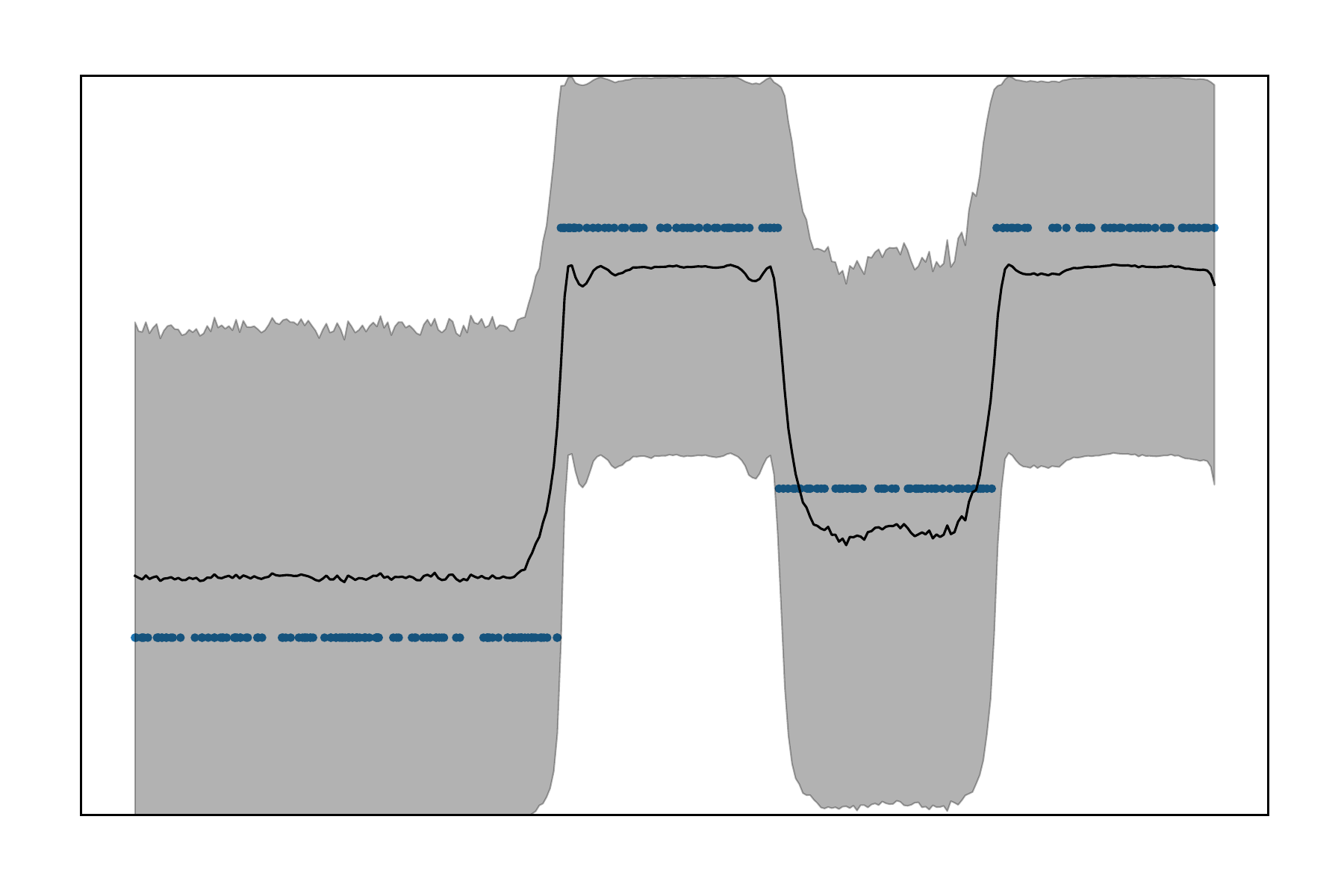}
        \end{subfigure}
        \hfill
        \begin{subfigure}[c]{0.32\linewidth}
            \includegraphics[width=\linewidth]{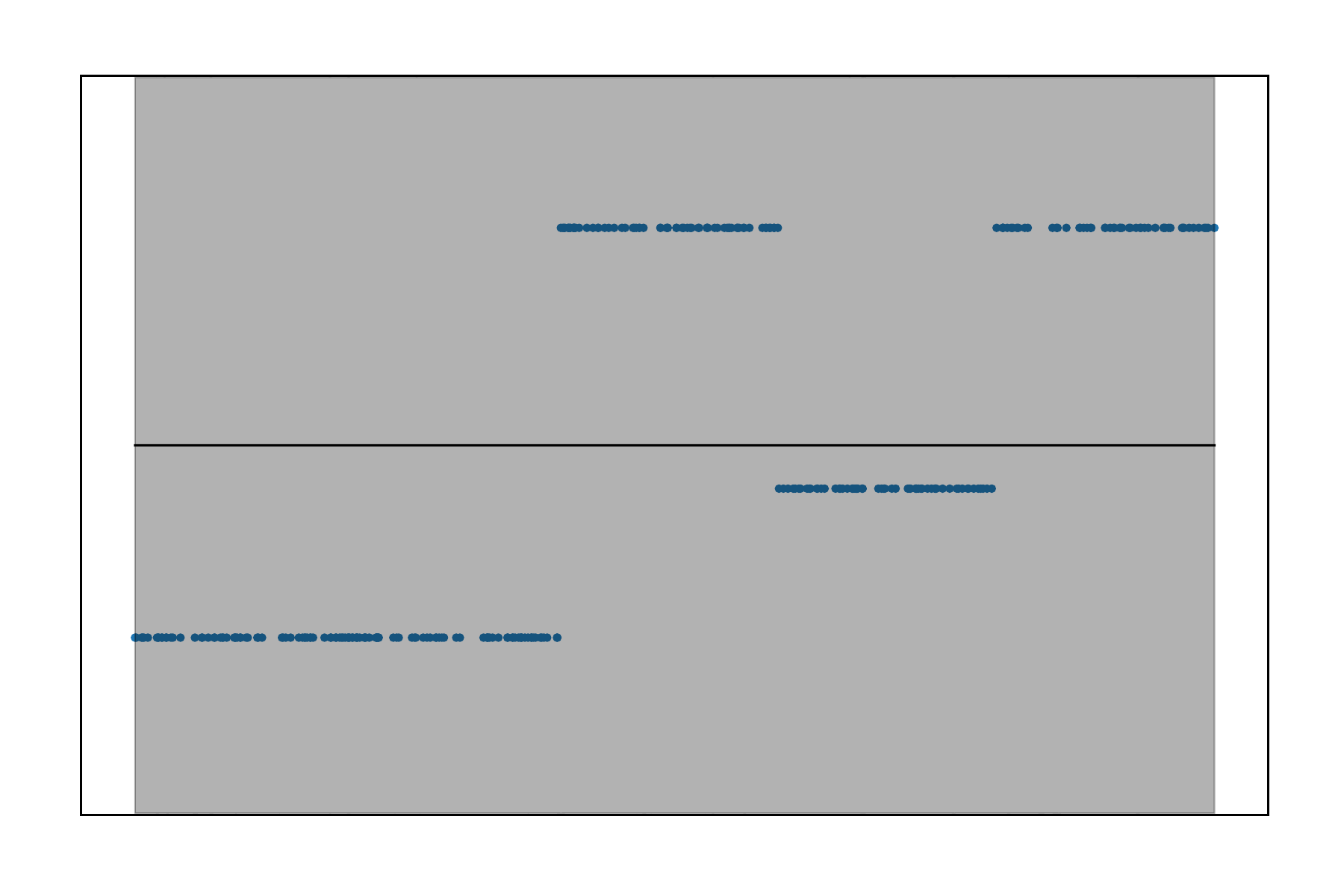}
        \end{subfigure}
        \hfill
        \begin{subfigure}[c]{0.32\linewidth}
            \includegraphics[width=\linewidth]{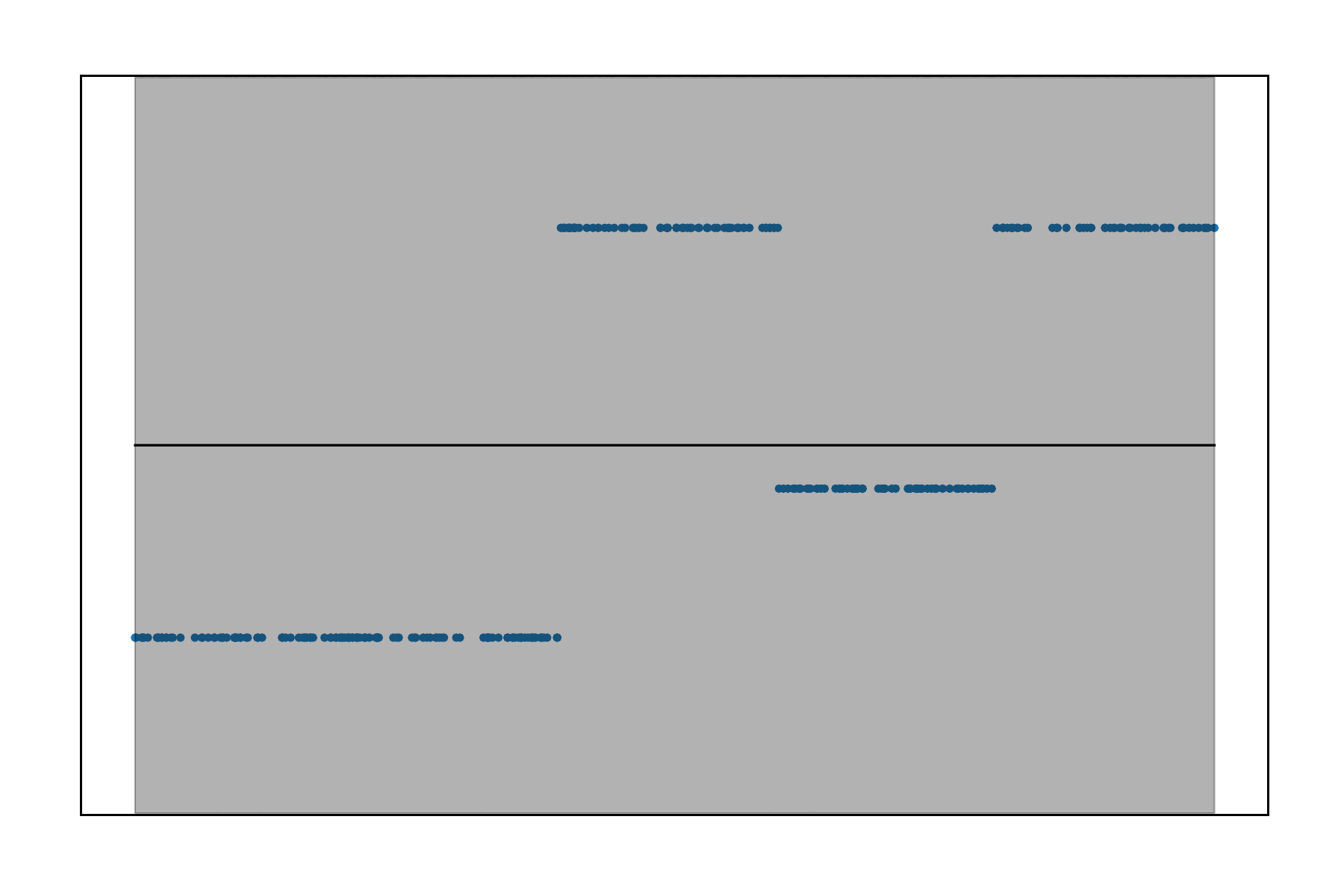}
        \end{subfigure}
    \end{minipage}
    \begin{minipage}[c]{\textwidth}
        \makebox[\linewidth][c]{\textbf{\ZEROW{}}}%
    \end{minipage}

    \begin{minipage}[c]{0.05\textwidth}
        \hfill 
    \end{minipage}%

    \begin{minipage}[c]{0.05\textwidth}
        \centering
        \rotatebox{90}{\textbf{5 Inducing}}
    \end{minipage}%
    \begin{minipage}[c]{0.93\textwidth}
        \centering
        \begin{subfigure}[c]{0.32\linewidth}
            \includegraphics[width=\linewidth]{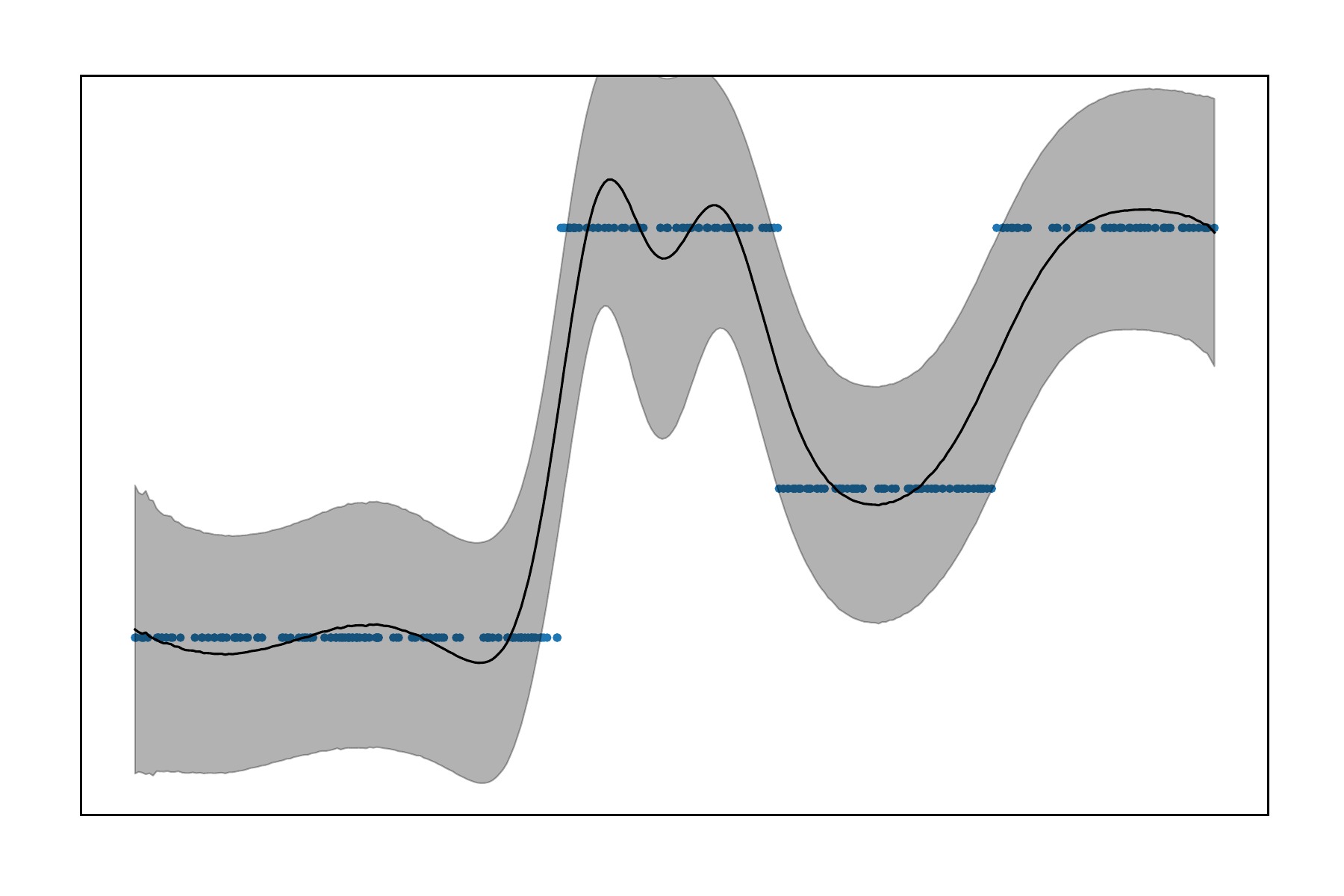}
        \end{subfigure}
        \hfill
        \begin{subfigure}[c]{0.32\linewidth}
            \includegraphics[width=\linewidth]{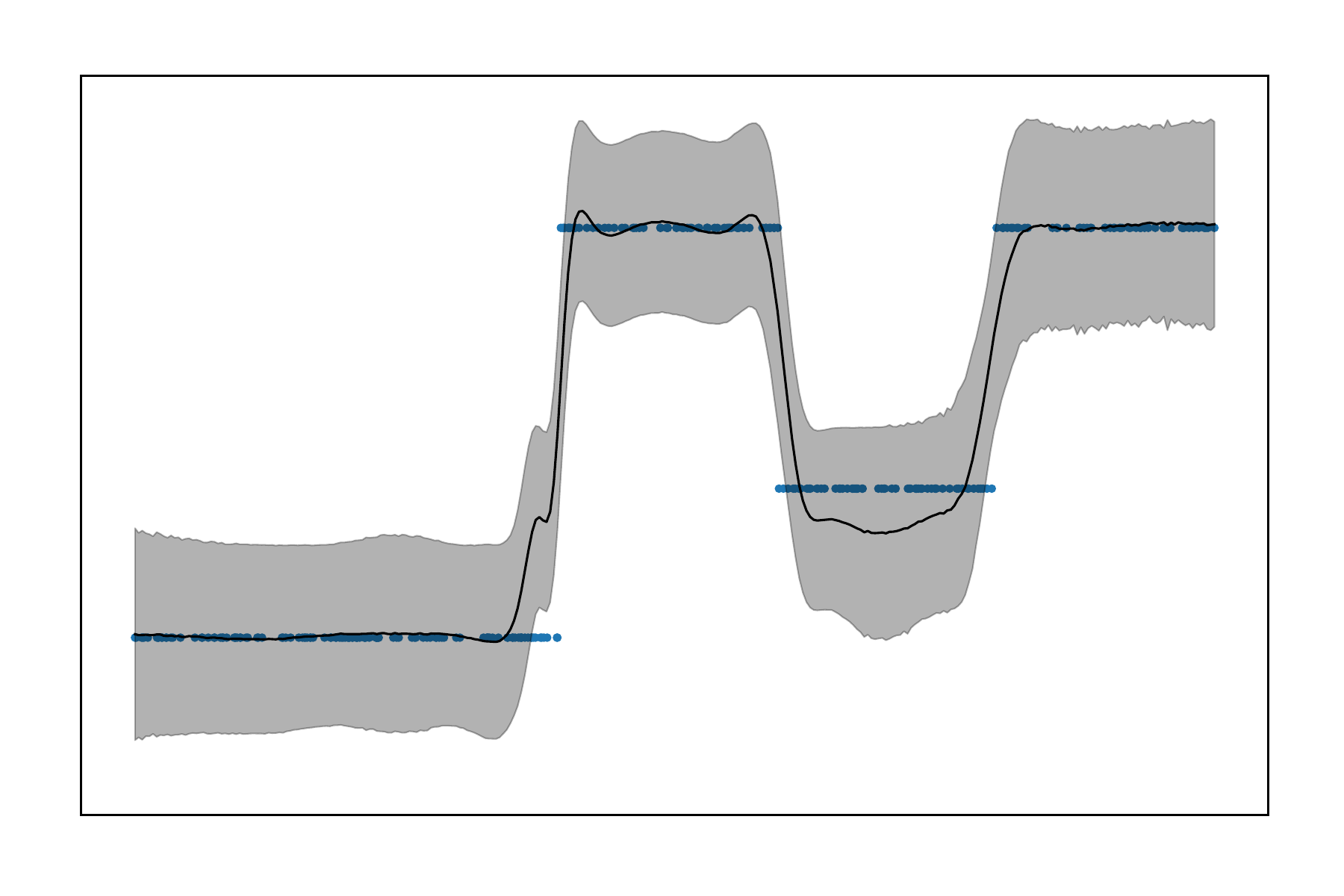}
        \end{subfigure}
        \hfill
        \begin{subfigure}[c]{0.32\linewidth}
            \includegraphics[width=\linewidth]{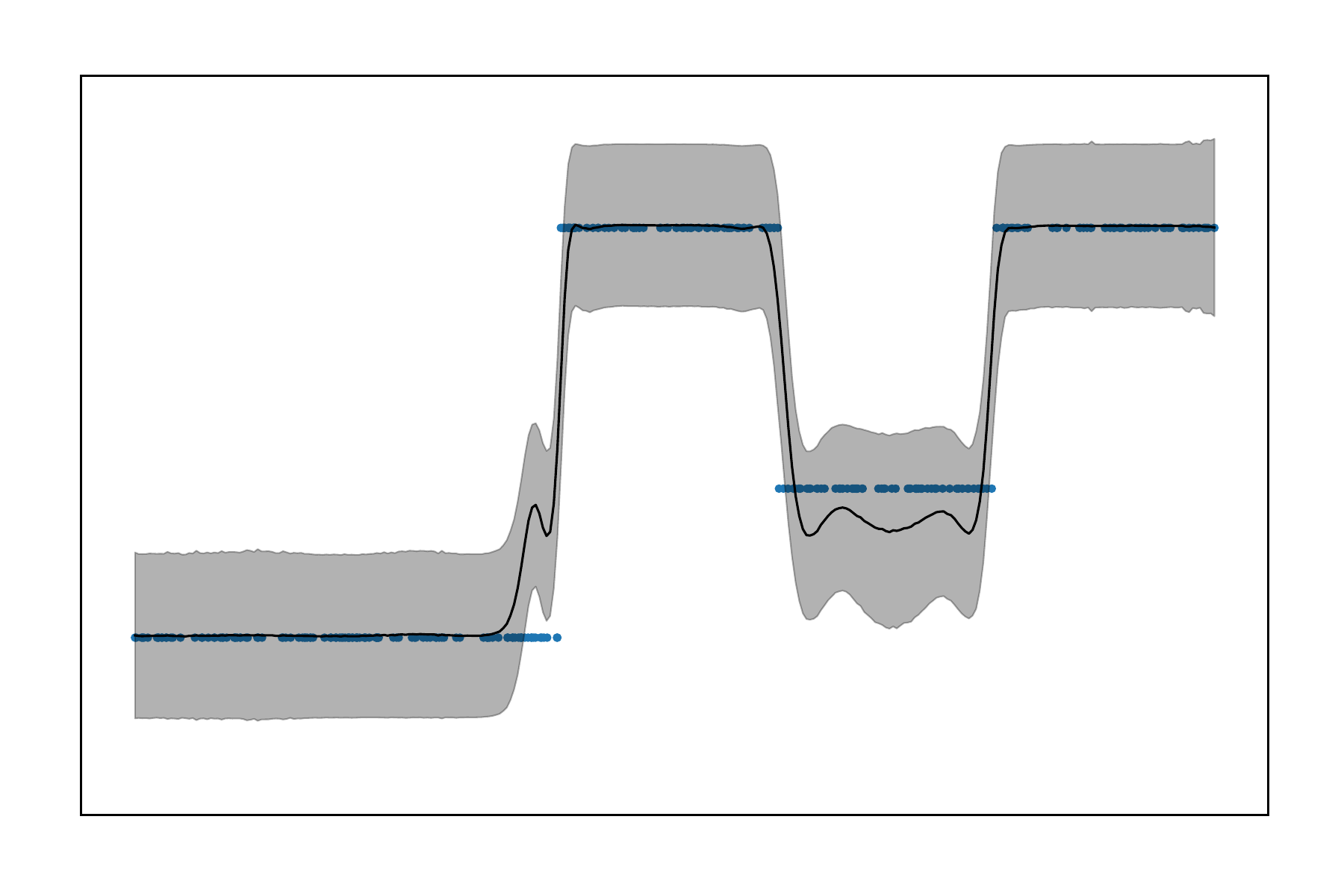}
        \end{subfigure}
    \end{minipage}

    \vspace{0.25cm} 
    
    \begin{minipage}[c]{0.05\textwidth}
        \centering
        \rotatebox{90}{\textbf{20 Inducing}}
    \end{minipage}%
    \begin{minipage}[c]{0.93\textwidth}
        \centering
        \begin{subfigure}[c]{0.32\linewidth}
            \includegraphics[width=\linewidth]{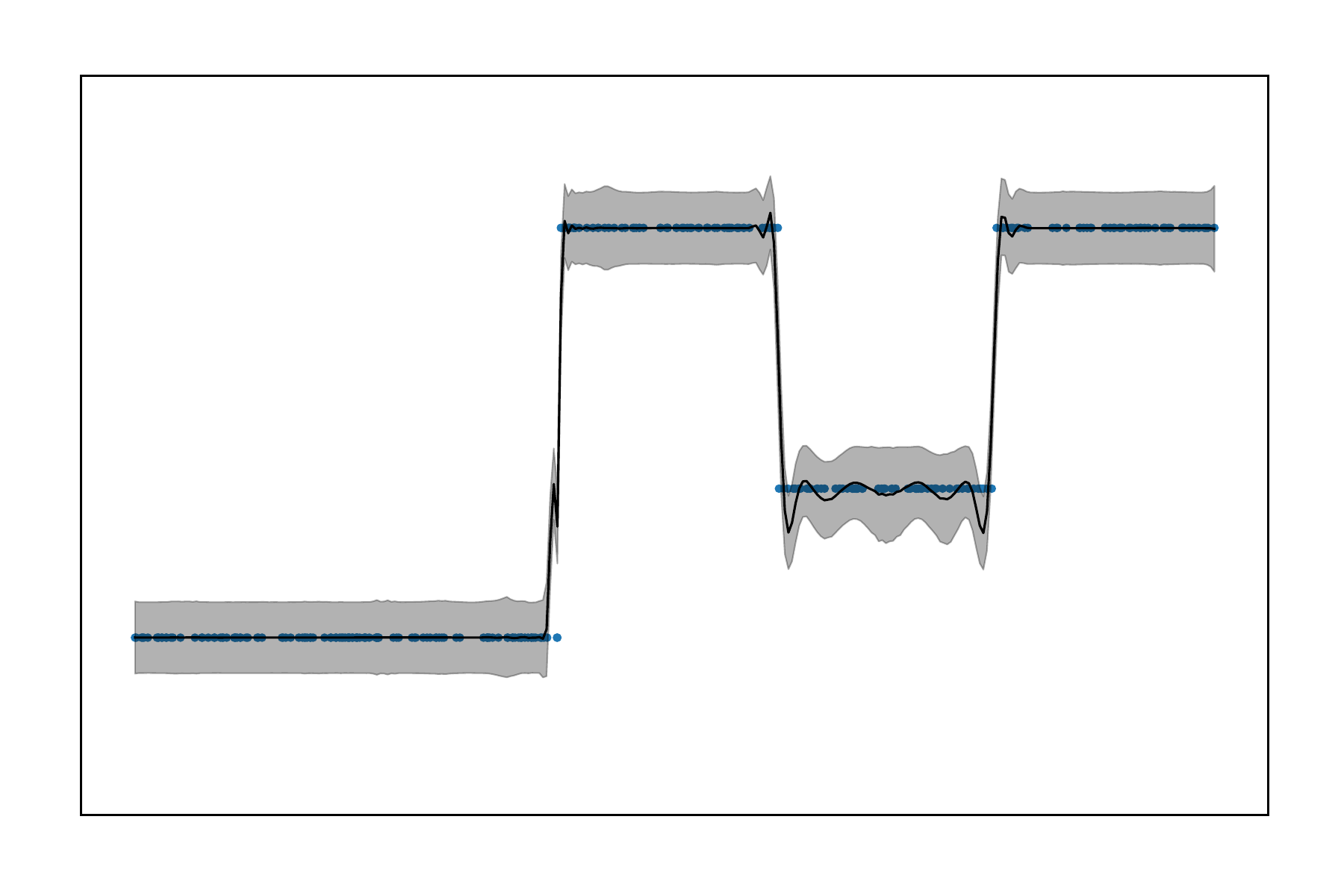}
        \end{subfigure}
        \hfill
        \begin{subfigure}[c]{0.32\linewidth}
            \includegraphics[width=\linewidth]{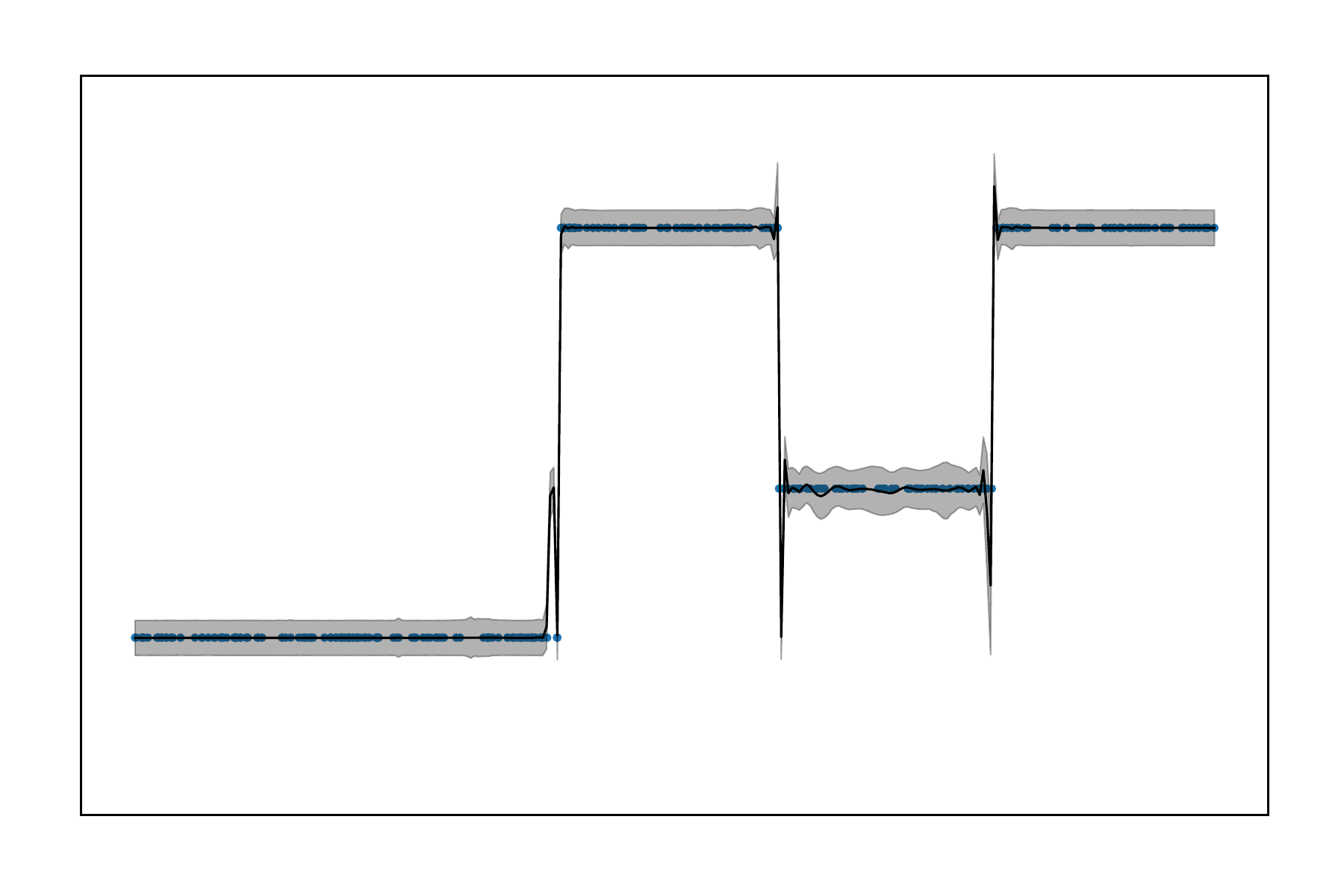}
        \end{subfigure}
        \hfill
        \begin{subfigure}[c]{0.32\linewidth}
            \includegraphics[width=\linewidth]{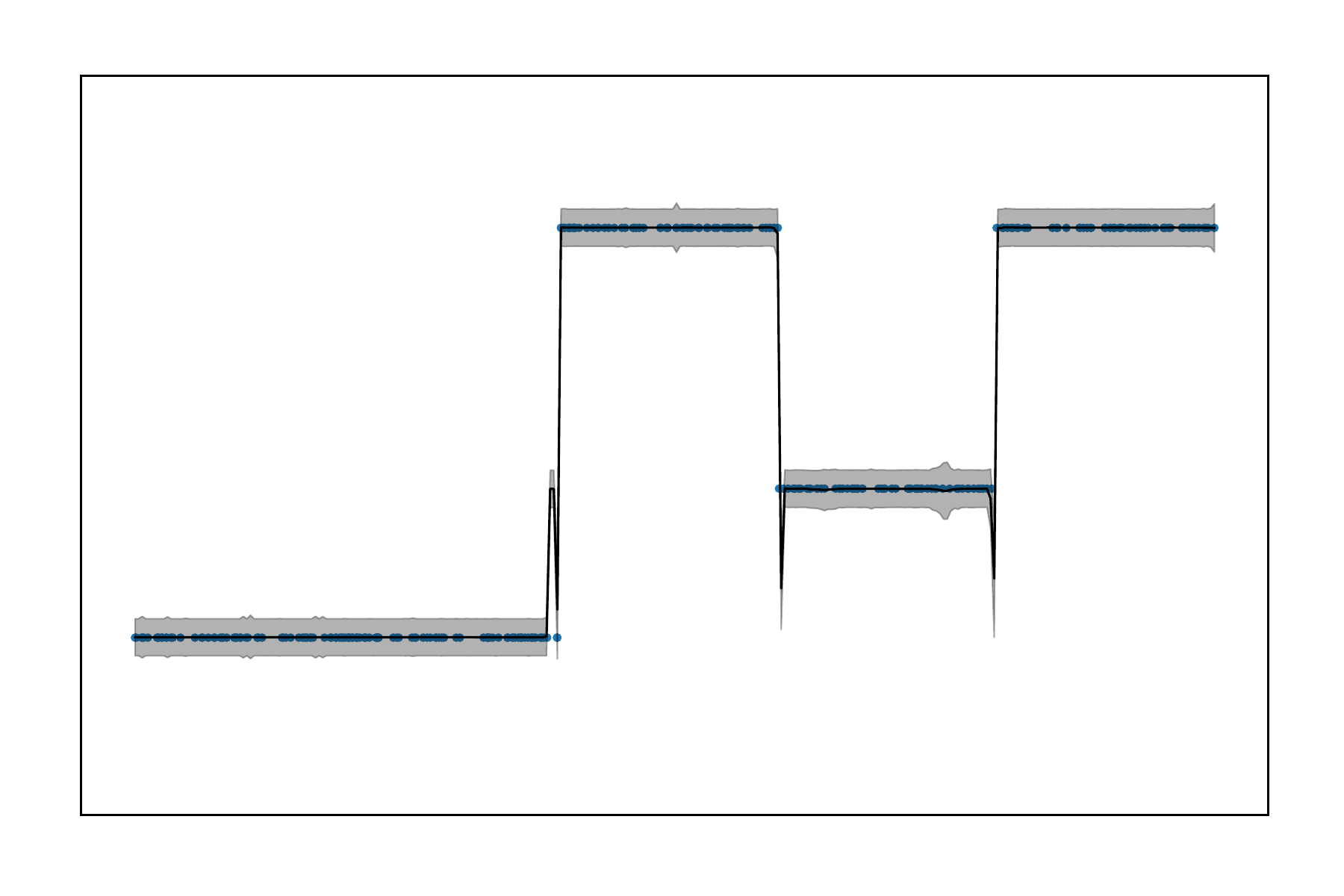}
        \end{subfigure}
    \end{minipage}

    \vspace{0.25cm} 

    \begin{minipage}[c]{0.05\textwidth}
        \centering
        \rotatebox{90}{\textbf{100 Inducing}}
    \end{minipage}%
    \begin{minipage}[c]{0.93\textwidth}
        \centering
        \begin{subfigure}[c]{0.32\linewidth}
            \includegraphics[width=\linewidth]{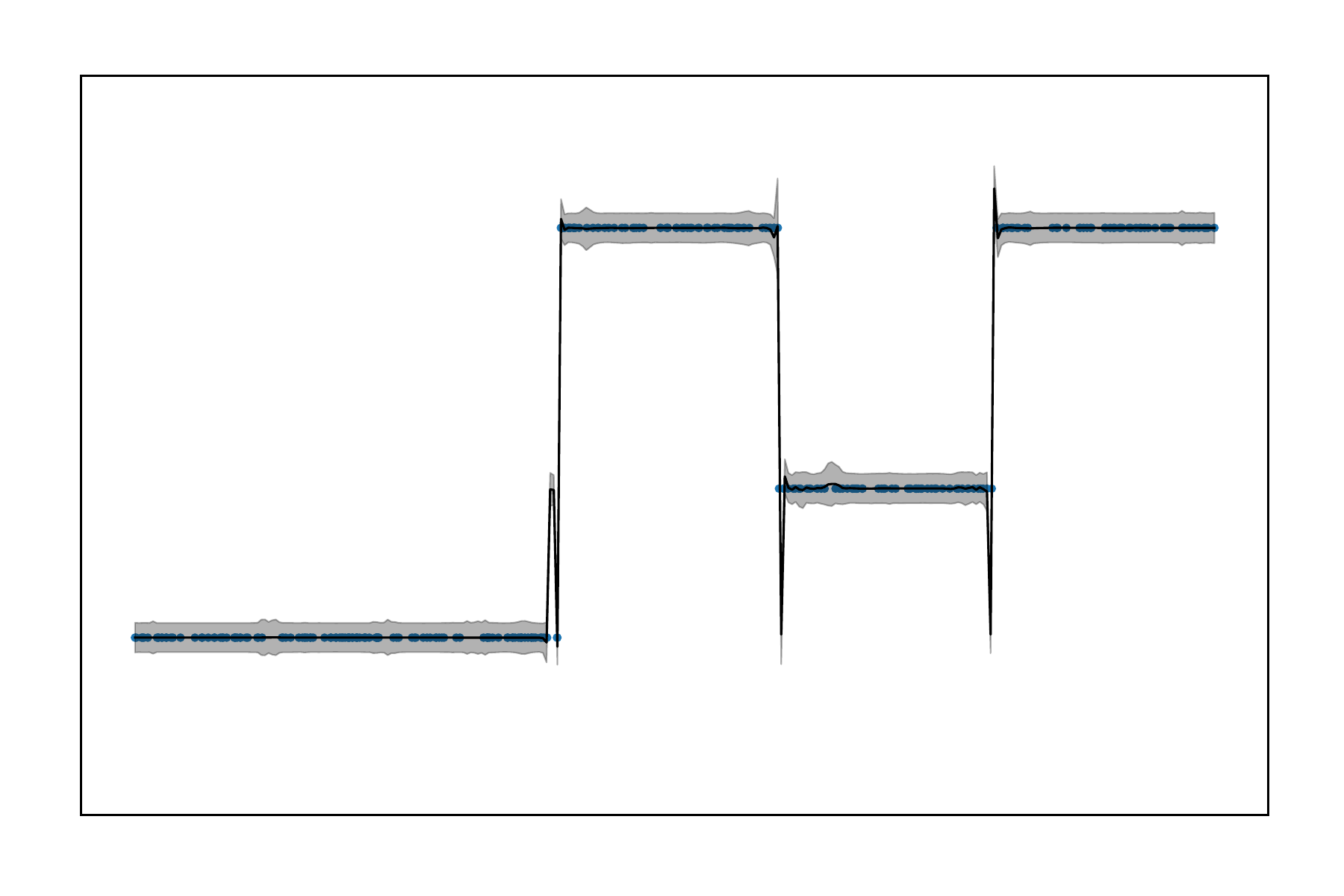}
        \end{subfigure}
        \hfill
        \begin{subfigure}[c]{0.32\linewidth}
            \includegraphics[width=\linewidth]{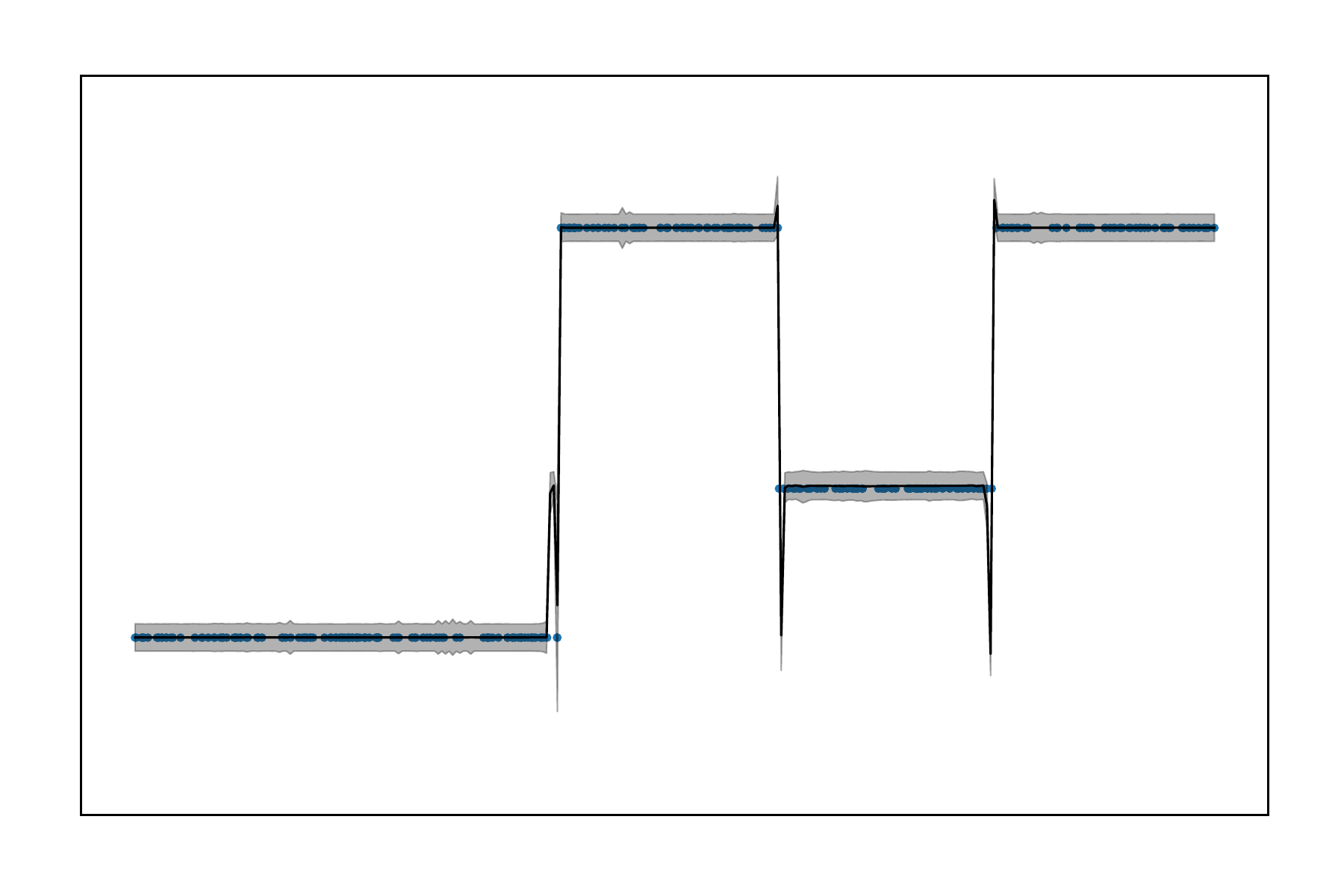}
        \end{subfigure}
        \hfill
        \begin{subfigure}[c]{0.32\linewidth}
            \includegraphics[width=\linewidth]{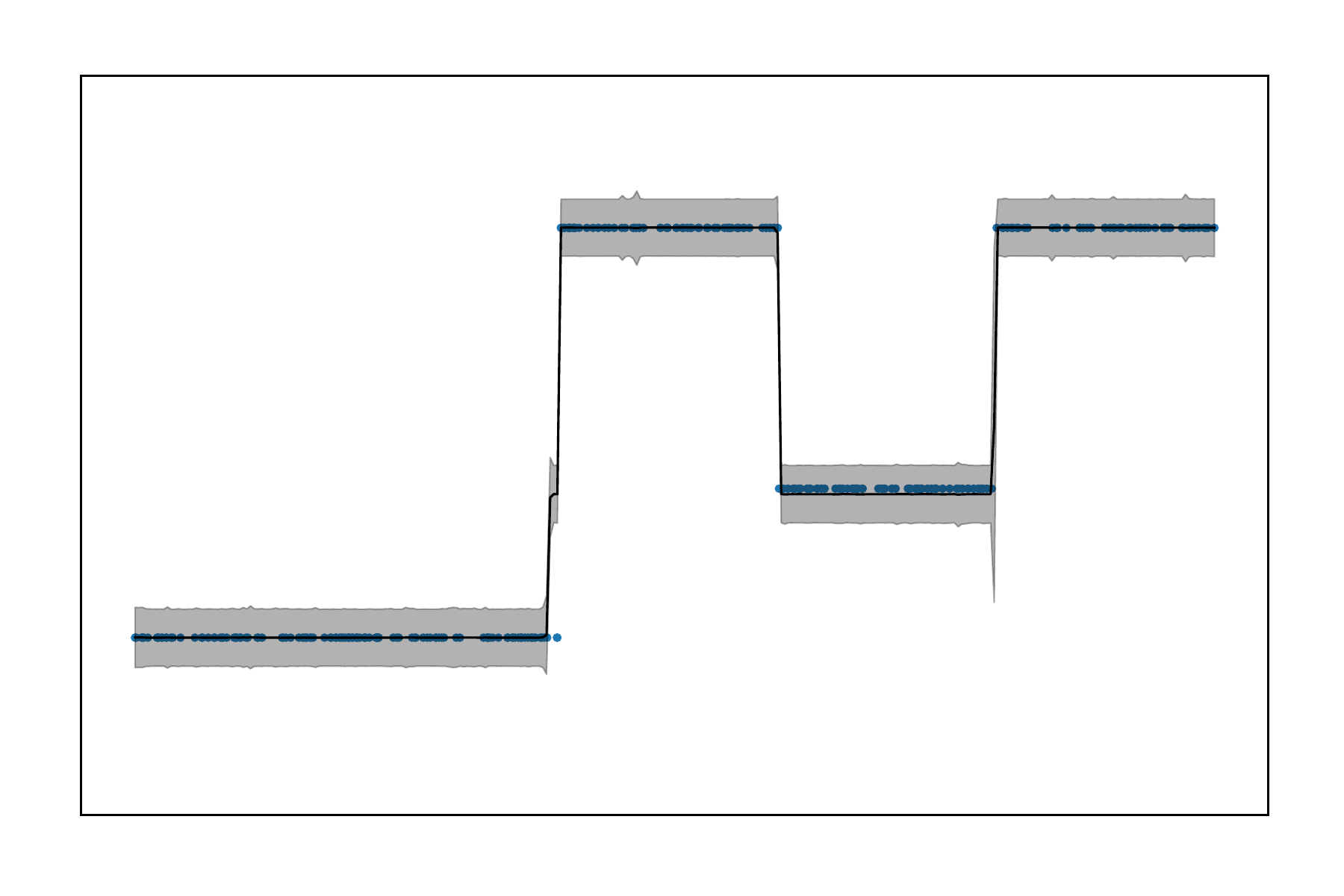}
        \end{subfigure}
    \end{minipage}
    
    \begin{minipage}[c]{\textwidth}
        \makebox[\linewidth][c]{\textbf{\PCAW{}}}%
    \end{minipage}
    \caption{Predictive distributions of the \ZEROW{} and \PCAW{} \DGP models when initialized with $\varSw{}{} =\mathbf{I}$ in all layers.}

    \label{fig:more_inducings_inner_high_output_high}
\end{figure}

\fig \ref{fig:more_inducings_inner_high_output_high} shows toy results using $\varSw{}{}=\matI$ in all layers. Following our previous analysis, we can observe how this configuration also yields poorer solutions in the \ZERO model and slightly higher predictive variance in the \PCA model. 

\begin{figure}[!htbp]
\centering
\begin{minipage}[c]{0.05\textwidth}
\end{minipage}%
\begin{minipage}[c]{0.93\textwidth}
\centering
\hfill
\makebox[0.32\linewidth][c]{\textbf{2 Layers}}%
\hfill
\makebox[0.32\linewidth][c]{\textbf{3 Layers}}%
\hfill
\makebox[0.32\linewidth][c]{\textbf{5 Layers}}
\end{minipage}
\begin{minipage}[c]{0.05\textwidth}
    \centering
    \rotatebox{90}{\textbf{5 Inducing}}
\end{minipage}%
\begin{minipage}[c]{0.93\textwidth}
    \centering
    \begin{subfigure}[c]{0.32\linewidth}        
    \includegraphics[width=\linewidth]{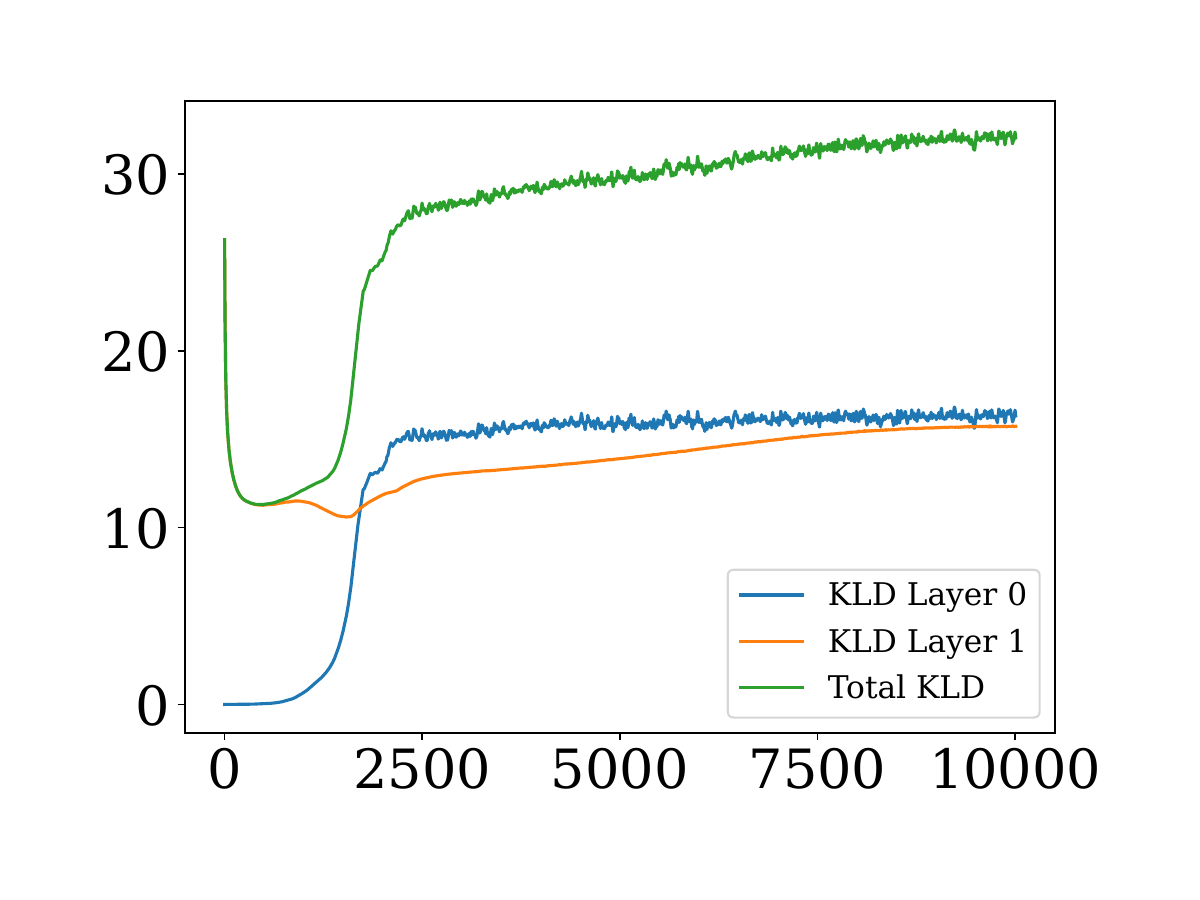}
    \end{subfigure}
    \begin{subfigure}[c]{0.32\linewidth}
\includegraphics[width=\linewidth]{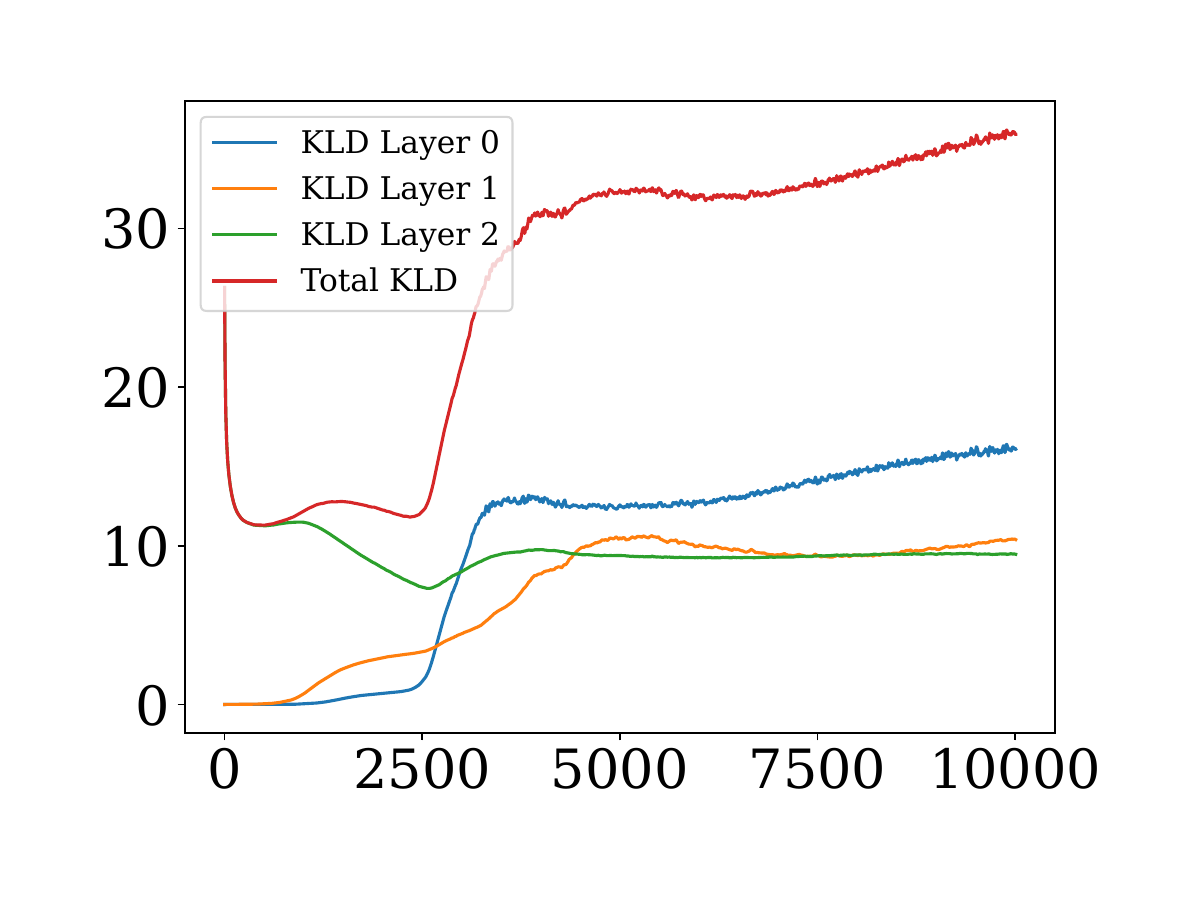}
    \end{subfigure}
    \begin{subfigure}[c]{0.32\linewidth}
\includegraphics[width=\linewidth]{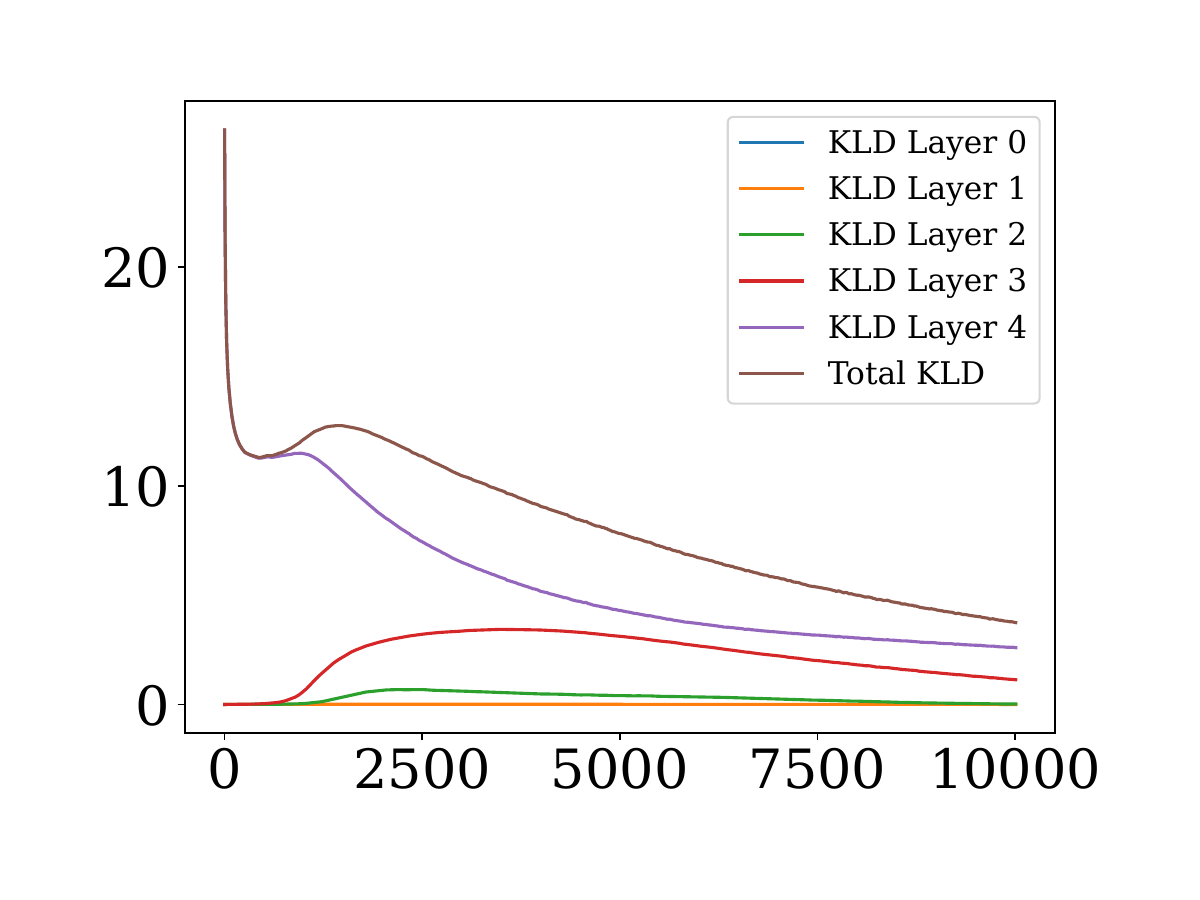}
    \end{subfigure}
\end{minipage}
\vspace{-2.5mm}
\begin{minipage}[c]{0.05\textwidth}
    \centering
    \rotatebox{90}{\textbf{20 Inducing}}
\end{minipage}%
\begin{minipage}[c]{0.93\textwidth}
    \centering
    \begin{subfigure}[c]{0.32\linewidth}
        \includegraphics[width=\linewidth]{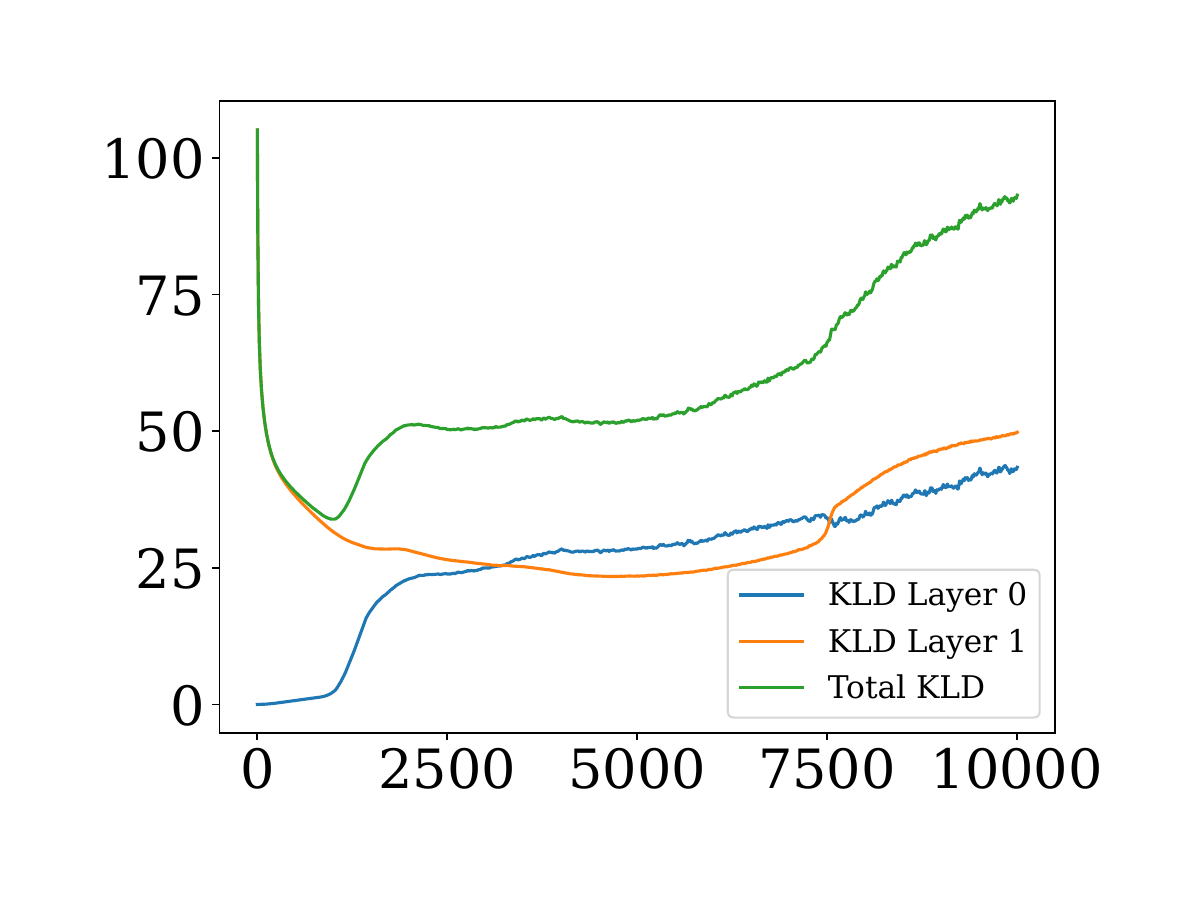}
    \end{subfigure}
    \begin{subfigure}[c]{0.32\linewidth}
        \includegraphics[width=\linewidth]{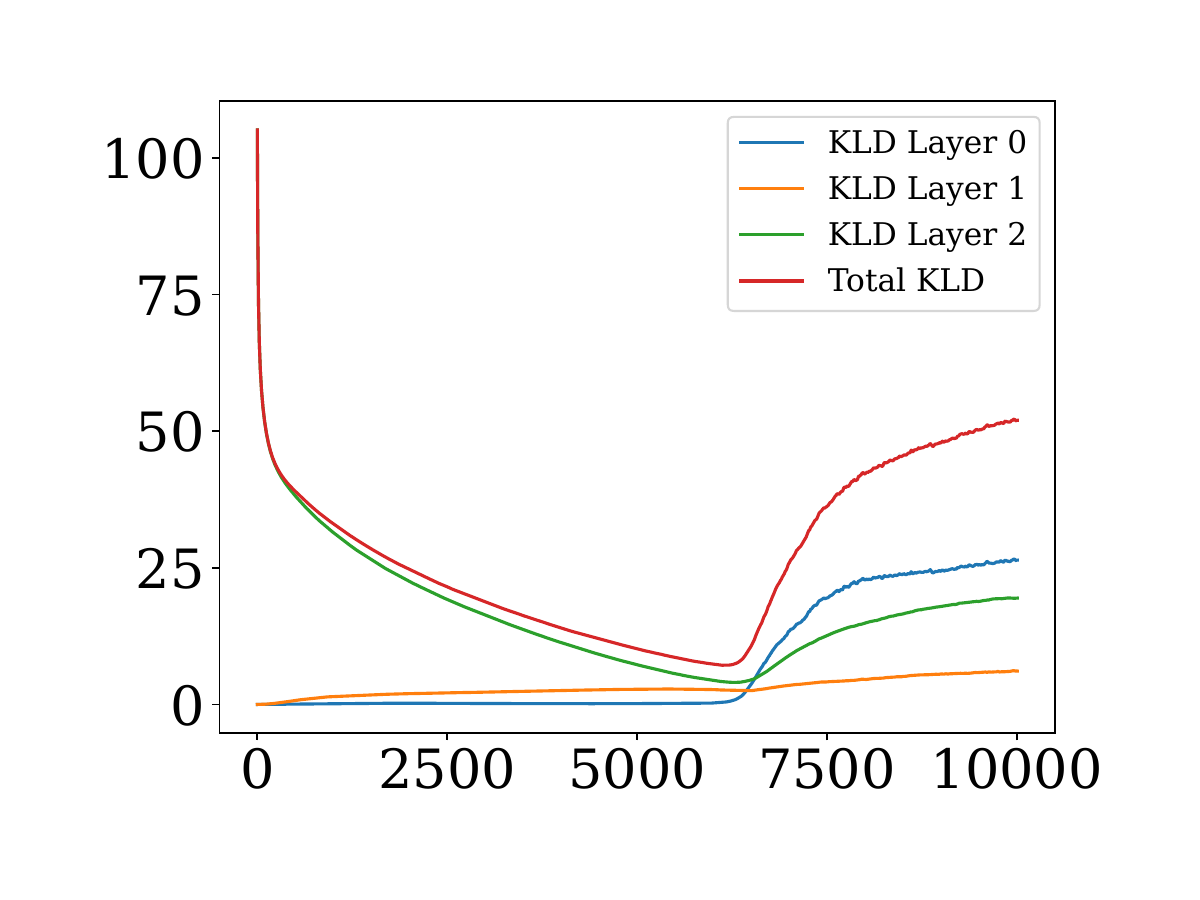}
    \end{subfigure}
    \begin{subfigure}[c]{0.32\linewidth}
        \includegraphics[width=\linewidth]{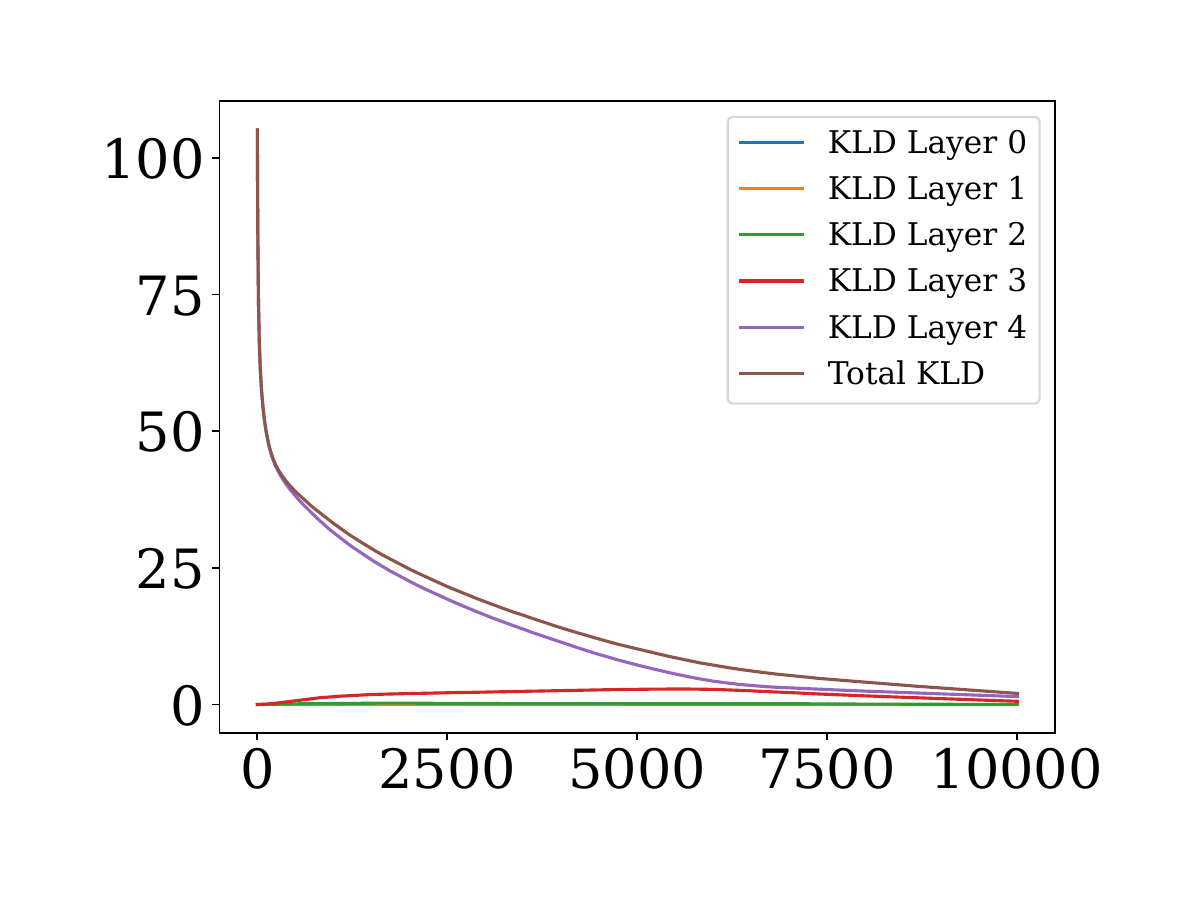}
    \end{subfigure}
\end{minipage}
\vspace{-2.5mm}
\begin{minipage}[c]{0.05\textwidth}
    \centering
    \rotatebox{90}{\textbf{100 Inducing}}
\end{minipage}%
\begin{minipage}[c]{0.93\textwidth}
    \centering
    \begin{subfigure}[c]{0.32\linewidth}
        \includegraphics[width=\linewidth]{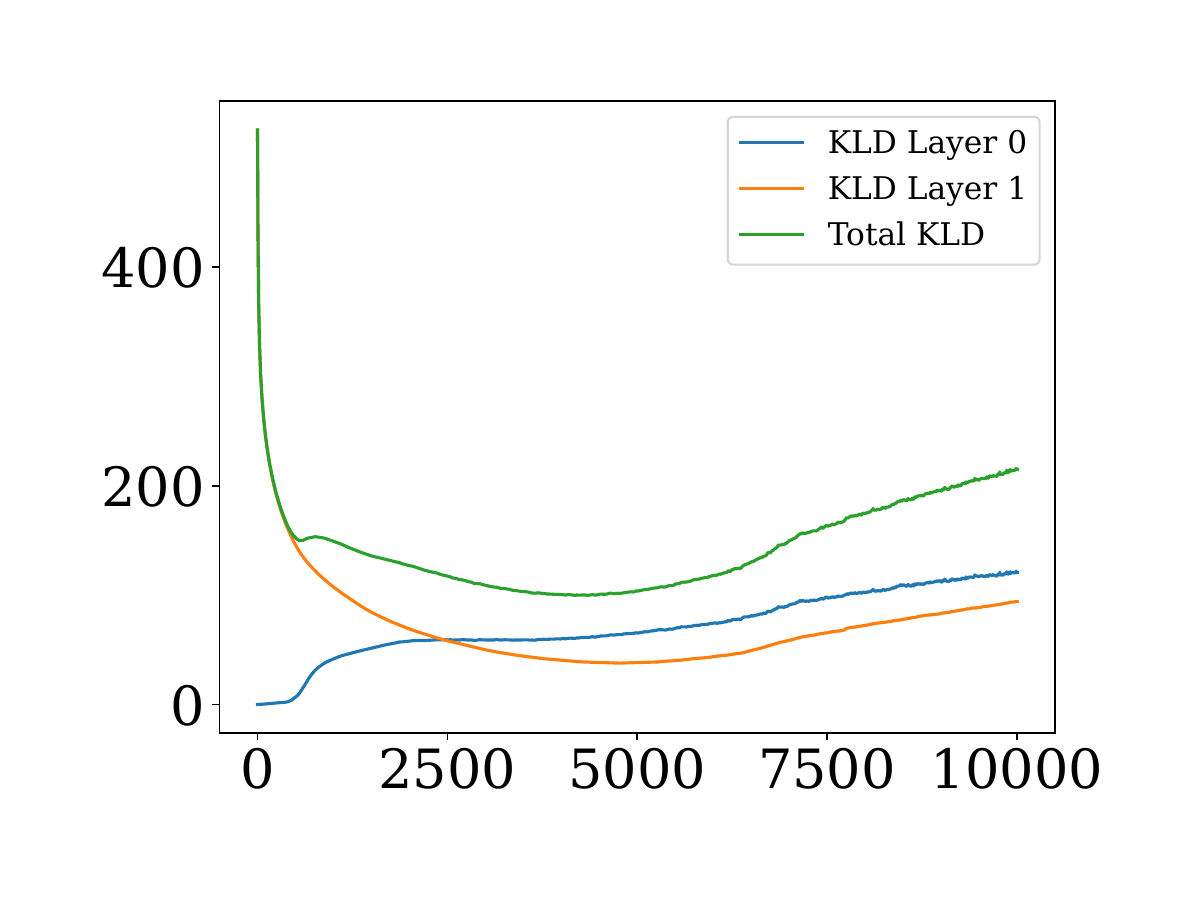}
    \end{subfigure}
    \begin{subfigure}[c]{0.32\linewidth}
        \includegraphics[width=\linewidth]{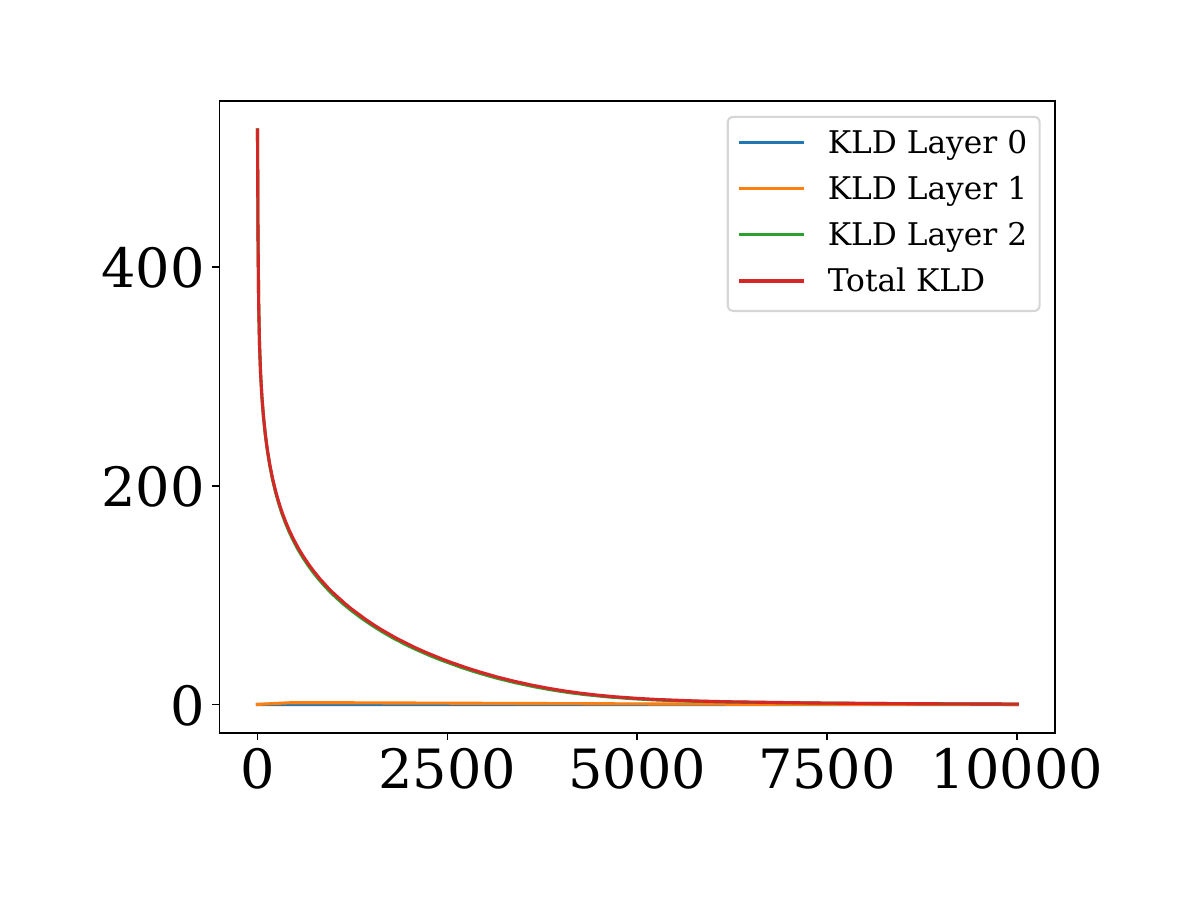}
    \end{subfigure}
    \begin{subfigure}[c]{0.32\linewidth}
        \includegraphics[width=\linewidth]{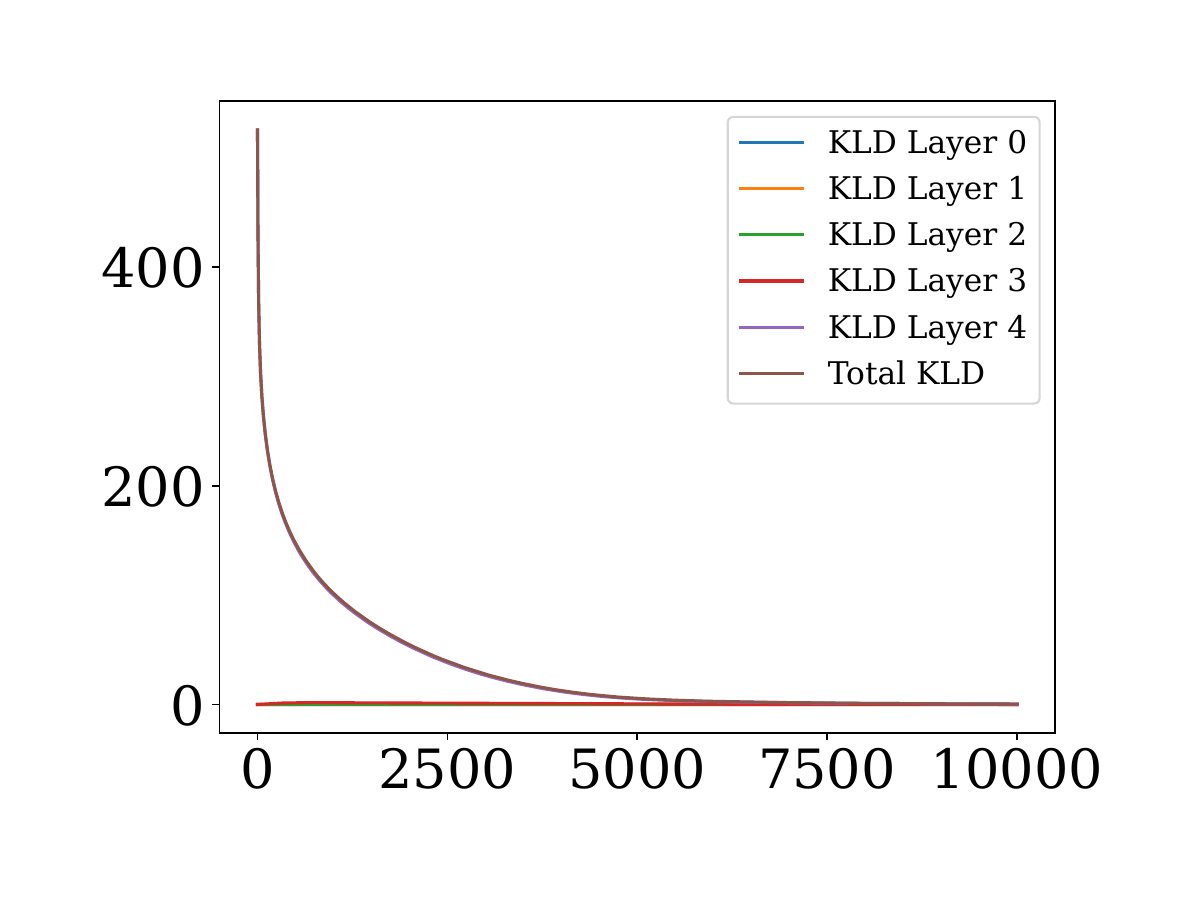}
    \end{subfigure}
\end{minipage}
\begin{minipage}[c]{\textwidth}
\makebox[\linewidth][c]{\textbf{\ZEROW{}}}
\end{minipage}
\vspace{-2.5mm}
\begin{minipage}[c]{0.05\textwidth}
    \centering
    \rotatebox{90}{\textbf{5 Inducing}}
\end{minipage}%
\begin{minipage}[c]{0.93\textwidth}
    \centering
    \begin{subfigure}[c]{0.32\linewidth}
        \includegraphics[width=\linewidth]{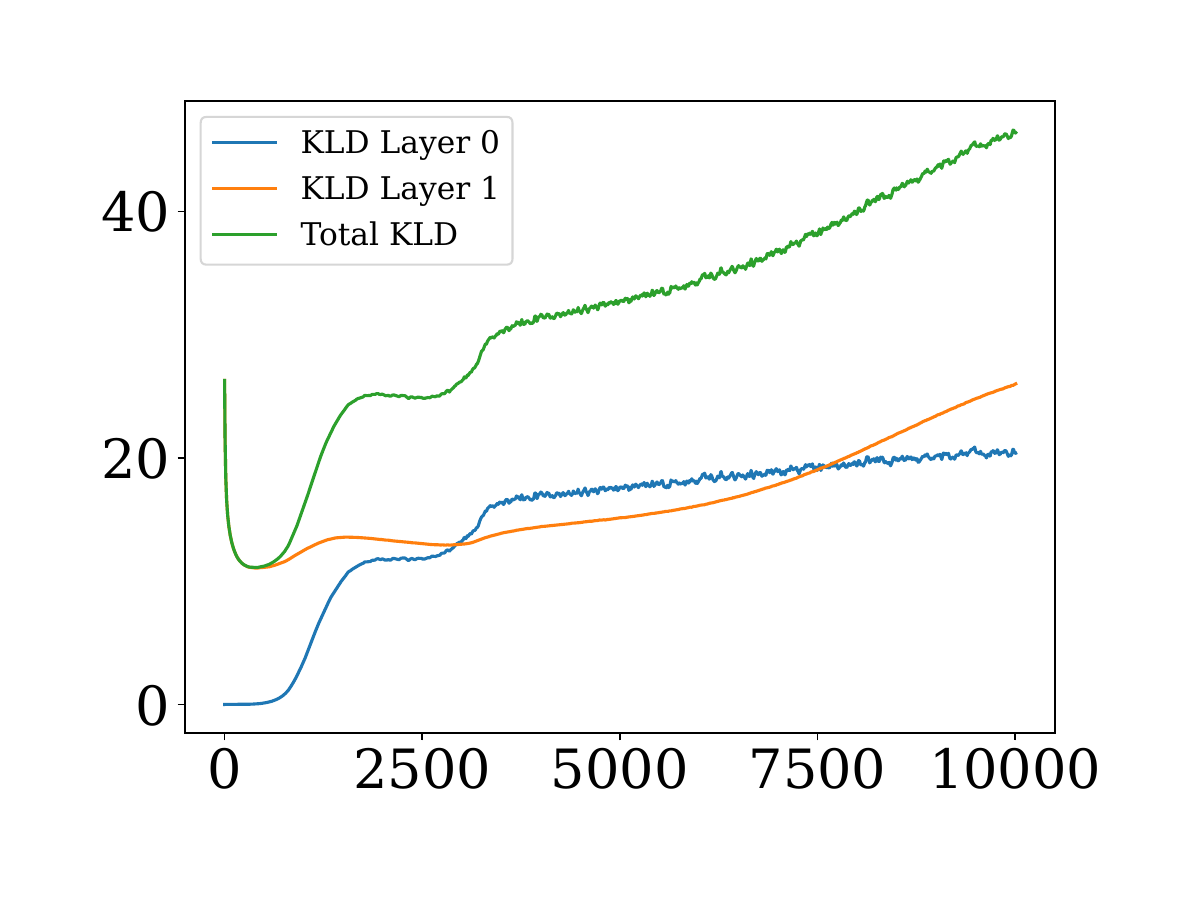}
    \end{subfigure}
    \begin{subfigure}[c]{0.32\linewidth}
        \includegraphics[width=\linewidth]{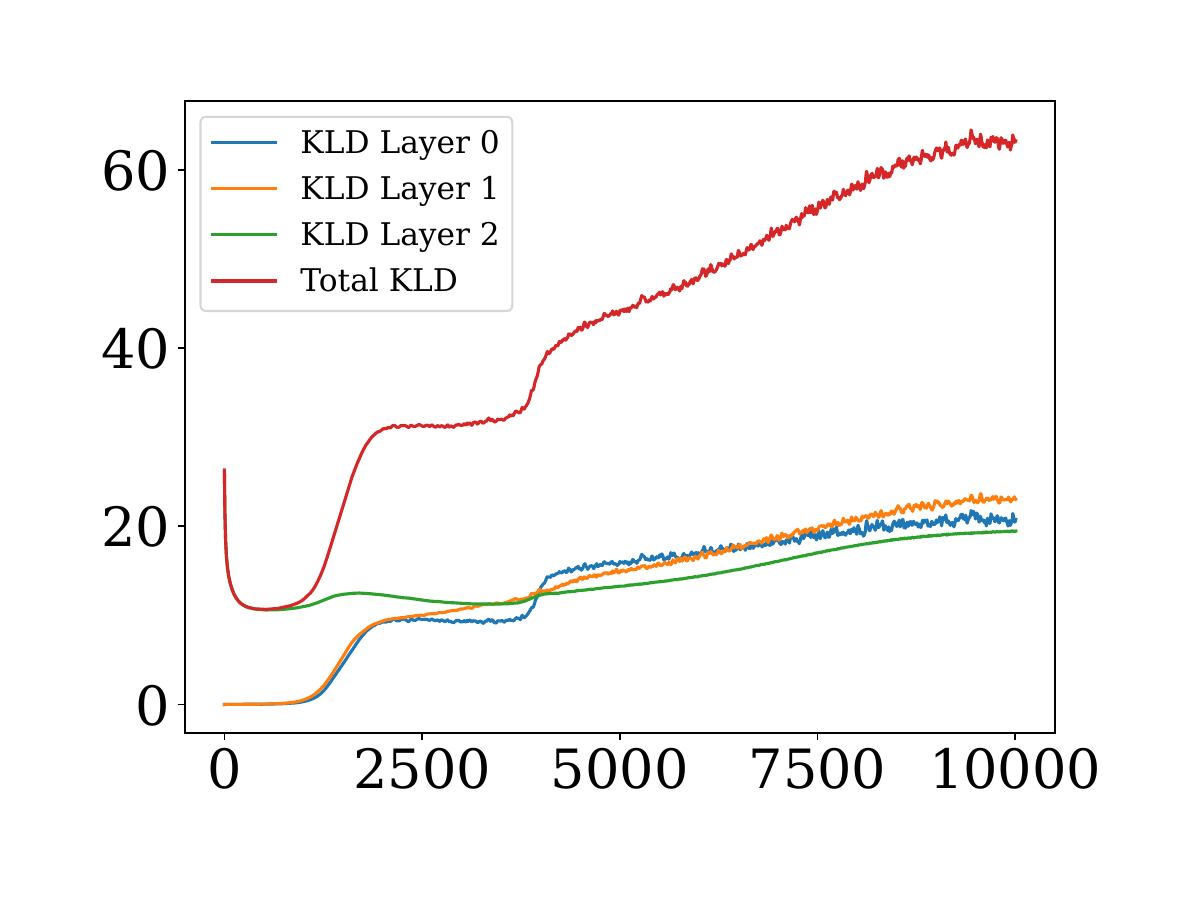}
    \end{subfigure}
    \begin{subfigure}[c]{0.32\linewidth}
        \includegraphics[width=\linewidth]{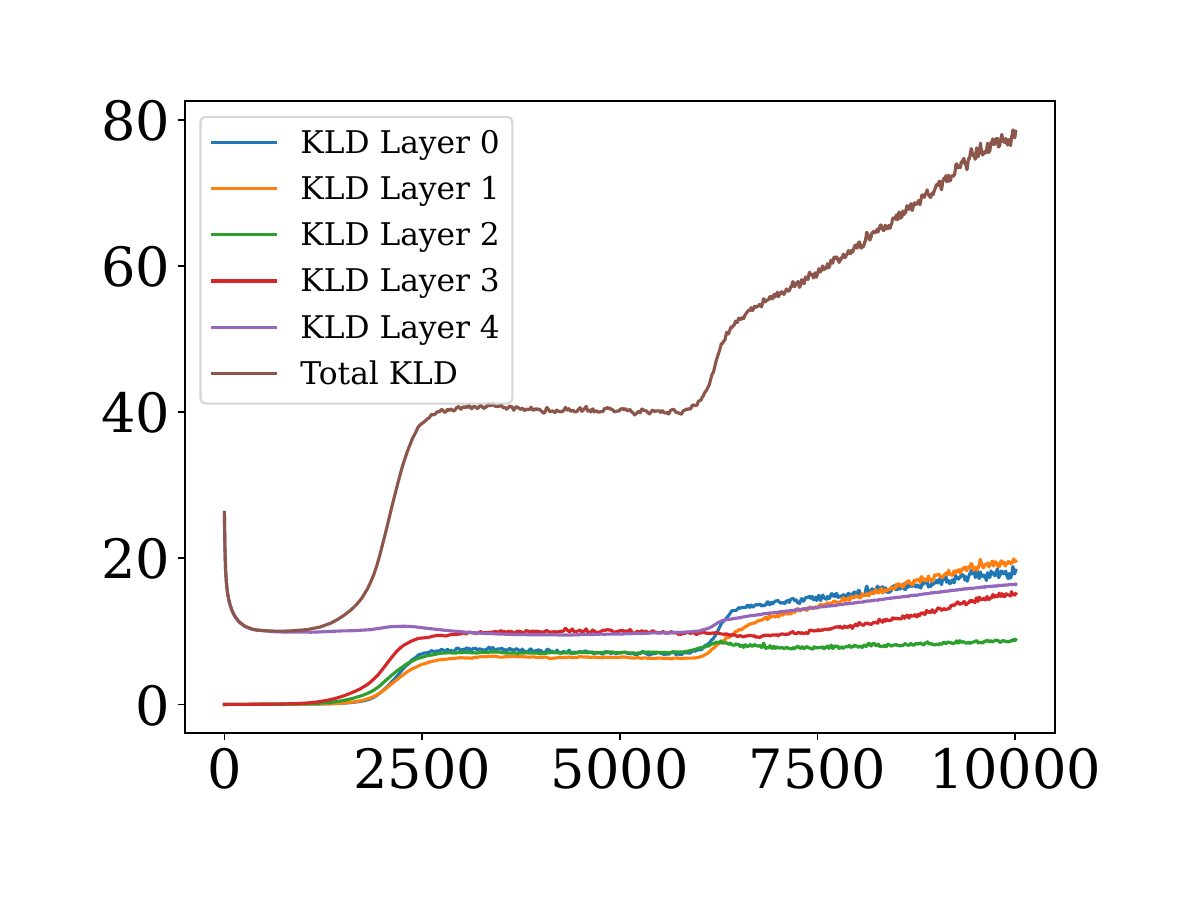}
    \end{subfigure}
\end{minipage}
\vspace{-2.5mm}
\begin{minipage}[c]{0.05\textwidth}
    \centering
    \rotatebox{90}{\textbf{20 Inducing}}
\end{minipage}%
\begin{minipage}[c]{0.93\textwidth}
    \centering
    \begin{subfigure}[c]{0.32\linewidth}
        \includegraphics[width=\linewidth]{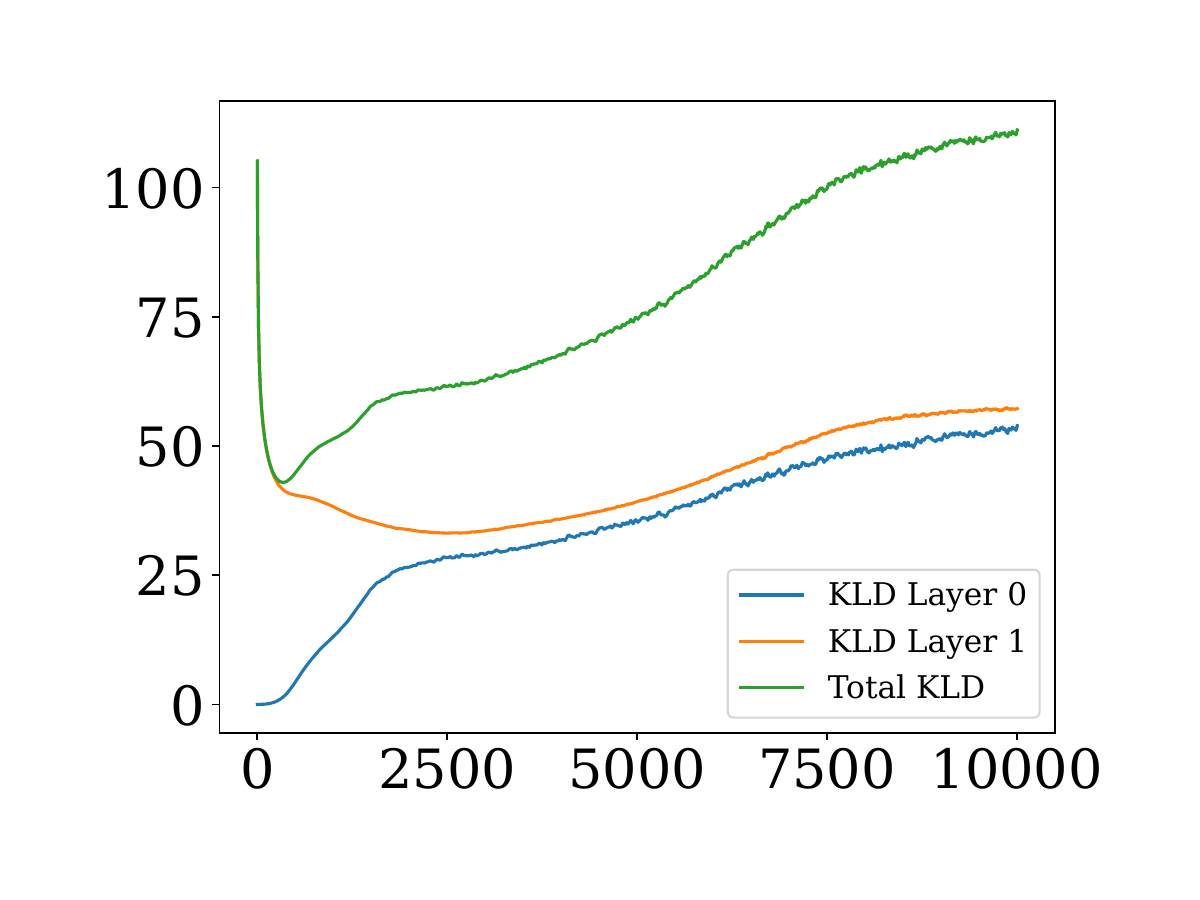}
    \end{subfigure}
    \begin{subfigure}[c]{0.32\linewidth}
        \includegraphics[width=\linewidth]{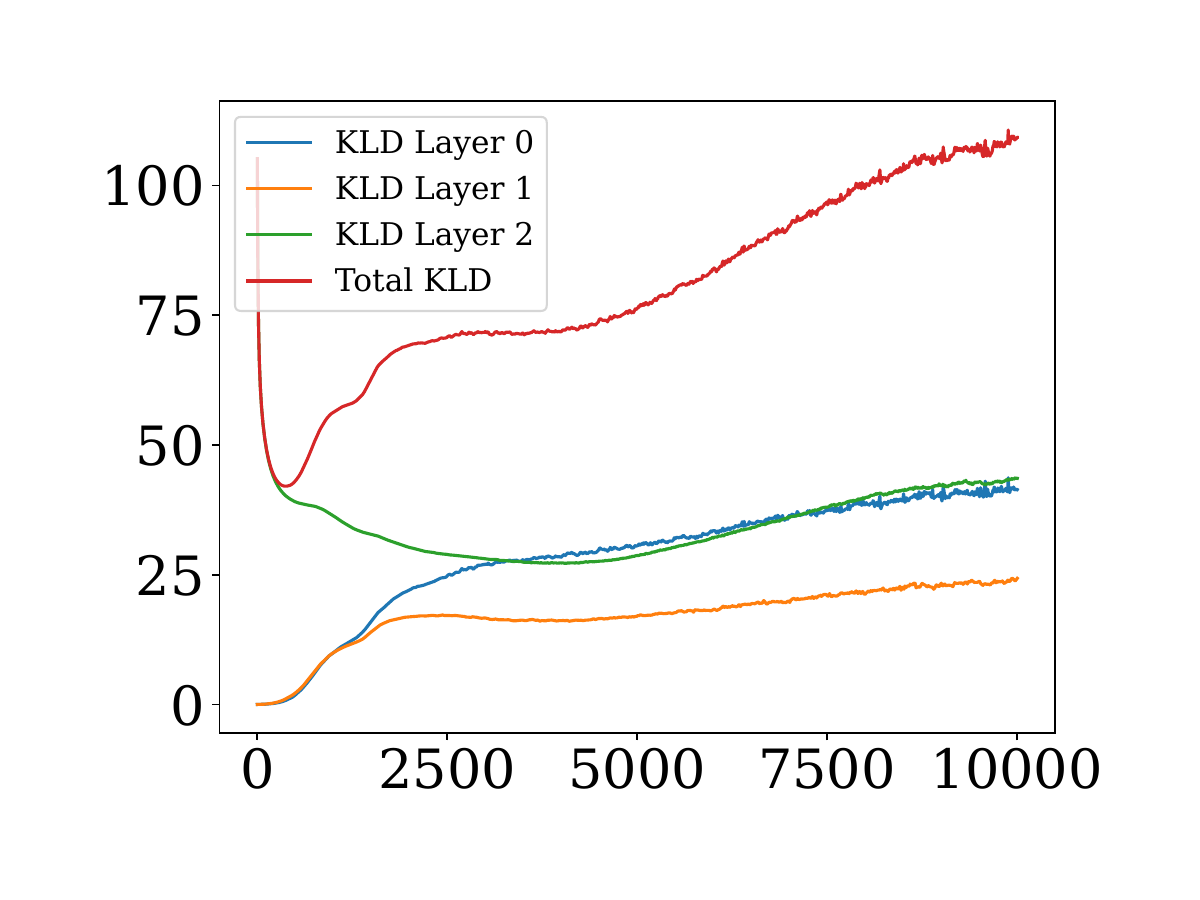}
    \end{subfigure}
    \begin{subfigure}[c]{0.32\linewidth}
        \includegraphics[width=\linewidth]{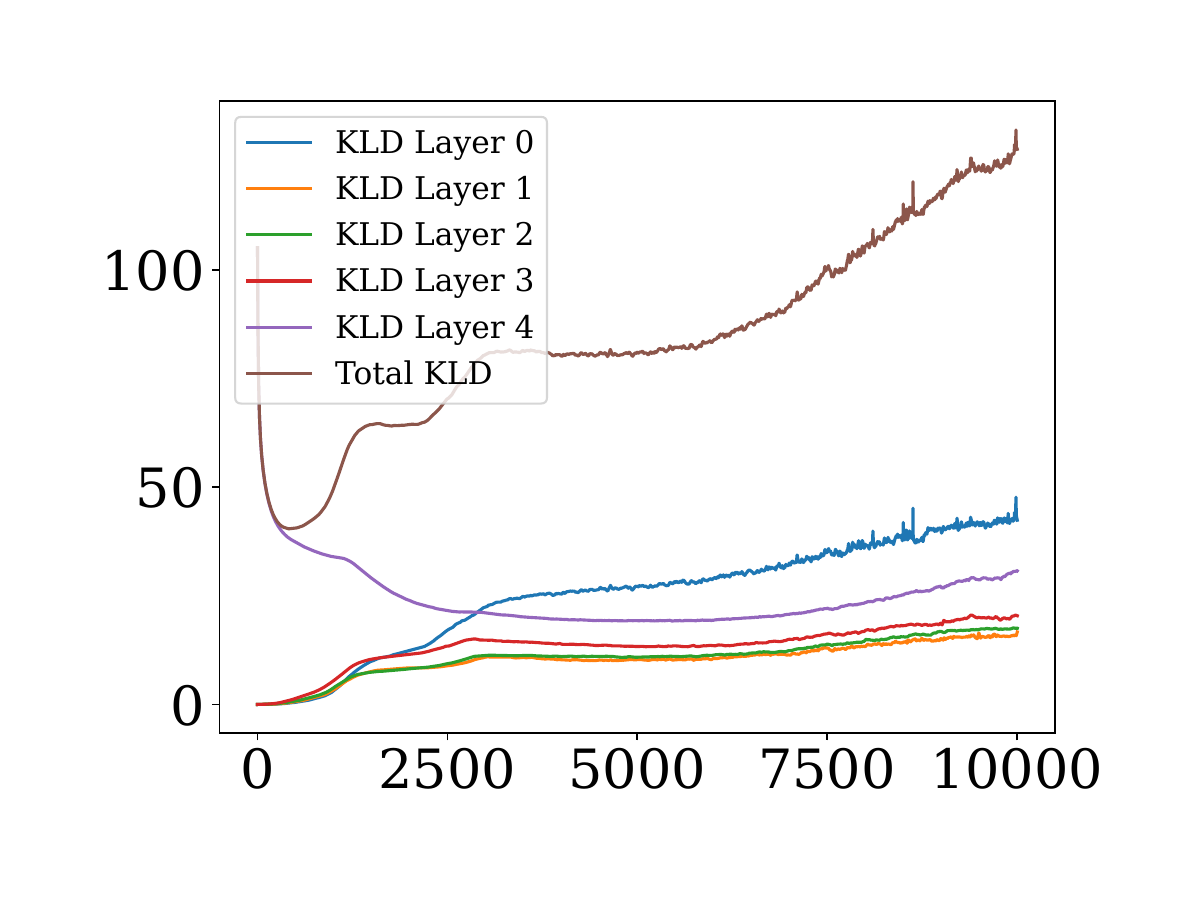}
    \end{subfigure}
\end{minipage}
\vspace{-2.5mm}
\begin{minipage}[c]{0.05\textwidth}
    \centering
    \rotatebox{90}{\textbf{100 Inducing}}
\end{minipage}%
\begin{minipage}[c]{0.93\textwidth}
    \centering

    \begin{subfigure}[c]{0.32\linewidth}
        \includegraphics[width=\linewidth]{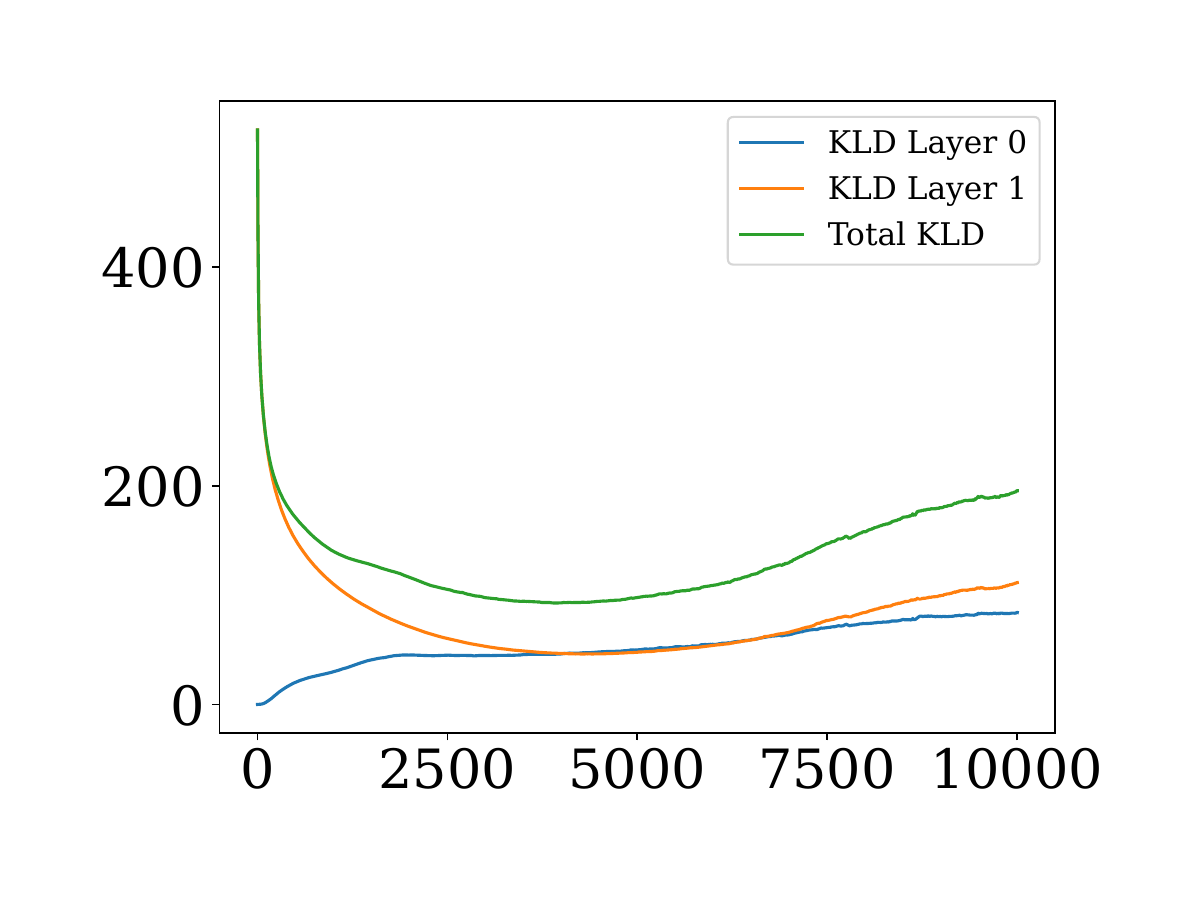}
    \end{subfigure}
    \begin{subfigure}[c]{0.32\linewidth}
        \includegraphics[width=\linewidth]{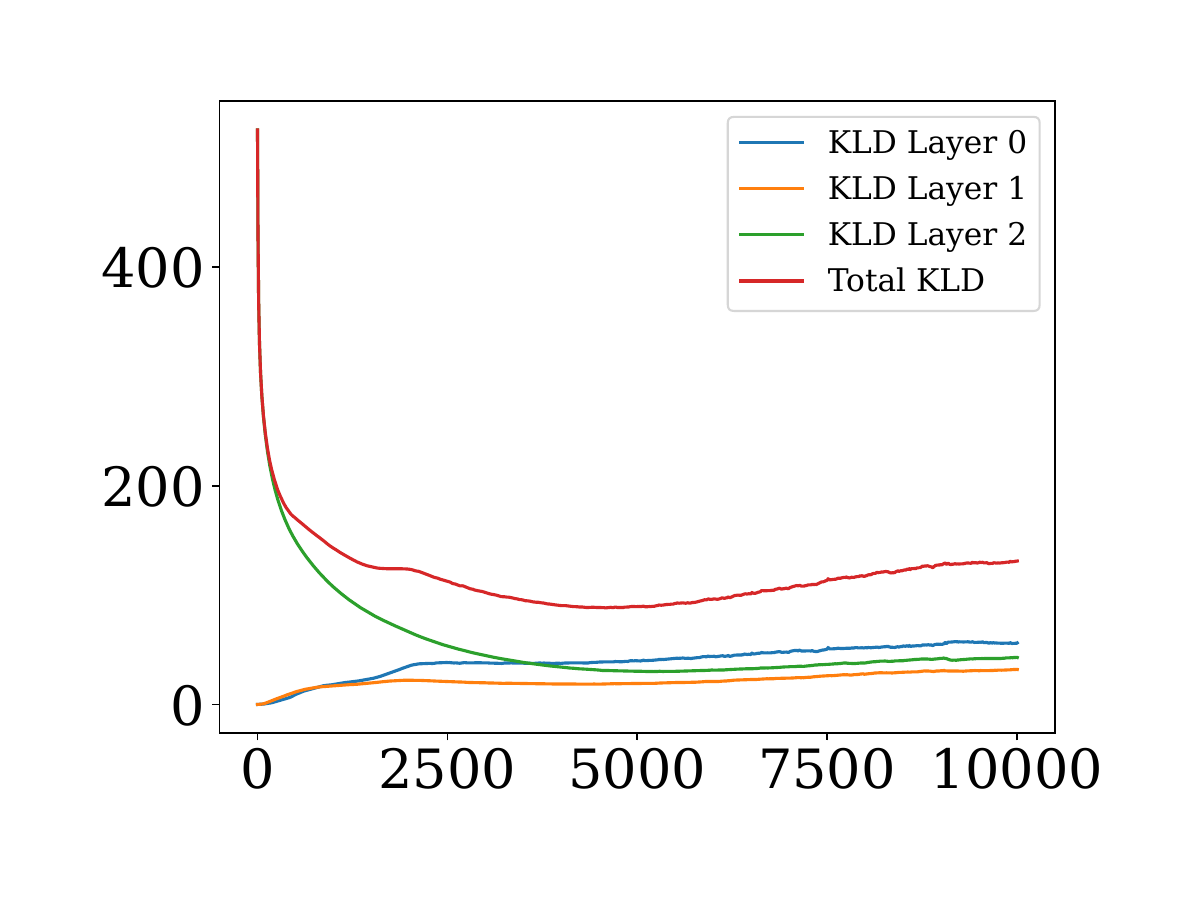}
    \end{subfigure}
    \begin{subfigure}[c]{0.32\linewidth}
        \includegraphics[width=\linewidth]{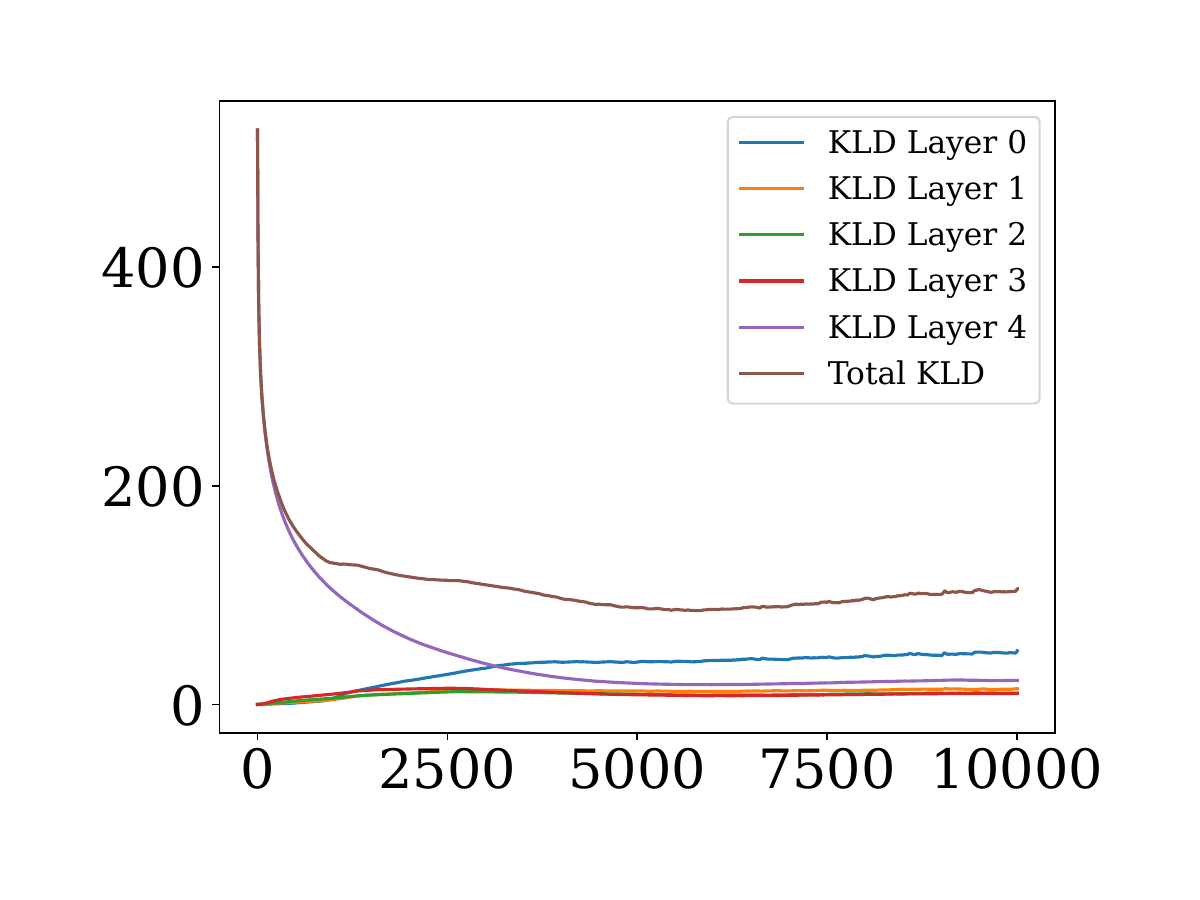}
    \end{subfigure}
\end{minipage}

\begin{minipage}[c]{\textwidth}
\makebox[\linewidth][c]{\textbf{\PCAW{}}}
\end{minipage}

\caption{Layer-wise \KLD{} during training of both \ZEROW{} and \PCAW{} models initialized with $\varSw{l}{} = \mathbf{I}$ in the inner layers and $\varSw{L}{} = 10^{-5}\mathbf{I}$ in the output layer. In some cases, the \ZEROW{} model can not escape from the local optimum of zero \KLD{} in some layers, especially with a big number of inducing points. With a lower number of inducing points, it is more likely that the \ZEROW{} escapes from the posterior collapse, as in the case of $2$ and $3$ layers and $5$ inducing points.}
\label{fig:more_inducings_klds_all_inner_high_out_low}
\end{figure}

For the experiment with $\varS{l}{}=\matI$ and $\varS{}{L}=10^{-5}\matI$,  \fig \ref{fig:more_inducings_klds_all_inner_high_out_low} shows the layer-wise \KLD{} of the model during training. We observe that in the cases with $20$ and $100$ inducing points and $3-5$ layers, the inner layers of the models essentially do not modify their \KLD{}, effectively not learning, which results in the final posterior collapse. Using a lower number of inducing points (in this figure, $5$) and $3$ layers, we observe how at the beginning of the training the model increases its \KLD{} to model the data, thus escaping from the posterior collapse effect. Nevertheless, the final solution is still suboptimal due to noisy optimization. With $5$  layers and $5$ inducing points, we observe how even if the model escapes from collapse, at the end all layers are tending to collapse. 
The \PCAW{} model never falls into the zero \KLD{} local minimum.

\clearpage

\subsection{Additional Toy Experiments Illustrating Optimization Instabilities Depending on the Parameterization}\label{sec:app:c:3:toy:results}

\begin{figure}[!ht]
    \centering
    \begin{subfigure}[t]{0.49\linewidth}
        \centering
        \includegraphics[width=\linewidth]{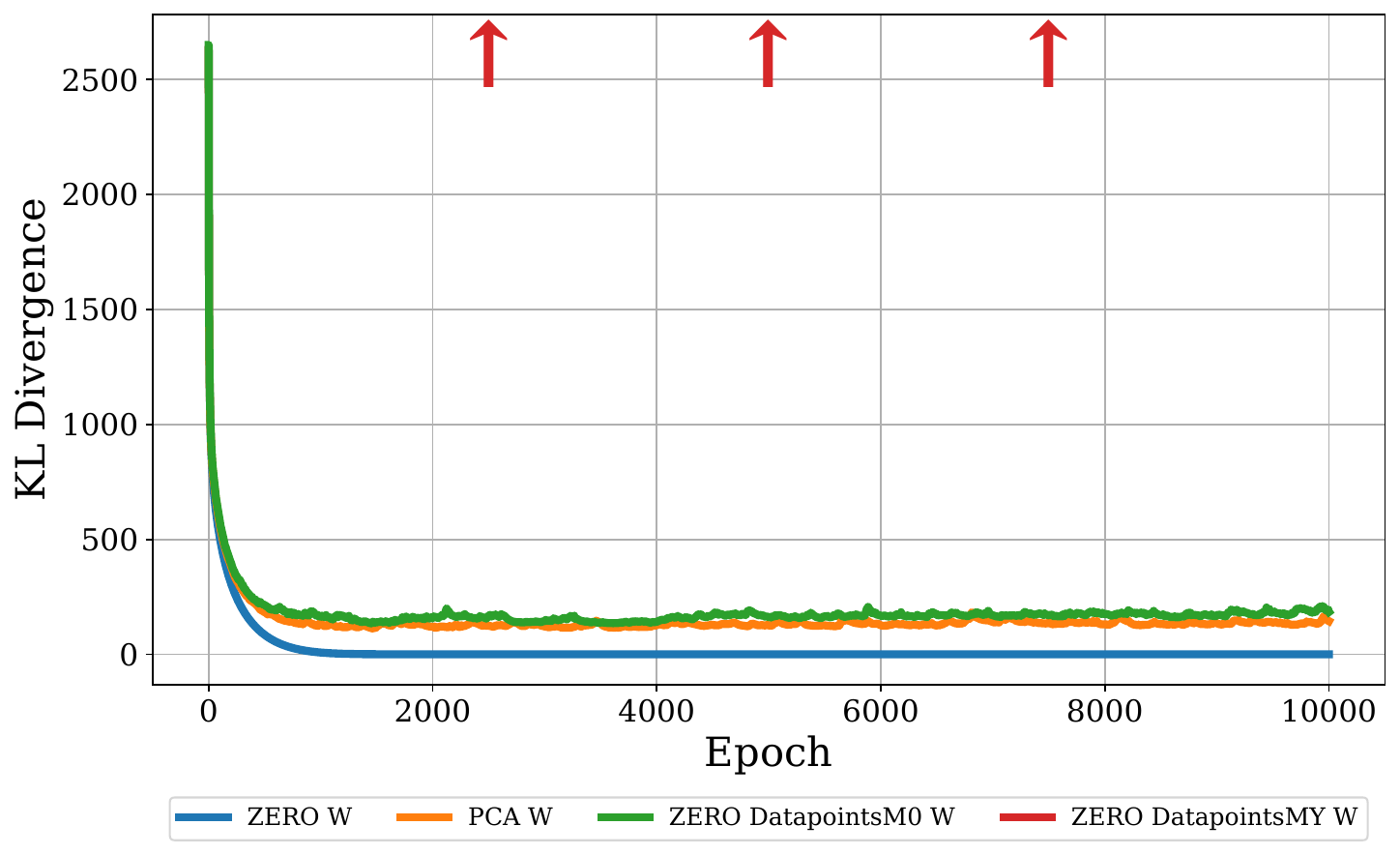}
        \caption{\KLD{} per epoch.}
        \label{fig:zero_w_collapsed:kl}
    \end{subfigure}
    \hfill
    \begin{subfigure}[t]{0.49\linewidth}
        \centering
        \includegraphics[width=\linewidth]{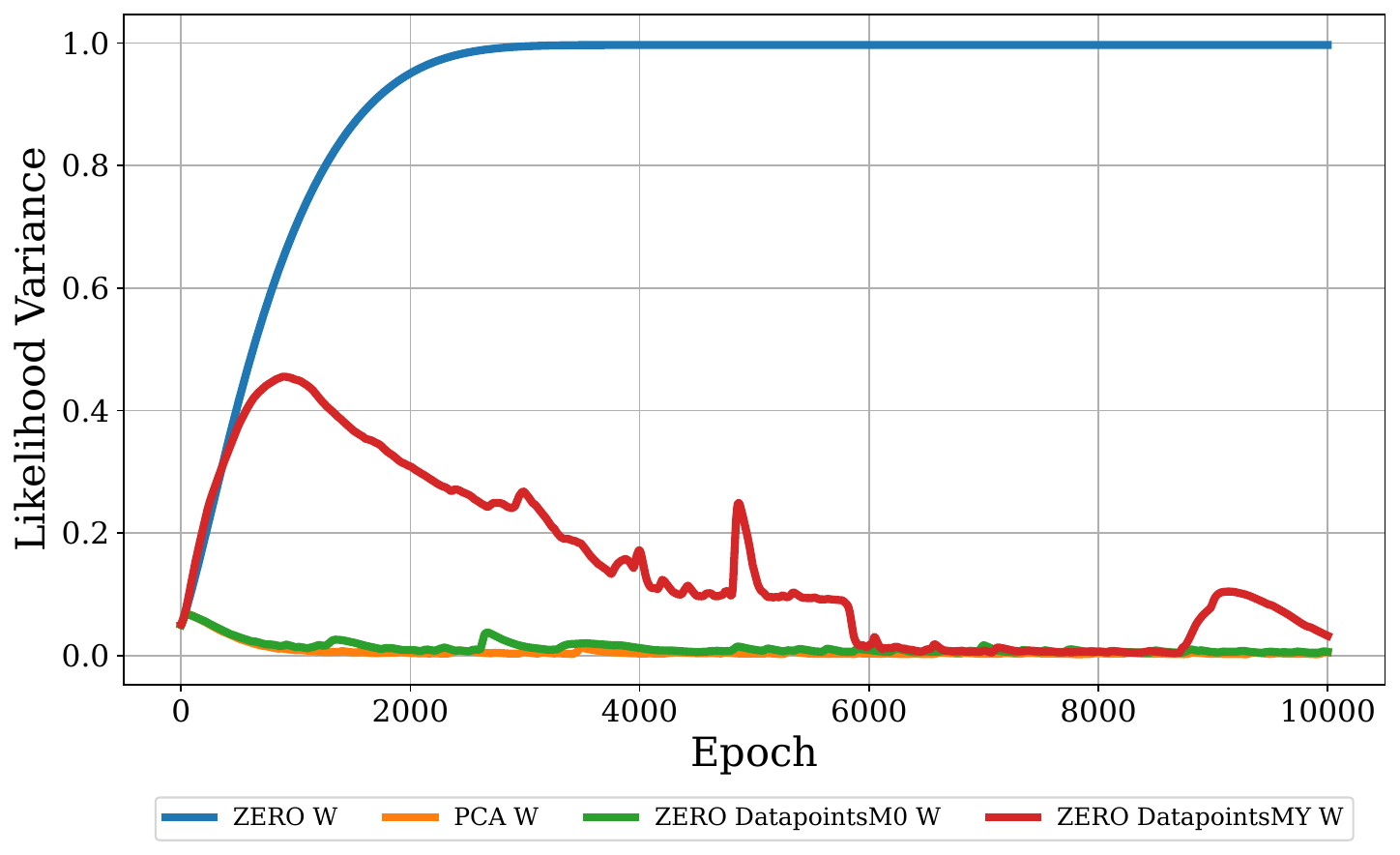}
        \caption{Likelihood variance per epoch.}
        \label{fig:zero_w_collapsed:likelihood_variance}
    \end{subfigure}
    \caption{\KLD{} and likelihood variance during training of the 5-layer whitened models in the toy dataset using $\lambda = 10^{-2}$.}
    \label{fig:zero_w_collapse:all}
\end{figure}

\begin{figure}[!ht]
    \centering
    \begin{subfigure}[t]{0.49\linewidth}
        \centering
        \includegraphics[width=\linewidth]{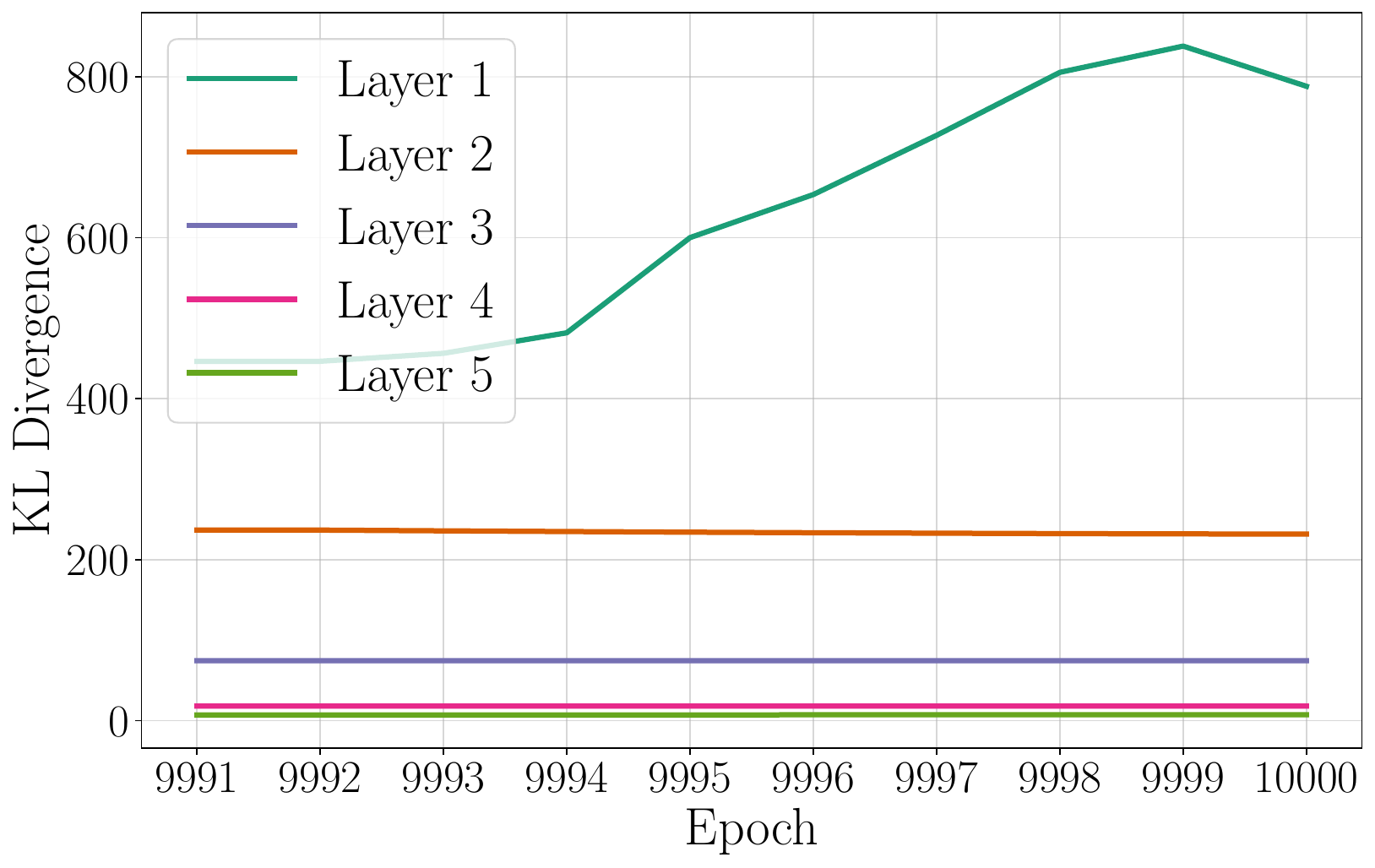}
        \caption{Layer-wise \KLD{} of the \PCANWR{}  in the last 10 epochs of training.}
        \label{fig:pca_nwr_collapsed:kl}
    \end{subfigure}
    \hfill
    \begin{subfigure}[t]{0.49\linewidth}
        \centering
        \includegraphics[width=\linewidth]{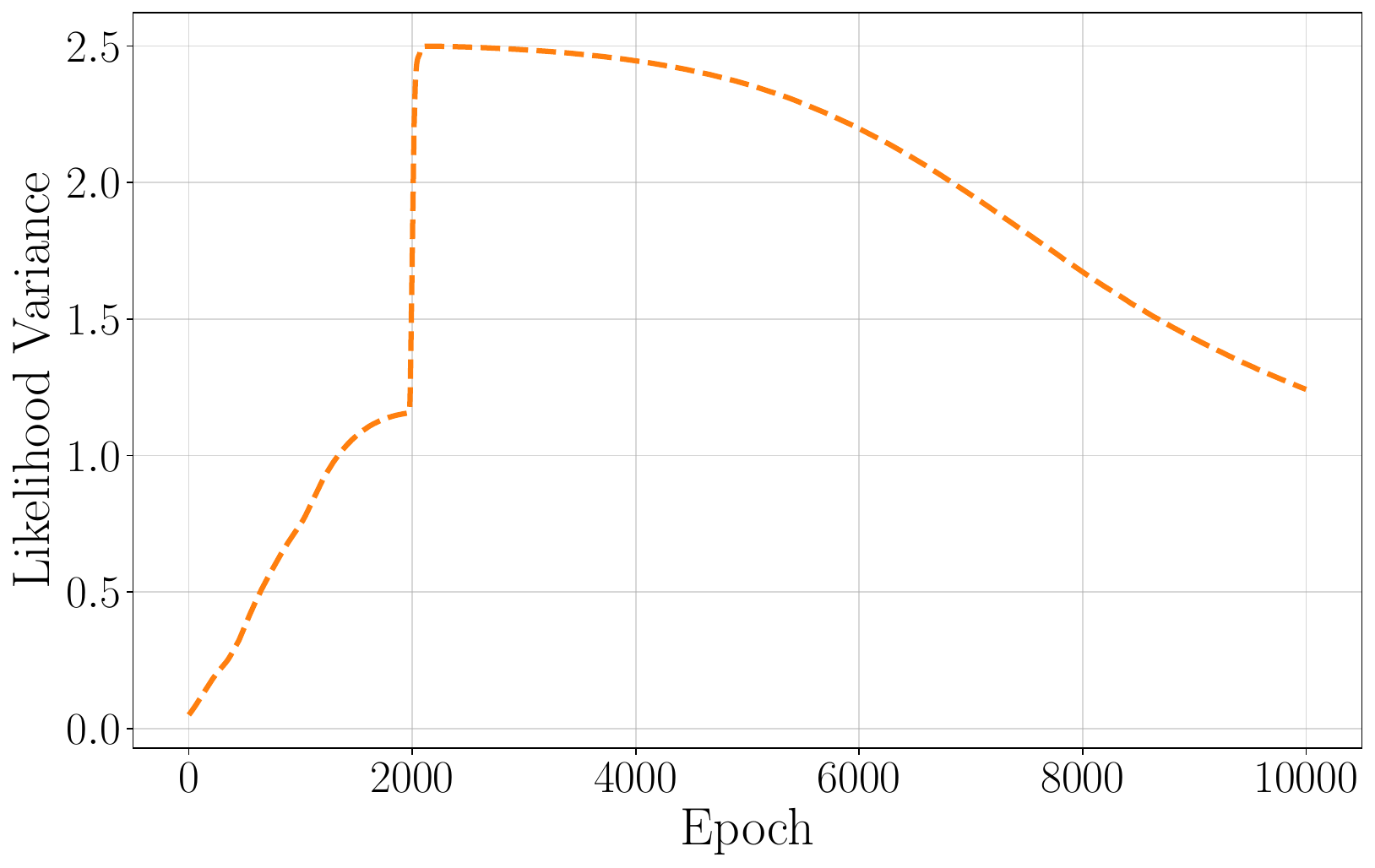}
        \caption{Likelihood variance per epoch of the \PCANWR{} model.}
        \label{fig:pca_nwr_collapsed:likelihood_variance}
    \end{subfigure}
    \caption{Layer-wise \KLD{} in the last 10 epochs of training and likelihood variance value during training of the 5-layer \PCANWR{} trained using $\lambda = 10^{-2}$.}
    \label{fig:pca_nwr_collapse:all}
\end{figure}

\begin{figure}[!b]
    \centering
    %
    \begin{subfigure}[t]{0.49\linewidth}
        \centering
        \includegraphics[width=\textwidth]{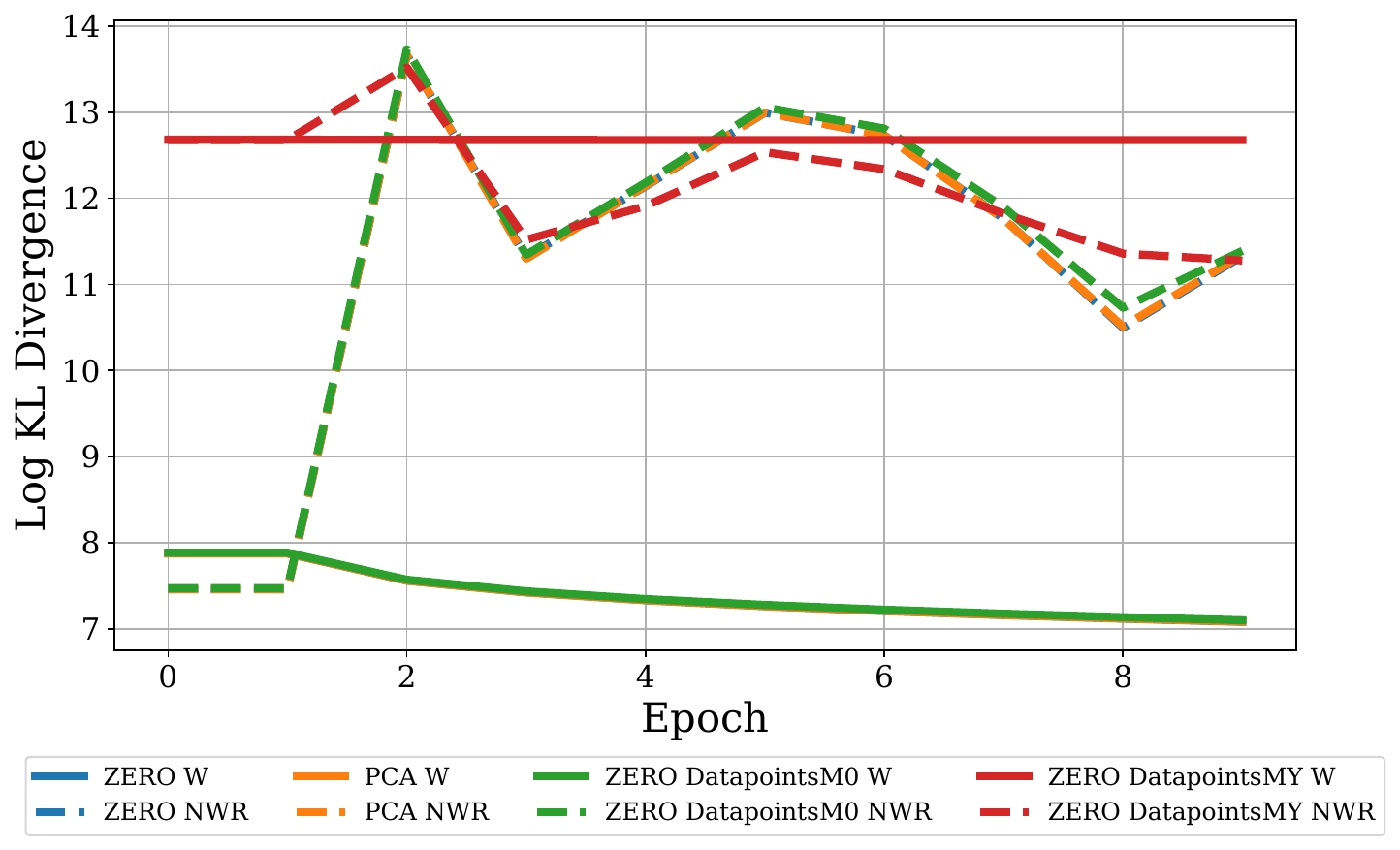}
        \caption{Learning rate: $10^{-2}$}
        \label{fig:toy:all:klds:0.01}
    \end{subfigure}
    \begin{subfigure}[t]{0.49\linewidth}
        \centering
    \includegraphics[width=\textwidth]{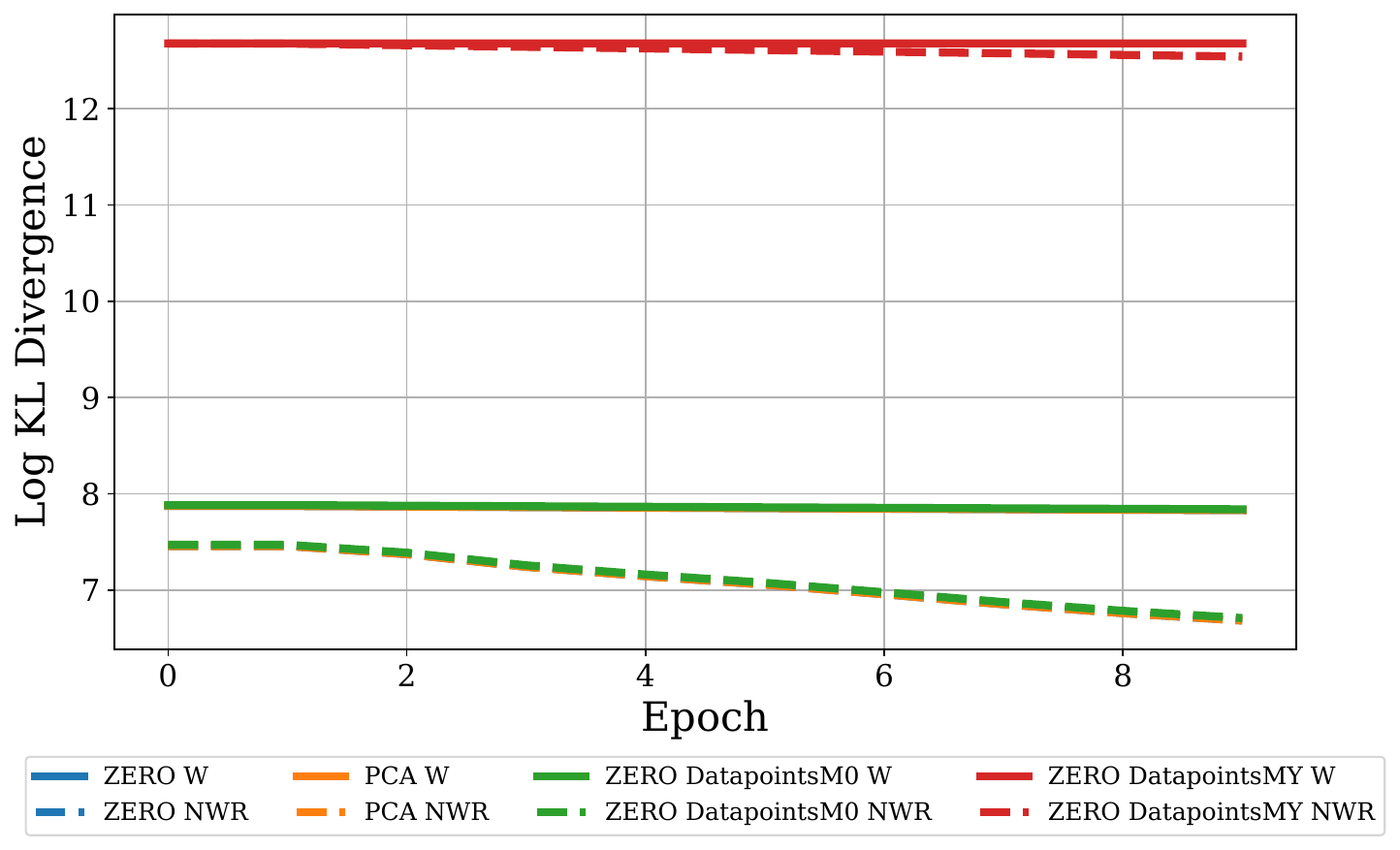}
        \caption{Learning rate: $10^{-4}$}
        \label{fig:toy:all:klds:0.0001}
    \end{subfigure}
    \caption{Log \KLD{} during the first $10$ epochs in the toy dataset. Non-visible curves overlap with visible ones. This figure shows how the \NWR model with $\lambda=10^{-2}$ results in a very unstable optimization, manifested through the peaks in the learning curve.}
    \label{fig:toy:klds}
\end{figure}

In the experimental section, we run the models using a learning rate of $10^{-4}$. We have also run the same experiments using $10^{-2}$. With this learning rate, as we see in \utab~\ref{tab:toy:test:results_lowerlr}, the \ZEROW{} model suffers posterior collapse, while the other models avoid it, see \fig~\ref{fig:zero_w_collapse:all}. In the case of \NWR models, all of them (including the \PCA) suffer from optimization instabilities that result in posterior collapse. These \NWR models clearly obtain worse results than their whitened version. While we do not observe exactly a \KLD of zero in the \PCANWR{} model, we do observe a high likelihood variance. By inspecting the \KLD at each layer, see \fig \ref{fig:pca_nwr_collapse:all}, we observe how the \PCA \NWR model collapses two of the layers and compensates this behavior by increasing the variance. This is an indicator of how even the \PCA model might result in a collapsed distribution when noise is injected in the optimization. Note that, in this work, we have presented two examples of a \PCA model suffering collapse: one coming from the \NWR parameterization, and another in a whitened parameterization where a small number of inducing points is used. Here, however, the \KLD from the \NWR models is higher than the whitened counterparts. This comes from the peaks in the optimization algorithm, see \fig \ref{fig:toy:klds}.

\begin{table}[!htp]
    \centering

\begin{tabular}{l|ccc}
\hline
    Model & \KLD{} & Lik. Var. & \RMSE{}  \\
    \hline
    \ZEROW{}       & $0.0010$ & $0.9967$ & $0.9896$  \\
    \ZERONWR{}     & $869.0811$ & $1.1559$ & $0.9895$  \\
    \PCAW{}        & $144.5761$ & $0.0053$ & $\mathbf{0.0420}$  \\
    \PCANWR{}      & $3310.1553$ & $1.1707$ & $0.9895$  \\
    \ZEROWMO{}     & $184.2085$ & $0.0058$ & $0.0580$  \\
    \ZERONWRMO{}   & $92017.6629$ & $1.0156$ & $0.9903$ \\
    \ZEROWMY{}     & $144253.7899$ & $0.0321$ & $0.0946$ \\
    \ZERONWRMY{}   & $2164.8801$ & $1.5375$ & $0.9901$  \\
    \hline
\end{tabular}
    
    \caption{Test metrics achieved by all the methods in the toy dataset using a higher learning rate ($\lambda = 10^{-2}$).}
    \label{tab:toy:test:results_lowerlr}
\end{table}

\clearpage
\subsection{Additional \UCI Results}
In this section, we present additional experimental results on \UCI{} datasets that complement the analysis performed in \usec~\ref{sec:experiments:uci}. We include the results of the models in terms of the \RMSE{}, compare the results between whitened and non-whitened parameterizations, and analyze the behavior of the models in some particular, interesting splits.
\label{sec:app:c:4:uci}
\subsubsection{\RMSE{} Results} \label{sec:app:c:1:rmse}
\begin{figure}[!p]
\vspace{-1.5cm}
    \centering
    \begin{subfigure}{0.9\linewidth}
        \centering
        \includegraphics[width=0.9\linewidth]{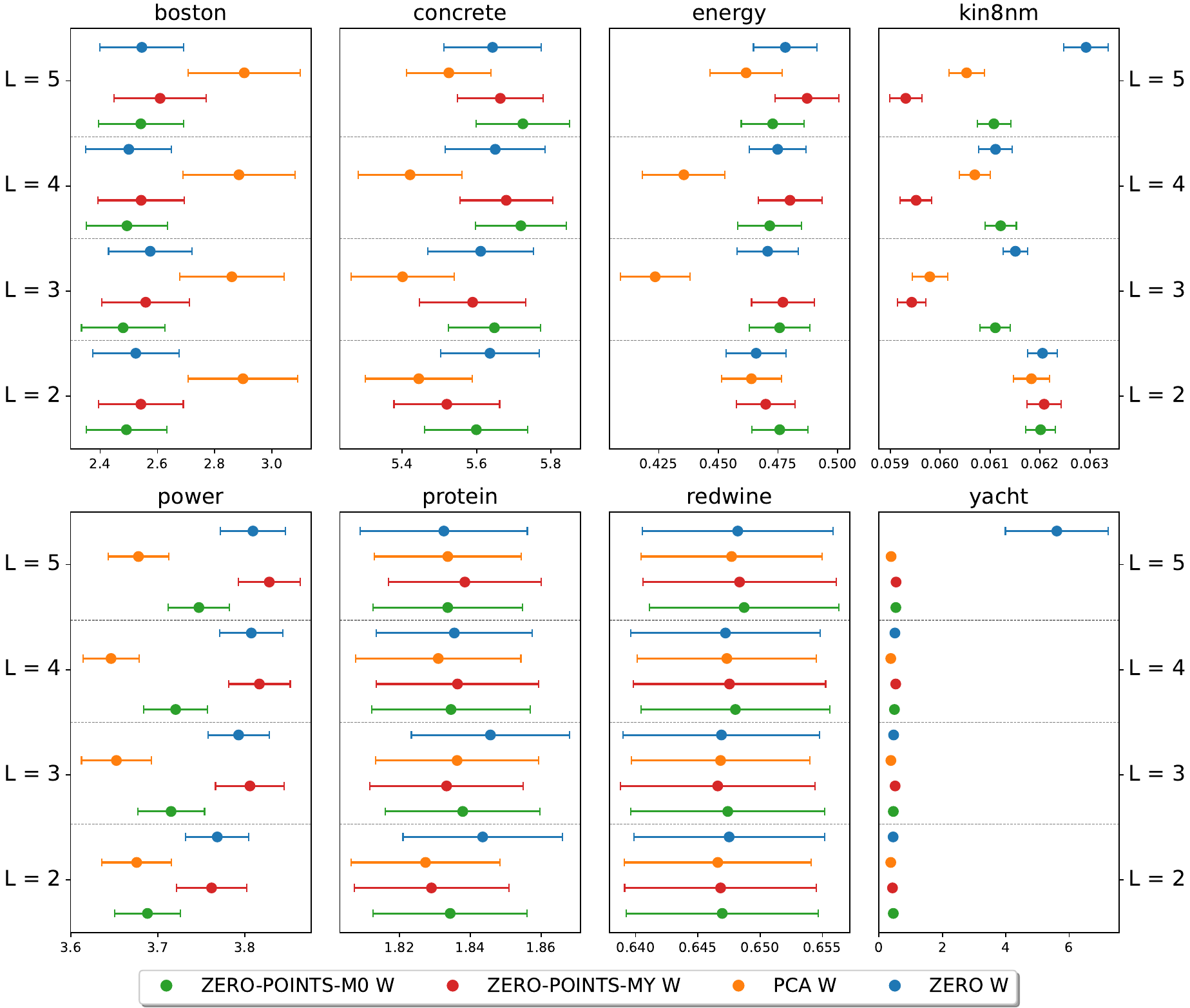}
        \caption{Whitened models.}
    \label{fig:results:rmse:whiten}
    \end{subfigure}
    \begin{subfigure}{0.9\linewidth}
        \centering
        \includegraphics[width=0.9\linewidth]{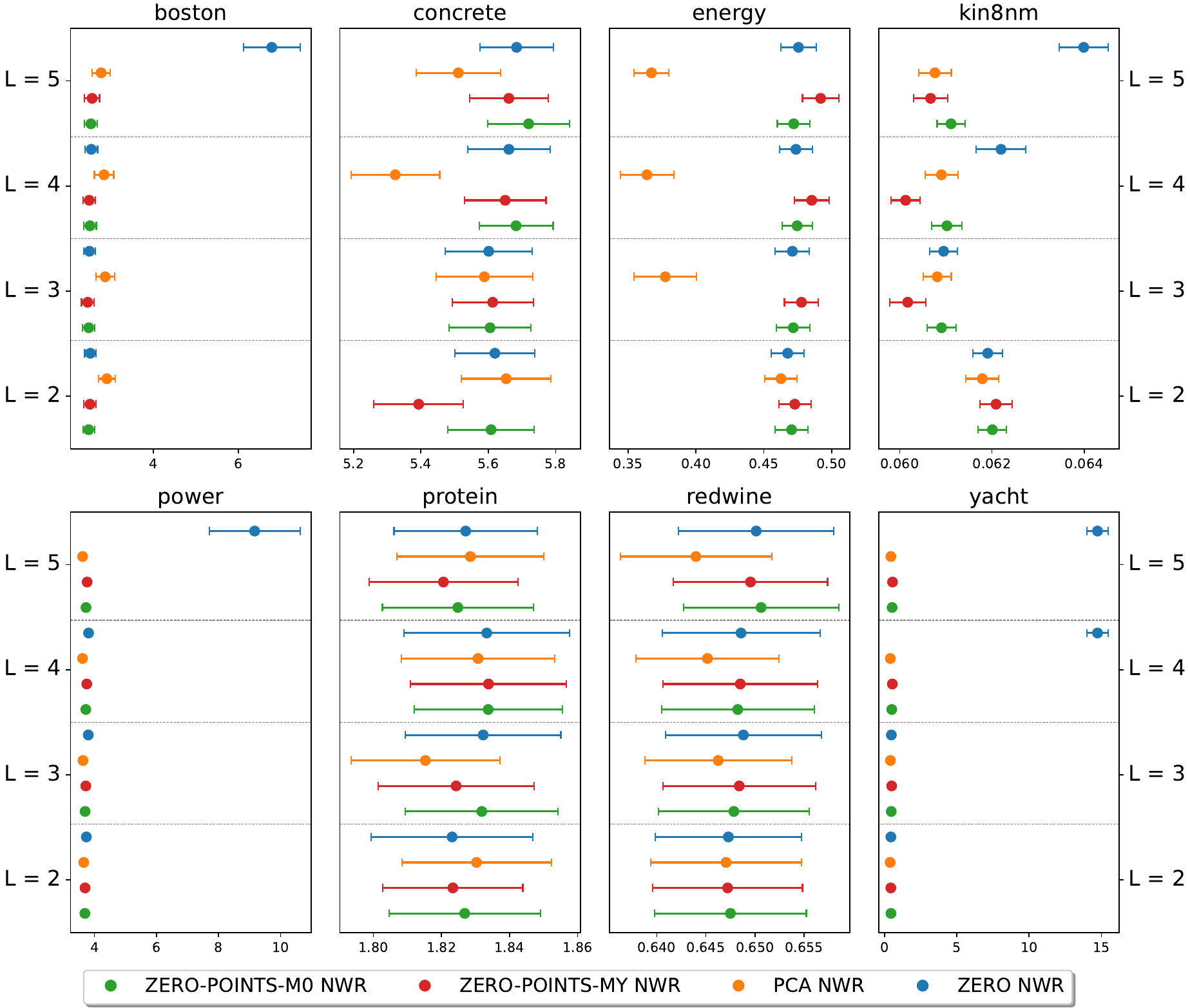}
        \caption{Non-whitened models.}
        \label{fig:results:rmse:nonwhiten}
    \end{subfigure}
    \caption{Test \RMSE (left is better) in all \UCI{} datasets.}
    \label{fig:results:uci:rmse}
\end{figure}

 \fig \ref{fig:results:rmse:whiten} and \fig\ref{fig:results:rmse:nonwhiten} shows \RMSE results for \UCI experiments on the whitened and non-whitened parameterizations. Furthermore, \utab~\ref{table:rankings:rmse:w} and \utab~\ref{table:rankings:rmse:nwr} show the average split-wise ranks of each model in terms of the \RMSE{}, with both parameterizations respectively.

\begin{table}[!htb]
      \centering
      \caption{Average rank of each \W{} model in terms of the test \RMSE{} for each dataset.}
  \label{table:rankings:rmse:w}
	\resizebox{\textwidth}{!}{
\begin{tabular}{lccccccccc}
\toprule
{} &  boston & concrete &  energy &  kin8nm &   power & protein & redwine &   yacht & Overall \\
\midrule
\PCA               &  $3.55$ &   $1.74$ &  $1.57$ &  $2.21$ &  $1.27$ &  $2.34$ &  $2.55$ &  $1.48$ &  $\mathbf{2.09}$ \\
\ZERO              &  $2.35$ &   $2.94$ &  $2.64$ &  $3.34$ &  $3.34$ &  $2.70$ &  $2.61$ &  $3.09$ &  $2.88$ \\
\ZEROMO &  $2.21$ &   $3.05$ &  $2.85$ &  $2.99$ &  $2.04$ &  $2.45$ &  $2.65$ &  $2.99$ &  $2.65$ \\
\ZEROMY &  $1.89$ &   $2.27$ &  $2.94$ &  $1.46$ &  $3.35$ &  $2.51$ &  $2.19$ &  $2.45$ &  \underline{$2.38$} \\
\bottomrule
\end{tabular}

}
\end{table}
\begin{table}[!htb]
      \centering
      \caption{Average rank of each \NWR{} model in terms of the test \RMSE{} for each dataset.}
  \label{table:rankings:rmse:nwr}
	\resizebox{\textwidth}{!}{
\begin{tabular}{lccccccccc}
\toprule
{} &  boston & concrete &  energy &  kin8nm &   power & protein & redwine &   yacht & Overall \\
\midrule
PCA               &  $3.36$ &   $1.91$ &  $1.31$ &  $2.25$ &  $1.35$ &  $2.48$ &  $2.21$ &  $1.82$ &  $\mathbf{2.09}$ \\
ZERO              &  $2.62$ &   $2.77$ &  $2.70$ &  $3.19$ &  $3.64$ &  $2.42$ &  $2.54$ &  $3.42$ &  $2.91$ \\
ZERO DatapointsM0 &  $2.23$ &   $2.89$ &  $2.74$ &  $2.77$ &  $2.31$ &  $2.64$ &  $2.64$ &  $2.65$ &  $2.61$ \\
ZERO DatapointsMY &  $1.79$ &   $2.42$ &  $3.25$ &  $1.79$ &  $2.70$ &  $2.46$ &  $2.61$ &  $2.10$ &  \underline{$2.39$} \\
\bottomrule
\end{tabular}

}
\end{table}

\subsubsection{Comparing Whiten and Non-Whiten Models}
\label{sec:app:c:4:w:vs:nwr}
\begin{figure}[!b]
    \centering
    %
    \begin{subfigure}{0.49\textwidth}
        \centering
        \includegraphics[width=\linewidth]{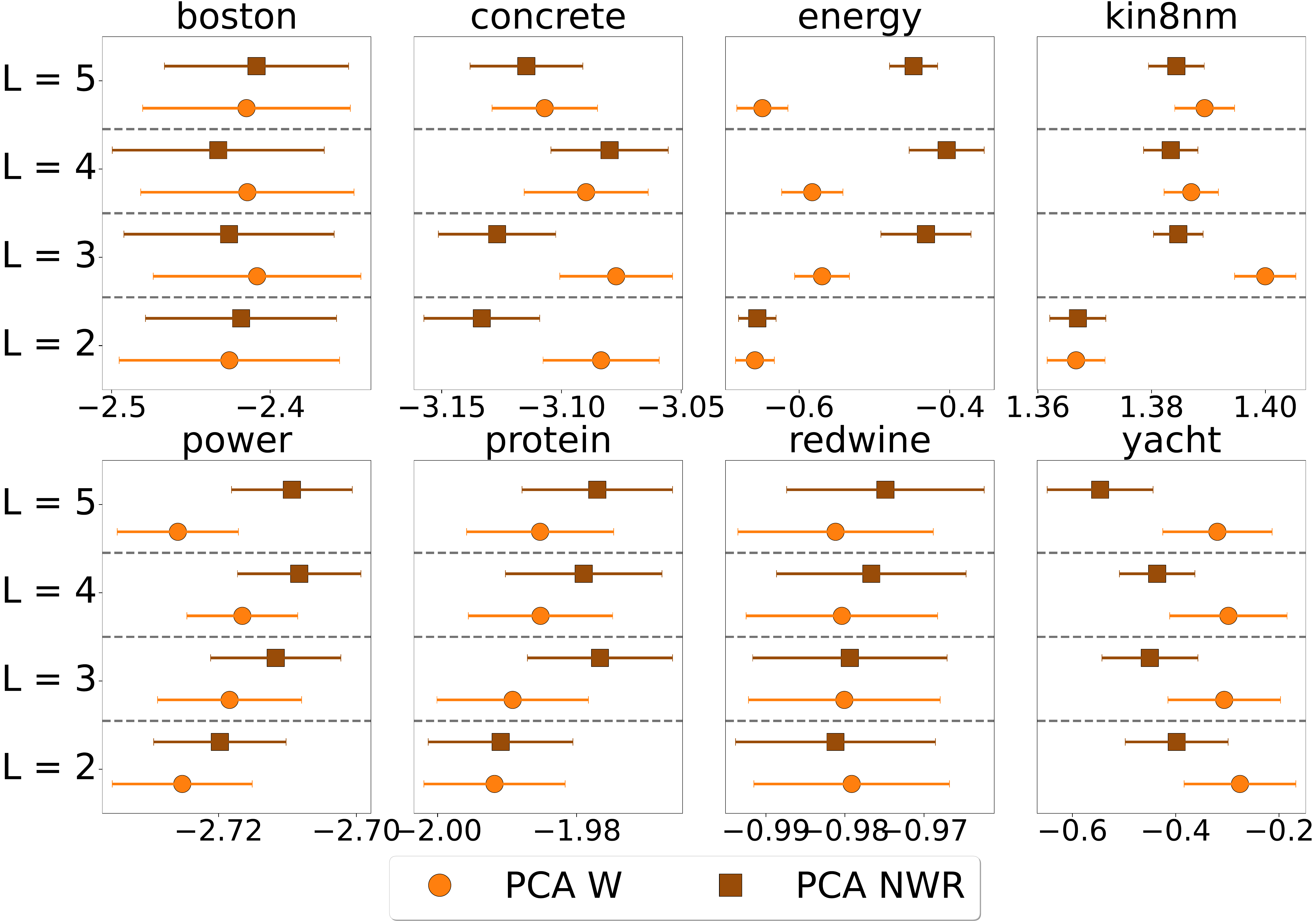}
        \caption{\PCA \DGP{}}
    \end{subfigure}
    \hfill
    \begin{subfigure}{0.49\textwidth}
        \centering
        \includegraphics[width=\linewidth]{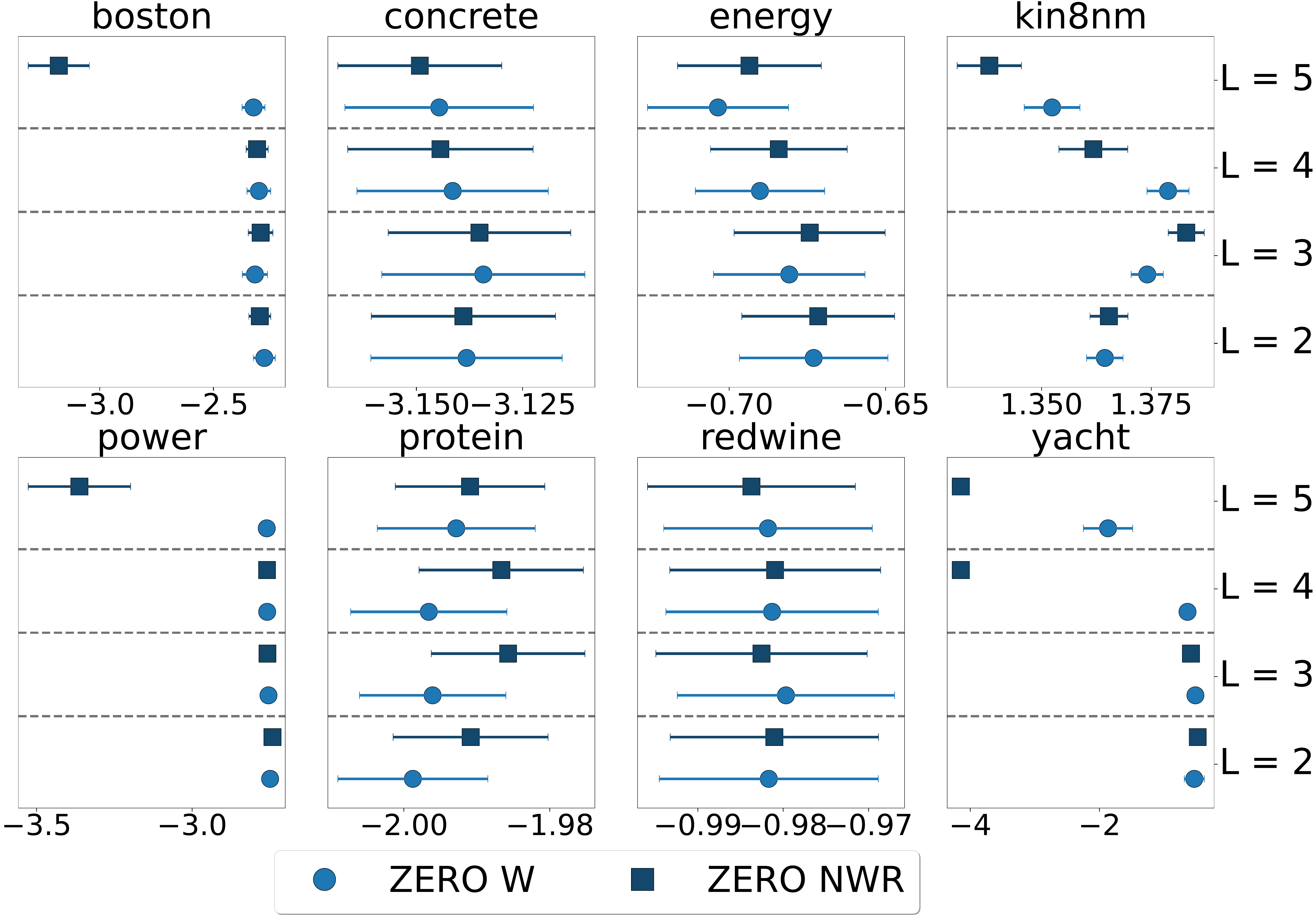}
        \caption{\ZERO \DGP{}}
    \end{subfigure}
    %
    \begin{subfigure}{0.49\textwidth}
        \centering
        \includegraphics[width=\linewidth]{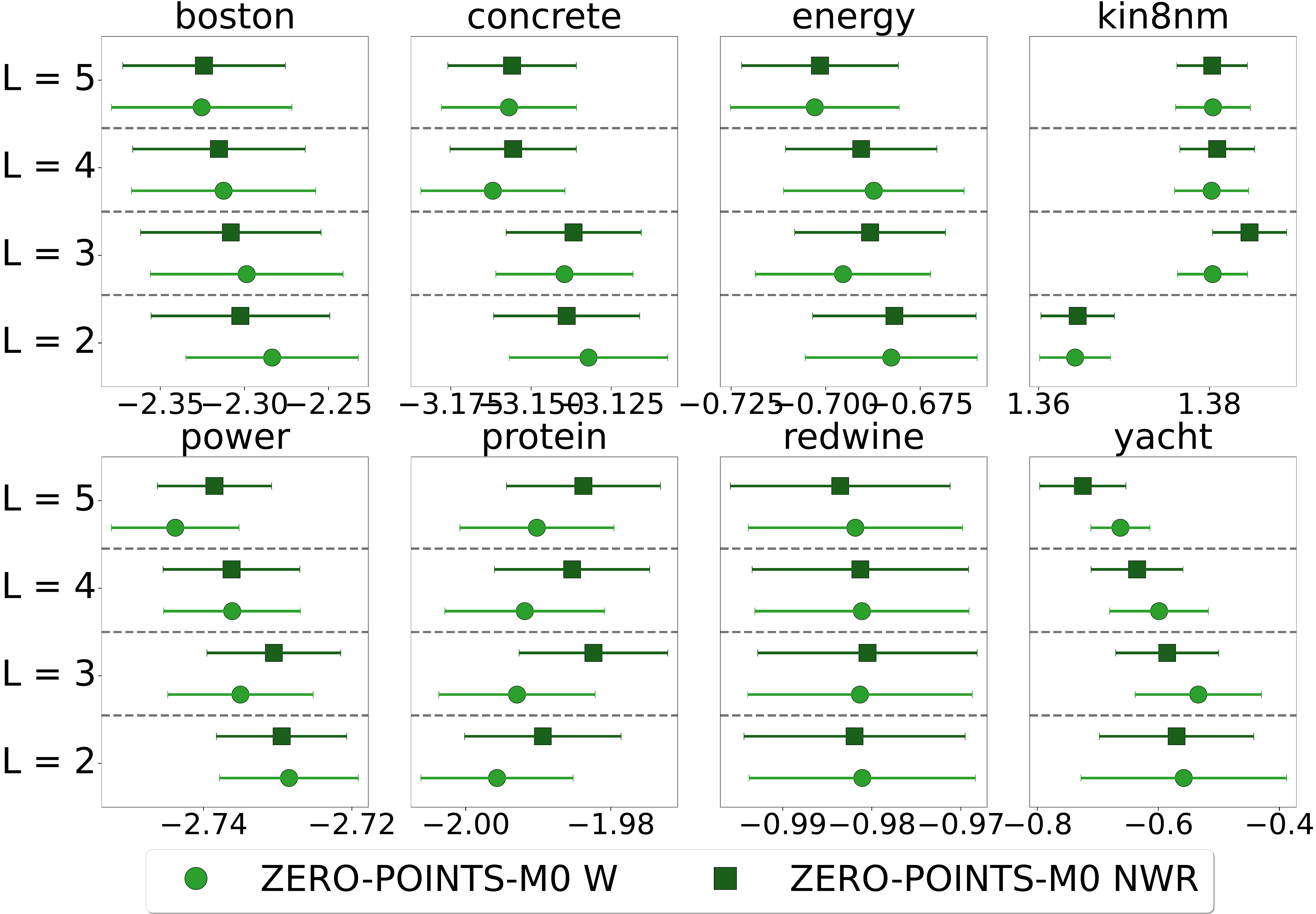}
        \caption{\MO \ZERO \DGP{}}
    \end{subfigure}
    \begin{subfigure}{0.49\textwidth}
        \centering
        \includegraphics[width=\linewidth]{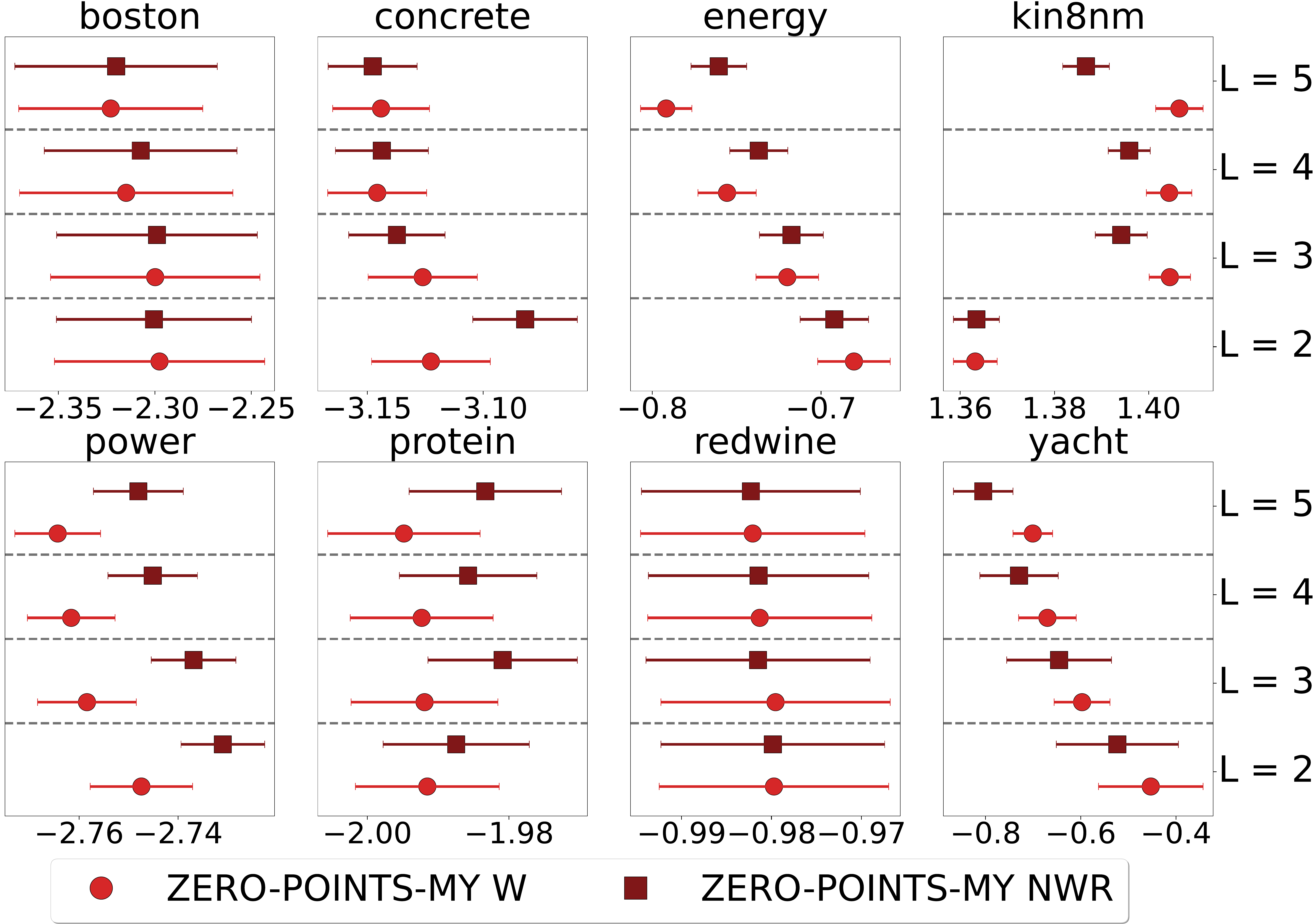}
        \caption{\MY \ZERO \DGP{}}
    \end{subfigure}
    \caption{Comparison of parameterization across different models and parameter initialization}
    \label{fig:uci:comparing:parameterizations}
\end{figure}
The results obtained in the toy dataset (see \usec~\ref{sec:experiments:toy}) showed that the \NWR in all the models achieved the worst performance. Moreover, these results are confirmed in the \UCI experiments (see \usec~\ref{sec:experiments:uci}) where we have observed that the \ZERO \NWR collapses more than its whitened version. We now compare both parameterizations for the rest of the models, namely \PCA, \MO, and \MY. 

We show the comparison of both parameterizations for each model in  \fig \ref{fig:uci:comparing:parameterizations}. The only conclusion we might draw is that the \ZERONWR is worse than the \ZEROW, with clear collapse in some cases (Power, Boston, Yacht, and Kin8nm). Only Protein benefits from the \NWR parameterization with this mean function, where it works better in all depths except $5$. In the rest of the parameterizations, there is no clear winner. In the \PCA model, we see that the Yacht dataset is better fitted using the whitened parameterization, in contrast to the Energy dataset. In the rest of the datasets, both parameterizations perform similarly, with one being the winner depending on the dataset. It must be noticed that, within a dataset, different \DGP{} depths perform differently depending on the parameterization, as in Concrete, where $2,3$ layers work better with the whitened, but $4$ works better with non-whitened. The reason behind performance differences for this mean function could rely on the different statistical models being implemented by each parameterization. This is a research direction for future work.

\begin{figure}[!tbp]
    \centering
    \begin{subfigure}{0.49\textwidth}
        \centering
        \includegraphics[width=\linewidth]{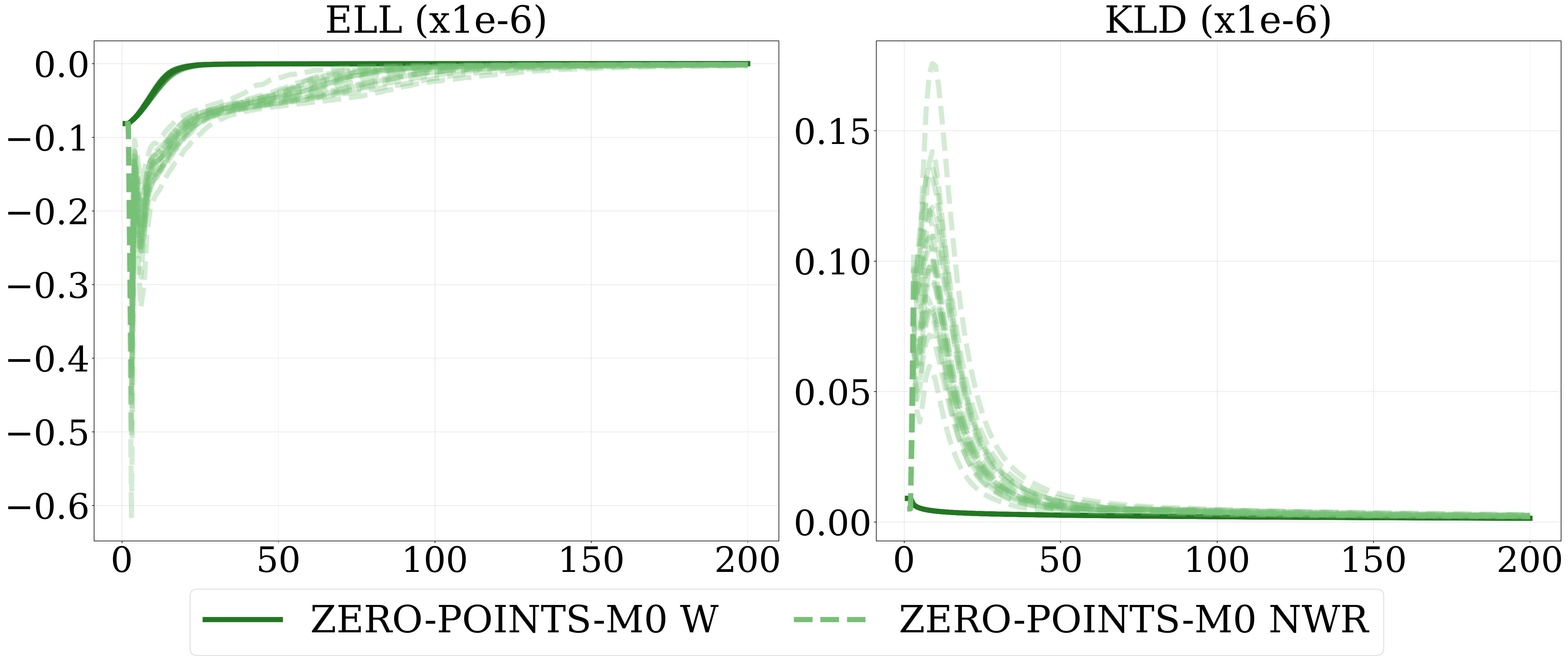}
    \end{subfigure}
     \begin{subfigure}{0.49\textwidth}
        \centering
        \includegraphics[width=\linewidth]{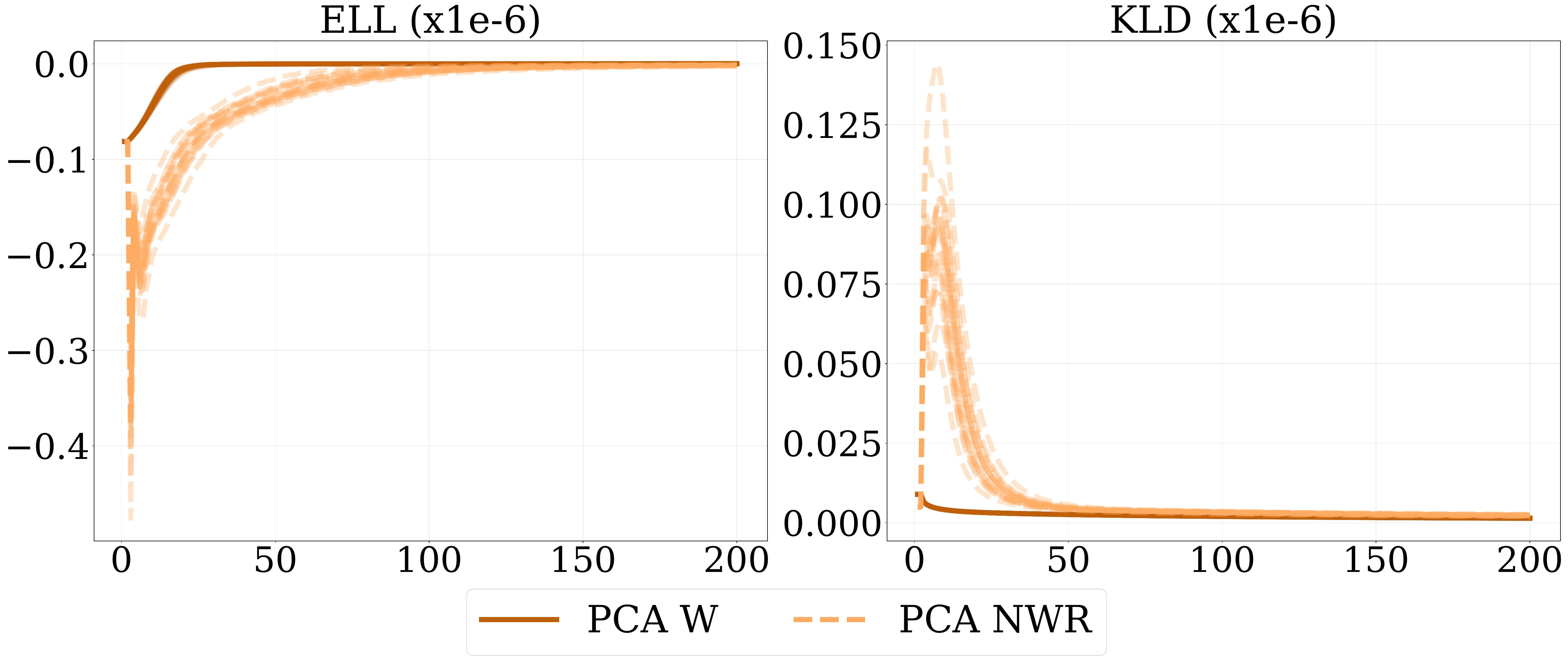}
    \end{subfigure}
\caption{Learning curves of both parameterizations, for each split, for the $5$ depth layer \DGP{}s trained in the Power dataset. Models displayed represent the \PCA and the \MO models.}
\label{fig:learn:curves:parameterization}
\end{figure}

For the \MO and \MY models, the tendency is similar. The \MY model performs better using the \NWR parameterization in Power or Protein. The whitened works better in Kin8nm. In the \MO both parameterizations work on pair. To confirm that this performance gap between parameterizations does not come from a poor solution coming from a noisy algorithm, we select (by observing \fig~\ref{fig:uci:comparing:parameterizations}) a model configuration where the \NWR parameterization outperforms the whitened one and plot its learning curves. In \fig~\ref{fig:learn:curves:parameterization}, we represent the \PCA and \ZEROMO models in a $5$ layer \DGP trained on the Power dataset. The figure shows that the \NWR parameterization presents a very noisy behavior, confirming that the improved results do not come from a stable optimization procedure. This confirms that, although the \NWR can achieve better results, it still yields to a poor optimization procedure. This suggests that \ZERO \NWR is more prone to posterior collapse due not only to its initialization, but also to the noisy gradients induced by the \NWR parameterization. The proposed initializations, \MO{} and \MY{}, do not mitigate the noisy optimization process itself, as expected, but they nevertheless help avoid convergence to poor solutions.

Lastly, in \utab~\ref{tab:w:nwr:difference}, we show the average difference between the performance of the \W{} and \NWR{} versions of each model. This table reveals that the proposed initializations yield the lowest average difference between the two different parameterizations of the model. This supports the previously indicated idea that our proposal induces a much more stable optimization process, which critically leads the model to avoid the posterior collapse pathology. 
\begin{table}[!t]
\caption{Average split-wise difference (in absolute value) between the performance of the whitened and \NWR{} parameterizations of each of the models. Lower values indicate higher similarity between the performance of both parameterizations.}
    \centering
    \resizebox{\textwidth}{!}{\begin{tabular}{llllllllll}
\toprule
 &     Boston &   Concrete &     Energy &     Kin8nm &      Power &    Protein &    Redwine &      Yacht &    Overall \\

\midrule
\PCA{}               &  $$0.053$$ &  $$0.041$$ &  $$0.213$$ &  $$0.011$$ &  $$0.021$$ &  $$0.016$$ &  $$0.009$$ &  $$0.227$$ &  $$0.074$$ \\
\ZERO{}              &  $$0.863$$ &  $$0.023$$ &  $$0.025$$ &  $$0.024$$ &  $$0.607$$ &  $$0.017$$ &  $$0.008$$ &  $$2.280$$ &  $$0.481$$ \\
\ZEROMO{} &  $$0.031$$ &  $$0.020$$ &  $$0.026$$ &  $$0.007$$ &  $$0.012$$ &  $$0.017$$ &  $$0.004$$ &  $$0.146$$ &  $\mathbf{0.033}$\\
\ZEROMY{} &  $$0.051$$ &  $$0.026$$ &  $$0.057$$ &  $$0.020$$ &  $$0.021$$ &  $$0.015$$ &  $$0.006$$ &  $$0.127$$ &  \underline{$0.040$} \\
\bottomrule
\end{tabular}
}
\label{tab:w:nwr:difference}
\end{table}

\subsubsection{Split-wise Differences}\label{sec:app:c:4:splits}

In the experimental section, we observed that the proposed initialization strategy sometimes outperforms or simply matches the performance of the \ZERO model. As we mentioned, since the statistical model or the inference family is not changed, we just shall expect improved performance of the proposed strategy when the \ZERO model with the initialization $\varm{}{}=\veczero$ cannot be correctly optimized. This section confirms this is the case most of the time, and that the reason behind suboptimal performance comes from a tendency towards posterior collapse.
We now investigate the performance gap.

Average results in \fig{}\ref{fig:results:ll:whiten},\ref{fig:results:ll:nwr} might hide particular splits in which the proposed solution solves the collapse problem. Only datasets with a large number of train-test splits collapsing, as in Yacht, will show an average degraded performance. This is because we are not expecting an improved performance of our proposal when the \ZEROW model optimizes correctly. Thus, for some datasets, we take the split with the lowest and the highest test log likelihood difference between the \PCA and \ZERO models and plot the corresponding \KLD and likelihood variance in \fig \ref{fig:KLD:LLH:several_splits}. We can observe the same tendency discussed. Models with the highest performance difference (left column) usually present a higher likelihood variance, and a smaller \KLD for the \ZEROW and \ZERONWR{} model's (blue line), showing this \emph{tendency} to collapse. We observe how the proposed initialization (green line) solves the problem, presenting a behavior similar to the \PCA (orange model). When the performance gap is the lowest (right column), we observe how the \ZERO model matches the \PCA and \MO. Interestingly, we observe how, at the beginning, the variance of the \ZERO model rapidly increases. At the end, however, the optimization algorithm is able to reduce it to a level comparable with the \MO  and \PCA models. Importantly, in the Concrete dataset, using $3$ layer models, the \ZEROW works better than the proposed solution, see \fig{}\ref{fig:results:ll:whiten}.  This confirms that, effectively, we shall only expect improved performance of our approach when the model presents collapse. This is because both statistical models are the same. Regarding the rest of the datasets and models, we have widely inspected most of the splits. We observe that not all of them present this behavior, and that sometimes our proposed solution cannot correct the problem or directly works worse. This might be because we are only initializing the \ZERO model close to the \PCA model at the inducing point locations. This is a limitation of our proposed solution, with further research going in this direction.

\begin{figure}[!htbp]
    \centering
    \begin{subfigure}[t]{0.49\linewidth}
        \centering
        \includegraphics[width=\linewidth]{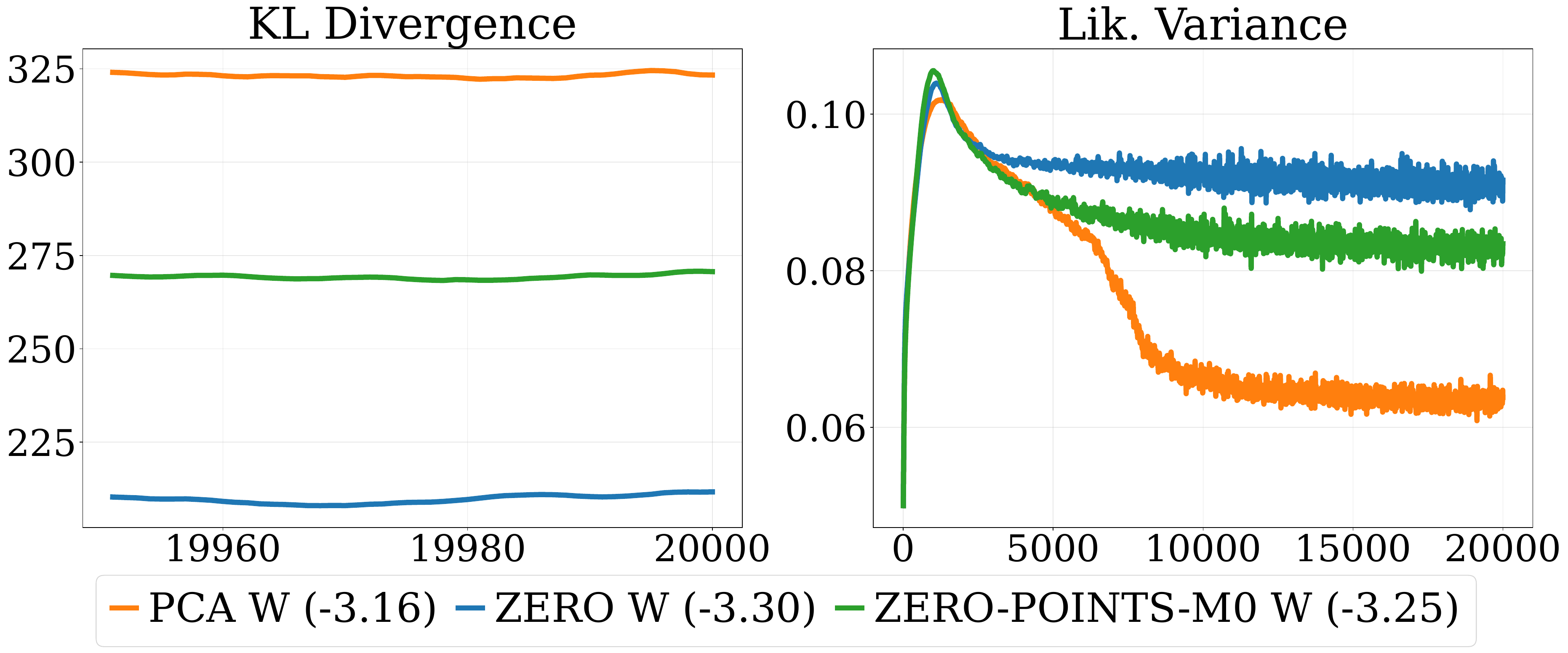}
    \end{subfigure}
    \begin{subfigure}[t]{0.49\linewidth}
        \centering
        \includegraphics[width=\linewidth]{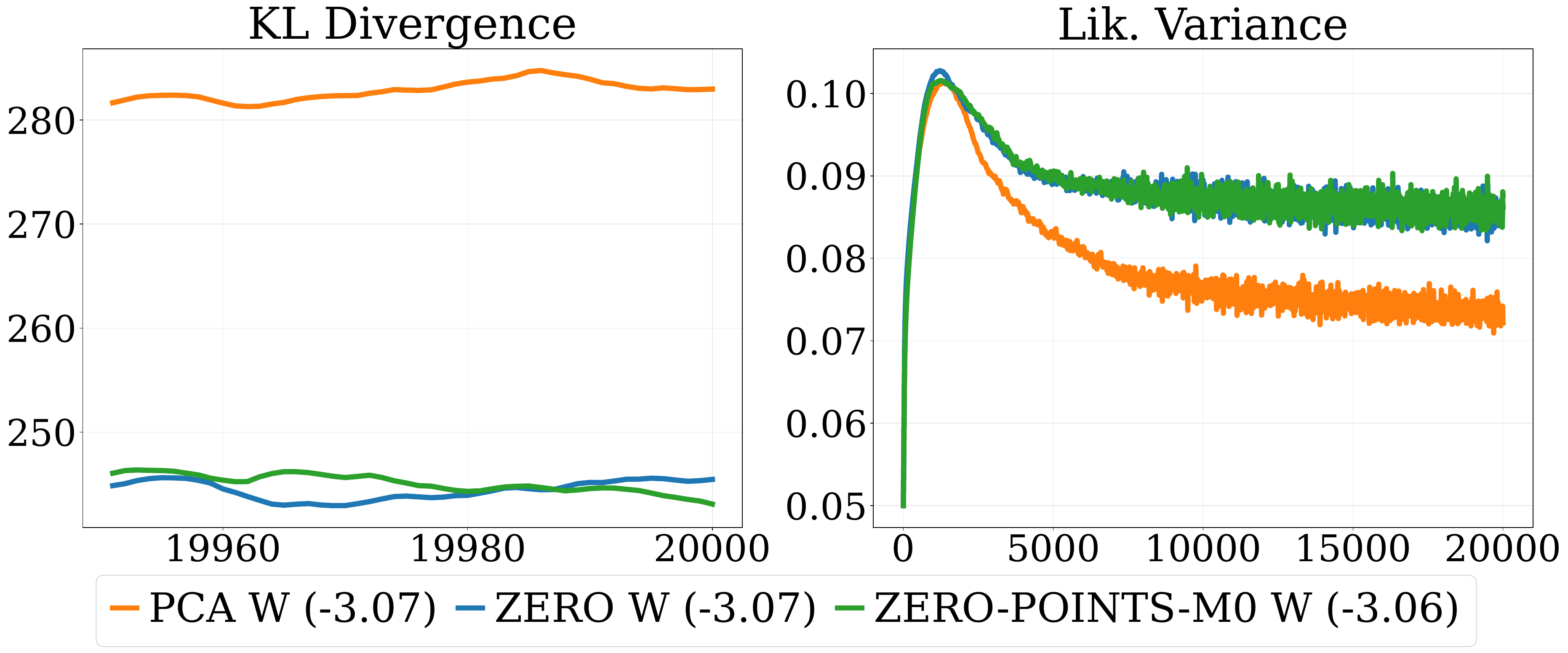}
    \end{subfigure}\\
    \caption*{Concrete dataset. 3 layers \DGP. Whitened parameterization.}
     \begin{subfigure}[t]{0.49\linewidth}
        \centering
        \includegraphics[width=\linewidth]{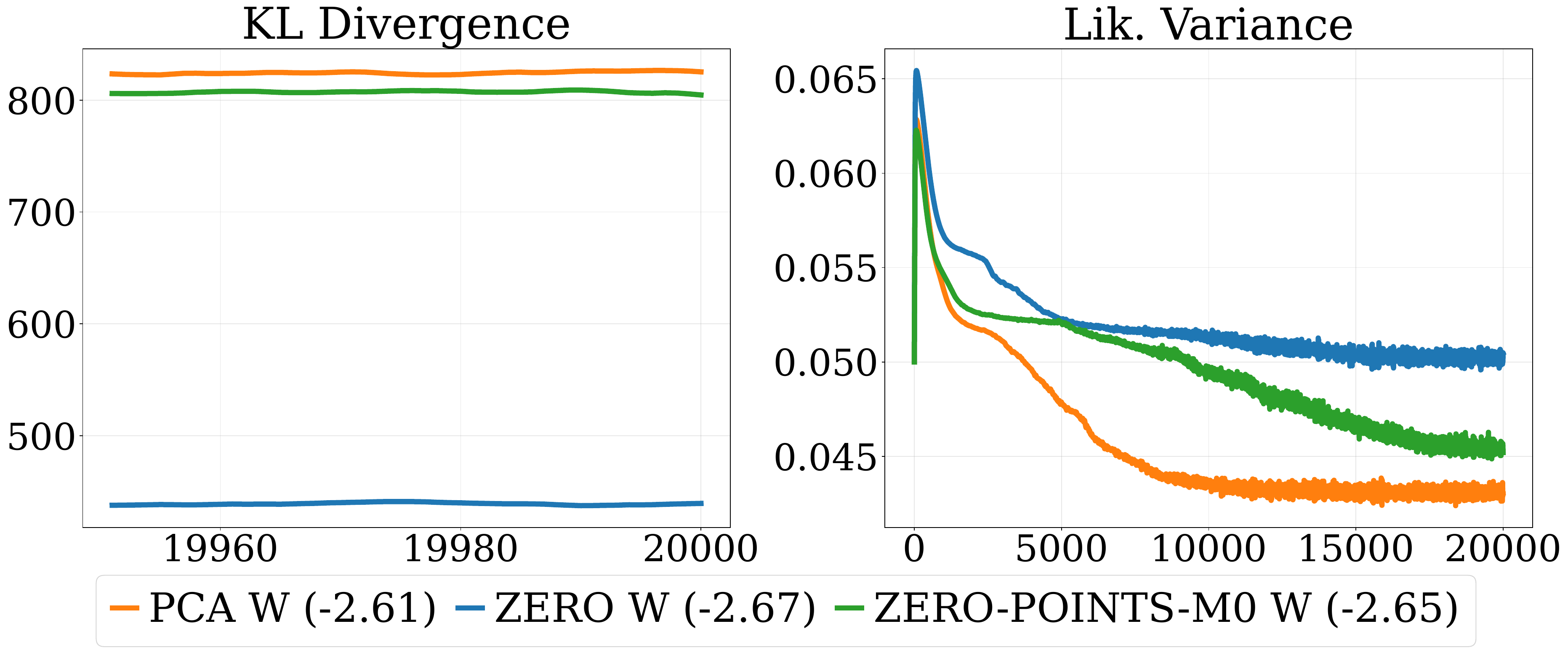}
    \end{subfigure}
    \begin{subfigure}[t]{0.49\linewidth}
        \centering
        \includegraphics[width=\linewidth]{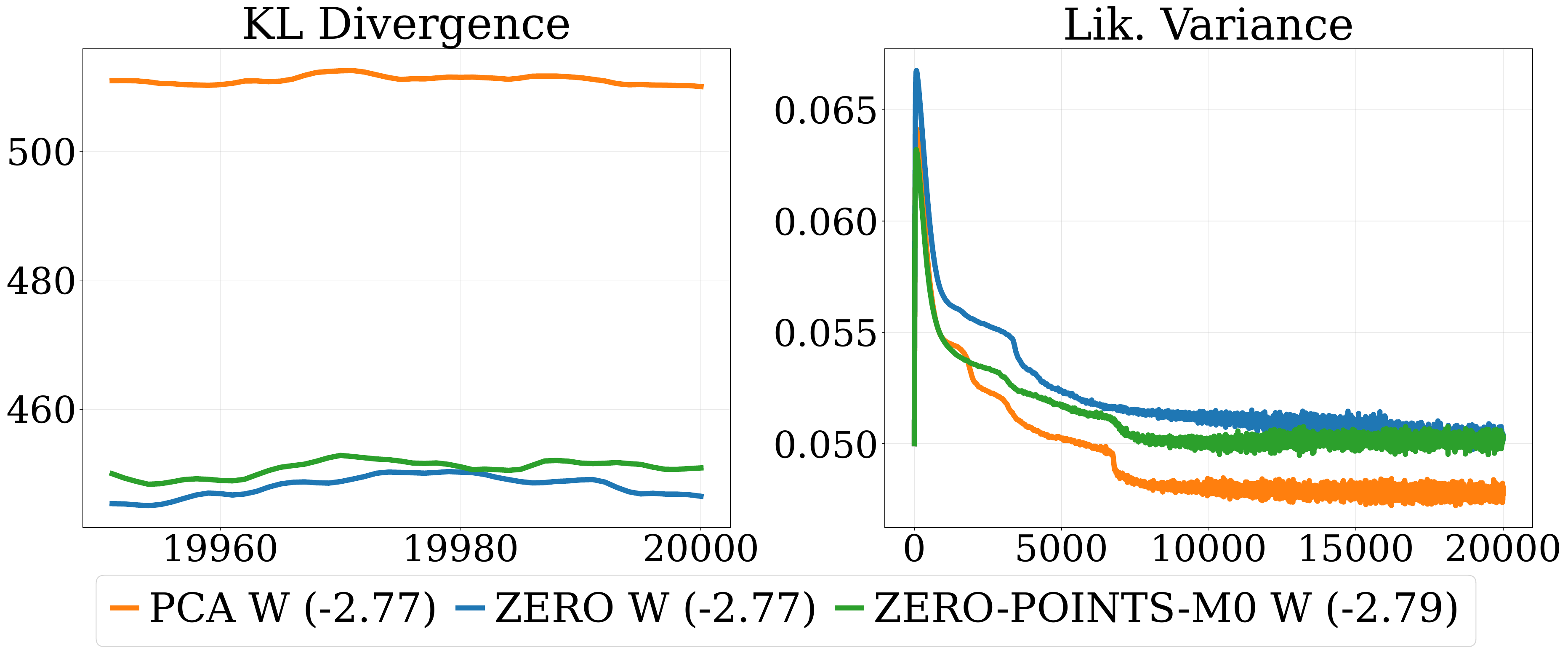}
    \end{subfigure}\\
    \caption*{Power dataset. 3 layers \DGP. Whitened parameterization.}
    \begin{subfigure}[t]{0.49\linewidth}
        \centering
        \includegraphics[width=\linewidth]{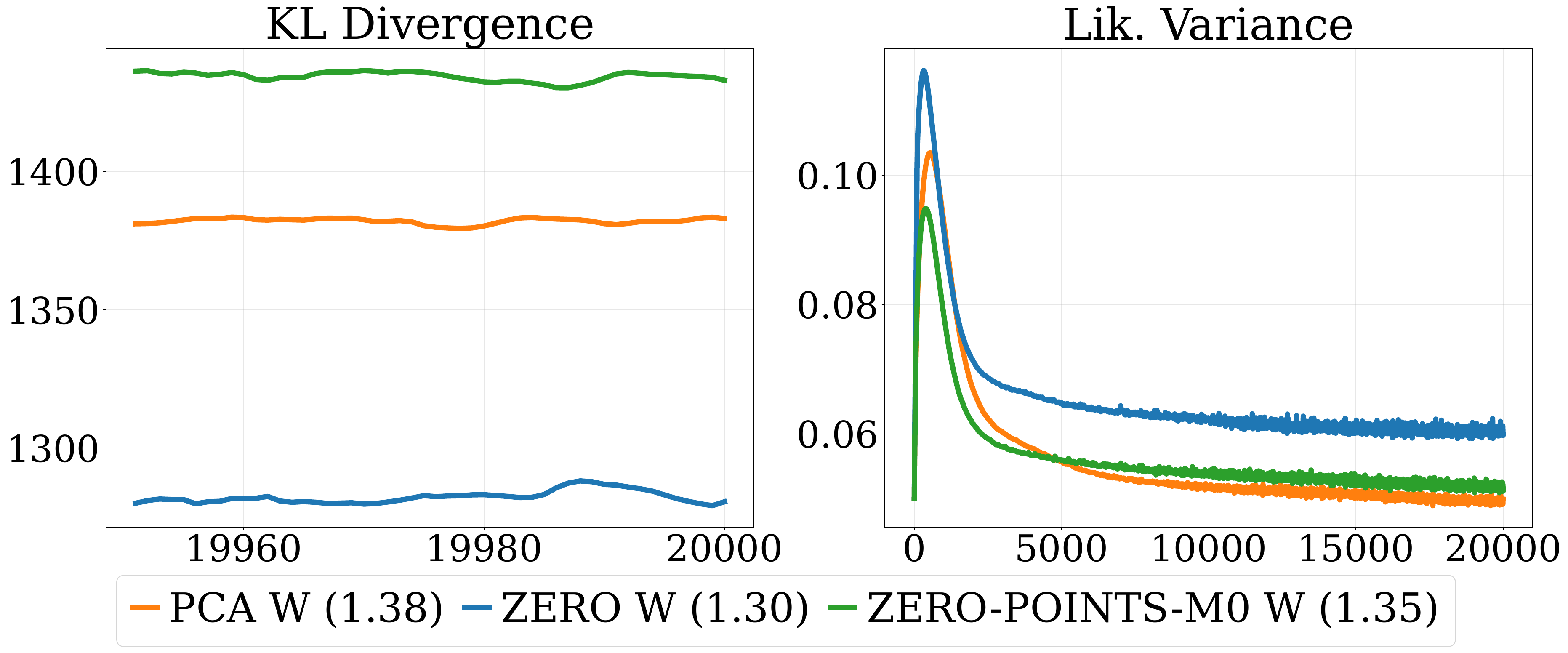}
    \end{subfigure}
    \begin{subfigure}[t]{0.49\linewidth}
        \centering
        \includegraphics[width=\linewidth]{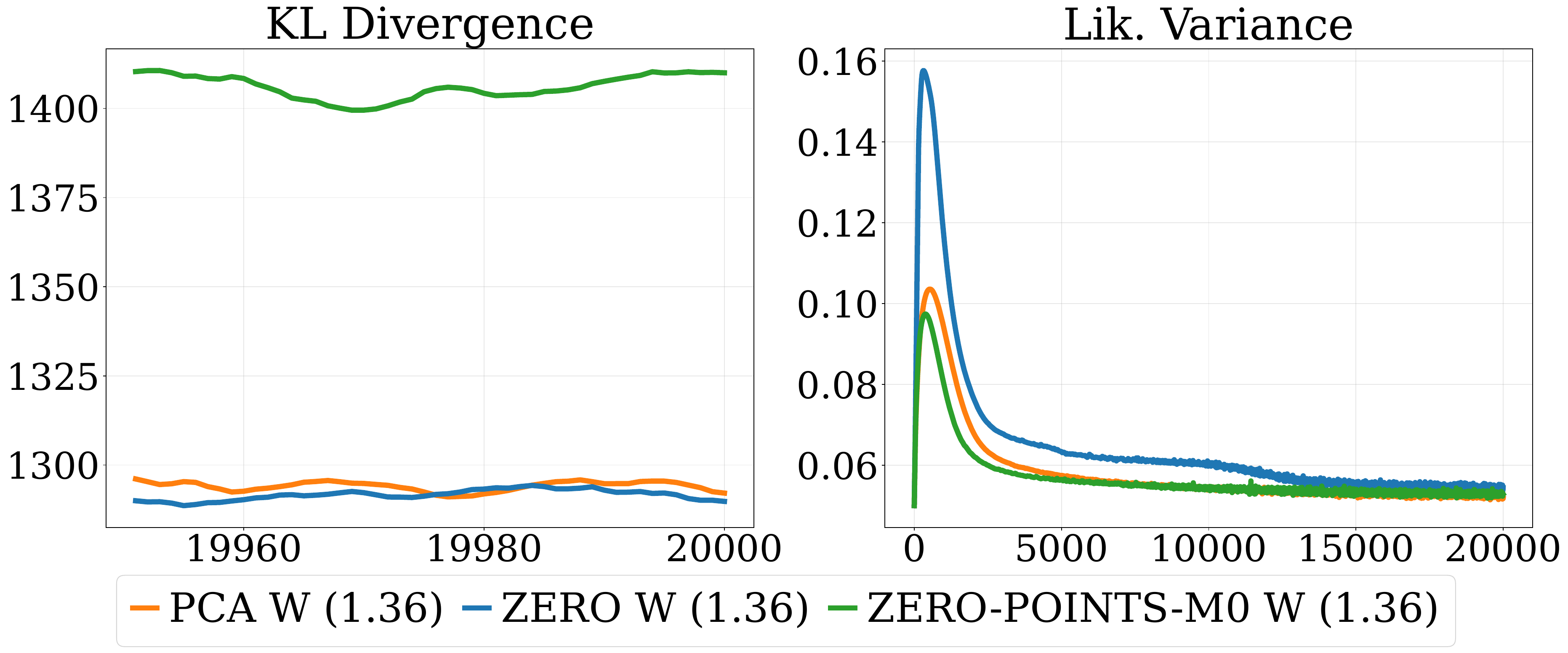}
    \end{subfigure}\\
    \caption*{Kin8nm dataset. 5 layers \DGP. Whitened parameterization.}
\end{figure}

\begin{figure}[!htbp]    
\ContinuedFloat
    \begin{subfigure}[t]{0.49\linewidth}
        \centering
        \includegraphics[width=\linewidth]{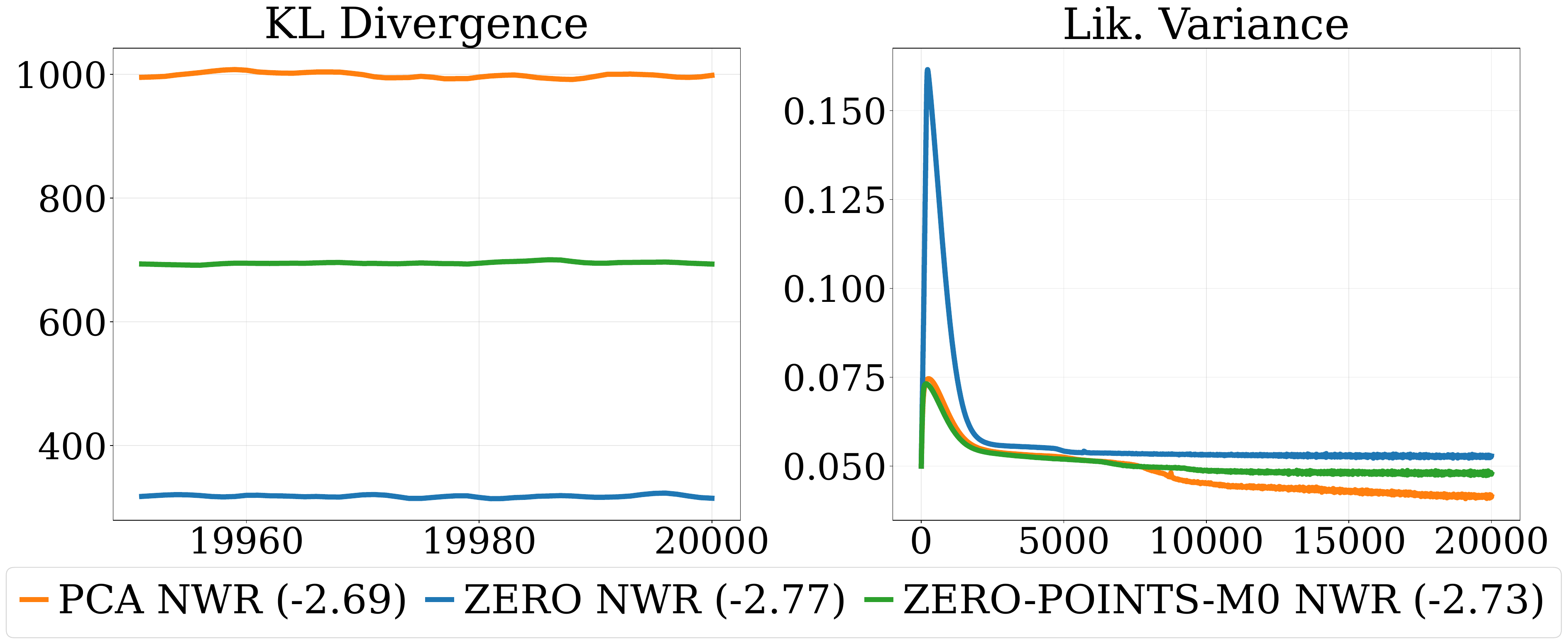}
    \end{subfigure}
    \begin{subfigure}[t]{0.49\linewidth}
        \centering
        \includegraphics[width=\linewidth]{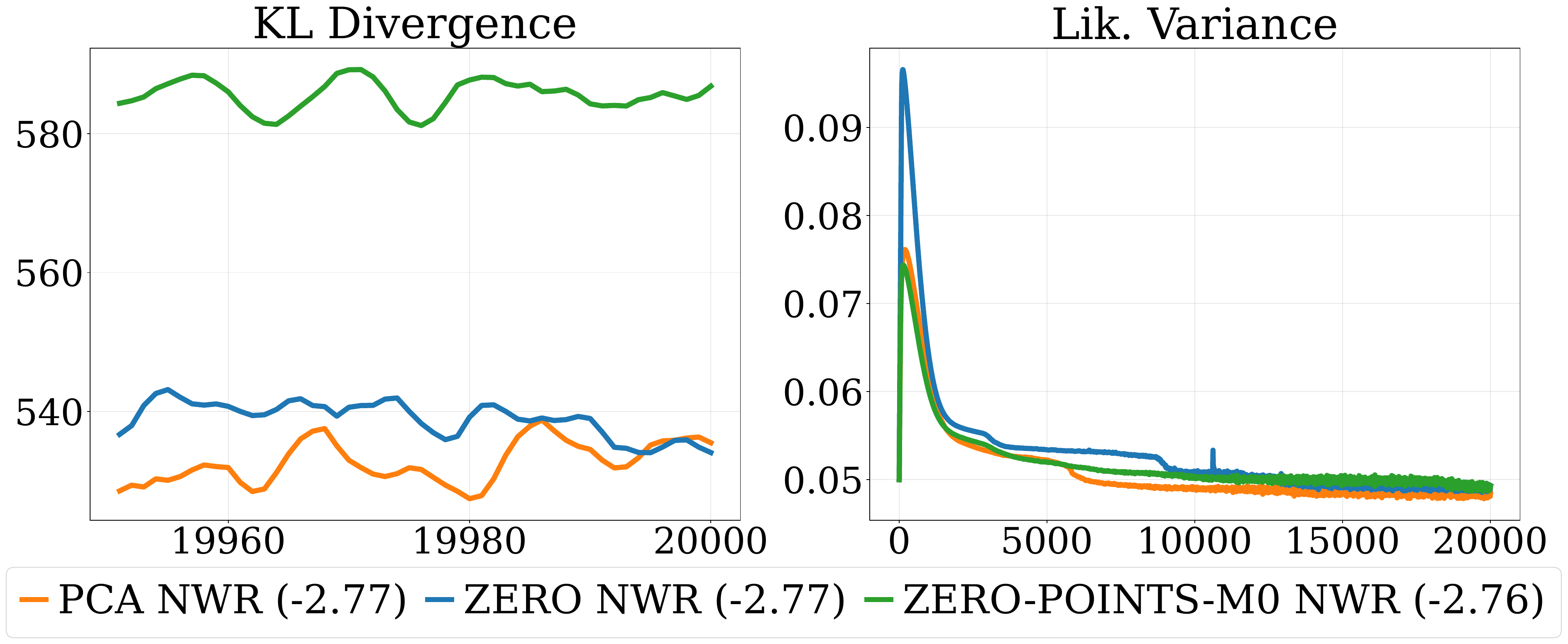}
    \end{subfigure}\\
    \caption*{Power dataset. 3 layers \DGP. Non-whitened parameterization.}
     \begin{subfigure}[t]{0.49\linewidth}
        \centering
        \includegraphics[width=\linewidth]{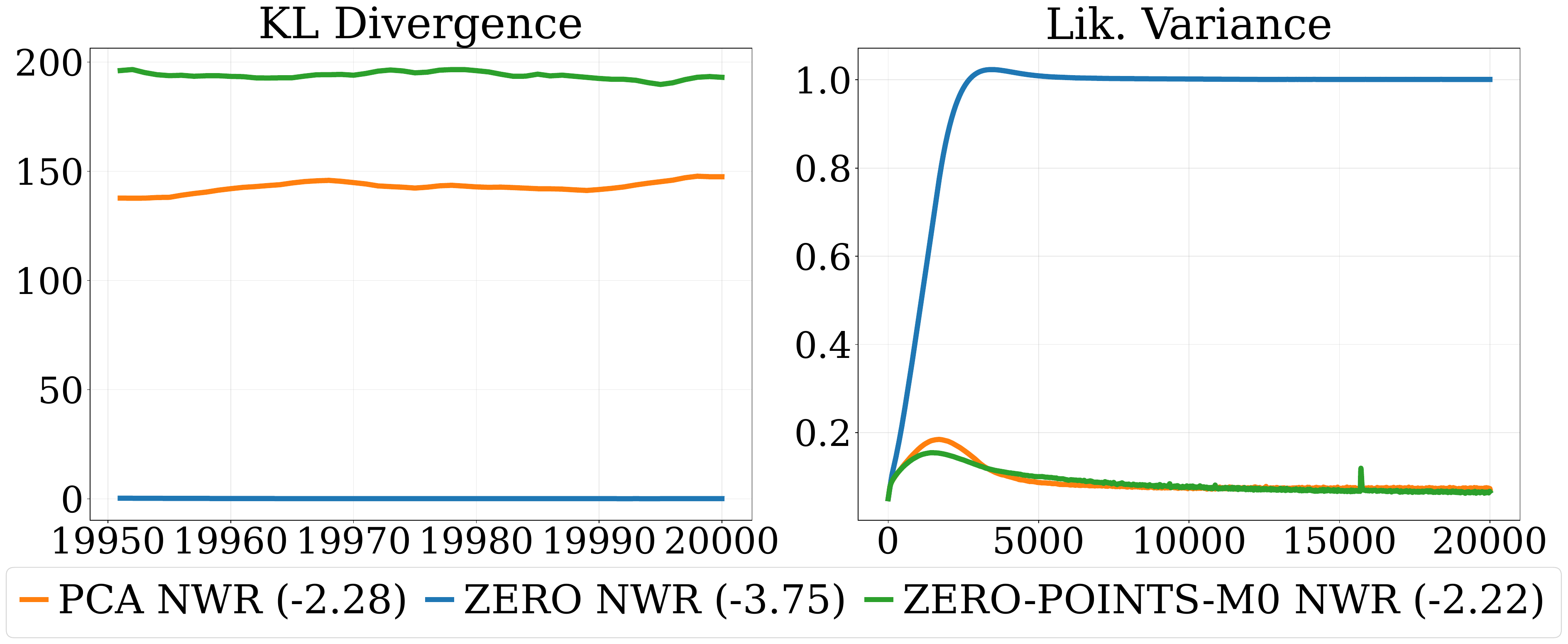}    
    \end{subfigure}
    \begin{subfigure}[t]{0.49\linewidth}
        \centering
        \includegraphics[width=\linewidth]{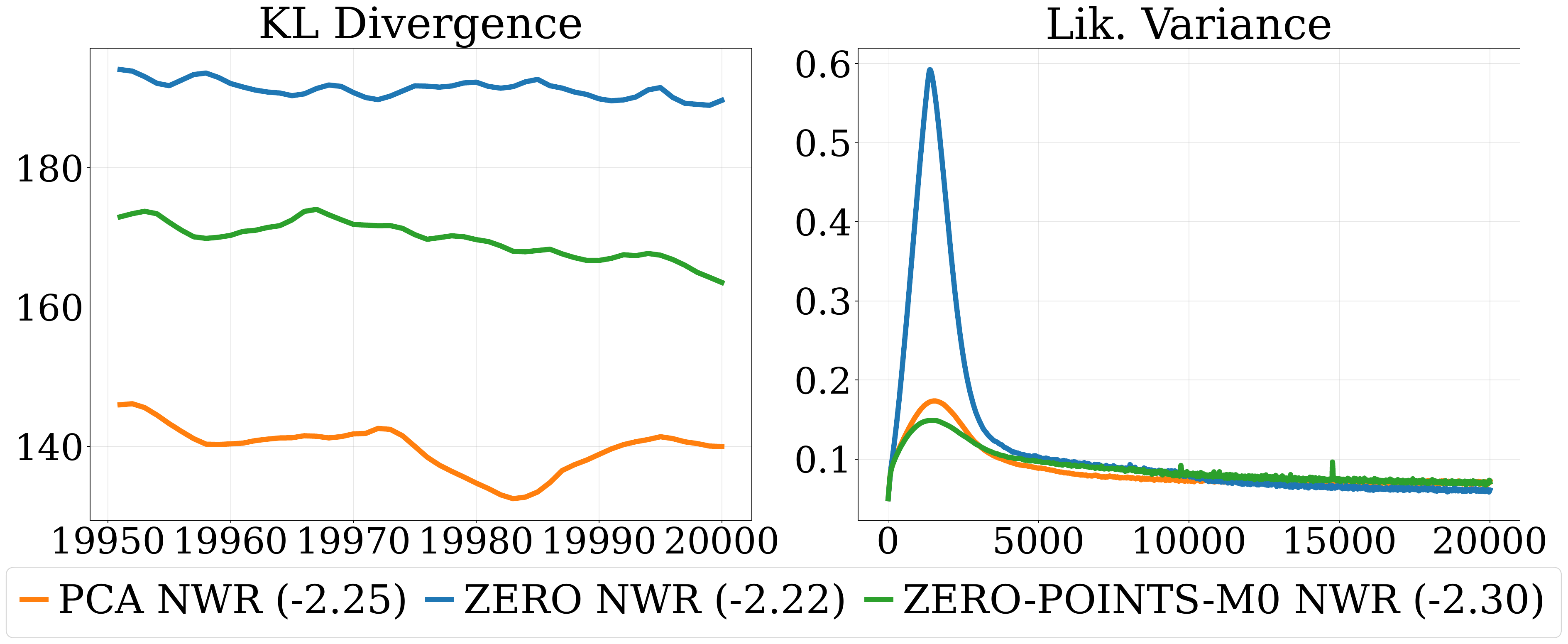}
    \end{subfigure}\\
    \caption*{Boston dataset. 5 layers \DGP. Non-whitened parameterization.}
     \begin{subfigure}[t]{0.49\linewidth}
        \centering
        \includegraphics[width=\linewidth]{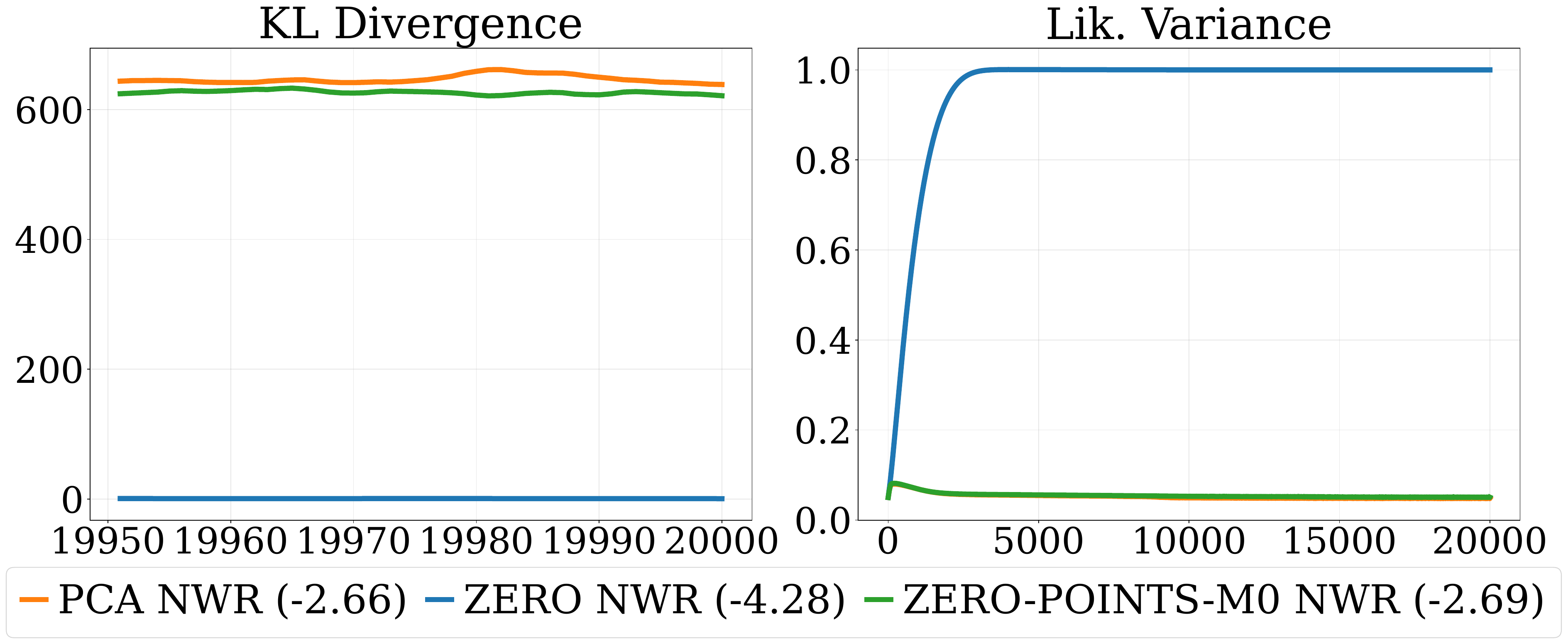}    
    \end{subfigure}
    \begin{subfigure}[t]{0.49\linewidth}
        \centering
        \includegraphics[width=\linewidth]{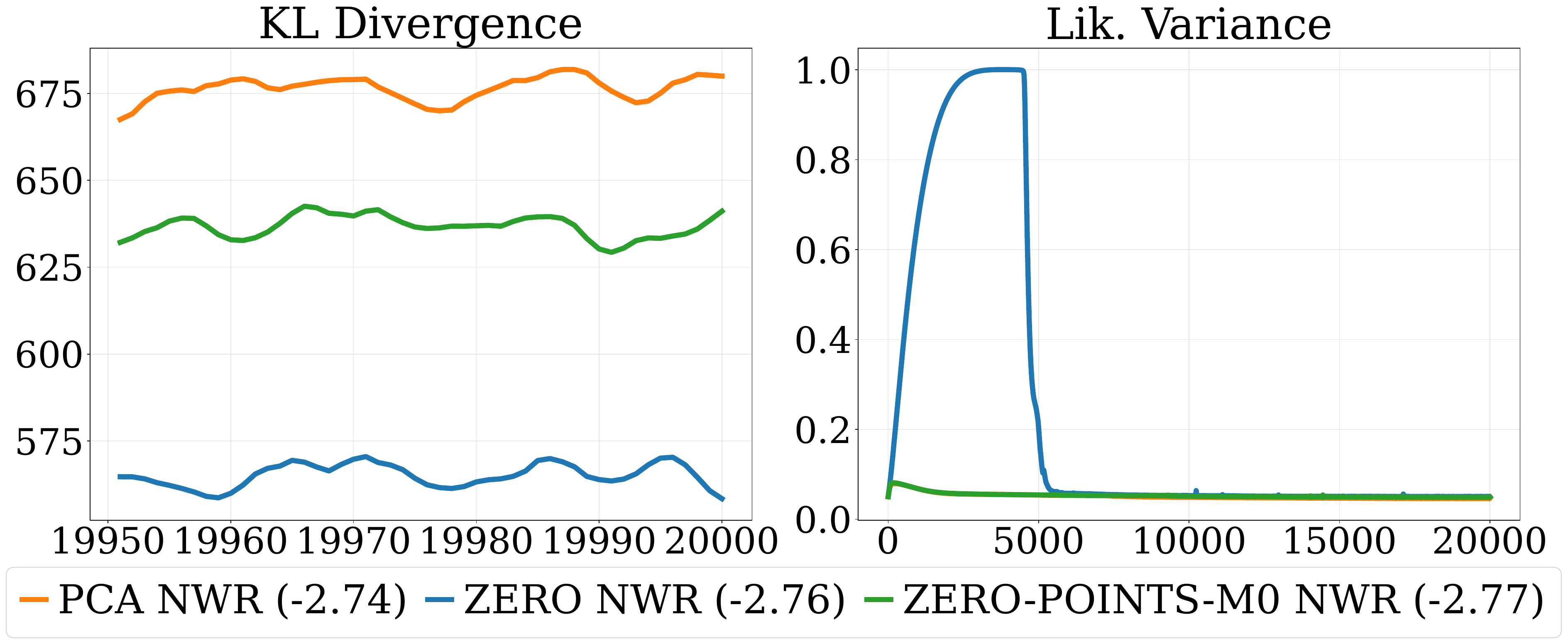}
    \end{subfigure}\\
    \caption*{Power dataset. 5 layers \DGP. Non-whitened parameterization.}
    \begin{subfigure}[t]{0.49\linewidth}
        \caption{Greatest split performance difference.}
    \end{subfigure}
    \begin{subfigure}[t]{0.49\linewidth}
        \caption{Lowest split performance difference.}
    \end{subfigure}
    \caption{ Final \KLD and likelihood variance parameter across several datasets, model deepness, and parameterization. The left column represents the curves for models in which the difference between the \PCA and \ZERO model performance is the highest across all the splits. The right column represents the same, but when the model performance is the lowest.}
    \label{fig:KLD:LLH:several_splits}
\end{figure}

\end{document}